\documentclass[10pt,twocolumn,letterpaper]{article}

\usepackage[pagenumbers]{cvpr} 
\usepackage{calligra}

\newcommand{\topic}[1]{\vspace{2mm}\noindent\textbf{#1.}}
\usepackage{stfloats}

\definecolor{cvprblue}{rgb}{0.21,0.49,0.74}
\usepackage[pagebackref,breaklinks,colorlinks,allcolors=cvprblue]{hyperref}

\def\confYear{2025}

\title{Illusion3D: 3D Multiview Illusion with 2D Diffusion Priors}
\author{
Yue Feng$^{1}$ \quad
Vaibhav Sanjay$^{1}$ \quad
Spencer Lutz$^{1}$ \quad
Badour AlBahar$^{2}$ \quad
Songwei Ge$^{1}$ \quad
Jia-Bin Huang$^{1}$\\
$^{1}$University of Maryland, College Park \quad
$^{2}$Kuwait University\\
\url{https://3D-multiview-illusion.github.io}
}

\begin{document}
\twocolumn[{
\renewcommand\twocolumn[1][]{#1}
\maketitle
\begin{center}
    \vspace{-9mm}
    \centering
    \includegraphics[width=\linewidth, trim=3 10 0 8, clip]{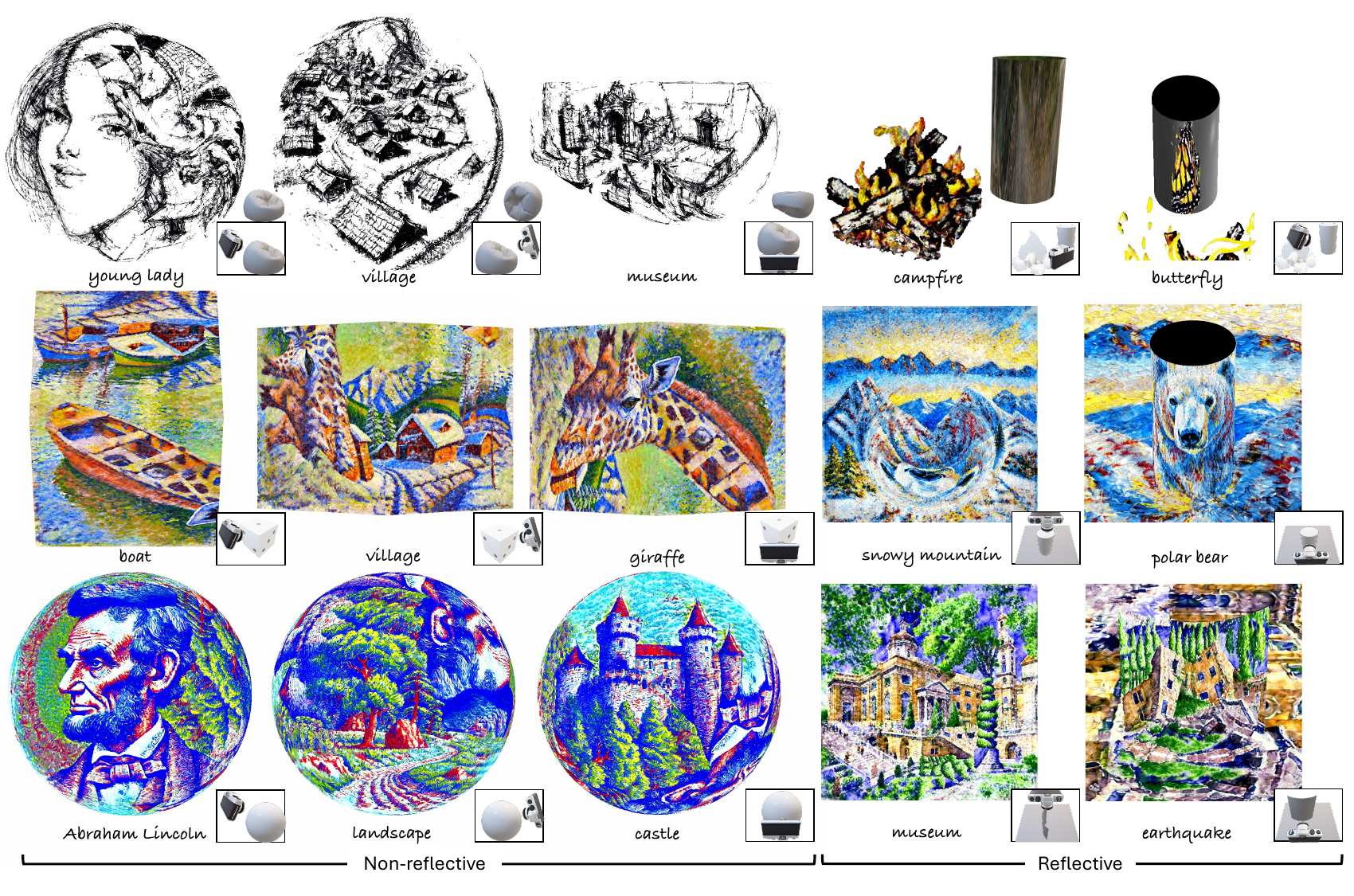}
    \vspace{-8.5mm}
    \captionof{figure}{\textbf{3D Multiview Illusion.} 
    Our work expands the capability of existing multiview illusions (based on shadow, wire, or 2D plane) to 3D surfaces with perspective views.
    Distinct visual interpretations can be observed when rendering our generated illusion from different perspectives on a consistent texture map.
    We showcase our 3D multiview illusions with different setups with text prompt inputs, including cubes, spheres, beanbags, and reflective cylinders/mirrors with 2D image grids and 3D shapes. For videos of the result gallery and real-world examples, please see \href{index.html}{index.html}.}
    \vspace{-1.5mm}
\label{fig:teaser}
\end{center}

}]
\begin{abstract}
Automatically generating a multiview illusion, where a single piece of visual content offers distinct interpretations from different viewing perspectives, is a compelling challenge.
Traditional methods, such as shadow art and wire art, create interesting 3D illusions but are limited to simple visual outputs (\ie, figure-ground or line drawing), restricting their artistic expressiveness and practical versatility. 
Recent diffusion-based illusion generation methods can generate more intricate designs but are confined to 2D images.
In this work, we present a simple yet effective approach for creating 3D multiview illusions based on user-provided text prompts or images.
Our method leverages a pre-trained text-to-image diffusion model to optimize the textures and geometry of neural 3D representations through differentiable rendering. 
When viewed from multiple angles, this produces different interpretations.
We develop several techniques to improve the quality of the generated 3D multiview illusions. 
We validate the effectiveness of our approach through extensive experiments and demonstrate 3D illusion with diverse 3D forms, showcasing its potential for different artistic expressions and practical applications.
\end{abstract}

\vspace{-8mm}
\section{Introduction}
\label{sec:intro}
From one angle, it’s a campfire; from another, a butterfly—such is the magic of multiview illusions, where a single object shifts its interpretation with every change of perspective~\cite{PerroniScharf2023ConstructingPS,Wu2022SurveyOC,Le2021OptimizedBF,Yang2019BinaryIC,Tian2023EvolvingTD}. 
This is an exciting form of art as the audience's angles affect their own visual experiences.
Creating an illusion is, however, not a trivial task. Various forms, including shadow art~\cite{Sadekar2021ShadowAR,ChaineHandSA,Mitra2009ShadowA,Min2017SoftSA}, wire art~\cite{Qu2023WiredPM, Xu2021ModelguidedED, Yeh2022GeneratingVW}, and reflective surface art, have been developed to produce appealing visual experiences yet limited expressiveness. Crafting these artworks also requires massive skill practice. 
Motivated by the success in using pretrained diffusion models to create 2D illusion~\cite{Burgert2023DiffusionIH, Geng2023VisualAG}, we aim to develop automatic methods for expressive 3D multiview illusion generation.

In this paper, we present an optimization-based framework for creating 3D multiview illusions that align with the given text prompts and images.
We harness the power of a pre-trained diffusion model to optimize 3D neural representations with Variational Score Distillation (VSD)~\cite{wang2023prolificdreamer}. Generating multiview 3D illusions, however, introduces unique challenges beyond those encountered in the 2D realm. The inherent ambiguity of 3D easily induces local minima that obstruct the optimization. To tackle this problem, we introduce a suite of novel techniques to enhance generation quality—scheduled patch-wise denoising, dynamic camera jittering, and progressively scaled render resolutions—all working in harmony to refine the final output.




Existing diffusion models are limited to specific resolution ranges. Nevertheless, we find that the model has been trained on different regions and views of the same object~\cite{podell2023sdxl}. This motivates us to optimize the local regions of an underlying 3D scene, pushing the upper bound of the supported resolutions. We introduce scheduled patch-wise denoising, enabling VSD-based methods to optimize efficiently at higher resolutions with a diffusion model pre-trained in low resolution. Our approach extends multiview illusions to produce detailed 3D outputs at \(1024\times1024\) and even \(2048\times2048\) resolution.


When optimizing the texture map with VSD to generate 3D illusions, we noticed several recurring artifacts. First, the outputs look noisier and less smooth than the results in optimizing 2D images. Second, the concepts appear multiple times within a single view or only occupy partial views. 

To mitigate the first issue, we introduce scheduled camera jittering.
Specifically, we perturb the rendering camera with a random Gaussian noise during optimization, facilitating seamless transitions when changing views. Perturbing the camera too much, however, could exacerbate the duplication issue. Therefore, we find a noise scheduling strategy helps, starting with small camera jittering and gradually increasing it throughout the denoising process. 

We also developed a novel technique that progressively increases render resolution to tackle the duplicate pattern issue.
Specifically, we start the optimization with a low-resolution rendering and progressively increase the render resolution of the optimized region throughout the optimization process.
This approach ensures that the main object is synthesized at the center of the view, achieving clearer and more focused illusions when combined with camera jittering and patch-wise denoising. We find that this technique also enforces the fusion of image content across different views and further reduces the duplicate pattern issue. 




Experiments show that our method is capable of generating a wide range of high-quality 3D multiview illusions in different formats, including cubes, spheres, complex shapes (such as beanbags and Legos) with neural texture representations, reflective surfaces, and ambiguous 3D structures. 

As a summary, our main contributions include:
\begin{itemize}
\item 
We explore a novel problem to generate 3D multiview illusions using pre-trained 2D diffusion models.
\item We introduce several techniques and design choices to improve the quality of the 3D multiview illusions.
\item We demonstrate that our approach can enhance and overcome the limitations of traditional artistic methods, with results applicable to real-world scenarios. 
\end{itemize}
\begin{figure*}[t]
    \vspace{-9mm}
    \centering
    \tiny
    \begin{subfigure}{0.22\linewidth}
        \includegraphics[width=\linewidth, trim={0 5 12 10}, clip]{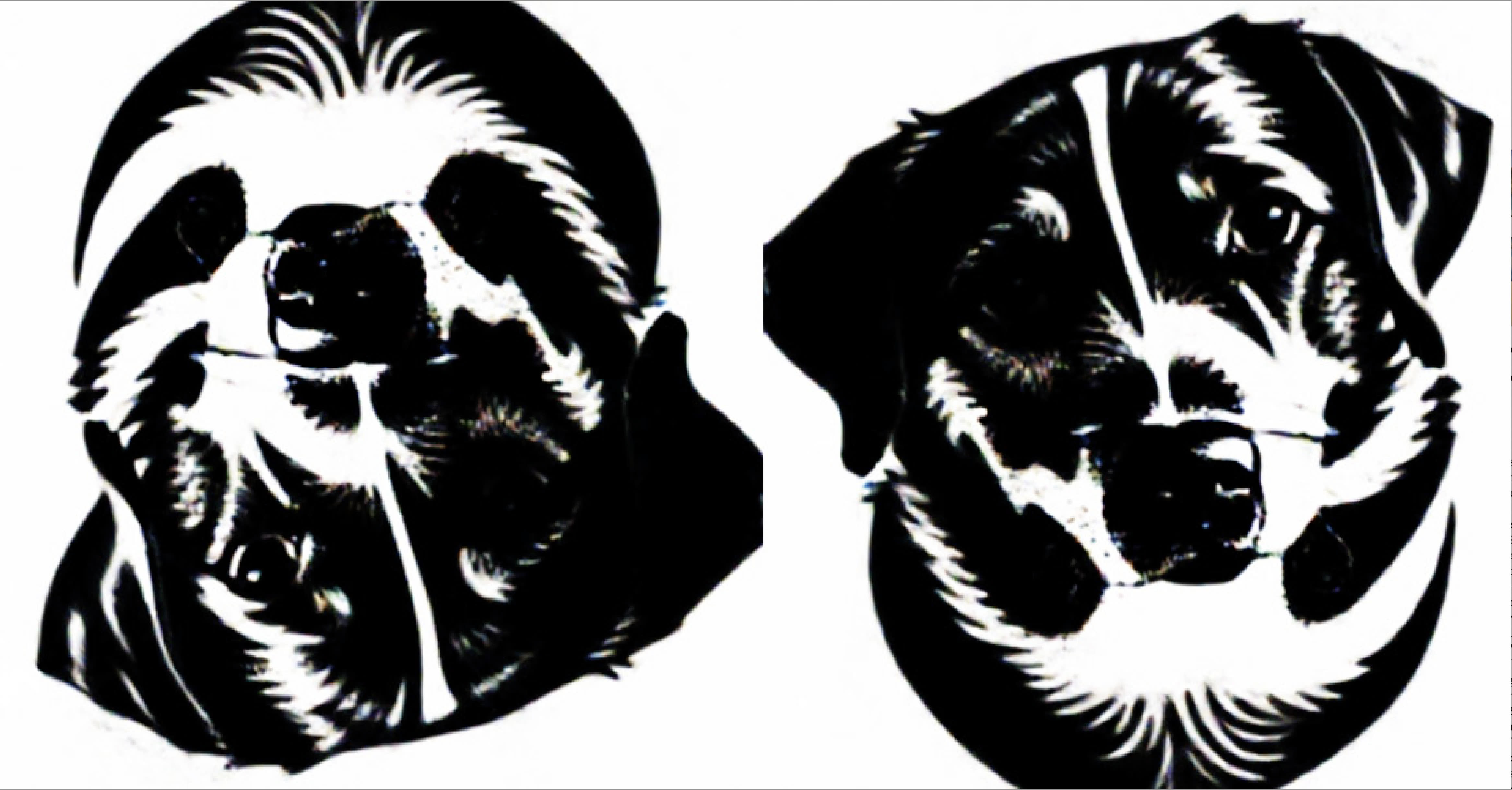}
        \caption{sloth  and  dog ~\cite{Rombach2021HighResolutionIS}.}
    \end{subfigure}
    \hfill
    \begin{subfigure}{0.24\linewidth}
        \includegraphics[width=\linewidth, trim={0 40 900 60}, clip]{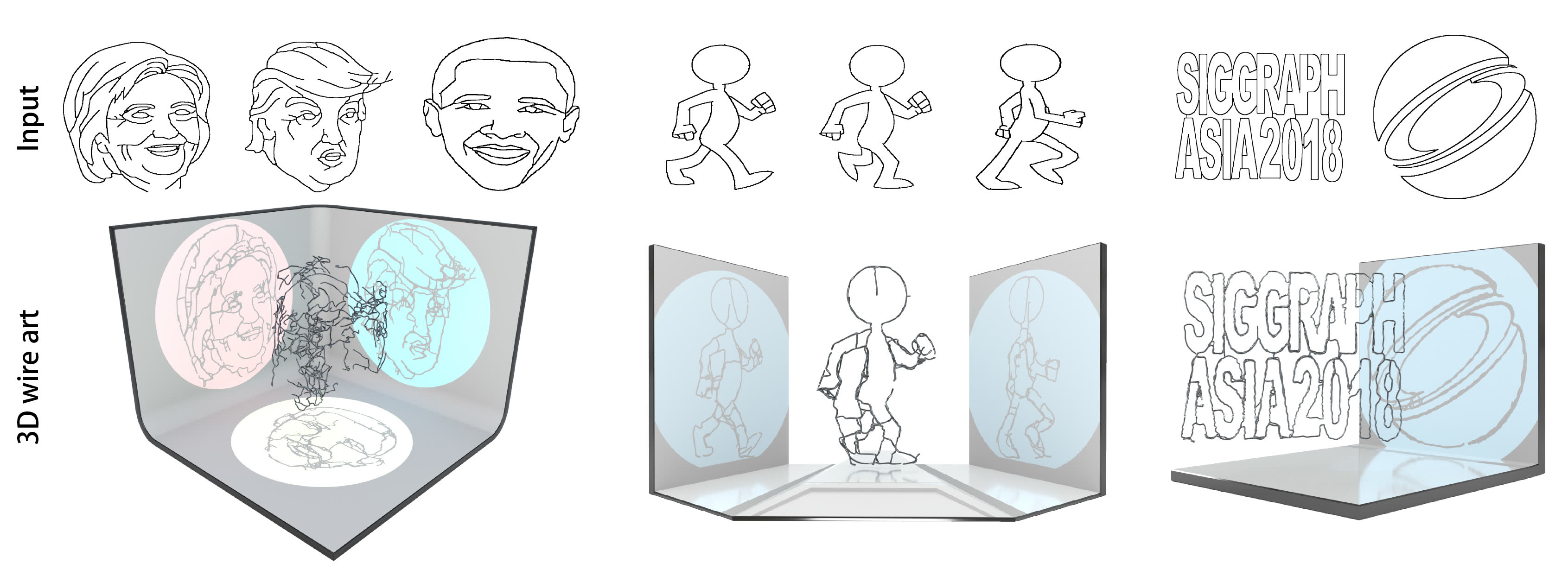}
        \caption{Multiview wire art \cite{Hsiao2018MultiviewWA}.}
    \end{subfigure}
    \hfill
    \begin{subfigure}{0.24\linewidth}
        \includegraphics[width=\linewidth, trim={70 27 30 17}, clip]{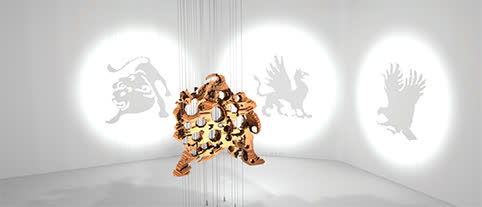}
        \caption{Shadow art~\cite{mitra2009shadow}.}
    \end{subfigure}
    \hfill
    \begin{subfigure}{0.24\linewidth}
        \includegraphics[width=\linewidth, trim={0 0 0 0}, clip]{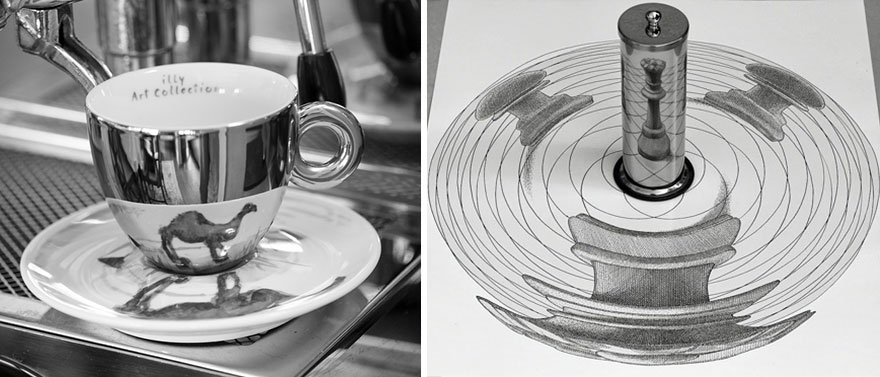}
        \caption{Reflective surface art~\cite{boredpanda_anamorphic}.}
    \end{subfigure}
    \vspace{-2mm}
    \caption{\textbf{2D and 3D Illusions. Each subfigure highlights different types of visual effects.}
    (a) 2D flip illusion ~\cite{Rombach2021HighResolutionIS}.
    Existing 3D illusions typically form line drawings~\cite{Hsiao2018MultiviewWA} (b) or figure-ground images like shadows~\cite{Mitra2009ShadowA} (c). 
    In contrast, our work expands the capability to generate 2D color images from different viewpoints. 
    (d) When placed on a textured surface, a reflective surface can reveal new content~\cite{boredpanda_anamorphic}. 
    However, it takes substantial time and effort for an artist to develop such a work. 
    Our method simply describes the process and supports generating more than one single view.}
    \label{fig:related}
    \vspace{-5mm}
\end{figure*}

\section{Related Work}
\label{sec:related}


\topic{2D illusion}
The ambiguous and illusory figures have long been studied by psychologists~\cite{Boring1930ANA, Long2004EnduringII, Bach2009ObjectPW, Kornmeier2011AmbiguousF}---human perception allows multiple interpretations of the same image~\cite{Nicholls2018PerceptionOA, Wimmer2007ChildrensPA}, influenced by factors such as stimulus duration~\cite{Wilton1985TheRE}, reaction time~\cite{Sperandio2010IsSR, Plewan2012TheIO}, critical features~\cite{Georgiades1997BiasingEI}, sensory memory~\cite{Pearson2008SensoryMF}, and predictive processing~\cite{Pepperell2023BeingAT}.
In computer vision, algorithms have been developed to create hybrid~\cite{Oliva2006HybridI} and mosaic images~\cite{Xu2019DiscernibleIM}, and recent work extends visual illusions to CNNs~\cite{Hertzmann2020VisualII}. Camouflage images, created using models like deep learning~\cite{Zhang2020DeepCI}, GANs~\cite{Guo2022GANmouflage3O}, and diffusion models~\cite{Luo2023CamDiffCI}, hide objects within scenes. Differentiable algorithms have also been used to design perceptual puzzles~\cite{Chandra2022DesigningPP}. In addition, there are also studies that quantify ambiguities in illusions~\cite{Wang2020TowardQA} and explore generative classifiers~\cite{Jaini2023IntriguingPO}. These studies~\cite{Ngo2023IsCF} reveal that models like CLIP~\cite{Ramesh2022HierarchicalTI} can also be misled by optical illusions.

Traditional approaches to making these illusions are highly time-consuming and require a high level of expertise. 
The development of generative AI, particularly diffusion models~\cite{Rombach2021HighResolutionIS}, has simplified the generation of high-quality images from text prompts. 
Recent works~\cite{Burgert2023DiffusionIH, Geng2023VisualAG} utilized diffusion models to generate 2D illusion with individual views aligned with different text prompts (Fig. ~\ref{fig:related}a). 
These methods utilize diffusion models to generate illusion artwork in two different ways. One~\cite{Geng2023VisualAG} simultaneously denoises the RGB image from different views using a pixel-based diffusion model~\cite{MikhailDeepfloyd2023}. However, applying a similar inference pipeline to 3D generation presents significant challenges. This limitation arises because of the lack of multiview consistent Gaussian noise. 
The other~\cite{Burgert2023DiffusionIH} uses an optimization-based method to update an RGB image representation with Score Distillation Sampling (SDS) loss~\cite{ruiz2023dreambooth} or Variational Score Distillation (VSD) loss~\cite{wang2023prolificdreamer}. 
We build upon an optimization-based framework and extend their capability to 3D illusion generation. 

\topic{3D illusion}
The perception of 3D objects may involve various forms of illusions. 
%
Different viewing angles can lead to varying interpretations of the same object~\cite{keiren2009constructability,sela2007generation,10.1145/3658231}.
Additionally, the interpretation of 2D shadows cast by 3D surfaces can vary based on the light source's position, affecting figure-ground perception~\cite{Sadekar2021ShadowAR,ChaineHandSA,Mitra2009ShadowA,Min2017SoftSA} (see Fig. \ref{fig:related}b).
Similarly, the interpretation of line drawings of 3D wires can also create illusions~\cite{Qu2023WiredPM, Xu2021ModelguidedED, Yeh2022GeneratingVW, Hsiao2018MultiviewWA} (see Fig. \ref{fig:related}c).
These methods typically rely on simple images—basic figure-ground forms, line drawings, or abstract shapes, often limited to just three views. In contrast, our work creates colorful 3D illusions, complex shapes, detailed interpretations, and extends to eight views (Fig. \ref{fig:8prompts}), advancing the craft of 3D illusion creation. Reflective art—like mirror and cylinder illusions (Fig. \ref{fig:related}d)—creates fascinating effects but is often confined to a single view. Our approach, however, generates multiple perspectives—top, side, and even two reflective cylinders/mirrors, offering three distinct views, transcending the limitations of human vision.

\topic{3D generation using diffusion models}
Diffusion models have demonstrated significant capability in generating photorealistic 2D images. Recently, they have been widely adopted for 3D generation, overcoming the need for large labeled 3D datasets~\cite{poole2022dreamfusion, lin2022magic3d, metzer2022latent, wang2023prolificdreamer, Chen_2023_ICCV,mcallister2024rethinking}.
DreamFusion~\cite{poole2022dreamfusion} leveraged text-to-image diffusion models for 3D synthesis by employing a probability density distillation (SDS) loss. 
However, the resulting images often have low resolution, excessive saturated colors, and over-smoothing issues.
Building on this, Magic3D~\cite{lin2022magic3d} achieved high-quality 3D mesh models through a two-stage optimization framework, significantly improving the quality of the generated 3D models.
ProlificDreamer~\cite{wang2023prolificdreamer} further advanced the field by introducing Variational Score Distillation (VSD) loss to address the saturation and smoothing problems associated with the SDS loss. This approach enhances the diversity and quality of the generated samples.
We utilize Variational Score Distillation (VSD) as our baseline to generate our 3D multiview illusions using text-to-image diffusion models, explicitly acknowledging that this approach is not a novel contribution of our work.

\begin{figure*}[t]
    \vspace{-8mm}
    \centering
    \includegraphics[width=\linewidth, trim=1 0 0 0 , clip]{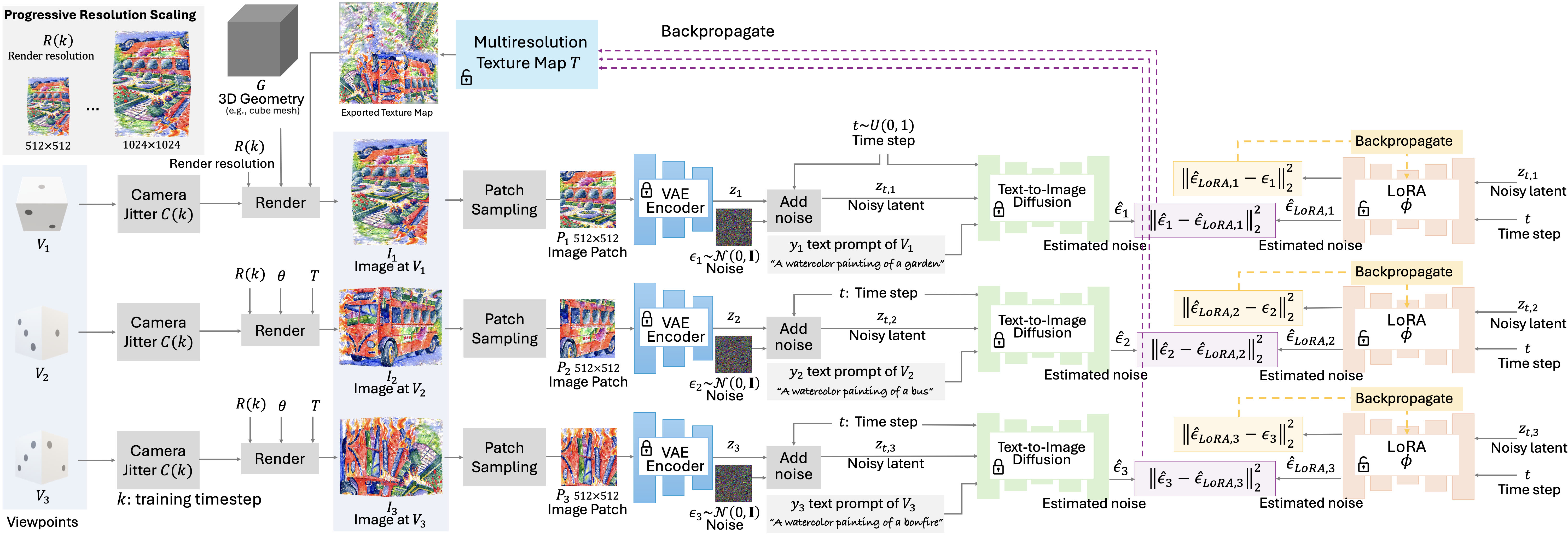}
    \vspace{-6mm}
    \caption{
    \textbf{Method overview.}
    We illustrate the process of generating a 3D multiview illusion from a cube with three varying interpretations from different viewpoints ($V_1$, $V_2$, and $V_3$) guided by text prompts ($y_1$, $y_2$, and $y_3$). 
    First, we render the cube from the target viewpoints $V_i$ applying scheduled camera jitter $C(k)$ and scheduled render size $R(k)$ with respect to the gradient flow time $k$. 
    Camera jitter improves generation quality and render size scheduling helps reduce the duplicate pattern issue issue. 
    We then utilize the multi-resolution texture field $T$ to obtain the images ($I_1$, $I_2$, and $I_3$) at resolution $R(k) \in[512, 1024]$.
    To increase resolution during training, we extract a random patch $P_i$ of size $512 \times 512$ from each rendered image $I_i$, which is then fed into a pre-trained VAE encoder.
    Given the 3D shape, we aim to optimize the parameters of the texture field. 
    The generation of these viewpoints is optimized by leveraging a text-to-image diffusion model guided by the text prompts ($y_1$, $y_2$, and $y_3$). 
    To avoid unnatural, saturated colors, we apply Variational Score Distillation (VSD) and LoRA model. We apply the same settings for spheres and scenes with reflective surfaces.}
    \label{fig:overview}
    \vspace{-5mm}
\end{figure*}
\section{Method}
\label{sec:method}

Given multiple text prompts or images, we aim to create 3D illusions that reveal distinct interpretations from different viewing angles, with each interpretation corresponding to a specific text prompt or reference image. 
To render different views of the same object, we either move the camera positions or use reflective surfaces.
Here, we present our approach for optimizing rendered views to align accurately with their corresponding text or visual descriptions. 
Our method leverages the capabilities of pretrained diffusion models with several novel techniques and design choices to enhance the quality of these 3D illusions.

\subsection{Background}
\label{sec:background}
Given a 3D representation, we optimize the rendering of each viewpoint $V_i$ to be aligned with a specific text prompt $y_i$. 
For example, to generate a 3D illusion with a cube, we texture the surface of a 3D cube as a neural representation to optimize. We begin by selecting a corner of the cube and identifying the three adjacent faces connected to it. Each view $V_i$ is configured to display any two adjacent faces, allowing us to associate three distinct text prompts with the corresponding views. 
This setup ensures that each face is visible in two different views, creating overlapping regions on the cube that are jointly optimized to align with two prompts.


We parameterize the texture field using a neural network with parameters \( \theta \) of the 3D scene. 
This network employs a hash encoding MLP proposed by InstantNGP~\cite{Mller2022InstantNG} to embed the texture features to support multi-resolution rendering. 
We denote the hashmap MLP as a texturing function \( f\), such that the queried RGB image \( T \) is given by  \( T = f(\mathcal{E}(q); \theta) \), where \( q \) is the query coordinate and \(\mathcal{E}\) is the embedded texture map feature. 
To optimize our texturing module \( f(\theta) \) for generating visual illusions, we leverage 2D diffusion priors from a pre-trained Stable Diffusion model~\cite{Rombach2021HighResolutionIS} via a 
Score Distillation Sampling~\cite{poole2022dreamfusion} is a common method for utilizing 2D diffusion priors, it often produces over-smoothing and color over-saturation artifacts. 
To mitigate these issues, we adopt the Variational Score Distillation (VSD) method~\cite{wang2023prolificdreamer}.
The VSD gradient is computed using a pre-trained diffusion model \( \phi_{\text{pretrained}} \) along with a trainable LoRA module \( \phi \):
\begin{equation}
L_{VSD}(\theta) = \mathbb{E}_{t, \epsilon}\left[w(t)(\epsilon_{\phi_{\text{pretrained}}} - \epsilon_{\phi}) \frac{\partial f(\theta)}{\partial \theta}\right],
\end{equation}
where $f(\theta)$ is the rendered image, and \( \epsilon_{\phi_{\text{pretrained}}} = \phi_{\text{pretrained}}(f(\theta); z, t) \) and \( \epsilon_{\phi} = \phi(f(\theta); z, t) \). 
The time step \( t \) is randomly sampled from \( t \sim \mathcal{U}(0.02, 0.98) \), and \( z \) represents the noisy input to the model with the injected noise following \( \epsilon \sim \mathcal{N}(0, 1) \).  
The weighting function \( w(t) \) adheres to the VSD configuration.

The optimization alternately updates the rendering module $\theta$ and the LoRA weights.
The VSD gradient above is used to update the texture map parameters $\theta$, while the LoRA module \( \phi \) is optimized with the following objective by fine-tuning the diffusion model on current renderings:

\begin{equation}
L_{LoRA}(\phi) = \min_{\phi}  \sum_{i=1}^{n} \mathbb{E}_{t, \epsilon}\left[\|\epsilon_{\phi}(f(\theta); z, t) - \epsilon\|_2^2\right].
\end{equation}


Generating 3D illusions with this baseline introduces several issues, as illustrated in Fig.~\ref{fig:comparison} column 2. These artifacts mainly arise from the under-constrained nature of 3D optimization. In the following, we discuss the primary challenges and the methods we developed to address these issues.

\subsection{Camera Jitter}
Optimizing the texture representation with VSD loss from a certain camera view can induce high-frequency color artifacts. 
Interestingly, such artifacts are rarely observed in text-to-image and also less observed in text-to-3D~\cite{wang2023prolificdreamer}. When computing the VSD gradient $\frac{\partial f(\theta)}{\partial \theta}$, we need to backpropagate the gradient to the texture map through the VAE encoder of the latent diffusion model. We hypothesize that these artifacts arise from the presence of a ``blind spot'' in the encoder. Specifically, the encoder's latent representation does not change significantly with or without these artifacts. In contrast, text-to-3D approaches suffer less from this issue due to dense camera view sampling during training. The artifacts in the ``blind spot'' of one view can be optimized in other multiple views.

\begin{figure}[htbp]
\vspace{-2mm}
\centering
\footnotesize

\begin{minipage}[c]{0.03\linewidth}
  \makebox[0pt][c]{\rotatebox{90}{\parbox{3cm}{\centering \textbf{w/o camera jitter}}}}
\end{minipage}%
\begin{minipage}[c]{0.97\linewidth}
  \centering
  \includegraphics[trim=70 70 70 70, clip, width=0.33\linewidth]{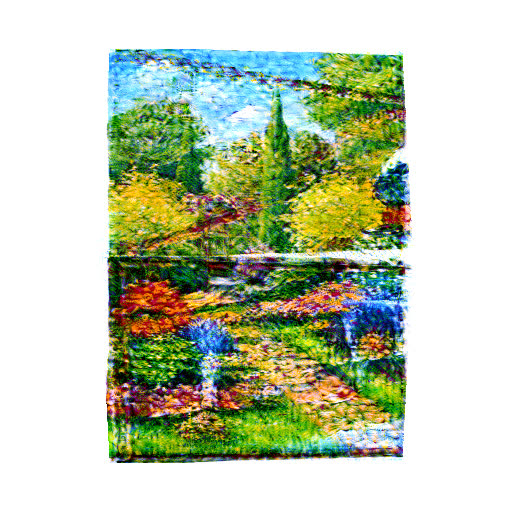}\hfill
  \includegraphics[trim=70 70 70 70, clip, width=0.33\linewidth]{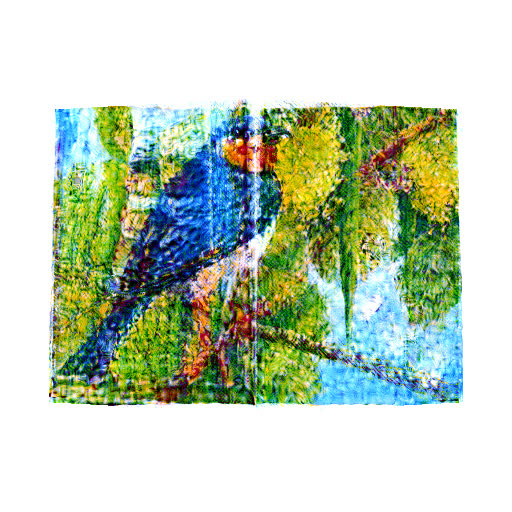}\hfill
  \includegraphics[trim=70 70 70 70, clip, width=0.33\linewidth]{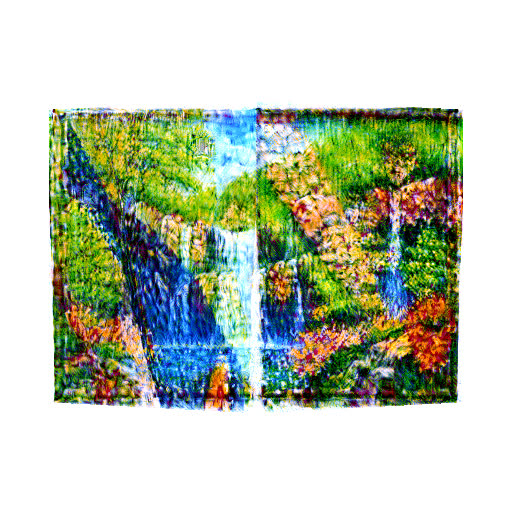}\hfill

\end{minipage}\\
\vspace{-.25cm}
\begin{minipage}[c]{0.03\linewidth}
  \makebox[0pt][c]{\rotatebox{90}{\parbox{3cm}{\centering \textbf{with camera jitter}}}}
\end{minipage}%
\begin{minipage}[c]{0.97\linewidth}
  \centering
  \includegraphics[trim=70 70 70 70, clip, width=0.33\linewidth]{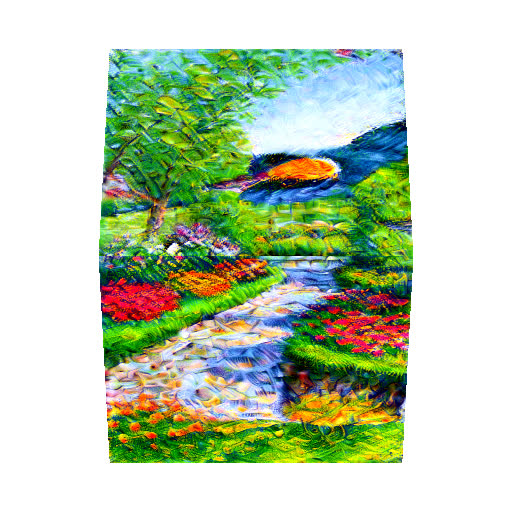}\hfill
  \includegraphics[trim=70 70 70 70, clip, width=0.33\linewidth]{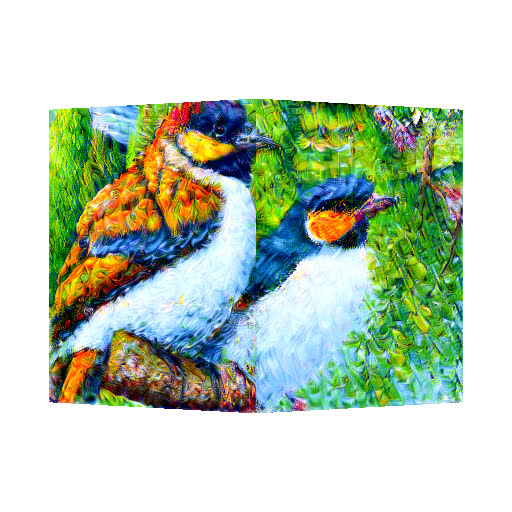}\hfill
  \includegraphics[trim=70 70 70 70, clip, width=0.33\linewidth]{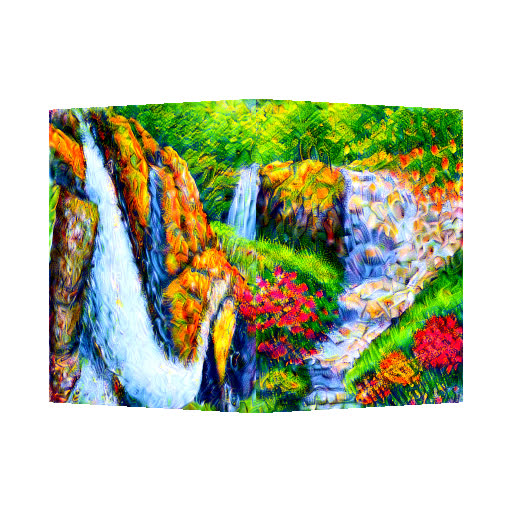}\hfill

\end{minipage}\\

\vspace{-1mm}
\begin{minipage}[c]{0.03\linewidth}
  \makebox[0pt][c]{}
\end{minipage}%
\begin{minipage}[c]{0.32\linewidth}
    \centering
    \includegraphics[width=0.6\linewidth, trim=490 20 490 30, clip]{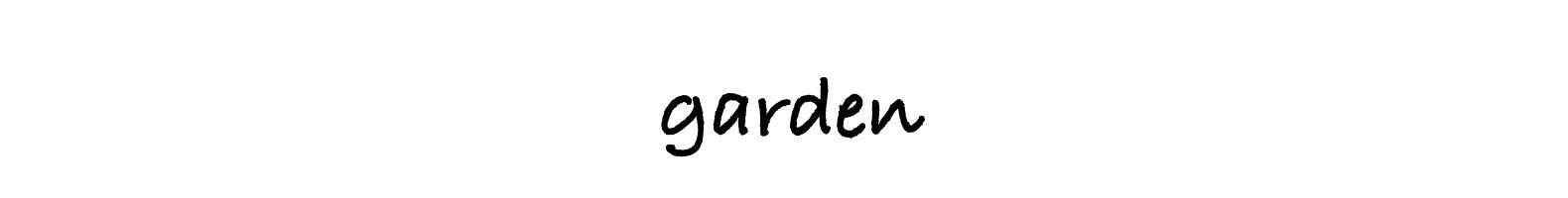}
\end{minipage}\hfill
\begin{minipage}[c]{0.32\linewidth}
    \centering
    \includegraphics[width=0.2\linewidth, trim=1030 20 400 30, clip]{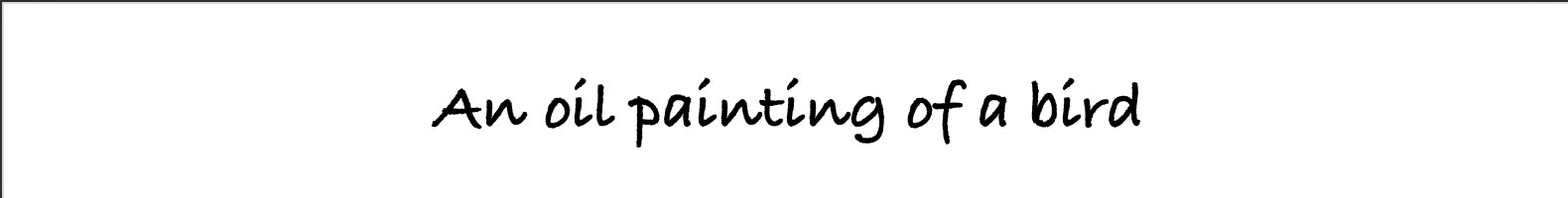}
\end{minipage}\hfill
\begin{minipage}[c]{0.32\linewidth}
    \centering
    \includegraphics[width=0.4\linewidth, trim=950 20 300 30, clip]{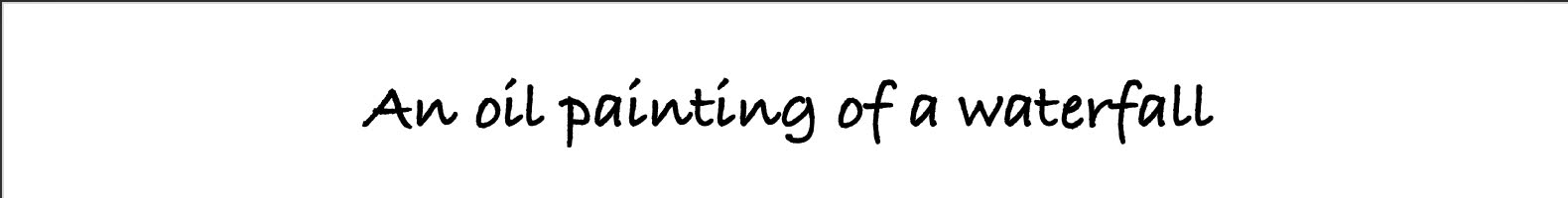}
\end{minipage}\\

\captionsetup{skip=-0.pt}
    \caption{\textbf{Camera jitter can make the result smoother.} Adding camera jitter to the baseline method can make the transition between views more seamless but may also introduce two primary contents in the generation.}
\vspace{-4mm}
\label{fig:camera-jitter}
\end{figure}

Based on this observation, we introduce a random offset to the camera parameters while optimizing our 3D illusion from certain views. 
As shown in Fig.~\ref{fig:camera-jitter}, the introduced camera shifts address the grid-like artifacts and produce smoother results.
Applying this camera jitter technique from the start of training, however, exacerbates another issue, where the optimization creates duplicate objects in the view.
To mitigate this issue, we constrain the camera jitter and \emph{gradually} increase its level as training progresses. We tested two scheduling strategies for increasing camera jitter, linear and sigmoid schedules, and found that a linear schedule generally performs better in practice.

We denote the camera jitter level as $C(k)$, which is a linear function of the current training timestep $k$:
\begin{equation}
C(k) = C_{\text{max}} \cdot \frac{k}{k_{\text{total}}}
\label{eq:camerajitter},
\end{equation}
where  \( C_{\text{max}} \) represents the maximum camera jitter level, and \( k_{\text{total}} \) is total number of training steps
The camera parameters, rotation $\mathbf{R}$, translation $\mathbf{T}$, and field of view FOV at each step are:
\begin{equation}
\begin{aligned}
\mathbf{R} &= \mathbf{R}_{\text{initial}} + \mathcal{N}(0, C(k) \cdot \sigma_\mathbf{R}), \\
\mathbf{T} &= \mathbf{T}_{\text{initial}} + \mathcal{N}(0, C(k) \cdot \sigma_\mathbf{T}), \\
\text{FOV} &= \text{FOV}_{\text{initial}} + \mathcal{N}(0, C(k) \cdot \sigma_{\text{FOV}}),
\end{aligned}
\label{eq:camera_params}
\end{equation}
where $\mathbf{R}_{\text{initial}}$, $\mathbf{T}_{\text{initial}}$ and  $\text{FOV}_{\text{initial}}$ are the initial camera parameters, and \( \sigma_\mathbf{R} \), \( \sigma_\mathbf{T} \), and \( \sigma_{\text{FOV}} \) denote the standard deviations for rotation, translation, and field of view. 

\subsection{Resolution scaling with patch denoising}


Pre-trained Stable Diffusion models~\cite{Rombach2021HighResolutionIS} are limited to single-stage resolutions of  \(512\times512\)  or  \(768\times768\). 
To achieve higher-resolution \(1024\times1024\) generation results, one way is to adopt a separate image super-resolution model. However, this approach induces artifacts even within the context of 2D illusions. Specifically, a super-resolution model would operate solely on one view and result in the distortion of content when viewed from a different perspective. More discussion is in Supp. Sec. \ref{sec:ablationsupp}.
We refer to the single-stage \(512\times512\) illusion generation with VSD loss as the baseline method.

\begin{figure*}[htbp]
    \vspace{-8mm}
    \centering
    \fontsize{5.5pt}{6.5pt}\selectfont
    \setlength{\tabcolsep}{1pt} 
    \renewcommand{\arraystretch}{1}

    \begin{tabular}{lccccccc}
        &\textbf{Burgert \textit{et al.} ~\cite{Burgert2023DiffusionIH}} & \textbf{Baseline} & \textbf{Random patch}  & \textbf{w/o resolution scaling} & \textbf{w/o resolution scaling} &  \textbf{w/o camera jitter} & \textbf{Ours}\\

        \vspace{-2mm}
        \rotatebox{90}{\includegraphics[width=0.125\textwidth, trim=168 20 185 30, clip]{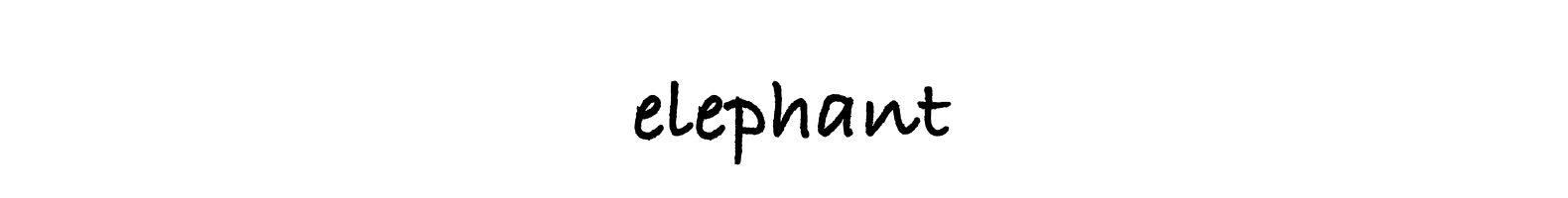}}
        & \includegraphics[width=0.125\textwidth, trim=70 70 70 70, clip]{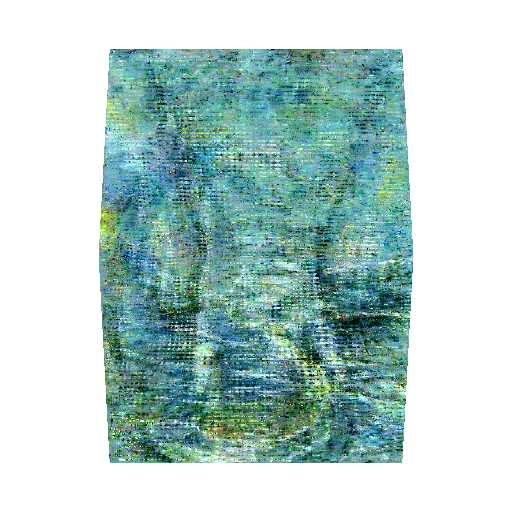} &
        \includegraphics[width=0.125\textwidth, trim=70 70 70 70, clip]{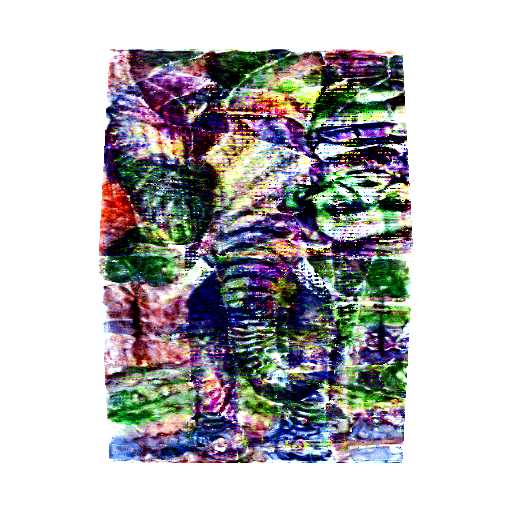} &
        \includegraphics[width=0.125\textwidth, trim=140 140 140 140, clip]{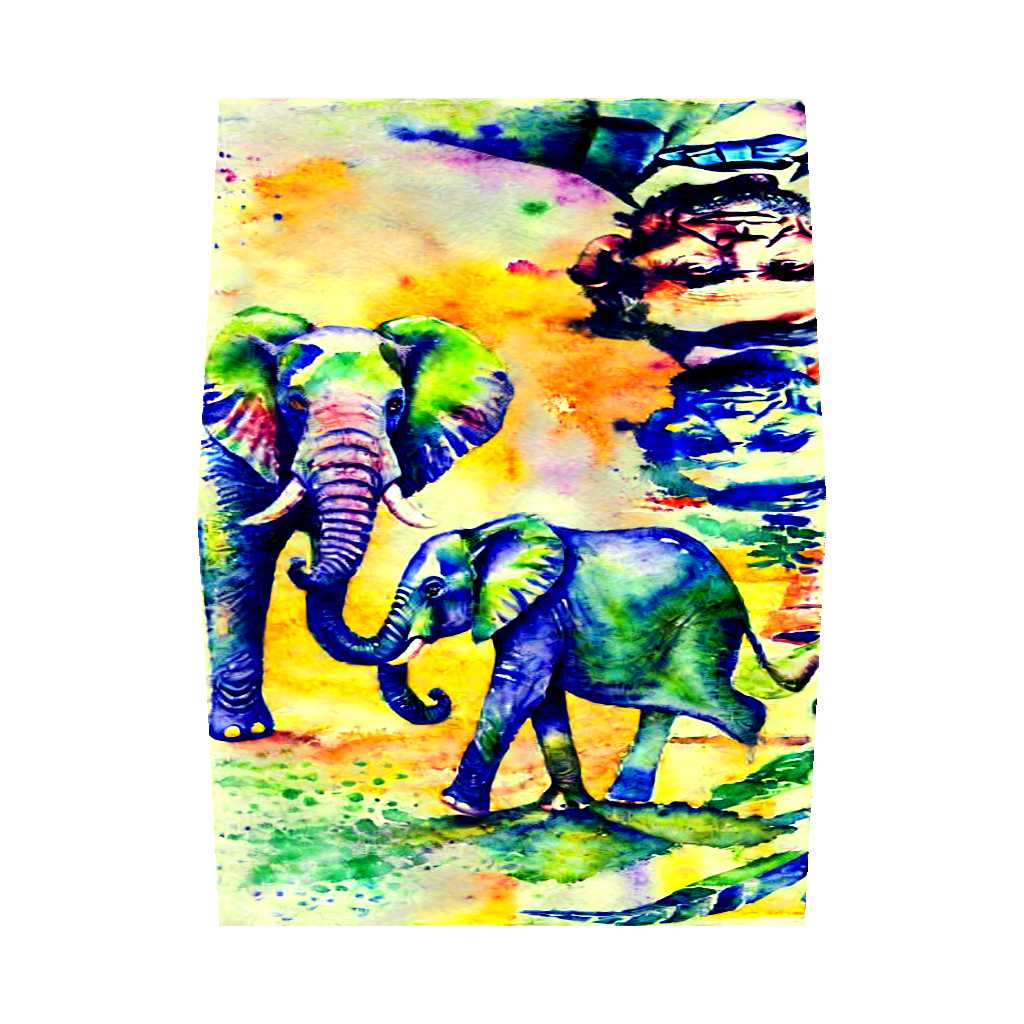} &
        \includegraphics[width=0.125\textwidth, trim=140 140 140 140, clip]{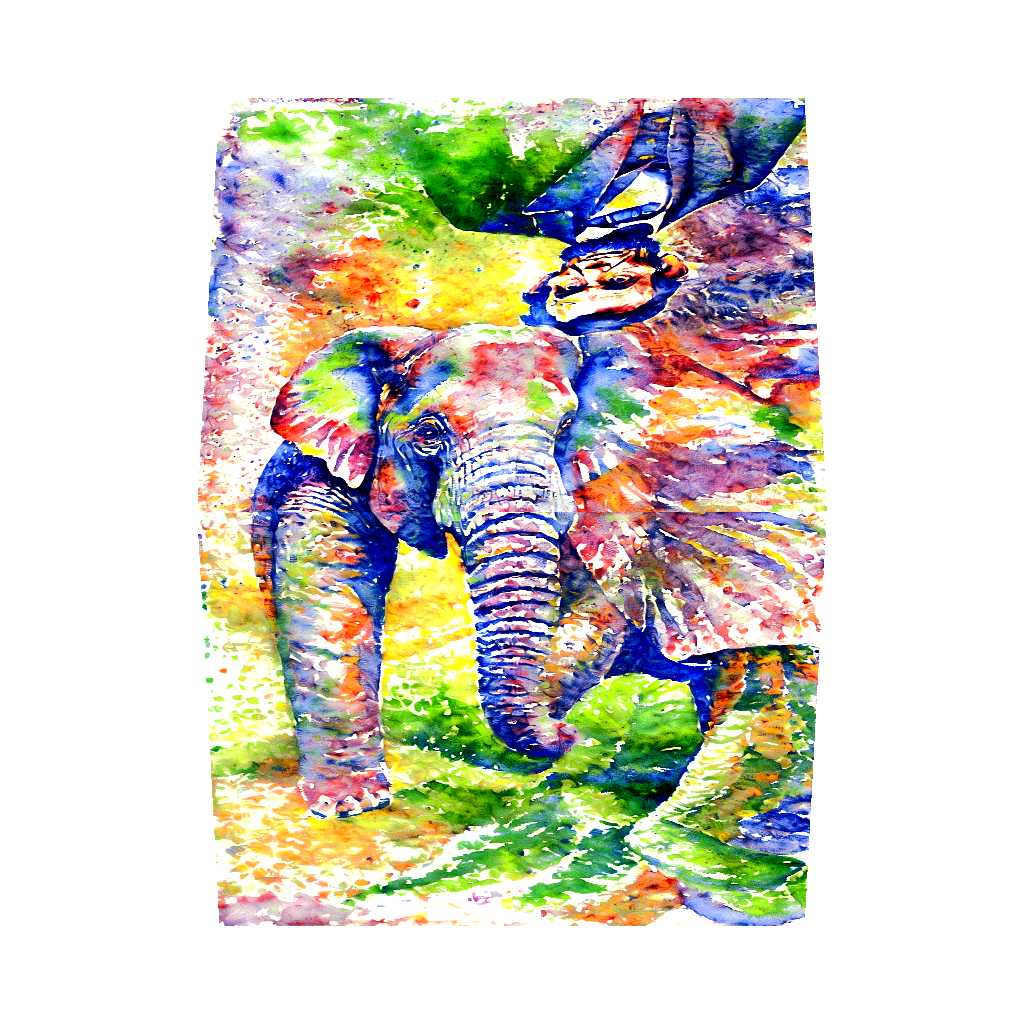} &
        \includegraphics[width=0.125\textwidth, trim=140 140 140 140, clip]{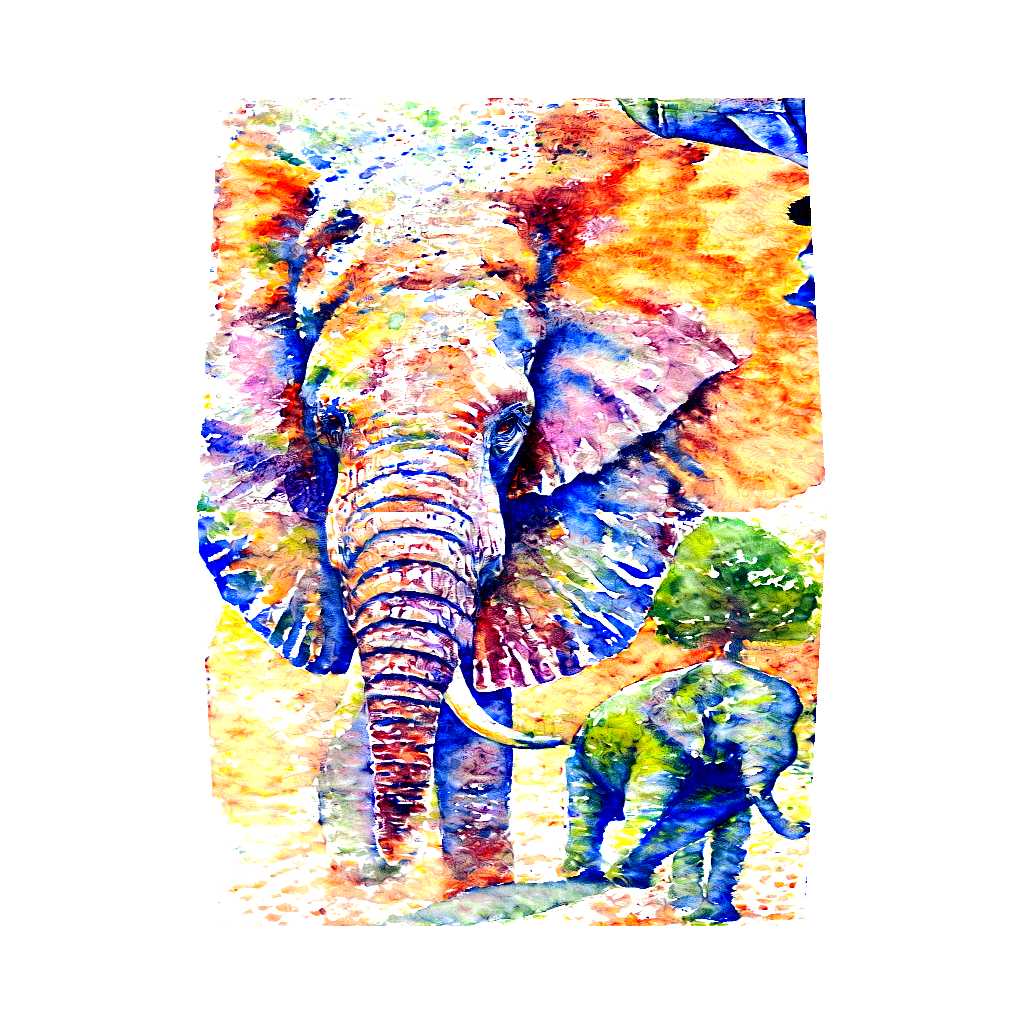} &
        \includegraphics[width=0.125\textwidth, trim=140 140 140 140, clip]{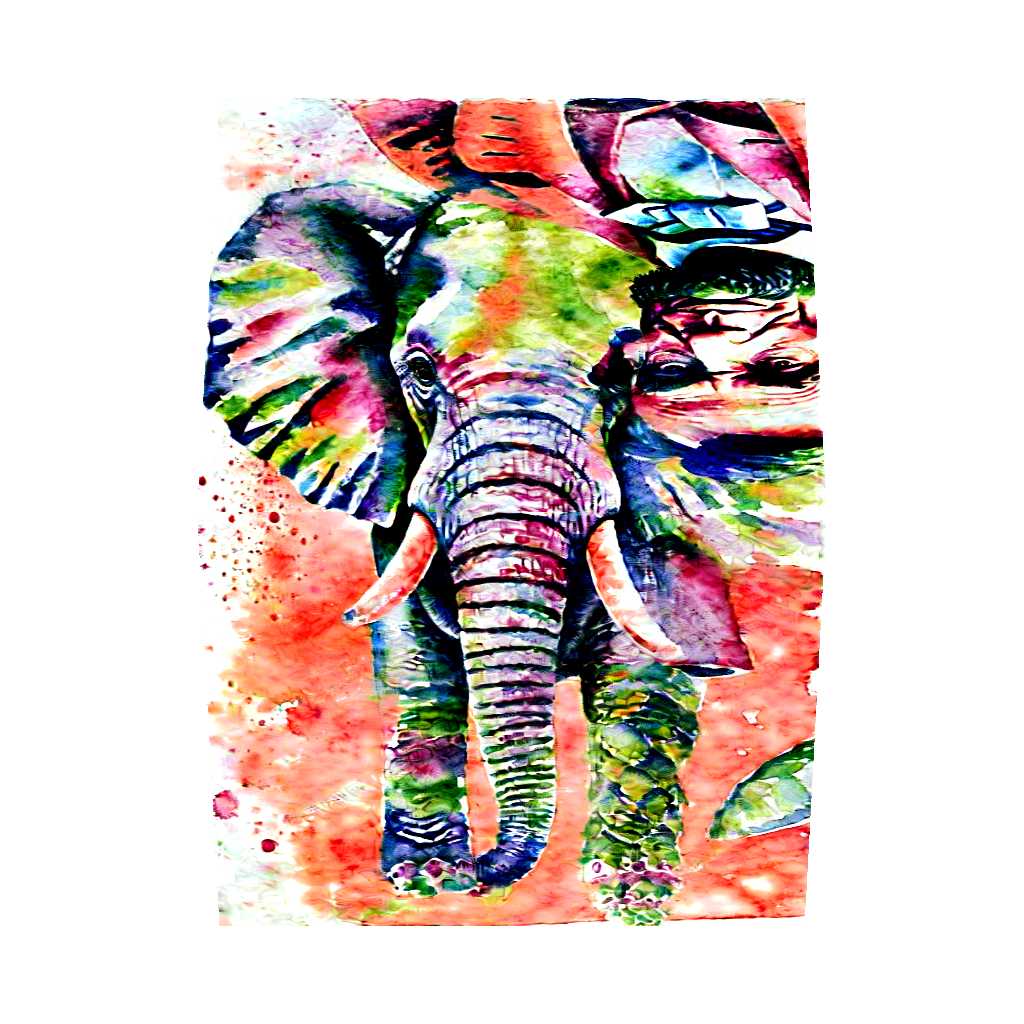} &
        \includegraphics[width=0.125\textwidth, trim=140 140 140 140, clip]{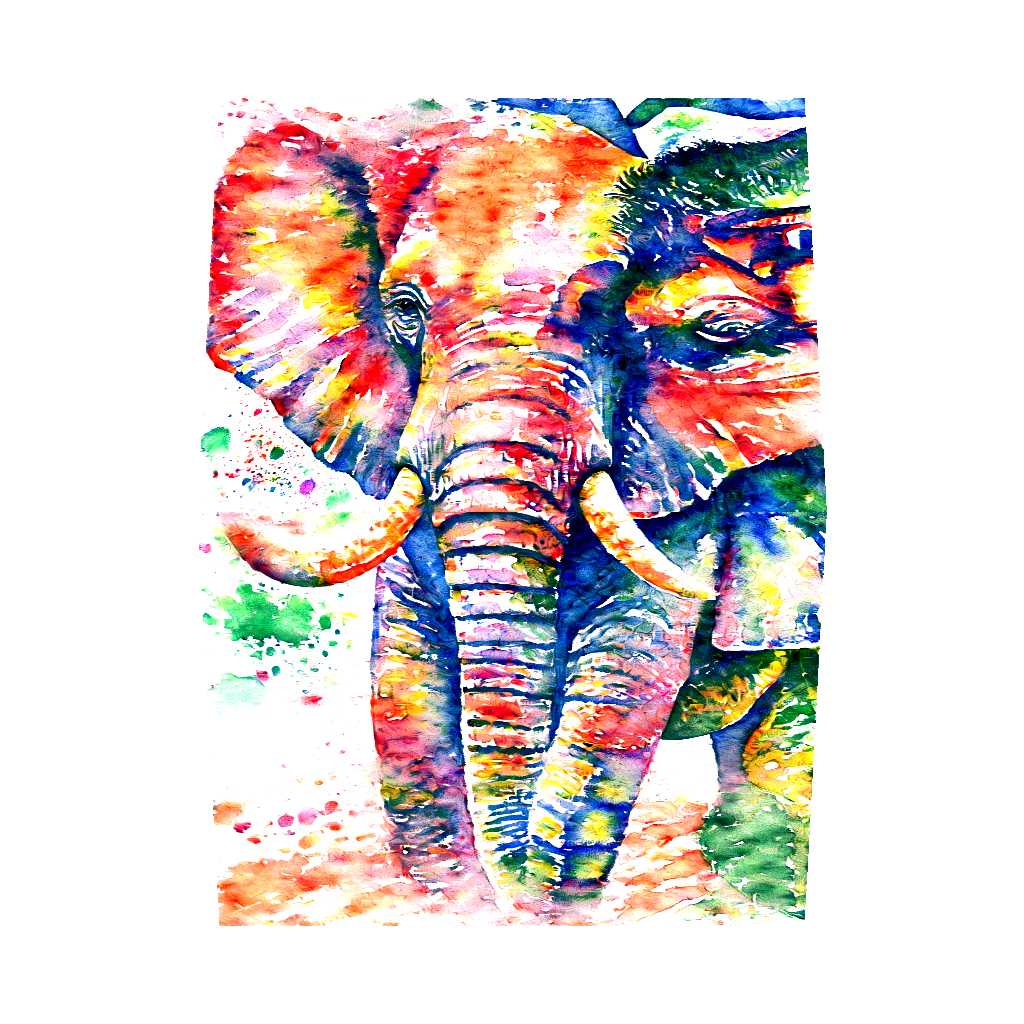} \\
        \vspace{-3mm}
        \rotatebox{90}{\includegraphics[width=0.125\textwidth, trim=168 20 185 30, clip]{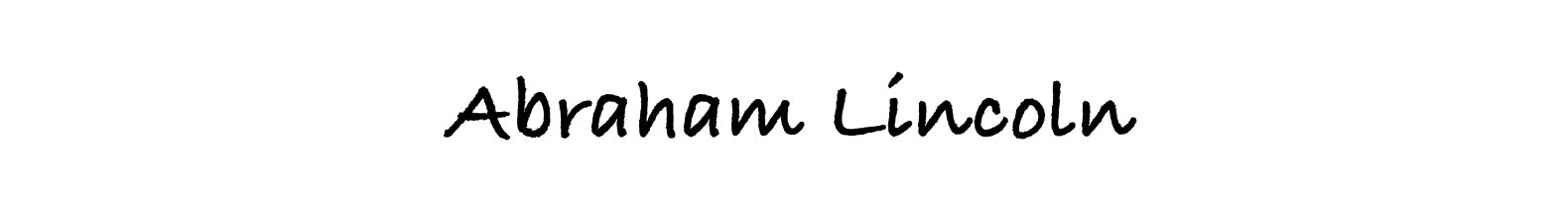}}
        & \includegraphics[width=0.125\textwidth, trim=70 90 70 90, clip]{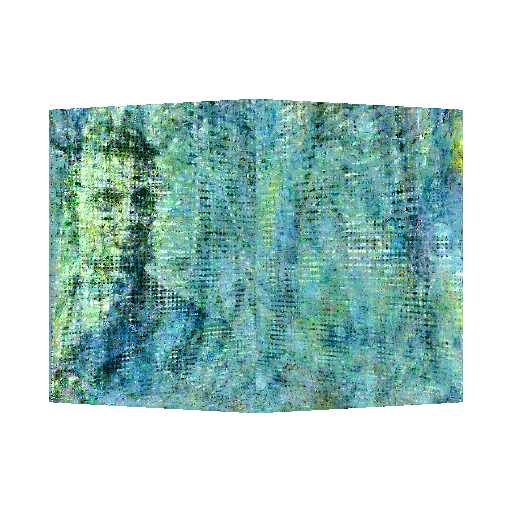} &
        \includegraphics[width=0.125\textwidth, trim=70 90 70 90, clip]{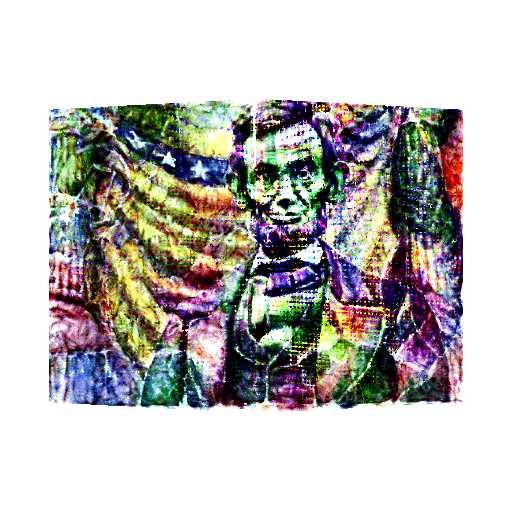} &
        \includegraphics[width=0.125\textwidth, trim=140 180 140 180, clip]{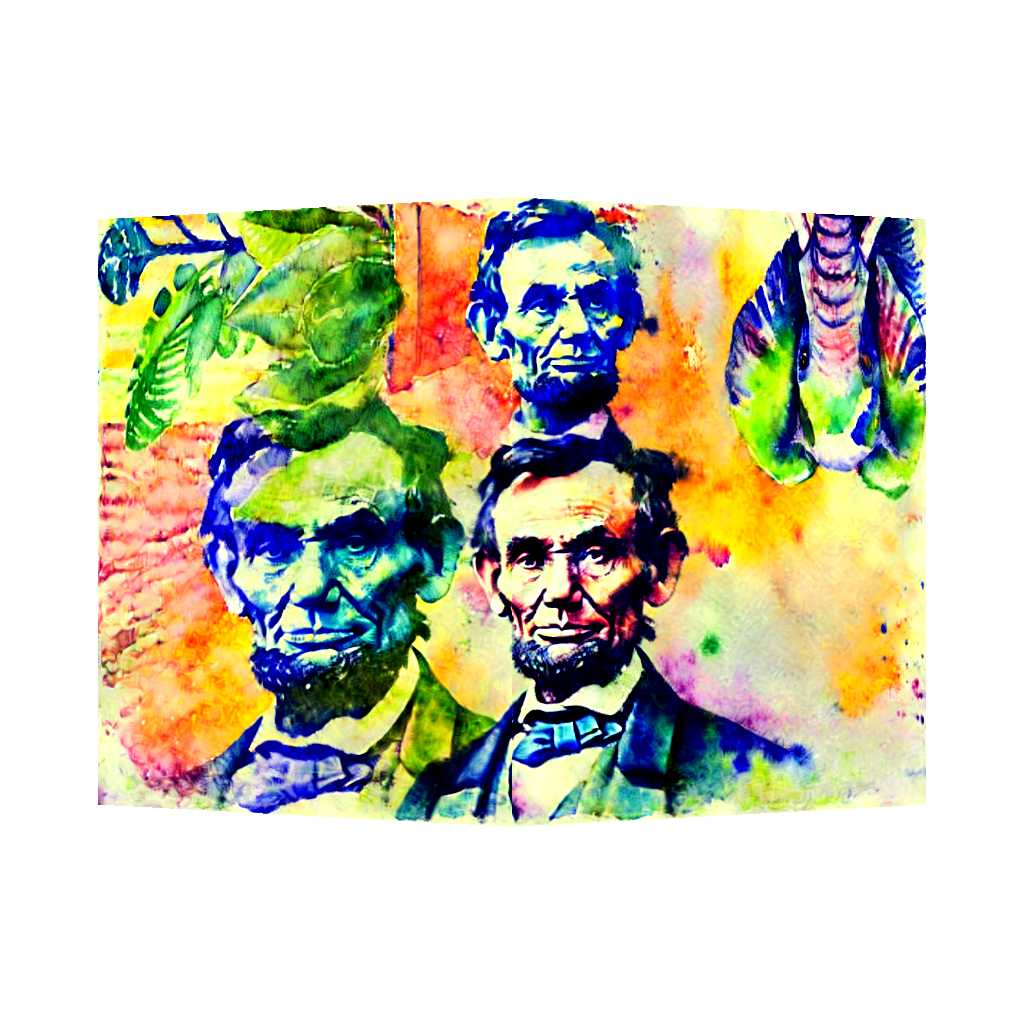} &
        \includegraphics[width=0.125\textwidth, trim=140 180 140 180, clip]{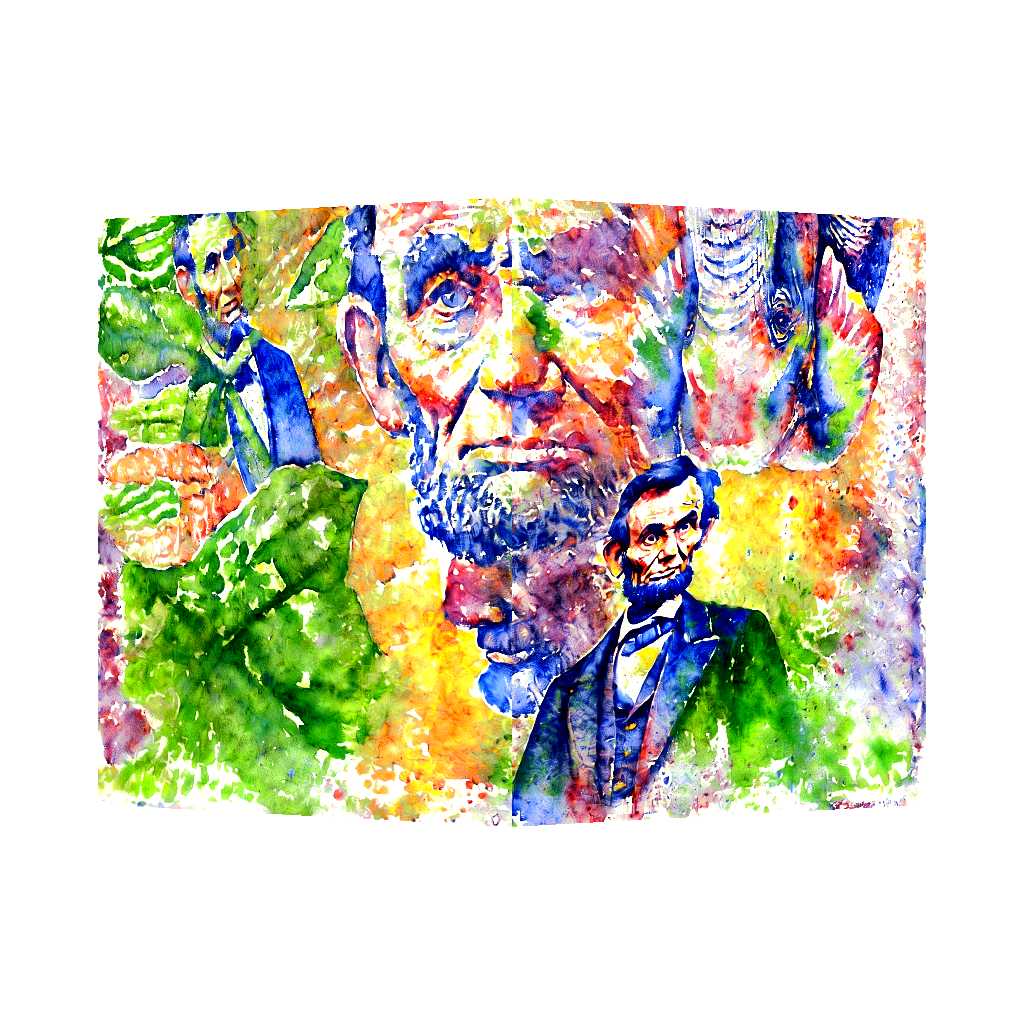} &
        \includegraphics[width=0.125\textwidth, trim=140 180 140 180, clip]{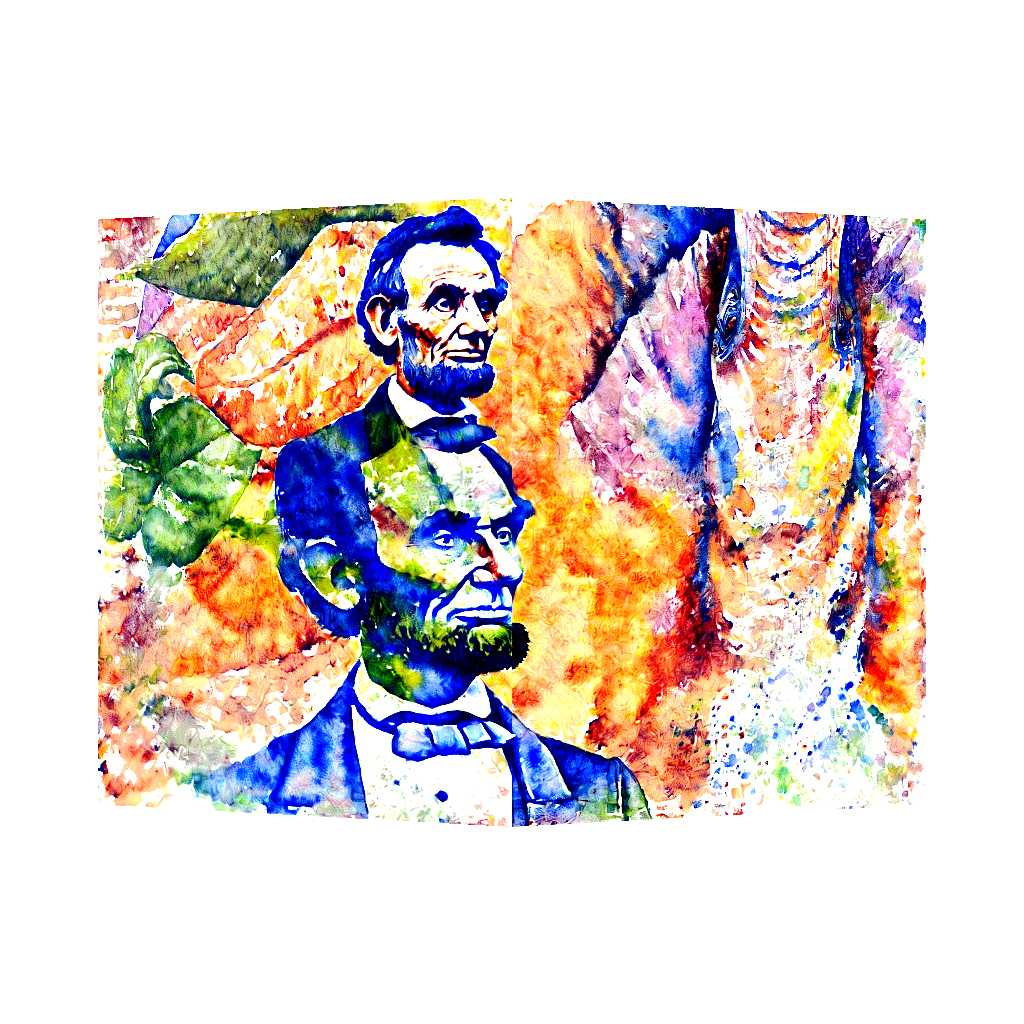} &
        \includegraphics[width=0.125\textwidth, trim=140 180 140 180, clip]{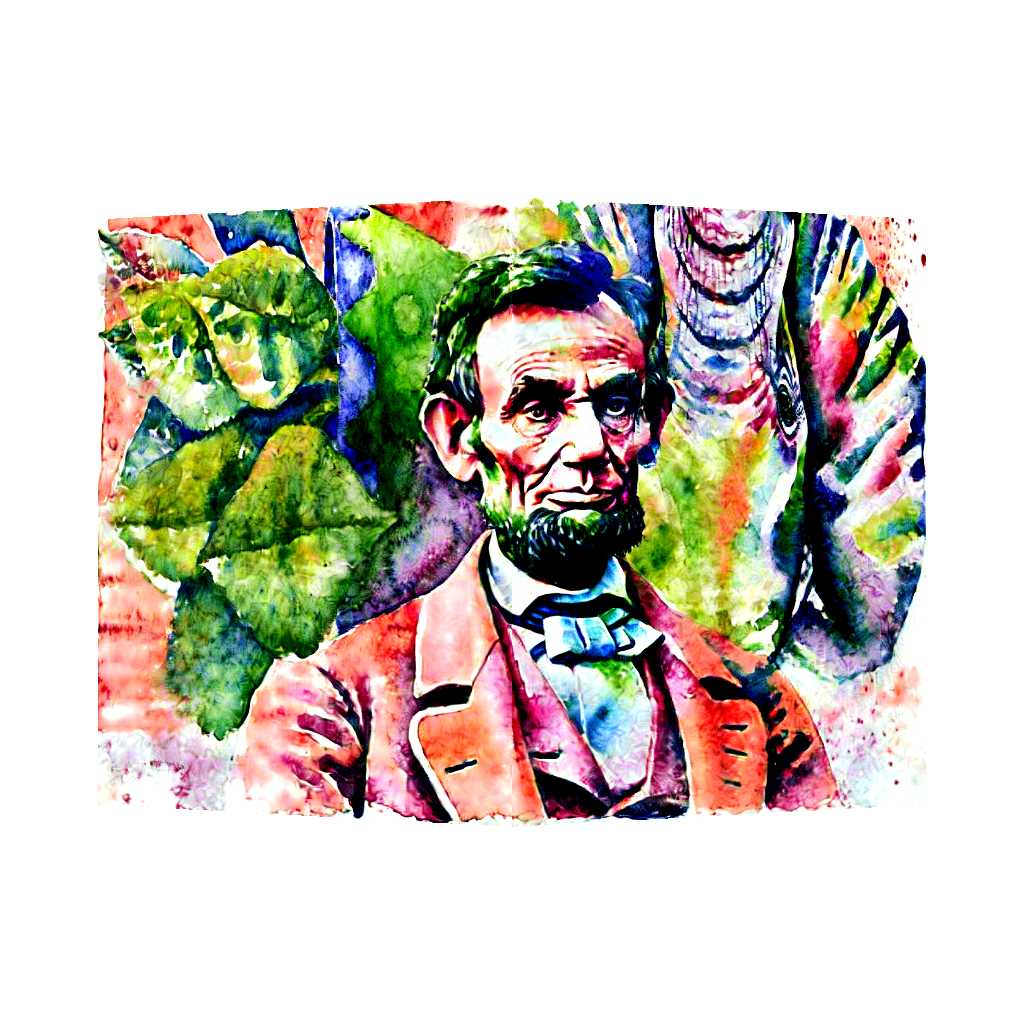} &
        \includegraphics[width=0.125\textwidth, trim=140 180 140 180, clip]{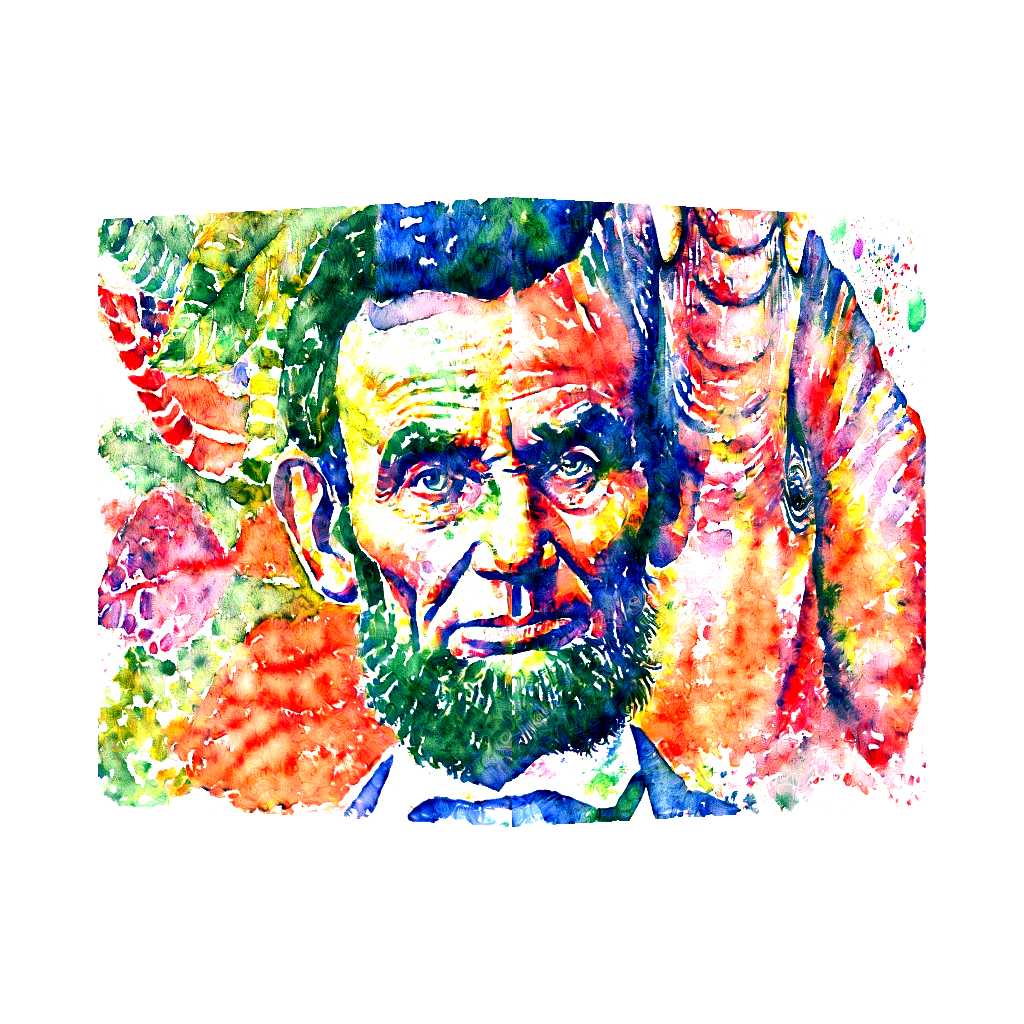} \\
        
        \rotatebox{90}{\includegraphics[width=0.125\textwidth, trim=168 20 185 30, clip]{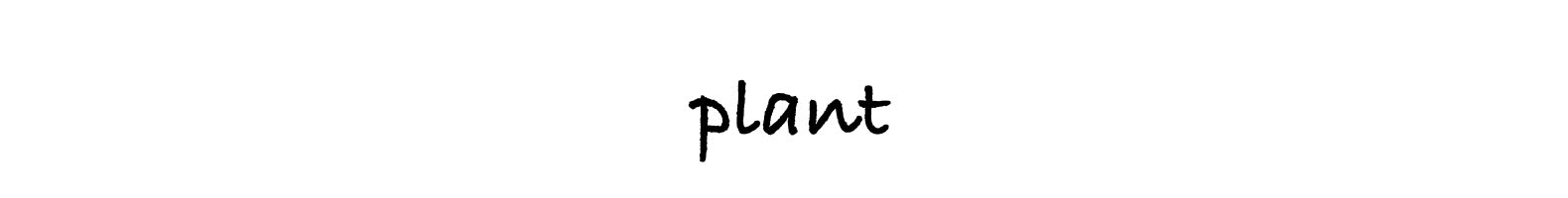}}
        & \includegraphics[width=0.125\textwidth, trim=70 90 70 90, clip]{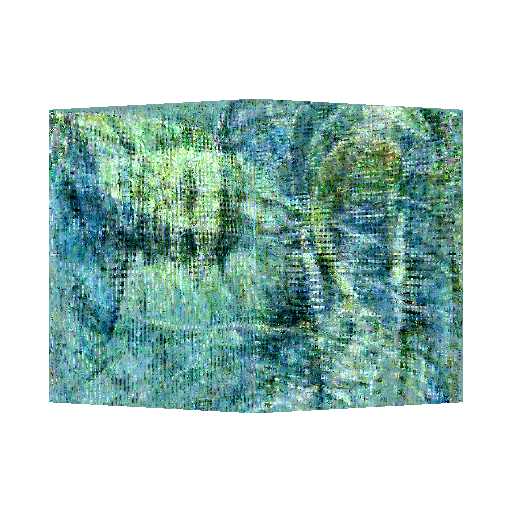} &
        \includegraphics[width=0.125\textwidth, trim=70 90 70 90, clip]{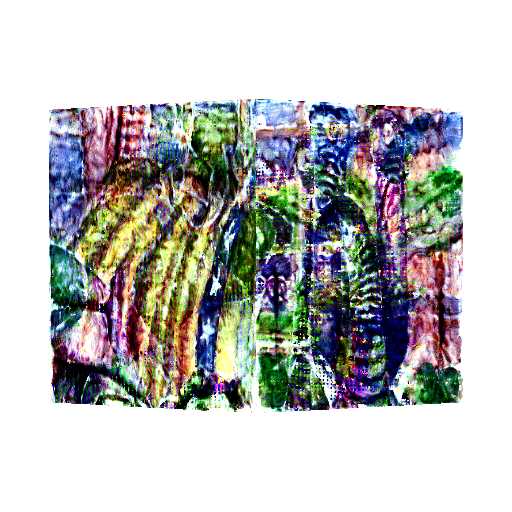} &
        \includegraphics[width=0.125\textwidth, trim=140 180 140 180, clip]{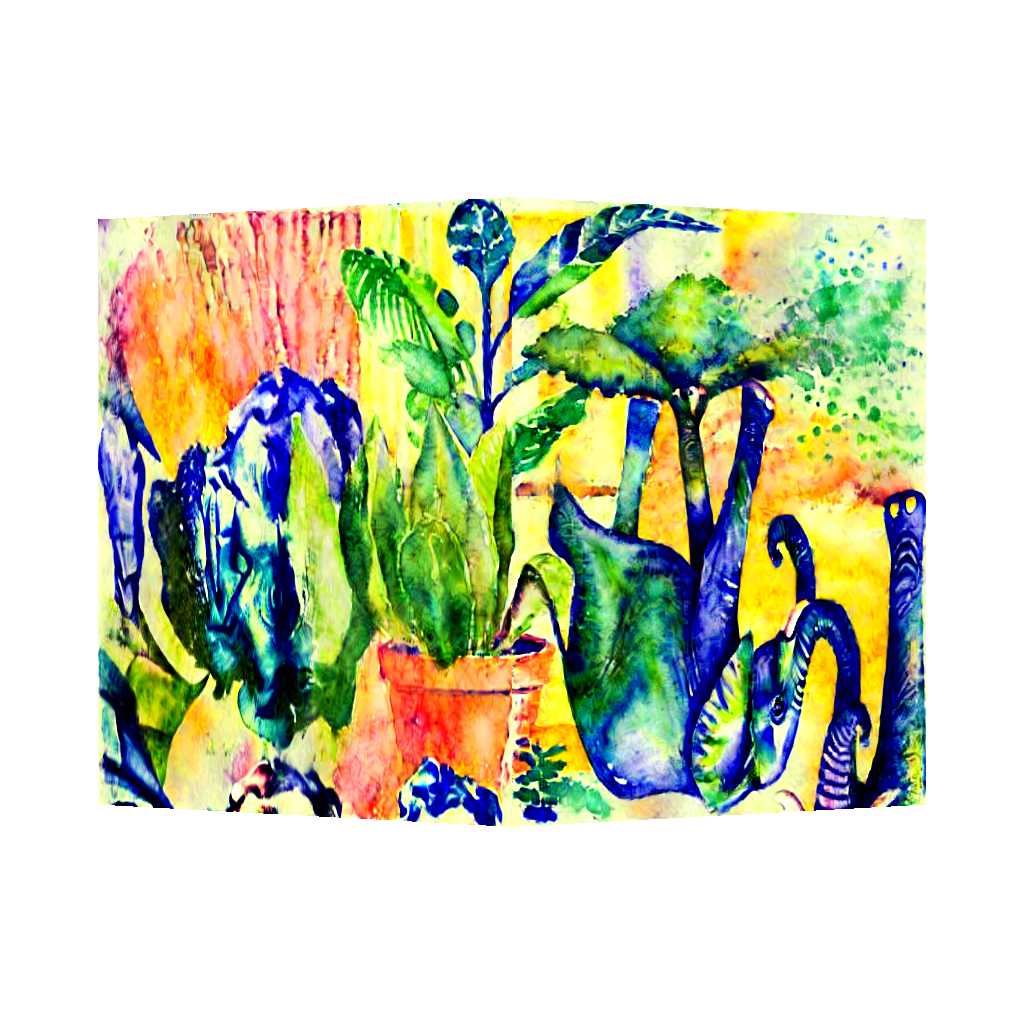} &
        \includegraphics[width=0.125\textwidth, trim=140 180 140 180, clip]{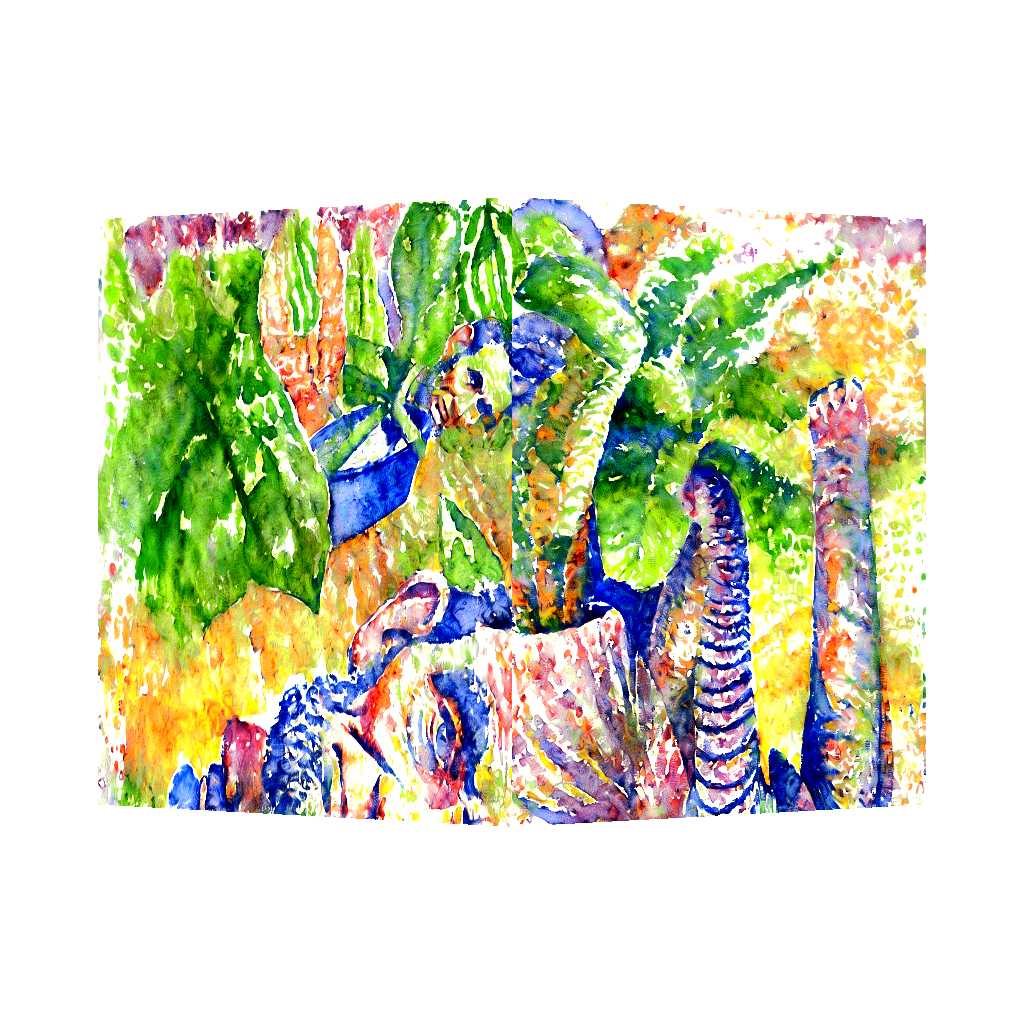} &
        \includegraphics[width=0.125\textwidth, trim=140 180 140 180, clip]{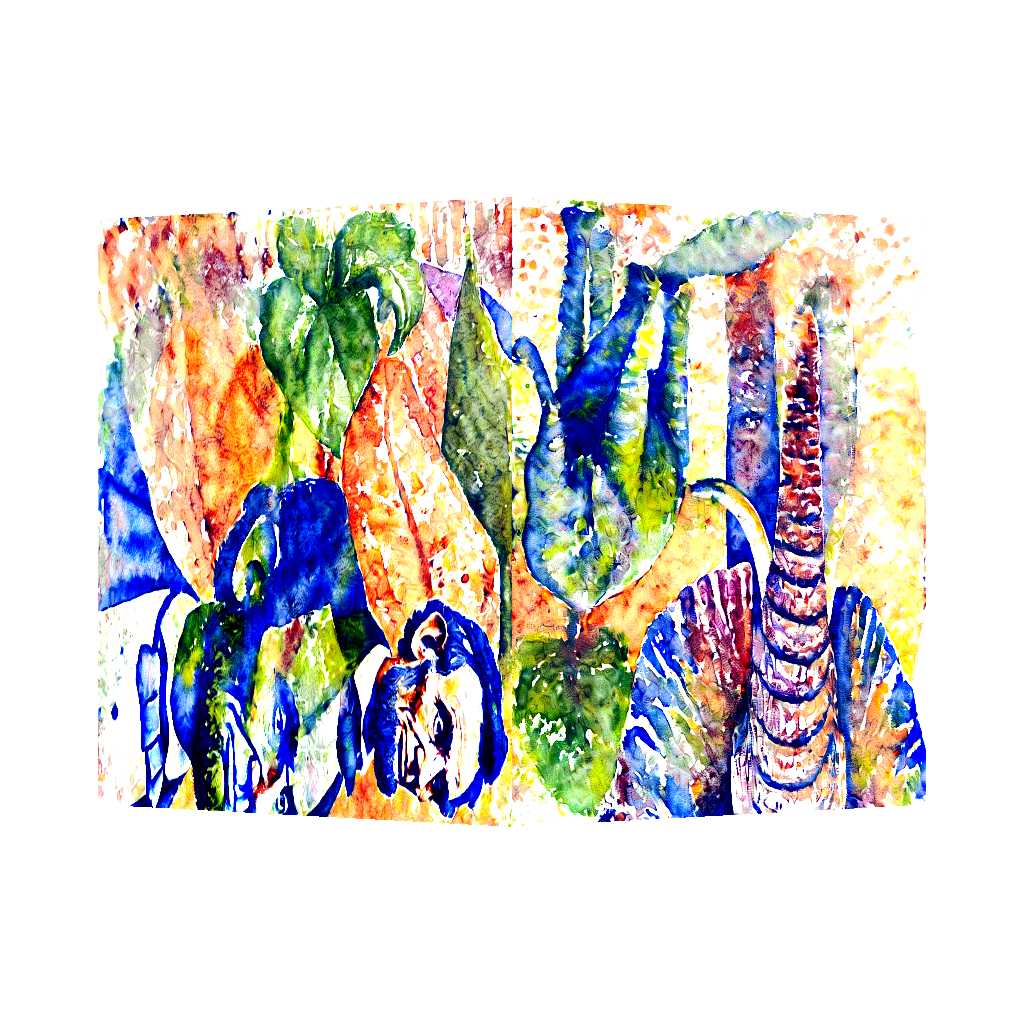} &
        \includegraphics[width=0.125\textwidth, trim=140 180 140 180, clip]{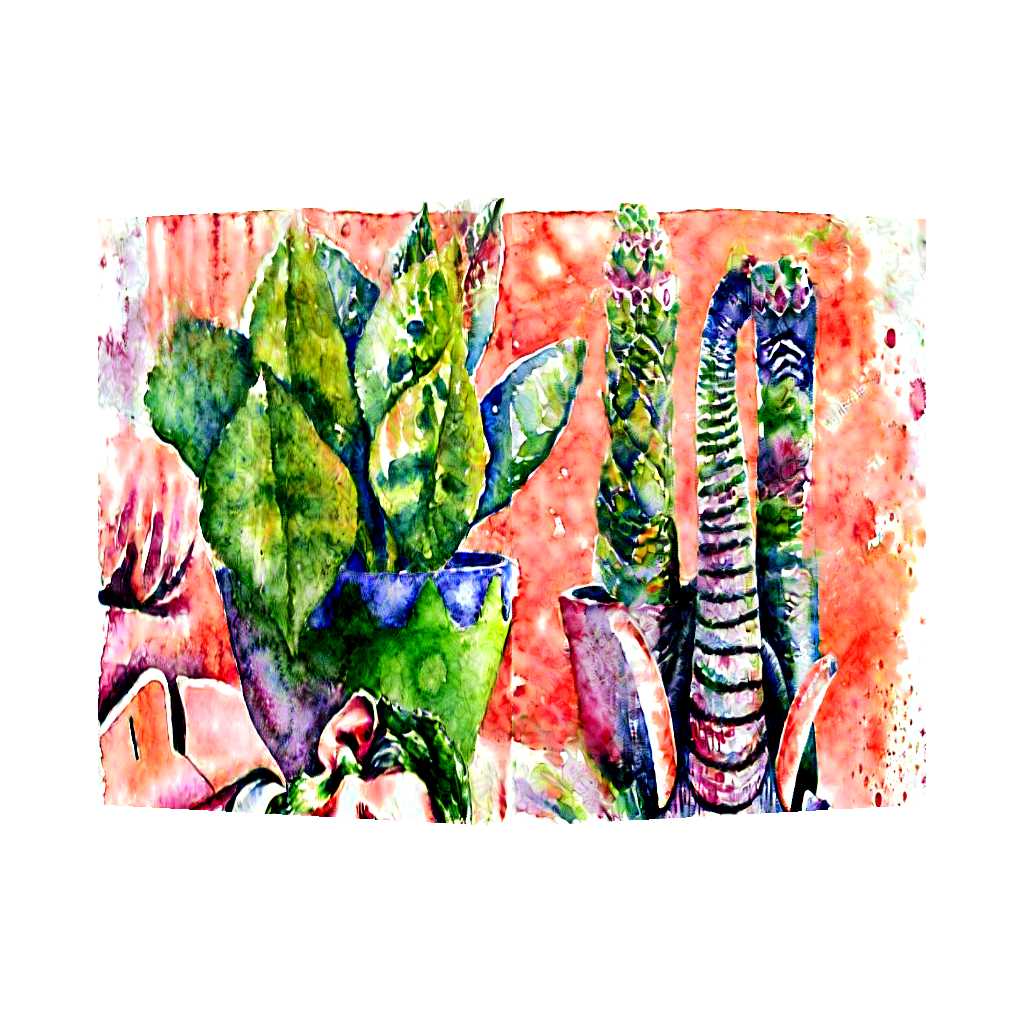} &
        \includegraphics[width=0.125\textwidth, trim=140 180 140 180, clip]{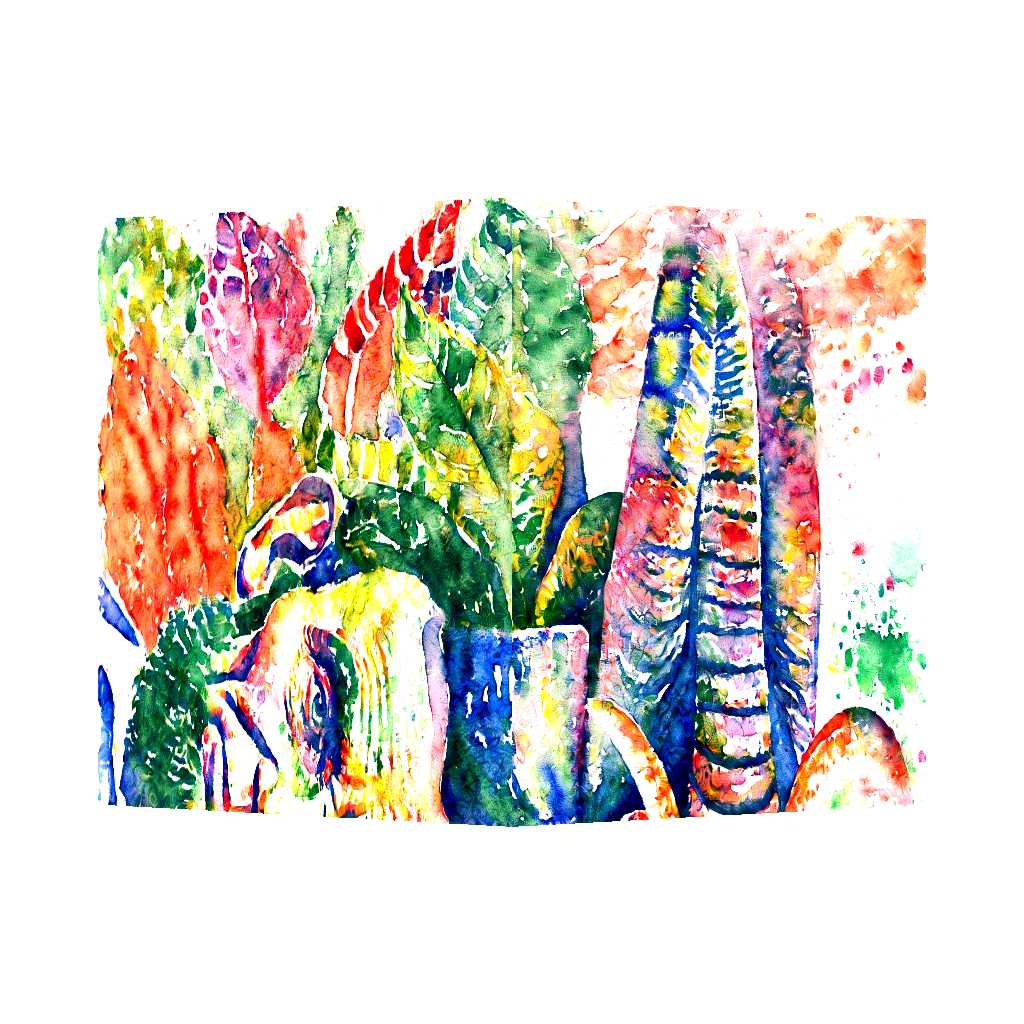} \\

        \toprule    
        \textbf{Camera jitter}        & \textcolor{red}{$\times$}& \textcolor{red}{$\times$}& \textcolor{red}{$\times$} & random & scheduled & \textcolor{red}{$\times$} & scheduled \\
        \textbf{Patch denoising}       &\textcolor{red}{$\times$} &\textcolor{red}{$\times$} & \textcolor{green}{$\checkmark$} & \textcolor{green}{$\checkmark$} & \textcolor{green}{$\checkmark$} & \textcolor{green}{$\checkmark$} & \textcolor{green}{$\checkmark$} \\
        \textbf{Resolution scaling}     & \textcolor{red}{$\times$}&\textcolor{red}{$\times$}& \textcolor{red}{$\times$} & \textcolor{red}{$\times$} & \textcolor{red}{$\times$} & \textcolor{green}{$\checkmark$} & \textcolor{green}{$\checkmark$}\\
        \bottomrule
    \end{tabular}

    \vspace{-2mm}
    \caption{\textbf{Comparison of different design choices.} 
    Our proposed method achieves optimal primary content fusion while maintaining high visual quality. In contrast, the baseline method struggles with the null space of the VAE encoder without camera jitter. Random patch denoising enhances resolution but introduces multiple instances of duplicate primary content. While random camera jitter facilitates smoother transitions, it still results in duplicate pattern artifacts if naively applied. Scheduled camera jitter mitigates these artifacts to some extent but does not fully eliminate them. Similarly, progressive resolution scaling improves smoothness but reduces primary content fusion and introduces minor duplicate pattern issues, particularly in columns 5 and 3. Our method integrates progressive resolution scaling with scheduled camera jitter to effectively center the primary content while minimizing duplicate pattern artifacts.}
    \label{fig:comparison}
\end{figure*}

Existing work like~\cite{Ding2023PatchedDD} and \cite{Wang2023PatchDF} employs patch-based denoising techniques with diffusion models to achieve higher-resolution generation. 
While effective for some tasks, using random patches introduces several failure cases due to the under-constrained optimization problem.
A common failure is that the rendered view only partially aligns with individual prompts. For example, in the cube optimization problem, prompt 1 should align with faces 1 and 3, prompt 2 with faces 1 and 2, and prompt 3 with faces 2 and 3. However, a local optimum can emerge that each prompt $i$ only shows up on face $i$, inhibiting aligning the entire view with the prompts. Similarly, another local optimum could occur in which each face contains multiple, separate illusion contents, leading to a duplicate pattern issue. For example, face 1 displays multiple objects indicated by prompt 1.
This issue is frequently observed in the baseline method (Fig.~\ref{fig:comparison}, columns 3 and 5) and significantly reduces the quality of the generated illusions.

To address these issues, we propose a progressive renderer resolution scaling strategy.
Specifically, for each view, we begin by denoising a single \(512\times512\) image, first establishing a clear and coherent baseline generation across faces and inhibiting repeated content from being generated. To achieve high resolutions (\eg, 1024$\times$1024), we then progressively enlarge the renderer resolution and sample a random \(512\times512\) image patch to refine the main content. 
Overall, this strategy effectively mitigates the duplicate pattern issue and enhances the centralized generation given the prompts while achieving high-resolution multiview illusions. 
\begin{figure}[t]
    \centering
    \footnotesize
    
    \setlength{\tabcolsep}{0.5pt} 
    \renewcommand{\arraystretch}{1}
        \begin{tabular}{p{0.04\linewidth} p{0.96\linewidth}}
        \begin{minipage}{\linewidth}
            \centering
            \rotatebox{90}{\textbf{\tiny Inverse Projection}}
        \end{minipage} & 
        \begin{tabular}{cccccc}
            \includegraphics[width=0.16\linewidth, trim=20 20 20 20 , clip]{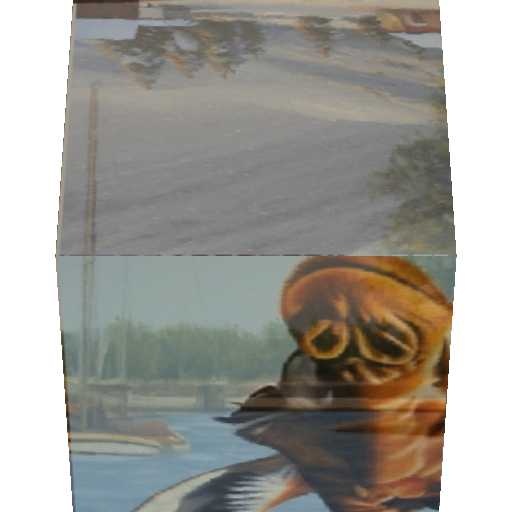} &
            \includegraphics[width=0.16\linewidth, trim=40 40 40 40 , clip]{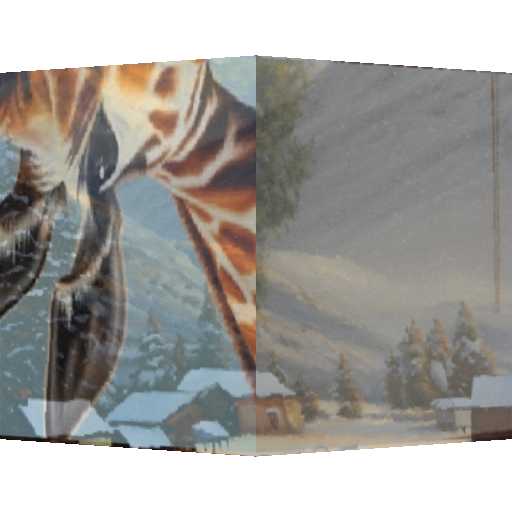} &
            \includegraphics[width=0.16\linewidth, trim=40 40 40 40 , clip]{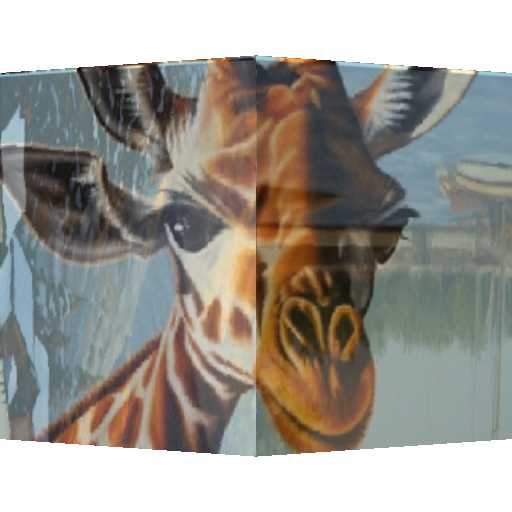} &
            \includegraphics[width=0.16\linewidth, trim=60 60 60 60, clip]{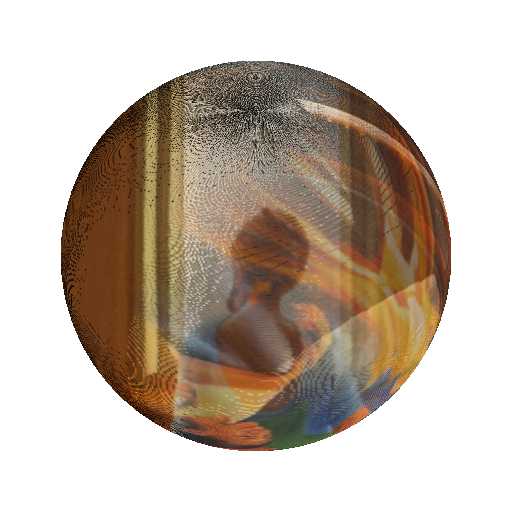} &
            \includegraphics[width=0.16\linewidth, trim=60 60 60 60, clip]{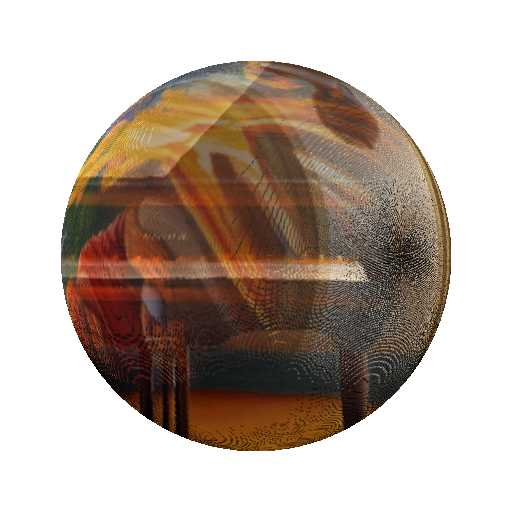} &
            \includegraphics[width=0.16\linewidth, trim=60 60 60 60, clip]{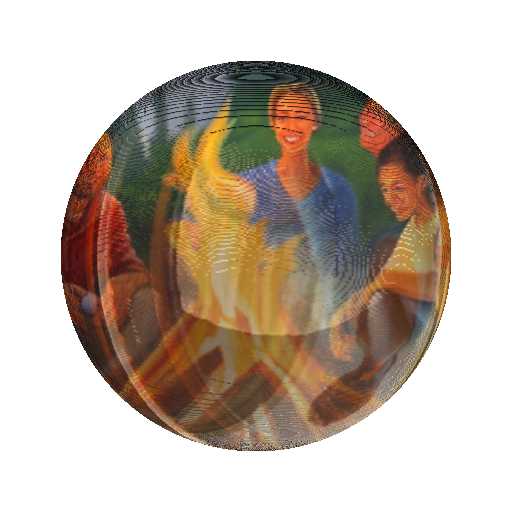}
            \\
        \end{tabular} \\

        \begin{minipage}{\linewidth}
            \centering
            \rotatebox{90}{\textbf{\tiny Latent Blending}}
        \end{minipage} & 
        \begin{tabular}{cccccc}
            \includegraphics[width=0.16\linewidth, trim=20 20 20 20 , clip]{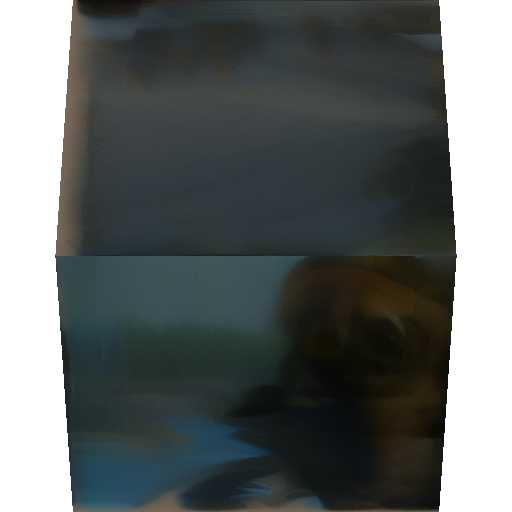} &
            \includegraphics[width=0.16\linewidth, trim=40 40 40 40 , clip]{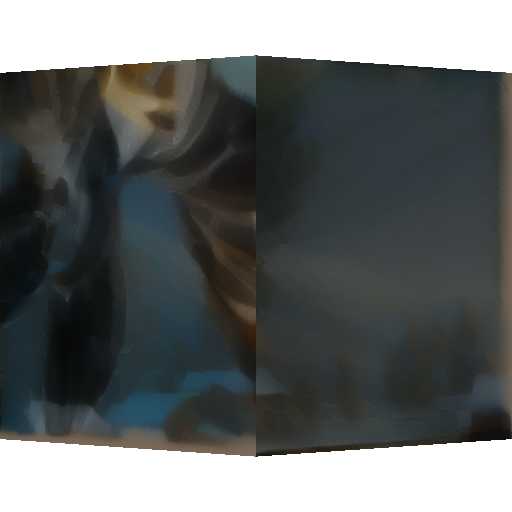} &
            \includegraphics[width=0.16\linewidth, trim=40 40 40 40 , clip]{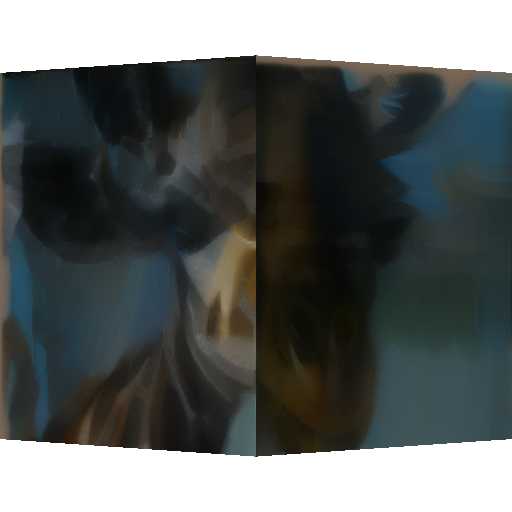} &
            \includegraphics[width=0.16\linewidth, trim=60 60 60 60, clip]{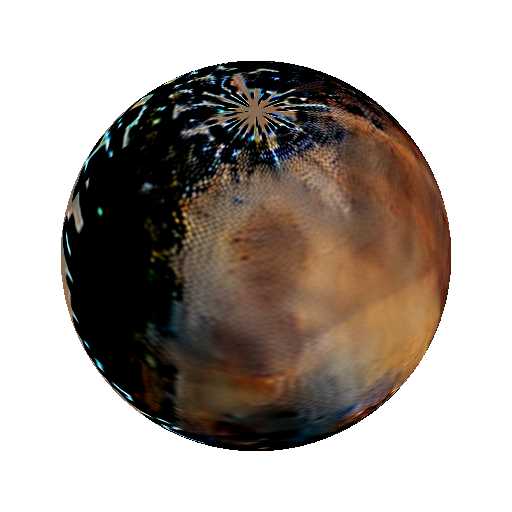} &
            \includegraphics[width=0.16\linewidth, trim=60 60 60 60, clip]{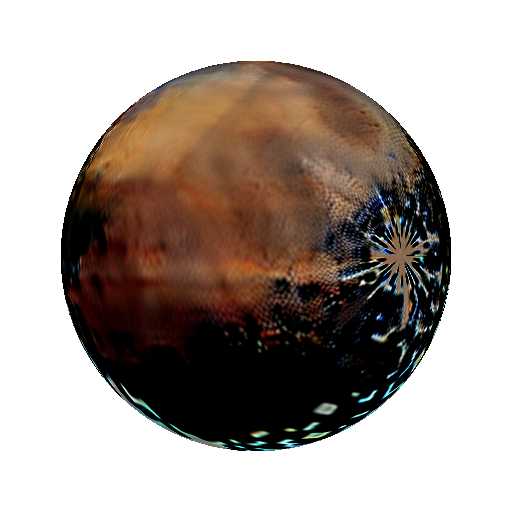} &
            \includegraphics[width=0.16\linewidth, trim=60 60 60 60, clip]{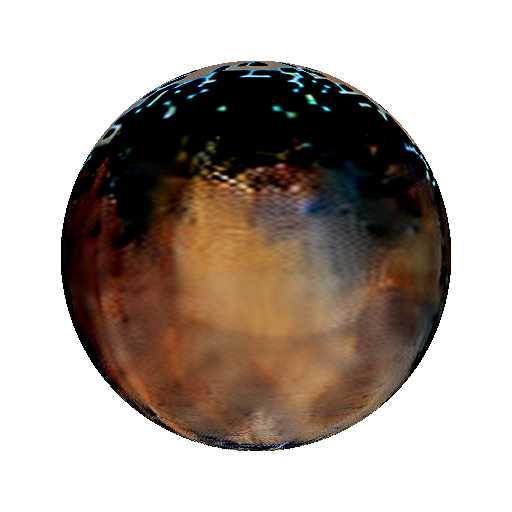}
            \\
        \end{tabular} \\

        \begin{minipage}{\linewidth}
            \centering
            \rotatebox{90}{\textbf{\tiny Burgert \textit{et al.} ~\cite{Burgert2023DiffusionIH}}}
        \end{minipage} & 
        \begin{tabular}{cccccc}
            \includegraphics[width=0.16\linewidth, trim=70 70 70 70, clip]{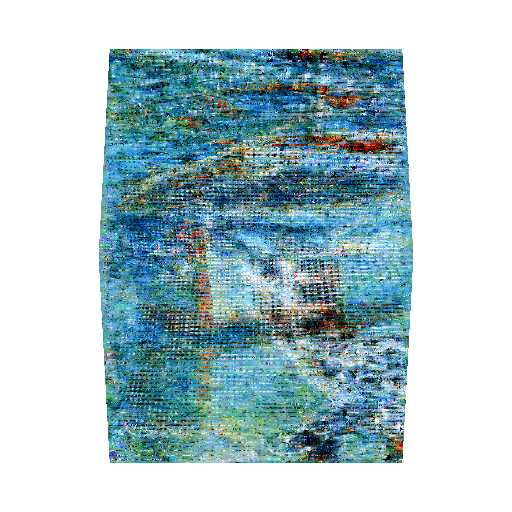} &
            \includegraphics[width=0.16\linewidth, trim=70 70 70 70, clip]{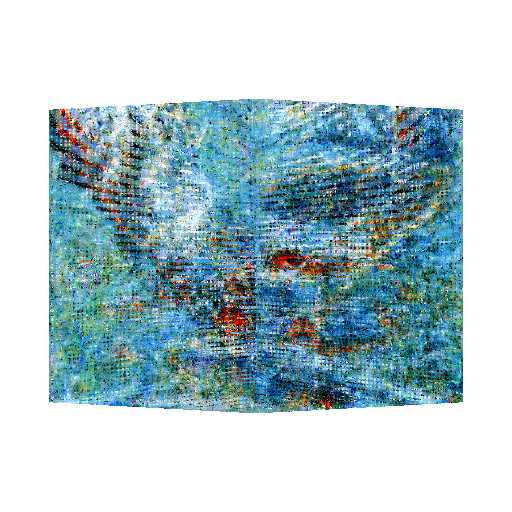} &
            \includegraphics[width=0.16\linewidth, trim=70 70 70 70, clip]{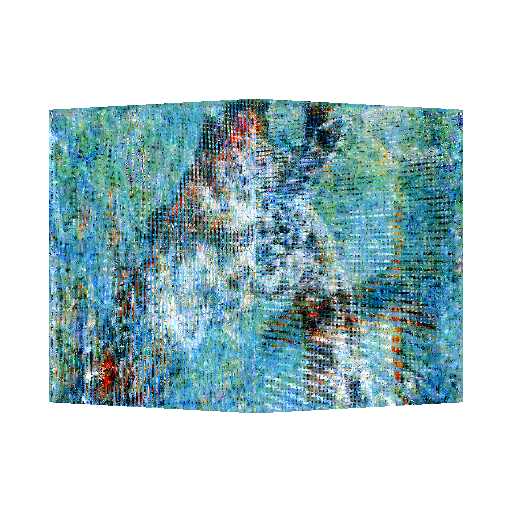} &
            \includegraphics[width=0.16\linewidth, trim=70 70 70 70, clip]{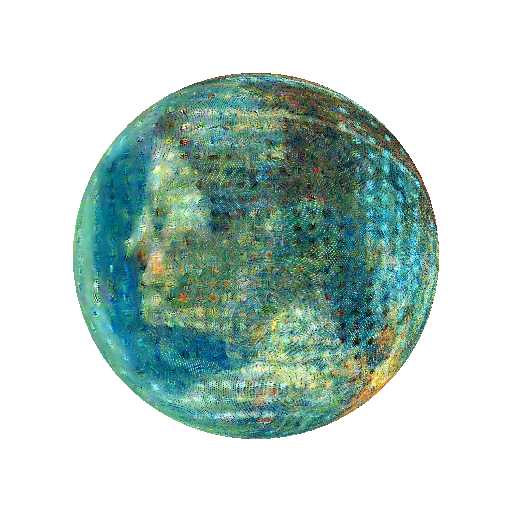} &
            \includegraphics[width=0.16\linewidth, trim=70 70 70 70, clip]{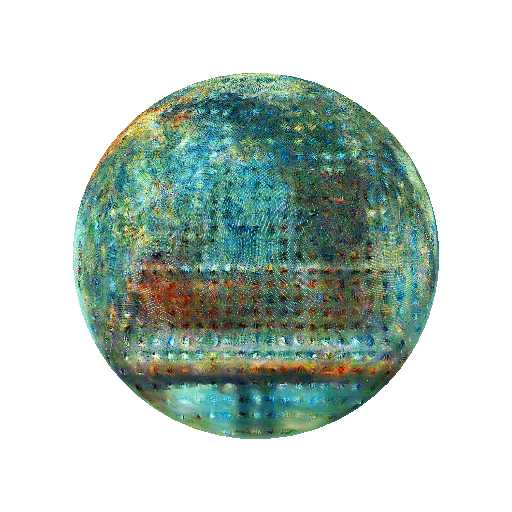} &
            \includegraphics[width=0.16\linewidth, trim=70 70 70 70, clip]{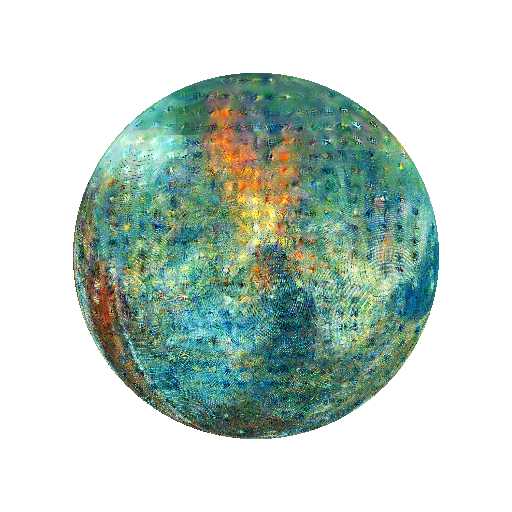}
            \\
        \end{tabular} \\

        \begin{minipage}{\linewidth}
            \centering
            \rotatebox{90}{\textbf{\tiny Ours}}
        \end{minipage} & 
        \begin{tabular}{cccccc}
            \includegraphics[width=0.16\linewidth, trim=140 140 140 140, clip]{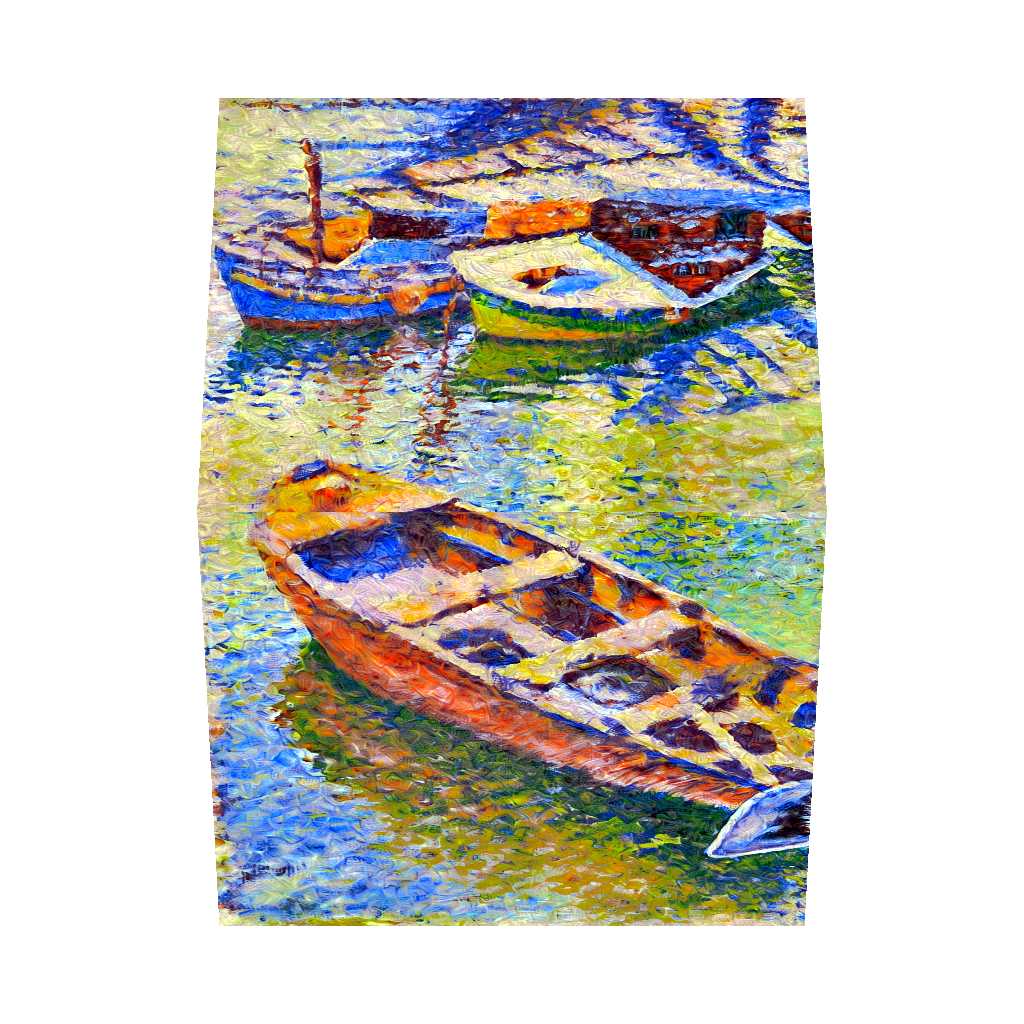} &
            \includegraphics[width=0.16\linewidth, trim=140 140 140 140, clip]{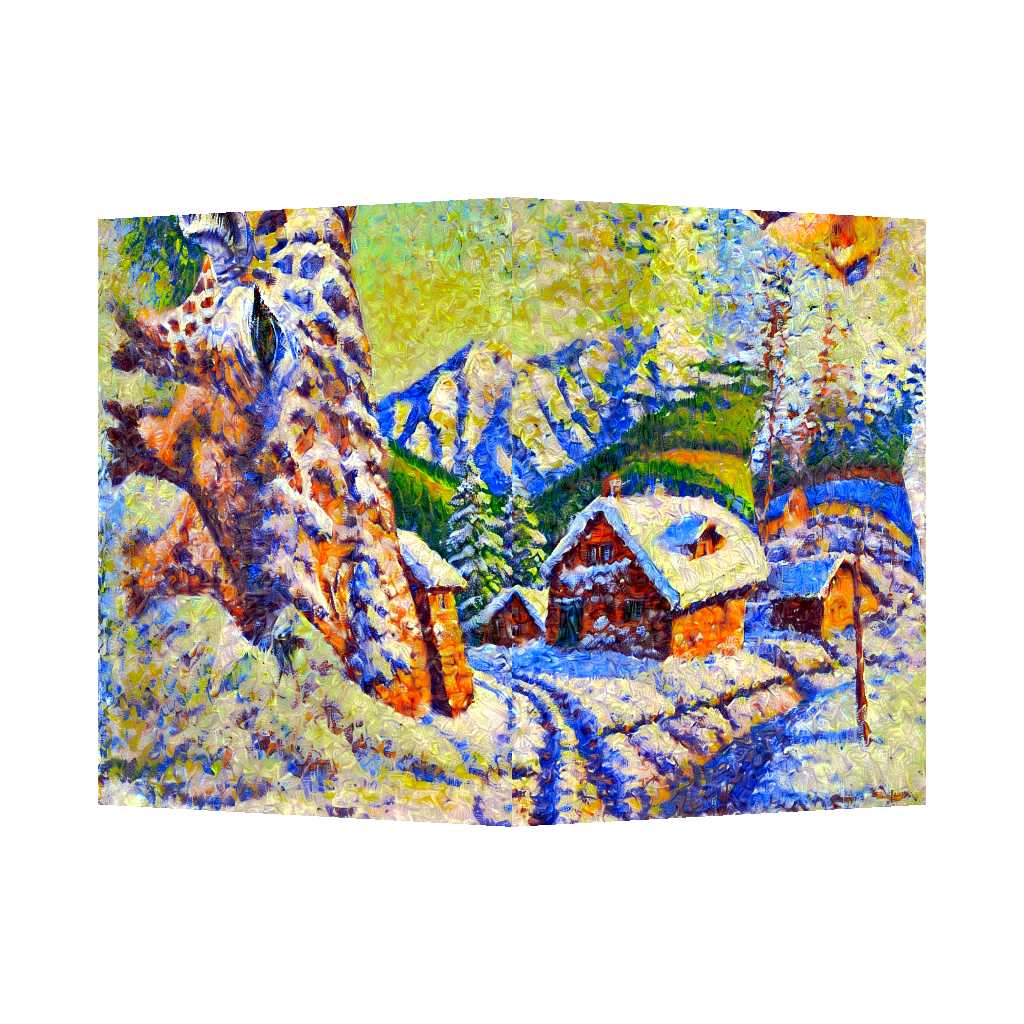} &
            \includegraphics[width=0.16\linewidth, trim=140 140 140 140, clip]{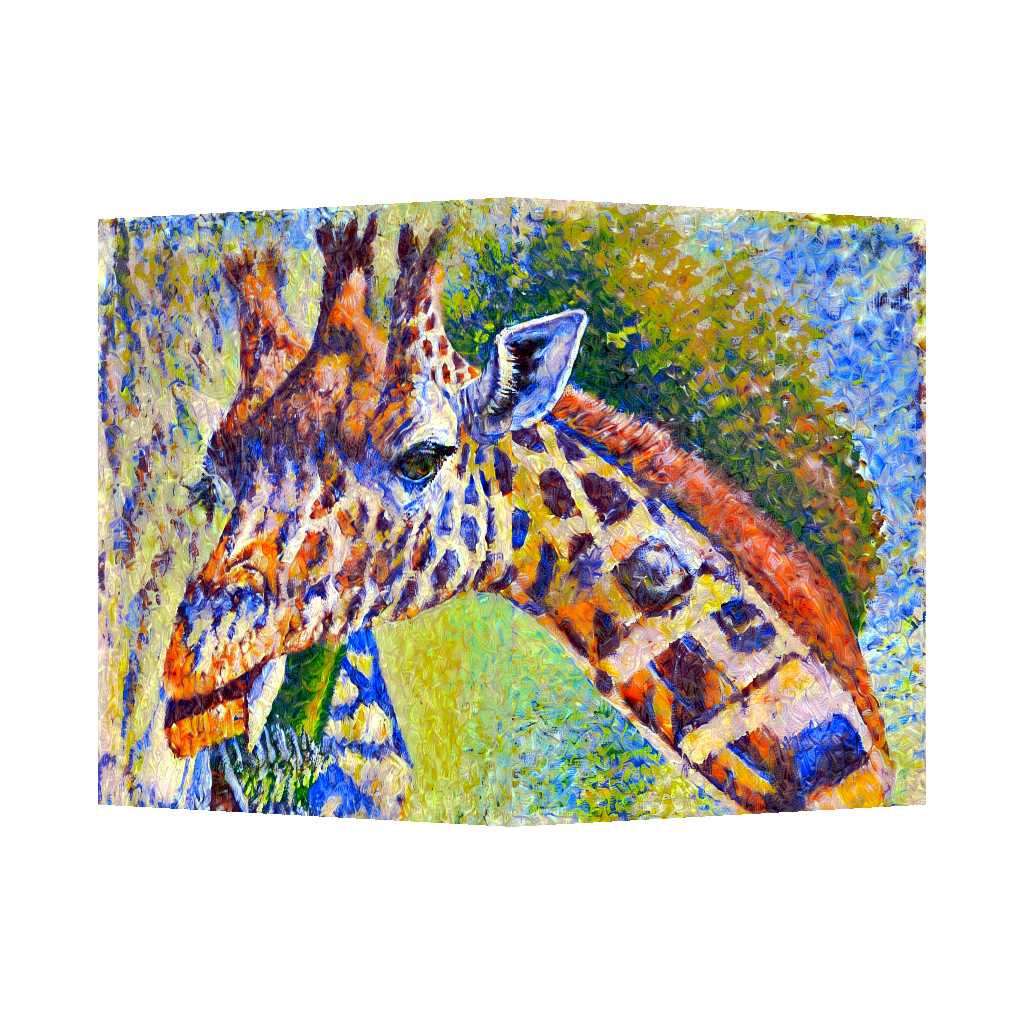}  &
            \includegraphics[width=0.16\linewidth, trim=140 140 140 140, clip]{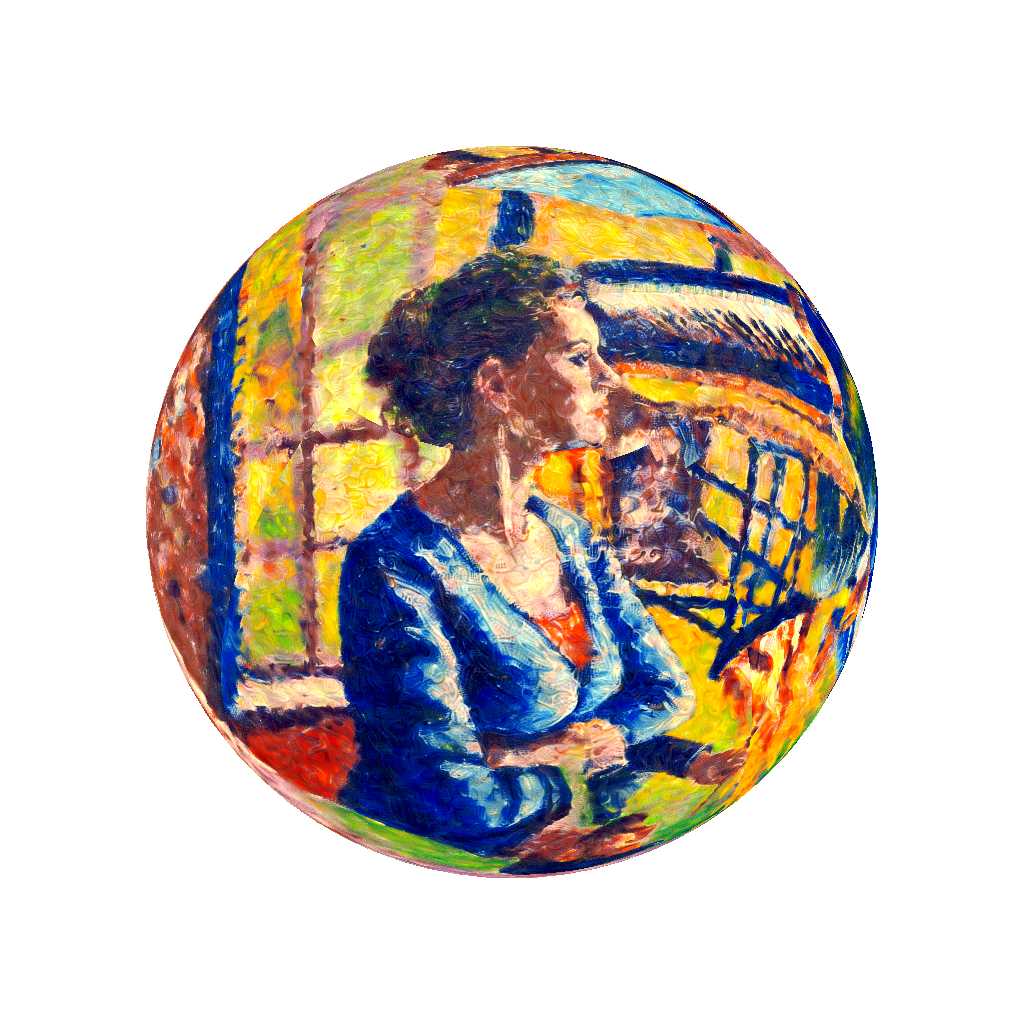} &
            \includegraphics[width=0.16\linewidth, trim=140 140 140 140, clip]{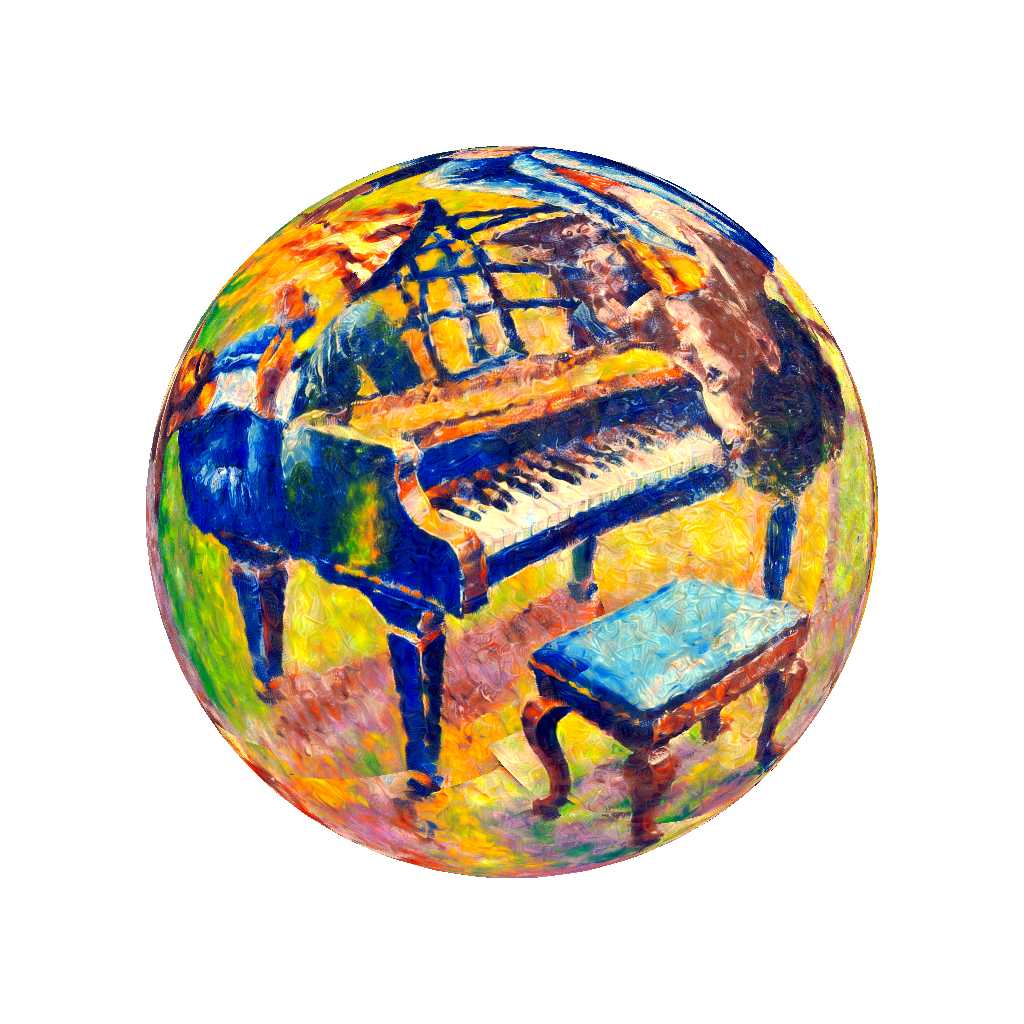} &
            \includegraphics[width=0.16\linewidth, trim=140 140 140 140, clip]{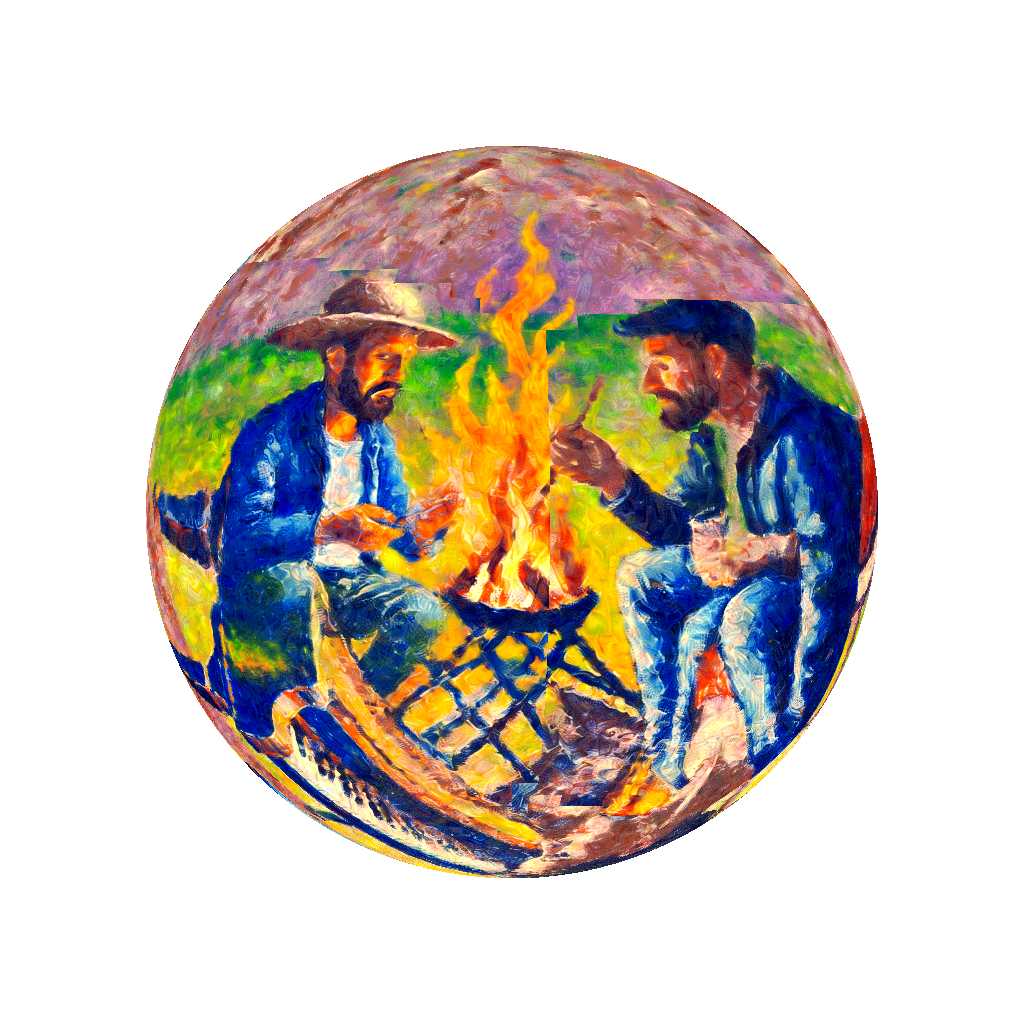}
            \\
        \end{tabular} \\

          & 
        \begin{tabular}{cccccc}
            \makebox[0.16\linewidth]{\includegraphics[scale=0.95, width=0.16\linewidth, trim=420 20 420 30, clip]{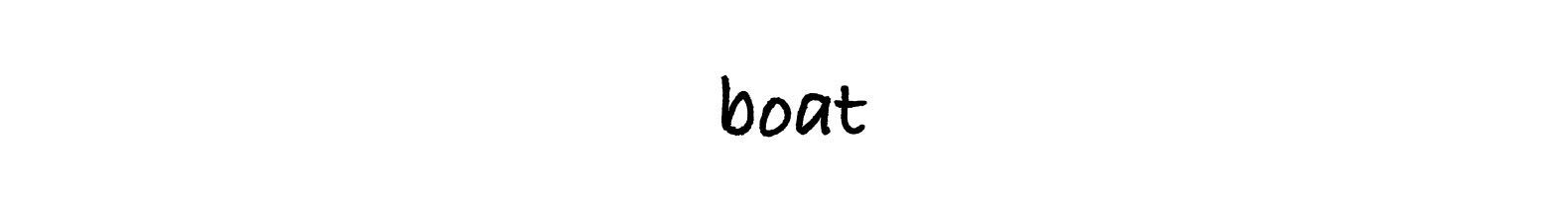}} &
            \makebox[0.16\linewidth]{\raisebox{-0.3mm}{\includegraphics[width=0.19\linewidth, trim=750 20 80 30, clip]{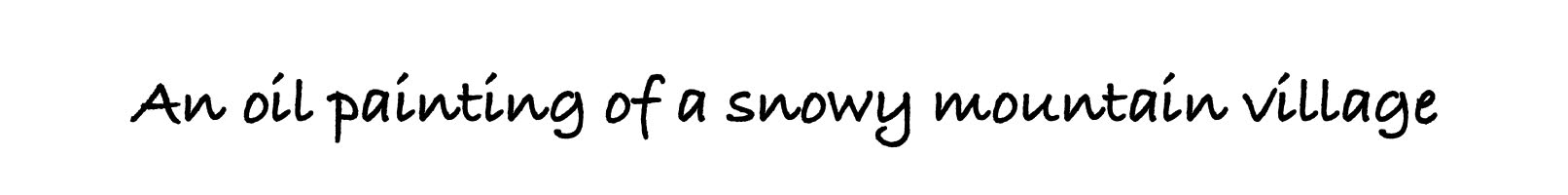}}}&
            \makebox[0.16\linewidth]{\includegraphics[scale=0.9, width=0.16\linewidth, trim=420 20 420 30, clip]{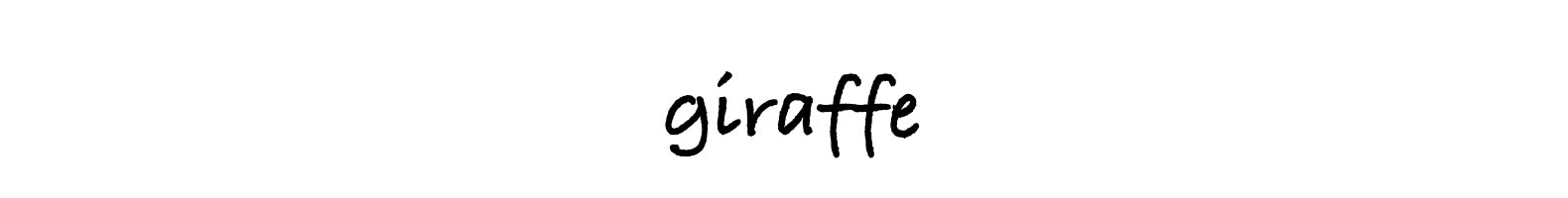}} &
            \makebox[0.16\linewidth]{\raisebox{0.2mm}{\includegraphics[width=0.2\linewidth, trim=250 20 150 30, clip]{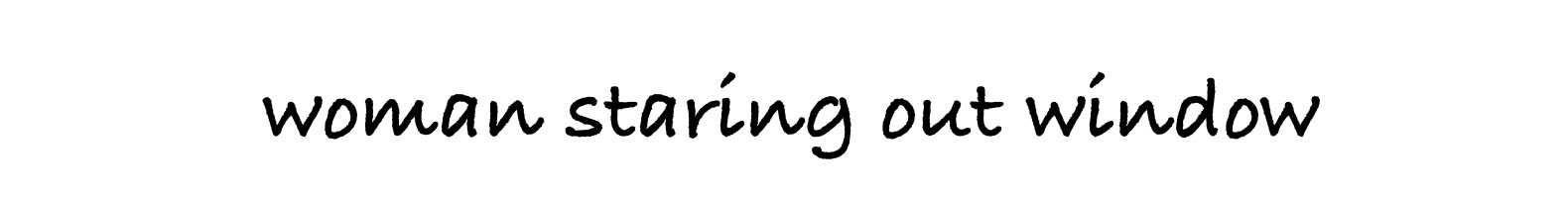}}} &
            \makebox[0.16\linewidth]{\raisebox{-0.55mm}{\includegraphics[scale=0.05, width=0.05\linewidth, trim=920 20 500 30, clip]{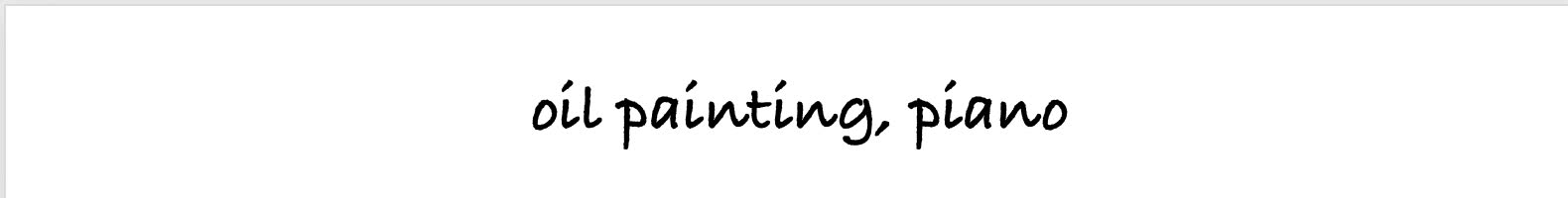}}}
             & 
            \makebox[0.16\linewidth]{\raisebox{-0.25mm}{\includegraphics[scale=0.99, width=0.16\linewidth, trim=790 20 120 30, clip]{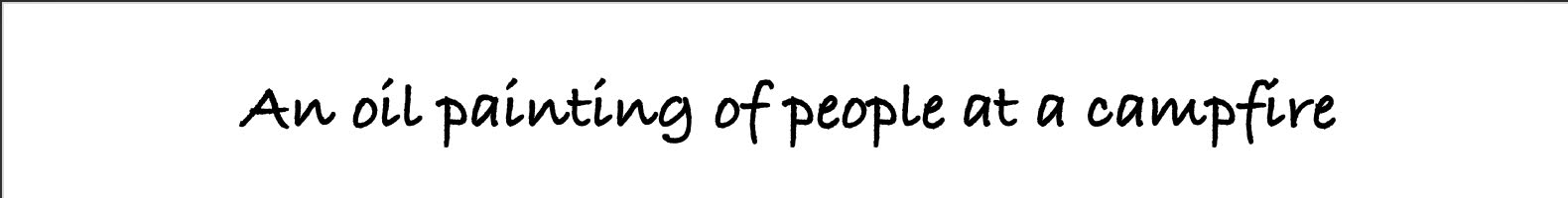}}} \\
        \end{tabular}

         \\
        
    \end{tabular}
    \vspace{-7mm}

    \caption{\textbf{Comparison with basic baselines.} Inverse projection can blend the images but cannot generate an illusion. Blending in latent space also fails. Burget \textit{et al.}'s ~\cite{Burgert2023DiffusionIH} method can not generate an illusion of 3D shape. Our method can blend the primary content while generating an illusion.}
    \label{fig:otherbaselines}
    \vspace{-3mm}
\end{figure}

We explore two strategies to progressively scale the rendered resolution \(R(k)\): linear and sigmoid schedules. We find that both schedules significantly reduce duplicate patterns in practice, while sigmoid mapping \(\sigma(k)\) generally demonstrated improved performance in our experiments with randomly sampled prompt pairs $y_i$. 
We define the sigmoid render resolution \(R(k)\) as follows:
\begin{equation}
\begin{aligned}
\sigma(k) &= \frac{1}{1 + \exp\left(-10 \cdot \left(\frac{k}{k_{\text{total}}} - 0.5\right)\right)} ,
\end{aligned}
\end{equation}
\begin{equation}
\begin{aligned}
R(k) &= a + \sigma\left(k\right) \cdot (b - a).
\end{aligned}
\label{eq:adjusted_equations}
\end{equation}
where \(a\) is the initial resolution of 512, \(b\) is the final resolution of 1024, and $k$ is the current training step.

\begin{figure}
    \vspace{-6mm}
    \centering
    
    \fontsize{5.5pt}{6.5pt}\selectfont

    \includegraphics[trim=0 0 0 0, clip, width=0.24\linewidth]{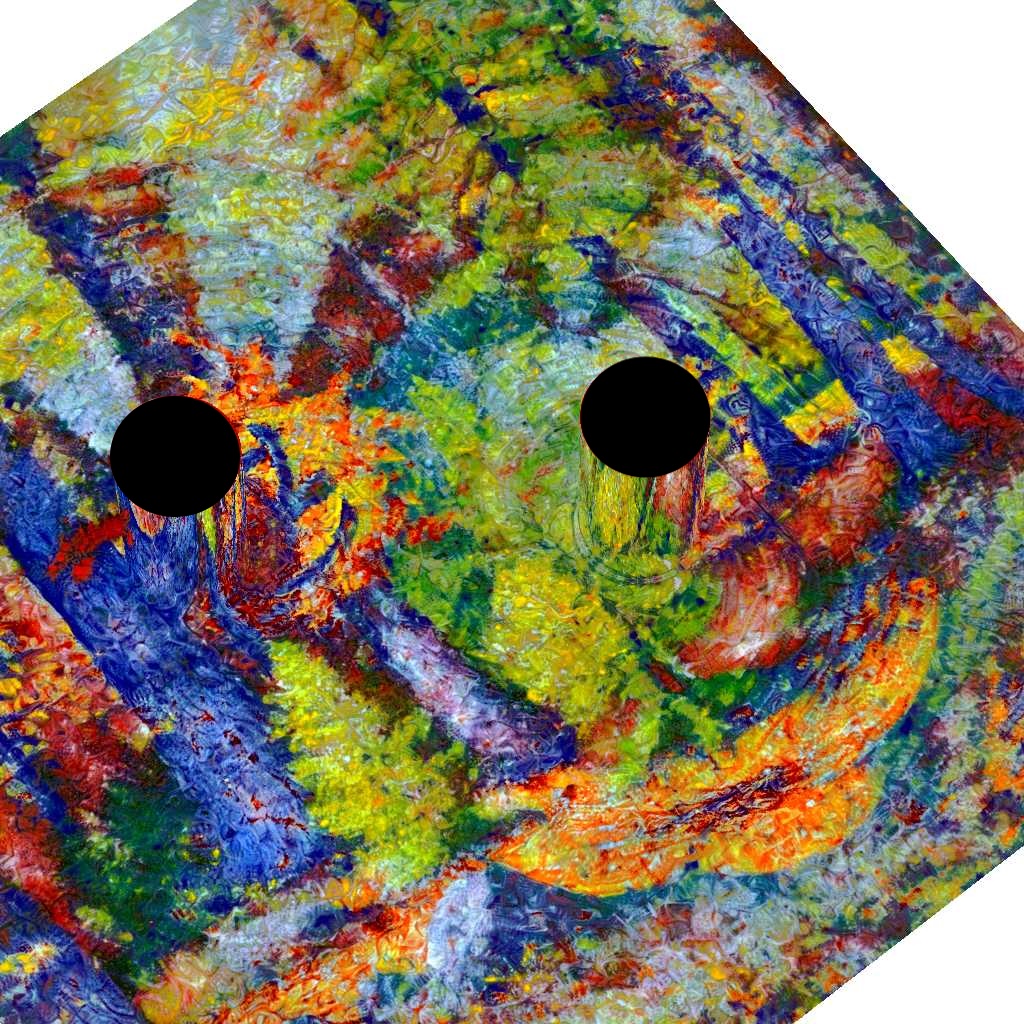}\hfill
    \includegraphics[trim=120 120 120 120, clip, width=0.24\linewidth]{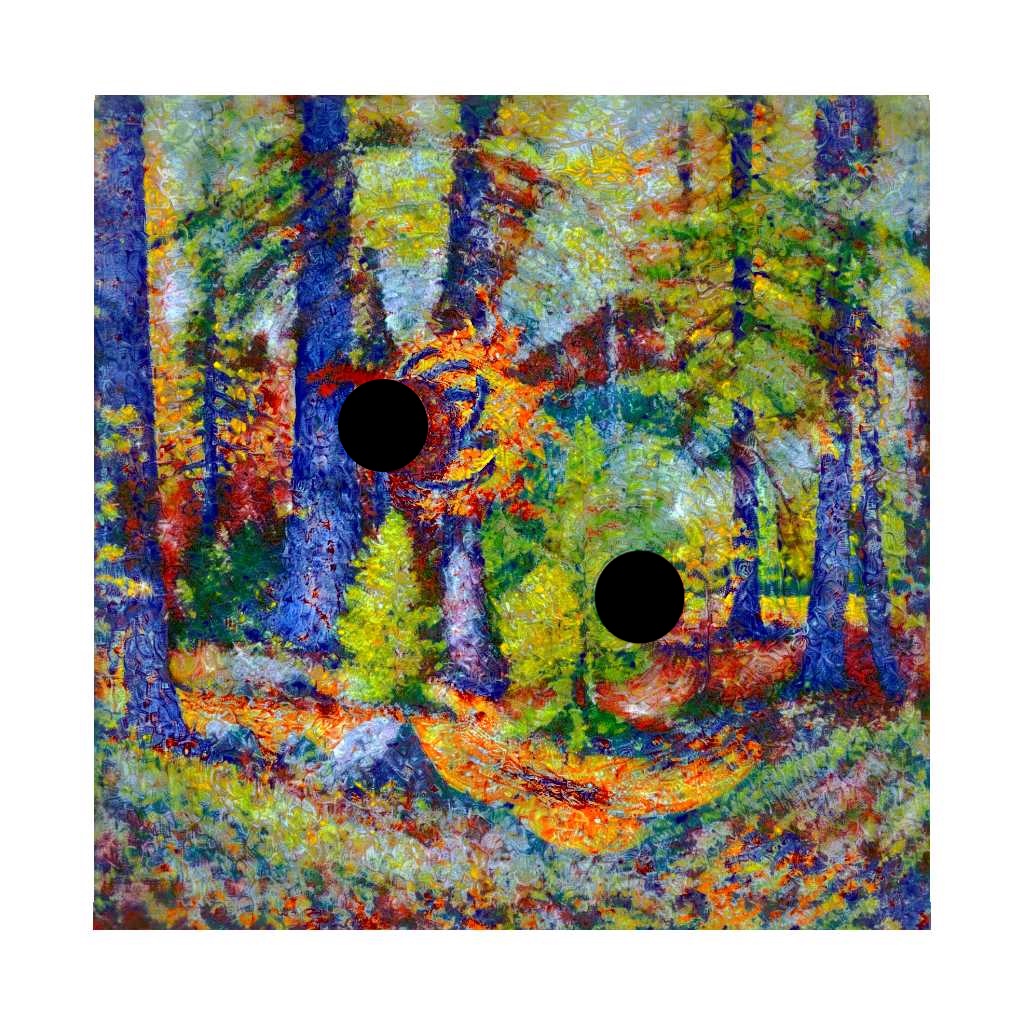}\hfill
    \includegraphics[trim=0 0 0 0, clip, width=0.24\linewidth]{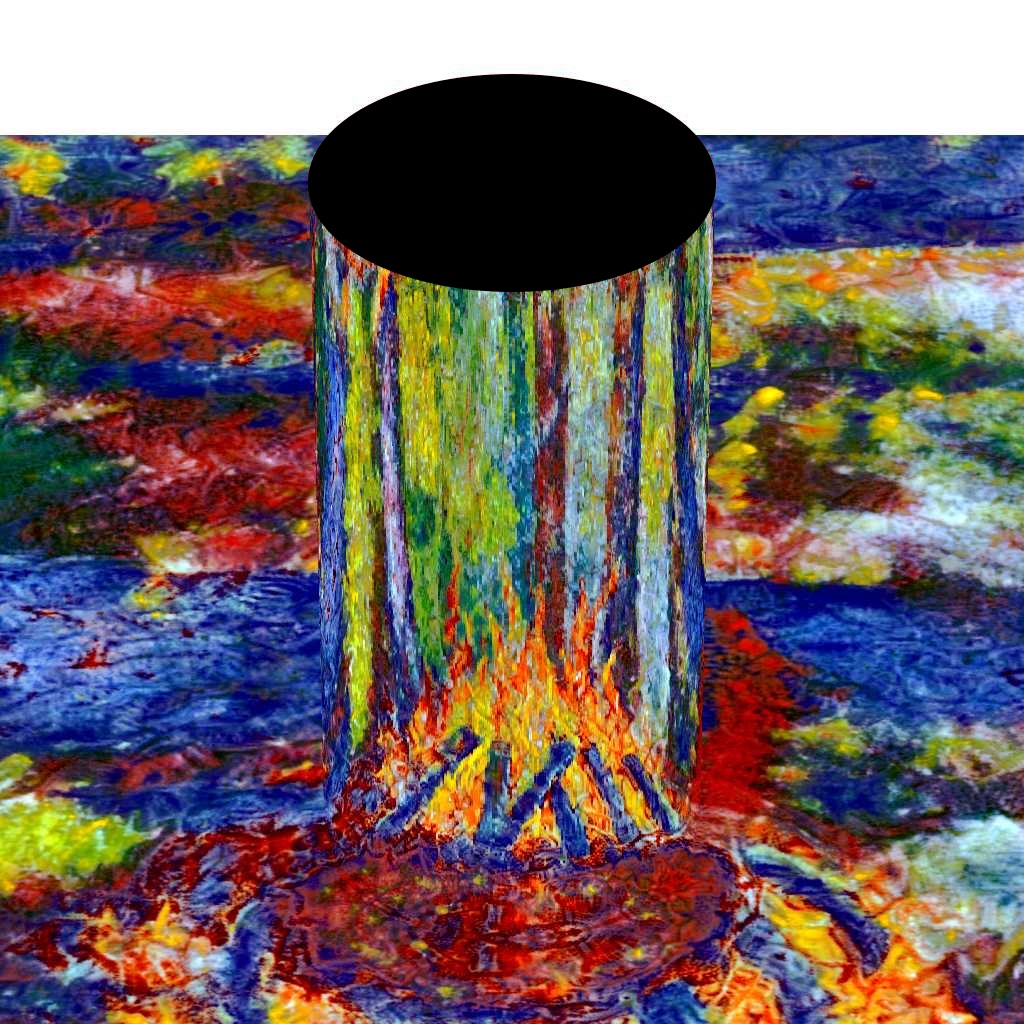}\hfill
    \includegraphics[trim=0 0 0 0, clip, width=0.24\linewidth]{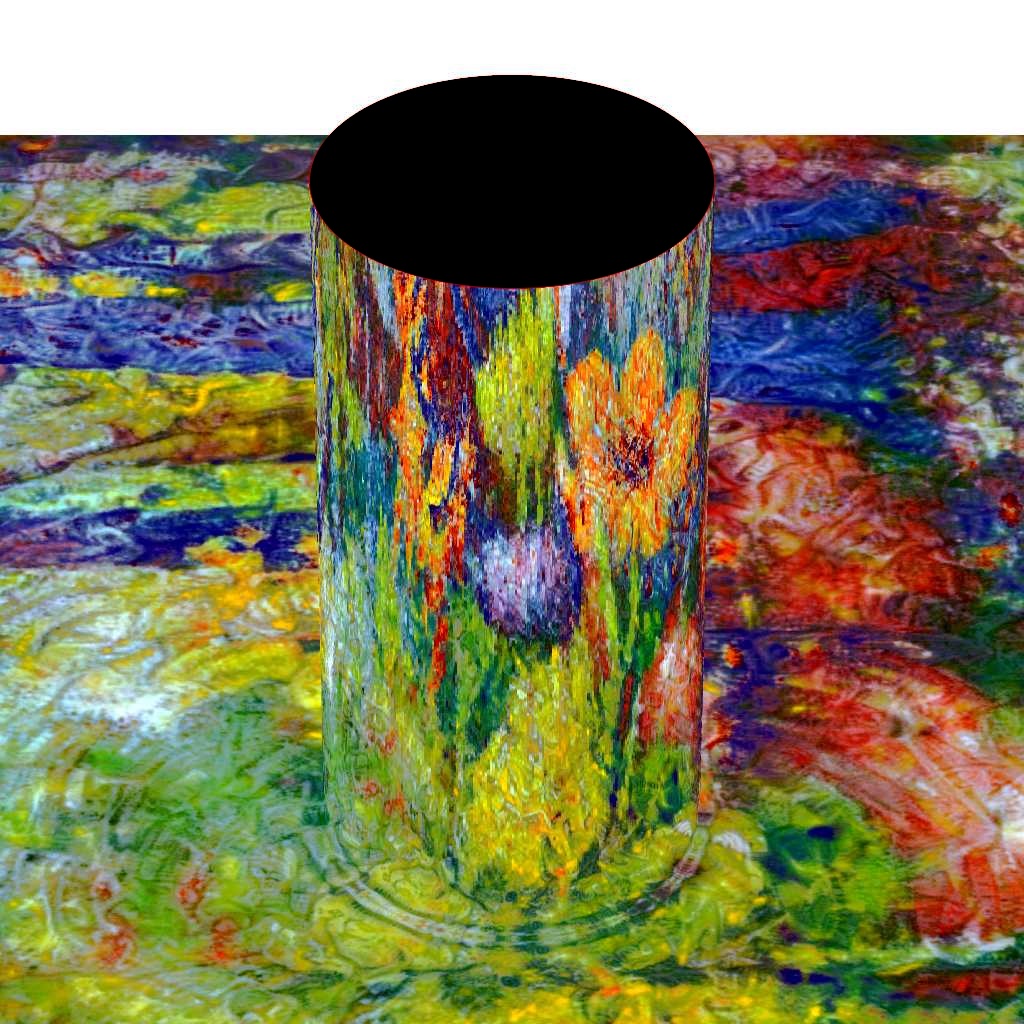}\\

    \begin{minipage}[c]{0.25\linewidth}
        \centering
        {Reflective Scene}
    \end{minipage}\hfill
    \begin{minipage}[c]{0.25\linewidth}
        \hspace{6mm}
        \includegraphics[width=\linewidth, trim=320 20 320 30, clip]{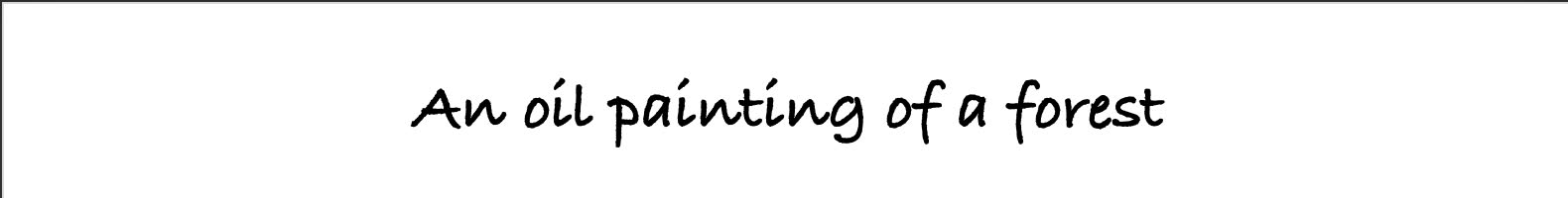}
    \end{minipage}\hfill
    \begin{minipage}[c]{0.25\linewidth}
        \centering
        \includegraphics[width=\linewidth, trim=320 20 320 30, clip]{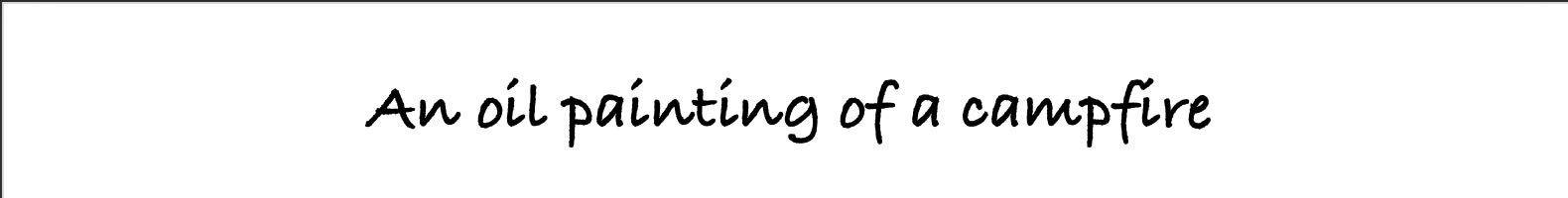}
    \end{minipage}\hfill
    \begin{minipage}[c]{0.25\linewidth}
        \hspace{6mm}
        \includegraphics[width=\linewidth, trim=320 20 320 30, clip]{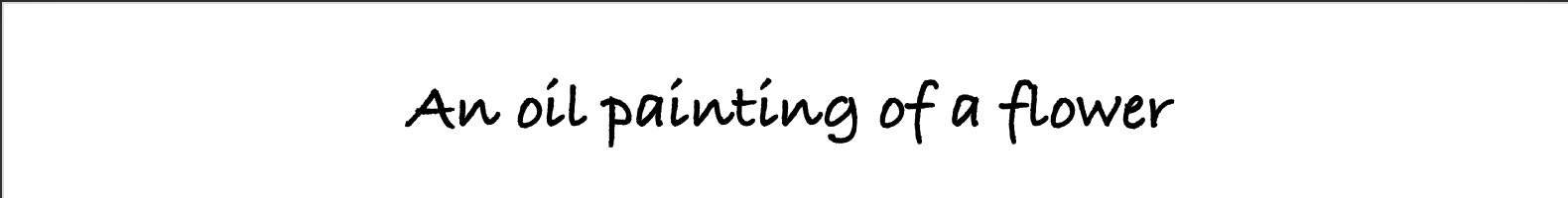}
    \end{minipage}\\

    \includegraphics[trim=0 0 0 0, clip, width=0.24\linewidth]{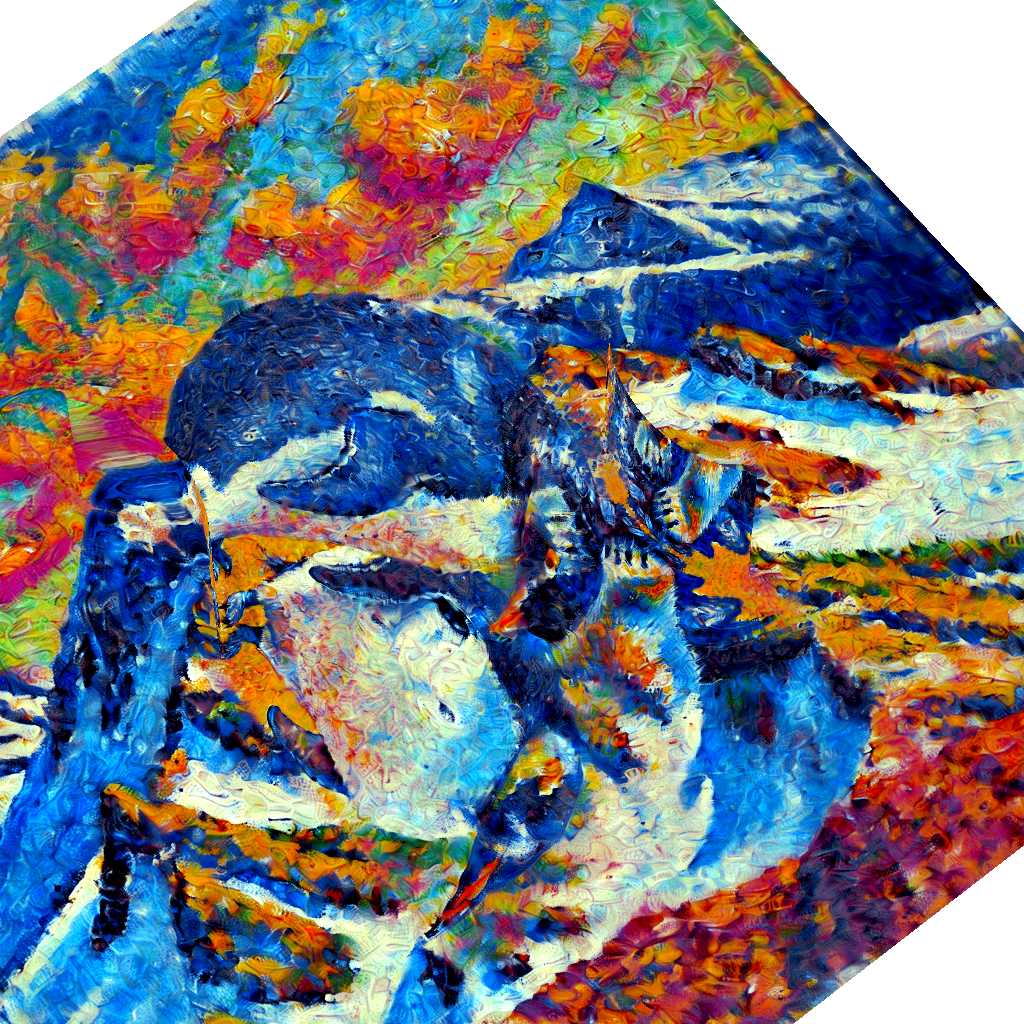}\hfill
    \includegraphics[trim=120 120 120 120, clip, width=0.24\linewidth]{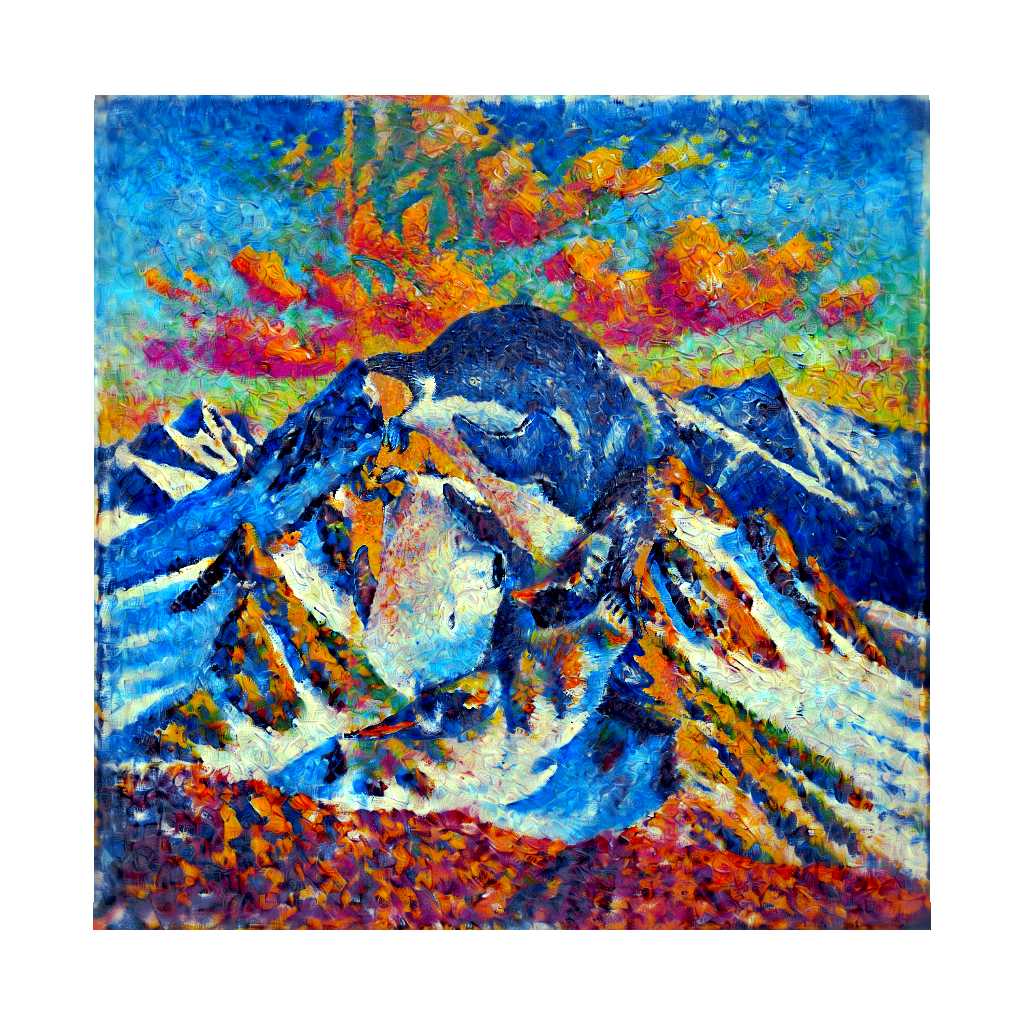}\hfill
    \includegraphics[trim=0 0 0 0, clip, width=0.24\linewidth]{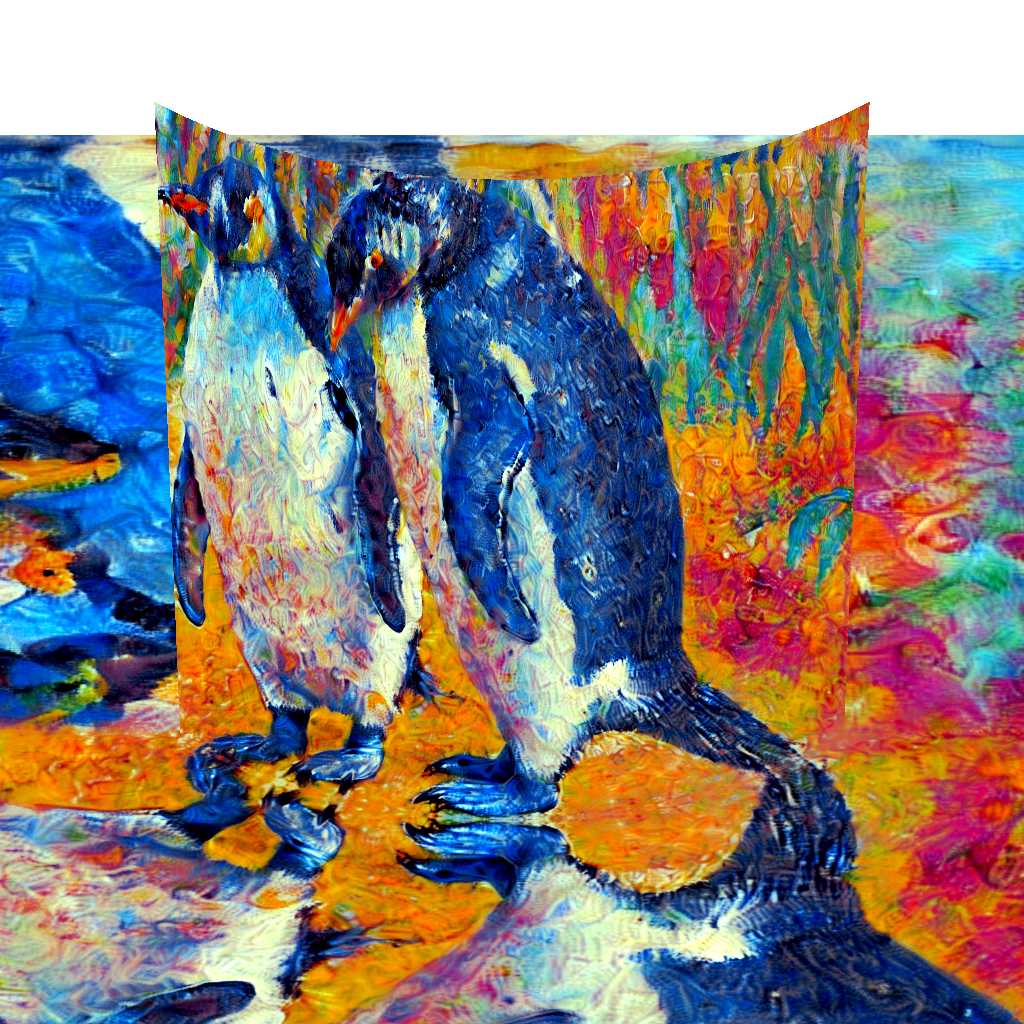}\hfill
    \includegraphics[trim=0 0 0 0, clip, width=0.24\linewidth]{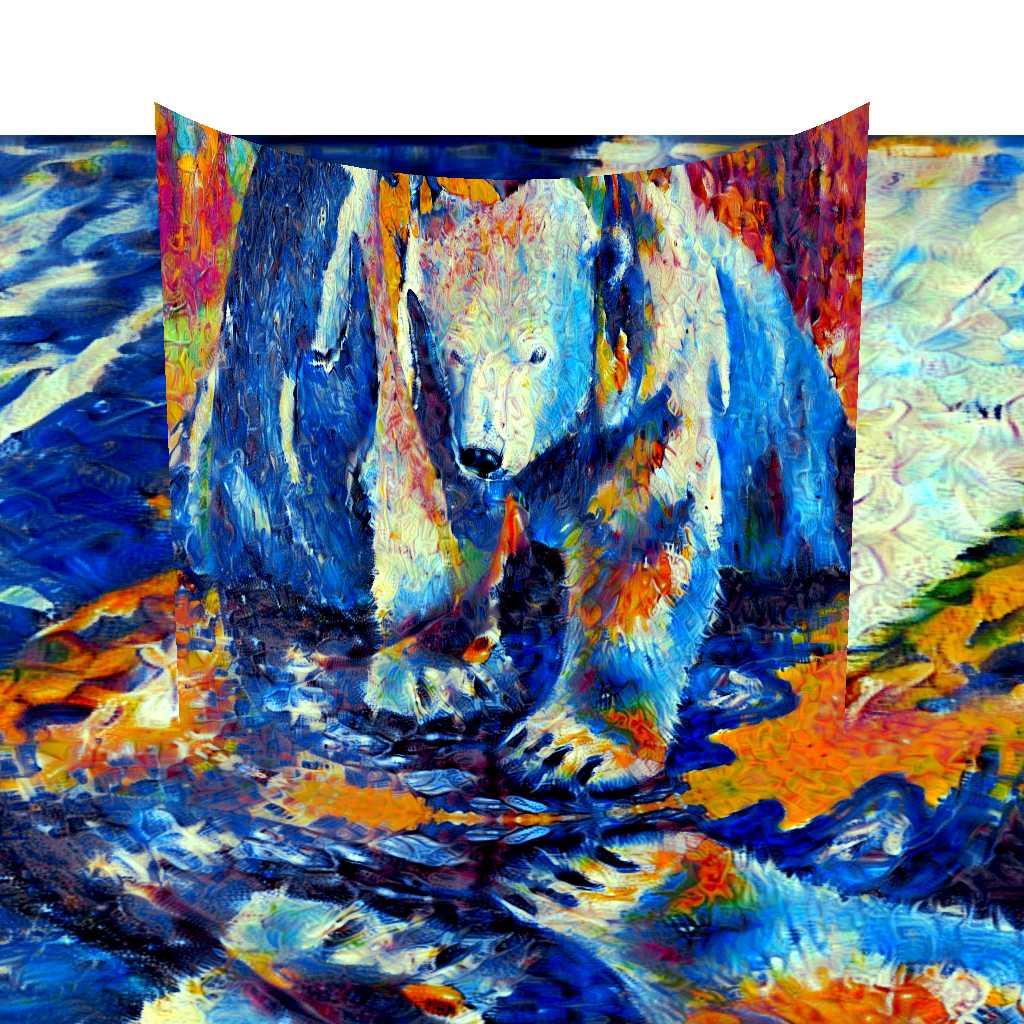}\\

    \begin{minipage}[c]{0.25\linewidth}
        \centering
        {Reflective Scene}
    \end{minipage}\hfill
    \begin{minipage}[c]{0.25\linewidth}
        \centering
        \includegraphics[width=\linewidth, trim=682 20 100 30, clip]{figures/prompts/village.jpg}
    \end{minipage}\hfill
    \begin{minipage}[c]{0.25\linewidth}
        \centering
        \includegraphics[width=\linewidth, trim=320 20 320 30, clip]{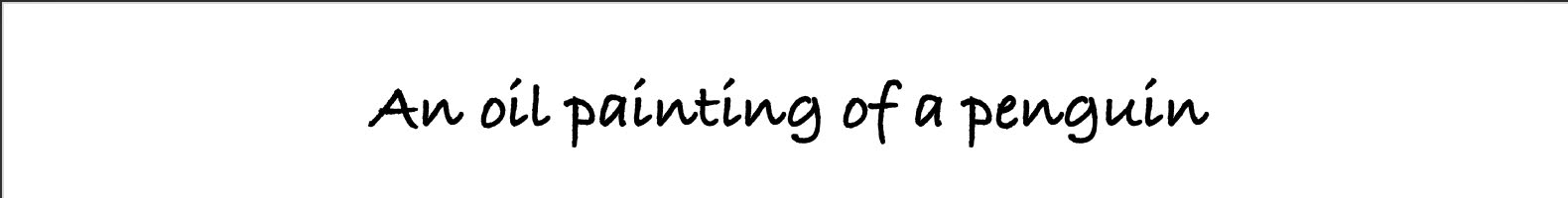}
    \end{minipage}\hfill
    \begin{minipage}[c]{0.25\linewidth}
        \centering
        \includegraphics[width=\linewidth, trim=320 20 320 30, clip]{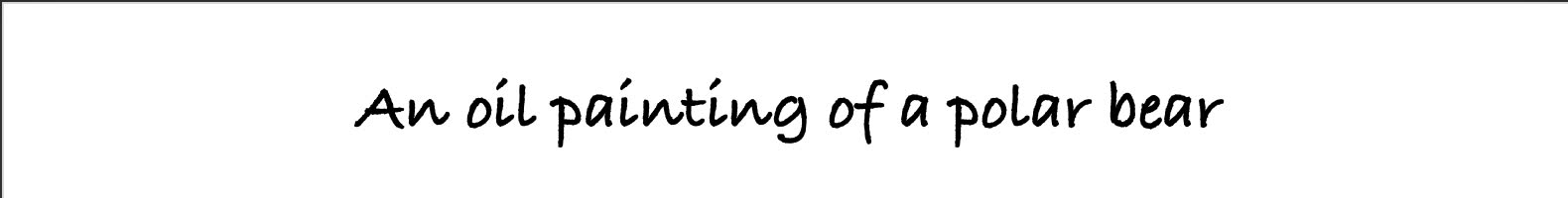}
    \end{minipage}\\
        \vspace{-3mm}
    \caption{
    \textbf{3D multiview illusion with two reflective cylinders/mirrors.} 
    Creating an illusion with a single reflective cylinder is already challenging for artists. 
    For traditional methods, generating artwork on two cylinders simultaneously is even more difficult, if not impossible.
    Here, using 2D diffusion priors, we can create intriguing examples of illusion generation on two reflective cylinders. 
    This showcases an extension of human capability in creating complex illusion artwork.
    }
    \vspace{-2mm}
    \label{fig:reflective-2cylinder}
\end{figure}
\begin{figure}
    \vspace{-2mm}
    \centering

    \begin{minipage}[c]{0.3\linewidth} 
        \includegraphics[trim=0 0 0 0, clip, width=\linewidth]{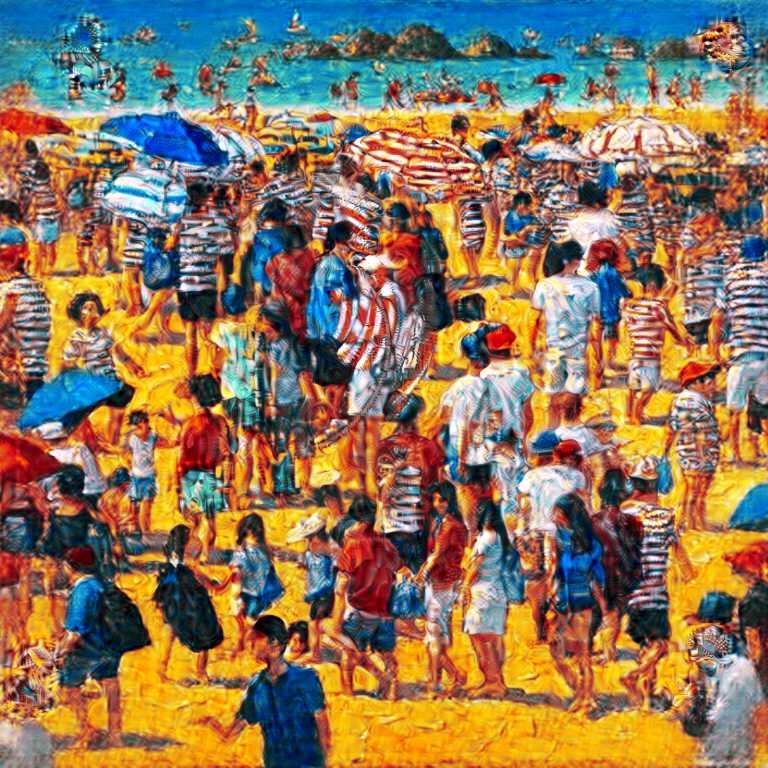}
    \end{minipage}\hspace{0.01\linewidth} 
    \begin{minipage}[c]{0.315\linewidth} 
        \includegraphics[trim=0 0 0 40, clip, width=\linewidth]{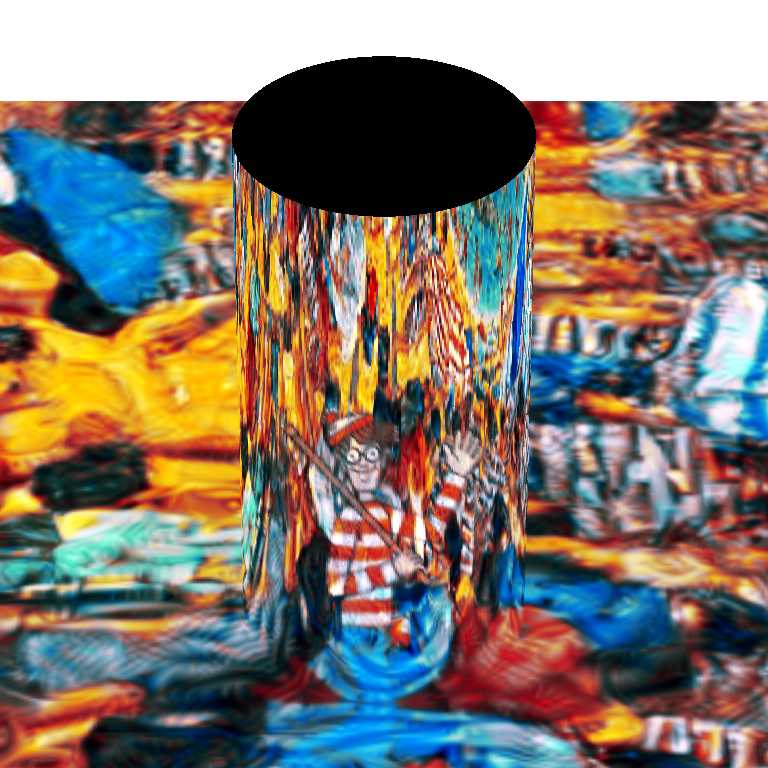}
    \end{minipage}\hspace{0.01\linewidth}
     \begin{minipage}[c]{0.312\linewidth} 
        \includegraphics[trim=0 0 0 40, clip, width=\linewidth]{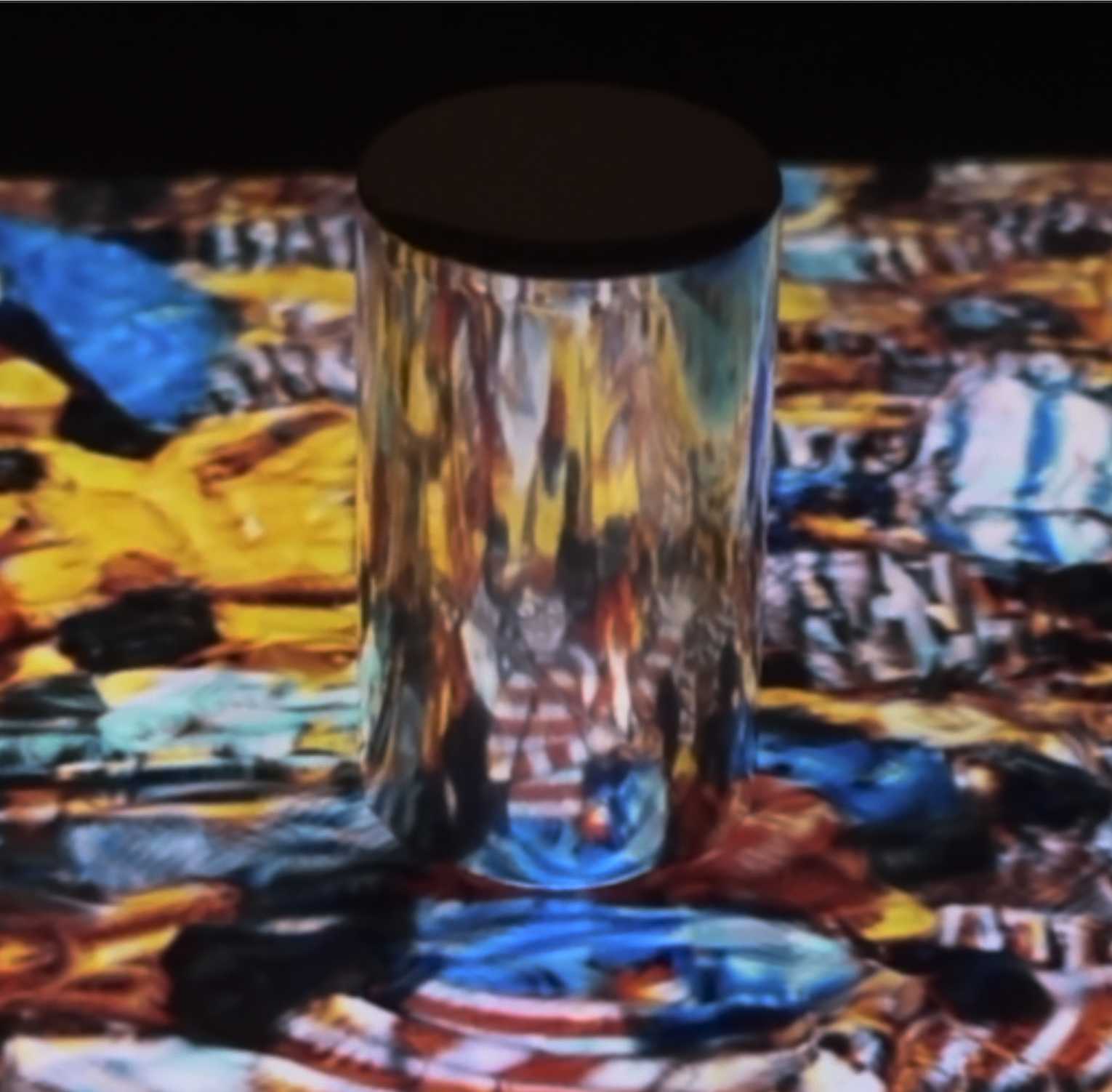}
    \end{minipage}
    \\

    \begin{minipage}[c]{0.3\linewidth} 
    \vspace{1.1mm}
    \hspace{-2mm}
        \includegraphics[width=\linewidth, trim=380 20 380 30, clip]{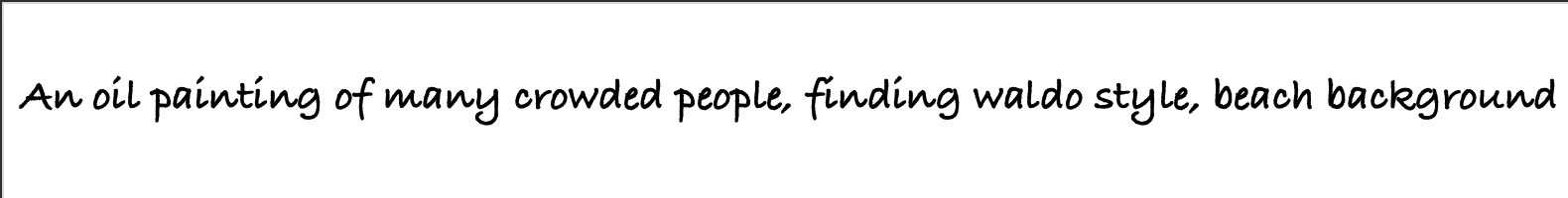}
    \end{minipage}\hspace{0.01\linewidth} 
    \begin{minipage}[c]{0.3\linewidth}
    \centering
        \scriptsize{Waldo image}
    \end{minipage}\hspace{0.01\linewidth}
    \begin{minipage}[c]{0.3\linewidth}
    \centering
    \vspace{-0.6mm}
        \scriptsize{Real}
    \end{minipage}\\
    \vspace{-4mm}\caption{\textbf{Personalized image illusion generation with reflective surfaces.} 
    Given an RGB image, we can supervise the generation of an image of one view by just using L2 loss and text to generate an image of another content. We reinvent the Finding Waldo game with a reflective cylinder. A real-world example is in the 3\textsuperscript{rd} image. 
    }
    \vspace{-4mm}
    \label{fig:waldo}
\end{figure}

\begin{figure}[htbp]
    \vspace{-3mm}
    \centering
    \tiny

    \begin{minipage}[c]{0.33\linewidth}
        \centering
        \includegraphics[trim=0 0 0 0, clip, width=\linewidth]{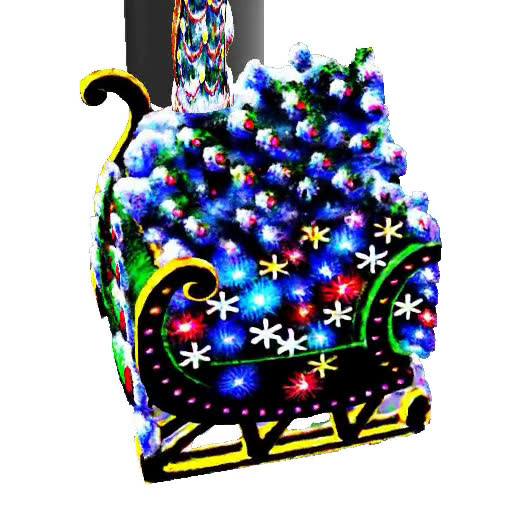}
    \end{minipage}%
    \begin{minipage}[c]{0.33\linewidth}
        \centering
        \includegraphics[trim=0 0 0 0, clip, width=\linewidth]{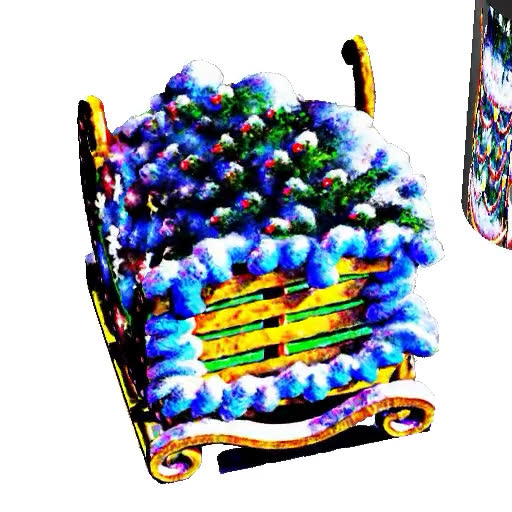}
    \end{minipage}%
    \begin{minipage}[c]{0.33\linewidth}
        \centering
        \includegraphics[trim=0 0 0 0, clip, width=\linewidth]{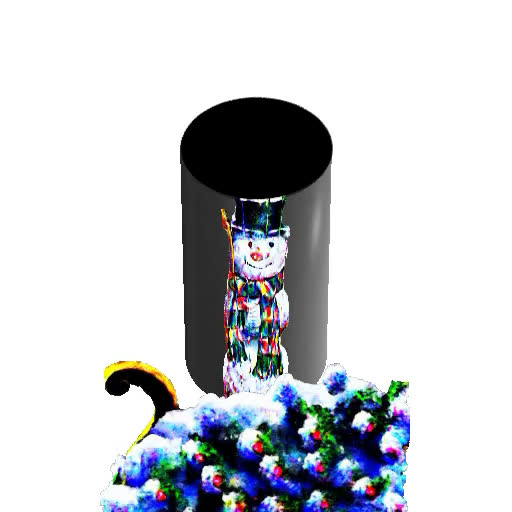}
    \end{minipage}\\

    \begin{minipage}[c]{0.33\linewidth}
        \centering
        \includegraphics[width=\linewidth, trim=130 20 130 30, clip]{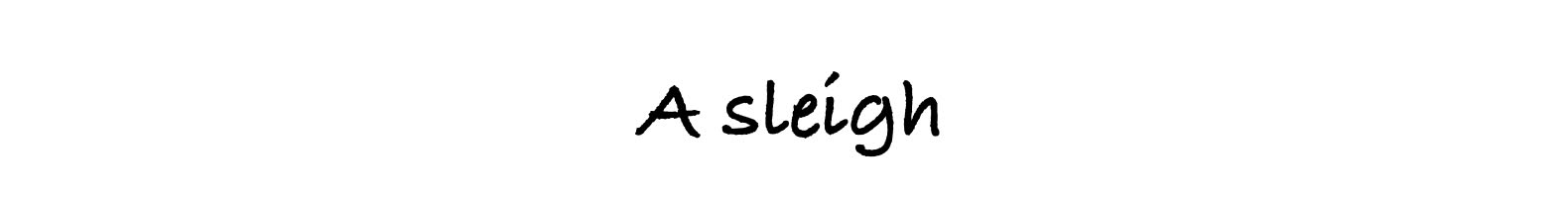}
    \end{minipage}\hfill
    \begin{minipage}[c]{0.33\linewidth}
        \centering
        \includegraphics[width=\linewidth, trim=130 20 130 30, clip]{figures/prompts/asleigh.jpg}
    \end{minipage}\hfill
    \begin{minipage}[c]{0.33\linewidth}
        \centering
        \includegraphics[width=\linewidth, trim=300 20 300 30, clip]{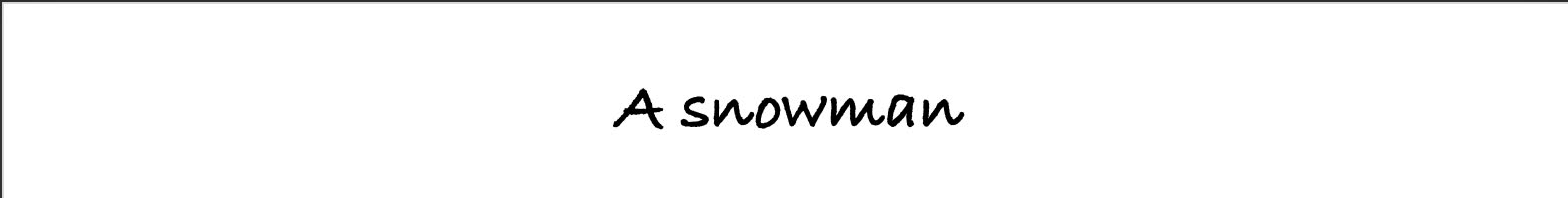}
    \end{minipage}\\

    \begin{minipage}[c]{0.33\linewidth}
        \centering
        \includegraphics[trim=0 0 0 0, clip, width=\linewidth]{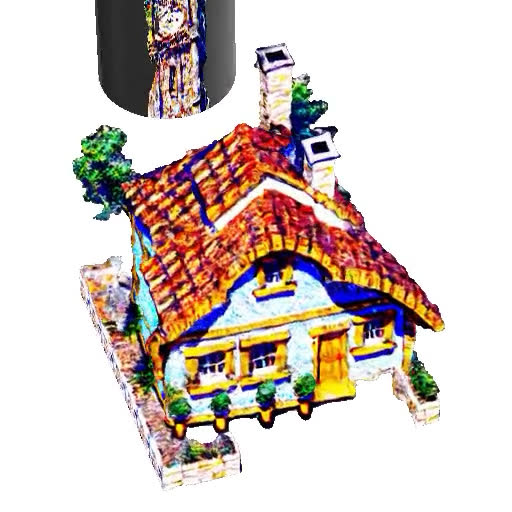}
    \end{minipage}%
    \begin{minipage}[c]{0.33\linewidth}
        \centering
        \includegraphics[trim=0 0 0 0, clip, width=\linewidth]{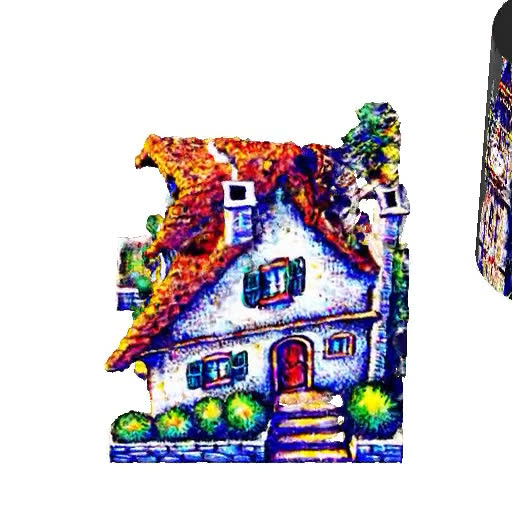}
    \end{minipage}%
    \begin{minipage}[c]{0.33\linewidth}
        \centering
        \includegraphics[trim=0 0 0 0 , clip, width=\linewidth]{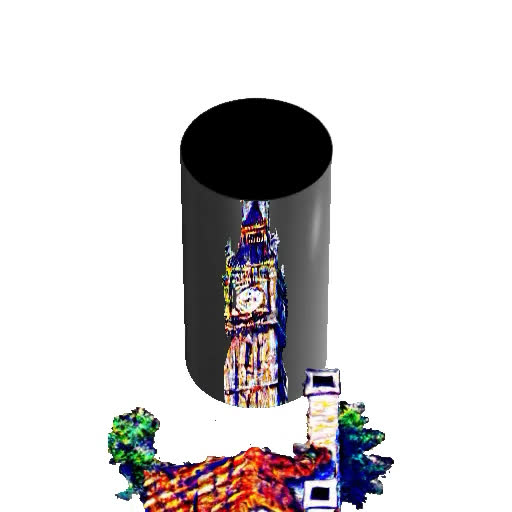}
    \end{minipage}\\

    \begin{minipage}[c]{0.33\linewidth}
        \centering
        \includegraphics[width=\linewidth, trim=300 20 300 30, clip]{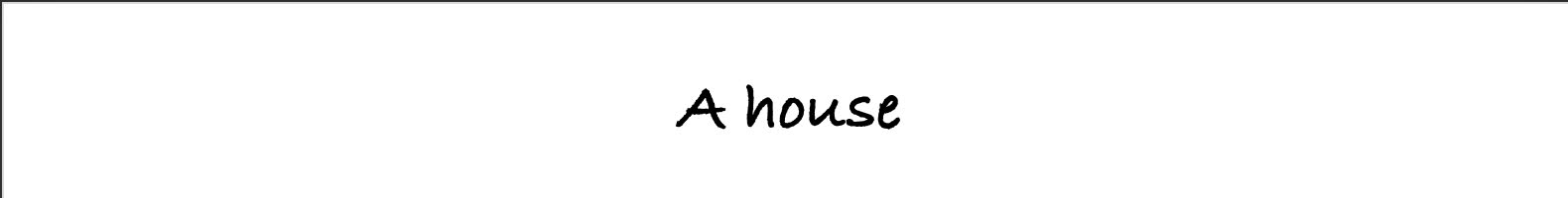}
    \end{minipage}\hfill
    \begin{minipage}[c]{0.33\linewidth}
        \centering
        \includegraphics[width=\linewidth, trim=300 20 300 30, clip]{figures/prompts/ahouse.jpg}
    \end{minipage}\hfill
    \begin{minipage}[c]{0.33\linewidth}
        \centering
        \includegraphics[width=\linewidth, trim=300 20 300 30, clip]{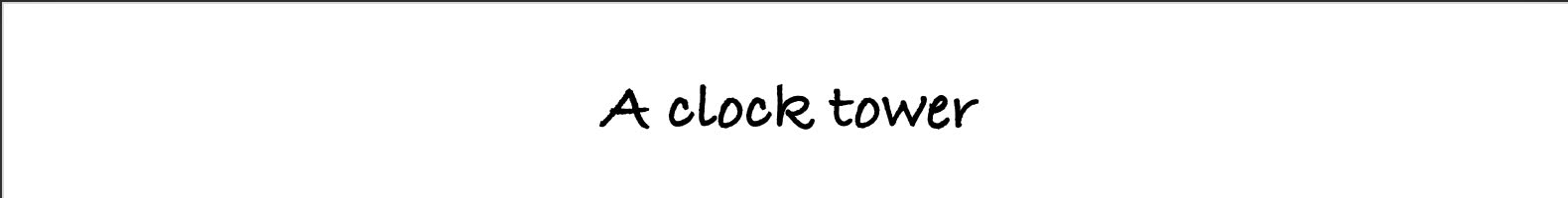}
    \end{minipage}\\
    \vspace{2mm}
  \captionsetup{skip=-0.1pt} 
    \caption{
    \textbf{3D shape illusion.} A 3D shape generation model is trained with a view from the reflected cylinder. Columns 1 and 2 are different views of the generated 3D object. Column 3 is a view of the object from the reflective cylinder.
    }
    \vspace{-4mm}
    \label{fig:result-3D}
\end{figure}

\begin{figure}[t]
\vspace{-3mm}
    \centering
    \fontsize{5.5pt}{6.5pt}\selectfont

    \begin{minipage}[c]{0.03\linewidth}
      \makebox[0pt][c]{\rotatebox{90}{\parbox{1.5cm}{\centering \textbf{Duplicate objects}}}}
    \end{minipage}
    \begin{minipage}[c]{0.32\linewidth}
        \centering
        \includegraphics[trim=150 75 150 75, clip, width=0.9\linewidth]{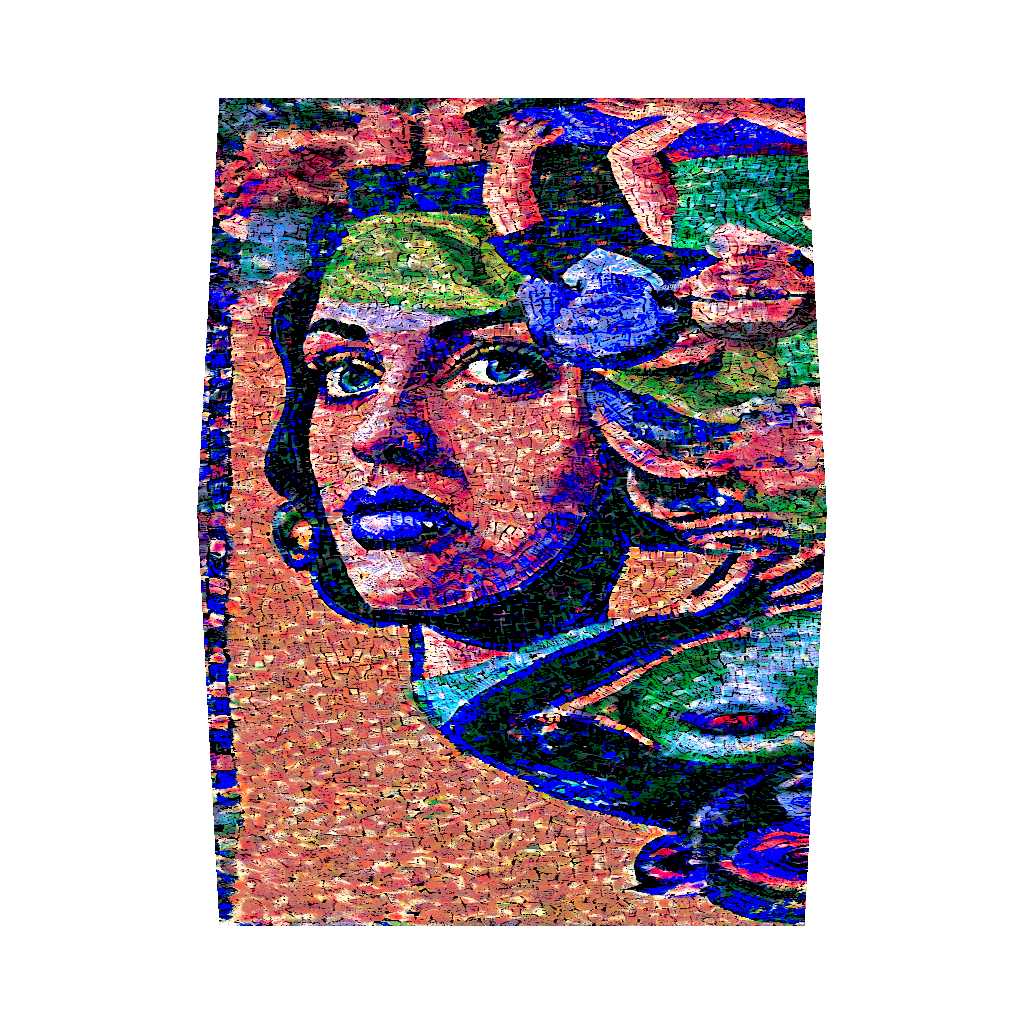}
    \end{minipage}%
    \begin{minipage}[c]{0.32\linewidth}
        \centering
        \includegraphics[trim=75 150 75 150, clip, width=\linewidth]{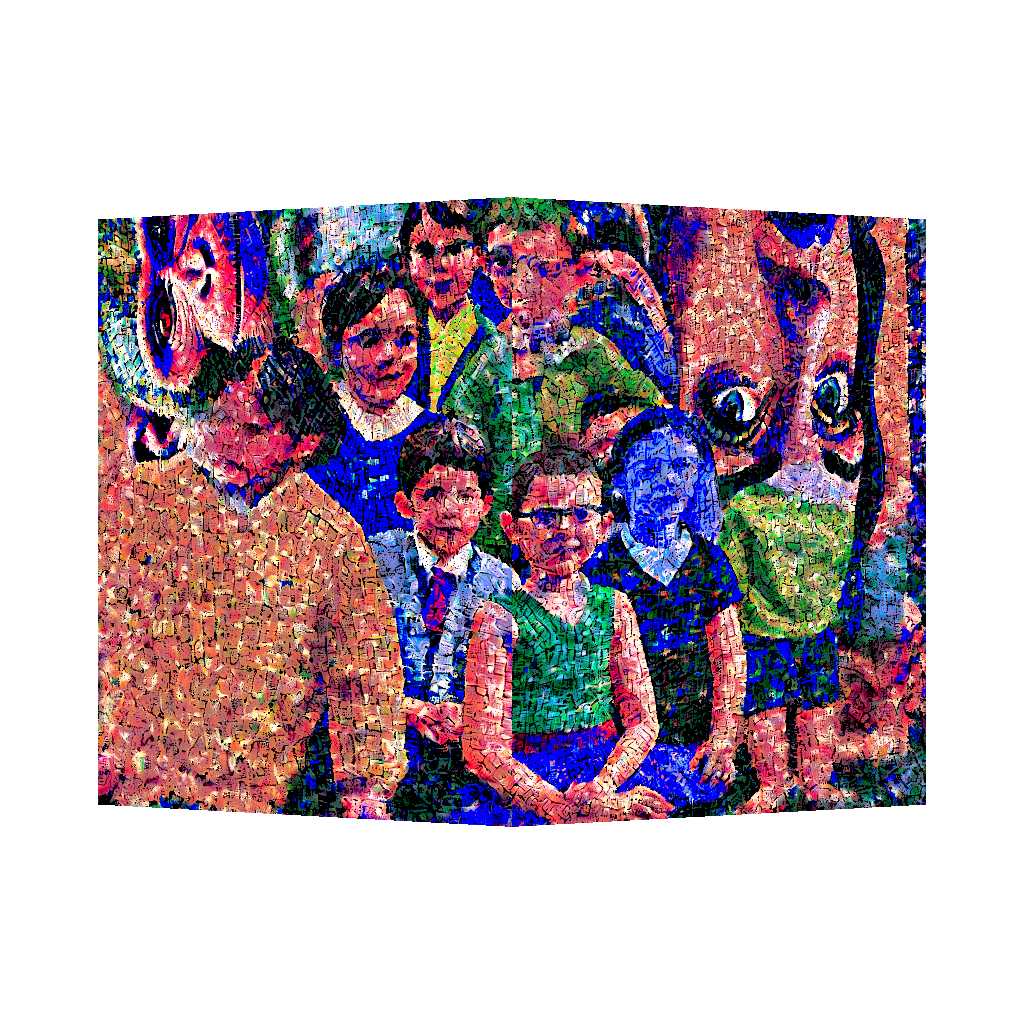}
    \end{minipage}%
    \begin{minipage}[c]{0.32\linewidth}
        \centering
        \includegraphics[trim=75 150 75 150, clip, width=\linewidth]{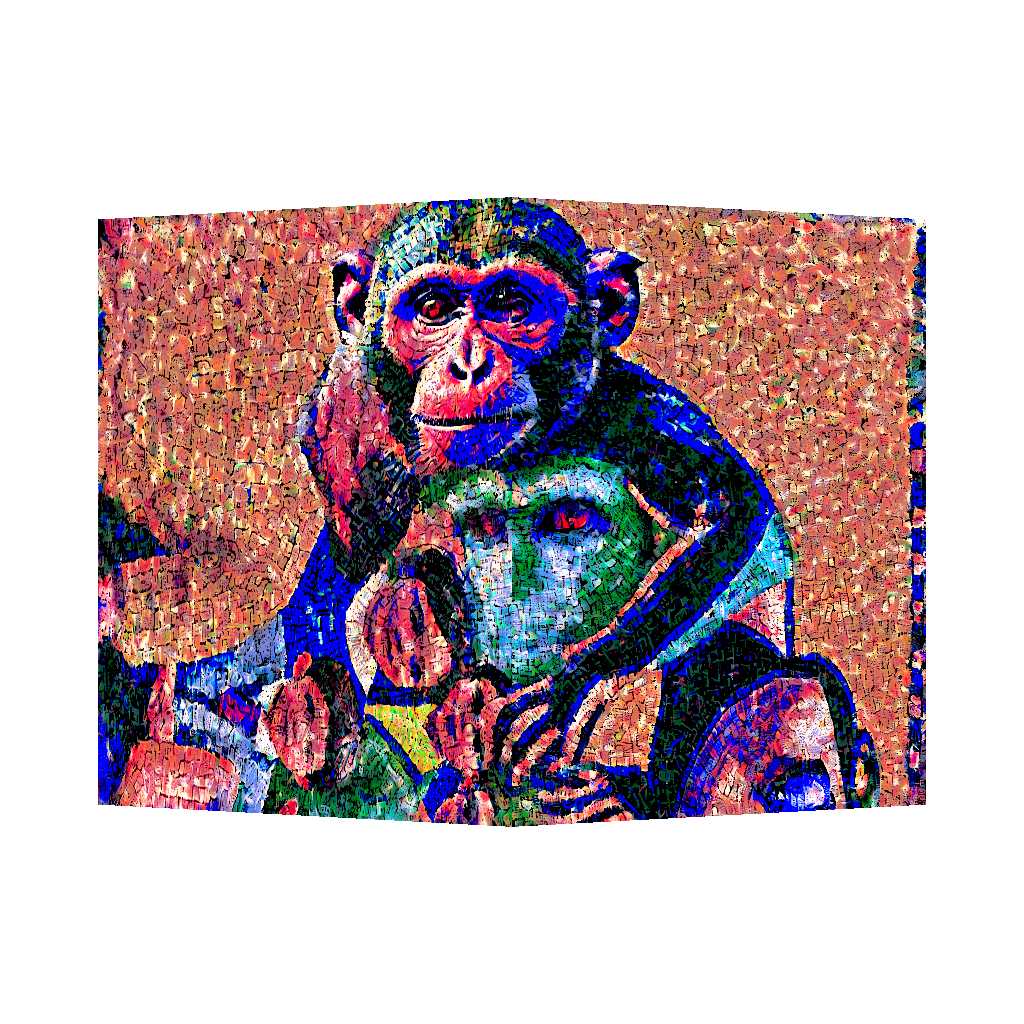}
    \end{minipage}\\

        \vspace{-1mm}
    \begin{minipage}[c]{0.03\linewidth}
      \makebox[0pt][c]{}
    \end{minipage}
    \begin{minipage}[c]{0.32\linewidth}
        \centering
        \includegraphics[width=\linewidth, trim=120 20 100 30, clip]{figures/prompt/womanwindow.jpg}
    \end{minipage}\hfill
    \begin{minipage}[c]{0.32\linewidth}
        \centering
        \includegraphics[width=0.85\linewidth, trim=220 20 220 30, clip]{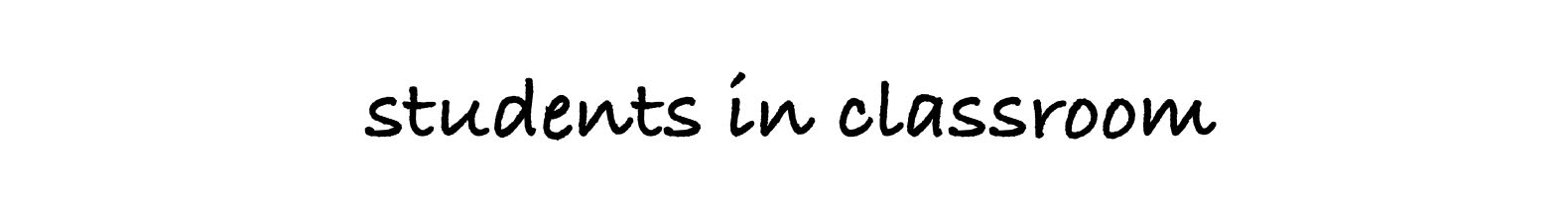}
    \end{minipage}\hfill
    \begin{minipage}[c]{0.32\linewidth}
        \centering
        \includegraphics[width=\linewidth, trim=300 20 300 30, clip]{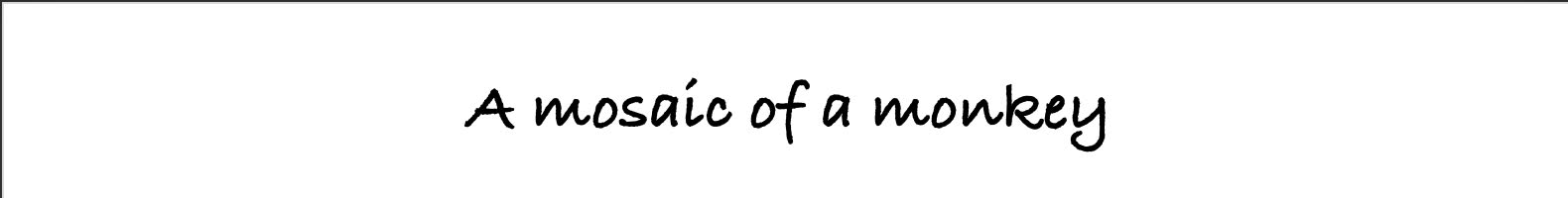}
    \end{minipage}\\

    \begin{minipage}[c]{0.03\linewidth}
      \makebox[0pt][c]{\rotatebox{90}{\parbox{1.5cm}{\centering \textbf{Blending objects}}}}
    \end{minipage}
    \begin{minipage}[c]{0.32\linewidth}
        \centering
        \includegraphics[trim=150 75 150 75, clip, width=0.9\linewidth]{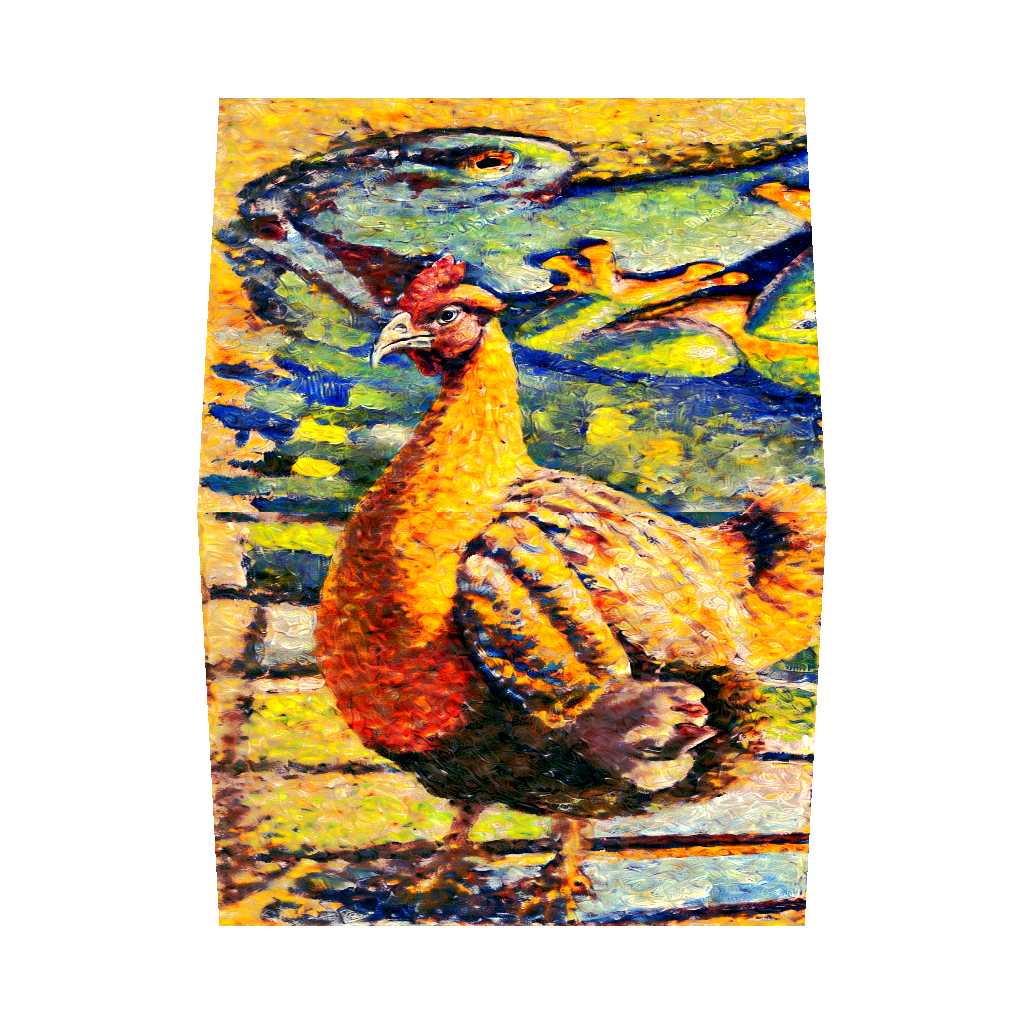}
    \end{minipage}%
    \begin{minipage}[c]{0.32\linewidth}
        \centering
        \includegraphics[trim=75 150 75 150, clip, width=\linewidth]{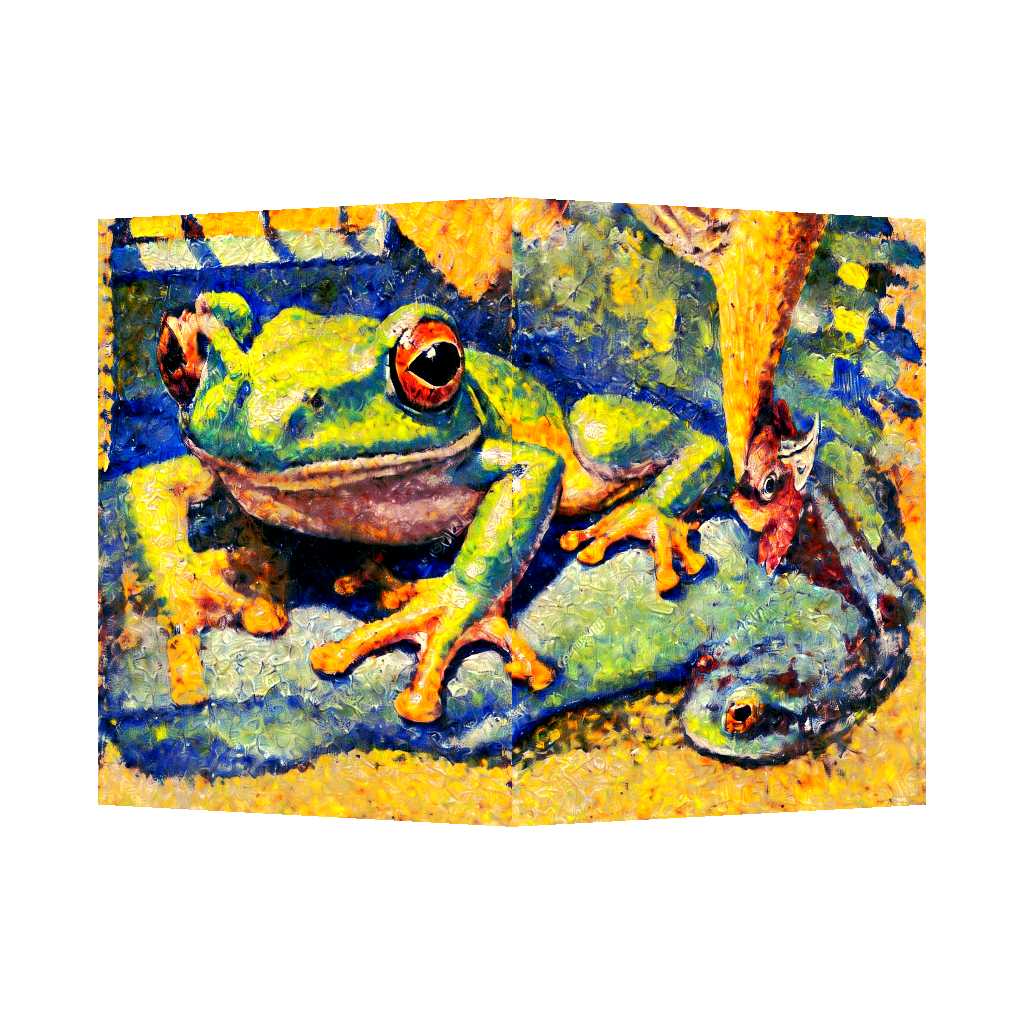}
    \end{minipage}%
    \begin{minipage}[c]{0.32\linewidth}
        \centering
        \includegraphics[trim=75 150 75 150, clip, width=\linewidth]{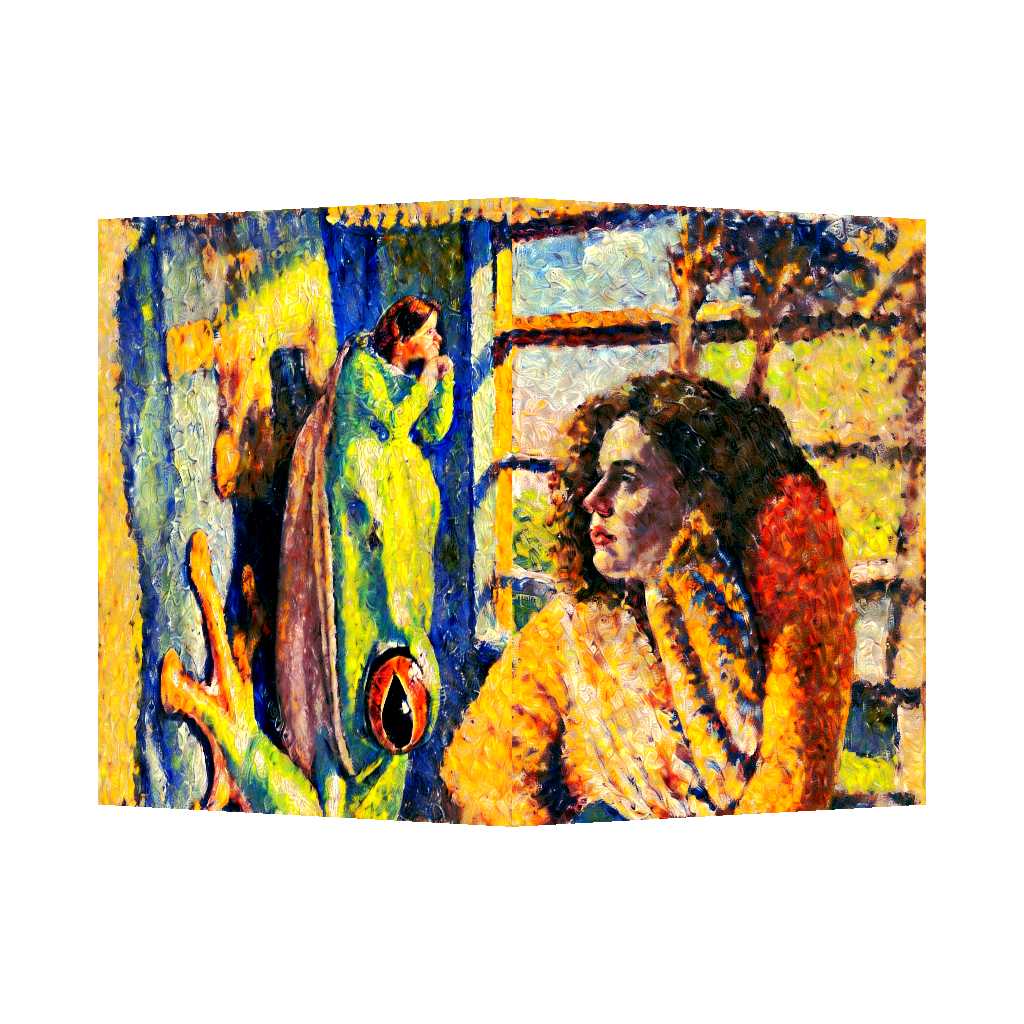}
    \end{minipage}\\

    \vspace{-1mm}
    \begin{minipage}[c]{0.03\linewidth}
      \makebox[0pt][c]{}
    \end{minipage}
    \begin{minipage}[c]{0.32\linewidth}
        \centering
        \includegraphics[width=0.3\linewidth, trim=975 20 370 30, clip]{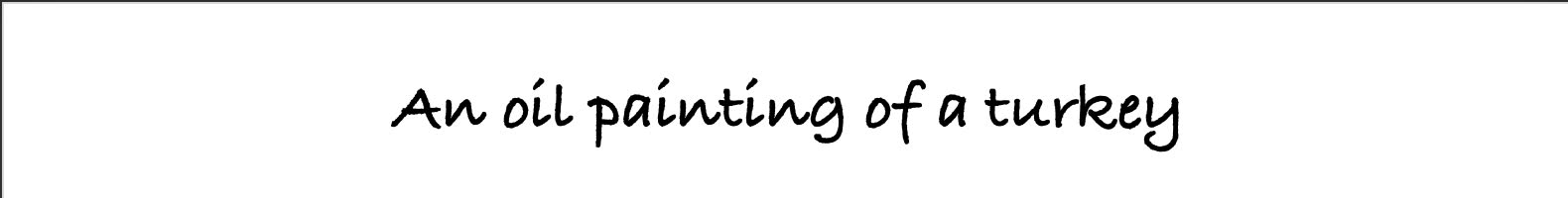}
    \end{minipage}\hfill
    \begin{minipage}[c]{0.32\linewidth}
        \centering
        \includegraphics[width=0.2\linewidth, trim=1030 20 410 30, clip]{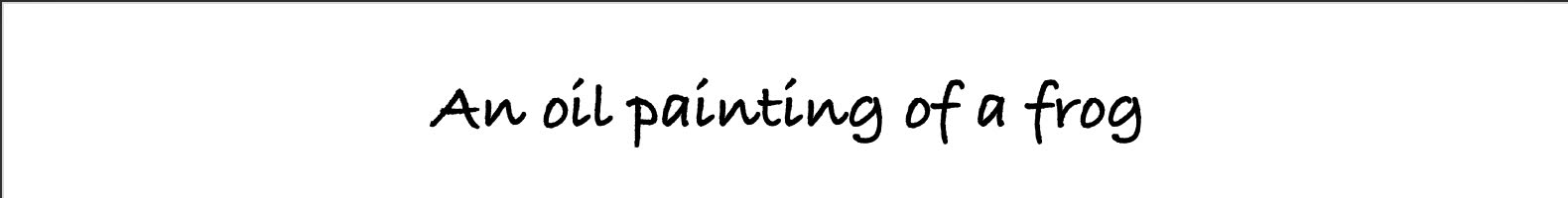}
    \end{minipage}\hfill
    \begin{minipage}[c]{0.32\linewidth}
        \centering
        \includegraphics[width=\linewidth, trim=120 20 100 30, clip]{figures/prompt/womanwindow.jpg}
    \end{minipage}\\
    \vspace{-2mm}
    \caption{
    \textbf{Failure cases.} We observe duplicate object generation and blending of objects among each view.  Row 1, duplicate monkey faces in the third view. Row 2, blending on primary objects.
    }
    \label{fig:result-failurecase}
    \vspace{-3mm}
\end{figure}

\begin{figure}[t]
\centering
\fontsize{5.5pt}{6.5pt}\selectfont

\begin{minipage}[c]{0.03\linewidth}
  \makebox[0pt][c]{\rotatebox{90}{\parbox{1.5cm}{\centering \textbf{Orthogonal views}}}}
\end{minipage}%
\begin{minipage}[c]{0.97\linewidth}
  \centering
  \includegraphics[trim=140 140 140 140, clip, width=0.2\linewidth]{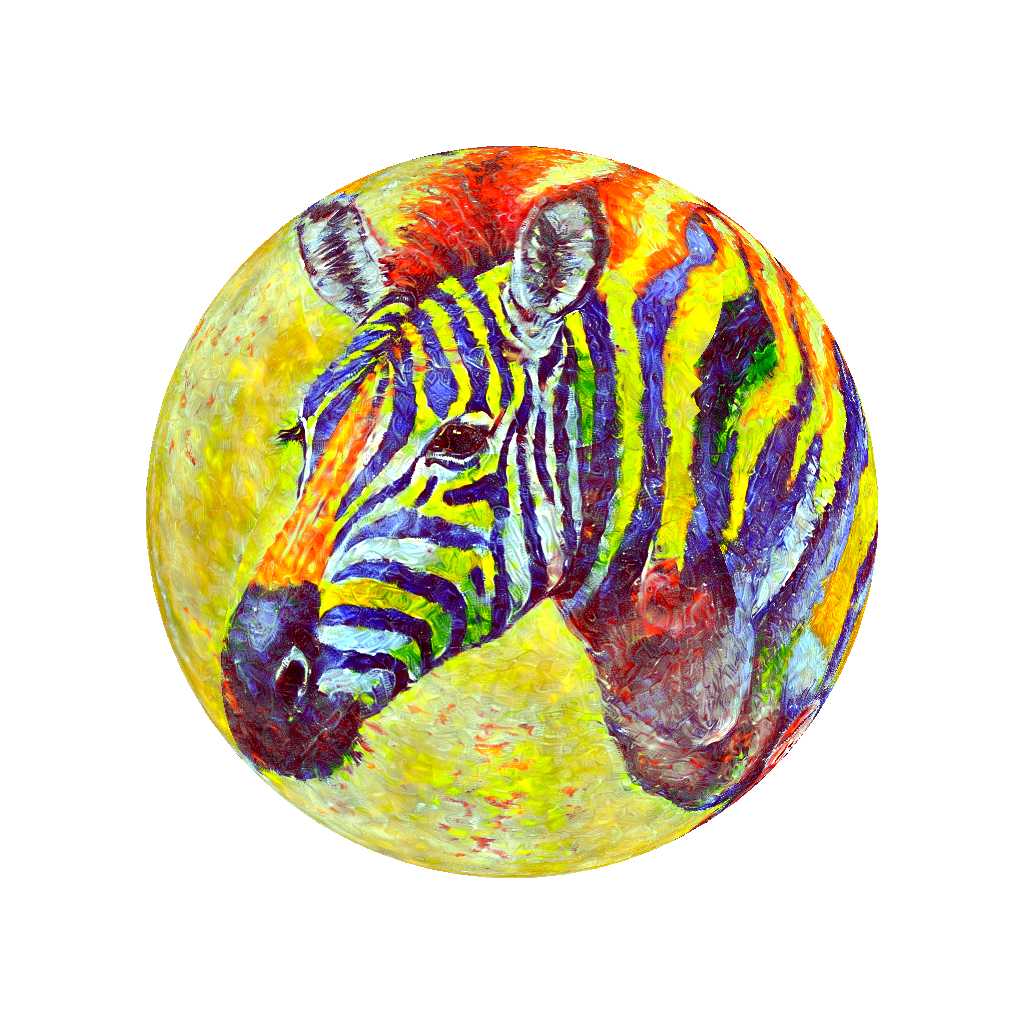}\hfill  
  \includegraphics[trim=140 140 140 140, clip, width=0.2\linewidth]{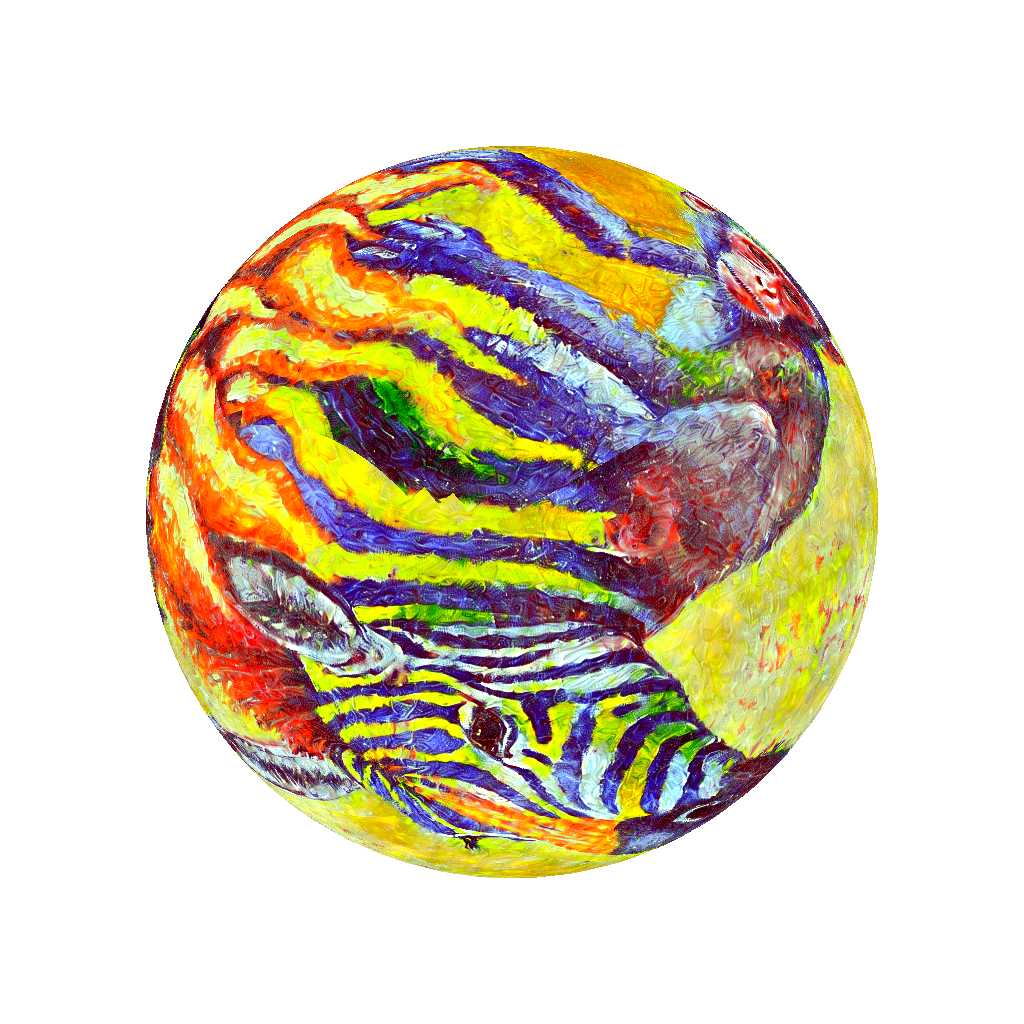}\hfill
  \includegraphics[trim=140 140 140 140, clip, width=0.2\linewidth]{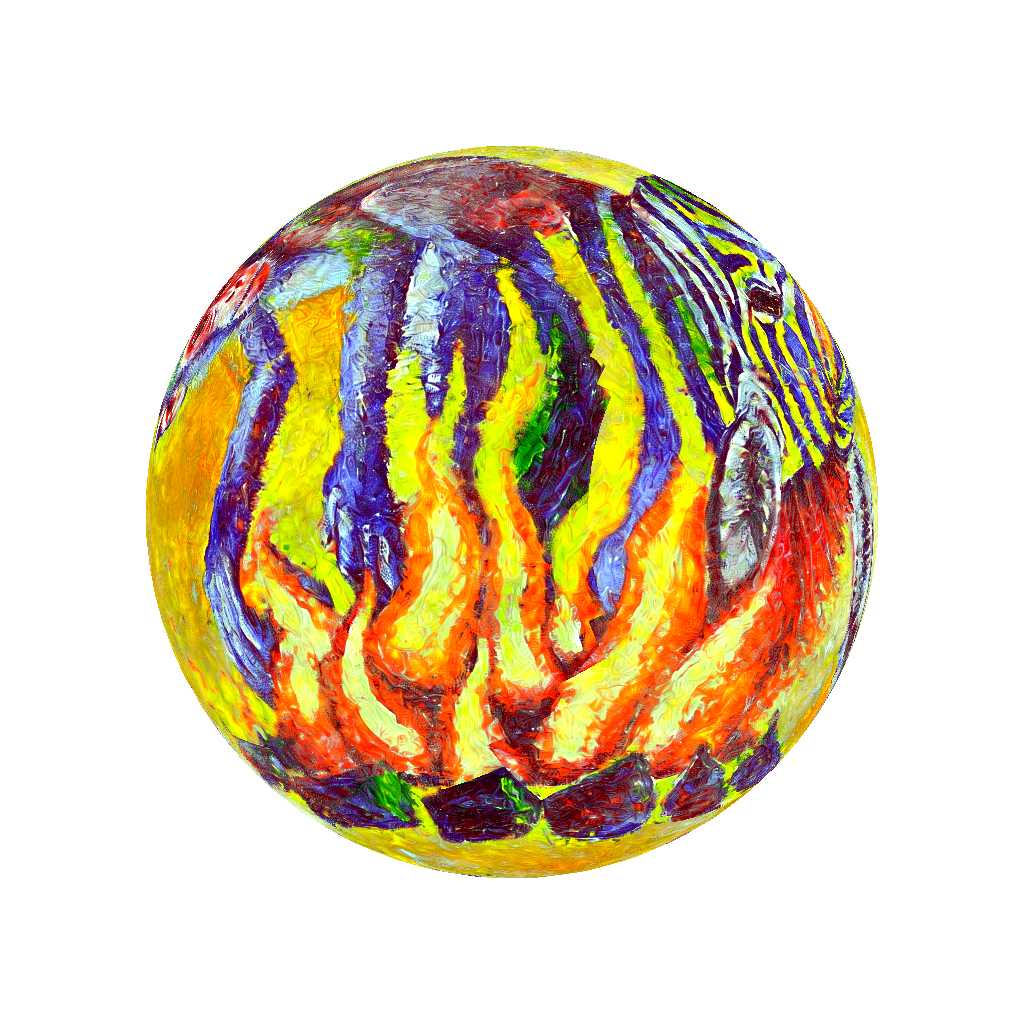}\hfill
  \includegraphics[trim=140 140 140 140, clip, width=0.2\linewidth]{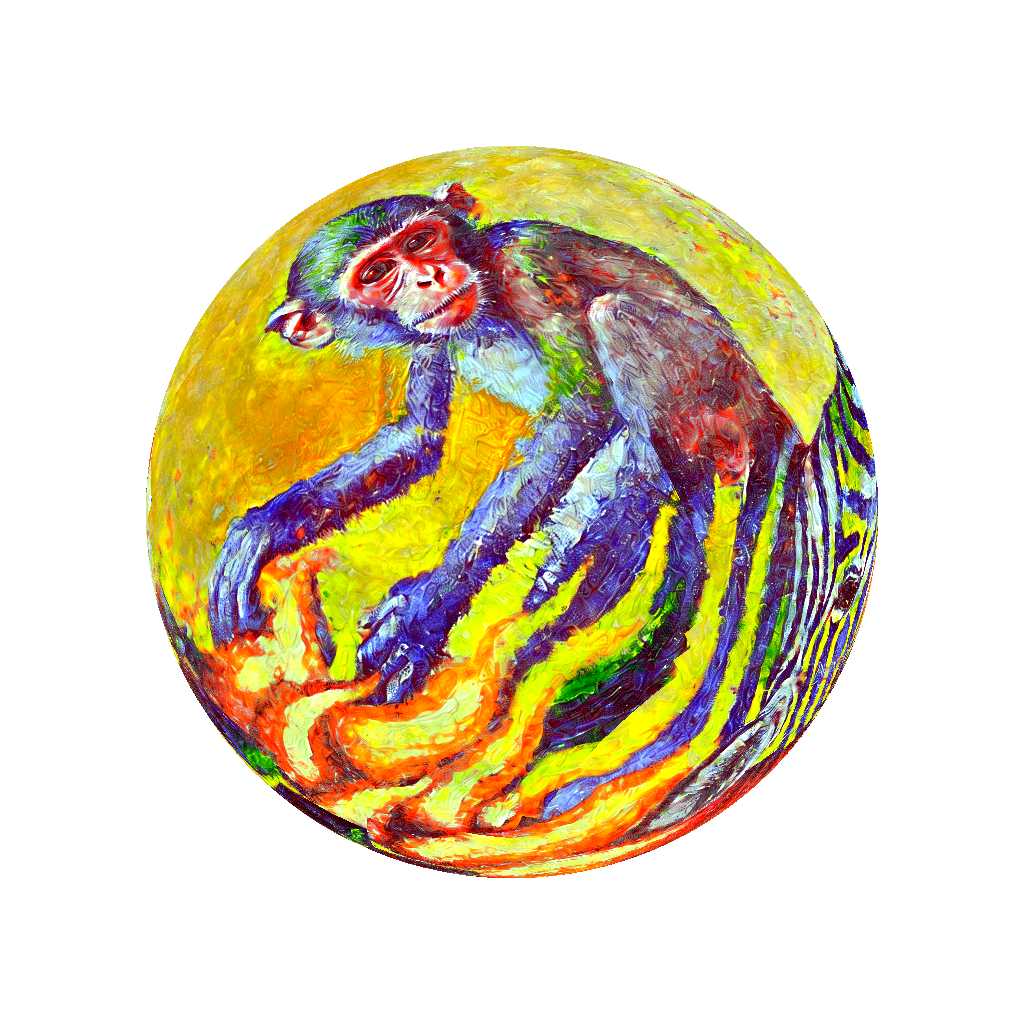}\hfill
  \includegraphics[trim=140 140 140 140, clip, width=0.2\linewidth]{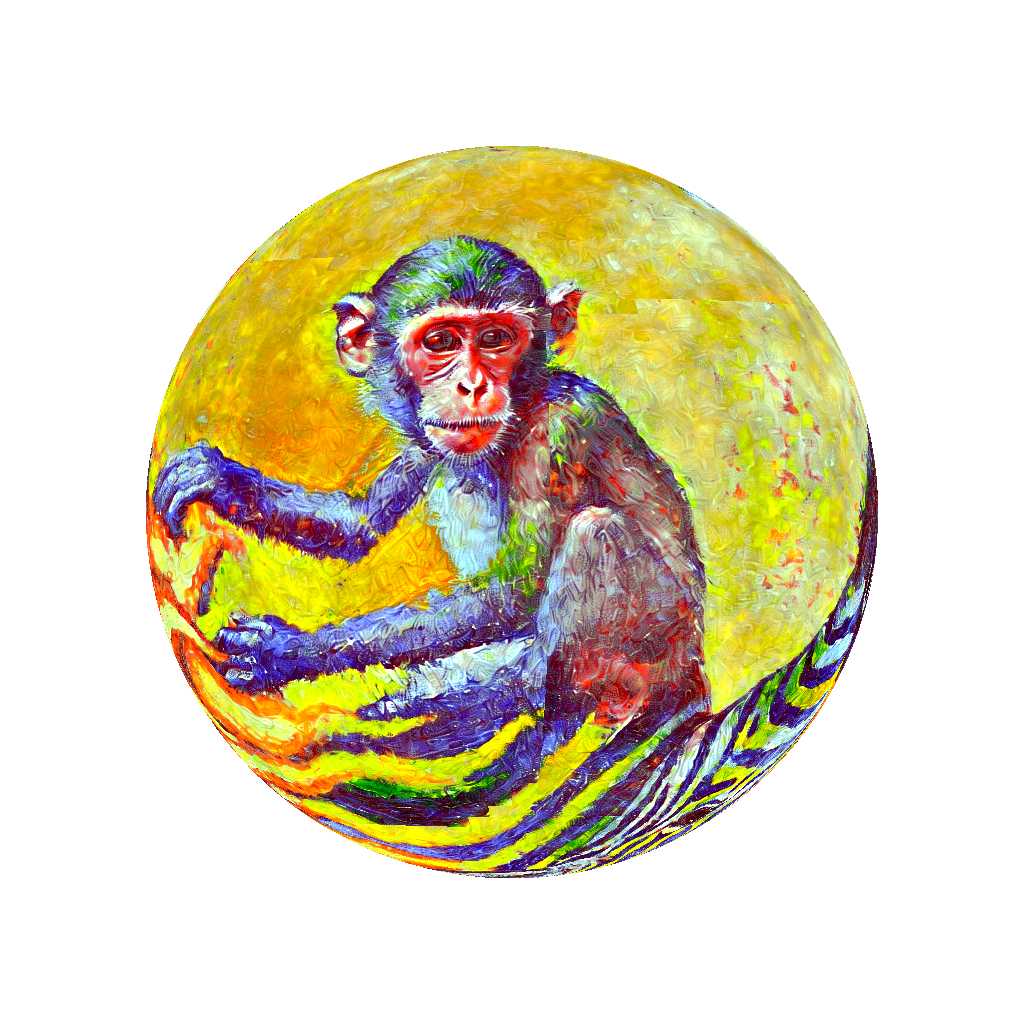}\hfill

\end{minipage}\\
\begin{minipage}[c]{0.03\linewidth}
  \makebox[0pt][c]{\rotatebox{90}{\parbox{1.5cm}{\centering \textbf{Shrink views}}}}
\end{minipage}%
\begin{minipage}[c]{0.97\linewidth}
  \centering
  \includegraphics[trim=140 140 140 140, clip, width=0.2\linewidth]{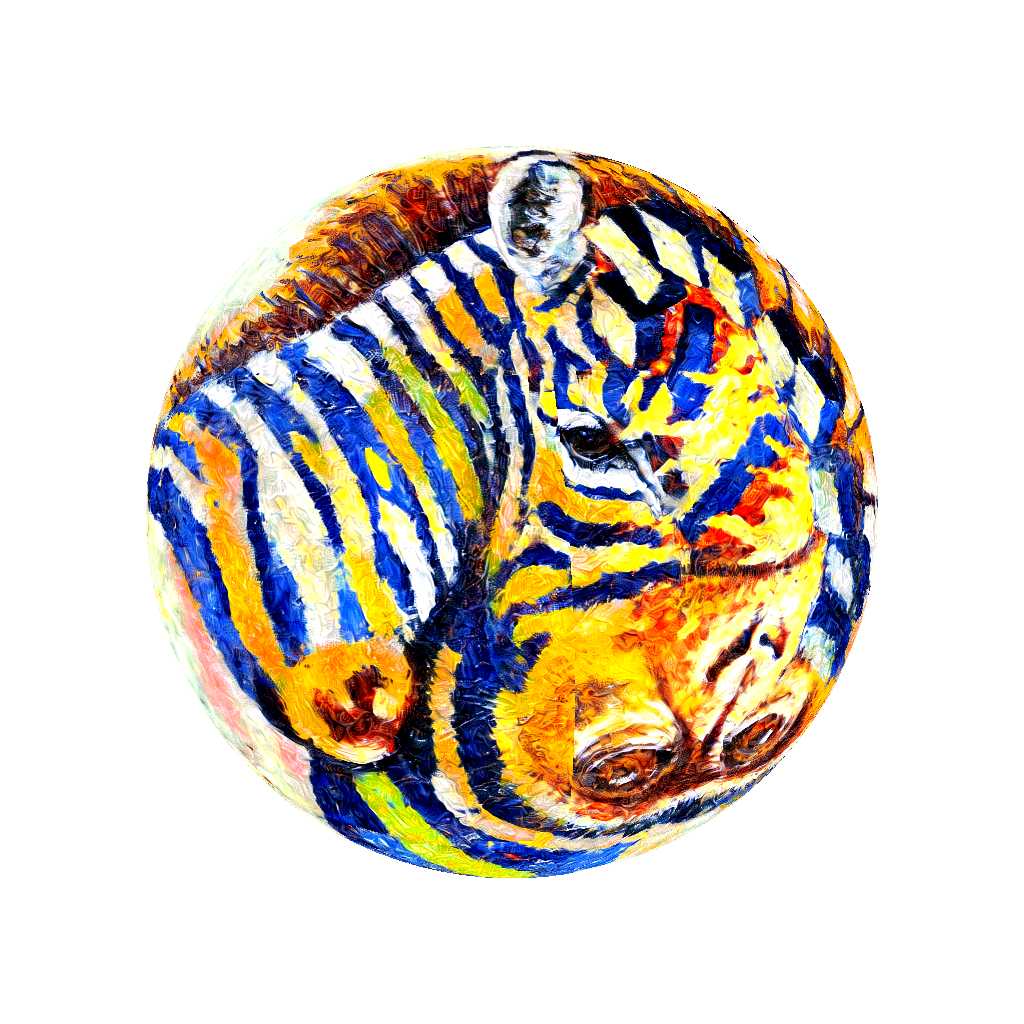}\hfill
  \includegraphics[trim=140 140 140 140, clip, width=0.2\linewidth]{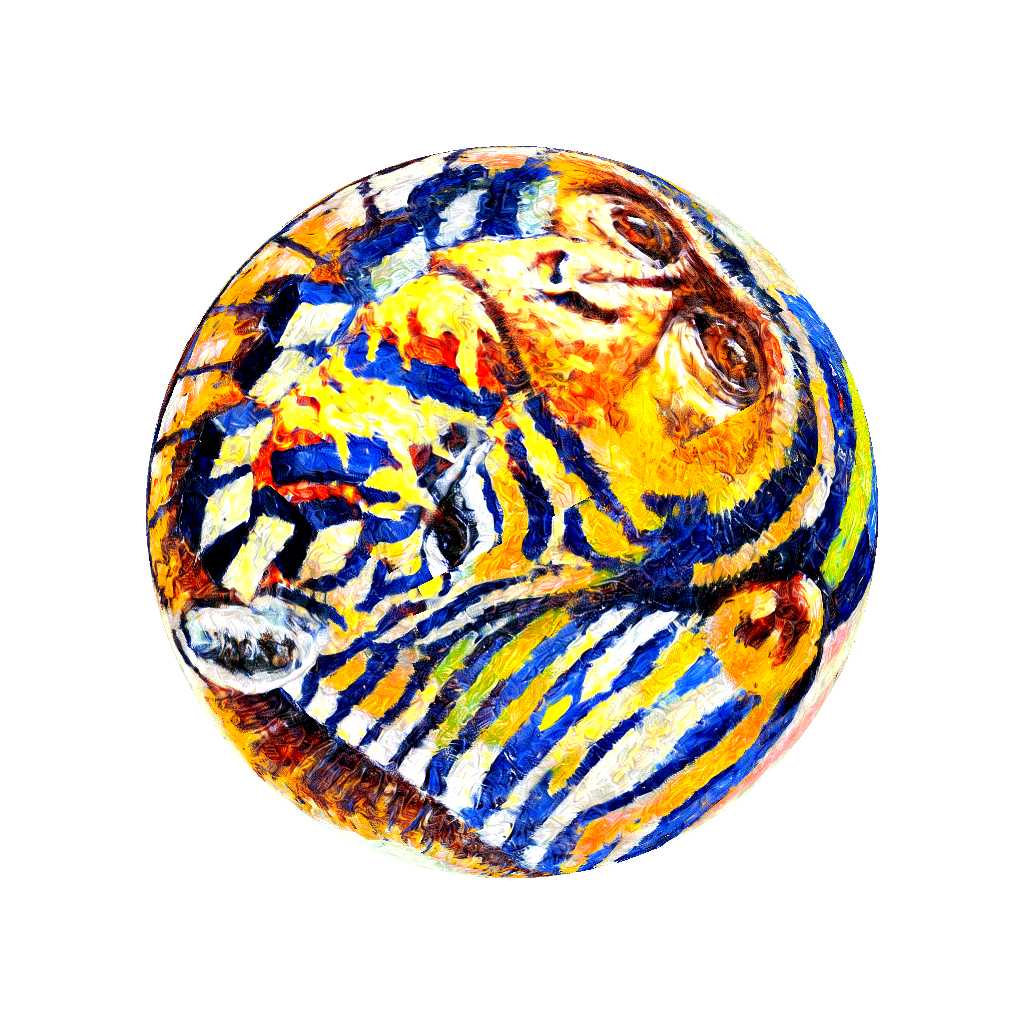}\hfill
  \includegraphics[trim=140 140 140 140, clip, width=0.2\linewidth]{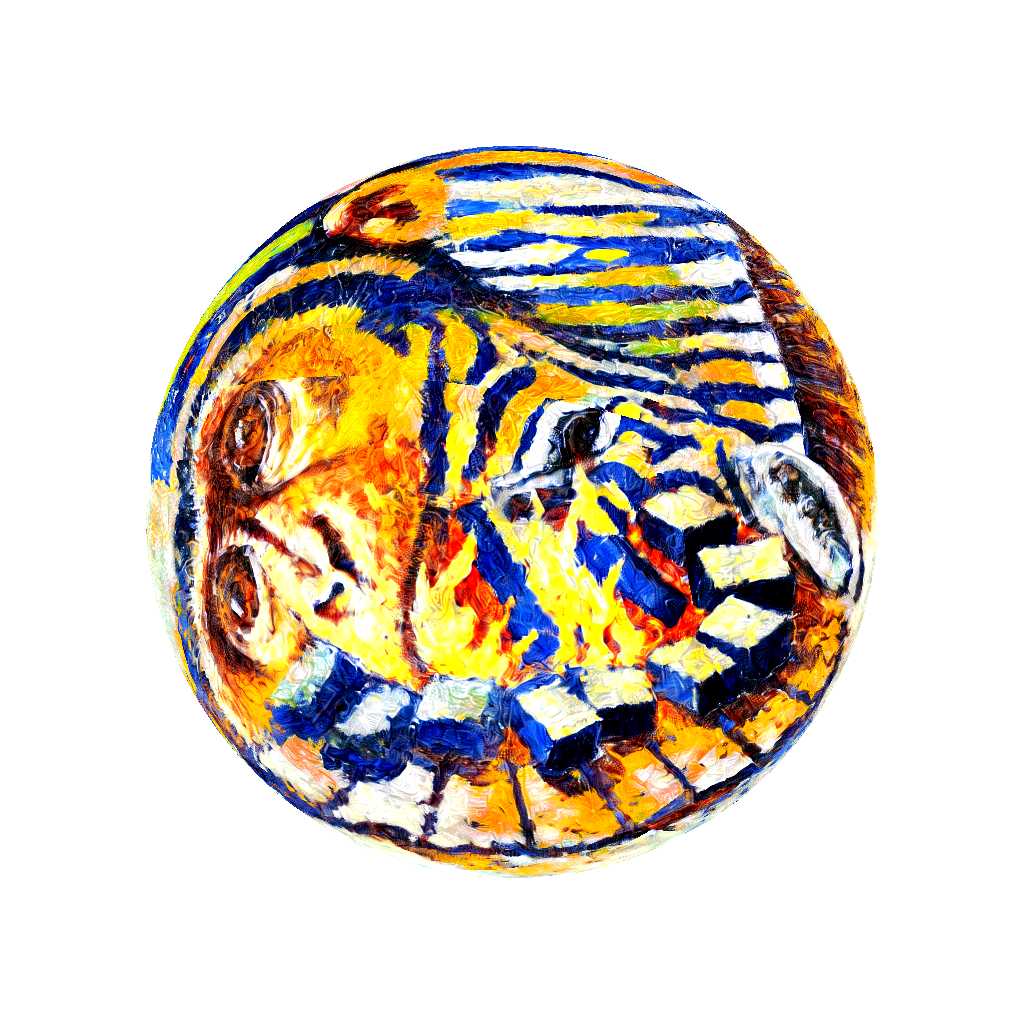}\hfill
  \includegraphics[trim=140 140 140 140, clip, width=0.2\linewidth]{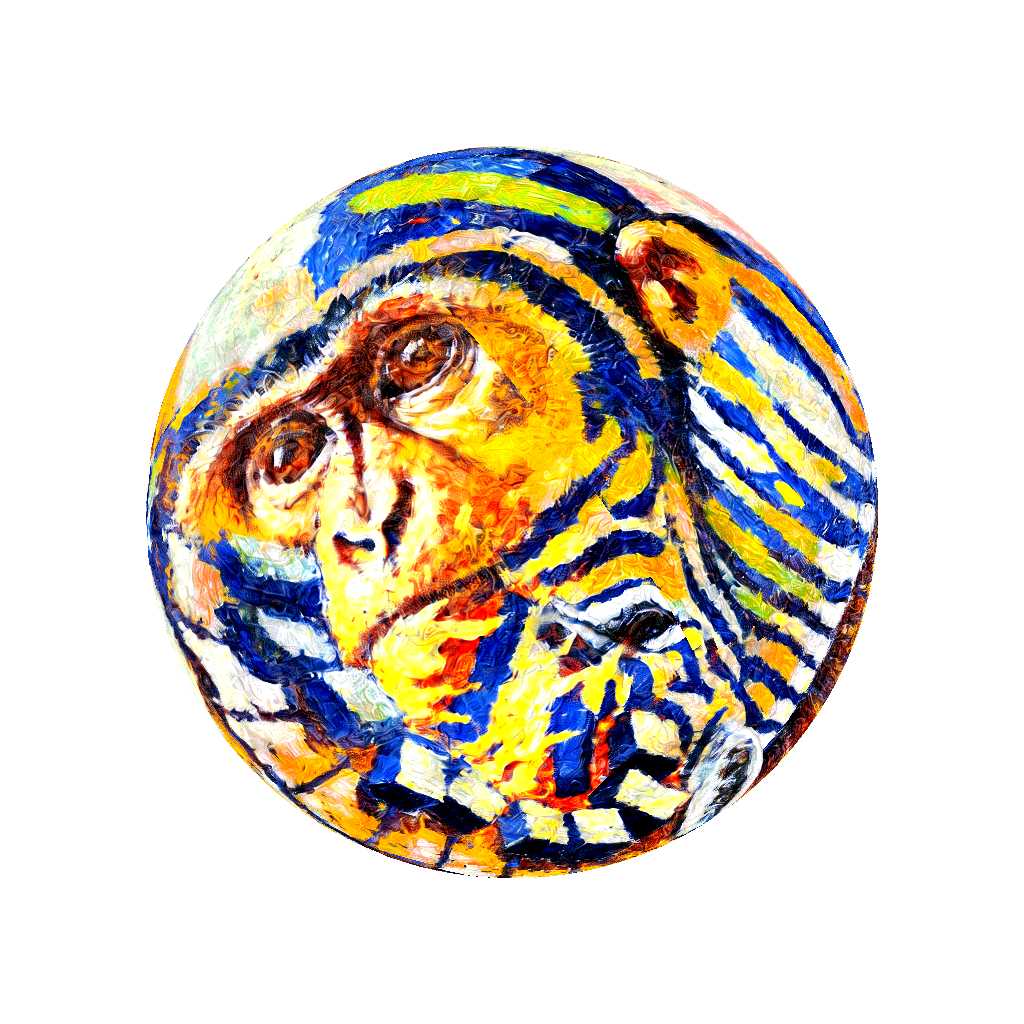}\hfill
  \includegraphics[trim=140 140 140 140, clip, width=0.2\linewidth]{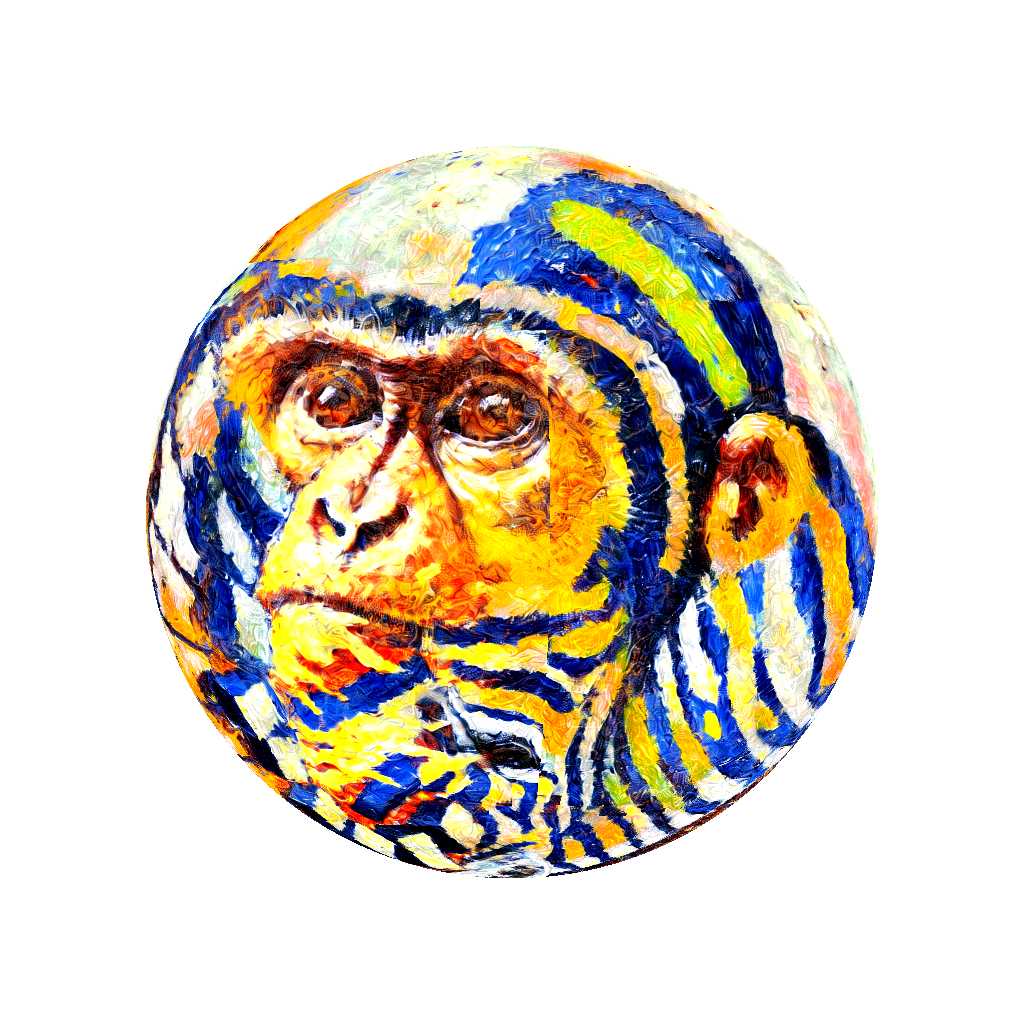}\hfill

\end{minipage}\\

\begin{minipage}[c]{0.03\linewidth}
  \makebox[0pt][c]{}
\end{minipage}%
\begin{minipage}[c]{0.2\linewidth}
    \centering
    \includegraphics[width=0.4\linewidth, trim=1000 20 390 30, clip]{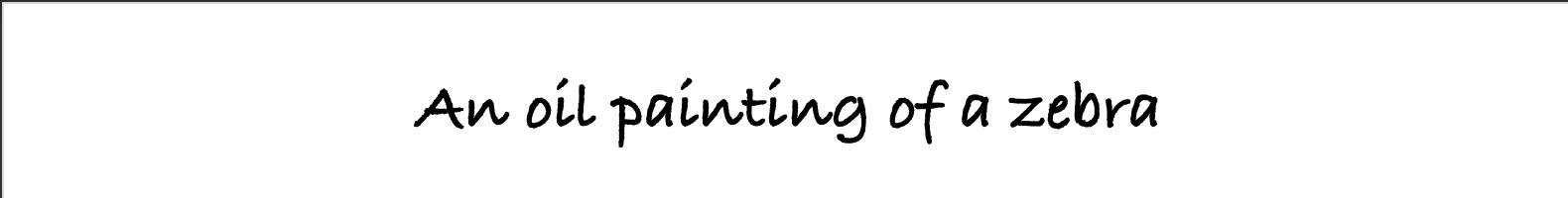}
\end{minipage}\hfill
\begin{minipage}[c]{0.2\linewidth}
    \centering    
\end{minipage}\hfill
\begin{minipage}[c]{0.2\linewidth}
    \centering
    \includegraphics[width=0.5\linewidth, trim=960 20 350 30, clip]{figures/prompts/campfire.jpg}
\end{minipage}\hfill
\begin{minipage}[c]{0.2\linewidth}
    \centering
    
\end{minipage}\hfill
\begin{minipage}[c]{0.2\linewidth}
    \centering
    \includegraphics[width=0.5\linewidth, trim=960 20 350 30, clip]{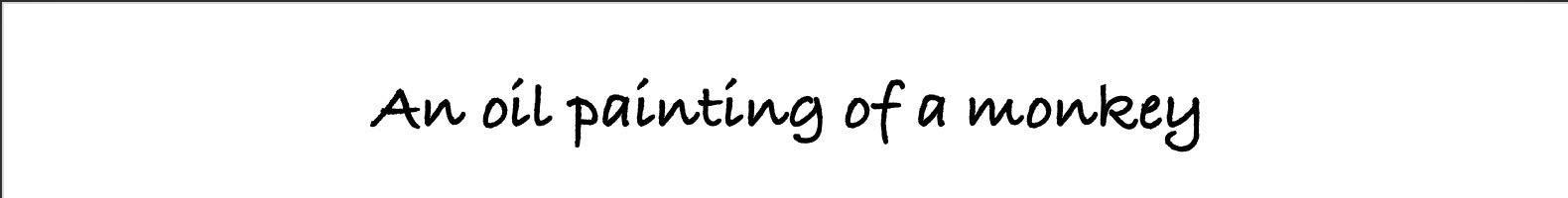}
\end{minipage}\\
\vspace{-3mm}
\caption{\textbf{Larger area overlapped by camera views on the sphere.} We make the camera views from three orthogonal views to more compact camera view positions (shrink views) on the sphere case. Intermediate views are presented in the 2\textsuperscript{nd} and 4\textsuperscript{th} columns.}

\label{fig:shrink-view}
\vspace{-4mm}
\end{figure}

\begin{figure}[t]
    \vspace{-1mm}
    \centering
    \includegraphics[width=\linewidth, trim=1 0 0 0 , clip]{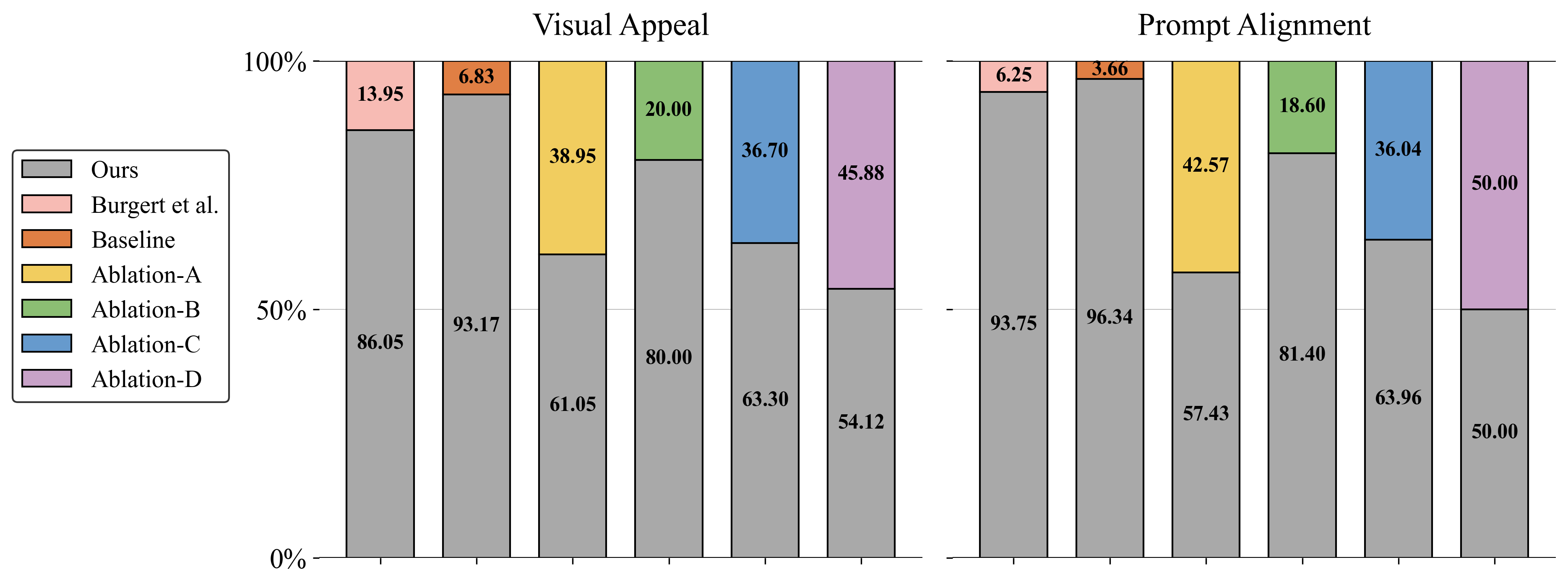}
    \vspace{-8mm}
    \caption{
    \textbf{User study: comparison with other design choices.}
    We compare our method pairwise with alternative design choices in Fig.~\ref{fig:comparison}. We calculate the percentage of cases where users prefer each design choice over ours based on visual appeal and prompt alignment. Study performed over 40 users comparing random sets of 24 results with corresponding results from our method.}
    \label{fig:userstudyfig}
    \vspace{-6mm}
\end{figure}

\section{Experiments}
\label{sec:Results}
We validate our method through comprehensive qualitative and quantitative evaluations, systematically ablating the impact of key design choices. We use a single set of parameters on all examples and do not cherry-pick results during optimization. More implementation details \ref{sec:implementation}, random results \ref{sec:addtional}, ablations \ref{sec:ablationsupp}, and discussions \ref{sec:discussionsupp}, are in the Supp..

\topic{Baselines}
We refer to the \(512 \times 512\) resolution setting described in Sec.~\ref{sec:background}, using the naive approach without patch denoising, camera jitter, or resolution scaling, as our baseline.
We compare our approach against the following baselines.
The inverse projection baseline utilizes Stable Diffusion~\cite{Rombach_2022_CVPR} to generate images independently for each of the three prompts, followed by an optimization of the cube’s texture map via inverse projection onto the 3D shape. 
Similarly, the latent blending baseline generates images independently for the three prompts using Stable Diffusion and then blends these views by averaging their latent representations in the overlapping regions. 
Additionally, we denote the baseline approach incorporating the Dream Target loss from 2D illusion~\cite{Burgert2023DiffusionIH} as Burgert \textit{et al.}~\cite{Burgert2023DiffusionIH}.
Fig.~\ref{fig:otherbaselines} presents a comparative analysis of our results against these baseline methods.


\topic{Datasets}
We randomly select a painting style and a primary object to construct 43 different prompt pairs to evaluate illusion generation on both spheres and cubes. This results in a total of 86 generated illusion examples. 
We evaluate the generated illusions using multiple metrics, including CLIP score~\cite{Radford2021LearningTV}, aesthetic and aesthetic artifact score~\cite{Murray2012AVAAL}, and alignment and concealment scores~\cite{Burgert2023DiffusionIH, Geng2023VisualAG}. We use these metrics to assess performance across baselines and various design choices. We include additional discussions on prompt selection in the Supp. Sec.~\ref{sec:discussionsupp}.

\begin{table}[ht]
\vspace{-3mm}
  \centering
   \footnotesize
            \caption{\textbf{Comparison with different baselines.}}
    \vspace{-3.5mm}
  \captionsetup{skip=4pt} 
  \begin{tabular}{@{}lccccc@{}}

    \toprule
    Method & CLIP $\uparrow$ & Aes $\uparrow$ & Aes artifact $\downarrow$ & $\mathcal{A} \uparrow$  & $\mathcal{C} \uparrow$  \\ 
    \midrule
    Inverse projection   & 0.271 & 2.303 & 3.505 & 0.229 & 0.385 \\ 
    Latent blending  & 0.193 & 2.528 & 3.413 & 0.160 & 0.272 \\ 
    Burgert \textit{et al.} ~\cite{Burgert2023DiffusionIH}& 0.179 & \textbf{2.958} & 3.309 & 0.144 & 0.206\\ 
    Ours & \textbf{0.288} &  2.791 & \textbf{3.247} & \textbf{0.251} & \textbf{0.461} \\ 
    \bottomrule
  \end{tabular}
  \vspace{-6mm}

  \label{tab:baseline}
\end{table}

\subsection{Quantitative Results}


We quantitatively compare our method with the baseline approaches in Tab.~\ref{tab:baseline}. 
While the baseline method by Burgert \textit{et al.}~\cite{Burgert2023DiffusionIH} achieves a higher aesthetic score, its visual results often produce evident artifacts. 
To support this observation, we conduct a user study (Tab.~\ref{tab:userstudy} and Fig.~\ref{fig:userstudyfig}) and provide qualitative comparisons (Fig.~\ref{fig:otherbaselines}). 
Beyond the aesthetic score, our method achieves superior performance compared to all baselines across the evaluation metrics.

\begin{table*}[t]
\vspace{-6mm}
  \centering
   \small
   \setlength{\tabcolsep}{3.1pt}
   \captionsetup{skip=3pt} 
   \caption{\textbf{Ablation on various design choices on CLIP sore, Aesthetic score, Alignment score and Concealment score.} Our method achieves the best among all the matrices.}
\vspace{-1mm}
  \begin{tabular}{@{}lccc|ccccc@{}}

    \toprule
    Method  & Patch denoising & Camera jitter  & Resolution scaling & CLIP  $\uparrow$ & Aes $\uparrow$ & Aes artifact $\downarrow$ & $\mathcal{A} \uparrow$  & $\mathcal{C} \uparrow$  \\ 
    \midrule
    Baseline  & \textcolor{red}{$\times$} & \textcolor{red}{$\times$} & \textcolor{red}{$\times$} & 0.184 & 2.526 & 3.481 & 0.143 & 0.195 \\ 
    Ablation-A \tiny{(random patch denoising)}  & \textcolor{green}{\checkmark} & \textcolor{red}{$\times$} & \textcolor{red}{$\times$} & 0.128 & 2.700 &  3.784 & 0.083 & 0.172 \\
    Ablation-B \tiny{(w/o resolution scaling, random camera jitter)} & \textcolor{green}{\checkmark} & random & \textcolor{red}{$\times$} & 0.163 & 2.682 &  3.594 & 0.125 & 0.169 \\
    Ablation-C \tiny{(w/o resolution scaling, scheduled camera jitter)} & \textcolor{green}{\checkmark} & scheduled & \textcolor{red}{$\times$} & 0.155 & 2.520 &  3.693 & 0.119 & 0.193 \\
    Ablation-D \tiny{(w/o camera jitter)} & \textcolor{green}{\checkmark} & \textcolor{red}{$\times$} & \textcolor{green}{\checkmark} & 0.165 & 2.663 &  3.695 & 0.129 & 0.159 \\
    Ours        & \textcolor{green}{\checkmark} & scheduled & \textcolor{green}{\checkmark} & \textbf{0.288} &  \textbf{2.791} & \textbf{3.247} & \textbf{0.251} & \textbf{0.461} \\ 
    \bottomrule
  \end{tabular}
  \centering
  \vspace{-3mm}

  \label{tab:designchoice}
\end{table*}

\begin{table*}[h]
\centering
\small
\setlength{\tabcolsep}{8.5pt}
\caption{\textbf{Percentage comparison of visual appeal and prompt alignment for different ablations.} Numbers indicate the percentage of users who preferred a given method over both alternatives in the user study.}
   \vspace{-3mm}
\begin{tabular}{l|ccccccc}
\toprule
 & Burgert \textit{et al.} ~\cite{Burgert2023DiffusionIH} & Baseline & Ablation-A & Ablation-B & Ablation-C & Ablation-D & Ours \\ 
\midrule
Visually Appealing? & 10.14\% & 5.26\% & 28.15\% & 17.57\% & 29.50\% & 39.80\% & \textbf{60.61\%} \\ 
Aligns with Prompt? & 4.73\% & 2.63\% & 30.37\% & 16.22\% & 28.78\% & 41.84\% & \textbf{63.17\%} \\ 
\bottomrule
\end{tabular}
\vspace{-3mm}

\label{tab:userstudy}
\vspace{-2mm}
\end{table*}
Our user study presents 40 participants with three rendered videos generated from different methods using the same prompts. 
After viewing the results, participants answered two questions: ``\textsl{Which result is the most visually appealing?}'' and ``\textsl{Which result aligns best with the text prompt?}''. 
We compare the result generated by our final method with two other results generated by other approaches. We compute the percentage of participants who preferred our result over the alternatives, normalizing by the number of times each result was presented. We provide additional details in the Supp. Sec.~\ref{sec:userstudy}.

\subsection{Qualitative Results}
We present qualitative comparisons between our method and baseline approaches in Fig.~\ref{fig:otherbaselines}.
Additionally, we analyze the impact of our design choices in Fig.~\ref{fig:comparison}. 
Our method effectively fuses content from different prompts, producing visually compelling results.  
We illustrate the setup of reflective surfaces by placing one and two reflective cylinders or curved mirrors on a plane in Fig.~\ref{fig:reflective-2cylinder} and the Supp. Fig.~\ref{fig:reflective-single}.

\topic{Image conditioning}
By supervising the cylinder view with an RGB image using L2 loss, we can generate a customized illusion, such as ``Finding Waldo" illusion, with a real-world example shown in Fig.~\ref{fig:waldo}. 

\topic{3D shape illusion}
Our approach also enables 3D shape illusions using reflective surfaces. As shown in Fig. \ref{fig:result-3D}, different content is revealed when we view the reflection of a complex 3D object on the reflective cylinder beside it.

\subsection{Ablation and Discussion}
\topic{Camera jittering}
Fig.~\ref{fig:camera-jitter} illustrates the smoothing effect of camera jittering.
Naively applying camera jittering can lead to repetitive patterns when combined with the patch denoising method.

\topic{Design choices}
Fig.~\ref{fig:comparison} and Supp. Fig.~\ref{fig:method-supp} illustrate that progressive resolution scaling, combined with scheduled camera jittering, effectively mitigates the repetitive patterns that arise from the naive application of random patch denoising and camera jittering.

\topic{More overlapped region}
To evaluate the performance in a more challenging setting, we increase the overlap between camera views. Fig.~\ref{fig:shrink-view} compares this setups with our orthogonal setting in the sphere case. The results demonstrate how our method effectively enforces seamless fusion between views.

\topic{More views}
We place cameras at all corners of a cube and present an 8-view 3D illusion in Supp. Fig.~\ref{fig:8prompts}. 

\topic{Failure cases}
\label{failure_case}
Rows 1 and 2 of Fig.~\ref{fig:result-failurecase} illustrate failure cases of our method. In row 1, we observe slight duplicate pattern artifacts in column 3, while row 2 shows a blending of image content. Nonetheless, our method substantially reduces the generation of duplicate primary objects compared to baseline methods, as demonstrated both quantitatively (Tab.~\ref{tab:designchoice}, \ref{tab:userstudy}) and qualitatively (Fig.~\ref{fig:comparison}).

\section{Limitations}

Our method significantly reduces duplicate pattern artifacts but does not entirely eliminate them. 
The illusion of an evenly shared surface is inherently ill-posed for the optimization process for each view. 
In the case of a cube, when optimizing two views that share the same area, once there is a principal component in that face that is not connected with the other face, we will have a repetitive pattern (the same content on every single face of a cube). 
We resolved duplicate pattern issues by introducing progressive resolution scaling with scheduled camera jitter and patch sampling.
This makes sense because we are experimenting with a random choice of prompt selections. And the model can still cheat the optimization criteria.



\section{Conclusions}
We introduce a novel approach for creating 3D multiview illusions guided by text prompts. 
Our work uses differentiable rendering and pre-trained diffusion models optimized with VSD loss to generate detailed 3D illusions viewable from different angles.
Key contributions include expanding multiview illusions to complex 3D shapes with detailed textures and introducing scheduled camera viewpoint jittering, patch sampling technique, and progressive resolution scaling, significantly improving quality and consistency.
The camera
viewpoint 
jitter improved the robustness and visual quality between the viewpoints. 
Scheduled patch denoising boosts the resolution of illusion results without introducing artifacts.
Progressive resolution scaling 
ensured more evident, focused illusions.
We also identified challenges in generating illusions for highly overlapping and minimally distorted shapes, suggesting future research directions.
Our work advances multiview illusions by combining text-to-image generation with 3D rendering, creating dynamic visual experiences. Future research can explore 
optimize efficiency and enhance realism, further unlocking the potential of 3D multiview illusions.

\newpage
{
    \small
    \bibliographystyle{ieeenat_fullname}
    \bibliography{main}
}

\clearpage
\setcounter{page}{1}
\maketitlesupplementary
\renewcommand{\thesection}{\Alph{section}}
\setcounter{section}{0}

Our results can be best viewed as videos; please see \href{index.html}{index.html}. We provide more examples and more ablation studies aside from the main paper and supplementary in the website.

\section{Implementation details}
\label{sec:implementation}
We implement the differentiable rendering and reflective render using Pytorch3D ~\cite{ravi2020pytorch3d}. Our method runs on an NVIDIA RTX A6000 GPU with 48GB RAM. We set the training time step for 2000, with timestep annealing to  \( t \sim \mathcal{U}(0.02, 0.5) \) after step 1000. We set the standard deviations in equation (\ref{eq:camera_params}) to 1, and \( C_{\text{max}}\) to 0.3. Adam optimizer ~\cite{kingma2017adammethodstochasticoptimization} was used, with learning rates \( 1 \times 10^{-3} \) for VSD loss and \( 1 \times 10^{-4} \) for the LoRA loss. Each training takes ~2 hours to converge. We use Stable Diffusion v2-1-base ~\cite{Rombach_2022_CVPR} as a pre-trained diffusion model. We use one set of parameters with all experiments (no fine-tuning, or selecting results of the same training process).

\section{Additional results}
\label{sec:addtional}

\begin{figure*}[htbp]
\vspace{-8mm}
    \centering
    \setlength{\tabcolsep}{1pt} 
    \renewcommand{\arraystretch}{1}

    \begin{tabular}{cccccc}
        \begin{minipage}[t]{0.16\textwidth}
        \includegraphics[width=\textwidth, trim=240 240 240 245, clip]{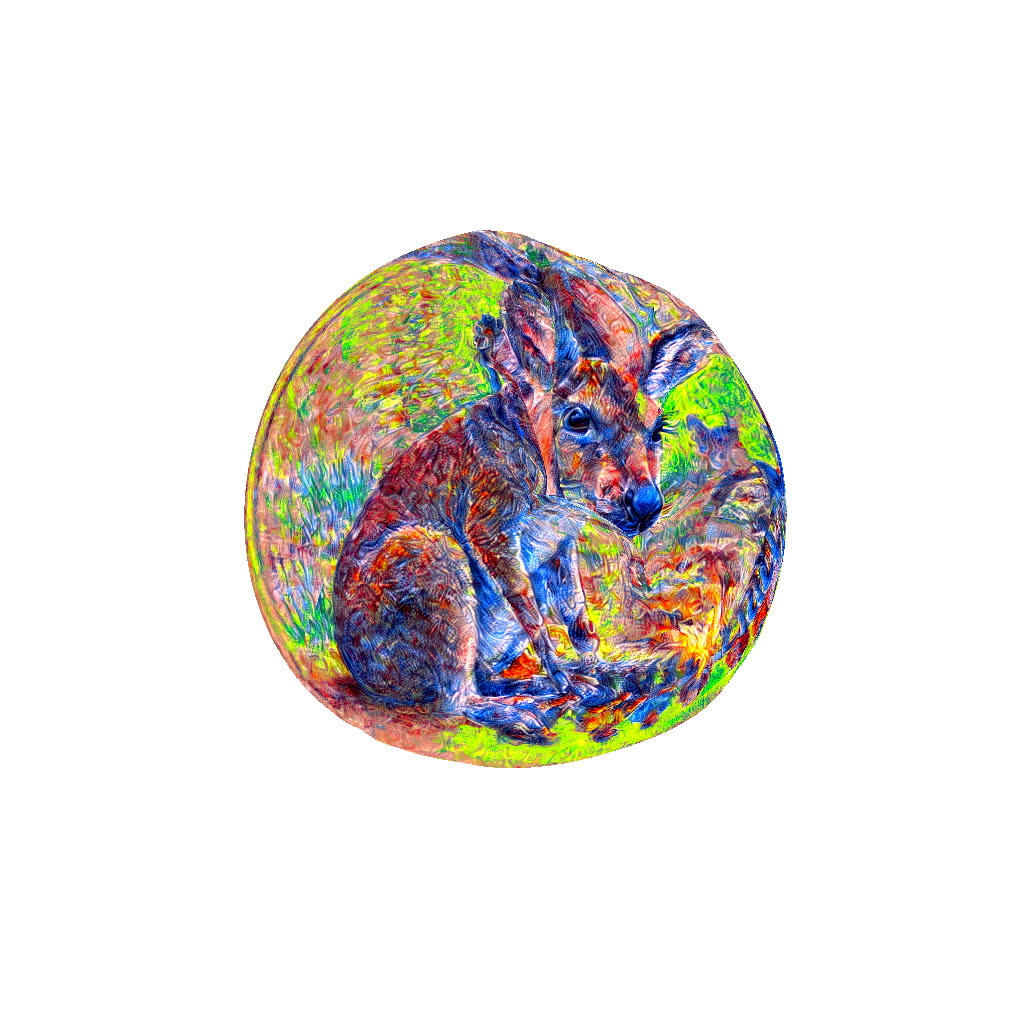}
    \end{minipage}%
    \begin{minipage}[t]{0.16\textwidth}
        \includegraphics[width=\textwidth, trim=240 210 240 220, clip]{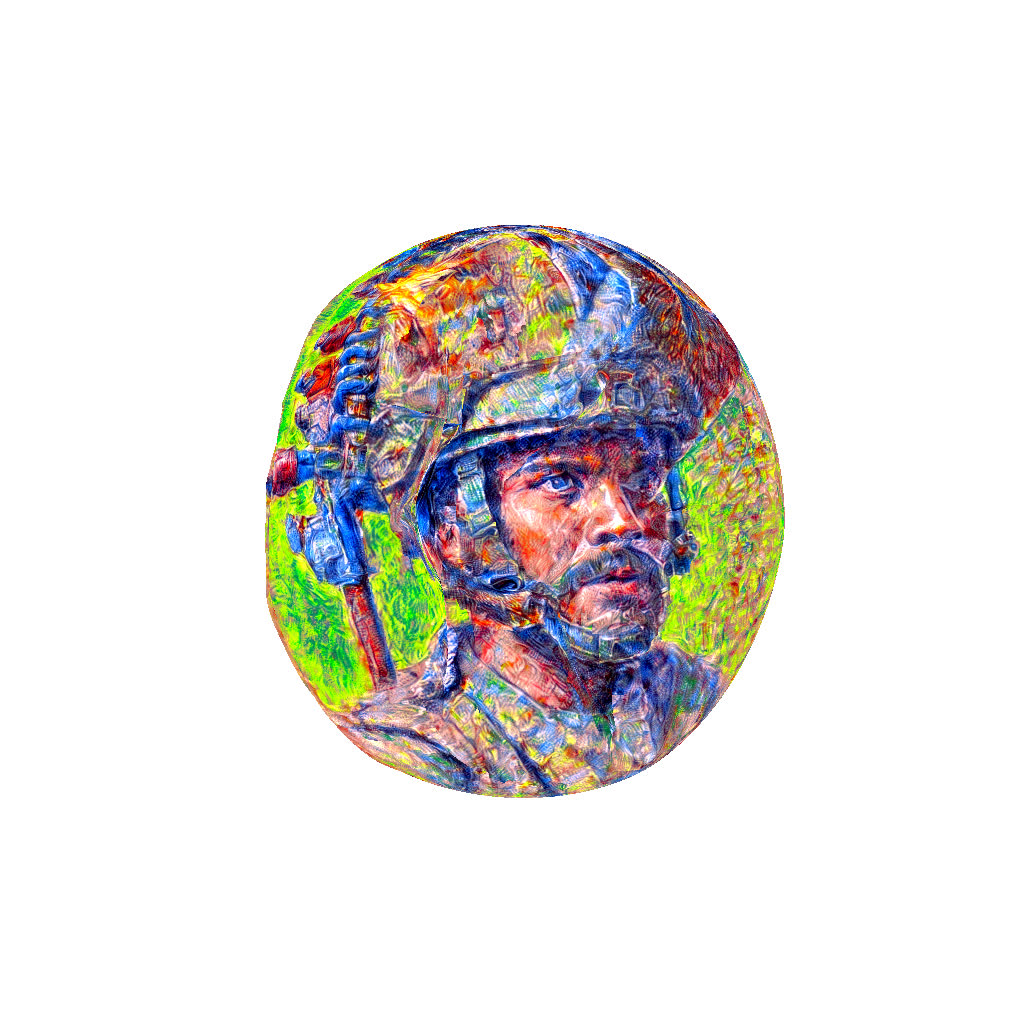}
    \end{minipage}%
    \begin{minipage}[t]{0.16\textwidth}
        \includegraphics[width=\textwidth, trim=220 240 220 240, clip]{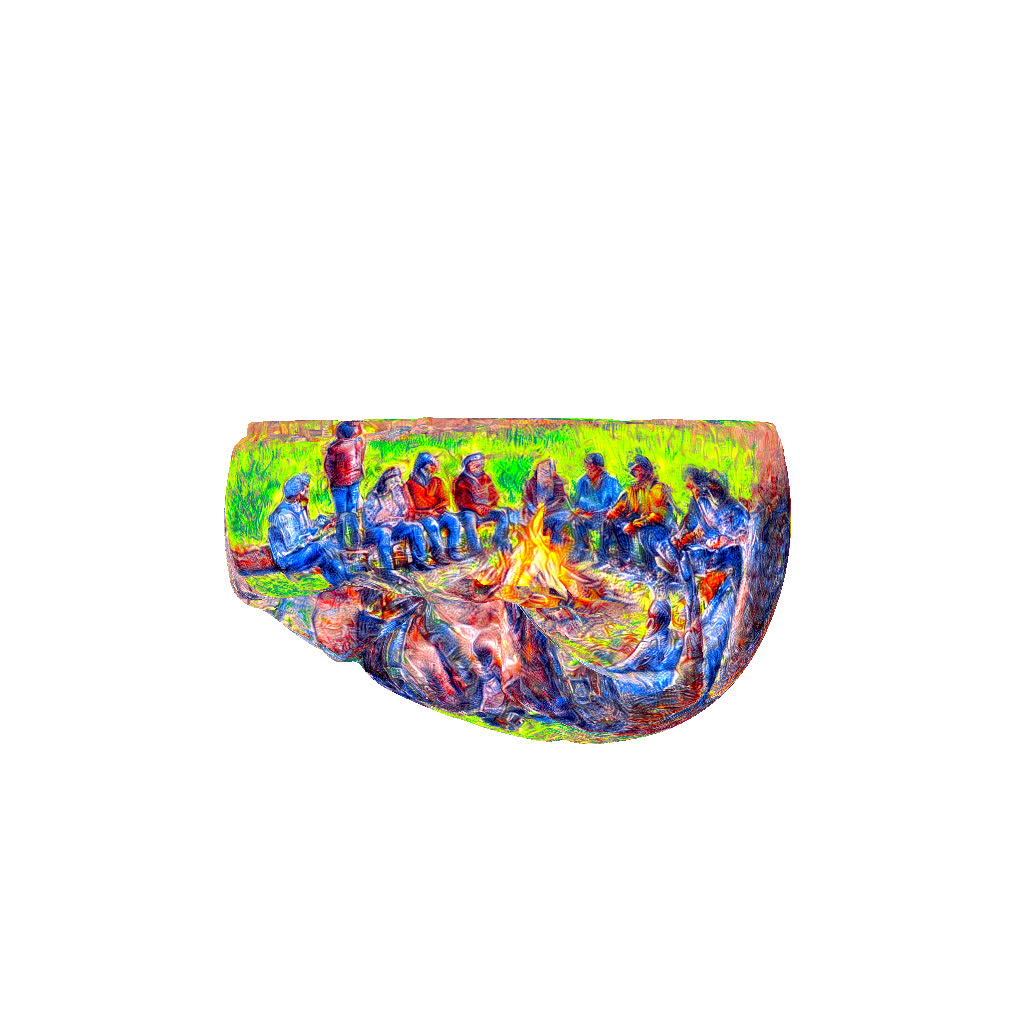}
    \end{minipage}%
    \begin{minipage}[t]{0.16\textwidth}
        \includegraphics[width=\textwidth, trim=240 240 240 245, clip]{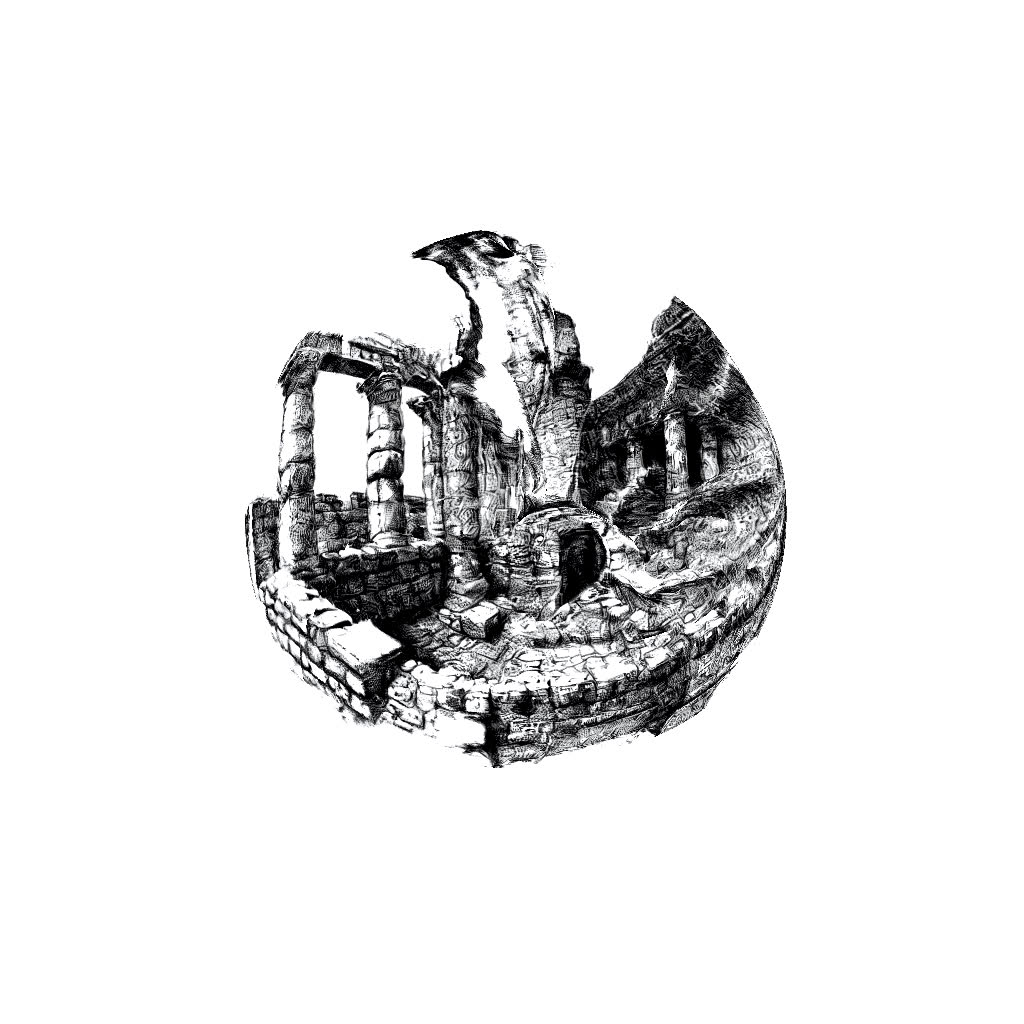}
    \end{minipage}%
    \begin{minipage}[t]{0.16\textwidth}
        \includegraphics[width=\textwidth, trim=240 210 240 220, clip]{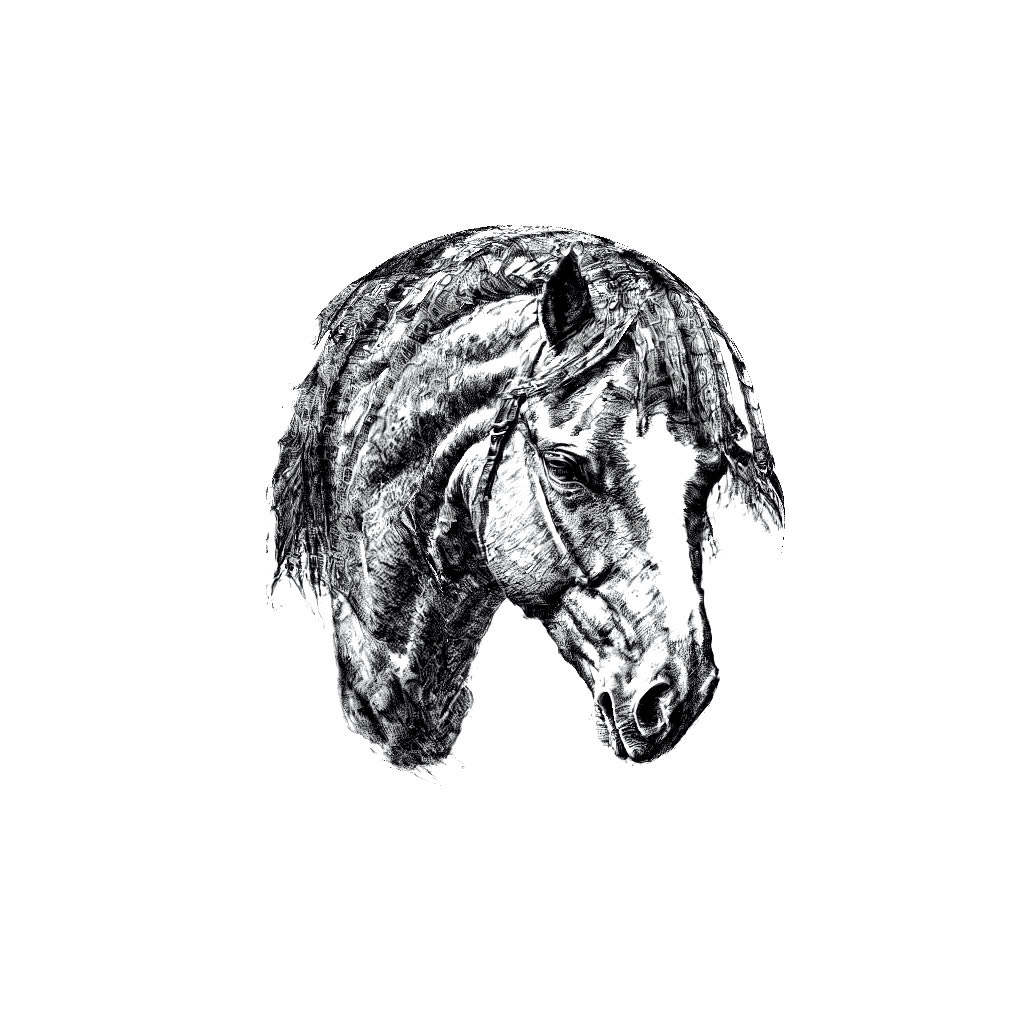}
    \end{minipage}%
    \begin{minipage}[t]{0.16\textwidth}
        \includegraphics[width=\textwidth, trim=220 240 220 240, clip]{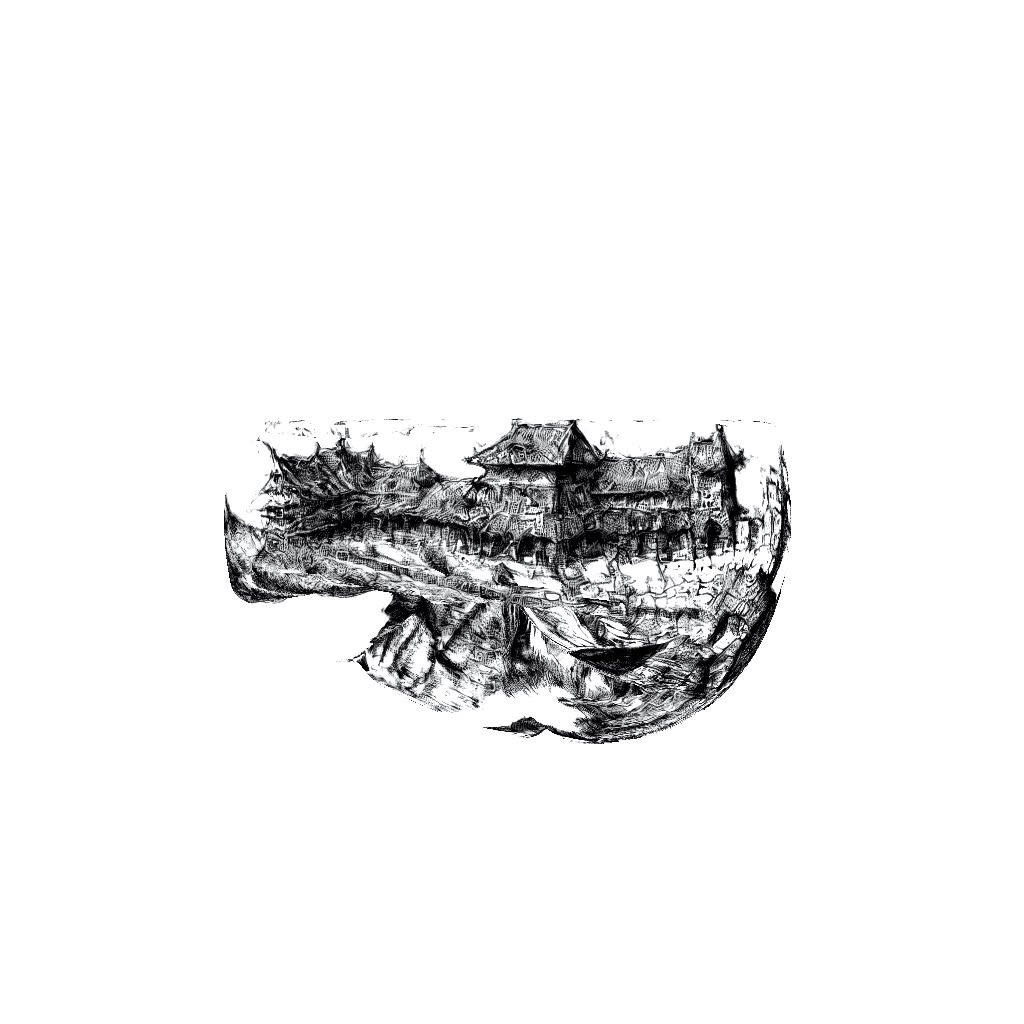}
    \end{minipage}\\

        \begin{minipage}[t]{0.16\textwidth}
        \centering
            \includegraphics[width=\linewidth, trim=185 15 185 30, clip]{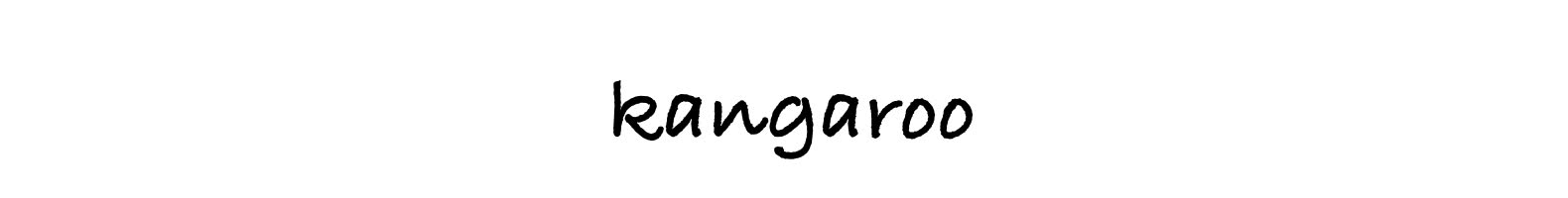}
            \end{minipage}\hfill
        \begin{minipage}[t]{0.16\textwidth}
            \centering
            \includegraphics[width=\linewidth, trim=185 20 185 30, clip]{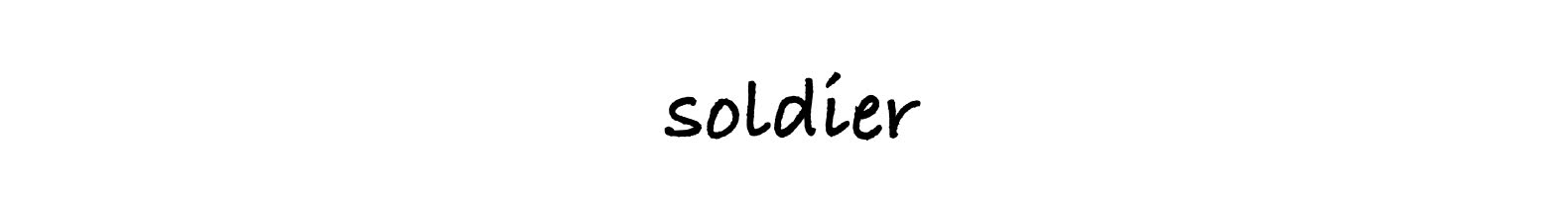}
            \end{minipage}\hfill
        \begin{minipage}[t]{0.16\textwidth}
            \centering
            \includegraphics[width=\linewidth, trim=185 20 185 30, clip]{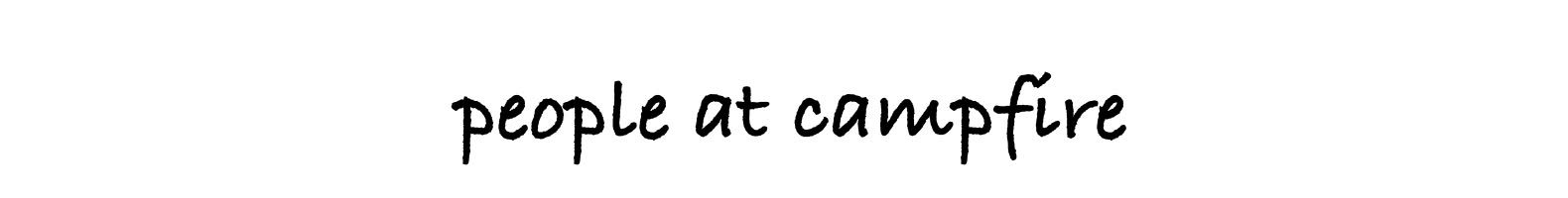}
        \end{minipage}\hfill
        \begin{minipage}[t]{0.16\textwidth}
            \centering
            \includegraphics[width=\linewidth, trim=185 20 185 30, clip]{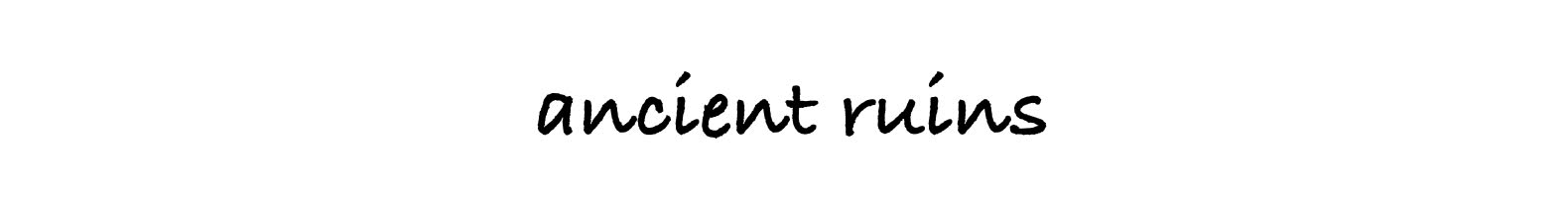}
        \end{minipage}
        \begin{minipage}[t]{0.16\textwidth}
            \centering
            \includegraphics[width=\linewidth, trim=185 20 185 30, clip]{figures/prompt/horse.jpg}
        \end{minipage}
        \begin{minipage}[t]{0.16\textwidth}
            \centering
            \includegraphics[width=\linewidth, trim=185 15 185 30, clip]{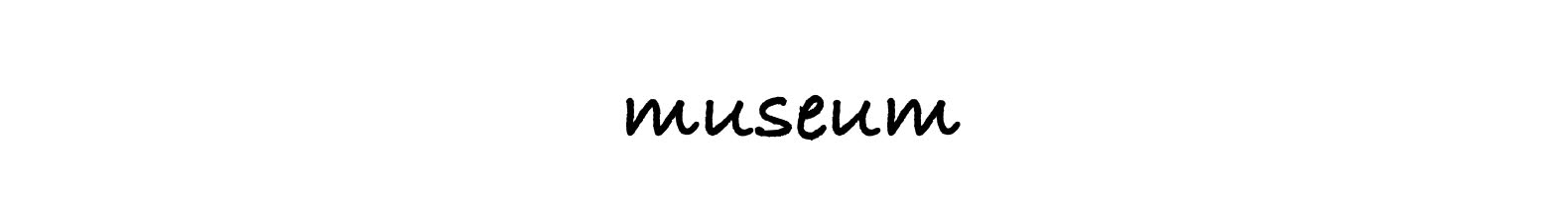}
        \end{minipage}\\    
    \begin{minipage}[t]{0.16\textwidth}
        \includegraphics[width=\textwidth, trim=240 240 240 245, clip]{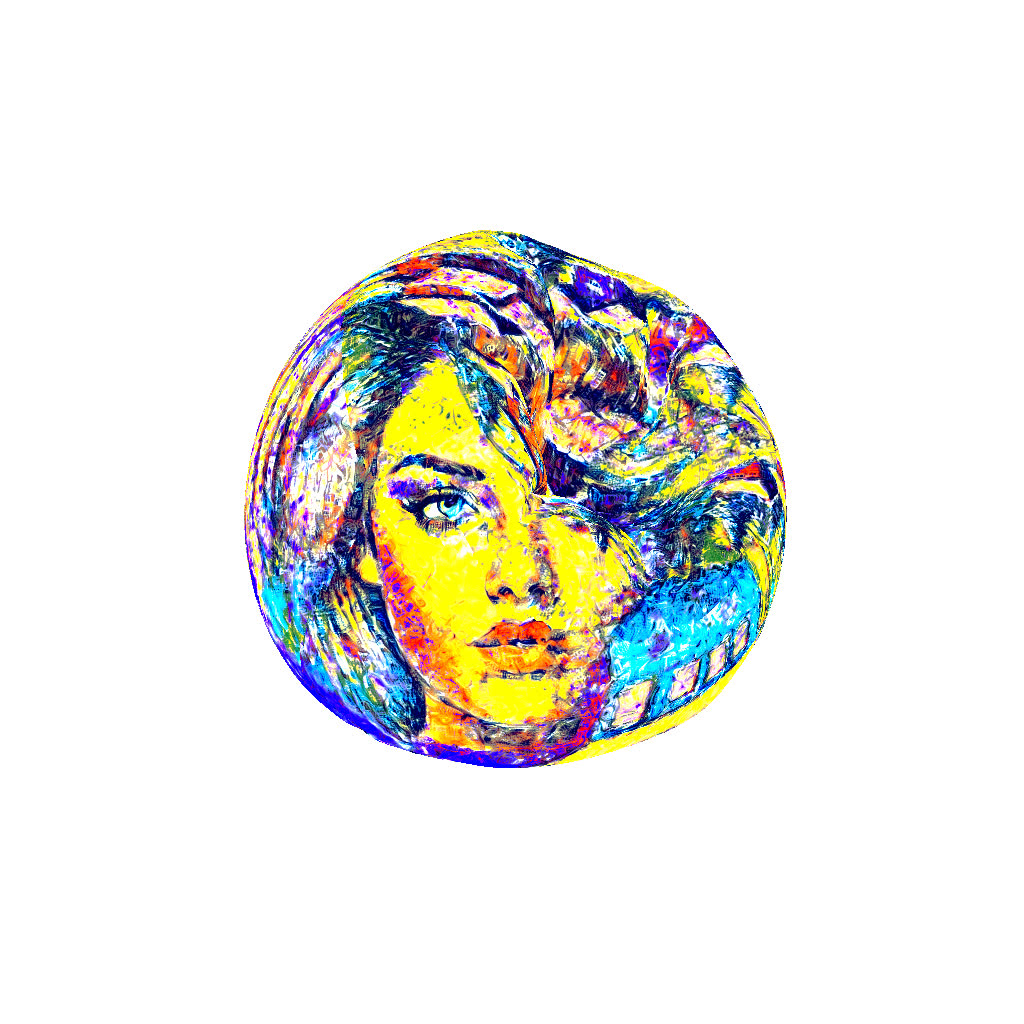}
    \end{minipage}%
    \begin{minipage}[t]{0.16\textwidth}
        \includegraphics[width=\textwidth, trim=240 210 240 220, clip]{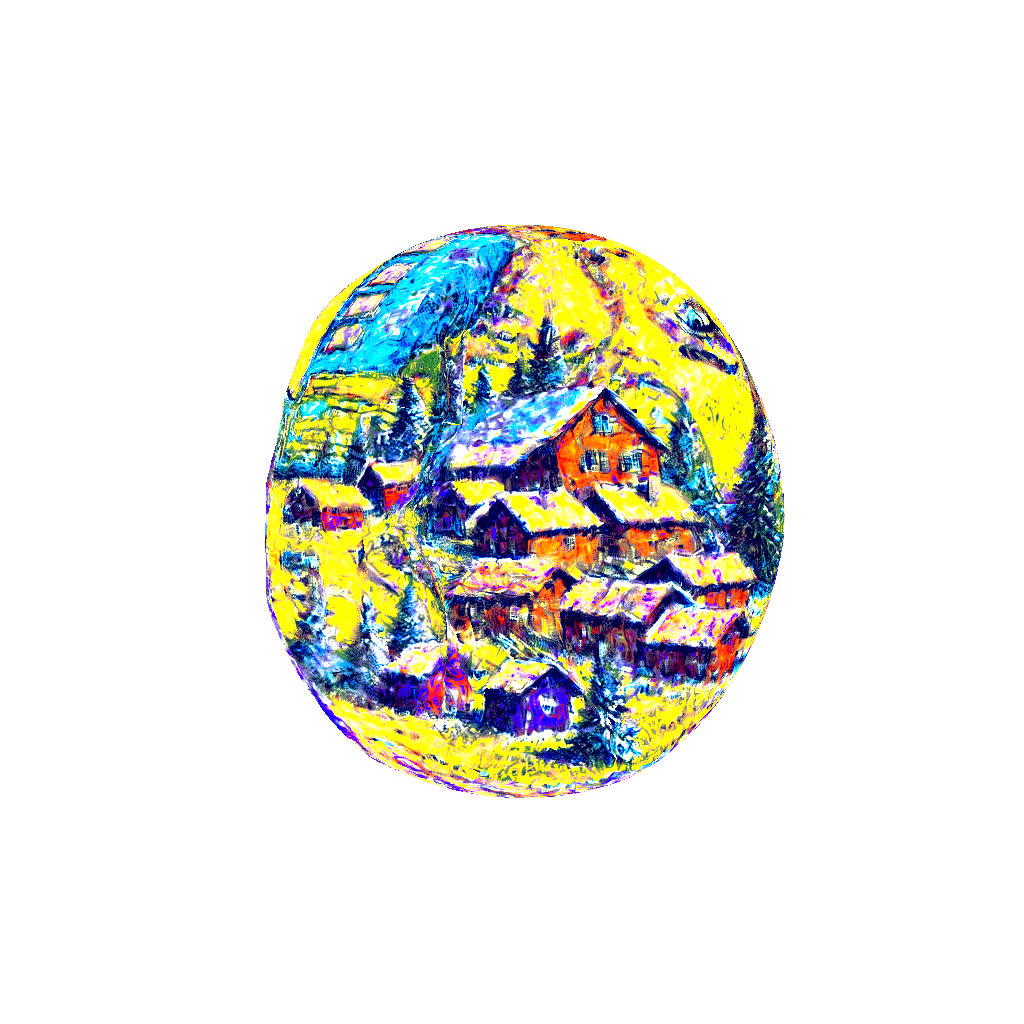}
    \end{minipage}%
    \begin{minipage}[t]{0.16\textwidth}
        \includegraphics[width=\textwidth, trim=220 240 220 240, clip]{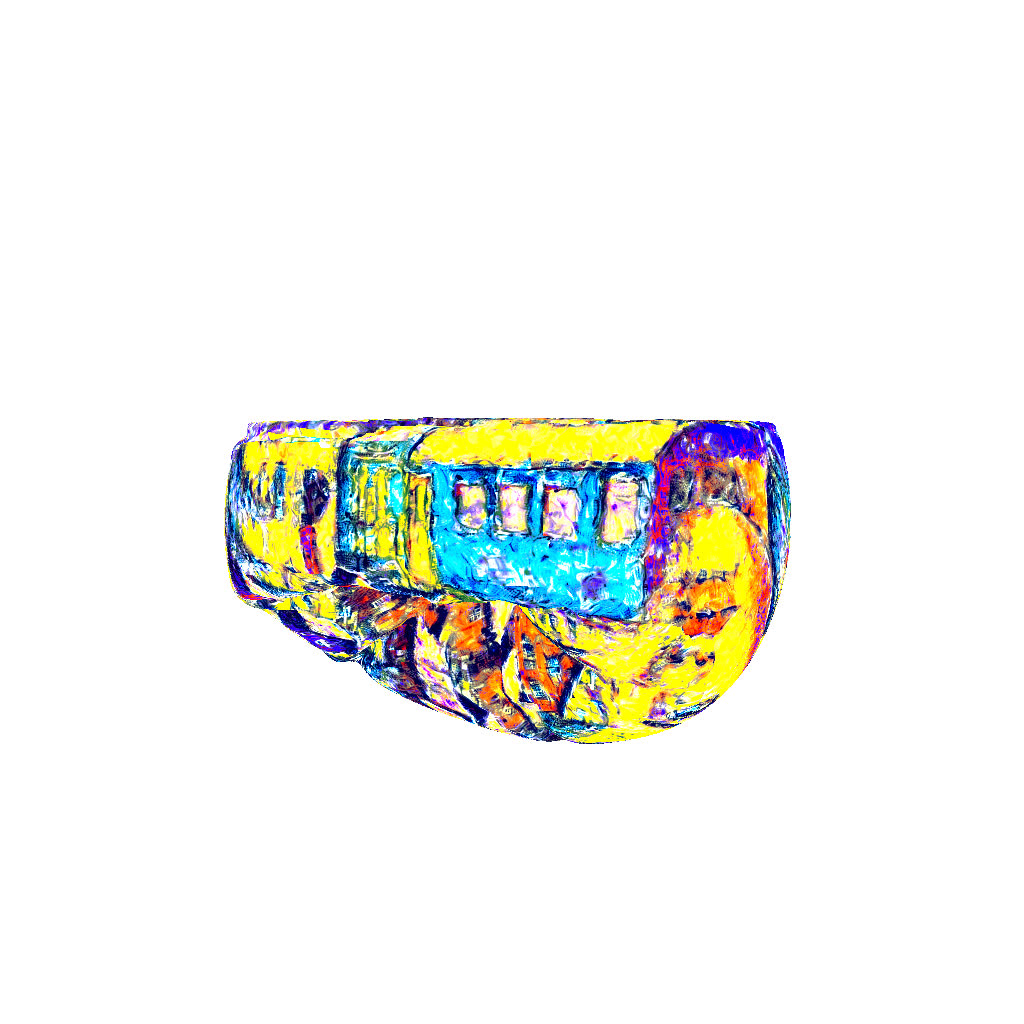}
    \end{minipage}%
    \begin{minipage}[t]{0.16\textwidth}
        \includegraphics[width=\textwidth, trim=240 240 240 245, clip]{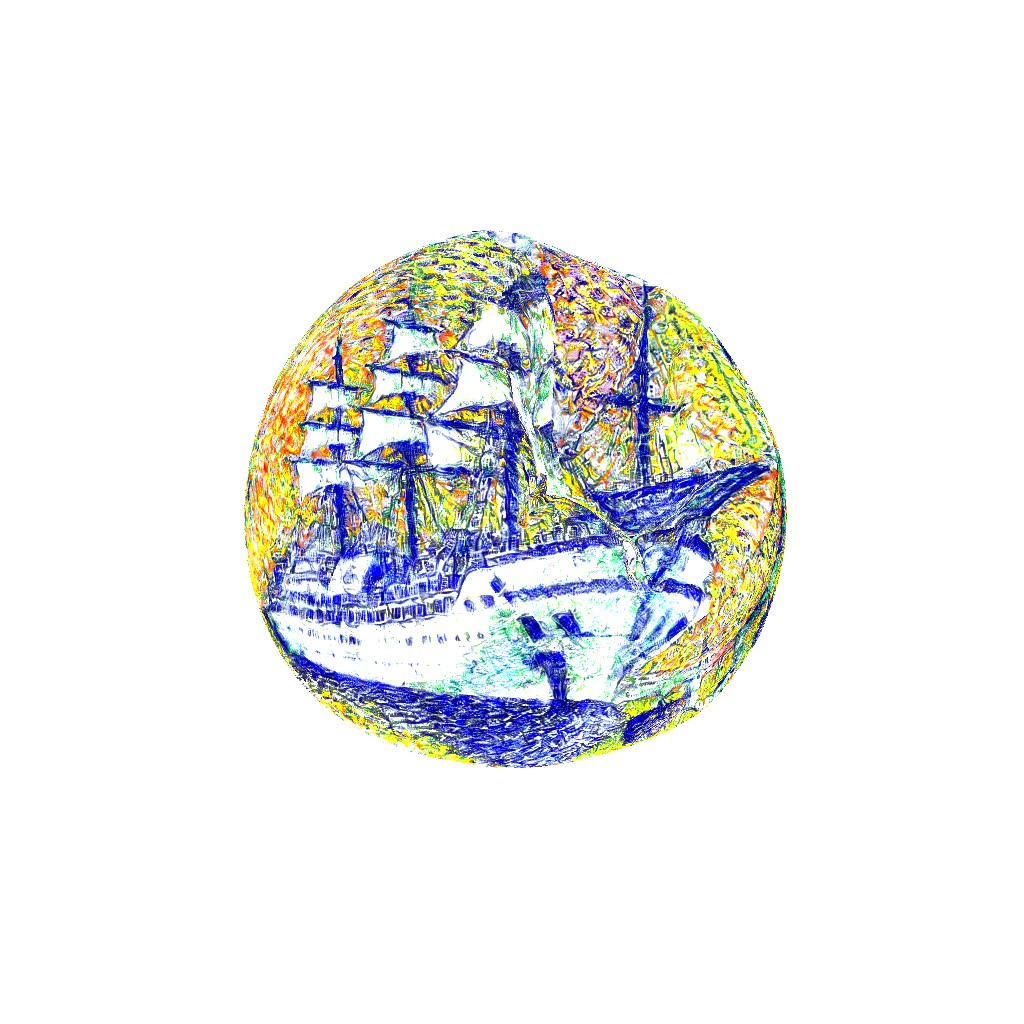}
    \end{minipage}%
    \begin{minipage}[t]{0.16\textwidth}
        \includegraphics[width=\textwidth, trim=240 210 240 220, clip]{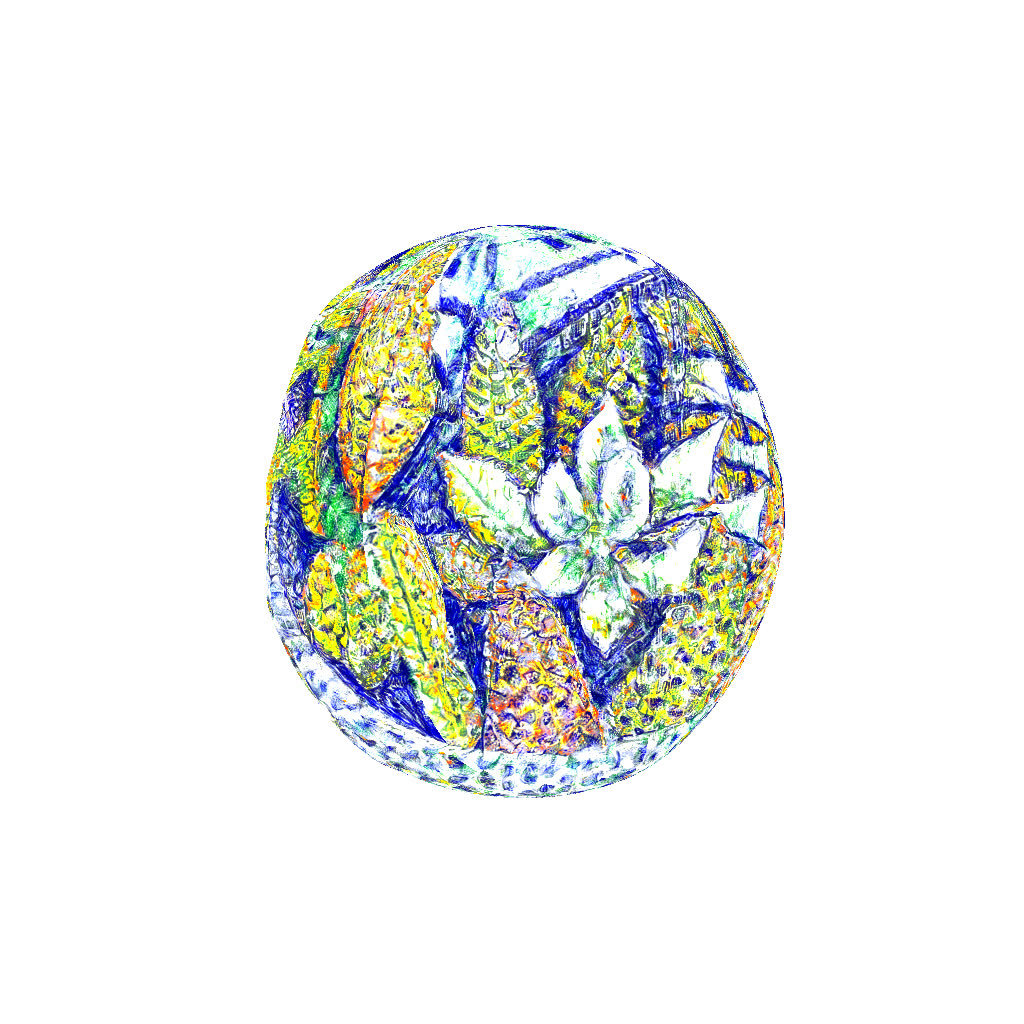}
    \end{minipage}%
    \begin{minipage}[t]{0.16\textwidth}
        \includegraphics[width=\textwidth, trim=220 240 220 240, clip]{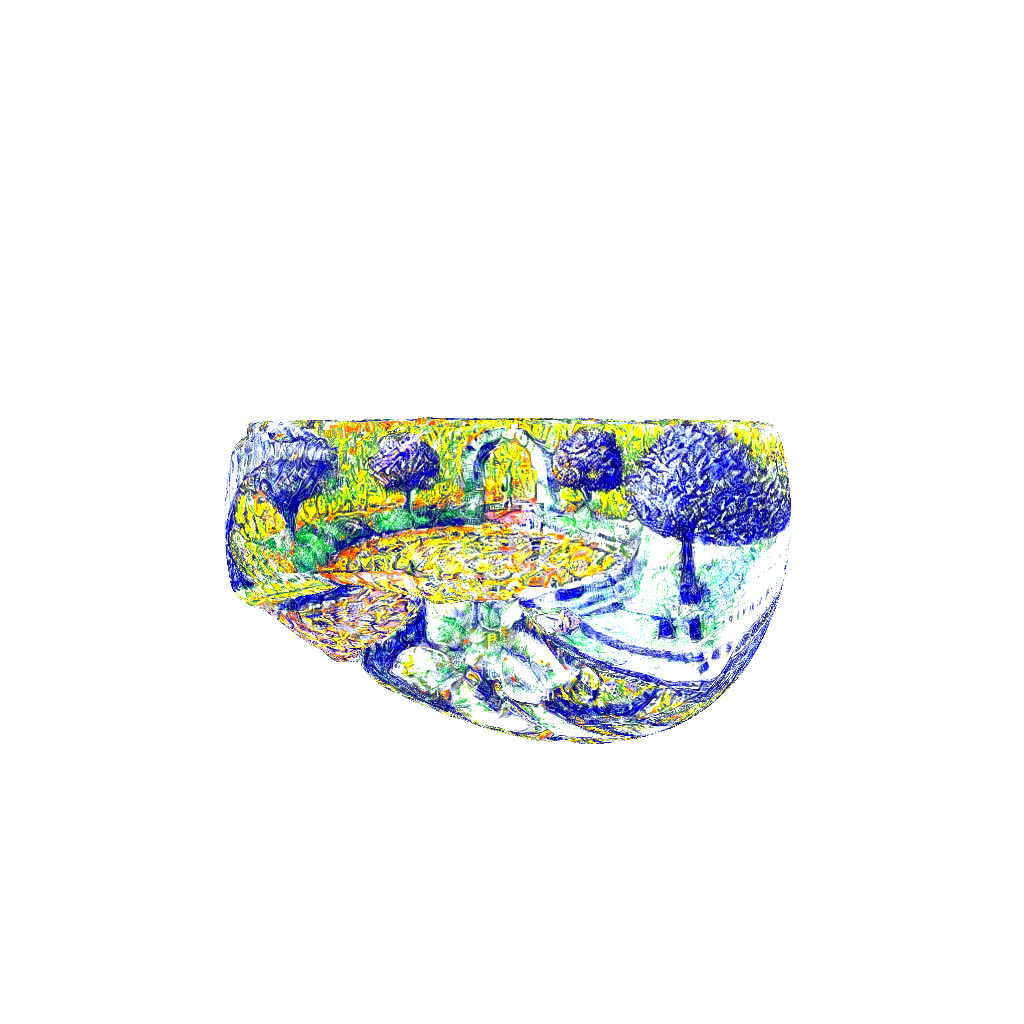}
    \end{minipage}\\

        \begin{minipage}[t]{0.16\textwidth}
        \centering
            \includegraphics[width=\linewidth, trim=185 15 185 30, clip]{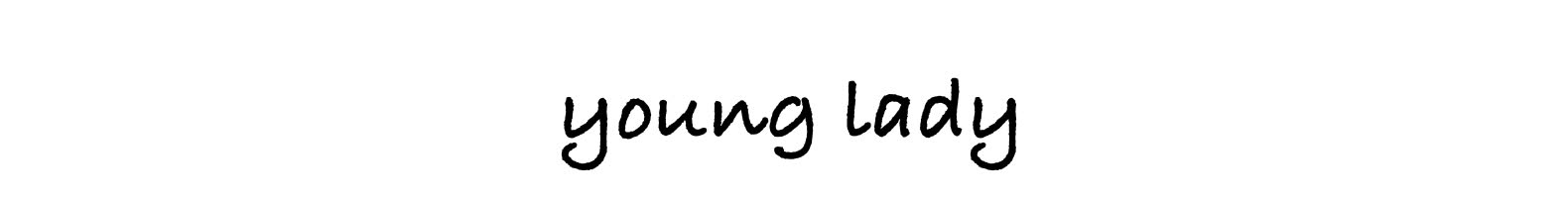}
            \end{minipage}\hfill
        \begin{minipage}[t]{0.16\textwidth}
            \centering
            \includegraphics[width=\linewidth, trim=185 20 185 30, clip]{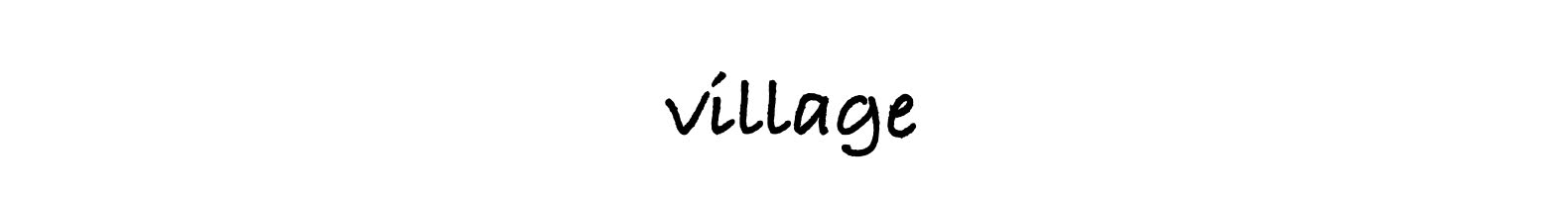}
            \end{minipage}\hfill
        \begin{minipage}[t]{0.16\textwidth}
            \centering
            \includegraphics[width=\linewidth, trim=185 20 185 30, clip]{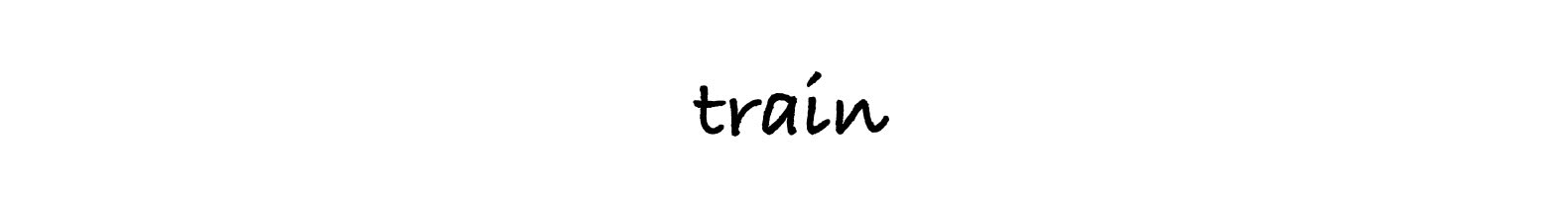}
        \end{minipage}\hfill
        \begin{minipage}[t]{0.16\textwidth}
            \centering
            \includegraphics[width=\linewidth, trim=185 20 185 30, clip]{figures/prompt/boat.jpg}
        \end{minipage}
        \begin{minipage}[t]{0.16\textwidth}
            \centering
            \includegraphics[width=\linewidth, trim=185 20 185 30, clip]{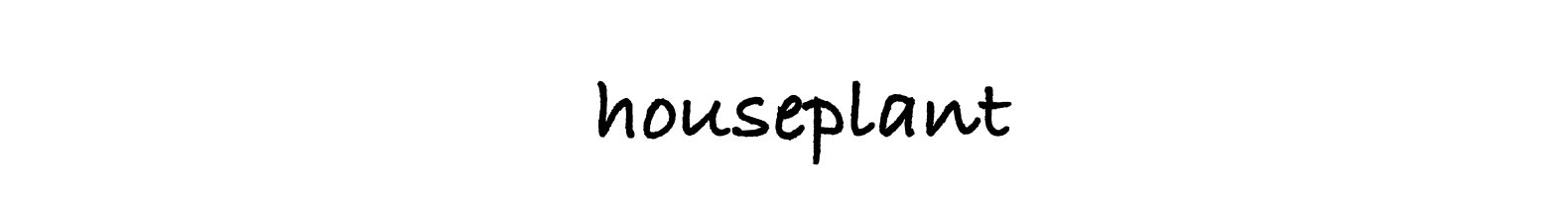}
        \end{minipage}
        \begin{minipage}[t]{0.16\textwidth}
            \centering
            \includegraphics[width=\linewidth, trim=185 15 185 30, clip]{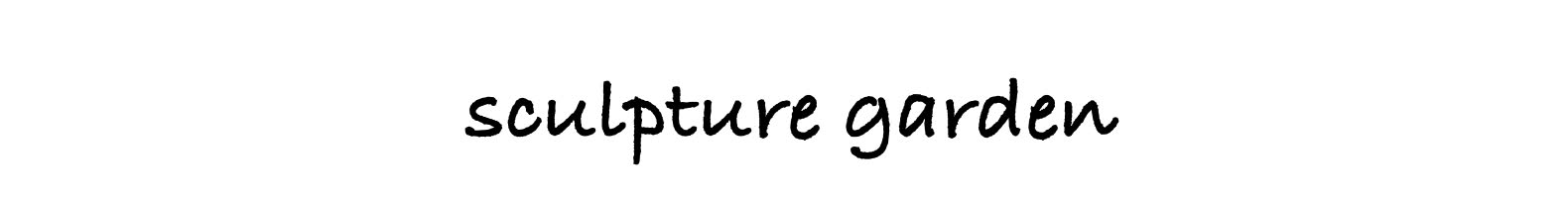}
        \end{minipage}\\

        \begin{minipage}[t]{0.16\textwidth}
            \includegraphics[width=\textwidth, trim=240 240 240 245, clip]{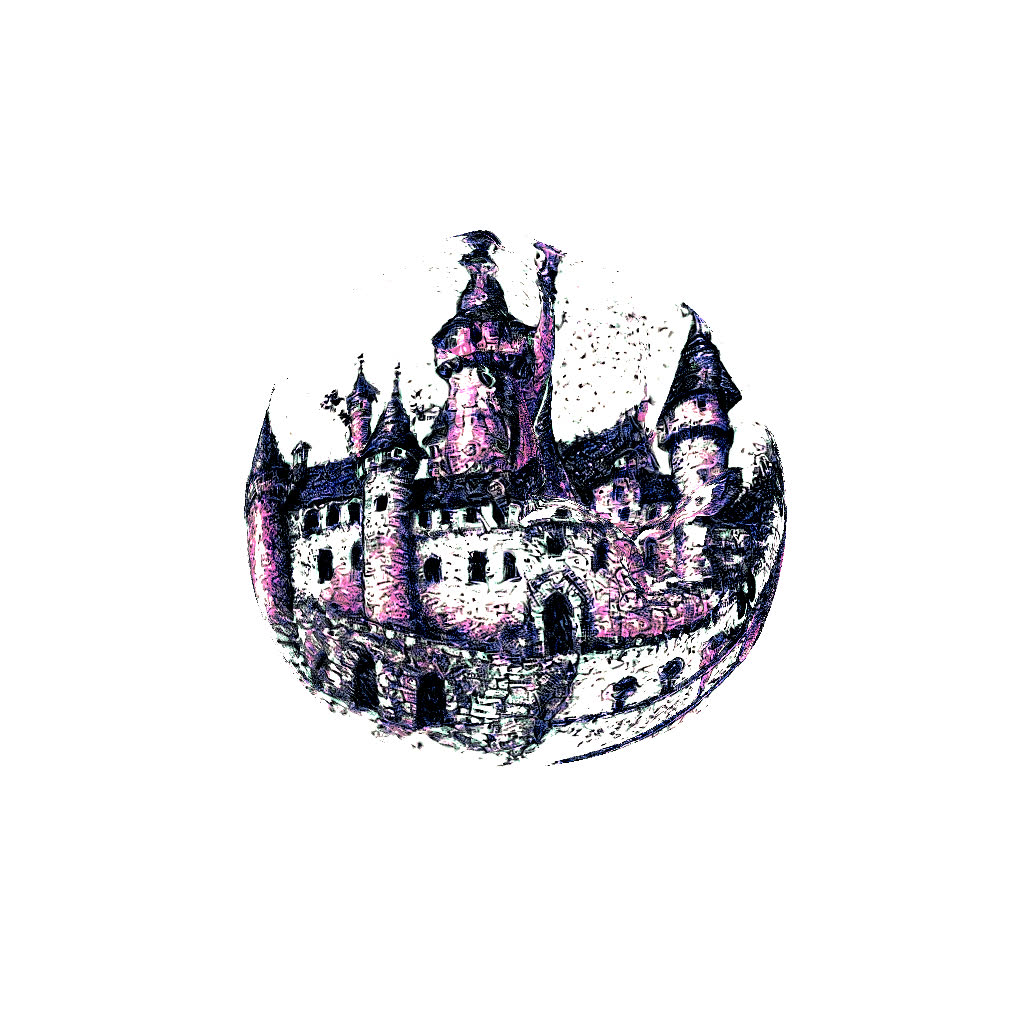}
        \end{minipage}%
        \begin{minipage}[t]{0.16\textwidth}
            \includegraphics[width=\textwidth, trim=240 210 240 220, clip]{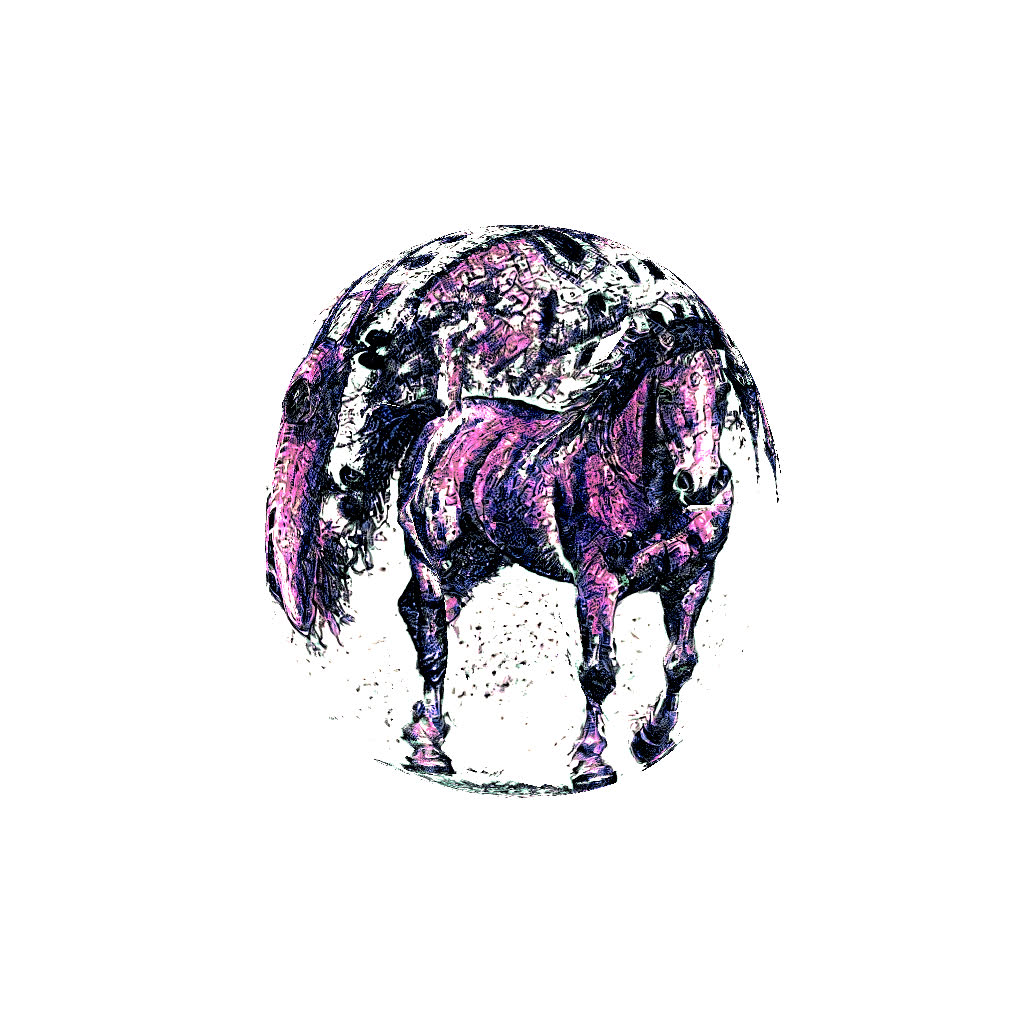}
        \end{minipage}%
        \begin{minipage}[t]{0.16\textwidth}
            \includegraphics[width=\textwidth, trim=220 240 220 240, clip]{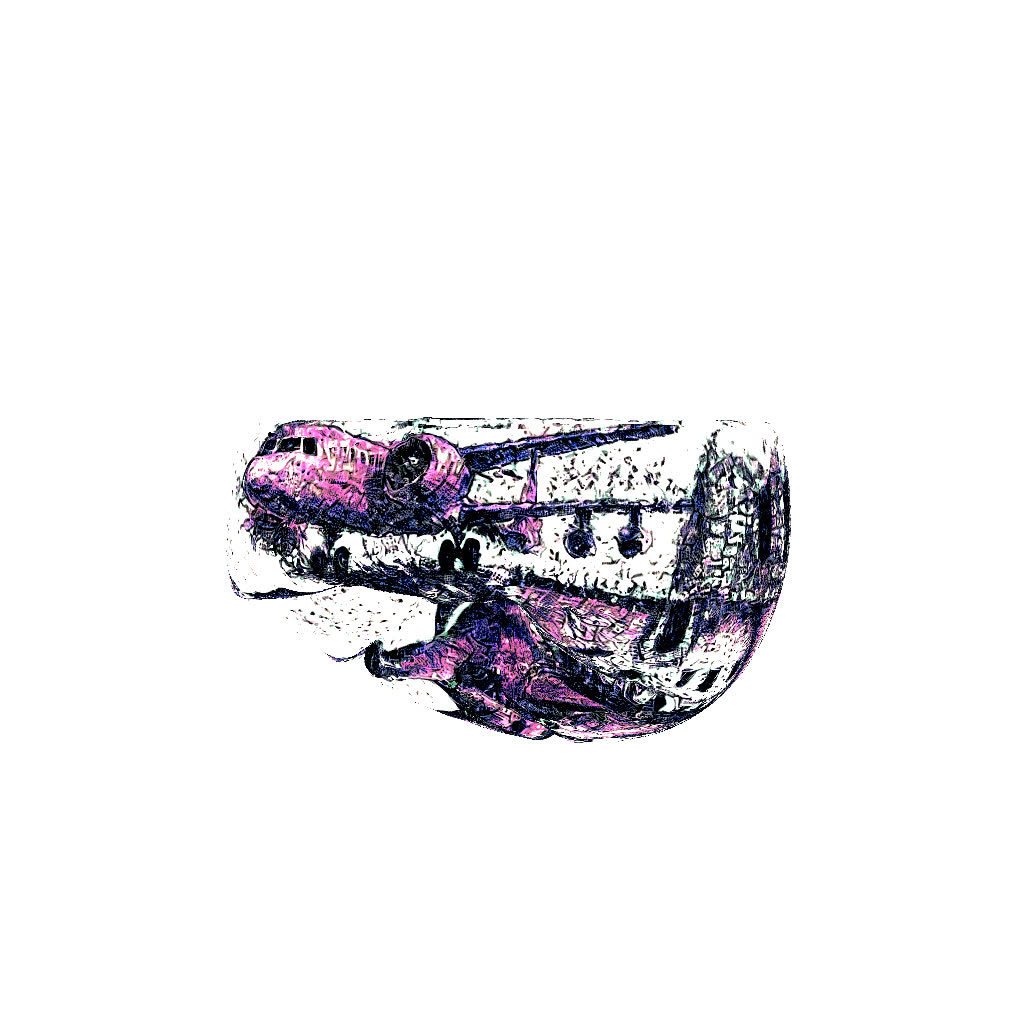}
        \end{minipage}%
        \begin{minipage}[t]{0.16\textwidth}
            \includegraphics[width=\textwidth, trim=240 240 240 245, clip]{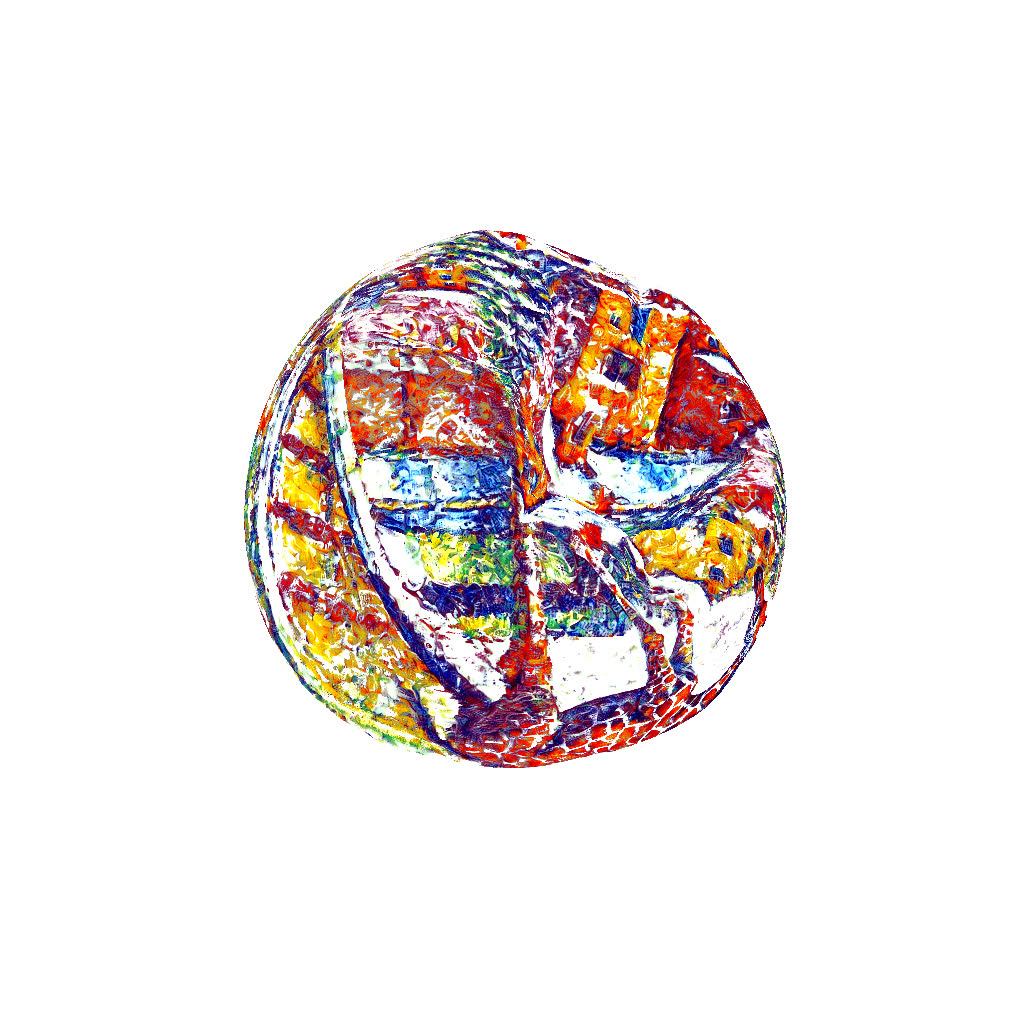}
        \end{minipage}%
        \begin{minipage}[t]{0.16\textwidth}
            \includegraphics[width=\textwidth, trim=240 210 240 220, clip]{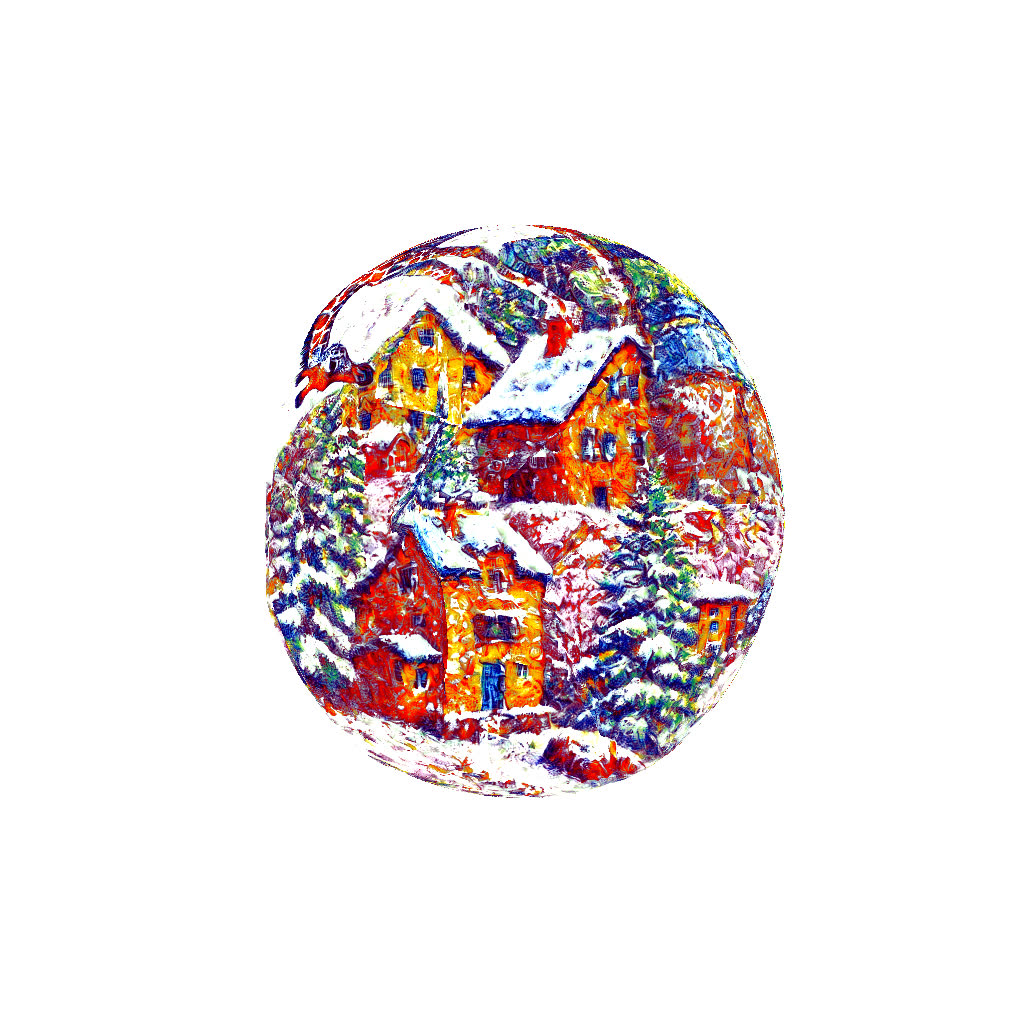}
        \end{minipage}%
        \begin{minipage}[t]{0.16\textwidth}
            \includegraphics[width=\textwidth, trim=220 240 220 240, clip]{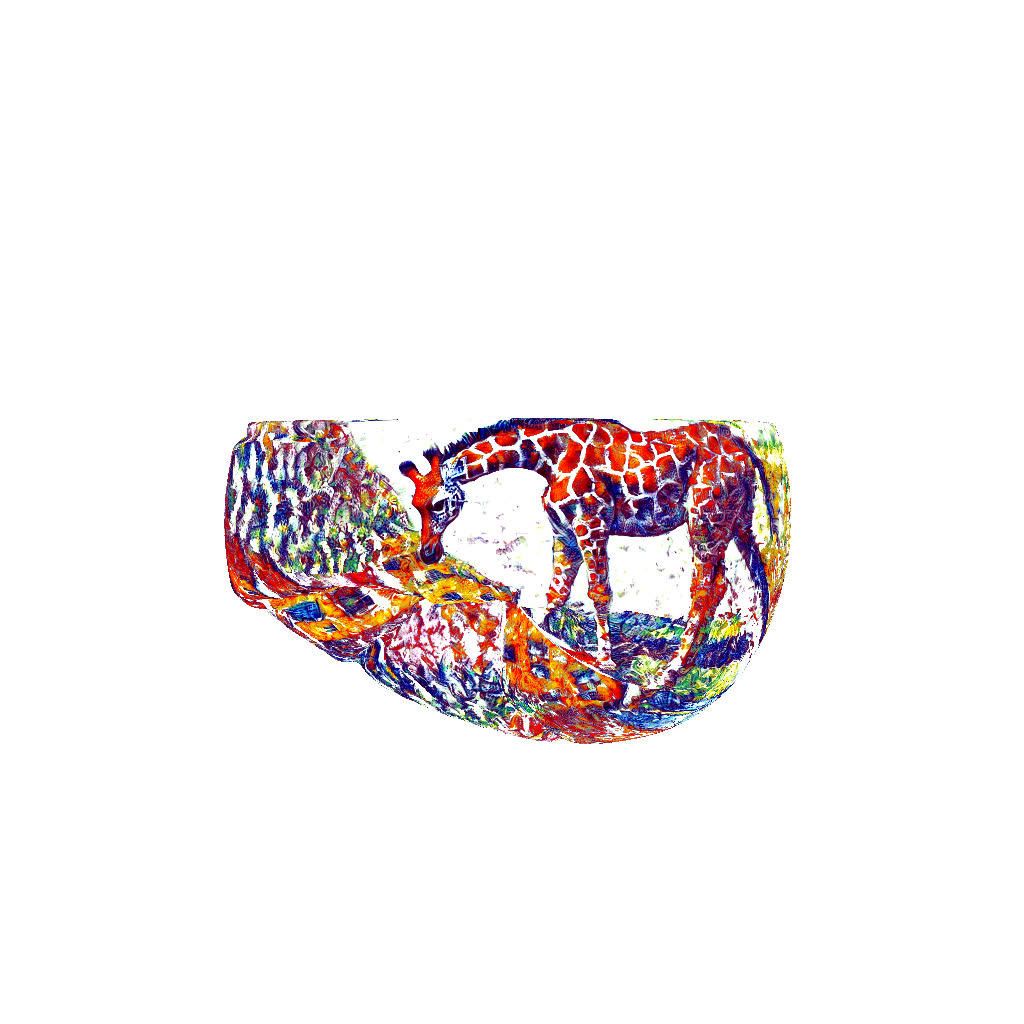}
        \end{minipage}\\

        \begin{minipage}[t]{0.16\textwidth}
        \centering
            \includegraphics[width=\linewidth, trim=185 20 185 30, clip]{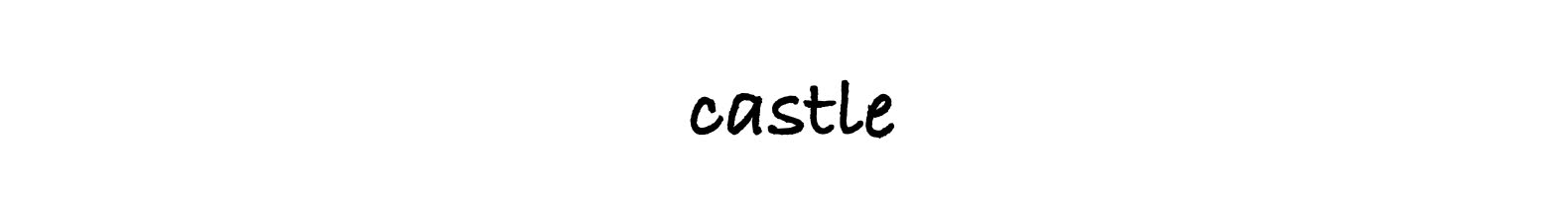}
            \end{minipage}\hfill
        \begin{minipage}[t]{0.16\textwidth}
            \centering
            \includegraphics[width=\linewidth, trim=185 15 185 30, clip]{figures/prompt/horse.jpg}
            \end{minipage}\hfill
        \begin{minipage}[t]{0.16\textwidth}
            \centering
            \includegraphics[width=\linewidth, trim=185 20 185 30, clip]{figures/prompt/train.jpg}
        \end{minipage}\hfill
        \begin{minipage}[t]{0.16\textwidth}
            \centering
            \includegraphics[width=\linewidth, trim=185 20 185 30, clip]{figures/prompt/boat.jpg}
        \end{minipage}
        \begin{minipage}[t]{0.16\textwidth}
            \centering
            \includegraphics[width=\linewidth, trim=185 20 185 30, clip]{figures/prompt/village.jpg}
        \end{minipage}
        \begin{minipage}[t]{0.16\textwidth}
            \centering
            \includegraphics[width=\linewidth, trim=185 20 185 30, clip]{figures/prompt/giraffe.jpg}
        \end{minipage}\\      
    
    \begin{minipage}{0.16\textwidth}
    \includegraphics[width=\linewidth, trim=240 240 240 245, clip]{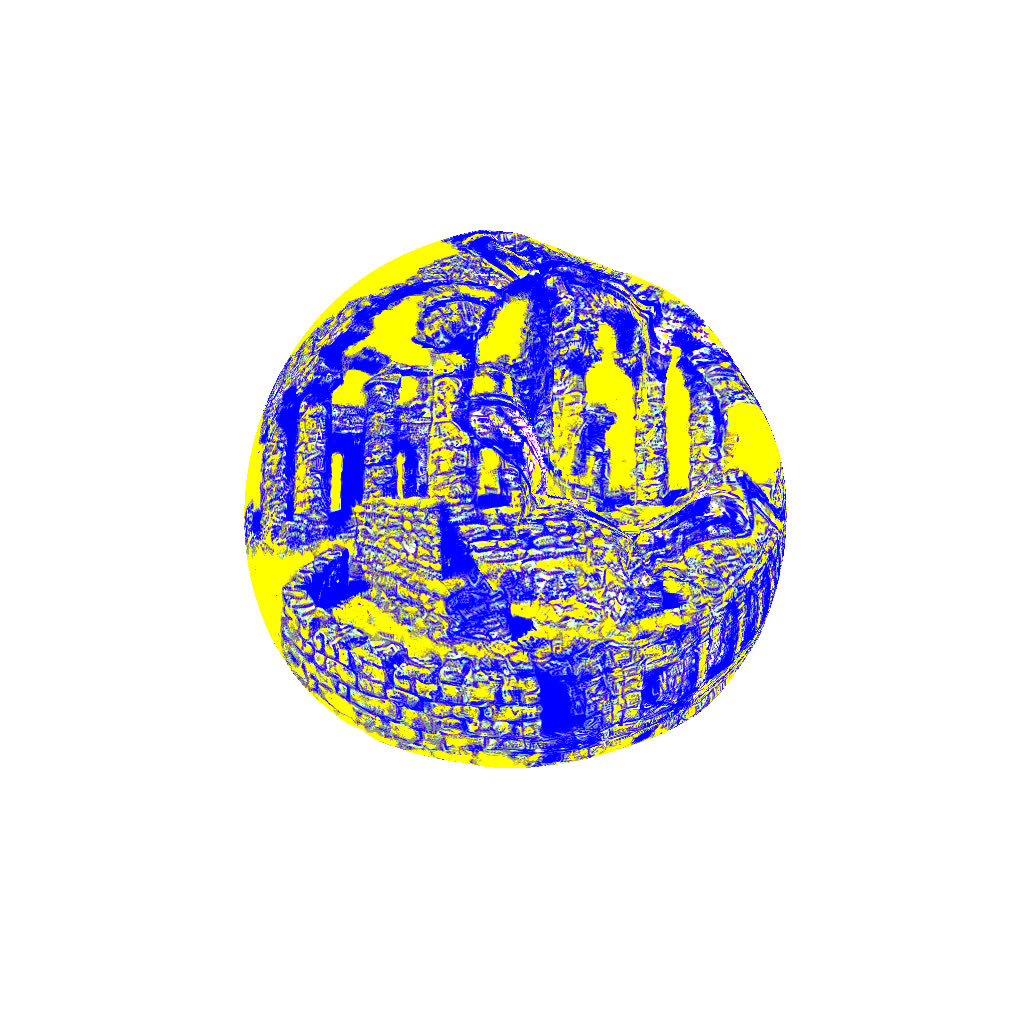}
    \end{minipage}%
    \begin{minipage}{0.16\textwidth}
        \includegraphics[width=\linewidth, trim=240 210 240 220, clip]{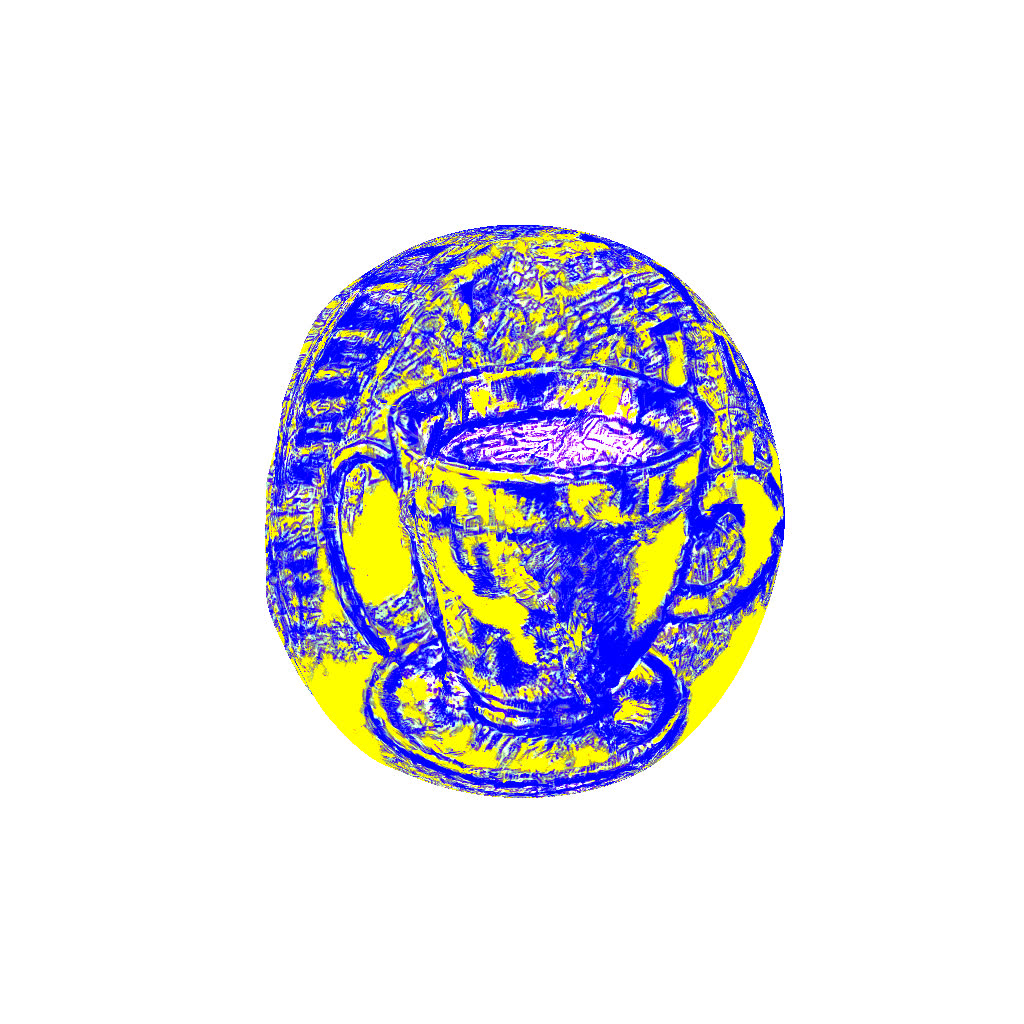}
    \end{minipage}%
    \begin{minipage}{0.16\textwidth}
        \includegraphics[width=\linewidth, trim=220 240 220 240, clip]{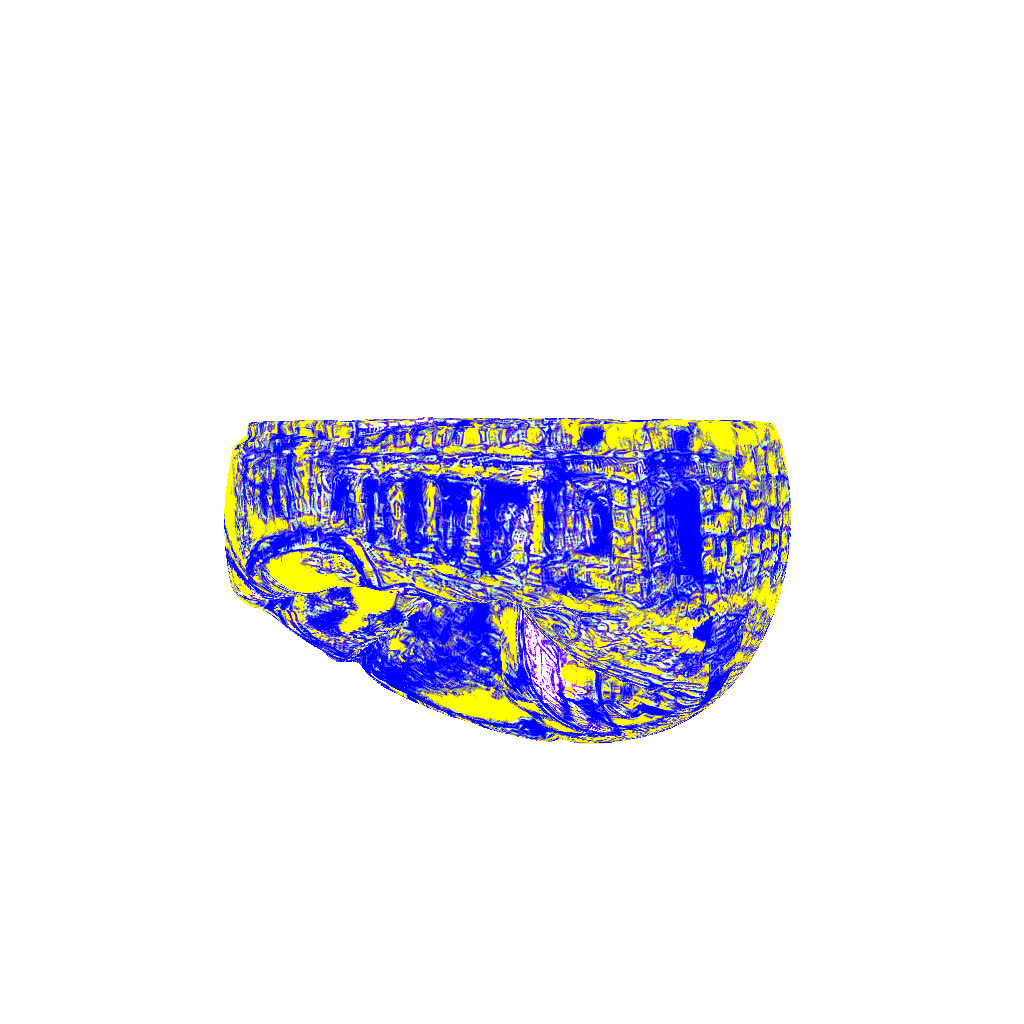}
    \end{minipage}%
    \begin{minipage}{0.16\textwidth}
        \includegraphics[width=\linewidth, trim=240 240 240 245, clip]{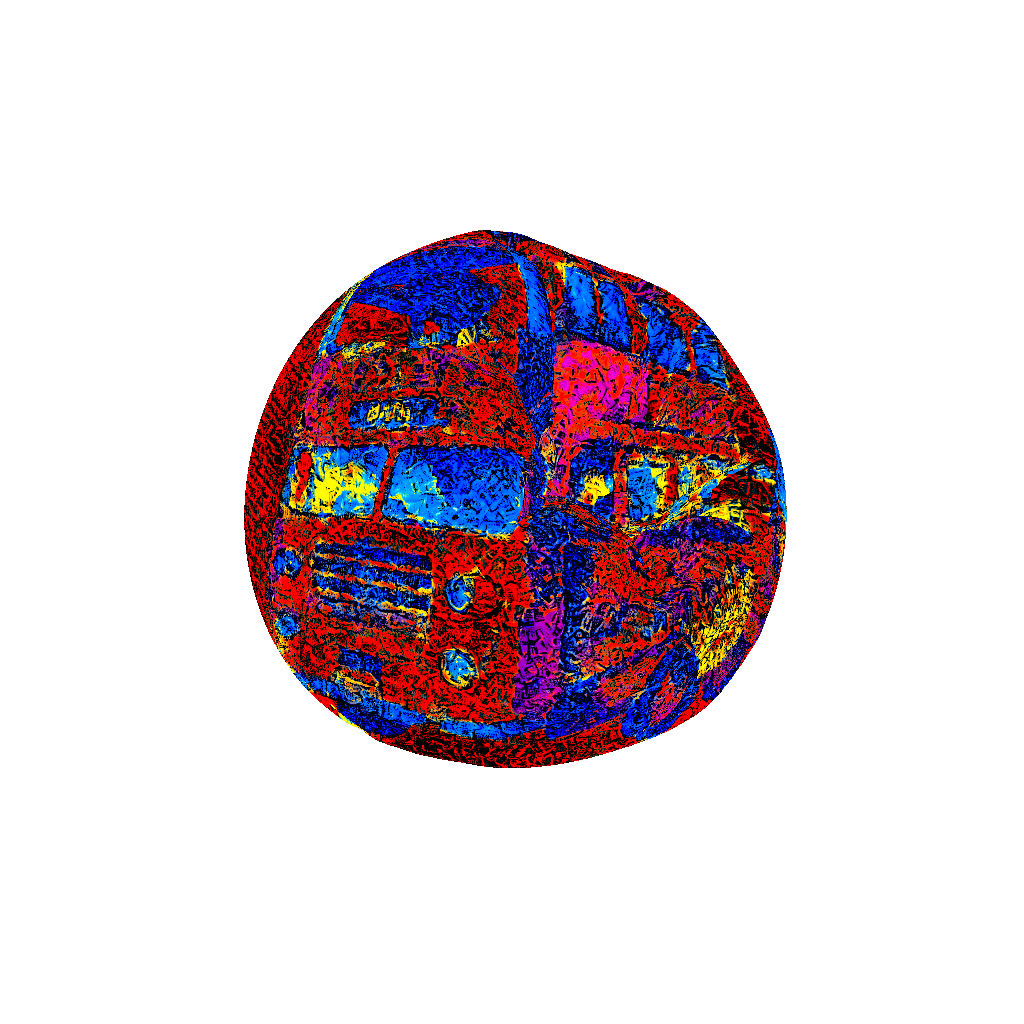}
    \end{minipage}%
    \begin{minipage}{0.16\textwidth}
        \includegraphics[width=\linewidth, trim=240 210 240 220, clip]{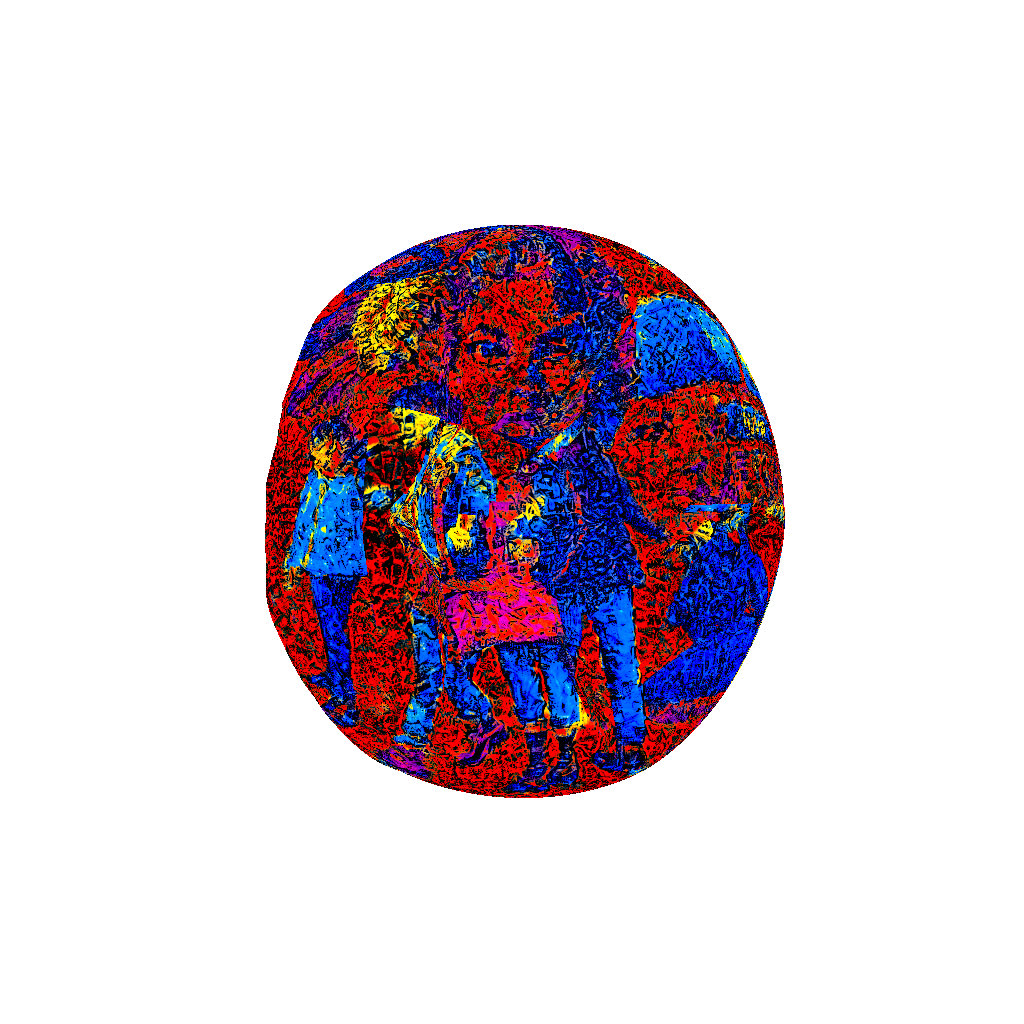}
    \end{minipage}%
    \begin{minipage}{0.16\textwidth}
        \includegraphics[width=\linewidth, trim=220 240 220 240, clip]{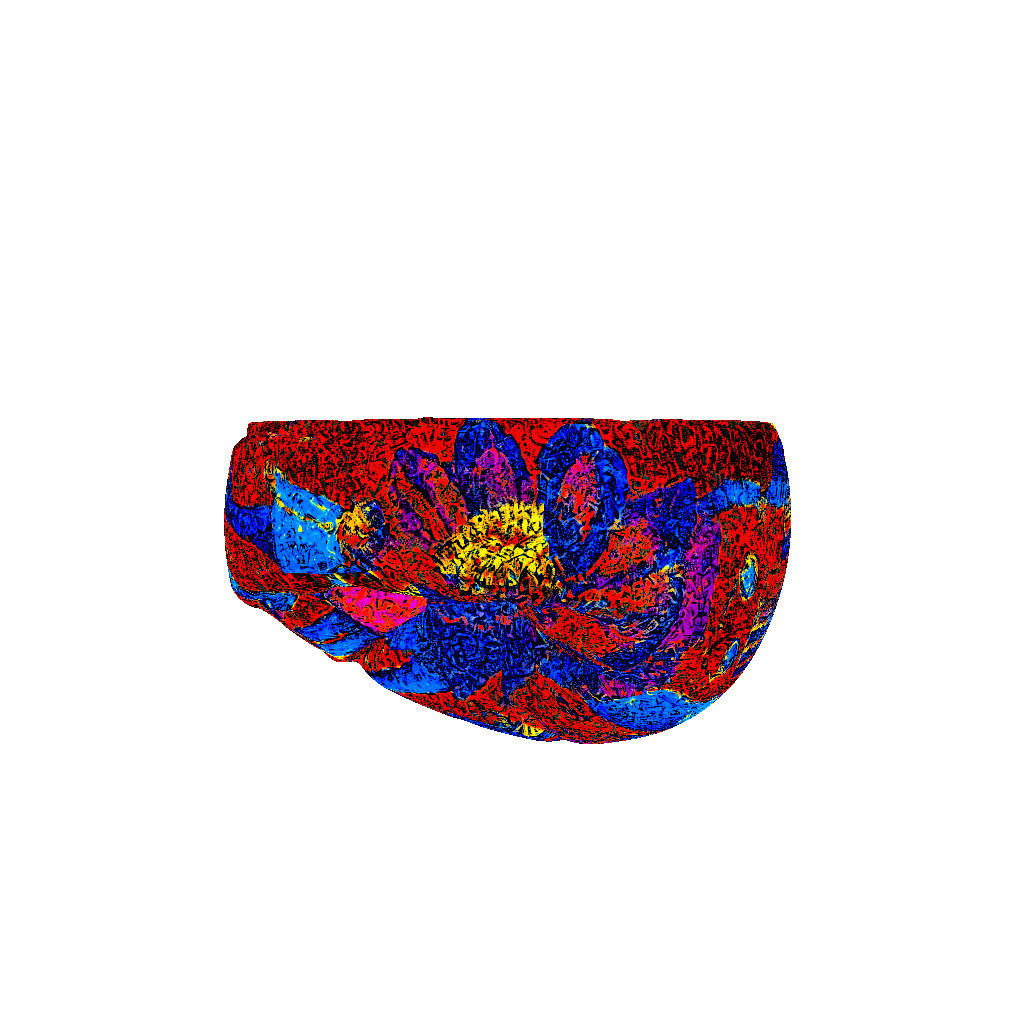}
    \end{minipage} \\

\vspace{1mm}
        \begin{minipage}[t]{0.16\textwidth}
        \centering
            \includegraphics[width=\linewidth, trim=185 20 185 30, clip]{figures/prompt/ar.jpg}
            \end{minipage}\hfill
        \begin{minipage}[t]{0.16\textwidth}
            \centering
            \includegraphics[width=\linewidth, trim=185 20 185 30, clip]{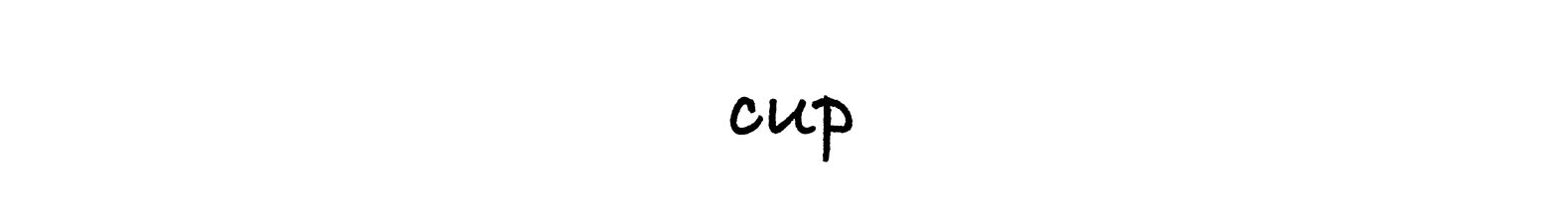}
            \end{minipage}\hfill
        \begin{minipage}[t]{0.16\textwidth}
            \centering
            \includegraphics[width=\linewidth, trim=185 20 185 30, clip]{figures/prompt/museum.jpg}
        \end{minipage}\hfill
        \begin{minipage}[t]{0.16\textwidth}
            \centering
            \includegraphics[width=\linewidth, trim=185 20 185 30, clip]{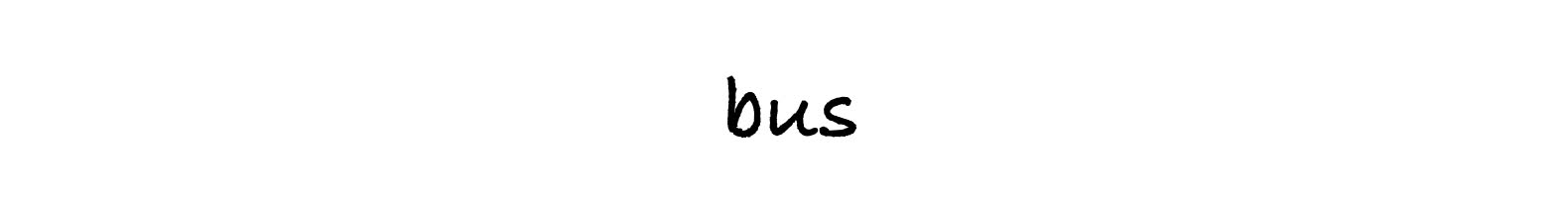}
        \end{minipage}
        \begin{minipage}[t]{0.16\textwidth}
            \centering
            \includegraphics[width=\linewidth, trim=165 20 165 30, clip]{figures/prompt/students.jpg}
        \end{minipage}
        \begin{minipage}[t]{0.16\textwidth}
            \centering
            \includegraphics[width=\linewidth, trim=185 20 185 30, clip]{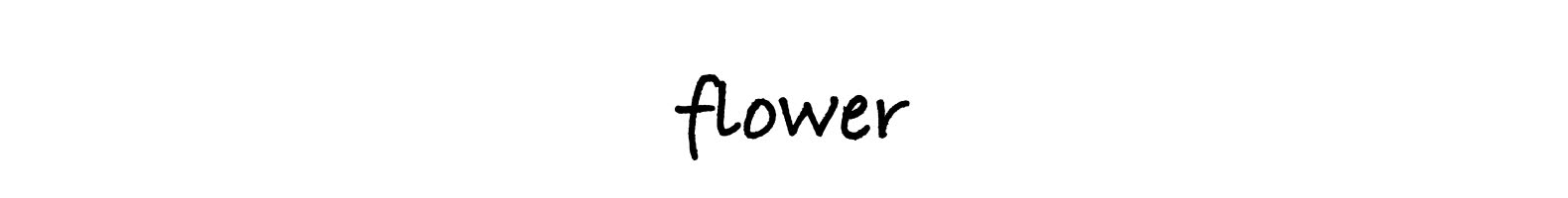}
        \end{minipage}\\

    \end{tabular}

    \vspace{-4mm}
    \caption{\textbf{Random samples on beanbag.} We present more random examples on beanbag (concave mesh). Styles (left to right): pencil sketch, pencil sketch, sketch, pencil sketch, ink drawing, monogram, pencil sketch, pop art.}
    \label{fig:randomsamplebeanbag}
\end{figure*}

\begin{figure*}[htbp]
\vspace{-6.5mm}
    \centering
    \setlength{\tabcolsep}{1pt} 
    \renewcommand{\arraystretch}{1}
    \vspace{-2mm}
    \begin{tabular}{cccccc}
      \begin{minipage}[t]{0.16\textwidth}
        \includegraphics[width=\textwidth, trim=100 120 100 180, clip]{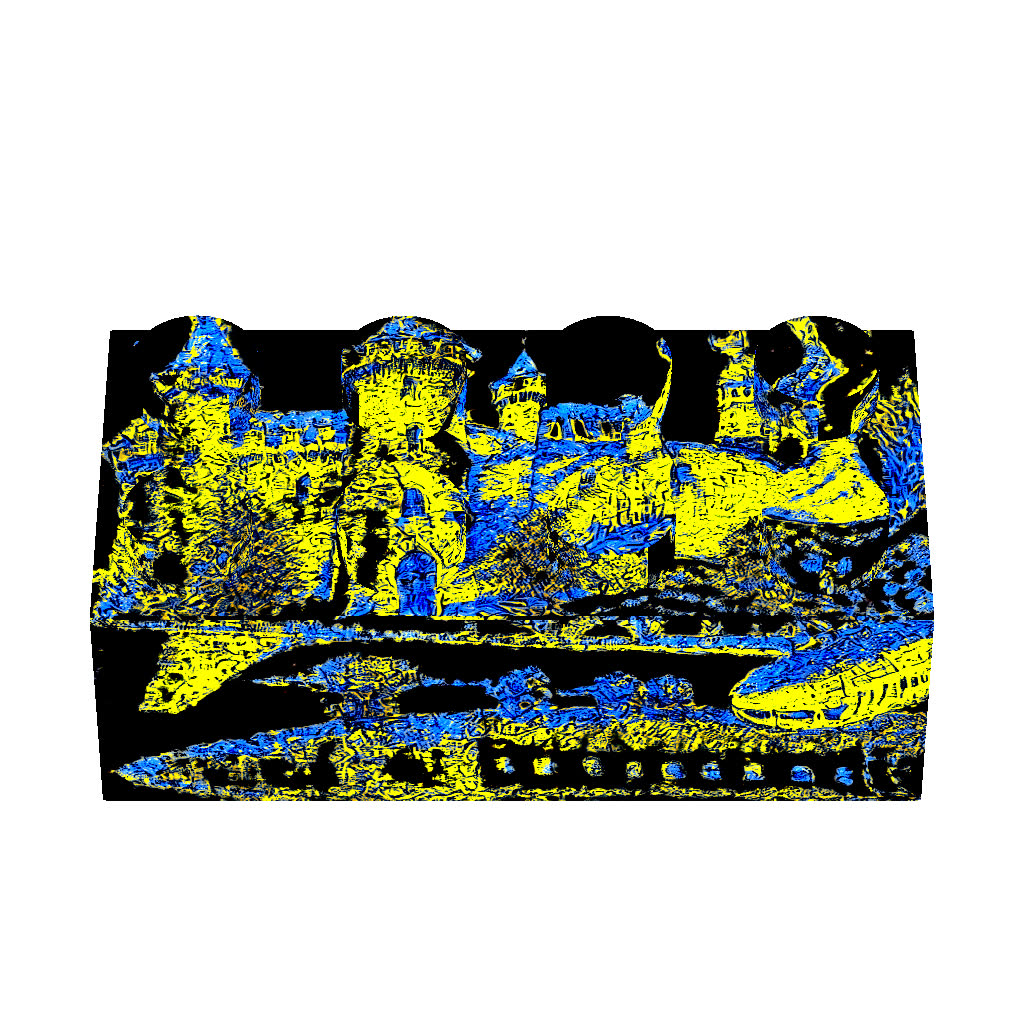}
    \end{minipage}%
    \begin{minipage}[t]{0.16\textwidth}
        \includegraphics[width=\textwidth, trim=45 170 155 130, clip]{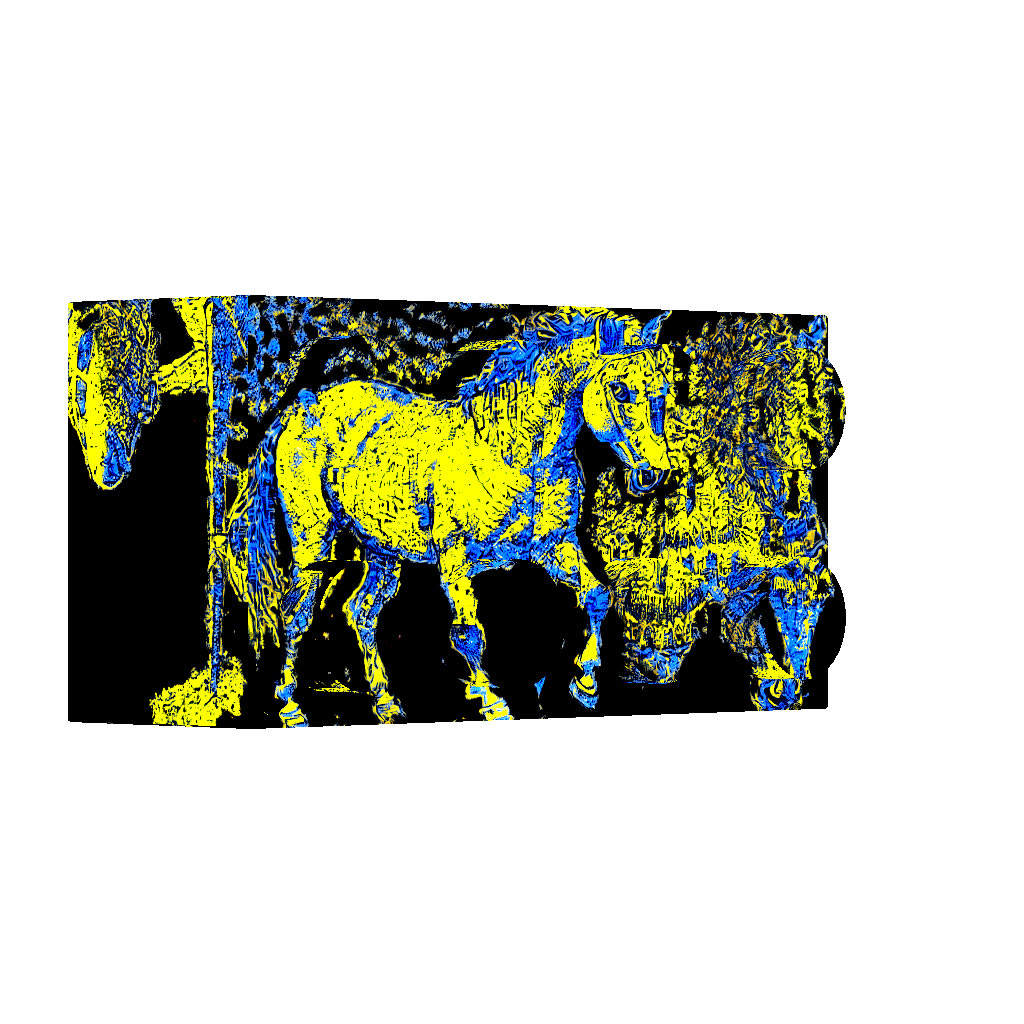}
    \end{minipage}%
    \begin{minipage}[t]{0.16\textwidth}
        \includegraphics[width=\textwidth, trim=140 260 140 40, clip]{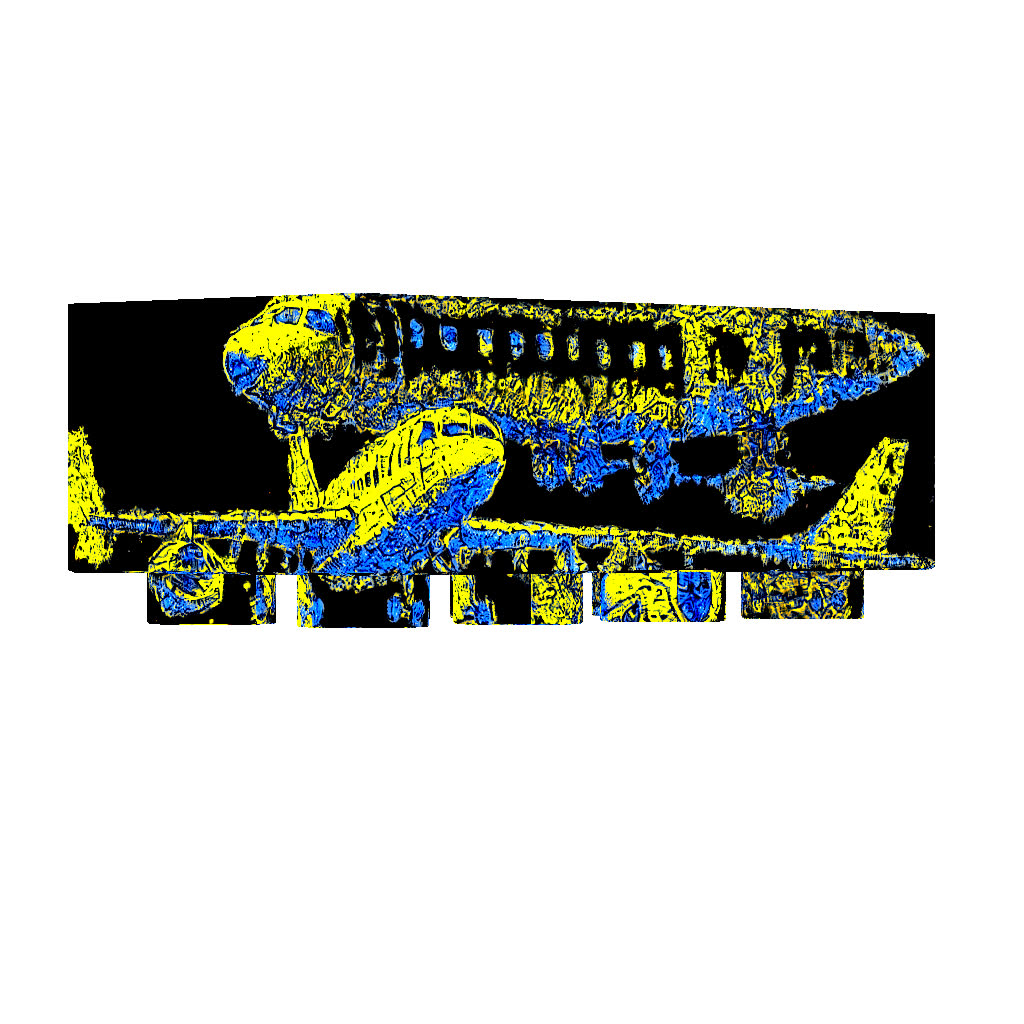}
    \end{minipage}%
    \begin{minipage}[t]{0.16\textwidth}
        \includegraphics[width=\textwidth, trim=100 120 100 180, clip]{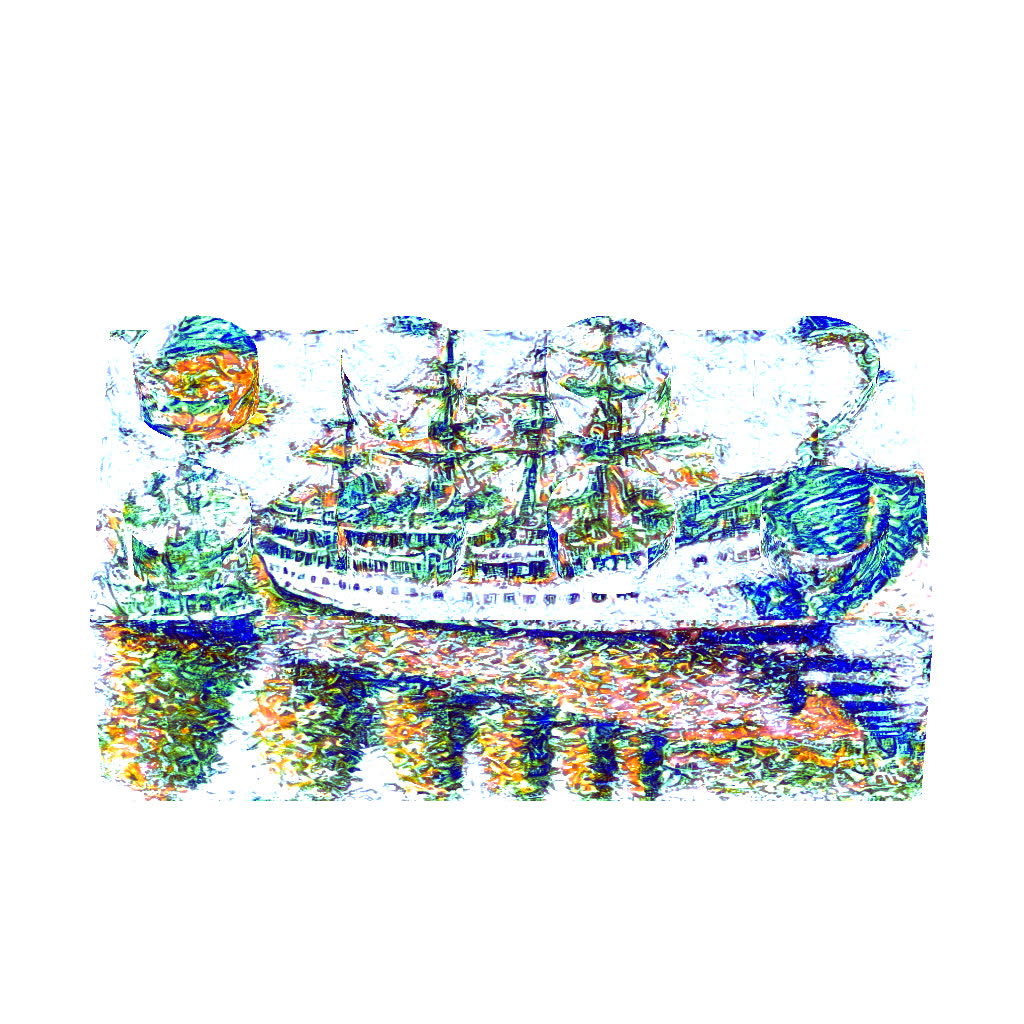}
    \end{minipage}%
    \begin{minipage}[t]{0.16\textwidth}
        \includegraphics[width=\textwidth, trim=45 170 155 130, clip]{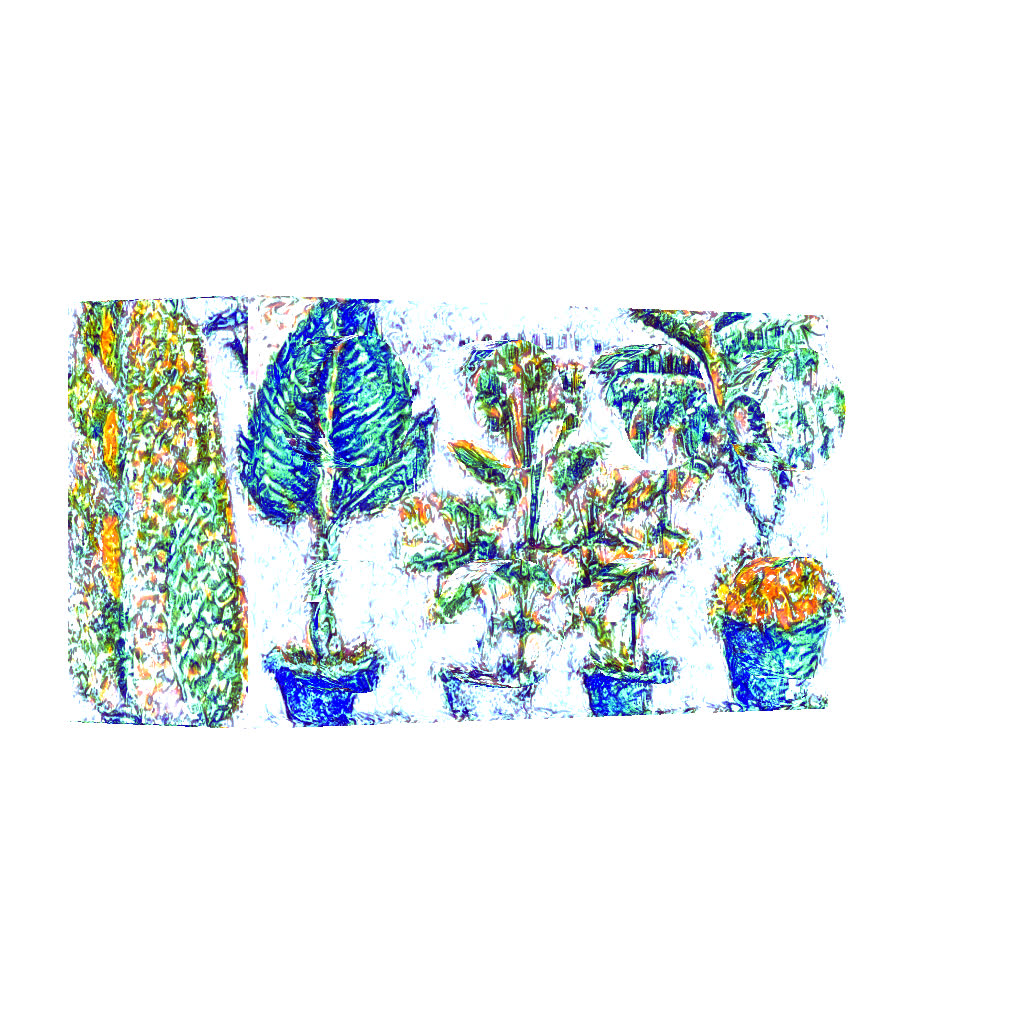}
    \end{minipage}%
    \begin{minipage}[t]{0.16\textwidth}
        \includegraphics[width=\textwidth, trim=140 260 140 40, clip]{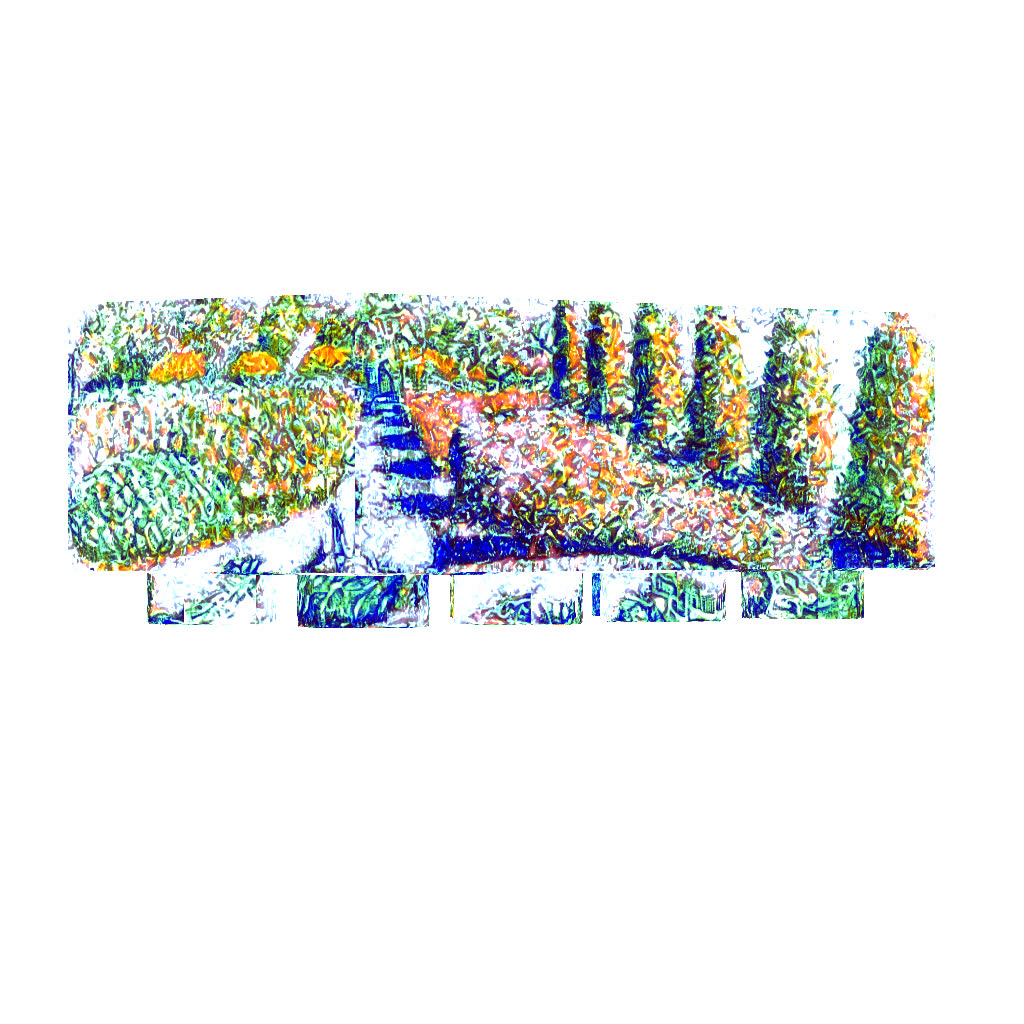}
    \end{minipage}\\
    \vspace{-1mm}
        \begin{minipage}[t]{0.16\textwidth}
        \centering
            \includegraphics[width=\linewidth, trim=185 20 185 30, clip]{figures/prompt/castle.jpg}
            \end{minipage}\hfill
        \begin{minipage}[t]{0.16\textwidth}
            \centering
            \includegraphics[width=\linewidth, trim=185 20 185 30, clip]{figures/prompt/horse.jpg}
            \end{minipage}\hfill
        \begin{minipage}[t]{0.16\textwidth}
            \centering
            \includegraphics[width=\linewidth, trim=185 20 185 30, clip]{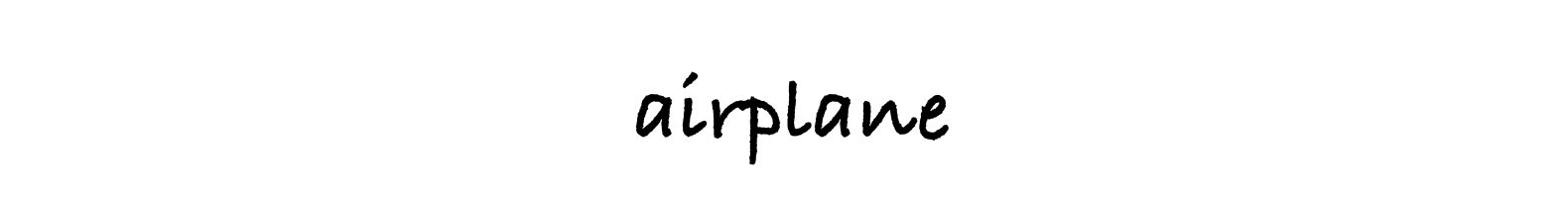}
        \end{minipage}\hfill
        \begin{minipage}[t]{0.16\textwidth}
            \centering
            \includegraphics[width=\linewidth, trim=185 20 185 30, clip]{figures/prompt/boat.jpg}
        \end{minipage}
        \begin{minipage}[t]{0.16\textwidth}
            \centering
            \includegraphics[width=\linewidth, trim=185 20 185 30, clip]{figures/prompt/houseplant.jpg}
        \end{minipage}
        \begin{minipage}[t]{0.16\textwidth}
            \centering
            \includegraphics[width=\linewidth, trim=185 15 185 30, clip]{figures/prompt/sculpturegarden.jpg}
        \end{minipage}\\
        \vspace{-3mm}
        \begin{minipage}[t]{0.16\textwidth}
            \includegraphics[width=\textwidth, trim=100 200 100 100, clip]{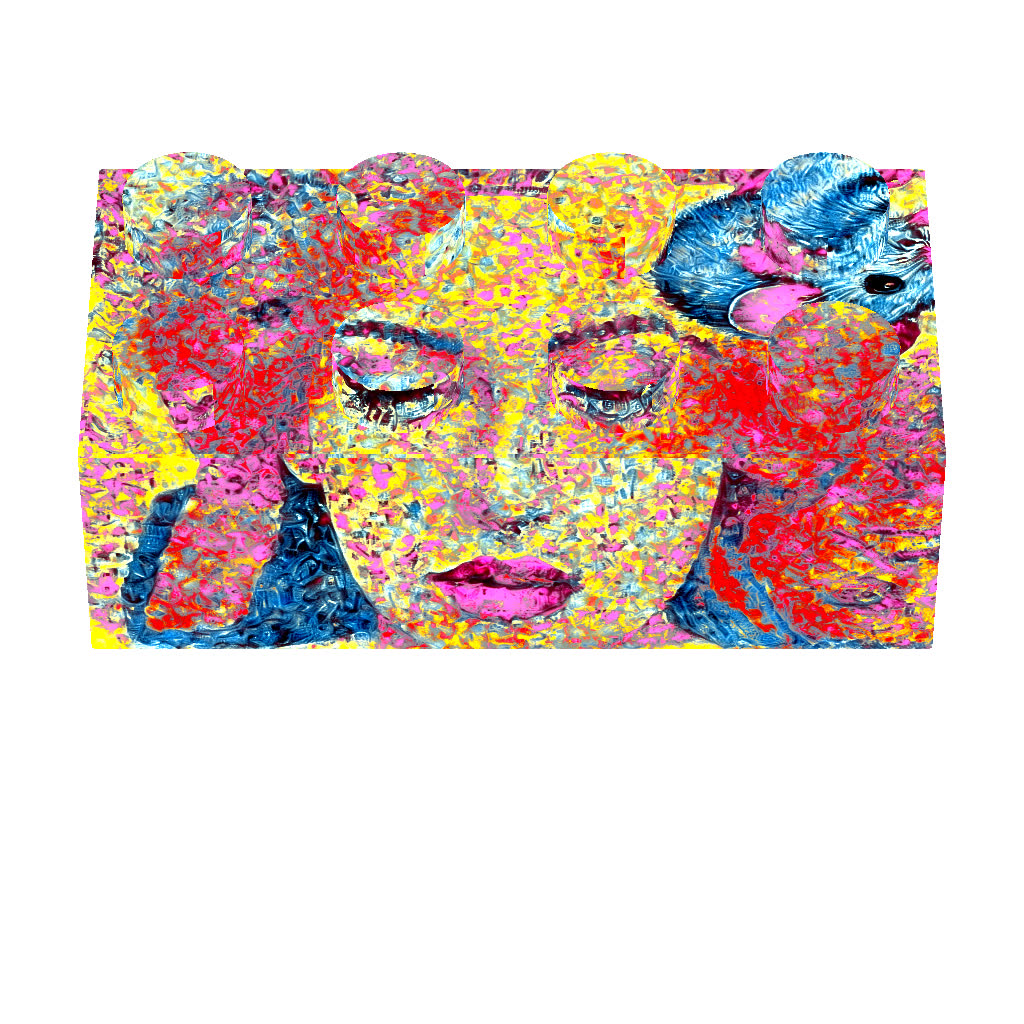}
        \end{minipage}%
        \begin{minipage}[t]{0.16\textwidth}
            \includegraphics[width=\textwidth, trim=200 120 0 180, clip]{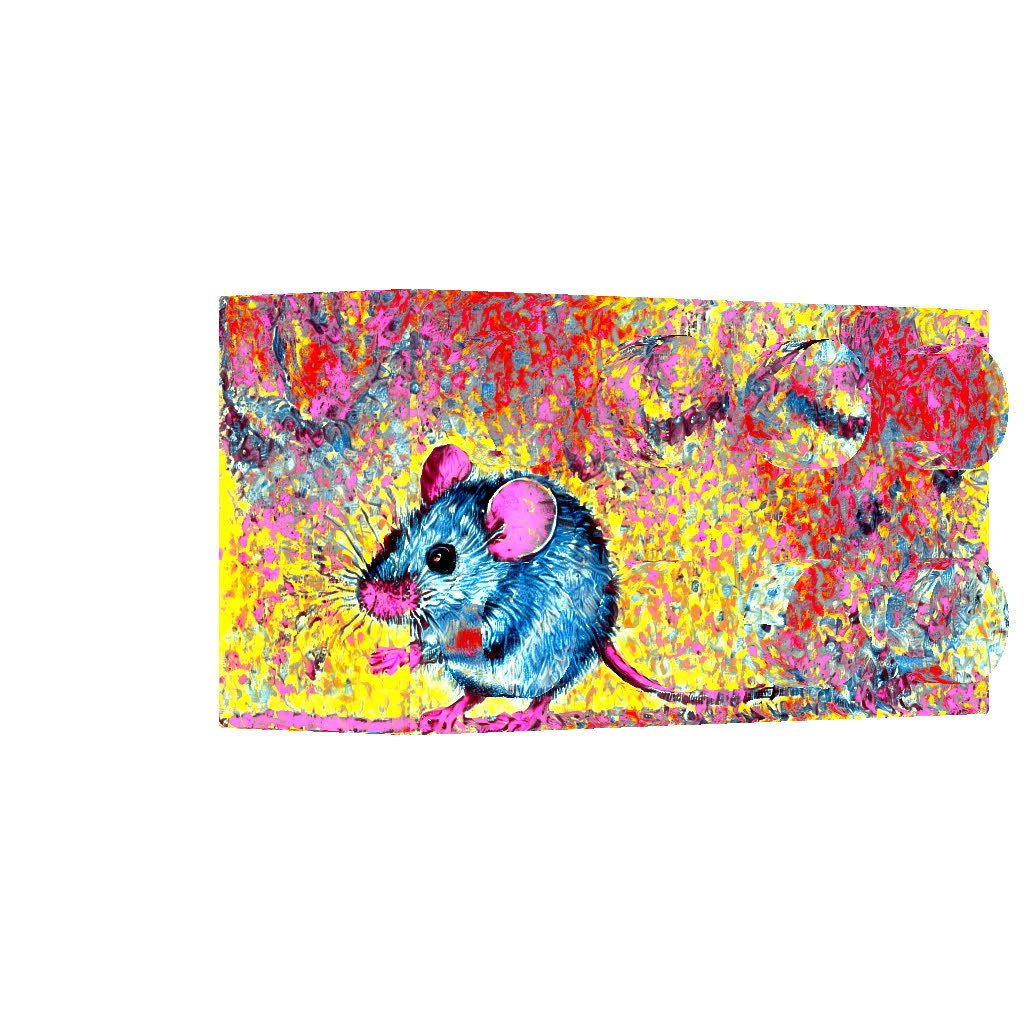}
        \end{minipage}%
        \begin{minipage}[t]{0.16\textwidth}
            \includegraphics[width=\textwidth, trim=140 0 140 300, clip]{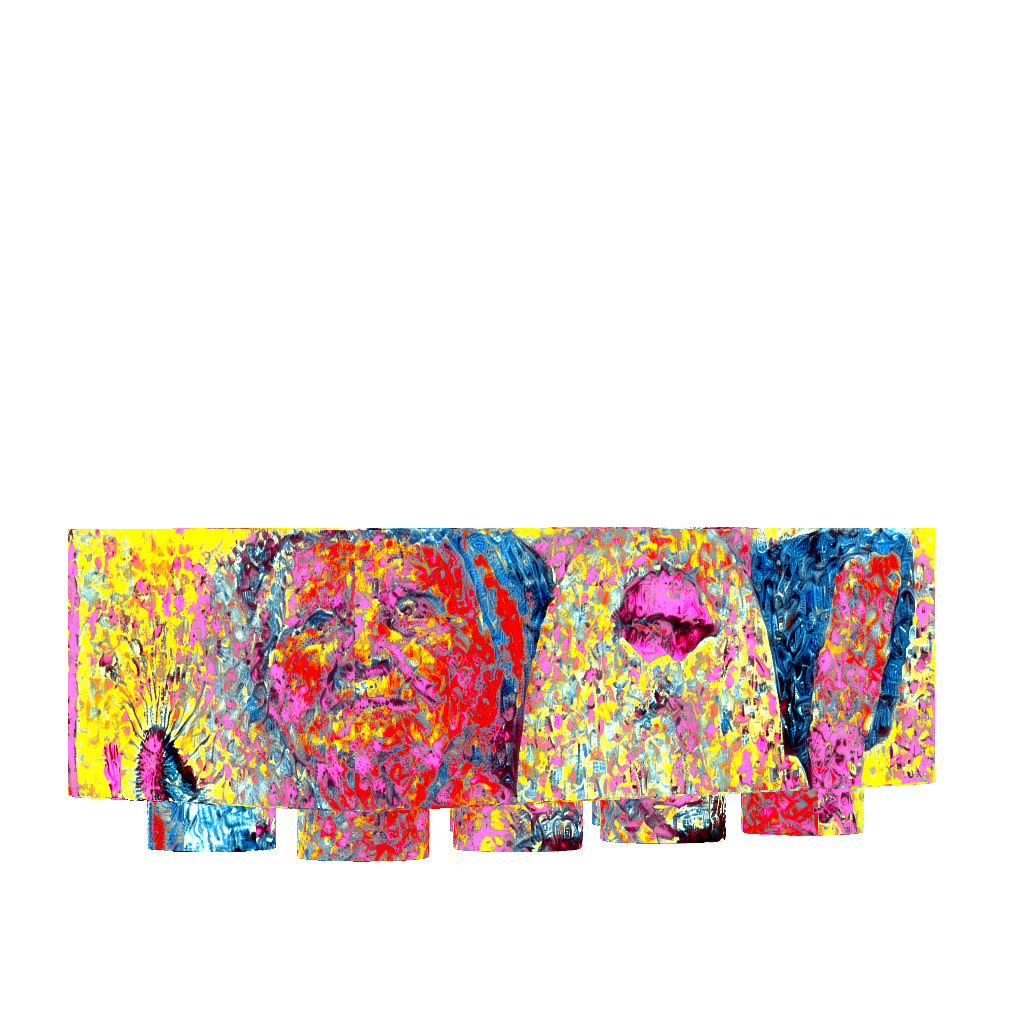}
        \end{minipage}%
        \begin{minipage}[t]{0.16\textwidth}
            \includegraphics[width=\textwidth, trim=100 200 100 100, clip]{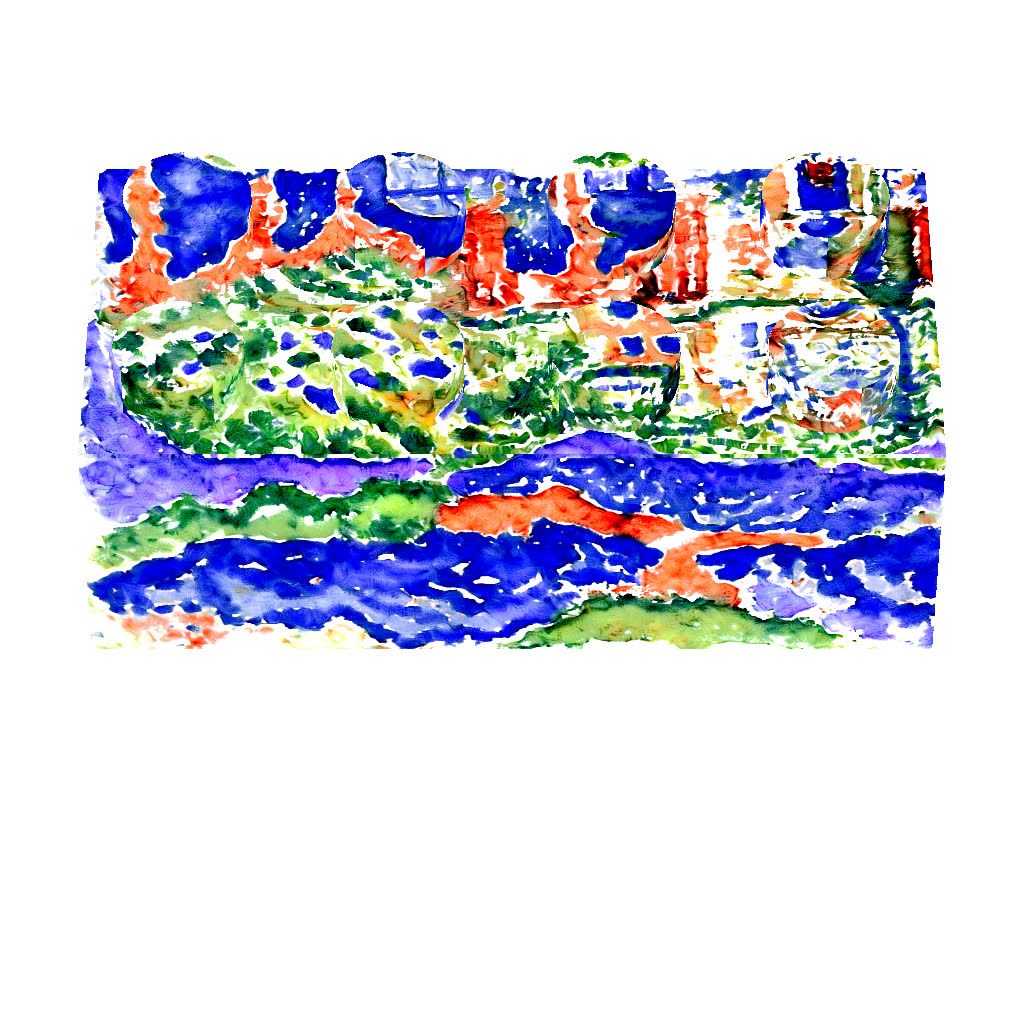}
        \end{minipage}%
        \begin{minipage}[t]{0.16\textwidth}
            \includegraphics[width=\textwidth, trim=200 120 0 180, clip]{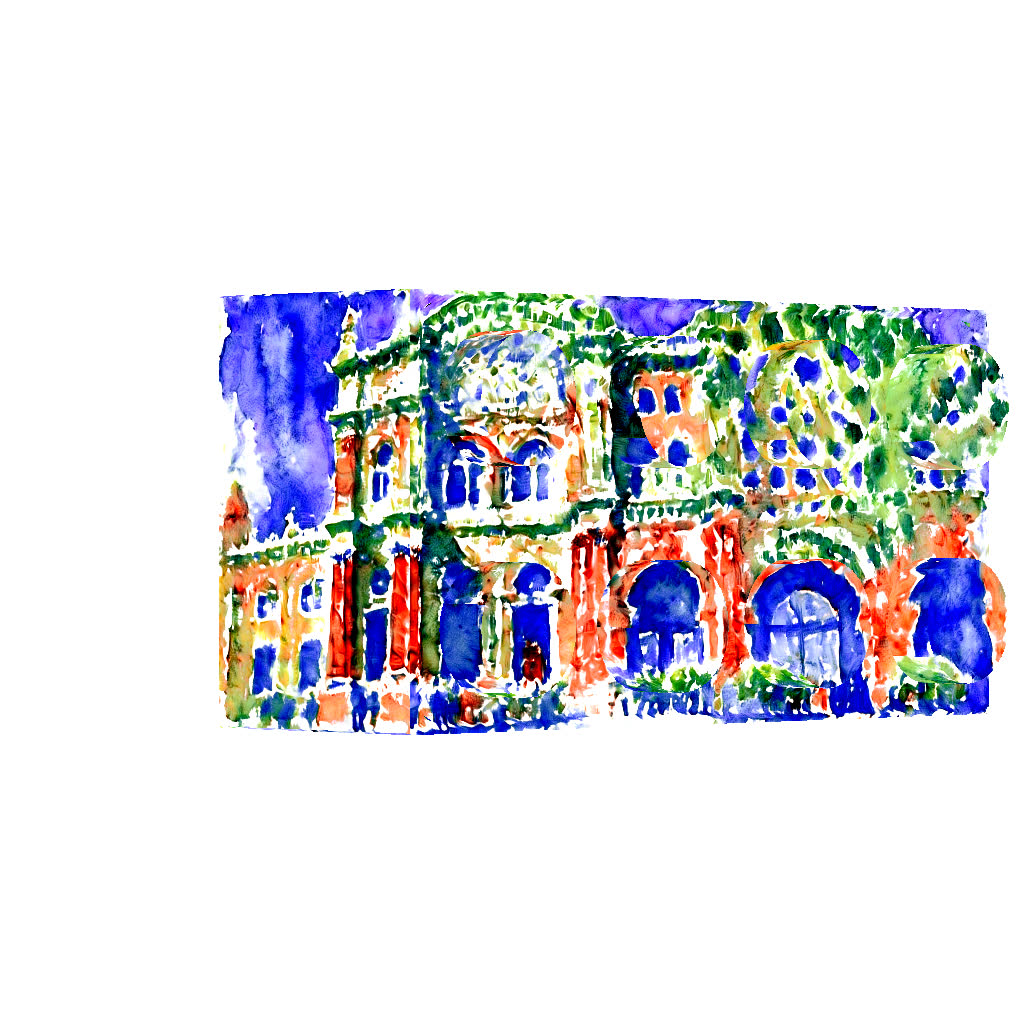}
        \end{minipage}%
        \begin{minipage}[t]{0.16\textwidth}
            \includegraphics[width=\textwidth, trim=140 0 140 300, clip]{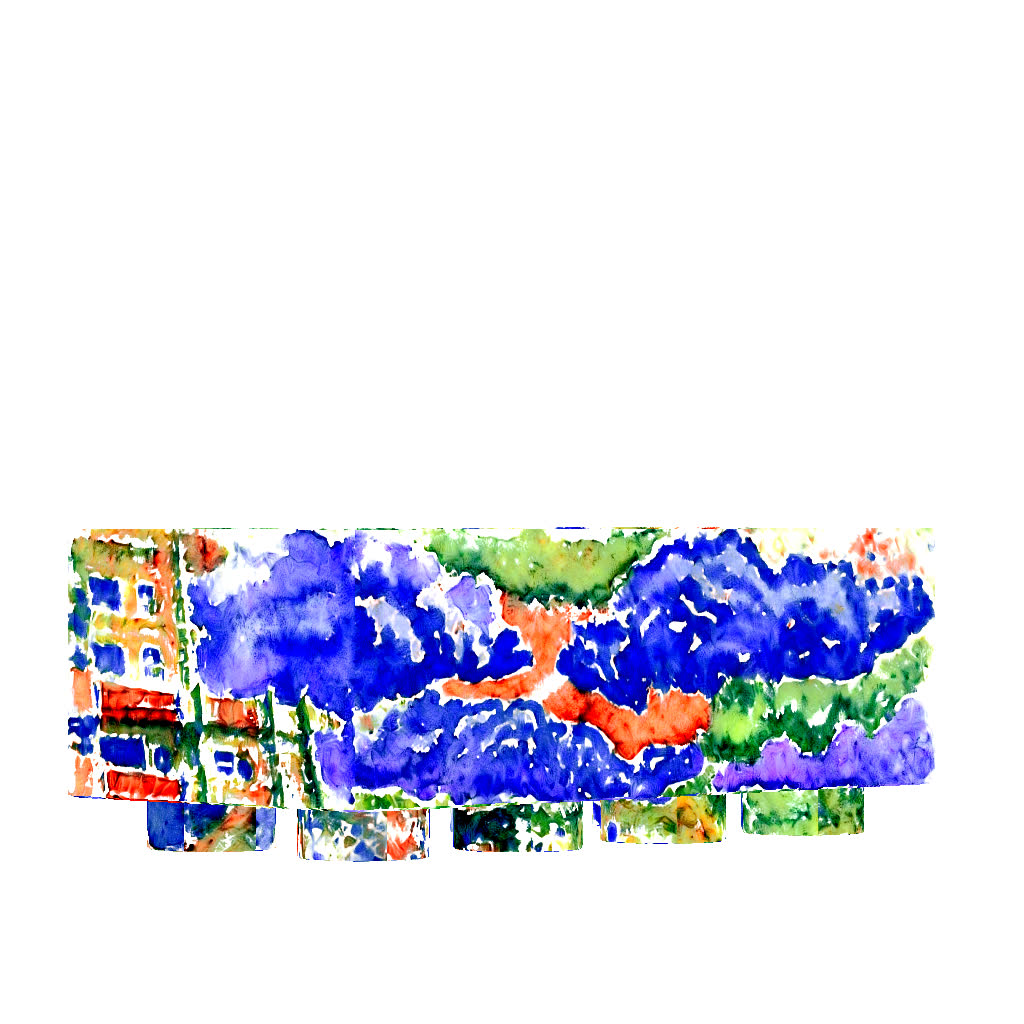}
        \end{minipage}\\

    \vspace{-2mm}
        \begin{minipage}[t]{0.16\textwidth}
        \centering
            \includegraphics[width=\linewidth, trim=185 20 185 30, clip]{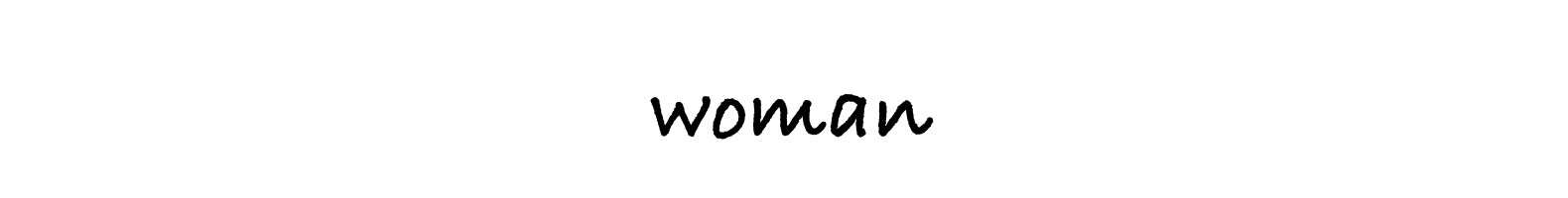}
            \end{minipage}\hfill
        \begin{minipage}[t]{0.16\textwidth}
            \centering
            \includegraphics[width=\linewidth, trim=185 15 185 30, clip]{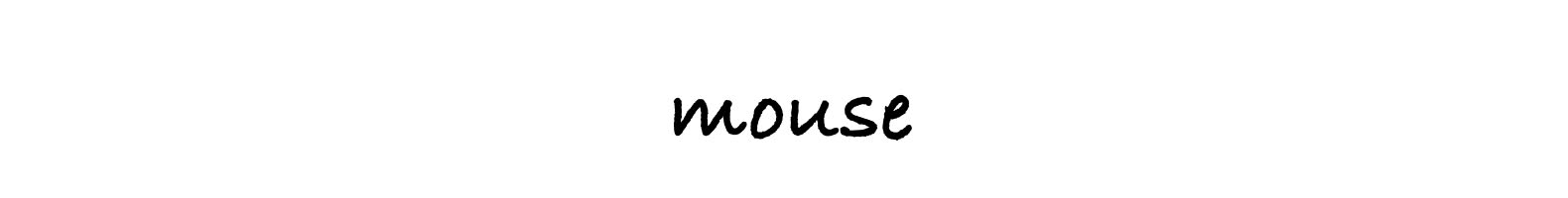}
            \end{minipage}\hfill
        \begin{minipage}[t]{0.16\textwidth}
            \centering
            \includegraphics[width=\linewidth, trim=185 20 185 30, clip]{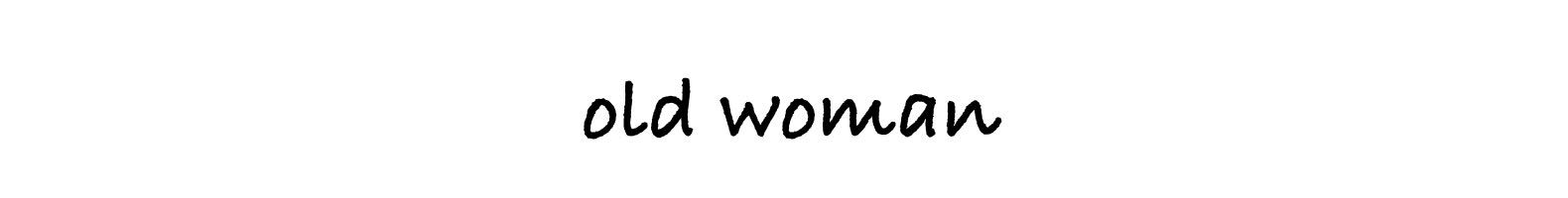}
        \end{minipage}\hfill
        \begin{minipage}[t]{0.16\textwidth}
            \centering
            \includegraphics[width=\linewidth, trim=185 20 185 30, clip]{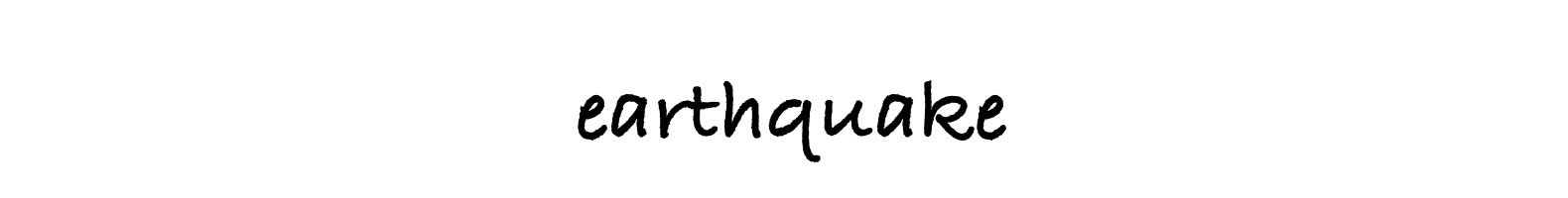}
        \end{minipage}
        \begin{minipage}[t]{0.16\textwidth}
            \centering
            \includegraphics[width=\linewidth, trim=185 20 185 30, clip]{figures/prompt/museum.jpg}
        \end{minipage}
        \begin{minipage}[t]{0.16\textwidth}
            \centering
            \includegraphics[width=\linewidth, trim=185 20 185 30, clip]{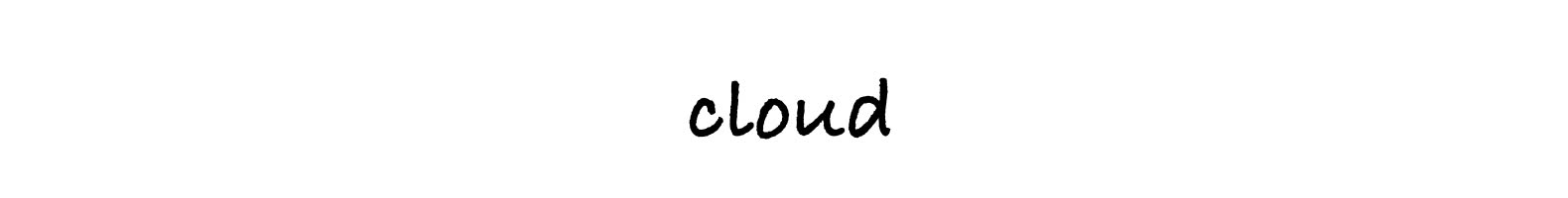}
        \end{minipage}\\      
    \vspace{-3mm}
    \begin{minipage}[t]{0.16\textwidth}
        \includegraphics[width=\textwidth, trim=100 200 100 100, clip]{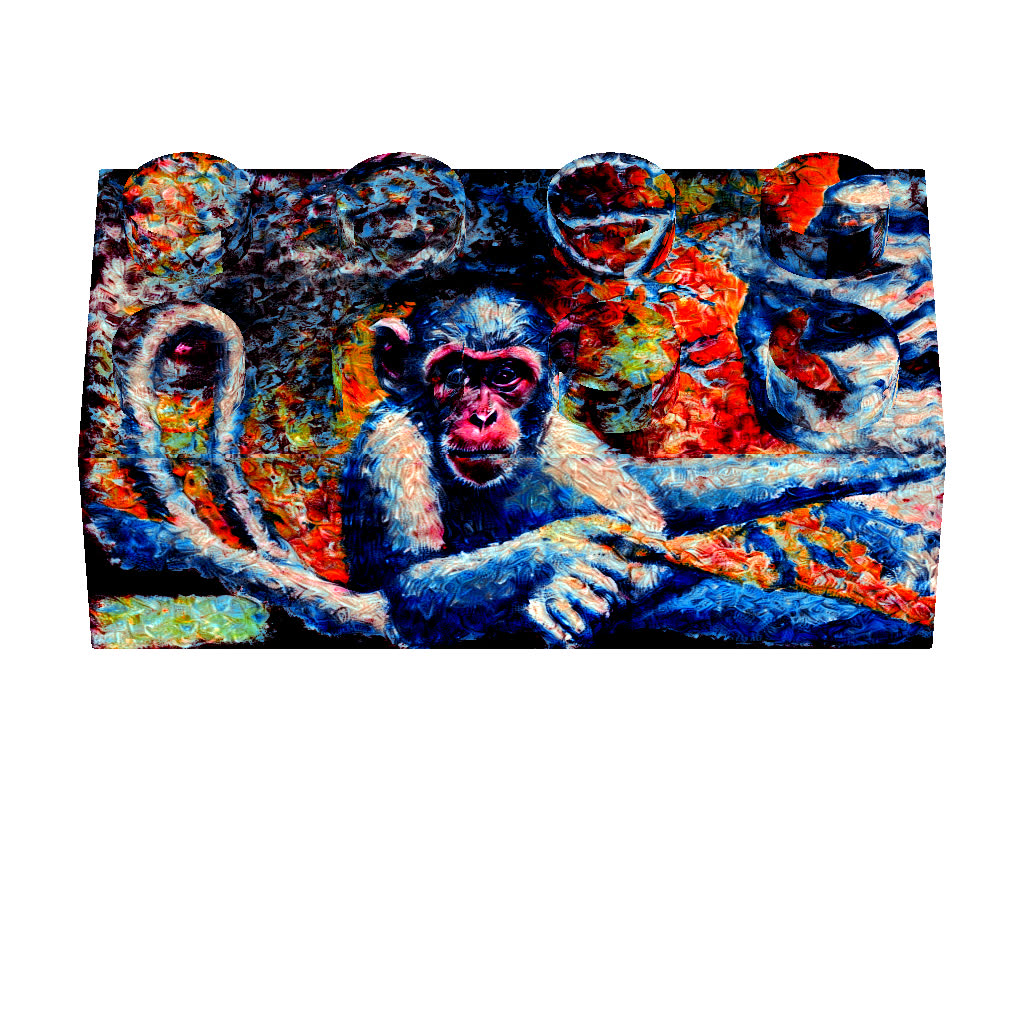}
    \end{minipage}%
    \begin{minipage}[t]{0.16\textwidth}
        \includegraphics[width=\textwidth, trim=200 120 0 180, clip]{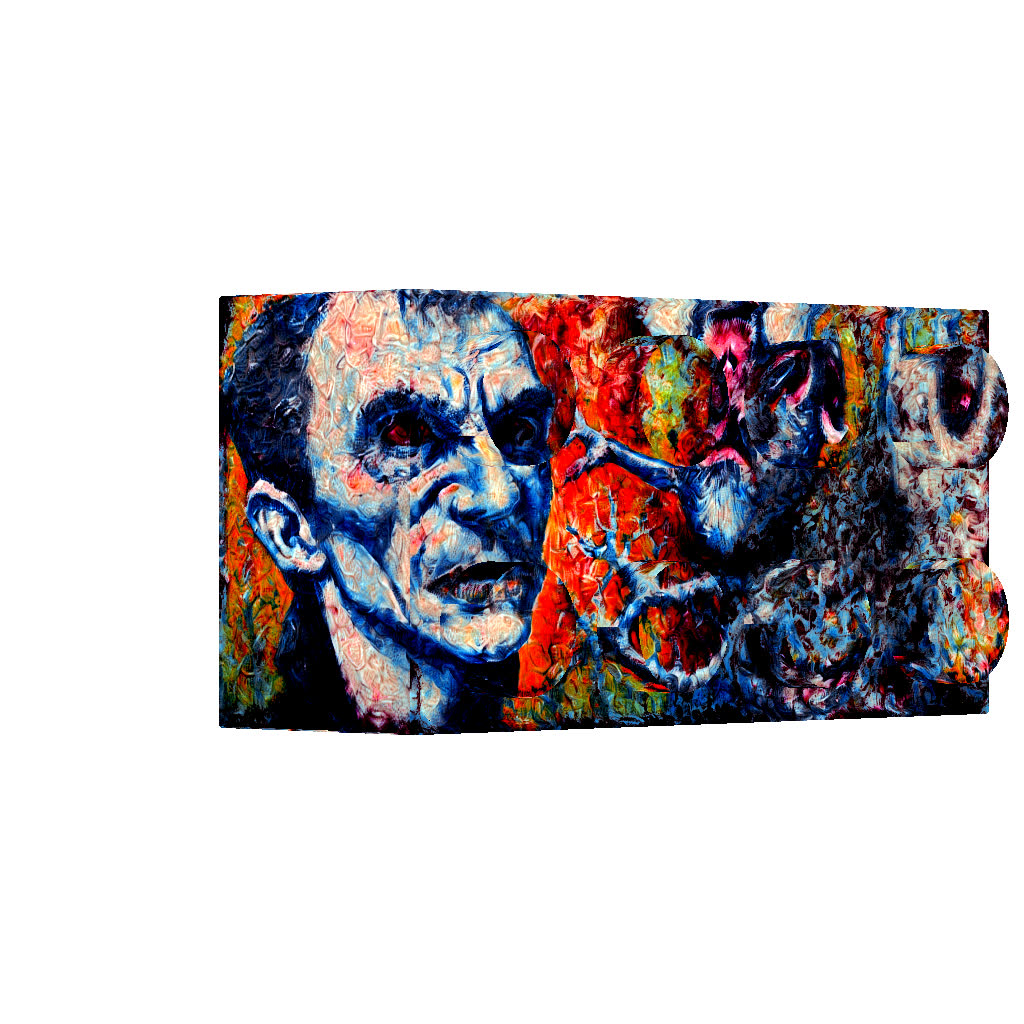}
    \end{minipage}%
    \begin{minipage}[t]{0.16\textwidth}
        \includegraphics[width=\textwidth, trim=140 0 140 300, clip]{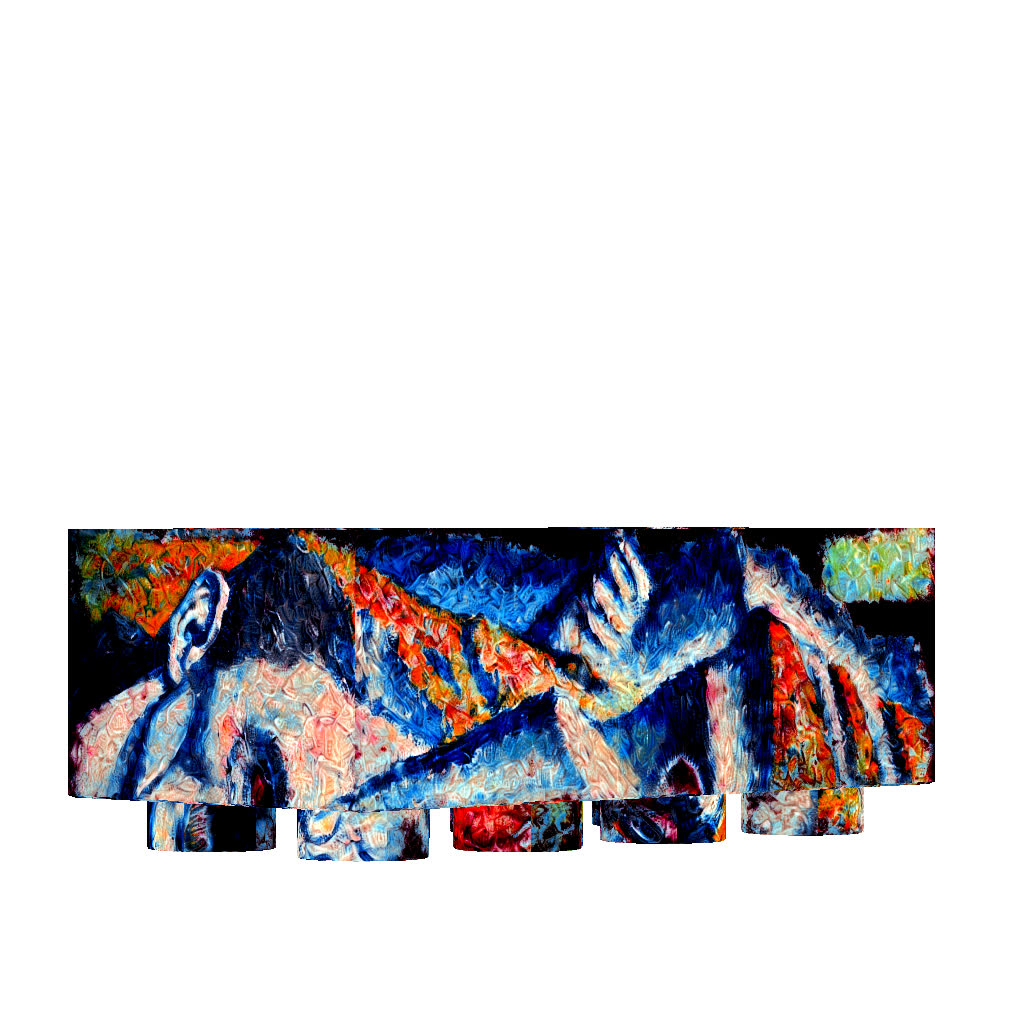}
    \end{minipage}%
    \begin{minipage}[t]{0.16\textwidth}
        \includegraphics[width=\textwidth, trim=100 200 100 100, clip]{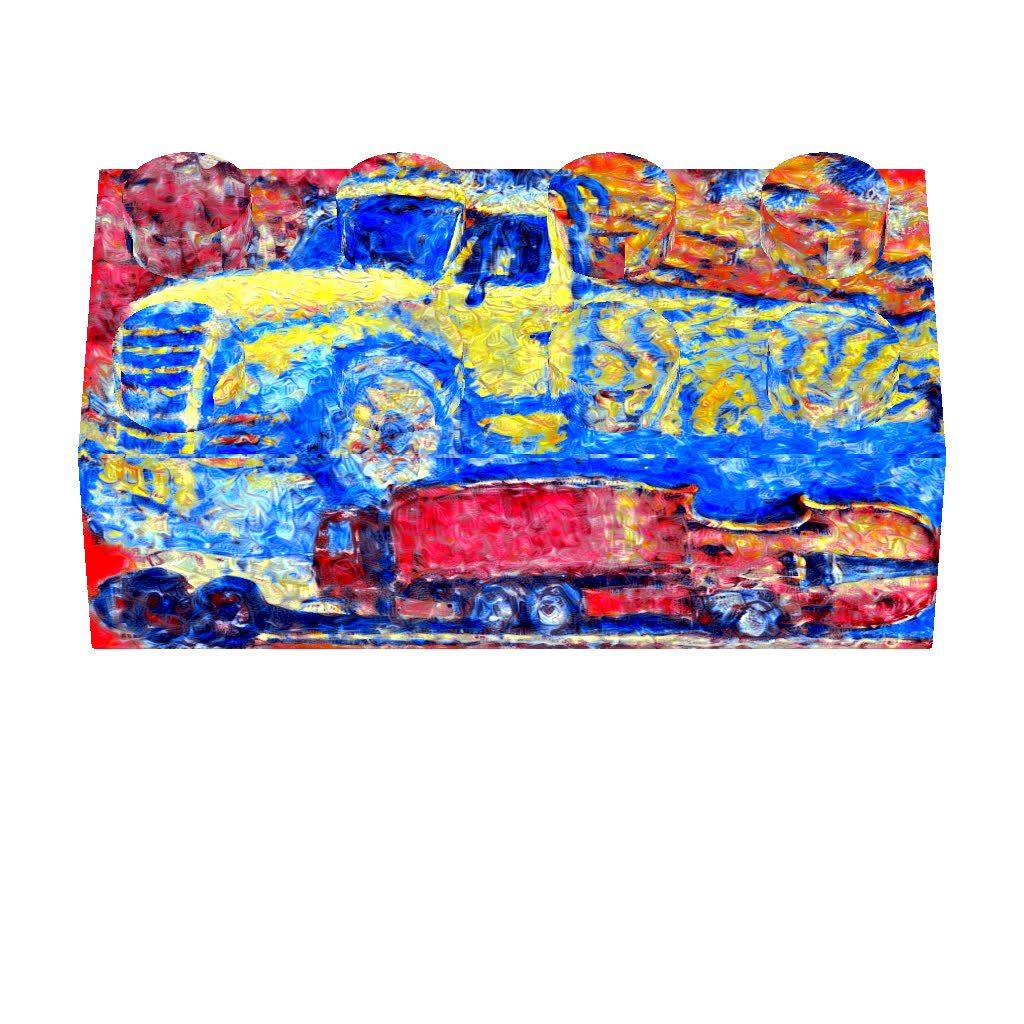}
    \end{minipage}%
    \begin{minipage}[t]{0.16\textwidth}
        \includegraphics[width=\textwidth, trim=200 120 0 180, clip]{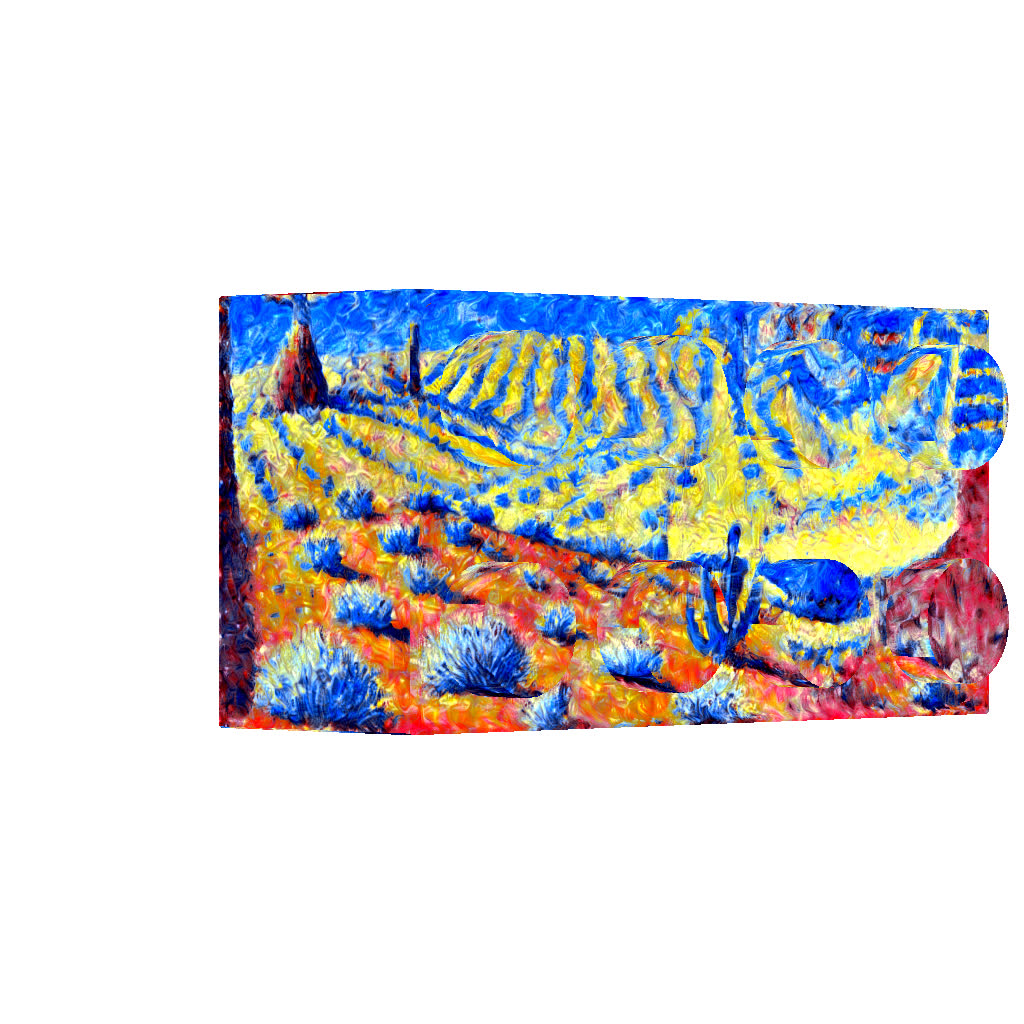}
    \end{minipage}%
    \begin{minipage}[t]{0.16\textwidth}
        \includegraphics[width=\textwidth, trim=140 0 140 300, clip]{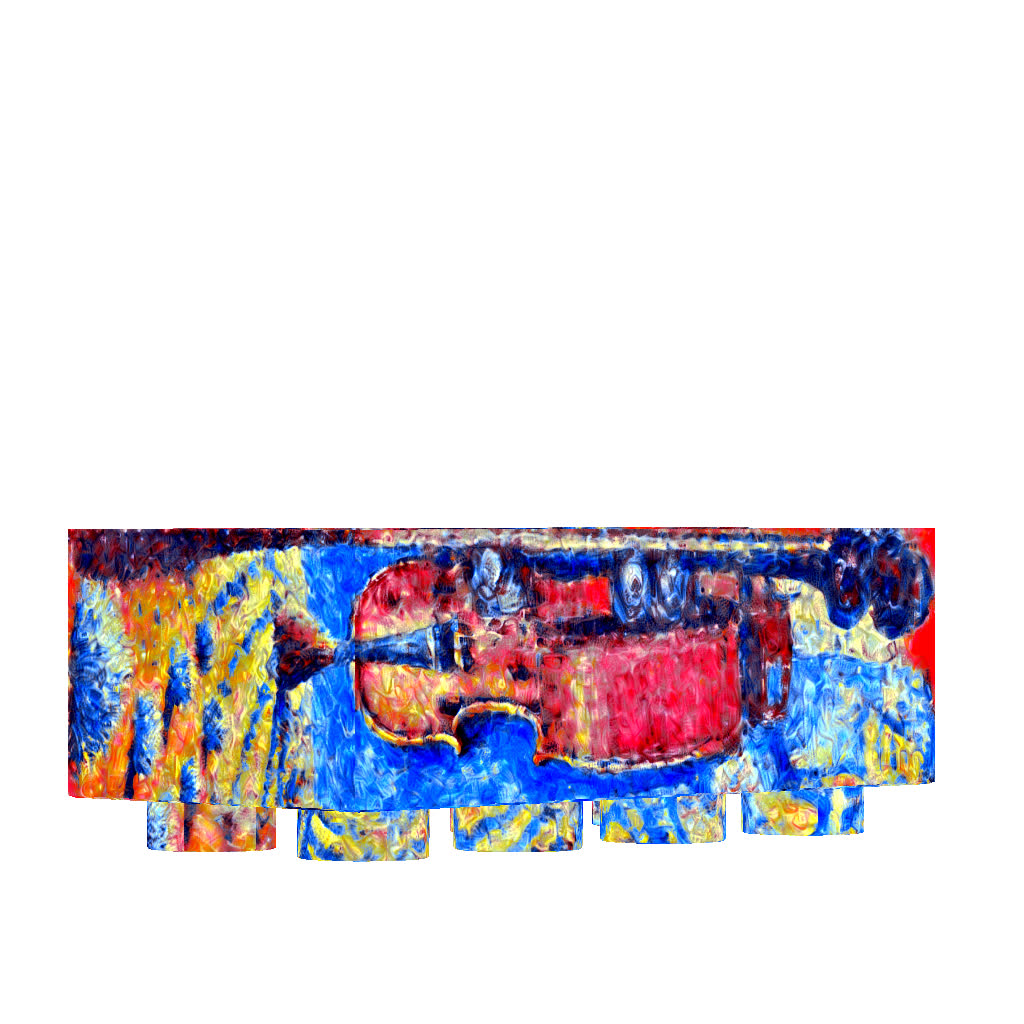}
    \end{minipage}\\
    \vspace{-1mm}
        \begin{minipage}[t]{0.16\textwidth}
        \centering
            \includegraphics[width=\linewidth, trim=185 15 185 30, clip]{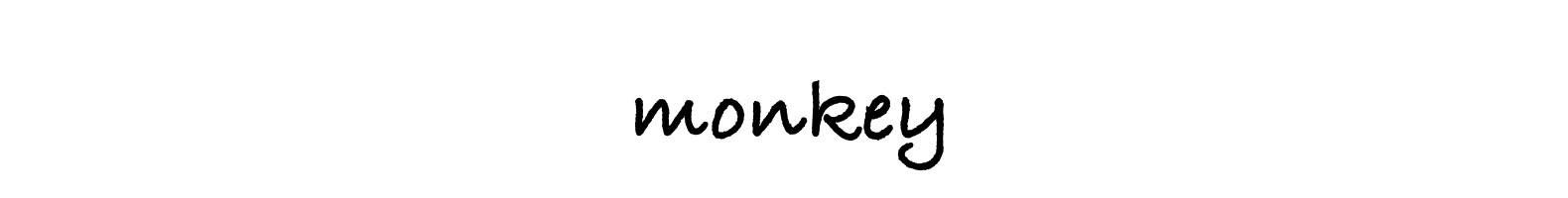}
            \end{minipage}\hfill
        \begin{minipage}[t]{0.16\textwidth}
            \centering
            \includegraphics[width=\linewidth, trim=185 20 185 30, clip]{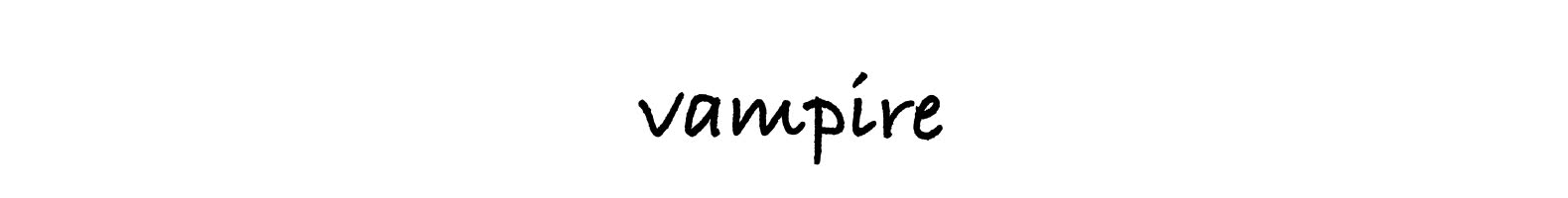}
            \end{minipage}\hfill
        \begin{minipage}[t]{0.16\textwidth}
            \centering
            \includegraphics[width=\linewidth, trim=185 20 185 30, clip]{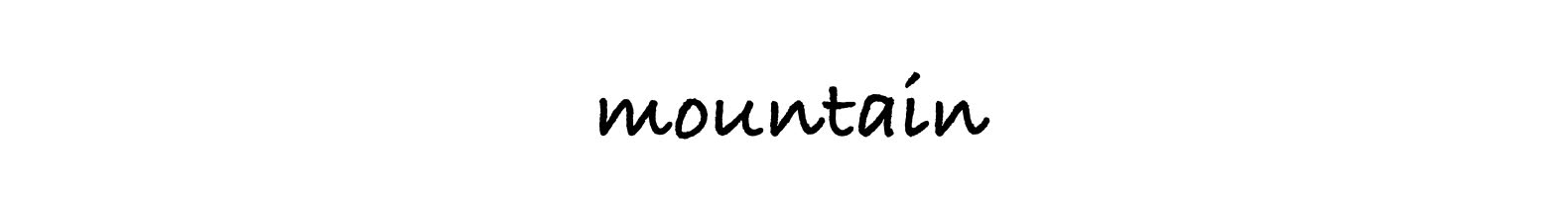}
        \end{minipage}\hfill
        \begin{minipage}[t]{0.16\textwidth}
            \centering
            \includegraphics[width=\linewidth, trim=185 15 185 30, clip]{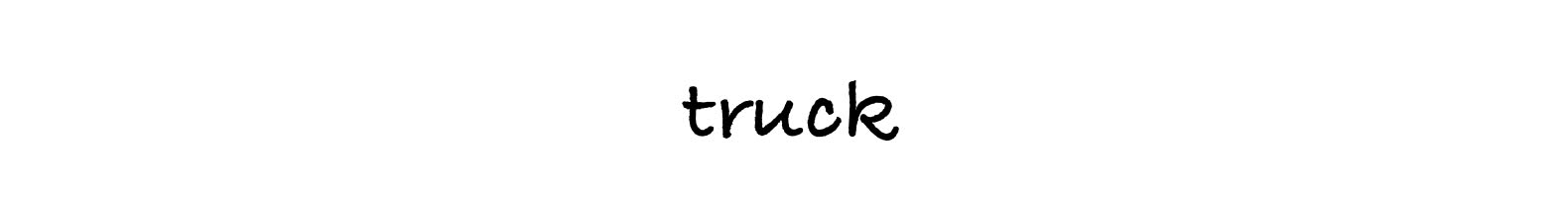}
        \end{minipage}
        \begin{minipage}[t]{0.16\textwidth}
            \centering
            \includegraphics[width=\linewidth, trim=185 20 185 30, clip]{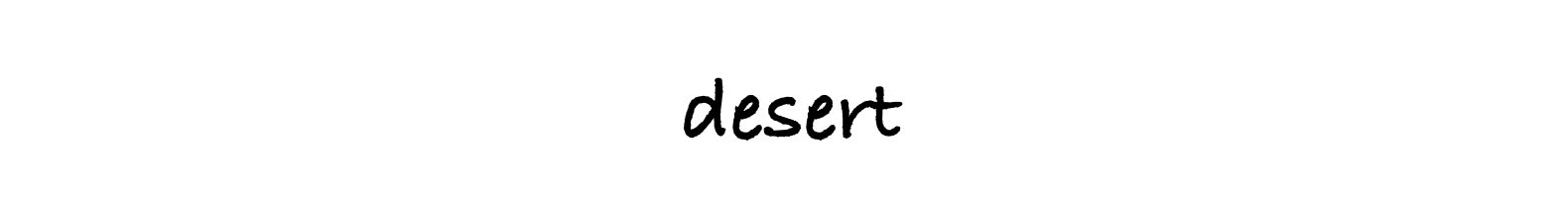}
        \end{minipage}
        \begin{minipage}[t]{0.16\textwidth}
            \centering
            \includegraphics[width=\linewidth, trim=185 20 185 30, clip]{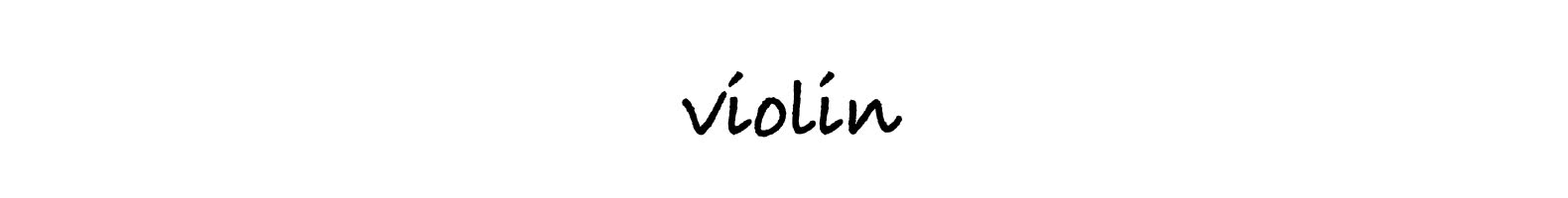}
        \end{minipage}\\
        
    \vspace{-3mm}
    \begin{minipage}{0.16\textwidth}
    \includegraphics[width=\linewidth, trim=100 200 100 100, clip]{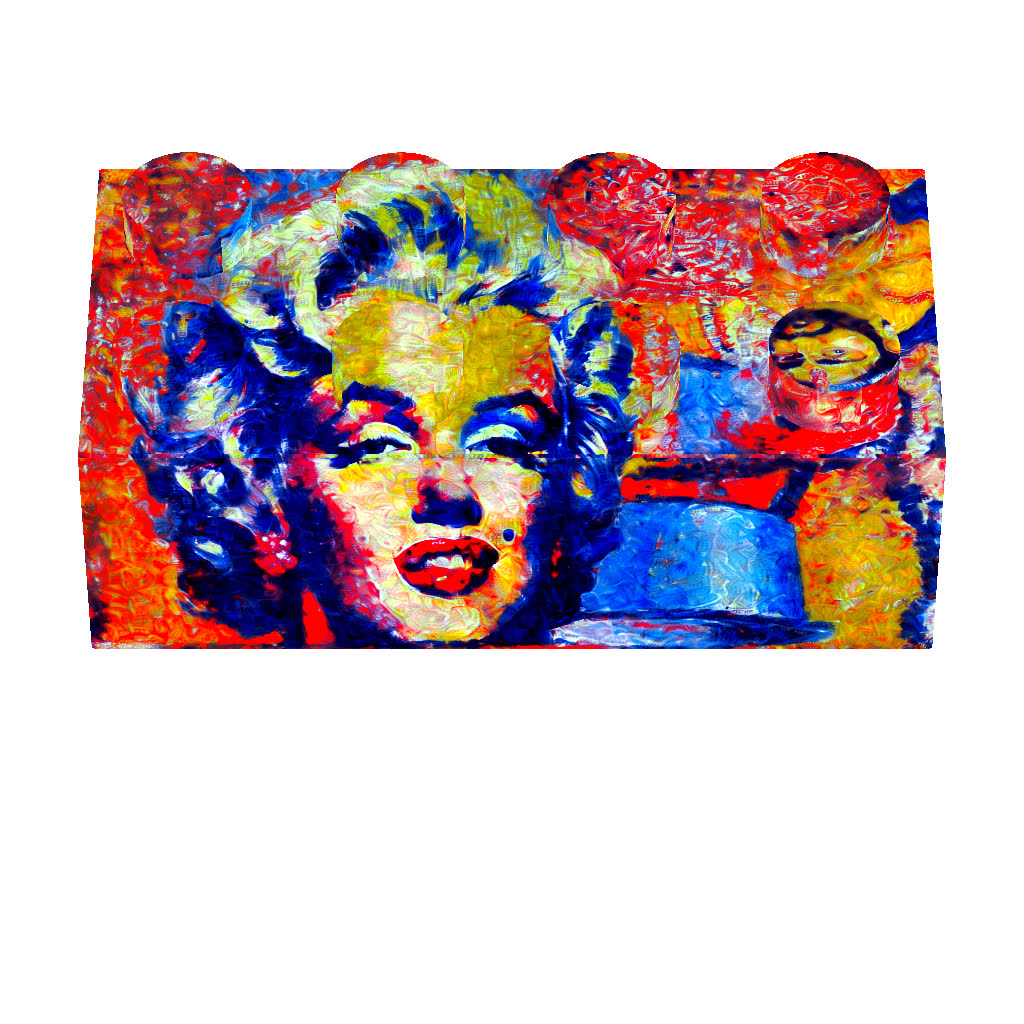}
    \end{minipage}%
    \begin{minipage}{0.16\textwidth}
        \includegraphics[width=\linewidth, trim=200 120 0 180, clip]{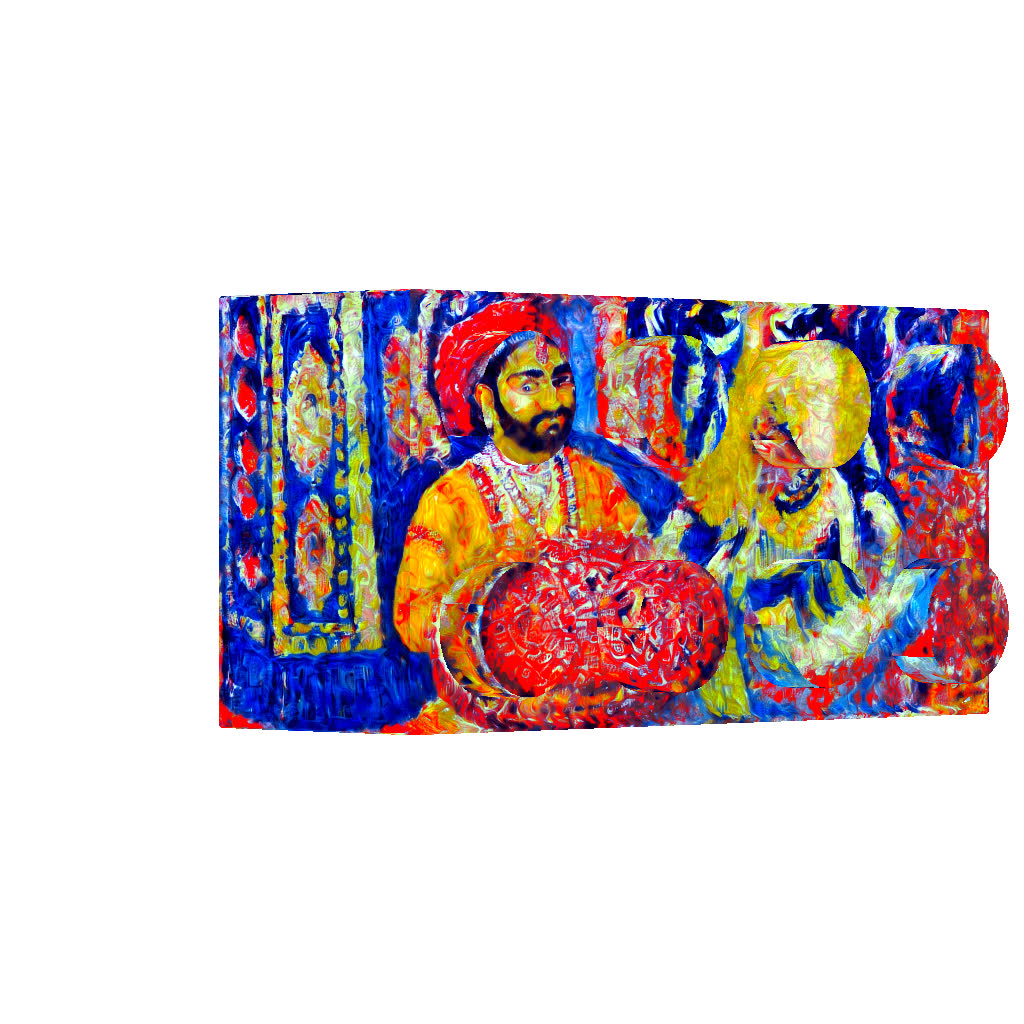}
    \end{minipage}%
    \begin{minipage}{0.16\textwidth}
        \includegraphics[width=\linewidth, trim=140 0 140 350, clip]{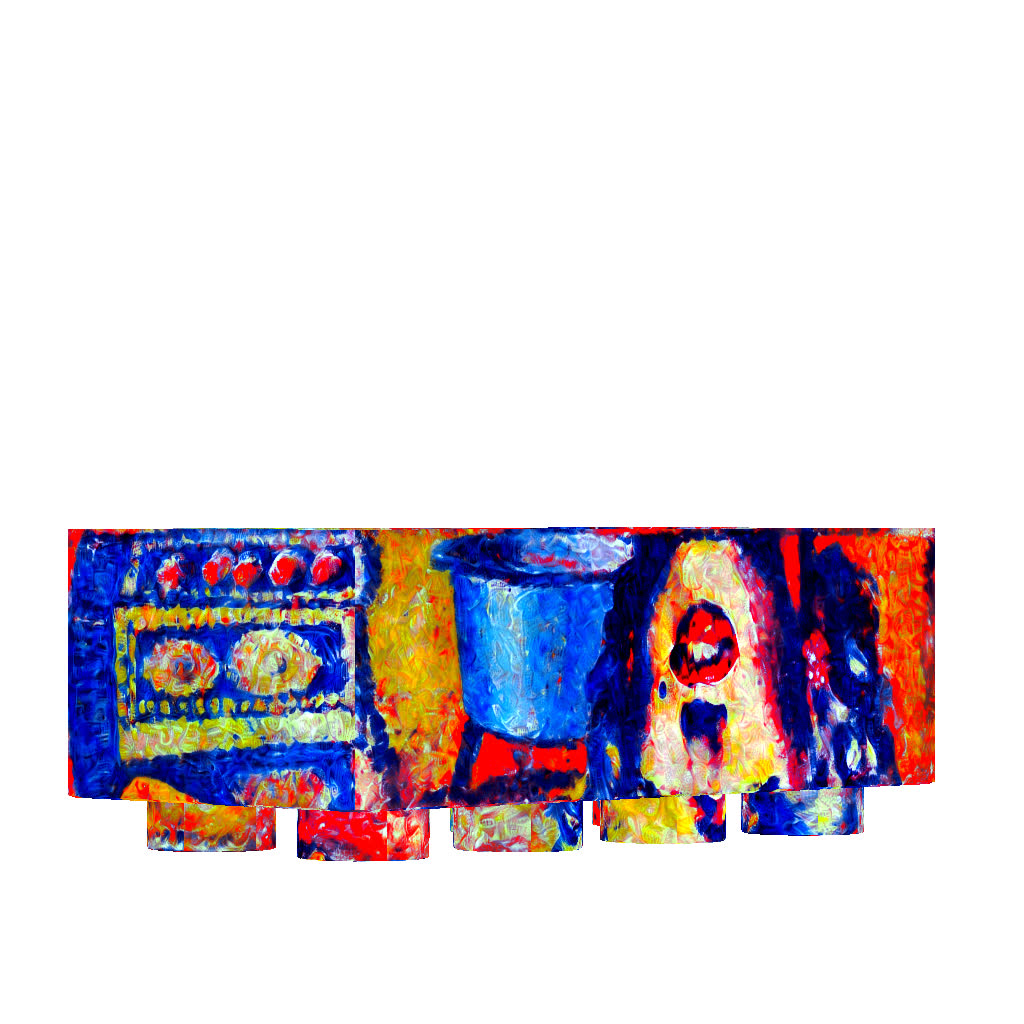}
    \end{minipage}%
    \begin{minipage}{0.16\textwidth}
        \includegraphics[width=\linewidth, trim=100 200 100 100, clip]{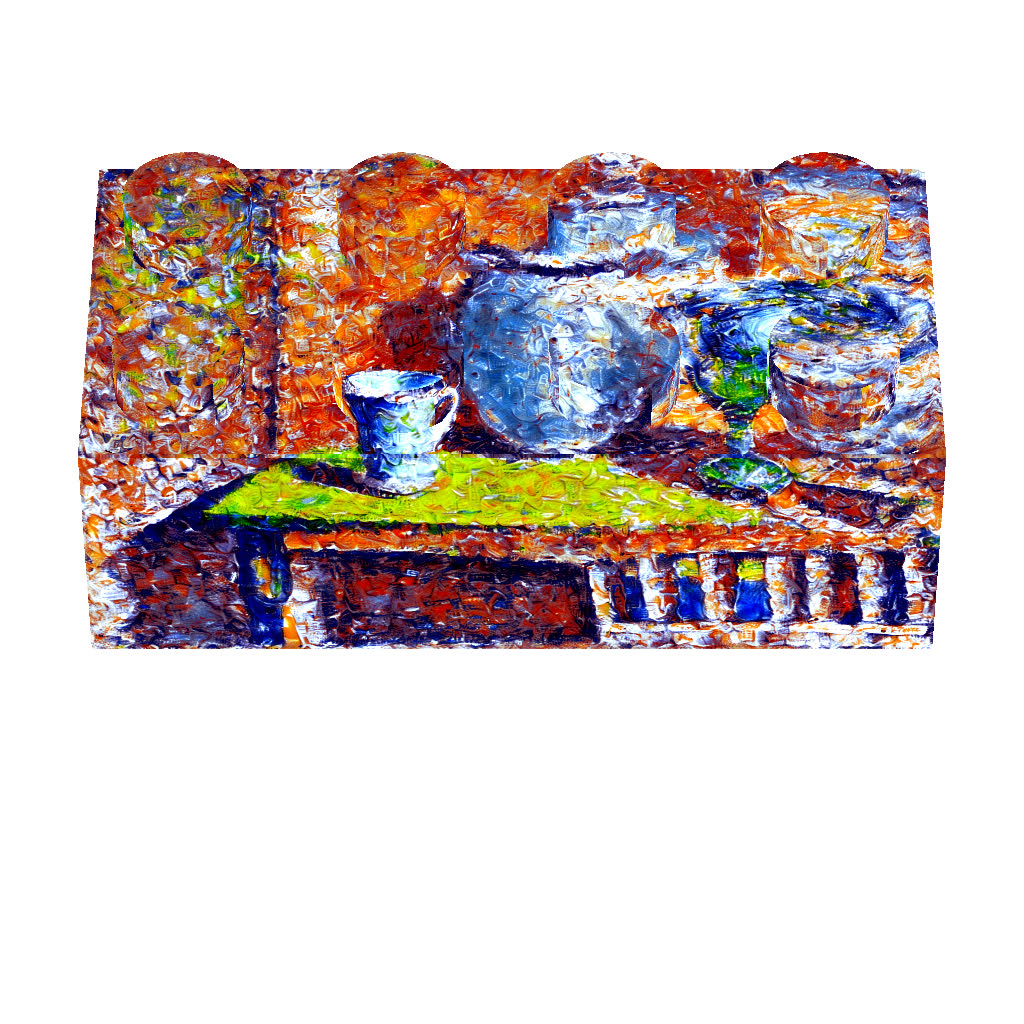}
    \end{minipage}%
    \begin{minipage}{0.16\textwidth}
        \includegraphics[width=\linewidth, trim=200 120 0 180, clip]{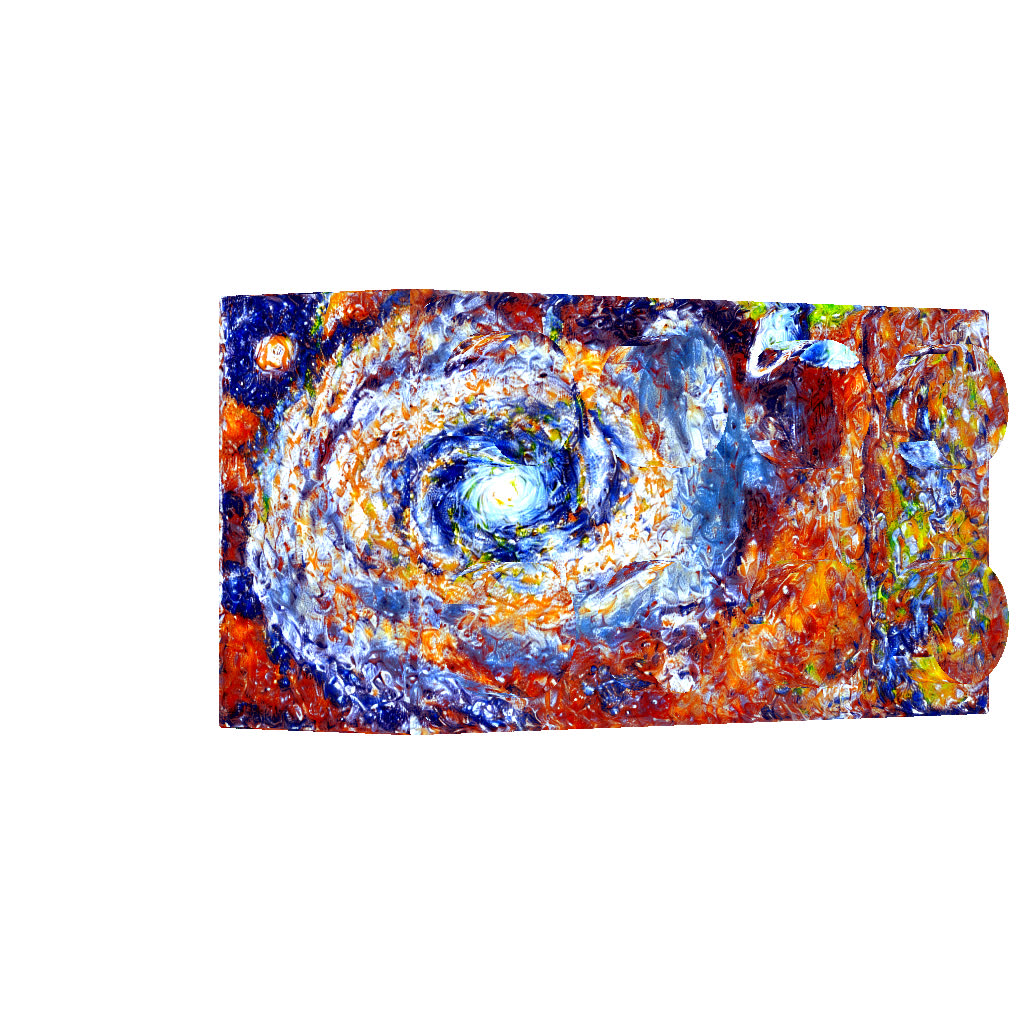}
    \end{minipage}%
    \begin{minipage}{0.16\textwidth}
        \includegraphics[width=\linewidth, trim=140 0 140 350, clip]{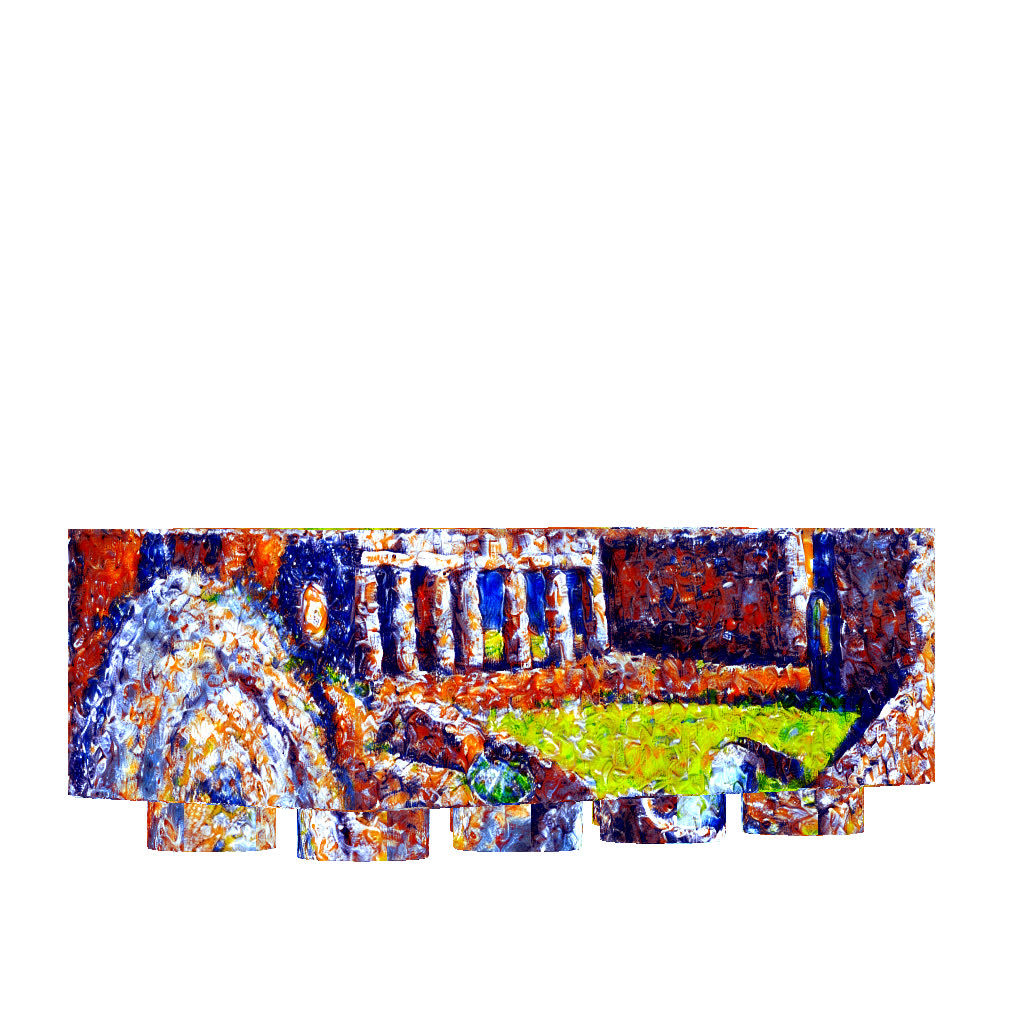}
    \end{minipage} \\

    \vspace{-2mm}
        \begin{minipage}[t]{0.16\textwidth}
        \centering
            \includegraphics[width=\linewidth, trim=185 15 185 30, clip]{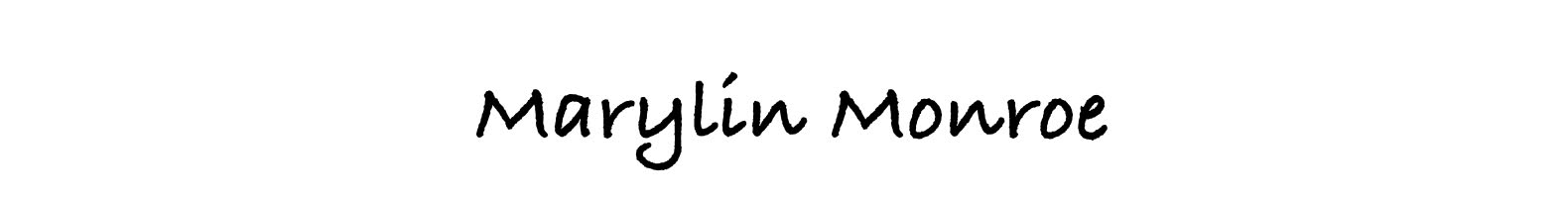}
            \end{minipage}\hfill
        \begin{minipage}[t]{0.16\textwidth}
            \centering
            \includegraphics[width=\linewidth, trim=185 20 185 30, clip]{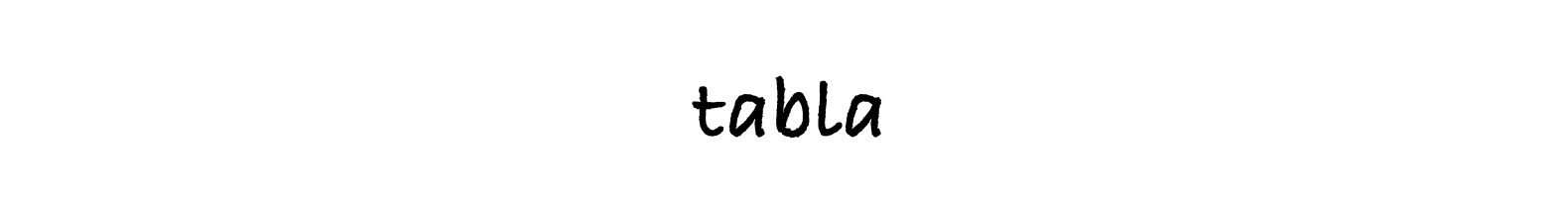}
            \end{minipage}\hfill
        \begin{minipage}[t]{0.16\textwidth}
            \centering
            \includegraphics[width=\linewidth, trim=185 20 185 30, clip]{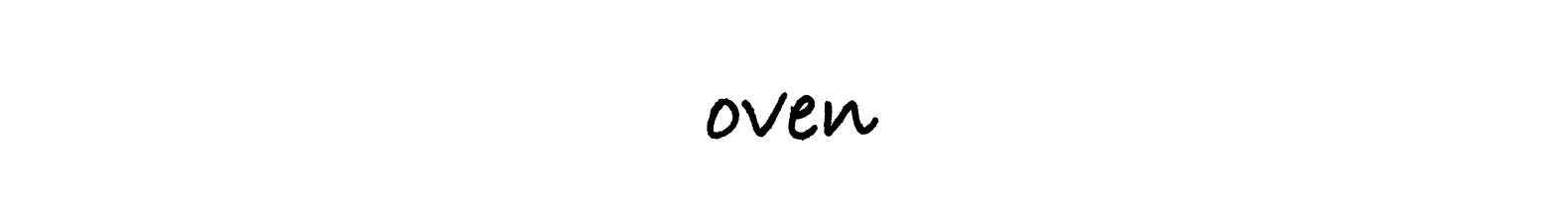}
        \end{minipage}\hfill
        \begin{minipage}[t]{0.16\textwidth}
            \centering
            \includegraphics[width=\linewidth, trim=185 20 185 30, clip]{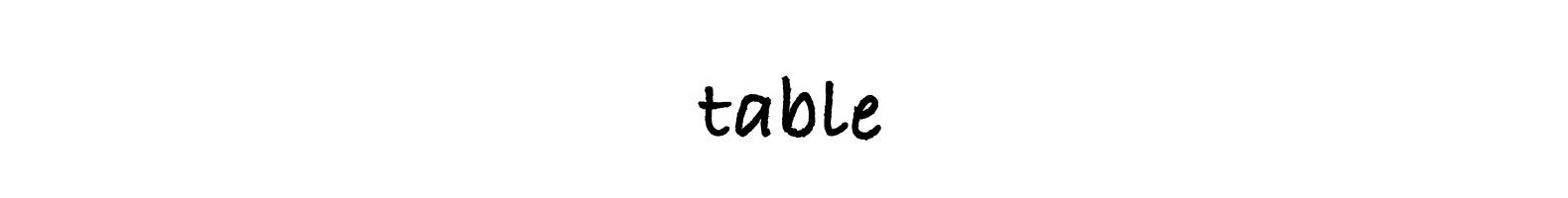}
        \end{minipage}
        \begin{minipage}[t]{0.16\textwidth}
            \centering
            \includegraphics[width=\linewidth, trim=185 15 185 30, clip]{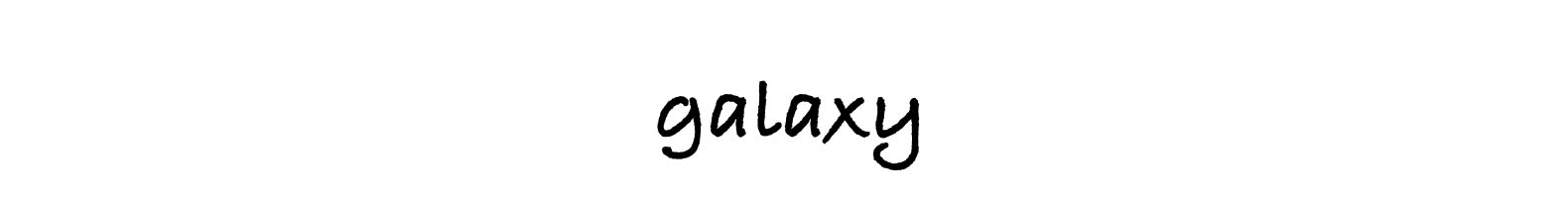}
        \end{minipage}
        \begin{minipage}[t]{0.16\textwidth}
            \centering
            \includegraphics[width=\linewidth, trim=185 20 185 30, clip]{figures/prompt/ar.jpg}
        \end{minipage}\\

    \end{tabular}
    \vspace{-1mm}
    
    \caption{\textbf{Random samples of Lego toy.} We present more random examples on a lego toy (convex surface). Styles (left to right): ink drawing, pencil sketch, painting, watercolor, oil painting($\times$4).}
    \label{fig:randomsamplelego}
    \vspace{-2mm}
\end{figure*}

\begin{figure*}[htbp]
    \centering
    \setlength{\tabcolsep}{1pt} 
    \renewcommand{\arraystretch}{1}

    \begin{tabular}{cccccc}
    
        \begin{minipage}{0.16\textwidth}
        \includegraphics[width=\linewidth, trim=140 140 140 140, clip]{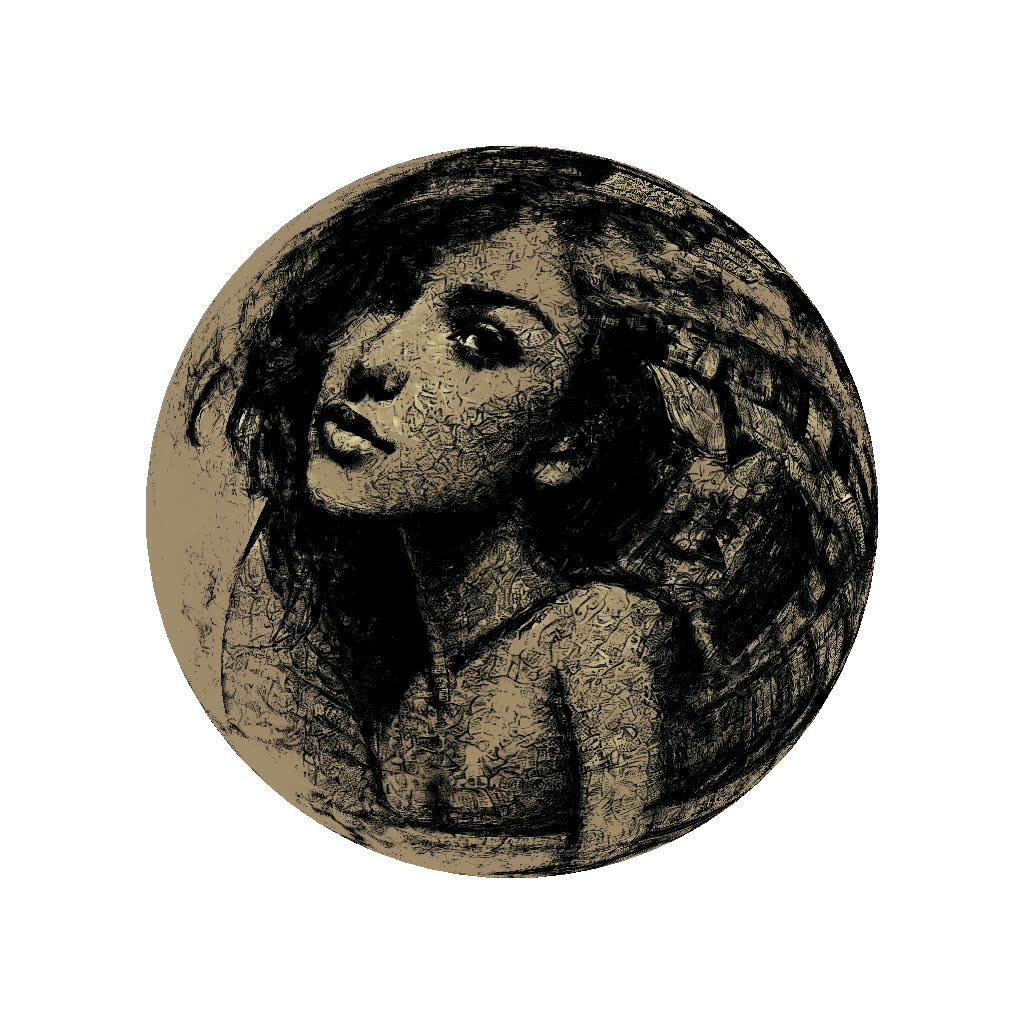}
    \end{minipage}%
    \begin{minipage}{0.16\textwidth}
        \includegraphics[width=\linewidth, trim=140 140 140 140, clip]{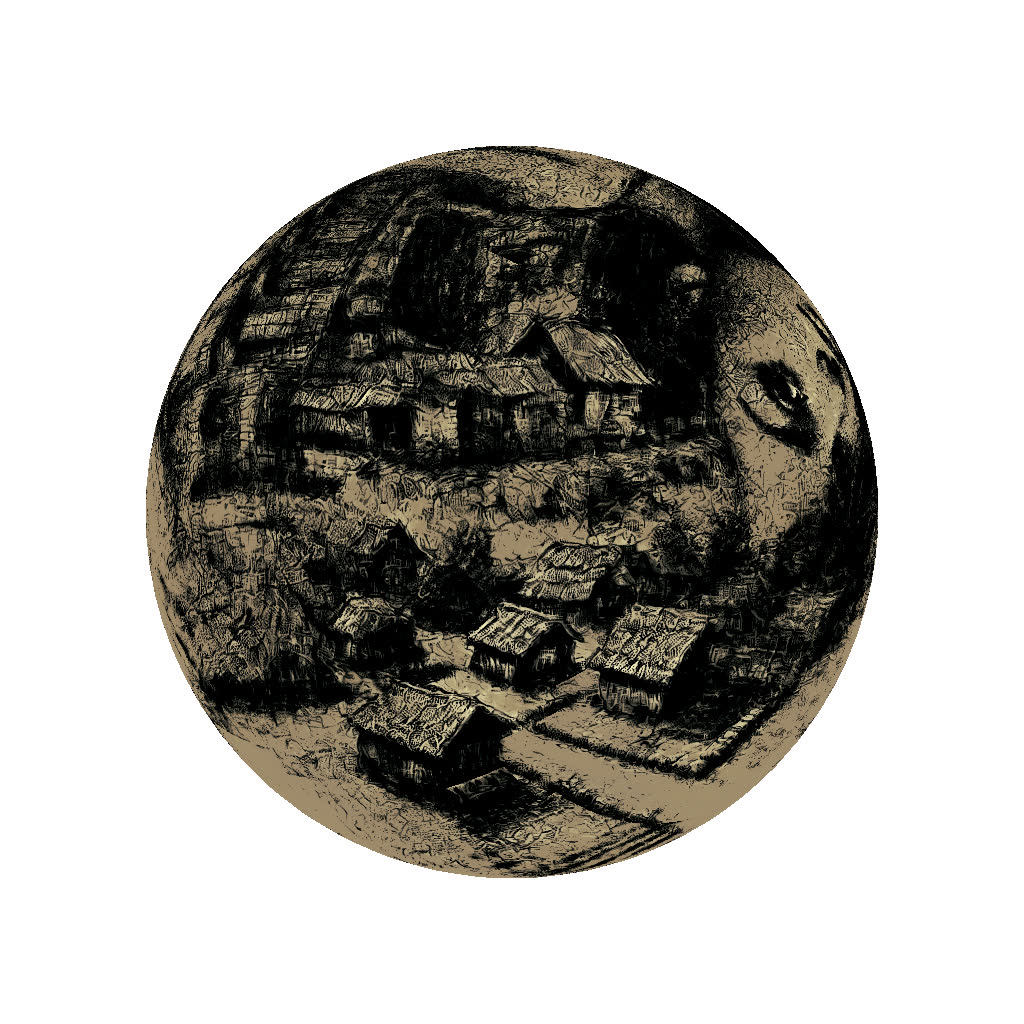}
    \end{minipage}%
    \begin{minipage}{0.16\textwidth}
        \includegraphics[width=\linewidth, trim=140 140 140 140, clip]{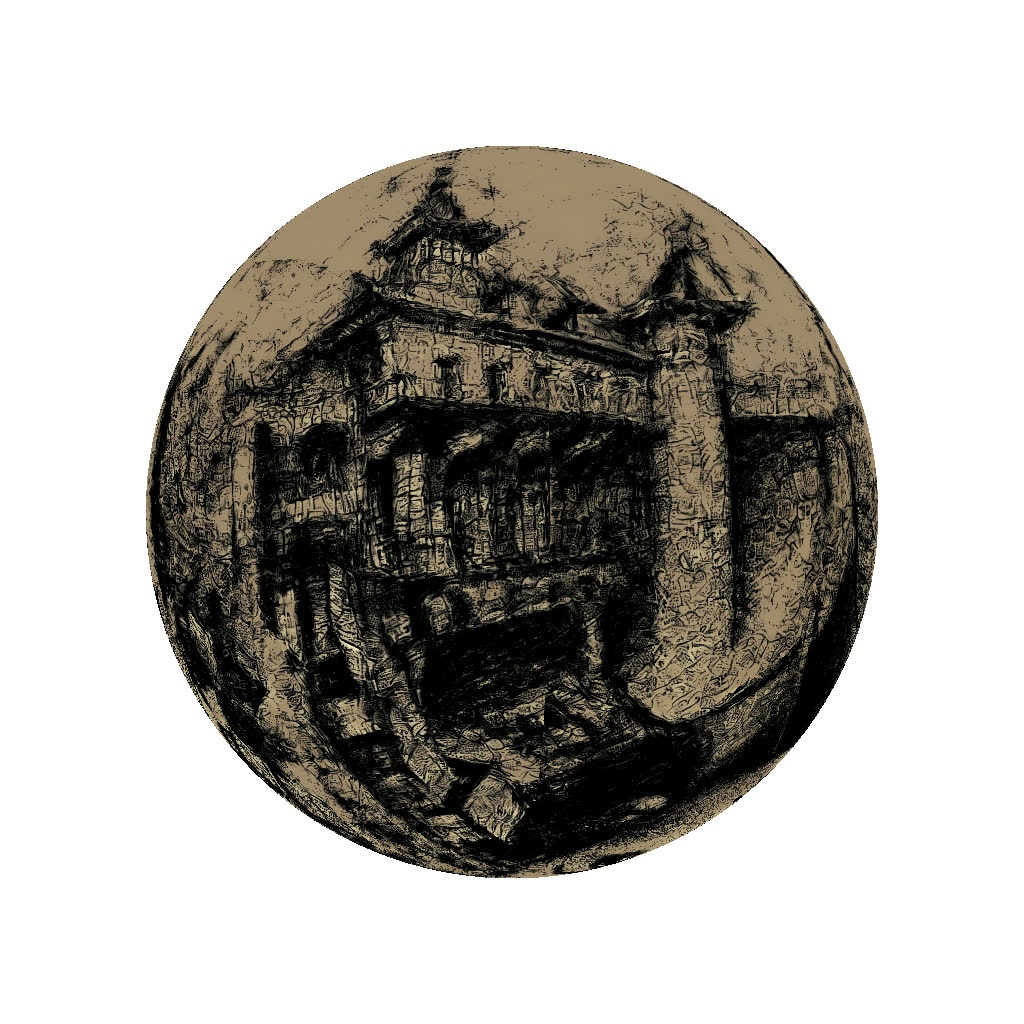}
    \end{minipage}%
    \begin{minipage}{0.16\textwidth}
        \includegraphics[width=\linewidth, trim=140 140 140 140, clip]{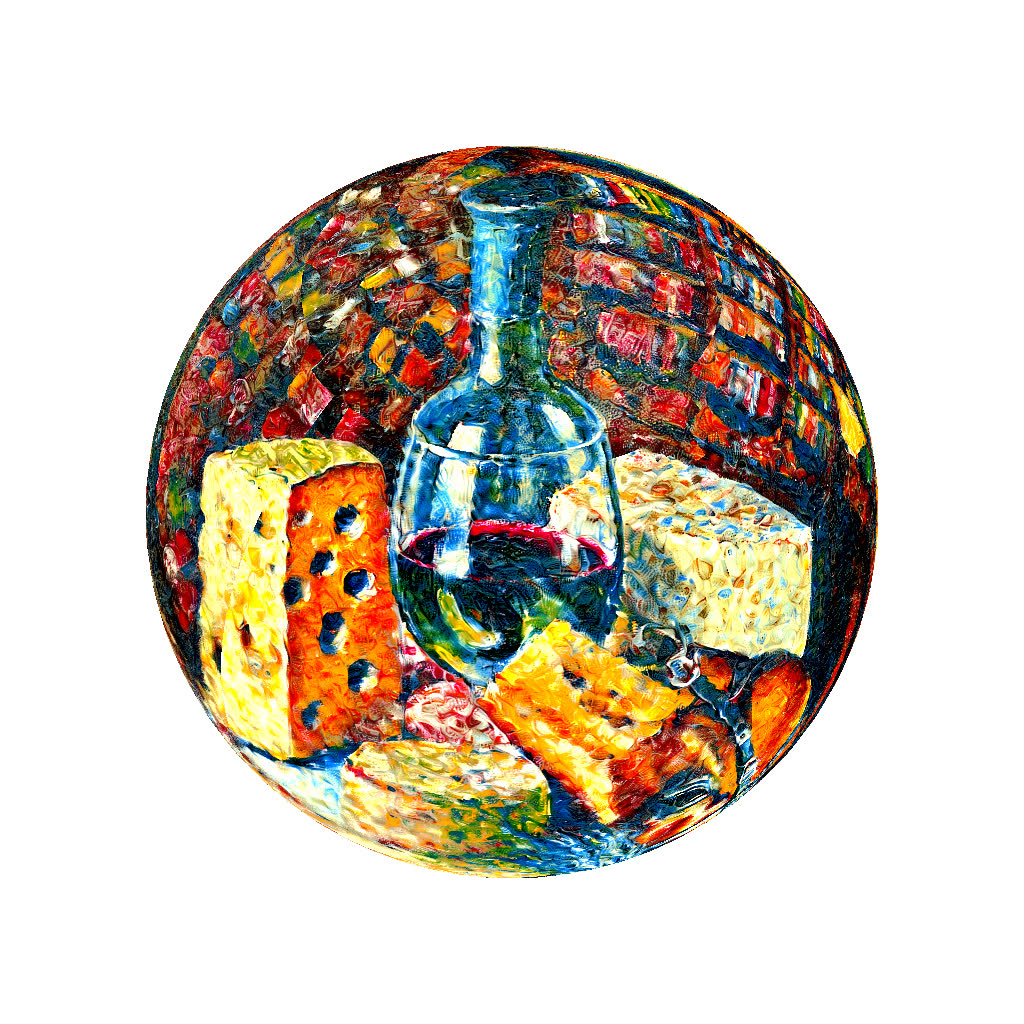}
    \end{minipage}%
    \begin{minipage}{0.16\textwidth}
        \includegraphics[width=\linewidth, trim=140 140 140 140, clip]{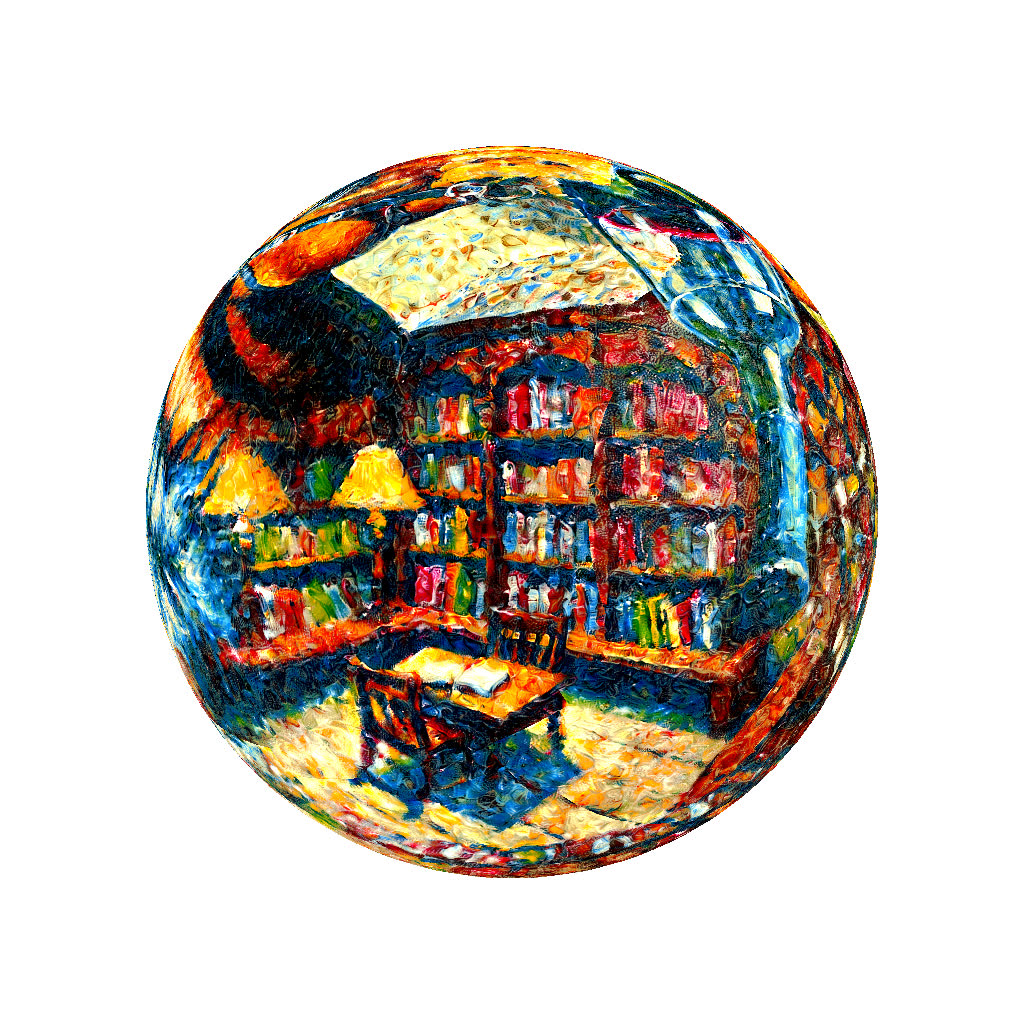}
    \end{minipage}%
    \begin{minipage}{0.16\textwidth}
        \includegraphics[width=\linewidth, trim=140 140 140 140, clip]{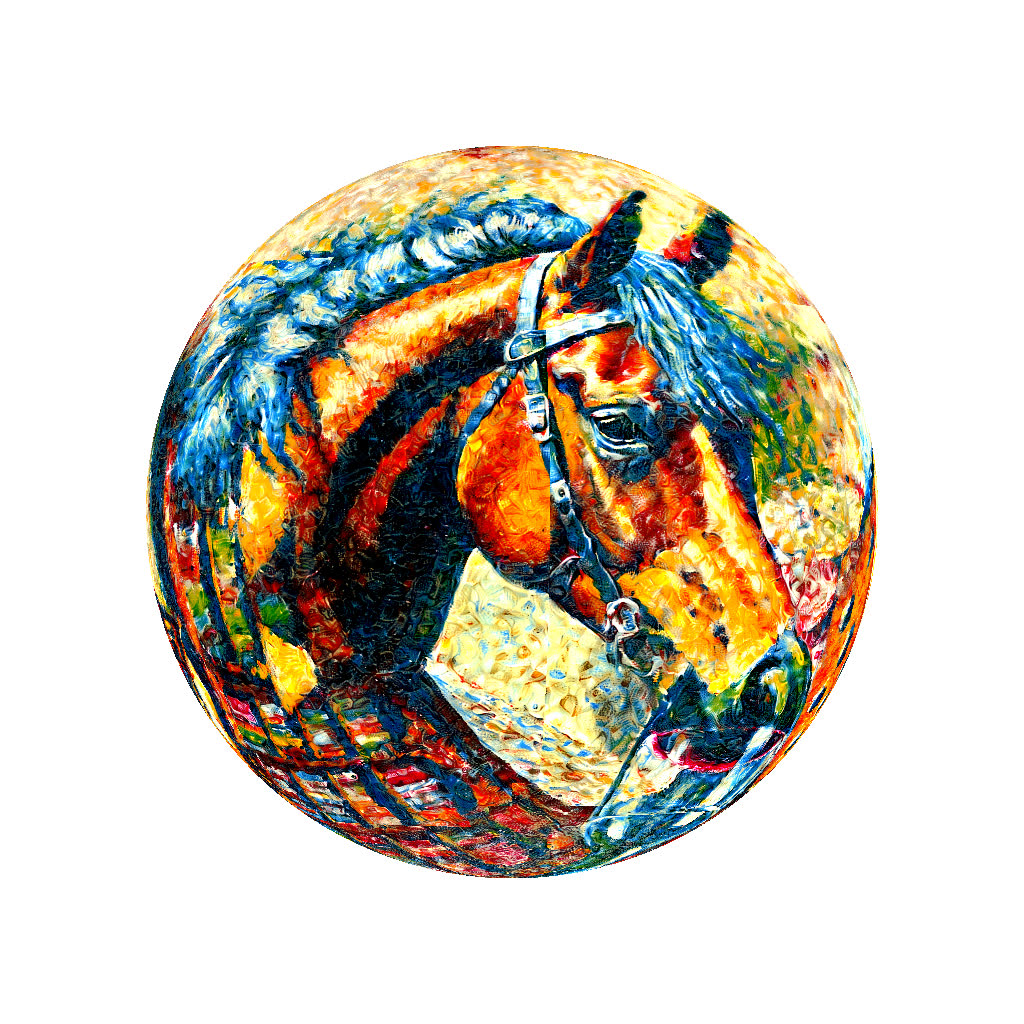}
    \end{minipage}\\
        \begin{minipage}[t]{0.16\textwidth}
        \centering
            \includegraphics[width=\linewidth, trim=185 15 185 30, clip]{figures/prompt/younglady.jpg}
            \end{minipage}\hfill
        \begin{minipage}[t]{0.16\textwidth}
            \centering
            \includegraphics[width=\linewidth, trim=185 15 185 30, clip]{figures/prompt/village.jpg}
            \end{minipage}\hfill
        \begin{minipage}[t]{0.16\textwidth}
            \centering
            \includegraphics[width=\linewidth, trim=185 20 185 30, clip]{figures/prompt/museum.jpg}
        \end{minipage}\hfill
        \begin{minipage}[t]{0.16\textwidth}
            \centering
            \includegraphics[width=\linewidth, trim=185 20 185 30, clip]{figures/prompt/wineandcheese.jpg}
        \end{minipage}
        \begin{minipage}[t]{0.16\textwidth}
            \centering
            \includegraphics[width=\linewidth, trim=165 15 165 30, clip]{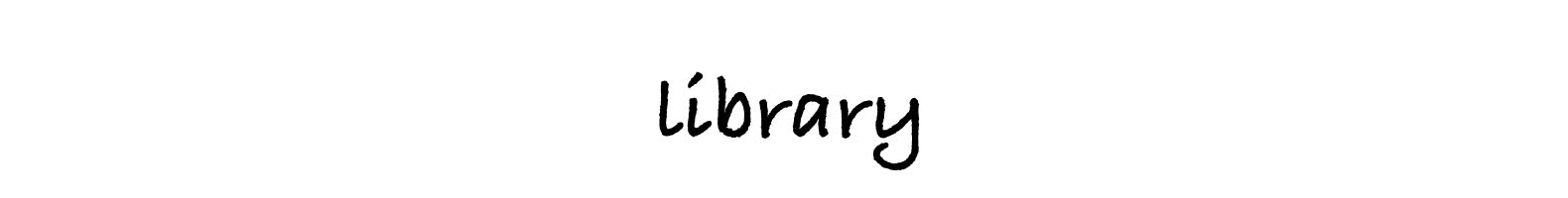}
        \end{minipage}
        \begin{minipage}[t]{0.16\textwidth}
            \centering
            \includegraphics[width=\linewidth, trim=185 15 185 30, clip]{figures/prompt/horse.jpg}
        \end{minipage}\\
        \begin{minipage}[t]{0.16\textwidth}
            \includegraphics[width=\textwidth, trim=140 140 140 140, clip]{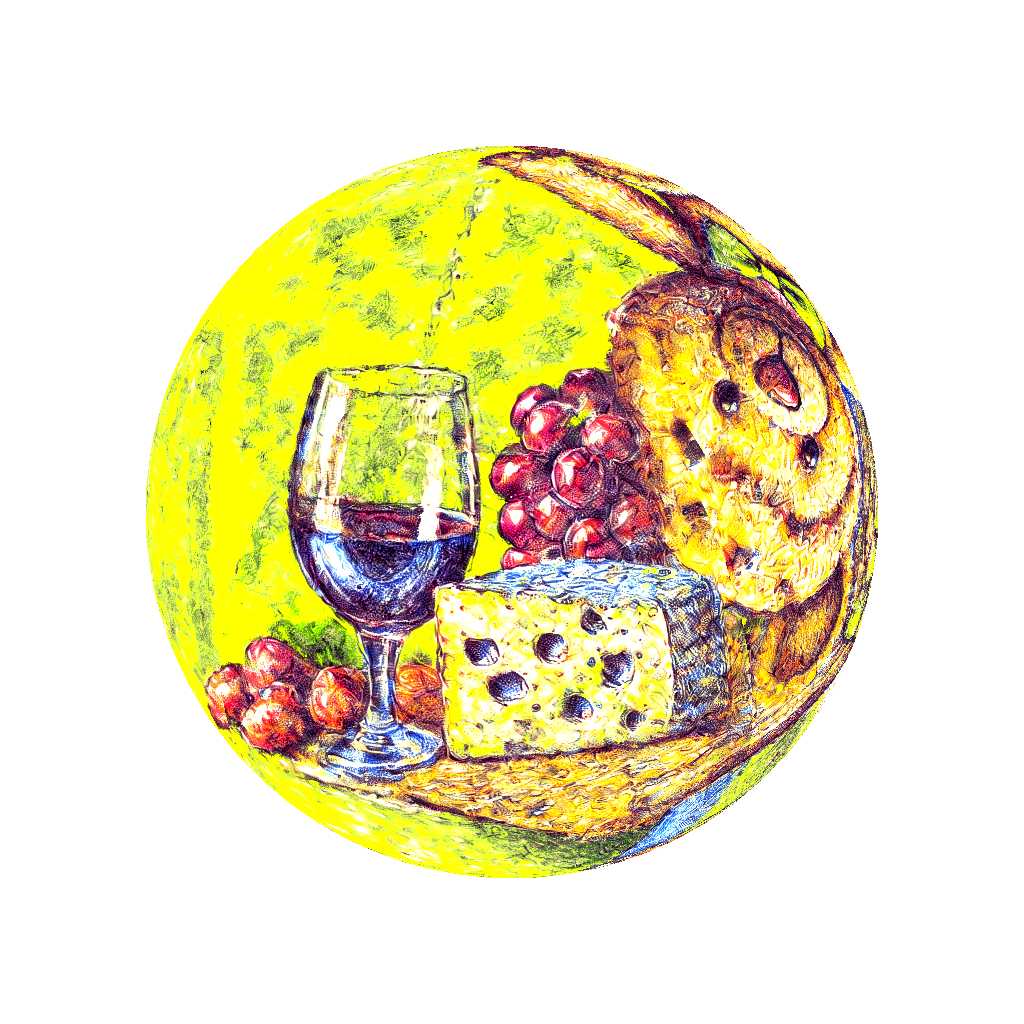}
        \end{minipage}%
        \begin{minipage}[t]{0.16\textwidth}
            \includegraphics[width=\textwidth, trim=140 140 140 140, clip]{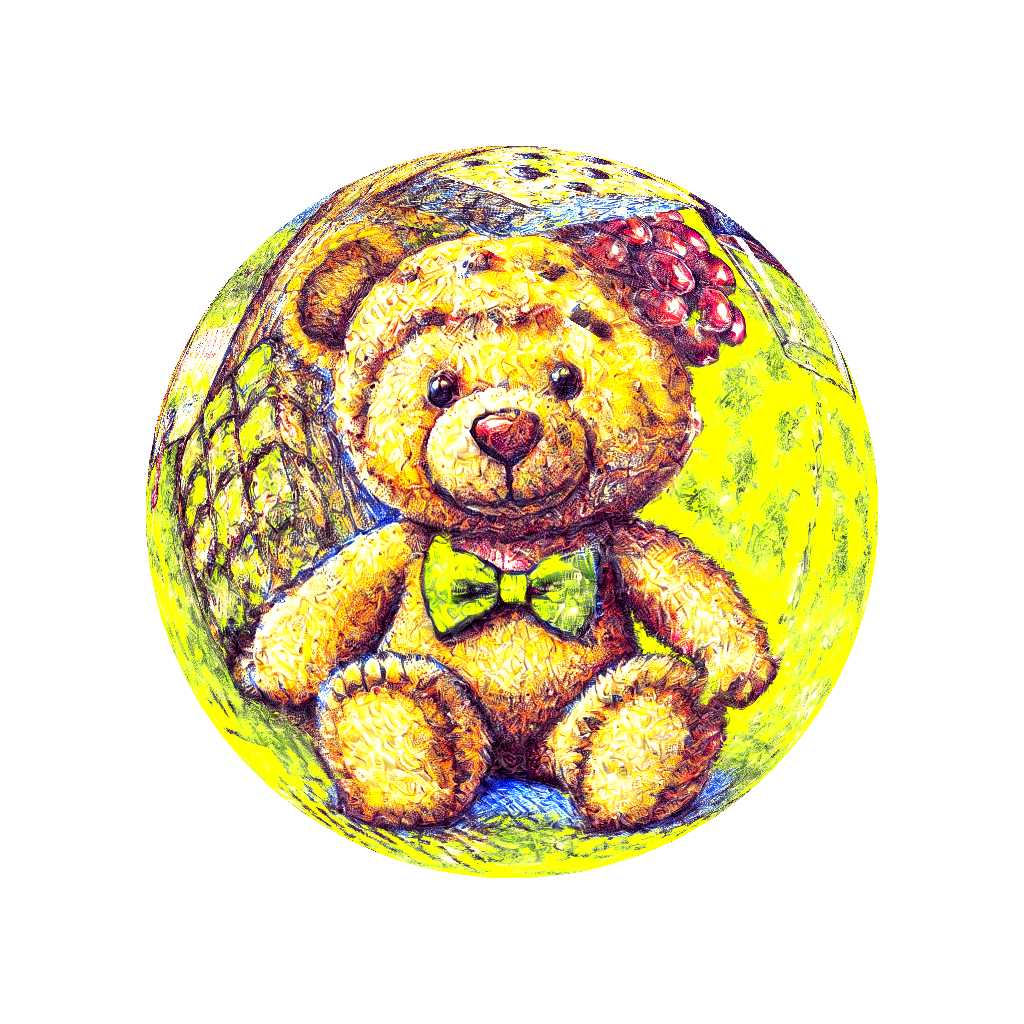}
        \end{minipage}%
        \begin{minipage}[t]{0.16\textwidth}
            \includegraphics[width=\textwidth, trim=140 140 140 140, clip]{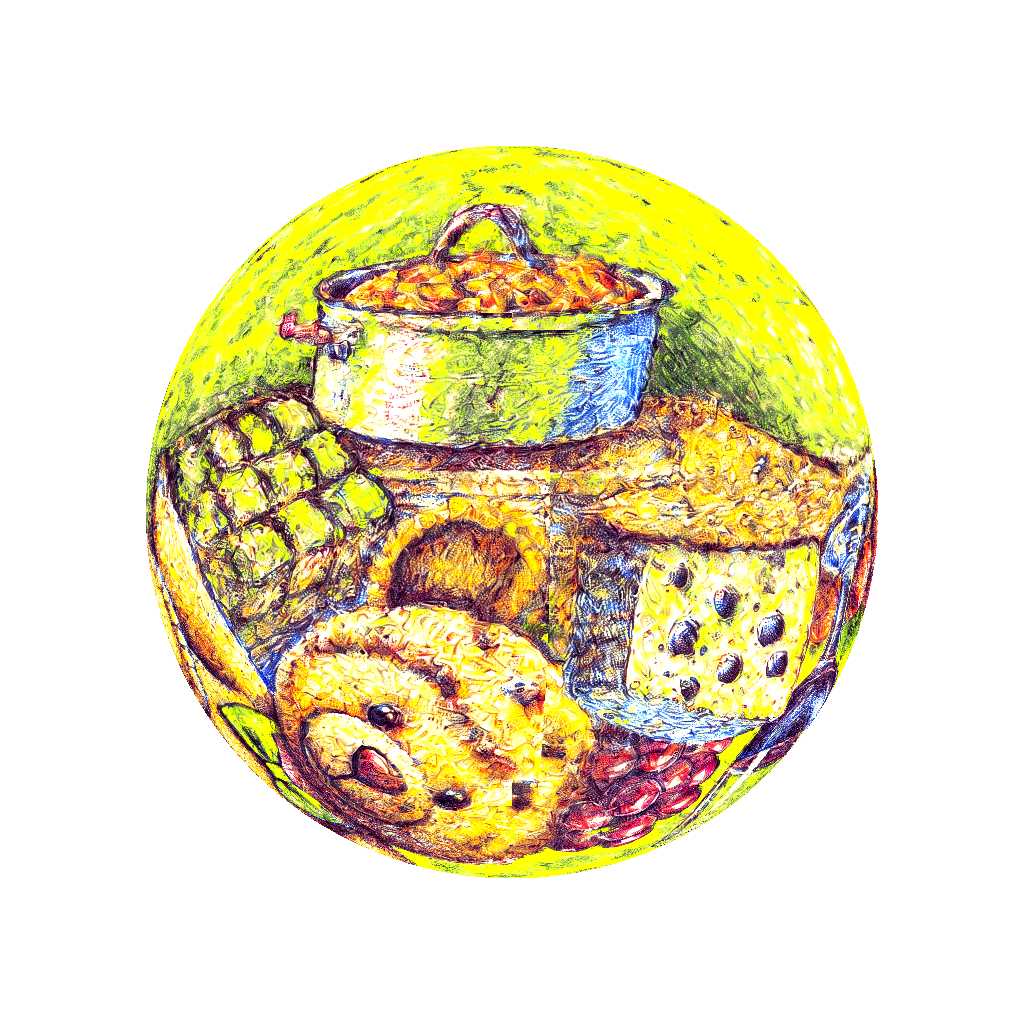}
        \end{minipage}%
        \begin{minipage}[t]{0.16\textwidth}
            \includegraphics[width=\textwidth, trim=140 140 140 140, clip]{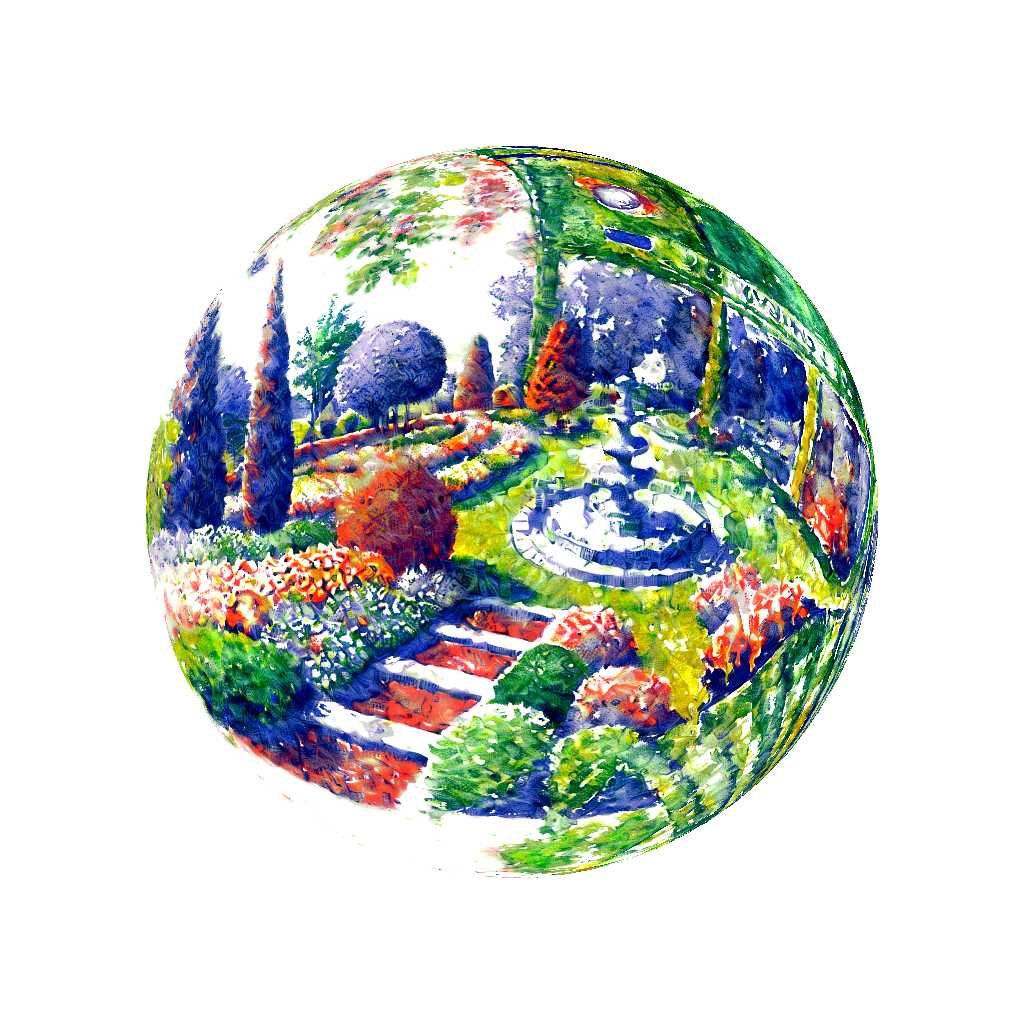}
        \end{minipage}%
        \begin{minipage}[t]{0.16\textwidth}
            \includegraphics[width=\textwidth, trim=140 140 140 140, clip]{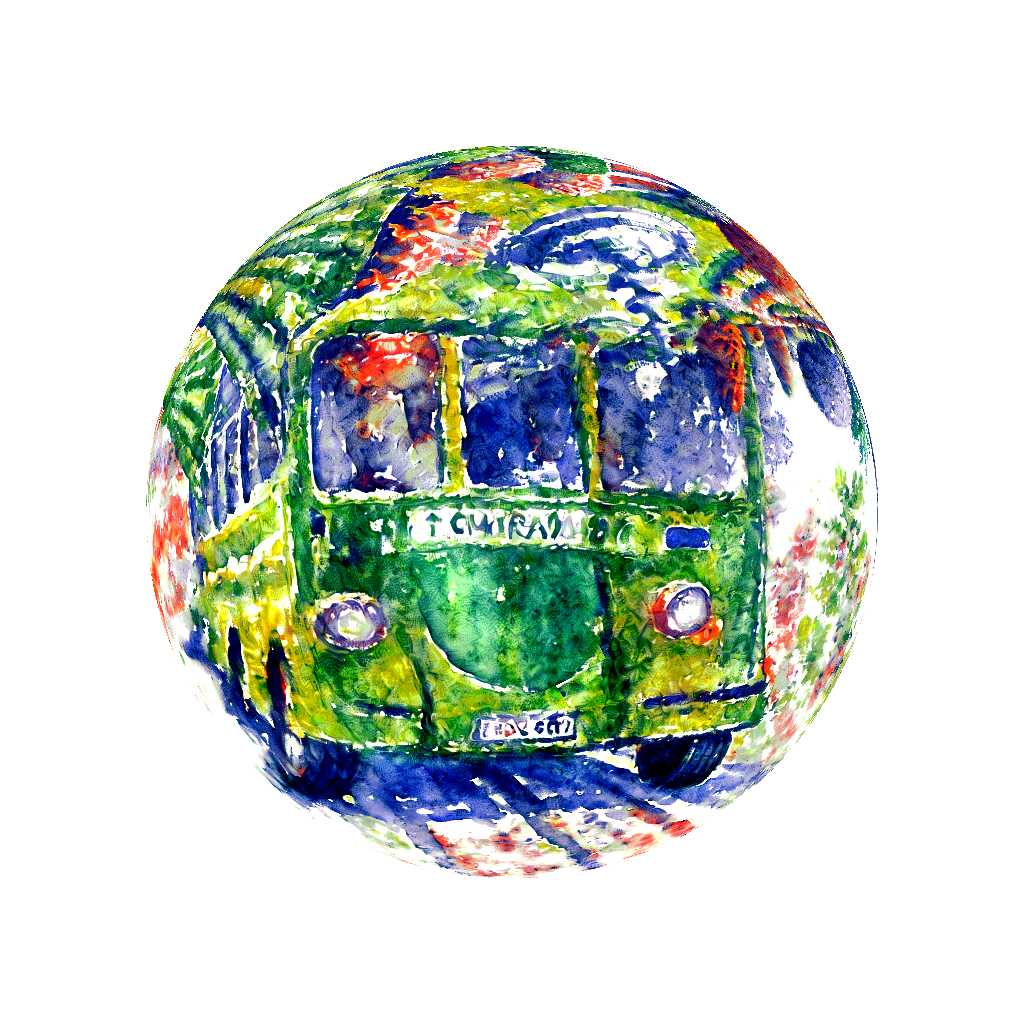}
        \end{minipage}%
        \begin{minipage}[t]{0.16\textwidth}
            \includegraphics[width=\textwidth, trim=140 140 140 140, clip]{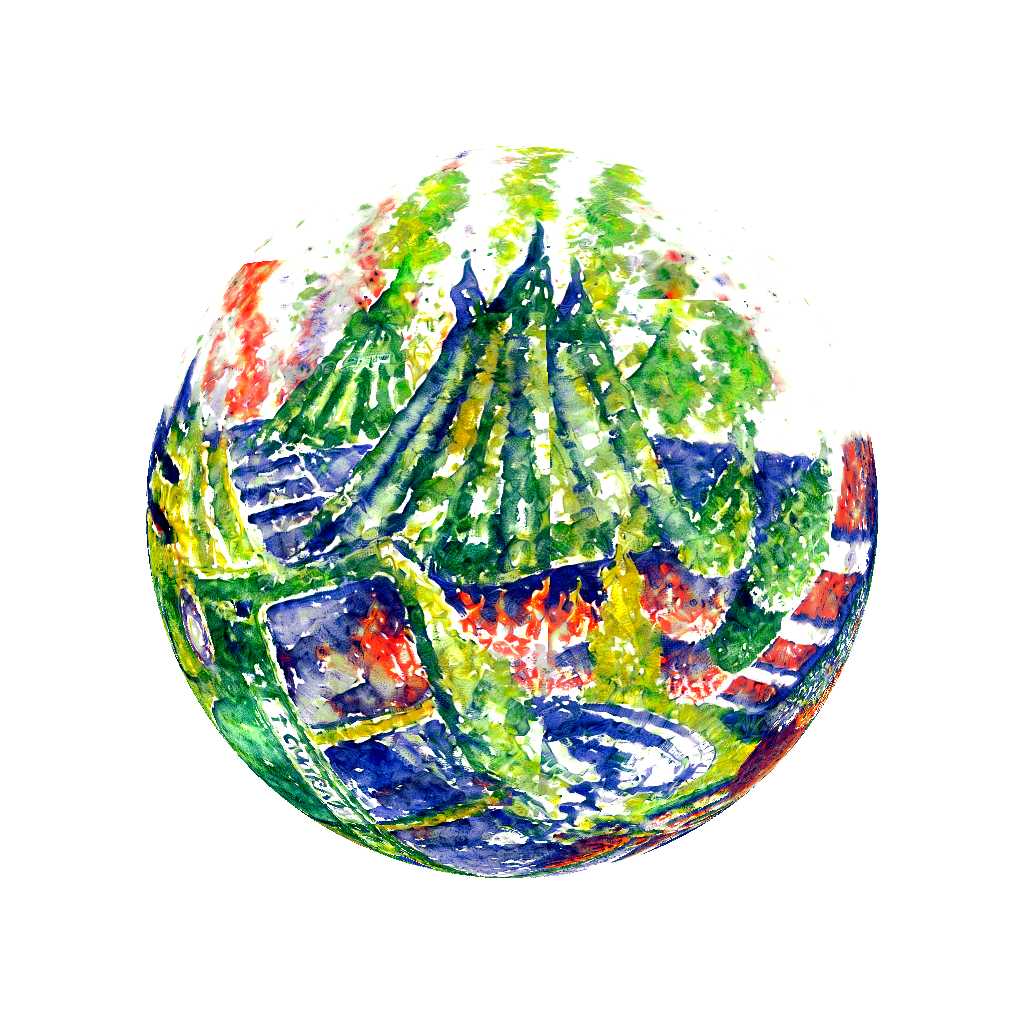}
        \end{minipage}\\

        \begin{minipage}[t]{0.16\textwidth}
        \centering
            \includegraphics[width=\linewidth, trim=185 20 185 30, clip]{figures/prompt/wineandcheese.jpg}
            \end{minipage}\hfill
        \begin{minipage}[t]{0.16\textwidth}
            \centering
            \includegraphics[width=\linewidth, trim=185 15 185 30, clip]{figures/prompt/teddybear.jpg}
            \end{minipage}\hfill
        \begin{minipage}[t]{0.16\textwidth}
            \centering
            \includegraphics[width=\linewidth, trim=185 20 185 30, clip]{figures/prompt/kitchenware.jpg}
        \end{minipage}\hfill
        \begin{minipage}[t]{0.16\textwidth}
            \centering
            \includegraphics[width=\linewidth, trim=185 20 185 30, clip]{figures/prompt/garden.jpg}
        \end{minipage}
        \begin{minipage}[t]{0.16\textwidth}
            \centering
            \includegraphics[width=\linewidth, trim=165 20 165 30, clip]{figures/prompt/bus.jpg}
        \end{minipage}
        \begin{minipage}[t]{0.16\textwidth}
            \centering
            \includegraphics[width=\linewidth, trim=185 20 185 30, clip]{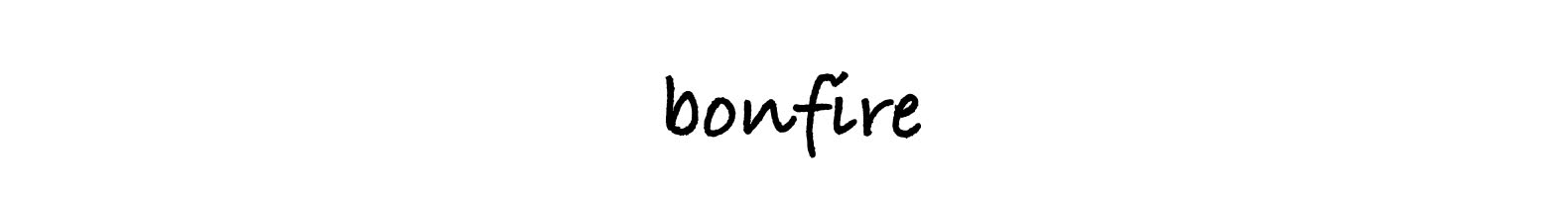}
        \end{minipage}\\      
    \begin{minipage}[t]{0.16\textwidth}
        \includegraphics[width=\textwidth, trim=140 140 140 140, clip]{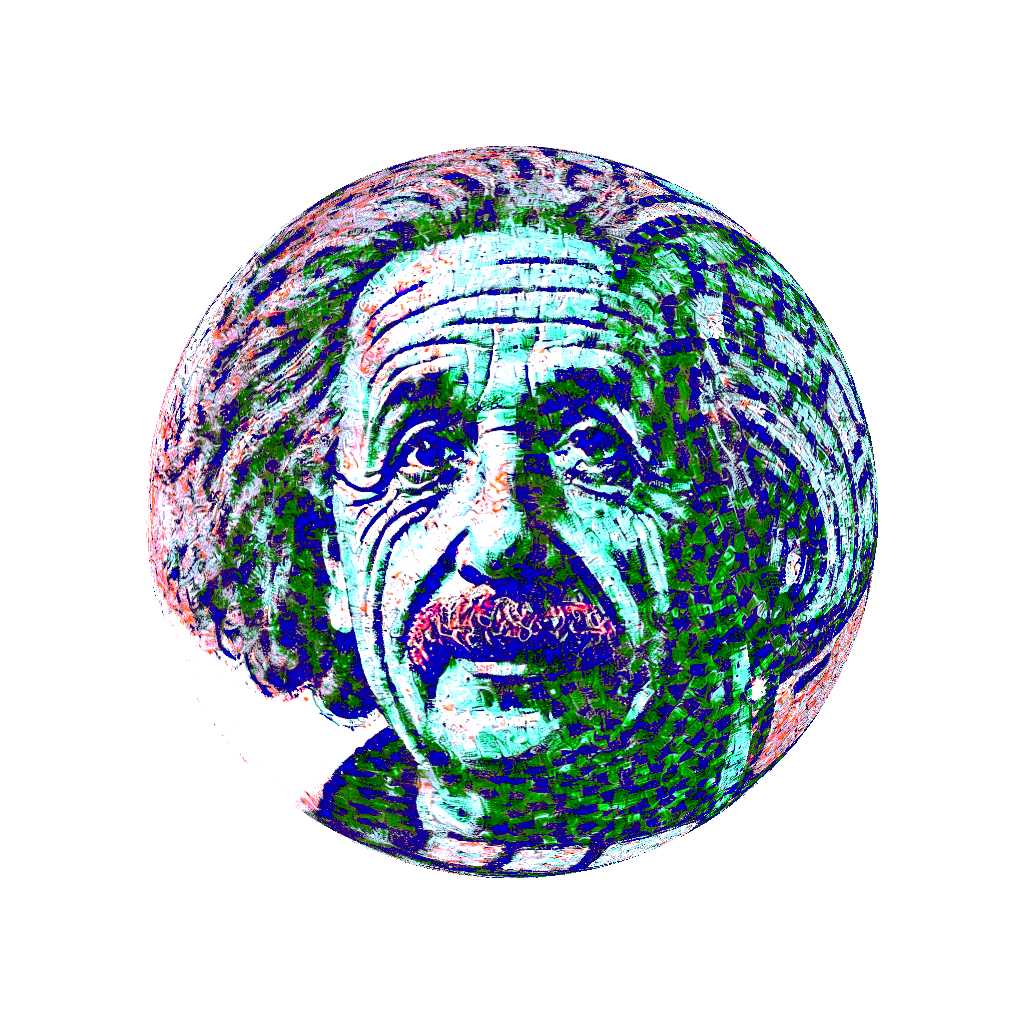}
    \end{minipage}%
    \begin{minipage}[t]{0.16\textwidth}
        \includegraphics[width=\textwidth, trim=140 140 140 140, clip]{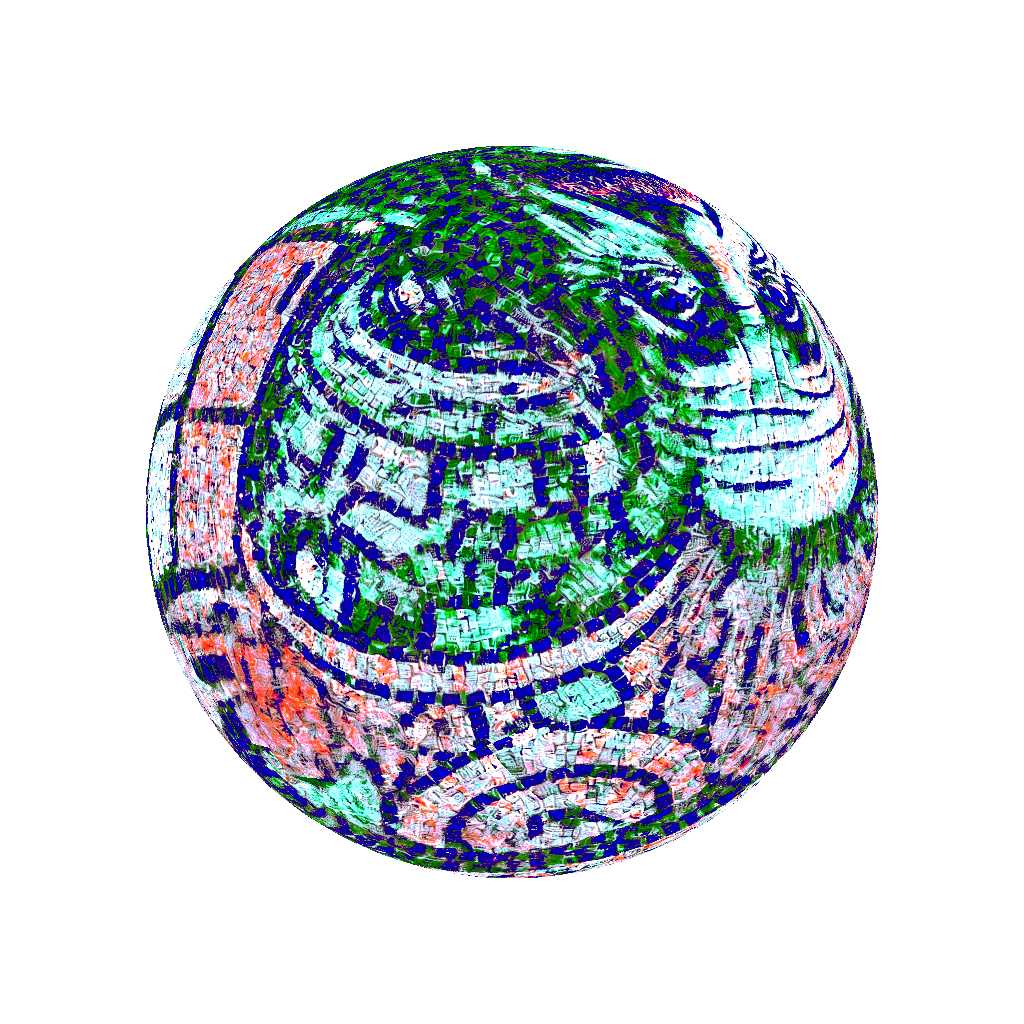}
    \end{minipage}%
    \begin{minipage}[t]{0.16\textwidth}
        \includegraphics[width=\textwidth, trim=140 140 140 140, clip]{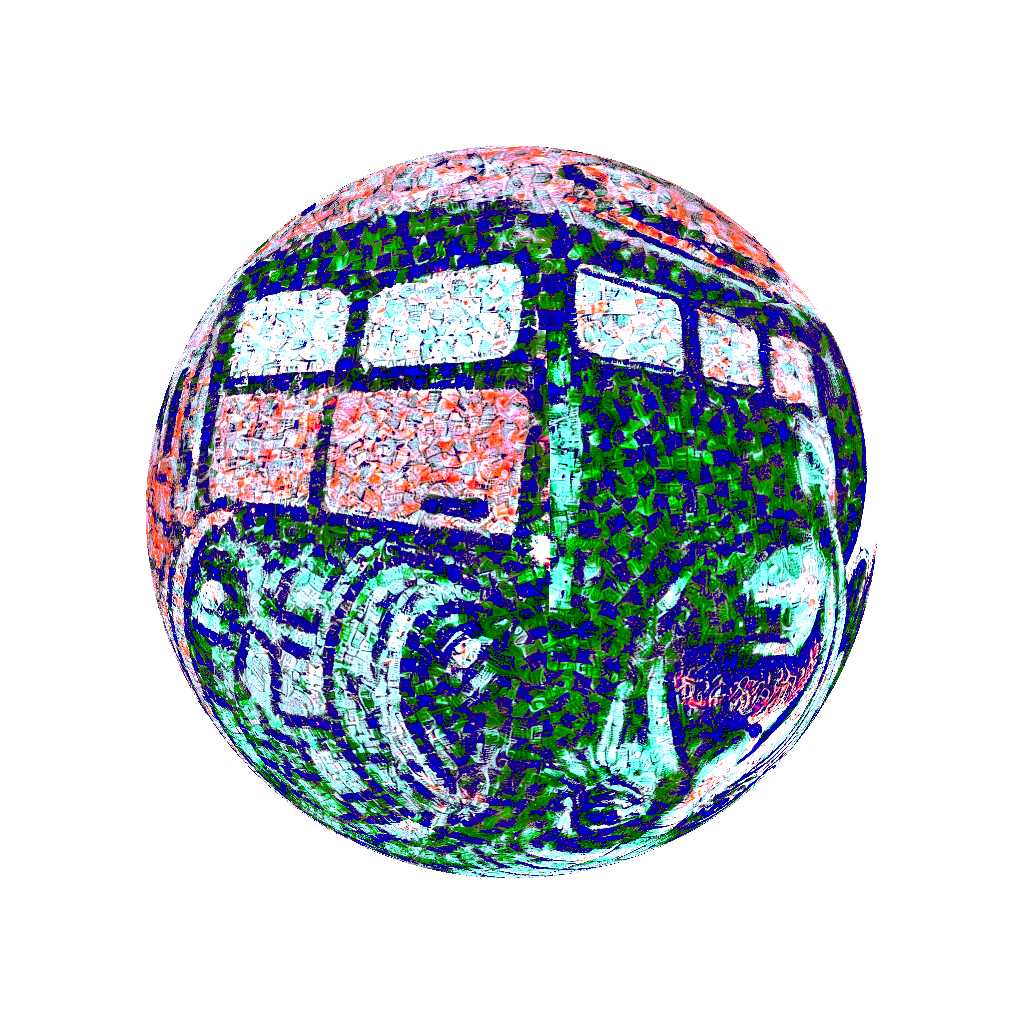}
    \end{minipage}%
    \begin{minipage}[t]{0.16\textwidth}
        \includegraphics[width=\textwidth, trim=140 140 140 140, clip]{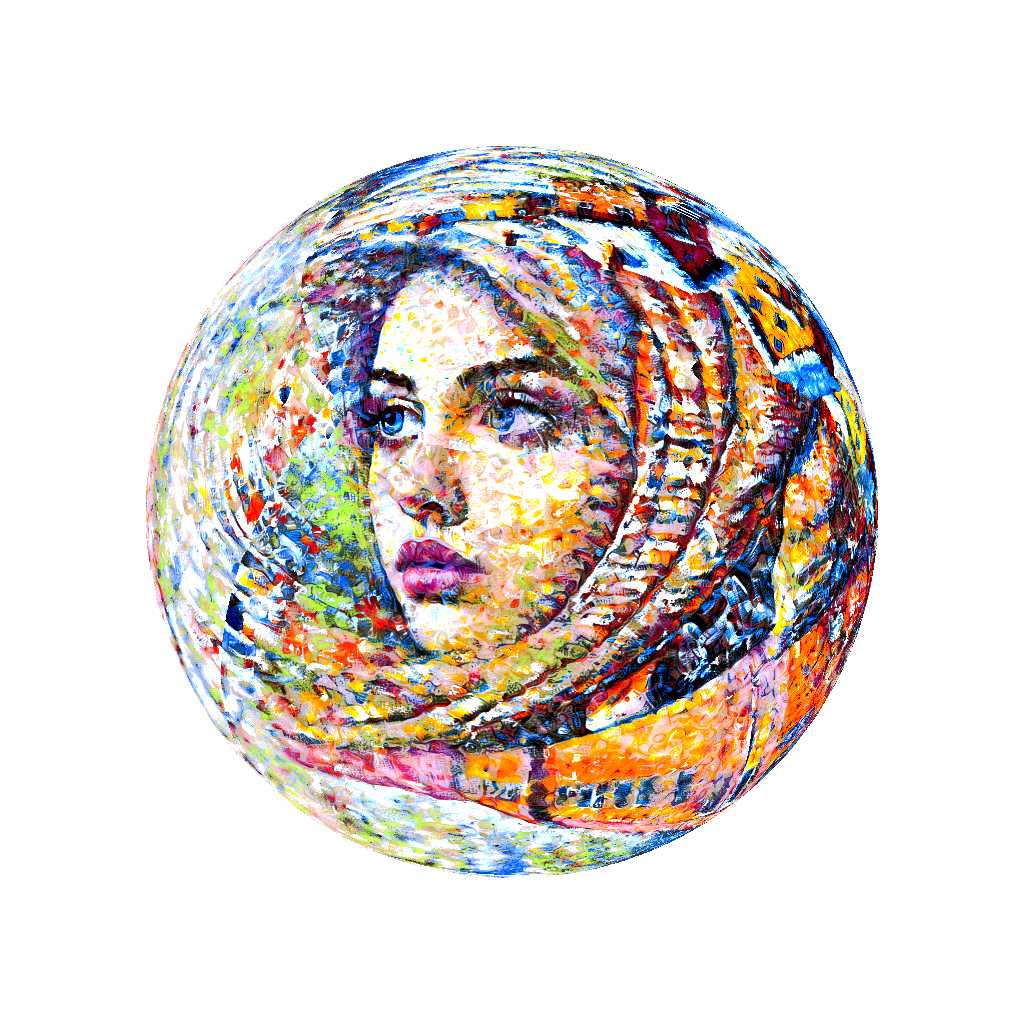}
    \end{minipage}%
    \begin{minipage}[t]{0.16\textwidth}
        \includegraphics[width=\textwidth, trim=140 140 140 140, clip]{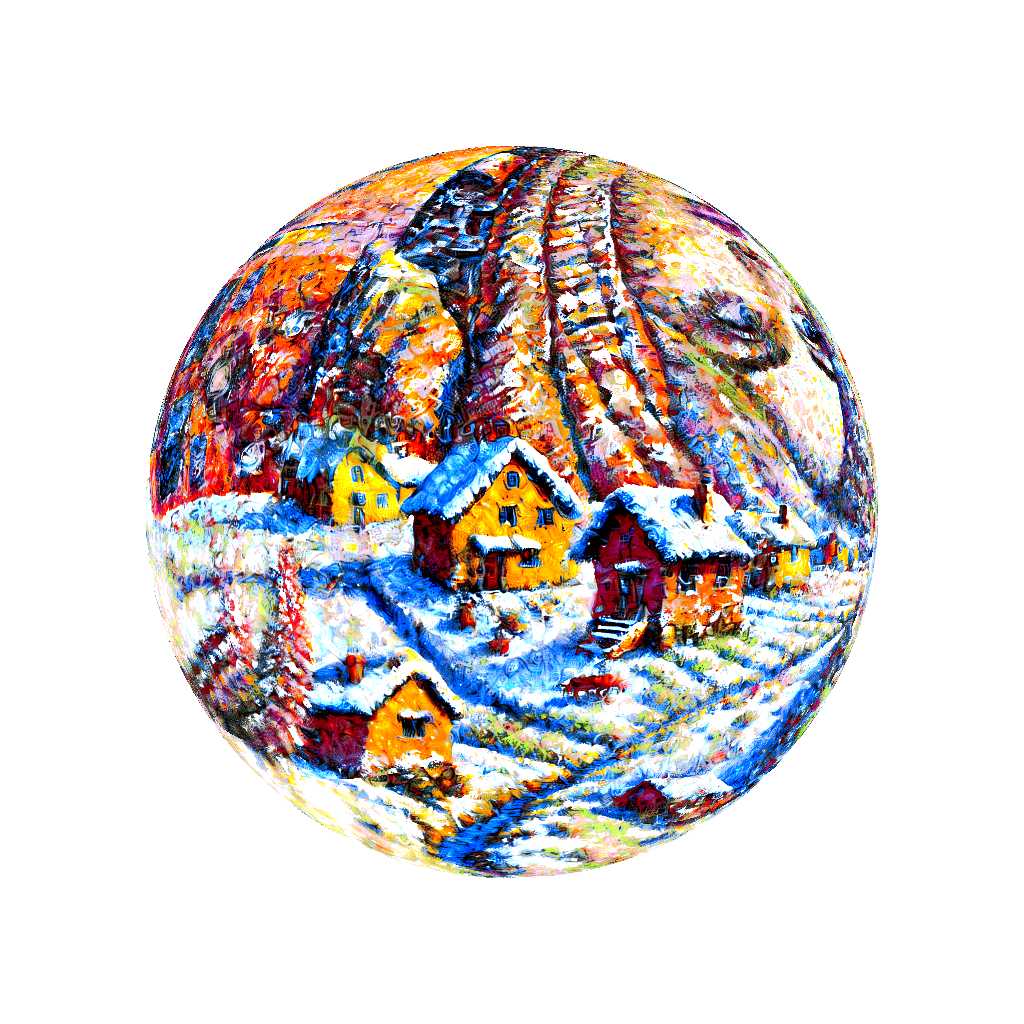}
    \end{minipage}%
    \begin{minipage}[t]{0.16\textwidth}
        \includegraphics[width=\textwidth, trim=140 140 140 140, clip]{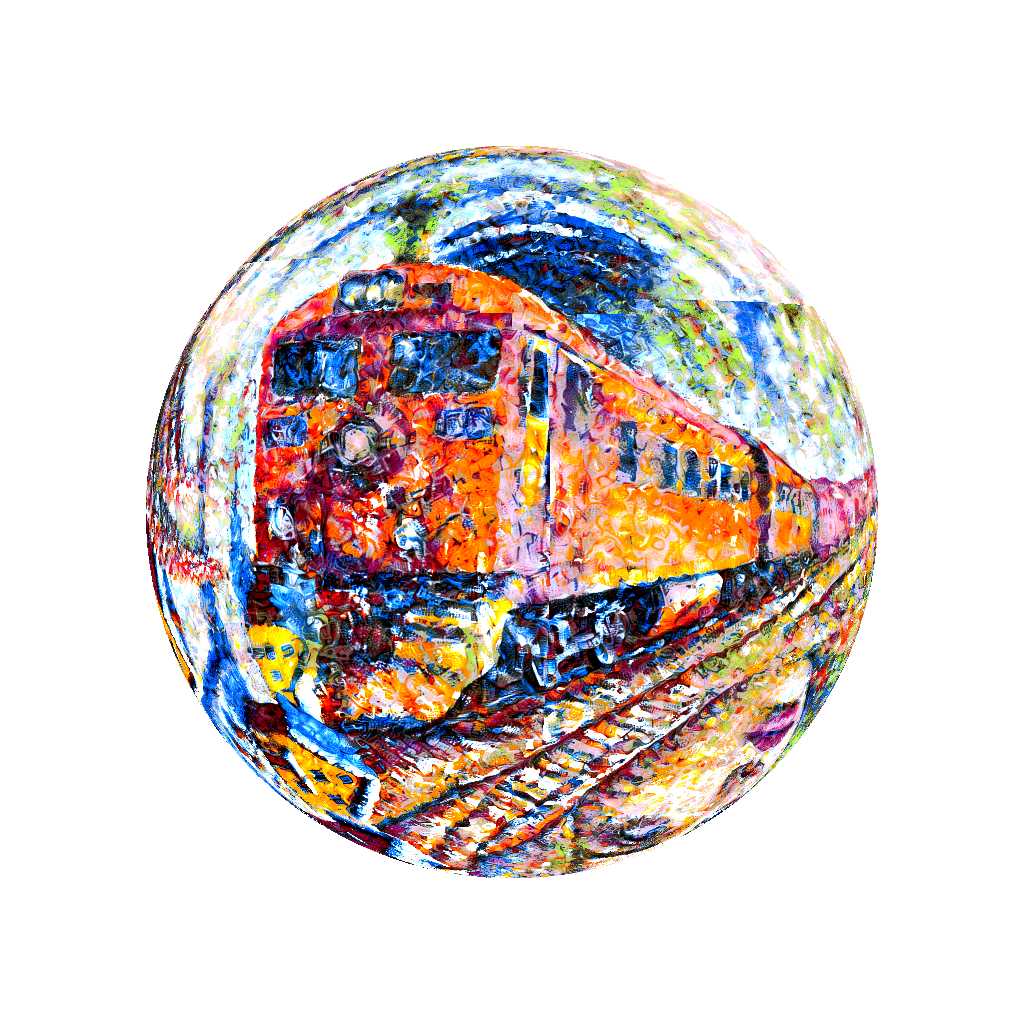}
    \end{minipage}\\
        \begin{minipage}[t]{0.16\textwidth}
        \centering
            \includegraphics[width=\linewidth, trim=185 20 185 30, clip]{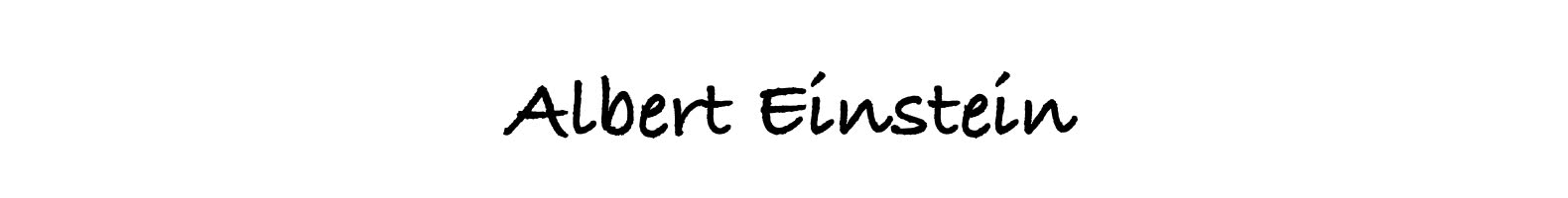}
            \end{minipage}\hfill
        \begin{minipage}[t]{0.16\textwidth}
            \centering
            \includegraphics[width=\linewidth, trim=185 20 185 30, clip]{figures/prompt/cup.jpg}
            \end{minipage}\hfill
        \begin{minipage}[t]{0.16\textwidth}
            \centering
            \includegraphics[width=\linewidth, trim=185 20 185 30, clip]{figures/prompt/bus.jpg}
        \end{minipage}\hfill
        \begin{minipage}[t]{0.16\textwidth}
            \centering
            \includegraphics[width=\linewidth, trim=185 15 185 30, clip]{figures/prompt/younglady.jpg}
        \end{minipage}
        \begin{minipage}[t]{0.16\textwidth}
            \centering
            \includegraphics[width=\linewidth, trim=165 20 165 30, clip]{figures/prompt/village.jpg}
        \end{minipage}
        \begin{minipage}[t]{0.16\textwidth}
            \centering
            \includegraphics[width=\linewidth, trim=185 20 185 30, clip]{figures/prompt/train.jpg}
        \end{minipage}\\

    \begin{minipage}{0.16\textwidth}
    \includegraphics[width=\linewidth, trim=140 140 140 140, clip]{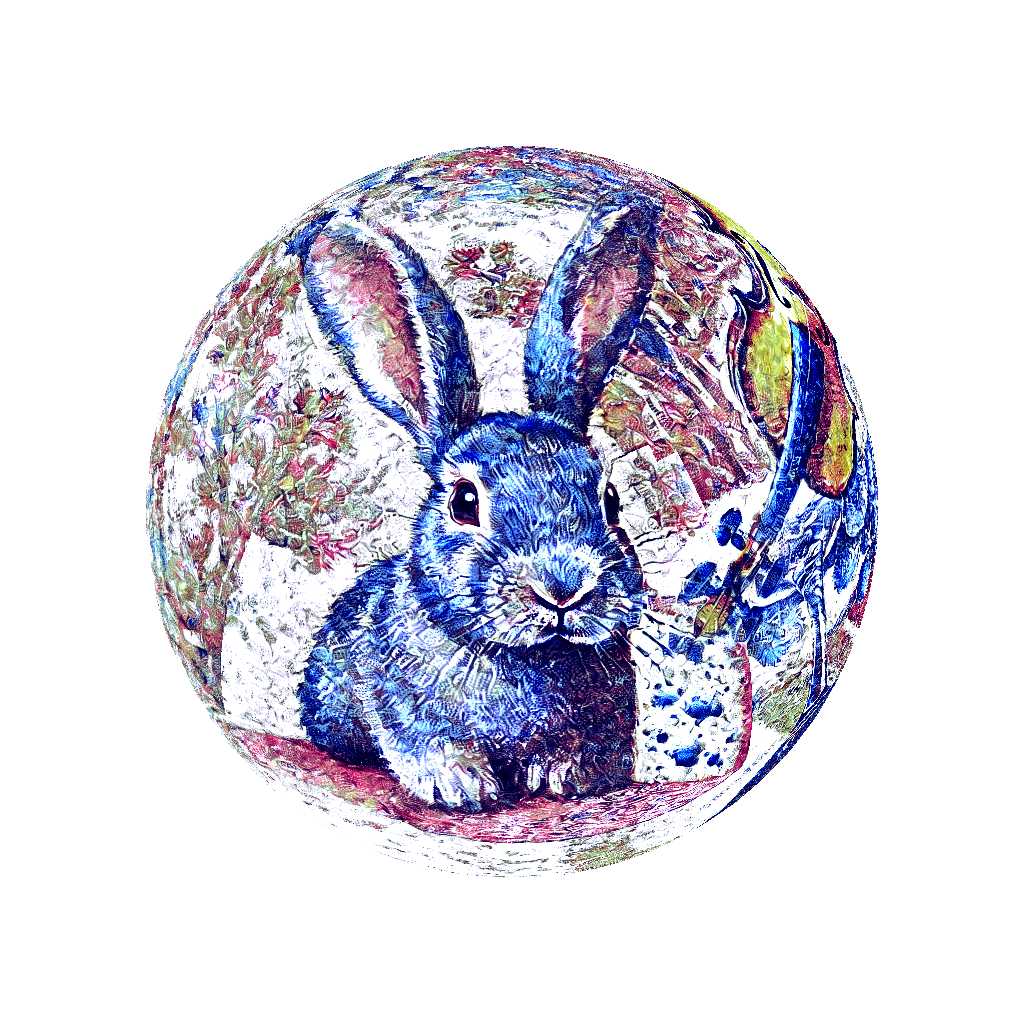}
    \end{minipage}%
    \begin{minipage}{0.16\textwidth}
        \includegraphics[width=\linewidth, trim=140 140 140 140, clip]{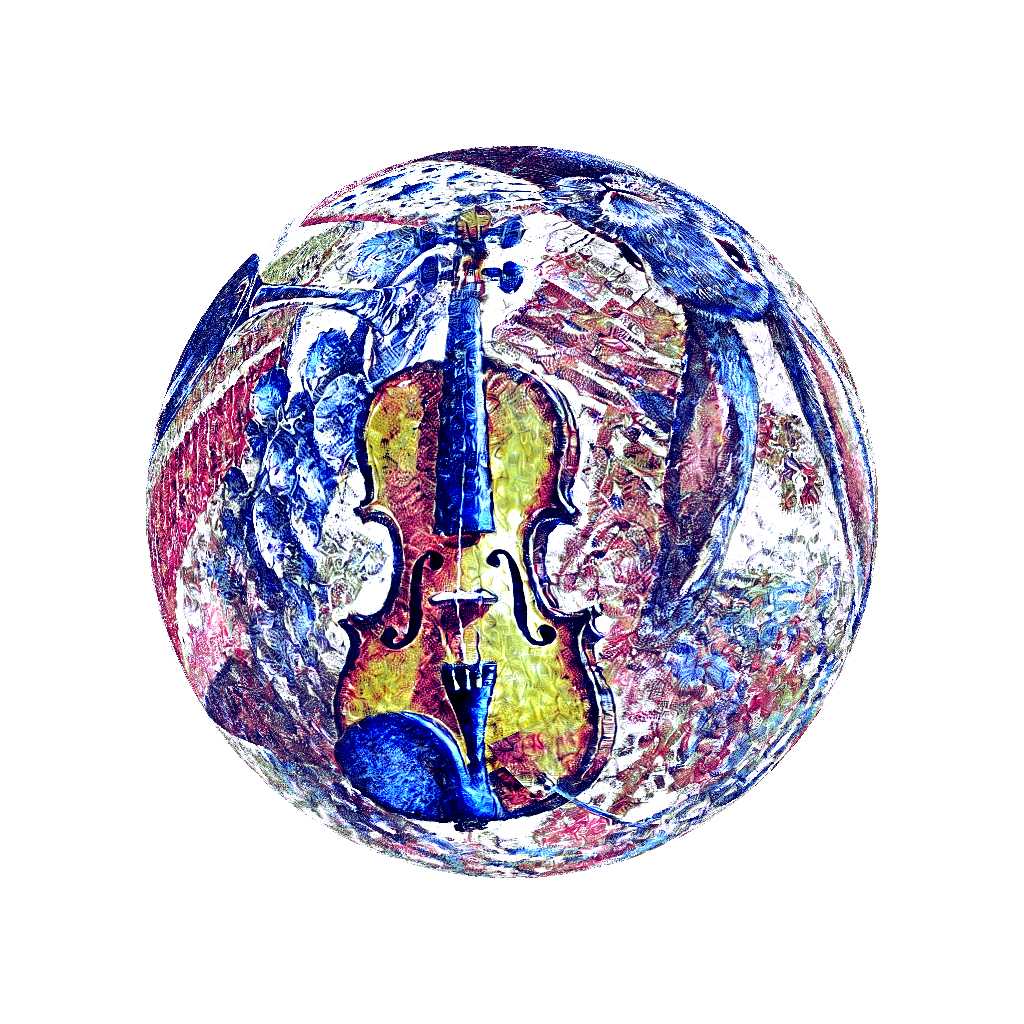}
    \end{minipage}%
    \begin{minipage}{0.16\textwidth}
        \includegraphics[width=\linewidth, trim=140 140 140 140, clip]{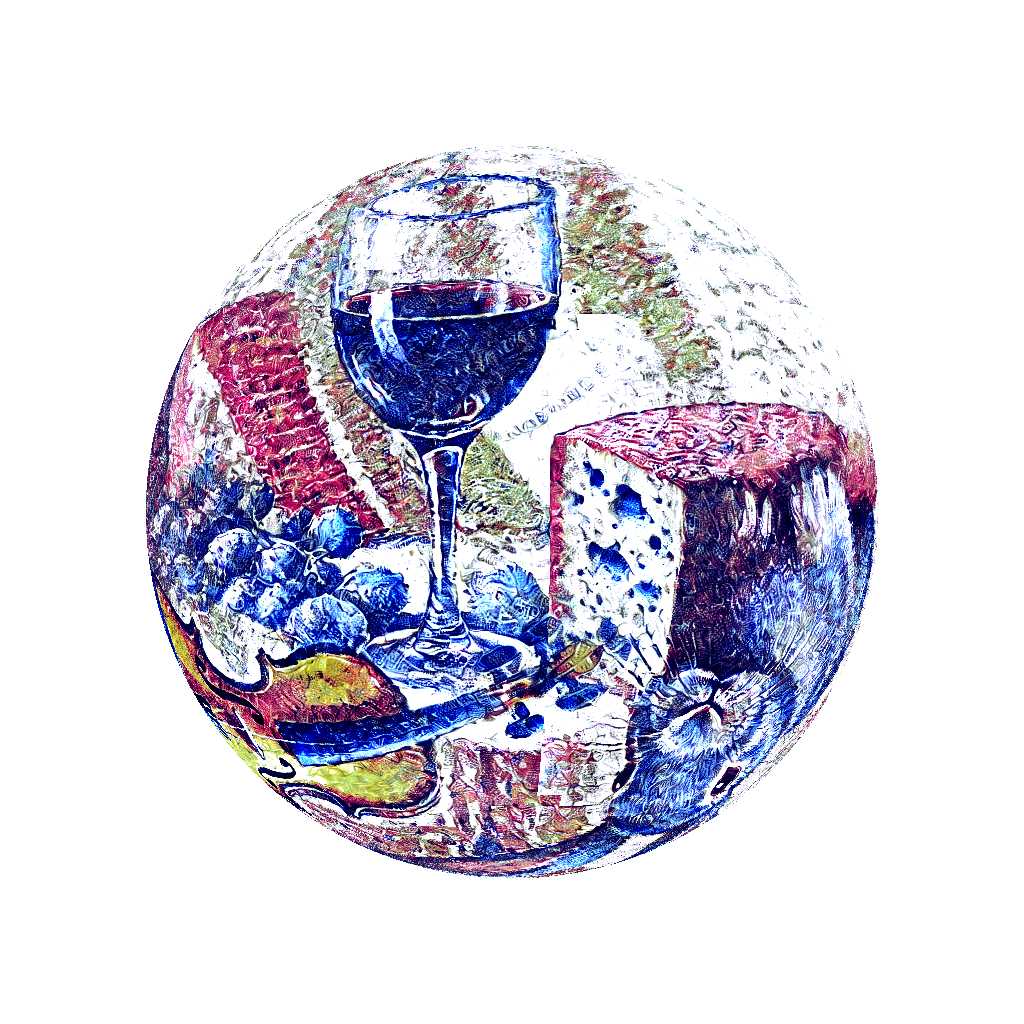}
    \end{minipage}%
    \begin{minipage}{0.16\textwidth}
        \includegraphics[width=\linewidth, trim=140 140 140 140, clip]{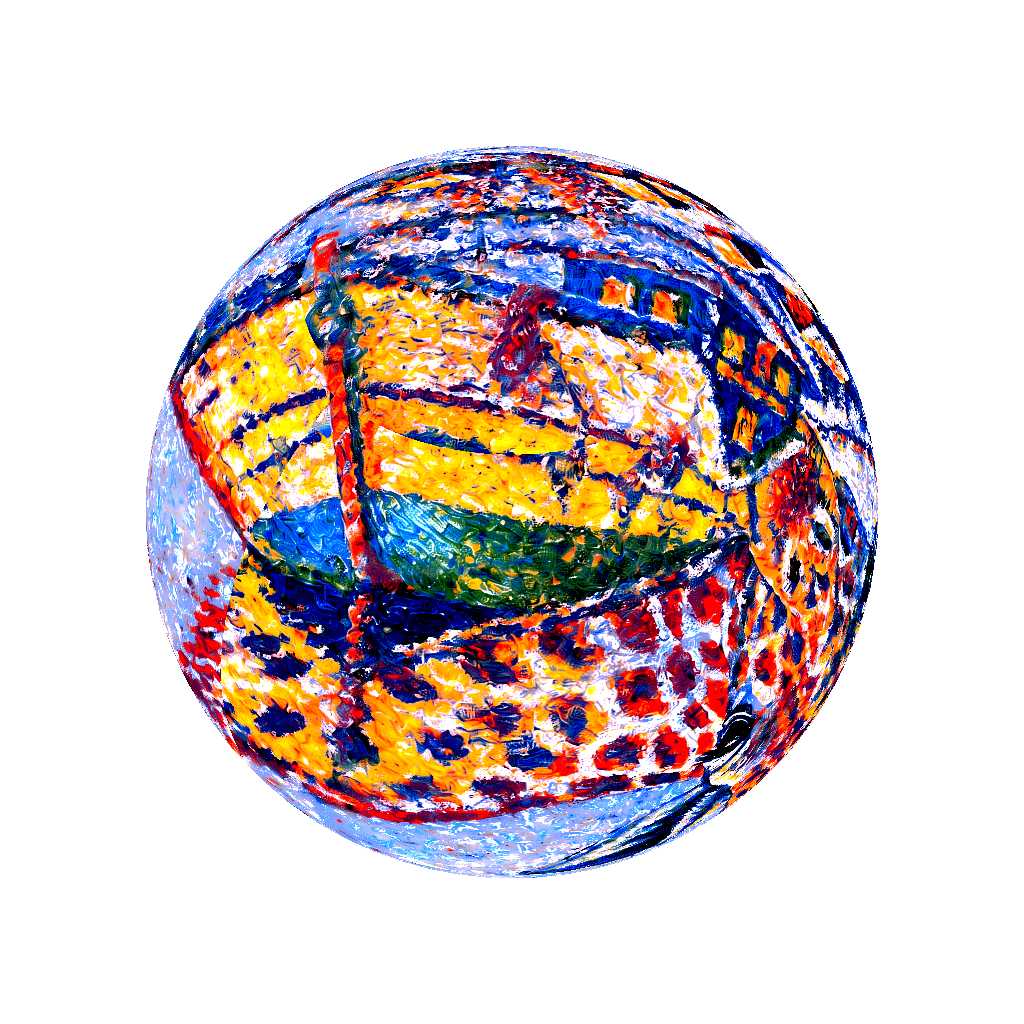}
    \end{minipage}%
    \begin{minipage}{0.16\textwidth}
        \includegraphics[width=\linewidth, trim=140 140 140 140, clip]{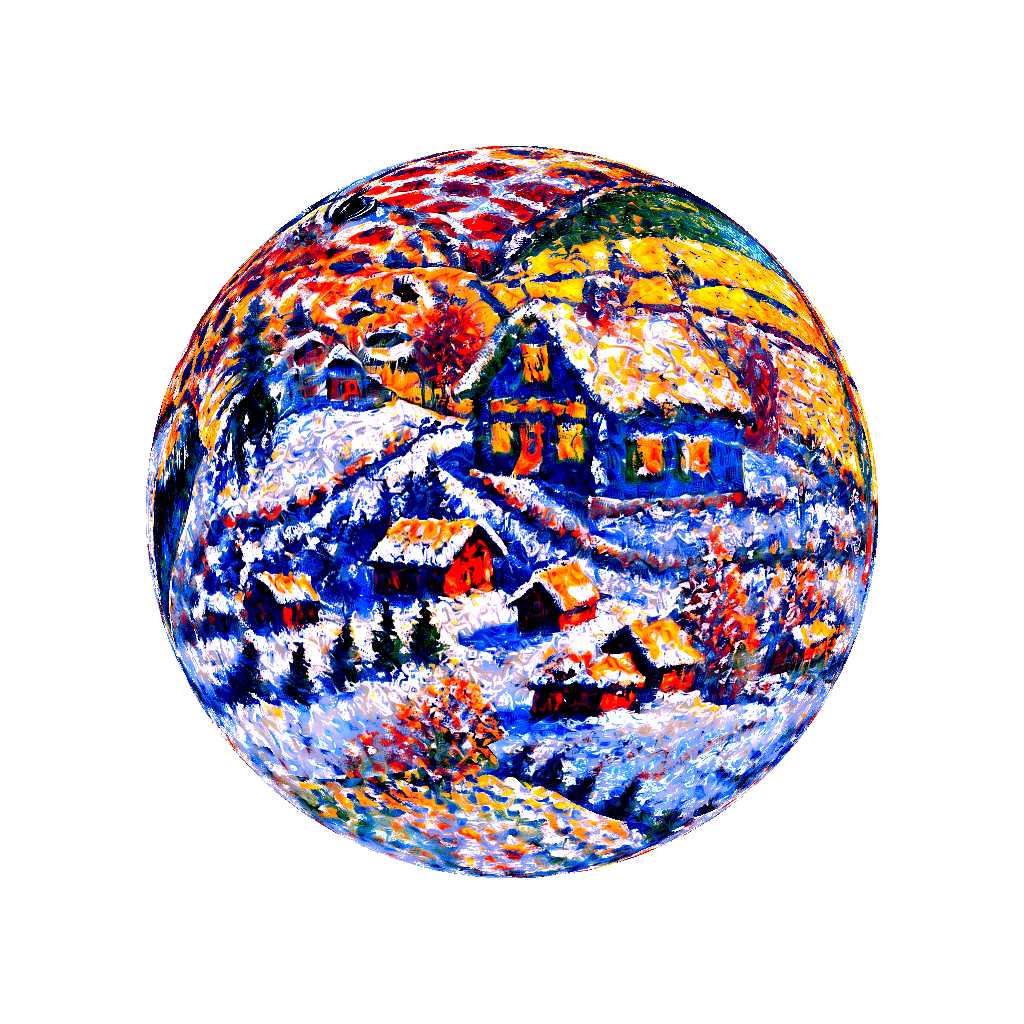}
    \end{minipage}%
    \begin{minipage}{0.16\textwidth}
        \includegraphics[width=\linewidth, trim=140 140 140 140, clip]{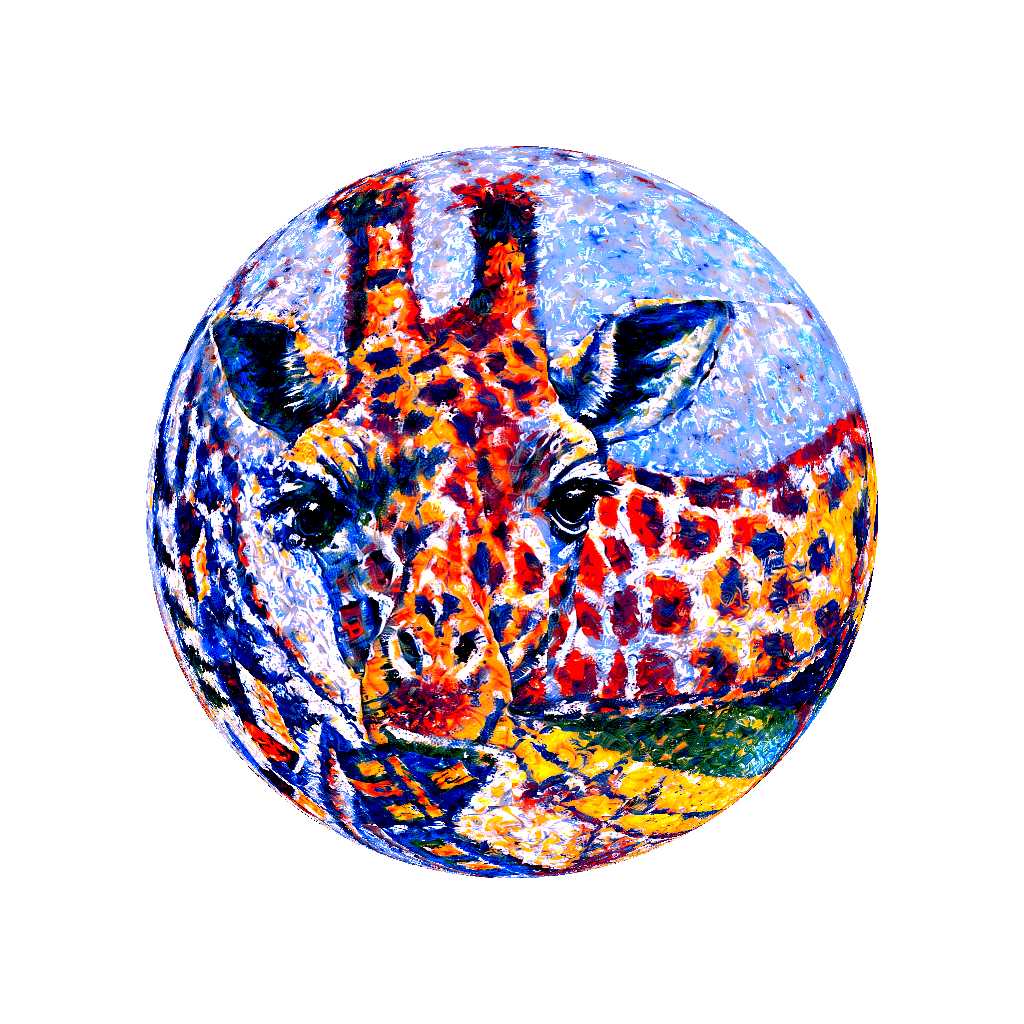}
    \end{minipage} \\

        \begin{minipage}[t]{0.16\textwidth}
        \centering
            \includegraphics[width=\linewidth, trim=185 20 185 30, clip]{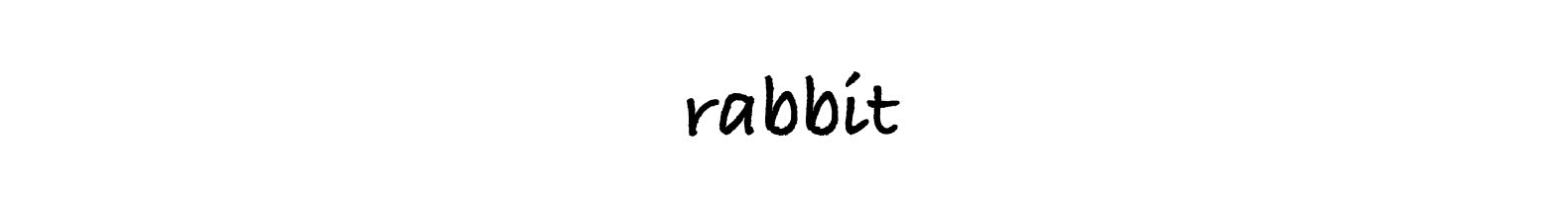}
            \end{minipage}\hfill
        \begin{minipage}[t]{0.16\textwidth}
            \centering
            \includegraphics[width=\linewidth, trim=185 20 185 30, clip]{figures/prompt/violin.jpg}
            \end{minipage}\hfill
        \begin{minipage}[t]{0.16\textwidth}
            \centering
            \includegraphics[width=\linewidth, trim=185 20 185 30, clip]{figures/prompt/wineandcheese.jpg}
        \end{minipage}\hfill
        \begin{minipage}[t]{0.16\textwidth}
            \centering
            \includegraphics[width=\linewidth, trim=185 20 185 30, clip]{figures/prompt/boat.jpg}
        \end{minipage}
        \begin{minipage}[t]{0.16\textwidth}
            \centering
            \includegraphics[width=\linewidth, trim=165 20 165 30, clip]{figures/prompt/village.jpg}
        \end{minipage}
        \begin{minipage}[t]{0.16\textwidth}
            \centering
            \includegraphics[width=\linewidth, trim=185 20 185 30, clip]{figures/prompt/giraffe.jpg}
        \end{minipage}\\

    \begin{minipage}{0.16\textwidth}
        \includegraphics[width=\linewidth, trim=140 140 140 140, clip]{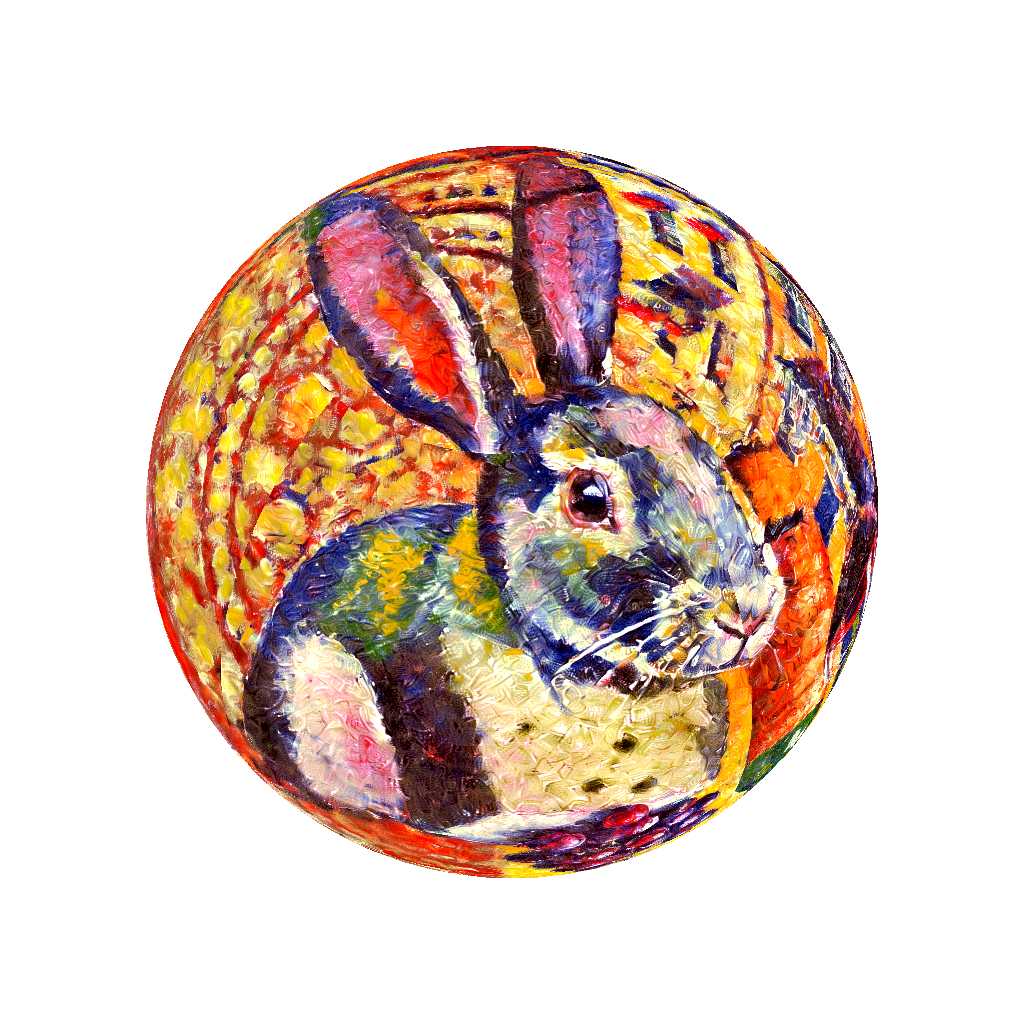}
    \end{minipage}%
    \begin{minipage}{0.16\textwidth}
        \includegraphics[width=\linewidth, trim=140 140 140 140, clip]{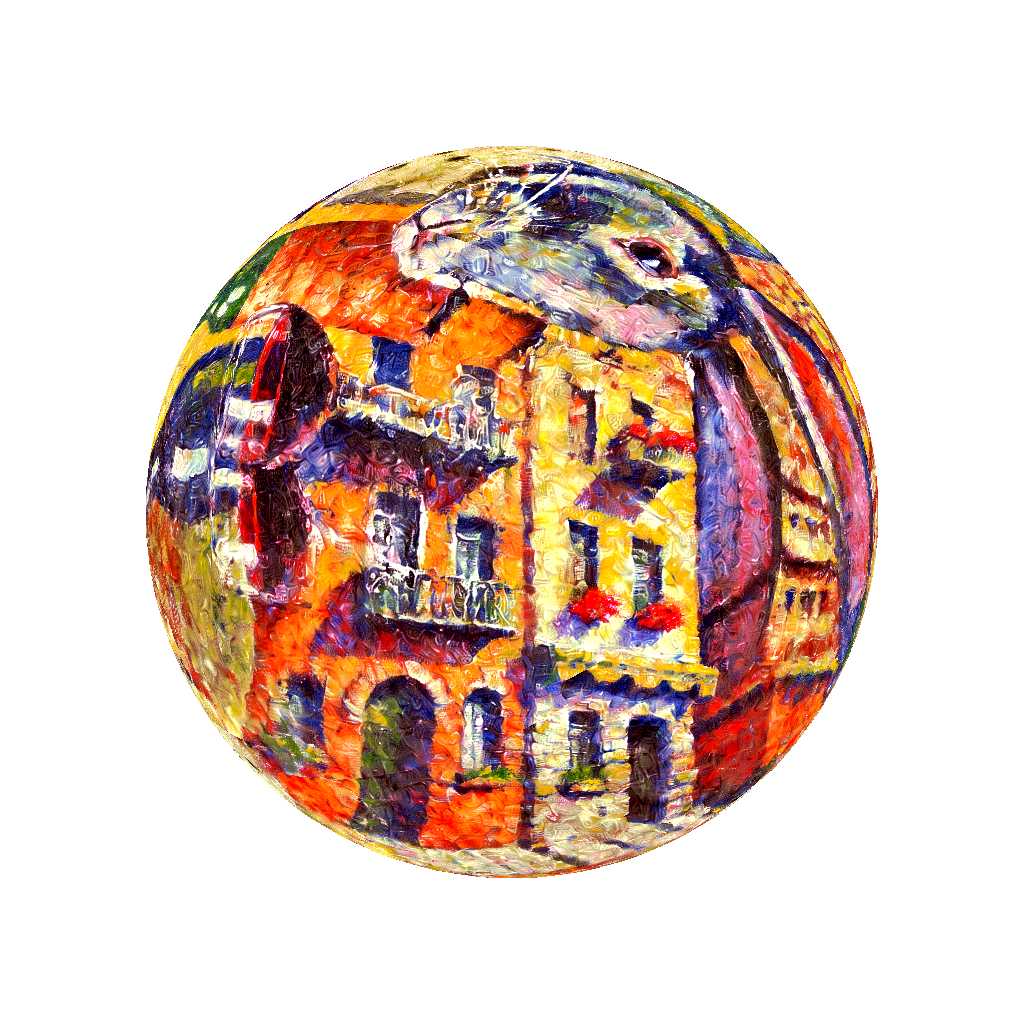}
    \end{minipage}%
    \begin{minipage}{0.16\textwidth}
        \includegraphics[width=\linewidth, trim=140 140 140 140, clip]{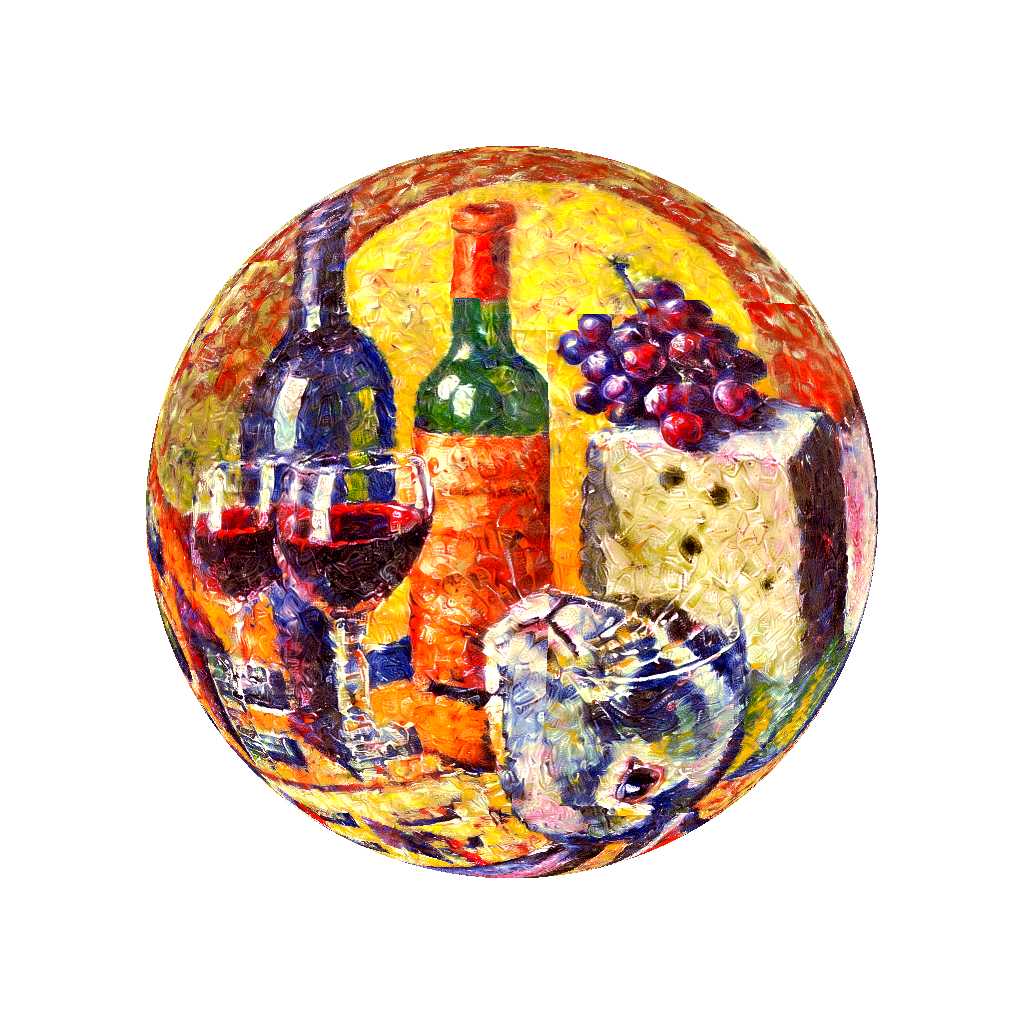}
    \end{minipage}%
    \begin{minipage}{0.16\textwidth}
        \includegraphics[width=\linewidth, trim=140 140 140 140, clip]{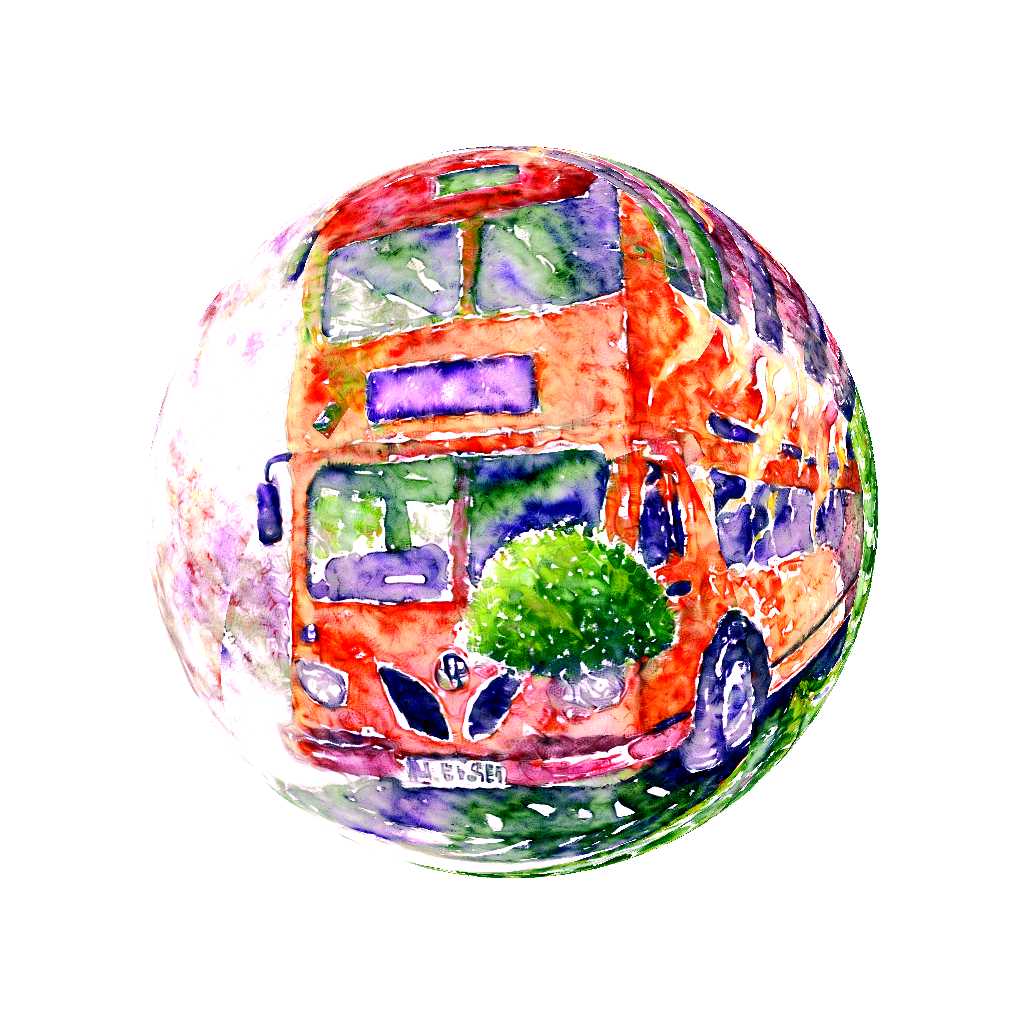}
    \end{minipage}%
    \begin{minipage}{0.16\textwidth}
        \includegraphics[width=\linewidth, trim=140 140 140 140, clip]{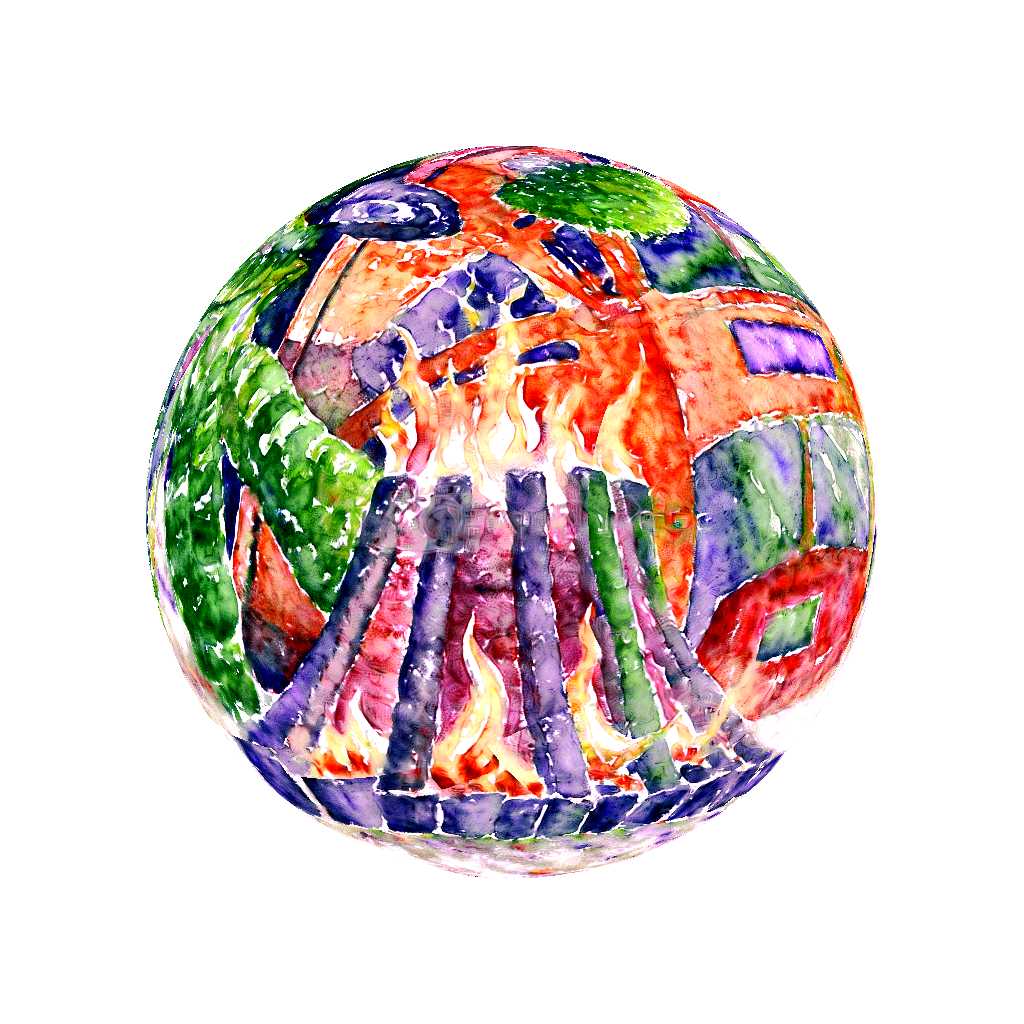}
    \end{minipage}%
    \begin{minipage}{0.16\textwidth}
        \includegraphics[width=\linewidth, trim=140 140 140 140, clip]{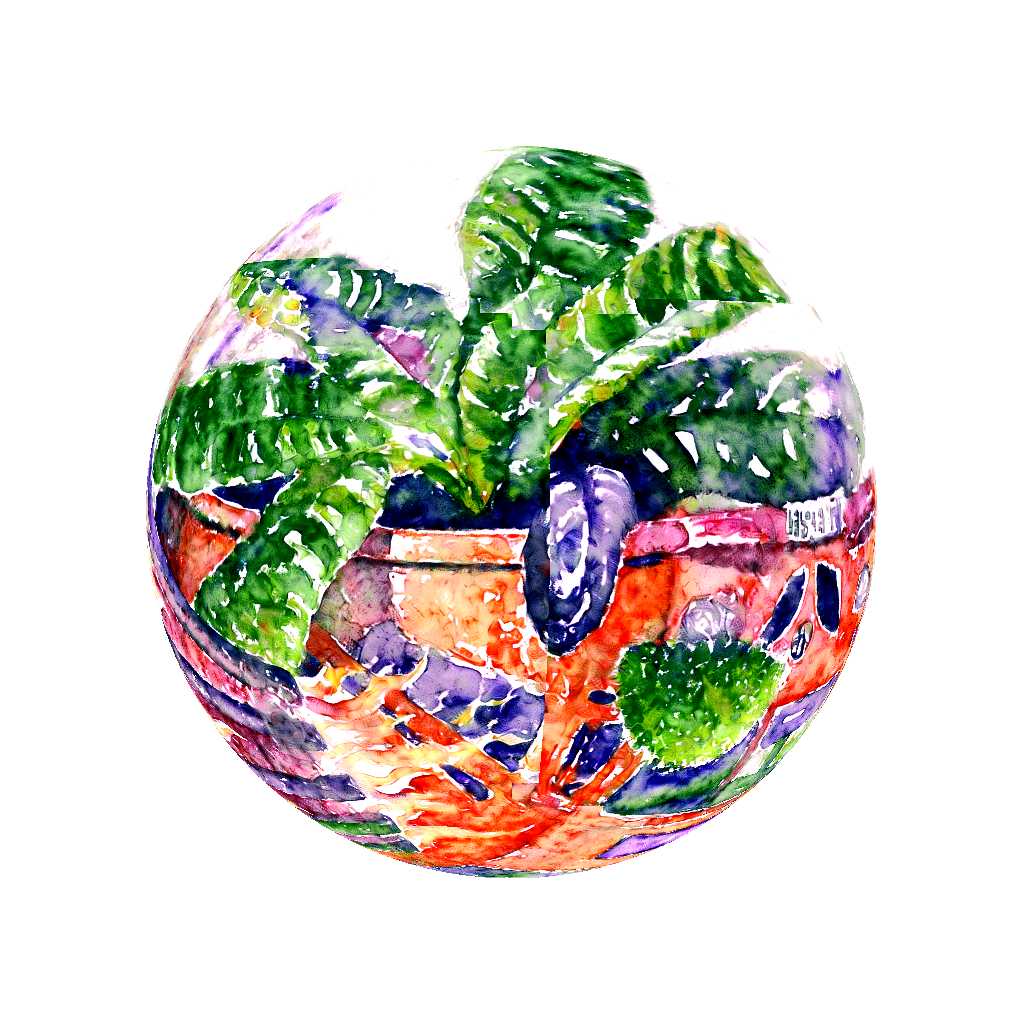}
    \end{minipage}\\
    
        \begin{minipage}[t]{0.16\textwidth}
        \centering
            \includegraphics[width=\linewidth, trim=185 20 185 30, clip]{figures/prompt/rabbit.jpg}
            \end{minipage}\hfill
        \begin{minipage}[t]{0.16\textwidth}
            \centering
            \includegraphics[width=\linewidth, trim=185 15 185 30, clip]{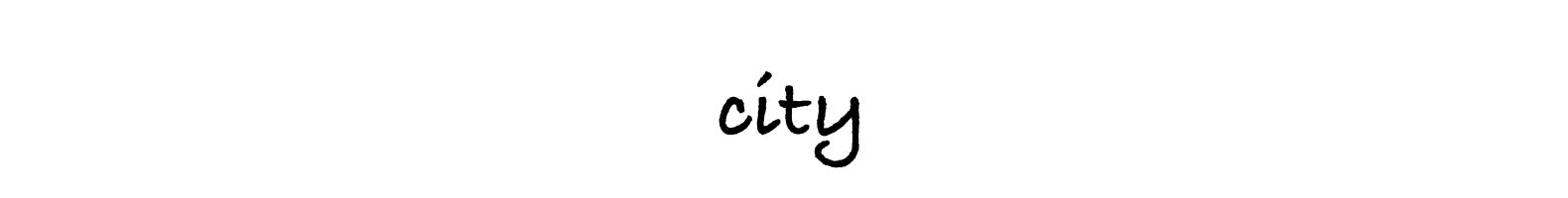}
            \end{minipage}\hfill
        \begin{minipage}[t]{0.16\textwidth}
            \centering
            \includegraphics[width=\linewidth, trim=185 20 185 30, clip]{figures/prompt/wineandcheese.jpg}
        \end{minipage}\hfill
        \begin{minipage}[t]{0.16\textwidth}
            \centering
            \includegraphics[width=\linewidth, trim=185 20 185 30, clip]{figures/prompt/bus.jpg}
        \end{minipage}
        \begin{minipage}[t]{0.16\textwidth}
            \centering
            \includegraphics[width=\linewidth, trim=165 20 165 30, clip]{figures/prompt/bonfire.jpg}
        \end{minipage}
        \begin{minipage}[t]{0.16\textwidth}
            \centering
            \includegraphics[width=\linewidth, trim=185 15 185 30, clip]{figures/prompt/teddybear.jpg}
        \end{minipage}\\

    \end{tabular}

    \caption{\textbf{Random samples.} We present more random examples on sphere case. }
    \label{fig:randomsamplesingle}
\end{figure*}

\topic{Random samples}
 Fig. \ref{fig:randomsamplebeanbag} (beanbag) and Fig. \ref{fig:randomsamplelego} (Lego) present random sample of complex shape results. We provide random samples of cube and spheres in Fig. \ref{fig:randomsample} and \ref{fig:randomsamplesingle}. All the prompts are randomly paired. We also have notched box examples in \href{index.html}{index.html} as they appear the same as cubes if we display them as figures.

\begin{figure}
    \centering

    \begin{minipage}[c]{0.24\linewidth}
        \includegraphics[trim=60 60 60 60, clip, width=\linewidth]{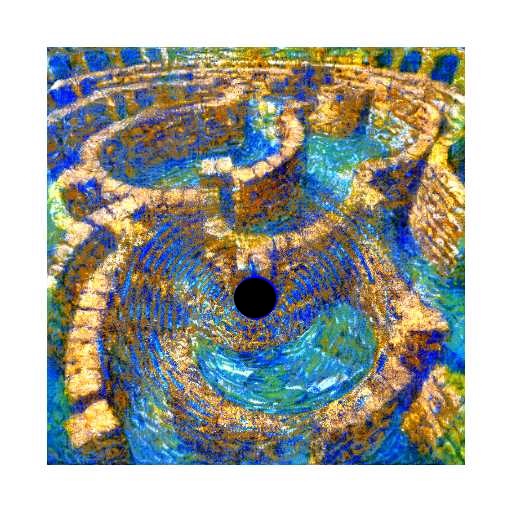}
    \end{minipage}\hfill
    \begin{minipage}[c]{0.24\linewidth}
        \includegraphics[trim=0 0 0 0, clip, width=\linewidth]{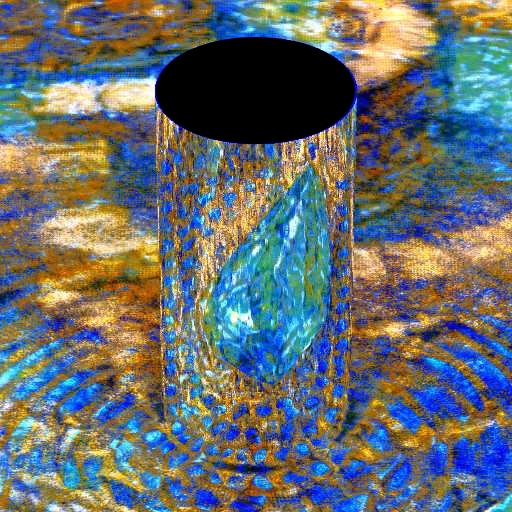}
    \end{minipage}\hfill
    \begin{minipage}[c]{0.24\linewidth}
        \includegraphics[trim=120 120 120 120, clip, width=\linewidth]{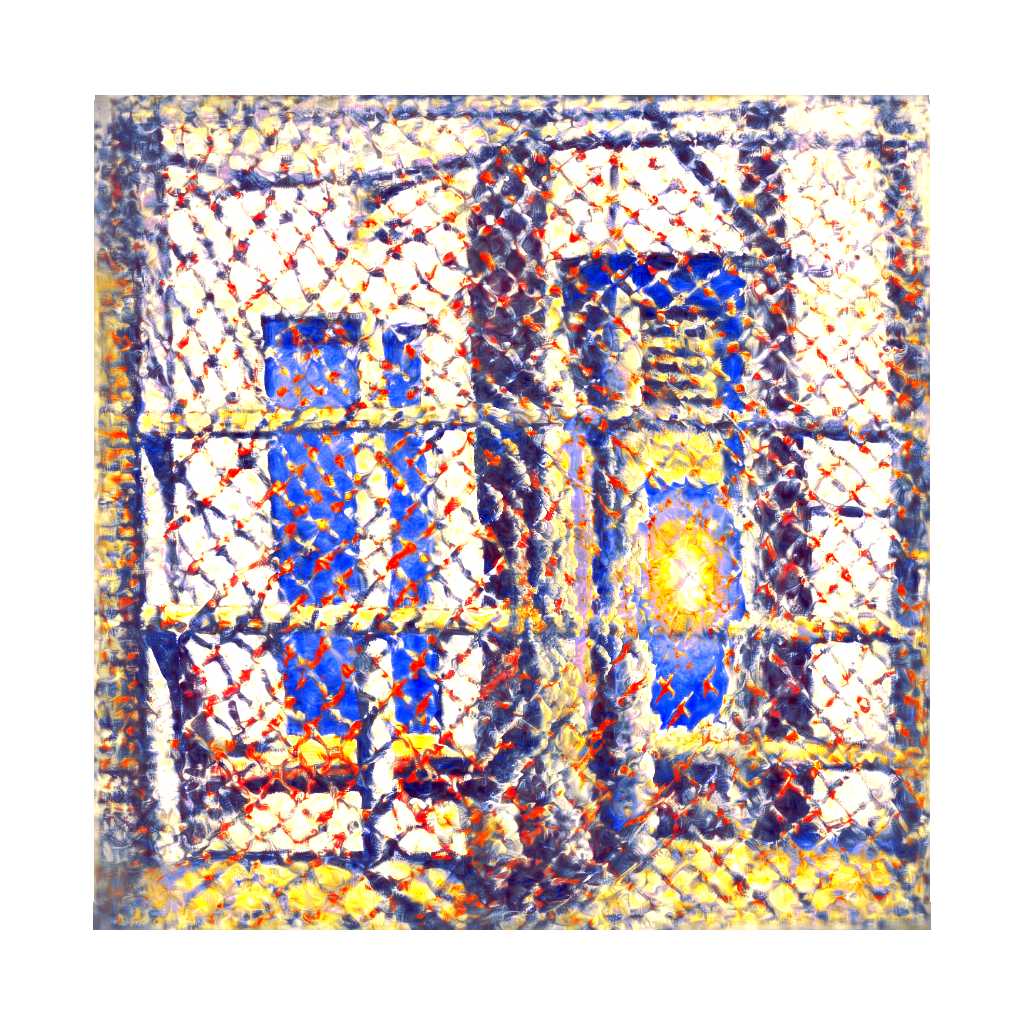}
    \end{minipage}\hfill
    \begin{minipage}[c]{0.24\linewidth}
        \includegraphics[trim=0 0 0 0, clip, width=\linewidth]{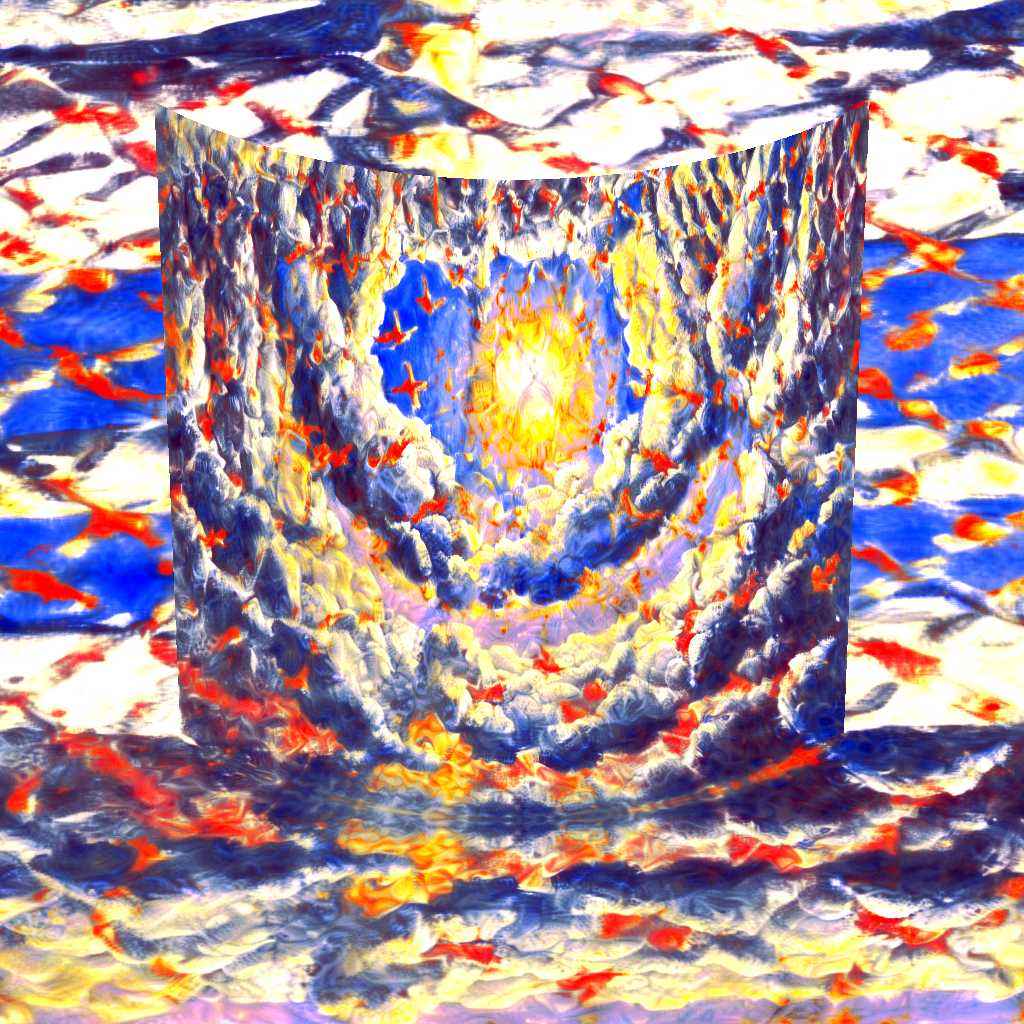}
    \end{minipage}\\

    \begin{minipage}[c]{0.23\linewidth}
        \hspace{5.5mm}
        \includegraphics[width=0.3\linewidth, trim=1020 20 400 30, clip]{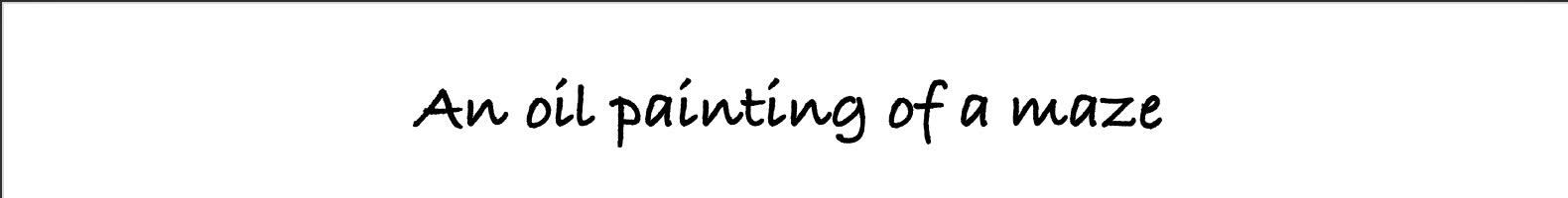}
    \end{minipage}\hfill
    \begin{minipage}[c]{0.23\linewidth}
        \hspace{6mm}
        \includegraphics[width=0.3\linewidth, trim=1030 20 400 30, clip]{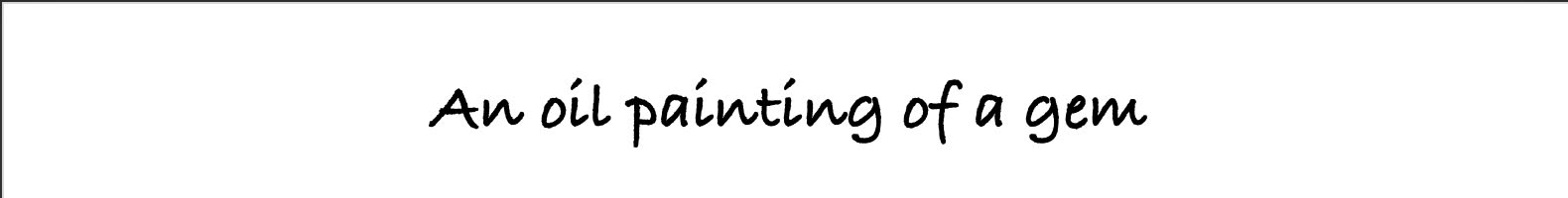}
    \end{minipage}\hfill
    \begin{minipage}[c]{0.23\linewidth}
        \hspace{6mm}
        \includegraphics[width=0.3\linewidth, trim=1030 20 400 30, clip]{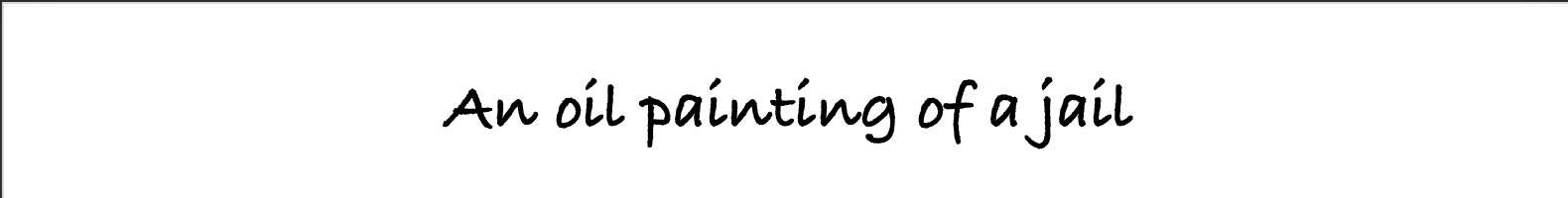}
    \end{minipage}\hfill
    \begin{minipage}[c]{0.23\linewidth}
        \hspace{4mm}
        \includegraphics[width=0.37\linewidth, trim=1000 20 400 30, clip]{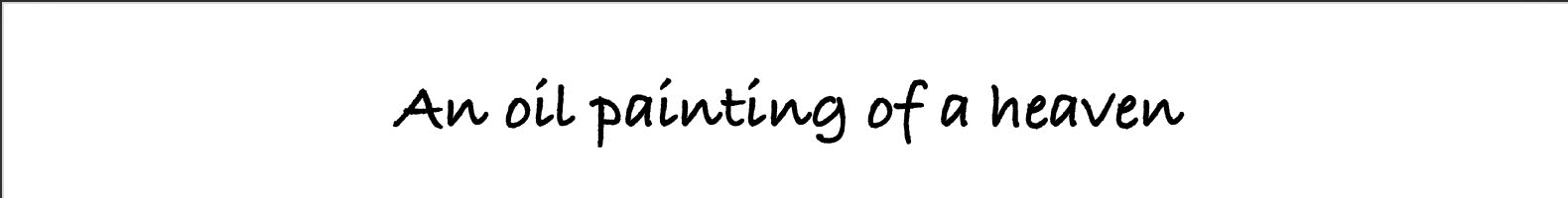}
    \end{minipage}\\

    \begin{minipage}[c]{0.24\linewidth}
        \includegraphics[trim=120 120 120 120, clip, width=\linewidth]{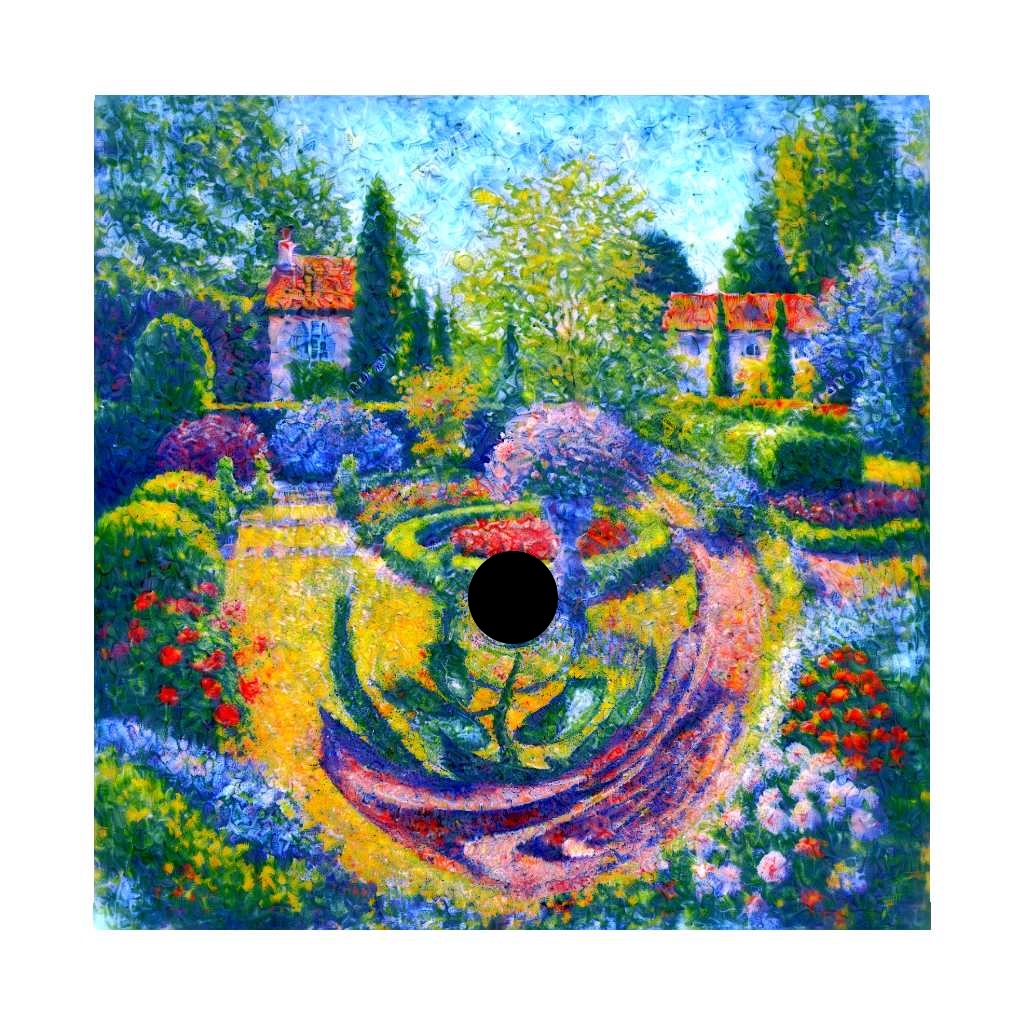}
    \end{minipage}\hfill
    \begin{minipage}[c]{0.24\linewidth}
        \includegraphics[trim=0 0 0 0, clip, width=\linewidth]{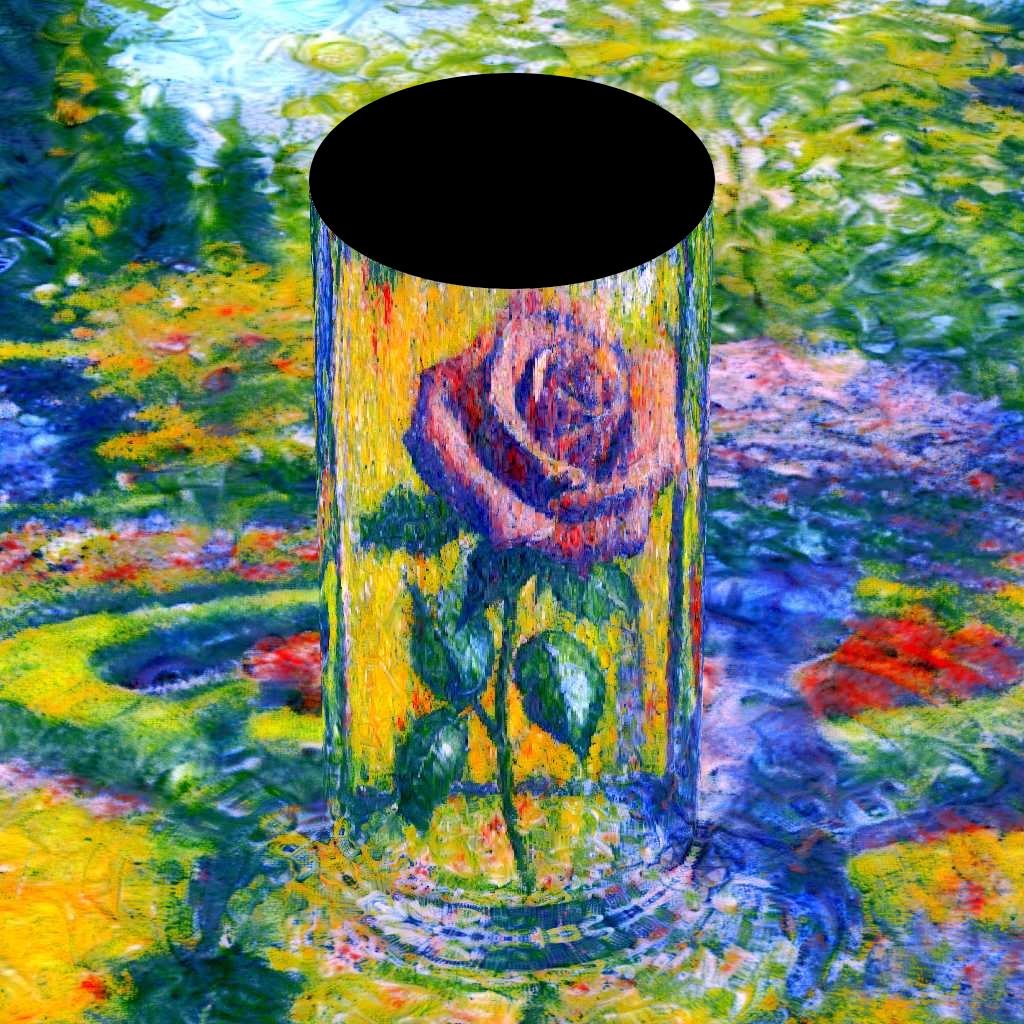}
    \end{minipage}\hfill
    \begin{minipage}[c]{0.24\linewidth}
        \includegraphics[trim=120 120 120 120, clip, width=\linewidth]{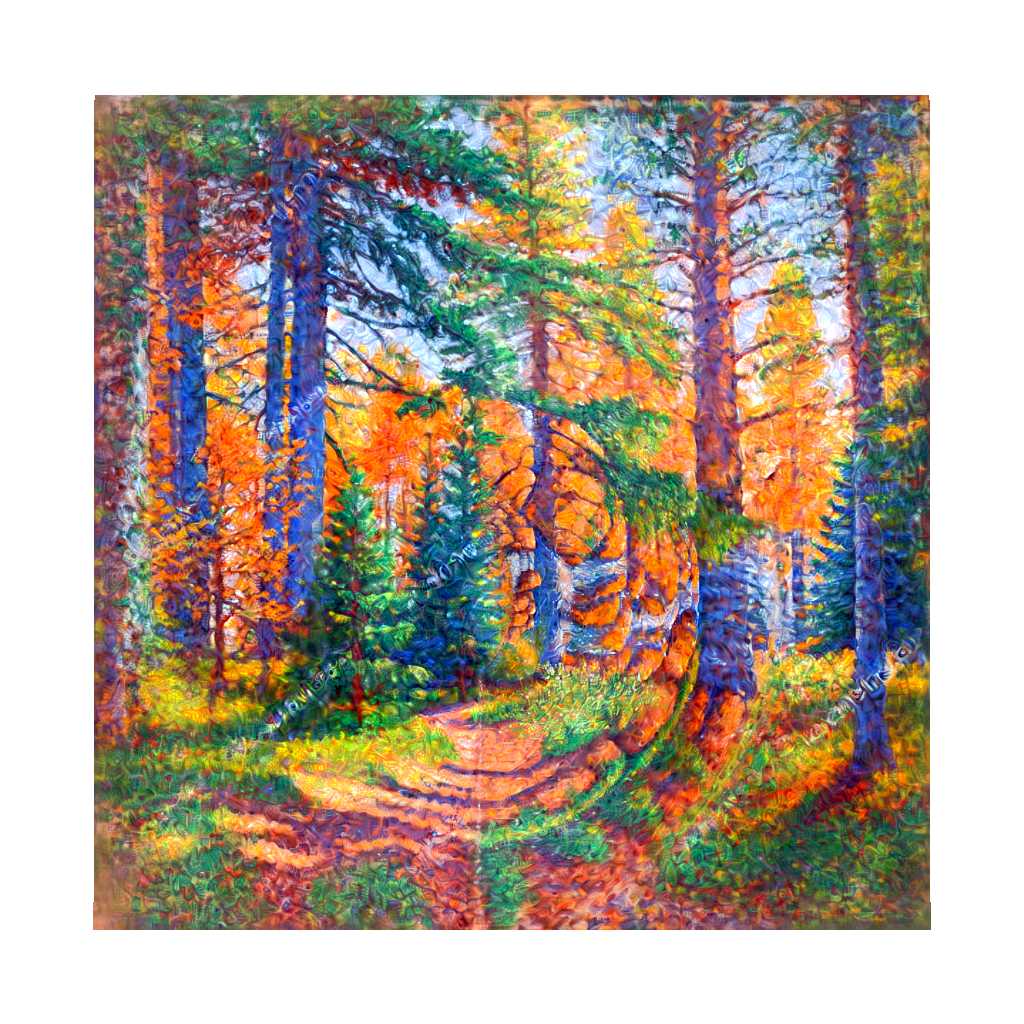}
    \end{minipage}\hfill
    \begin{minipage}[c]{0.24\linewidth}
        \includegraphics[trim=0 0 0 0, clip, width=\linewidth]{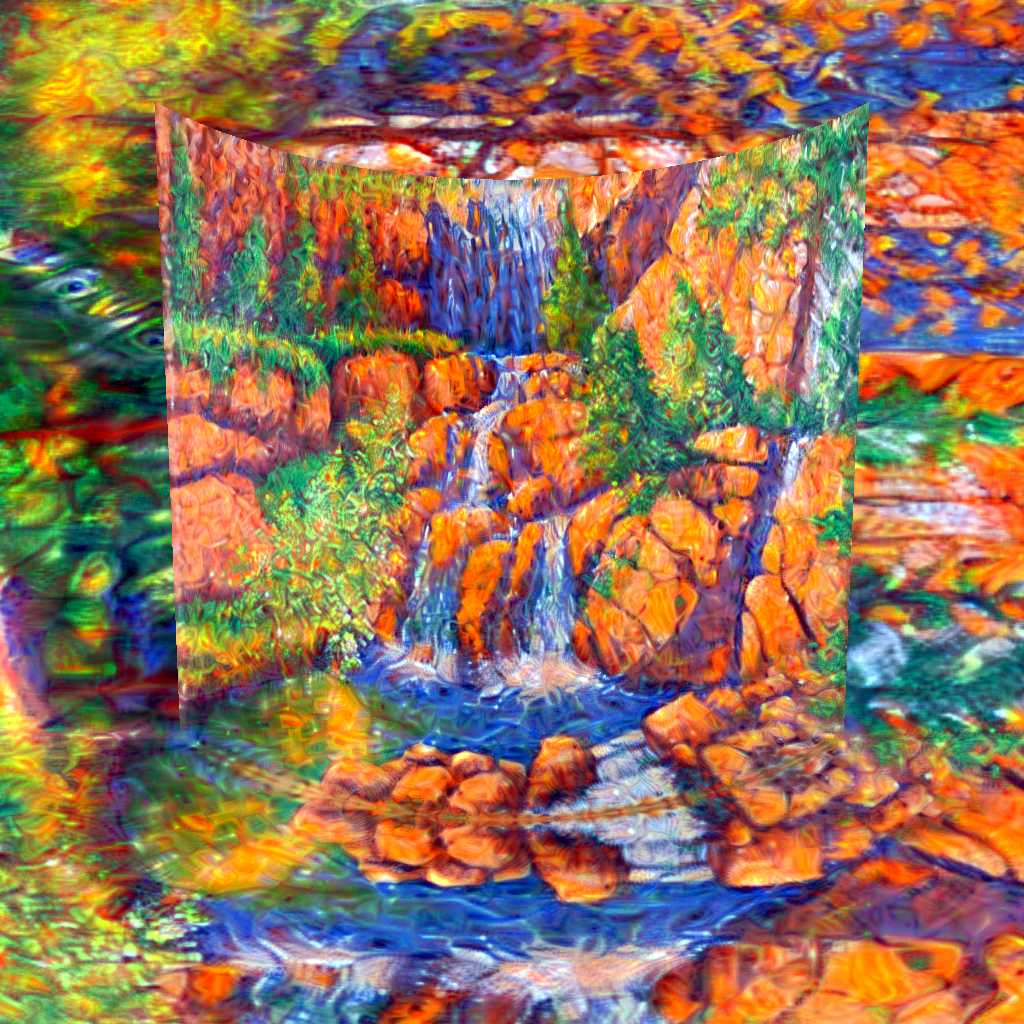}
    \end{minipage}\\

    \begin{minipage}[c]{0.23\linewidth}
        \hspace{5mm}
        \includegraphics[width=0.4\linewidth, trim=1080 20 300 30, clip]{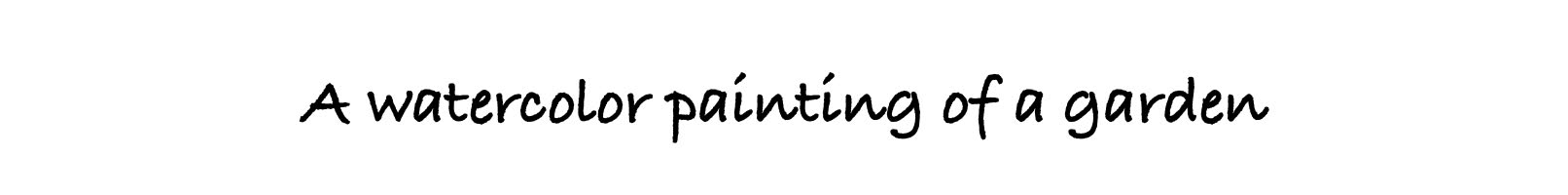}
    \end{minipage}\hfill
    \begin{minipage}[c]{0.23\linewidth}
        \hspace{6mm}
        \includegraphics[width=0.3\linewidth, trim=1030 20 400 30, clip]{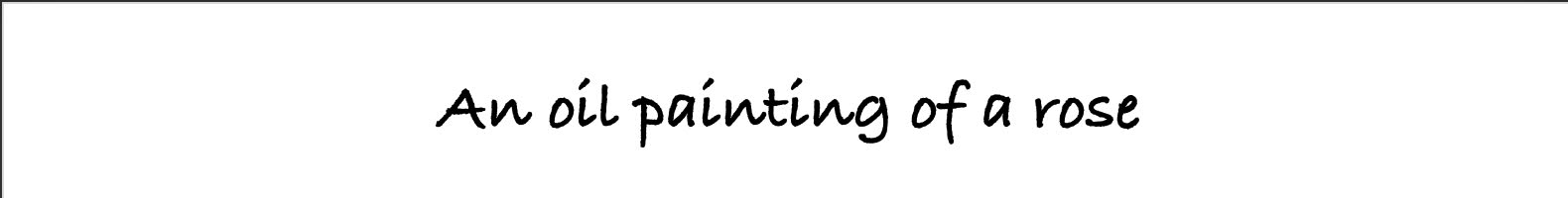}
    \end{minipage}\hfill
    \begin{minipage}[c]{0.23\linewidth}
        \hspace{5mm}
        \includegraphics[width=0.35\linewidth, trim=1010 20 400 30, clip]{figures/prompts/forest.jpg}
    \end{minipage}\hfill
    \begin{minipage}[c]{0.23\linewidth}
        \centering
        \includegraphics[width=0.5\linewidth, trim=970 20 360 30, clip]{figures/prompts/waterfall.jpg}
    \end{minipage}\\
    \vspace{-2mm}
    \caption{\textbf{3D multiview illusion with reflective surfaces.} 
    We demonstrate illusion generation on a reflective cylinder (left) and a curved mirror (right). 
    }
    \label{fig:reflective-single}
    \vspace{-4mm}
\end{figure}

For reflective case, a single reflective surface example is in Fig. \ref{fig:reflective-single}. Fig. \ref{fig:reflective-2cylinder} illustrates two reflective objects results. 

\topic{Failure cases}
We present more examples of failure cases videos and analysis at \href{index.html}{index.html}.  The human brain has objectified human faces in some sense, so it is harder to hide a human face from another view in the illusion generation process. 

\topic{Real examples}
More real-world examples videos can be found at \href{index.html}{index.html}. We use an iPad and a perfume cap/reflective paper card to make the example. We just grab what we have to make the examples and have not tested other materials. Note that the surface of the perfume cap and paper card is not smooth. If the reflective material is more specular, the result will be more pretty.

\topic{3D shape illusion}
We provide more examples of 3D shape illusion generation at \href{index.html}{index.html}.

\section{Ablation}
\label{sec:ablationsupp}

\begin{figure*}[h]
    \centering
    \fontsize{5.5pt}{6.5pt}\selectfont
    \setlength{\tabcolsep}{1pt} 
    \renewcommand{\arraystretch}{1}

    \begin{tabular}{lccccccc}
        &\textbf{Burgert \textit{et al.} ~\cite{Burgert2023DiffusionIH}} & \textbf{Baseline} & \textbf{Random patch}  & \textbf{w/o resolution scaling} & \textbf{w/o resolution scaling} &  \textbf{w/o camera jitter} & \textbf{Ours}\\

        \rotatebox{90}{\includegraphics[width=0.125\textwidth, trim=130 20 125 30, clip]{figures/prompt/womanwindow.jpg}}
        & \includegraphics[width=0.125\textwidth, trim=70 70 70 70, clip]{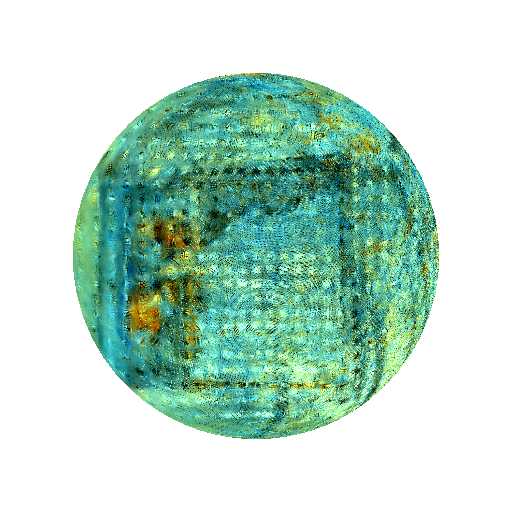} &
        \includegraphics[width=0.125\textwidth, trim=70 70 70 70, clip]{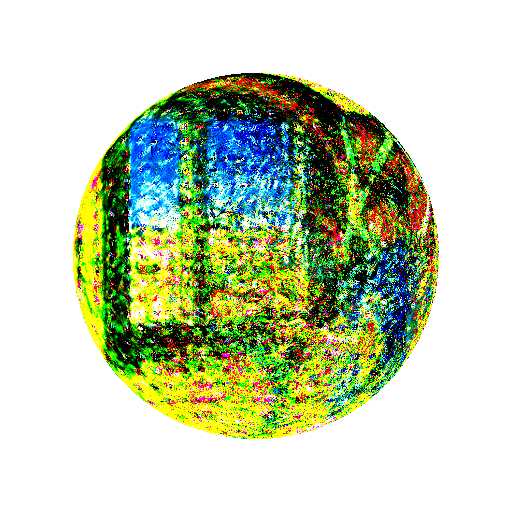} &
        \includegraphics[width=0.125\textwidth, trim=140 140 140 140, clip]{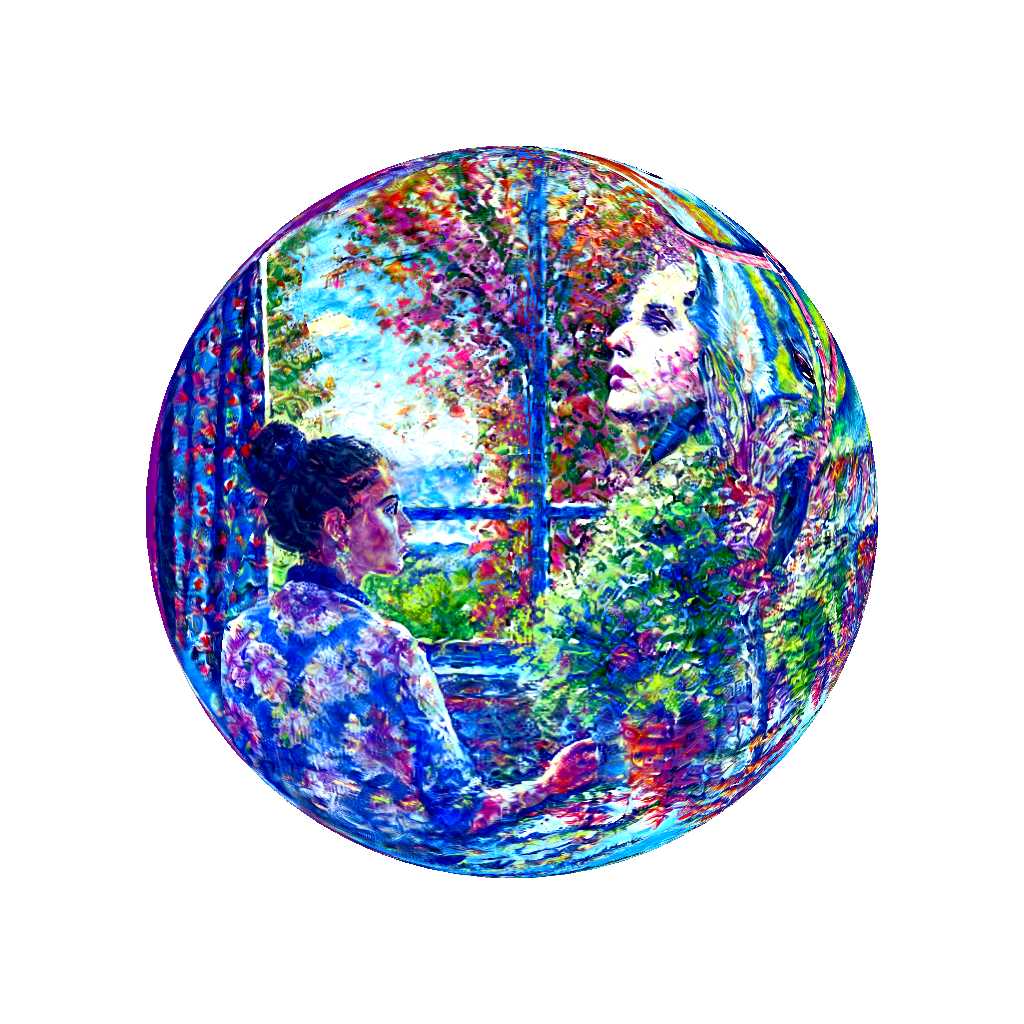} &
        \includegraphics[width=0.125\textwidth, trim=140 140 140 140, clip]{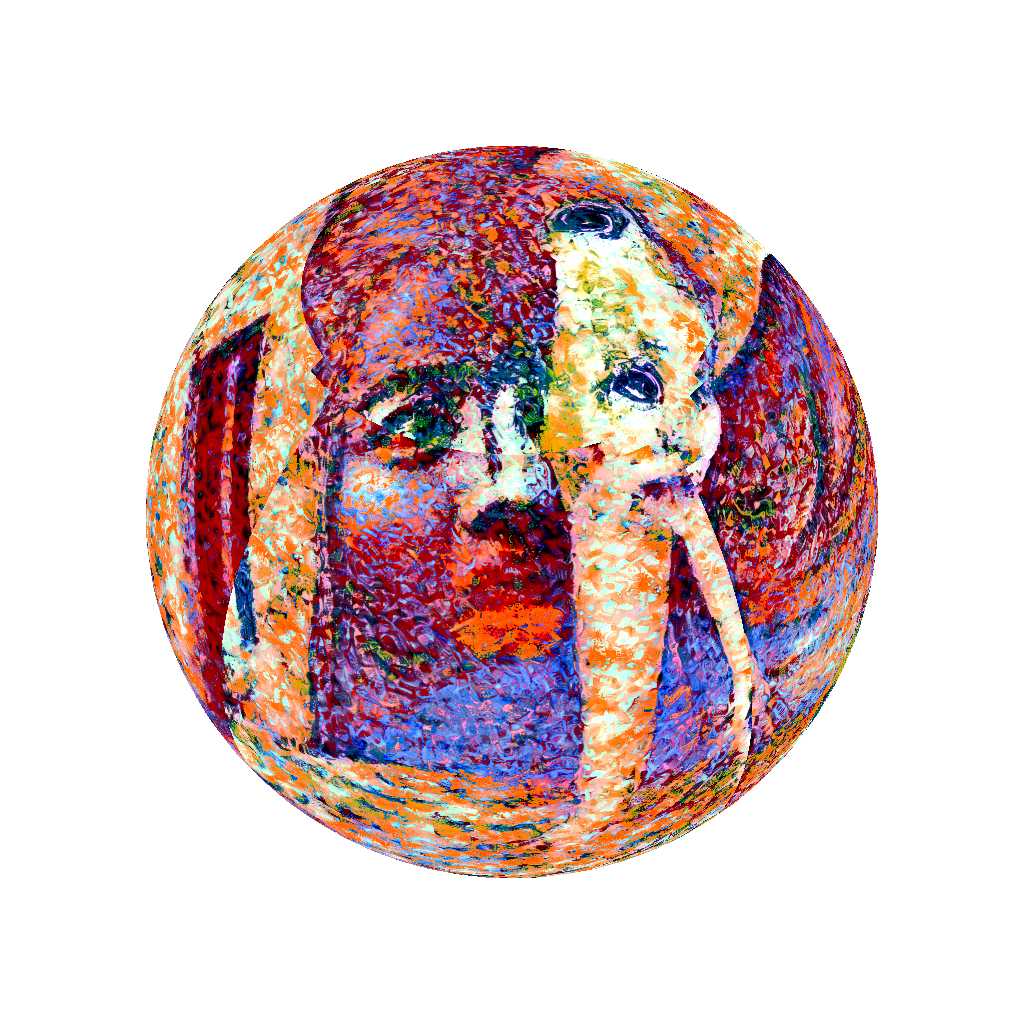} &
        \includegraphics[width=0.125\textwidth, trim=140 140 140 140, clip]{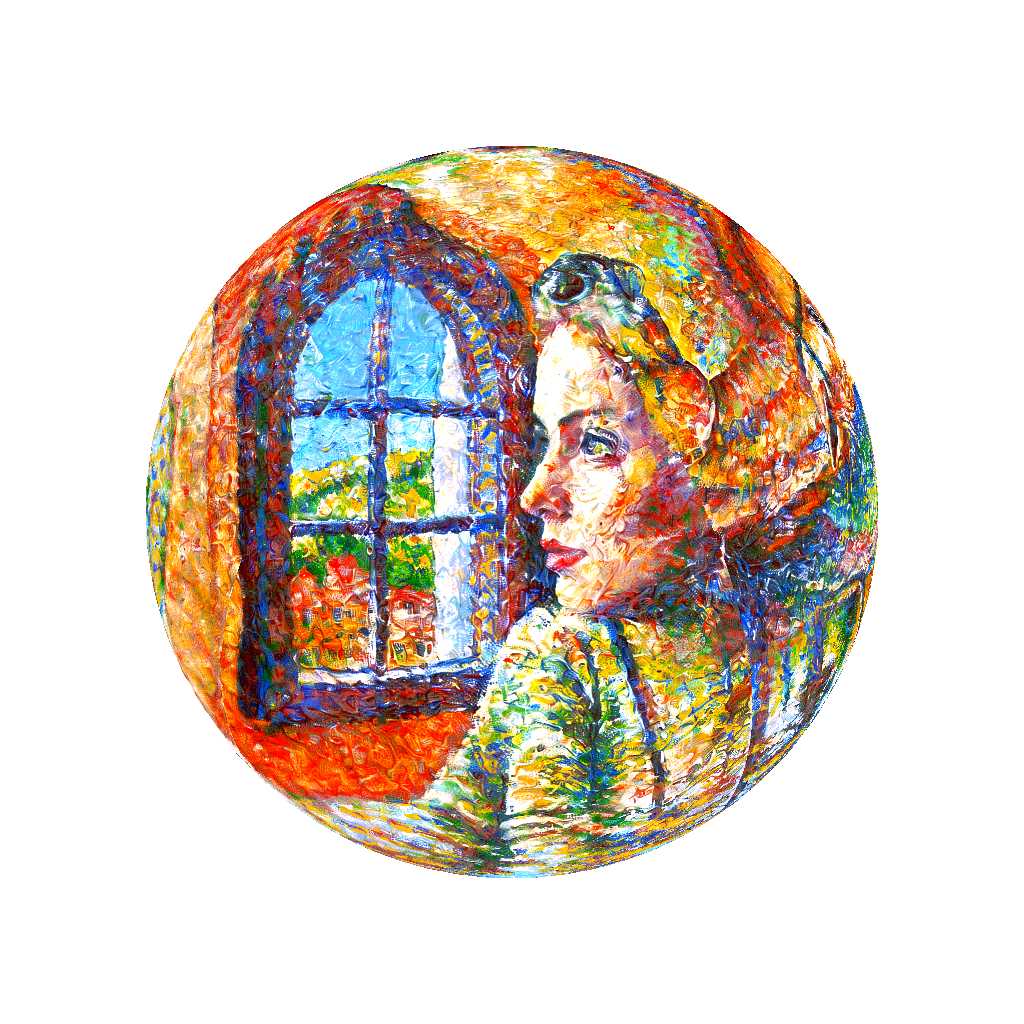} &
        \includegraphics[width=0.125\textwidth, trim=140 140 140 140, clip]{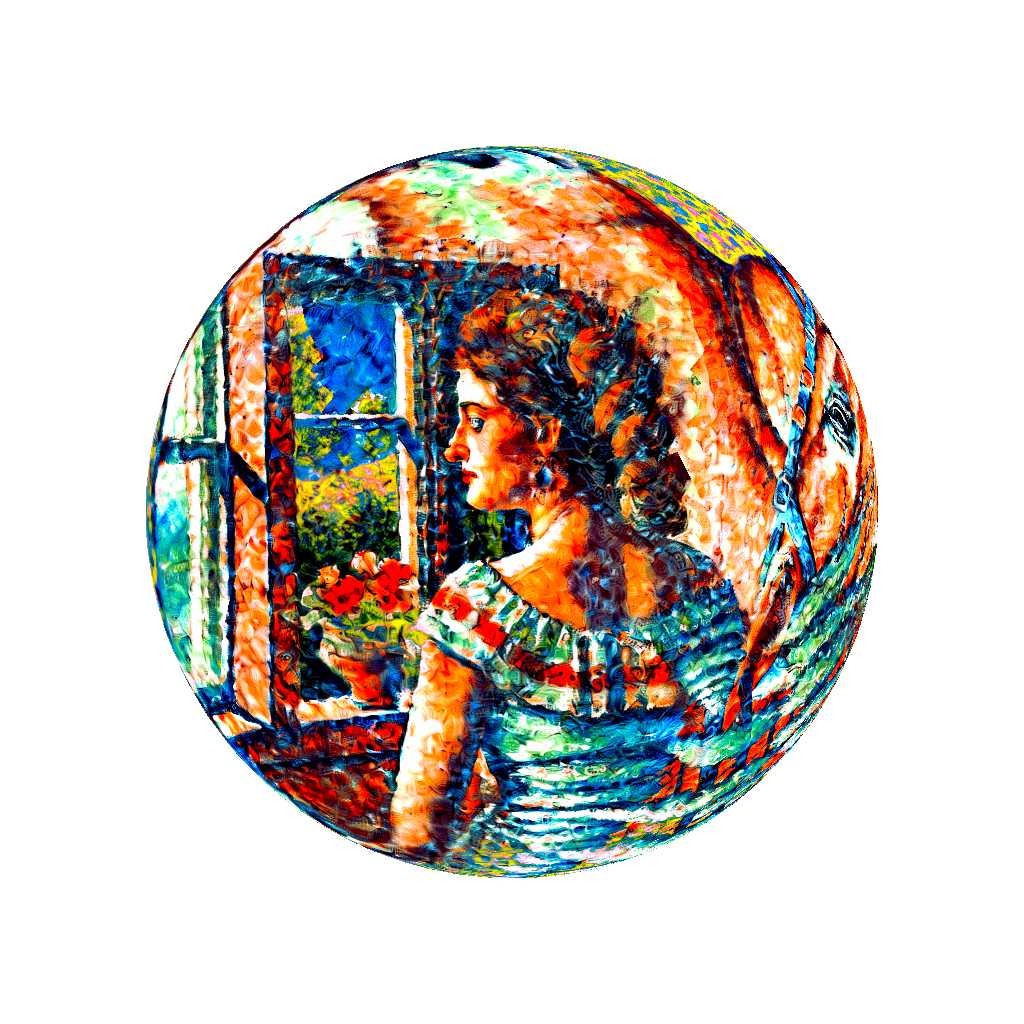} &
        \includegraphics[width=0.125\textwidth, trim=140 140 140 140, clip]{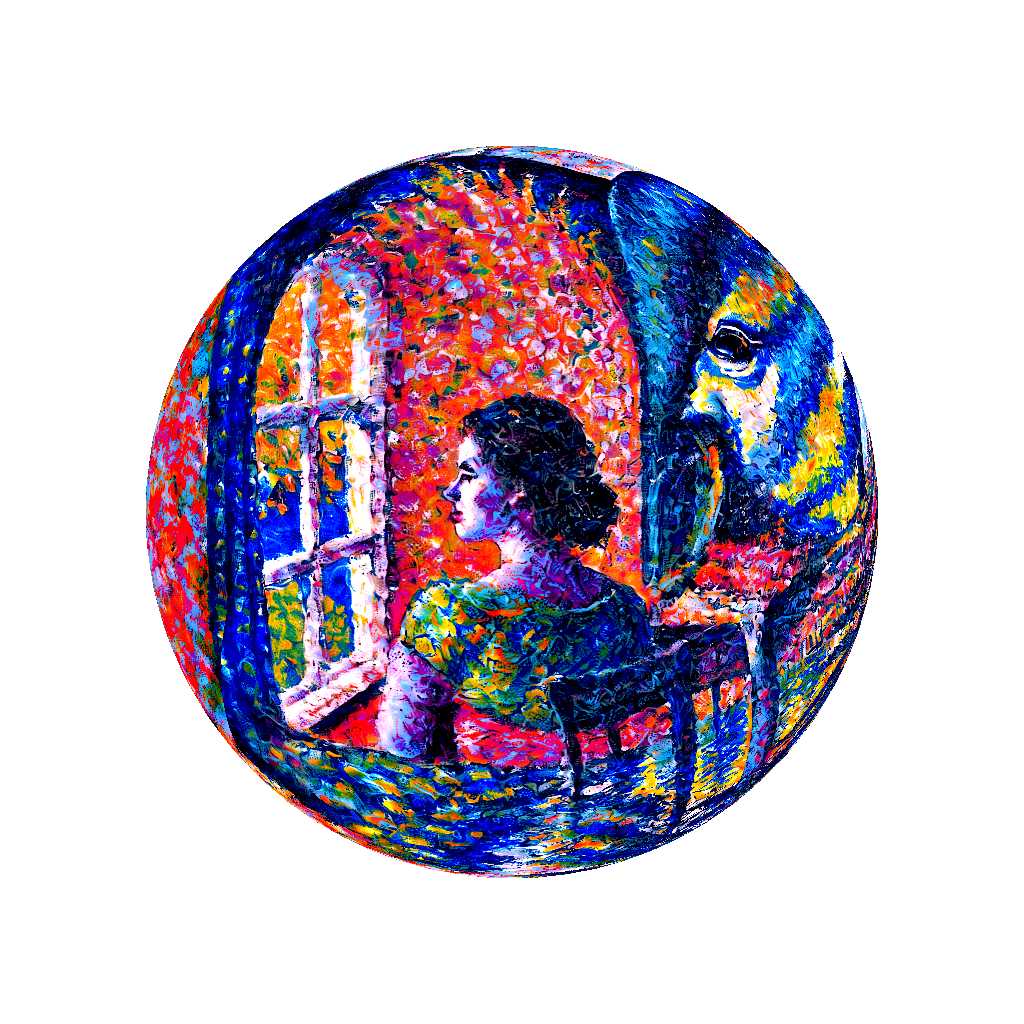} \\
        
        \rotatebox{90}{\includegraphics[width=0.125\textwidth, trim=150 20 150 30, clip]{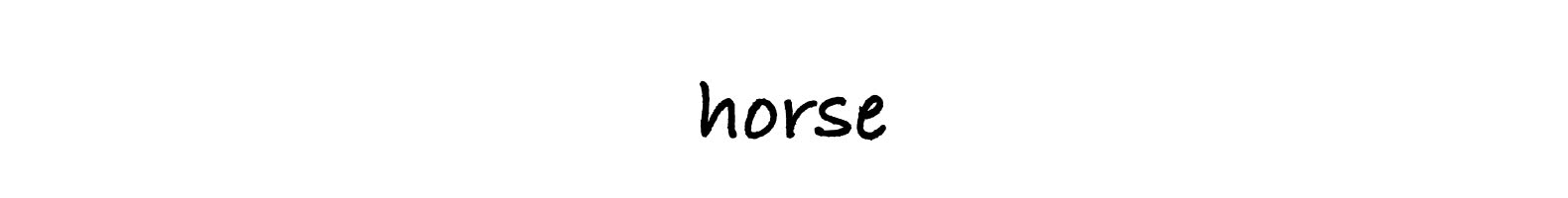}}
        & \includegraphics[width=0.125\textwidth, trim=70 70 70 70, clip]{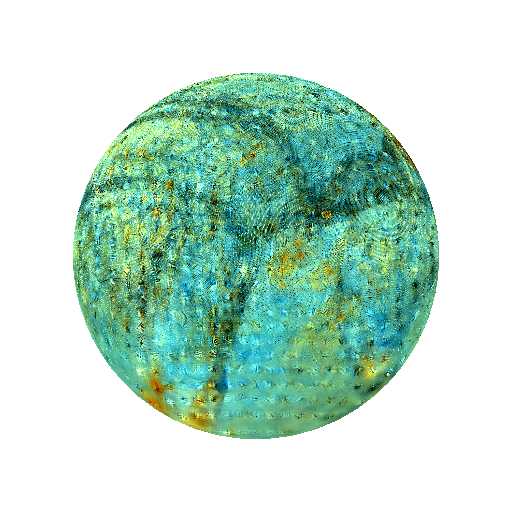} &
        \includegraphics[width=0.125\textwidth, trim=70 70 70 70, clip]{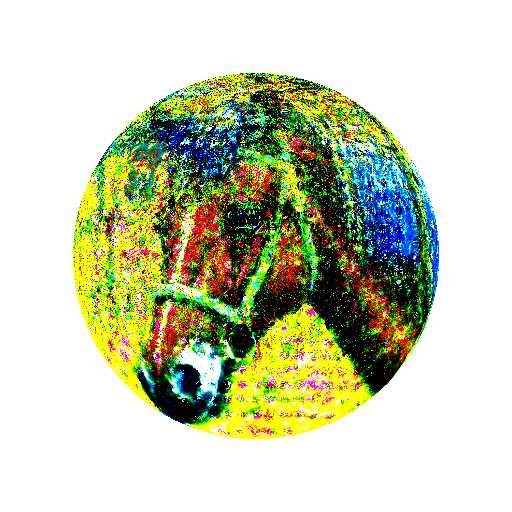} &
        \includegraphics[width=0.125\textwidth, trim=140 140 140 140, clip]{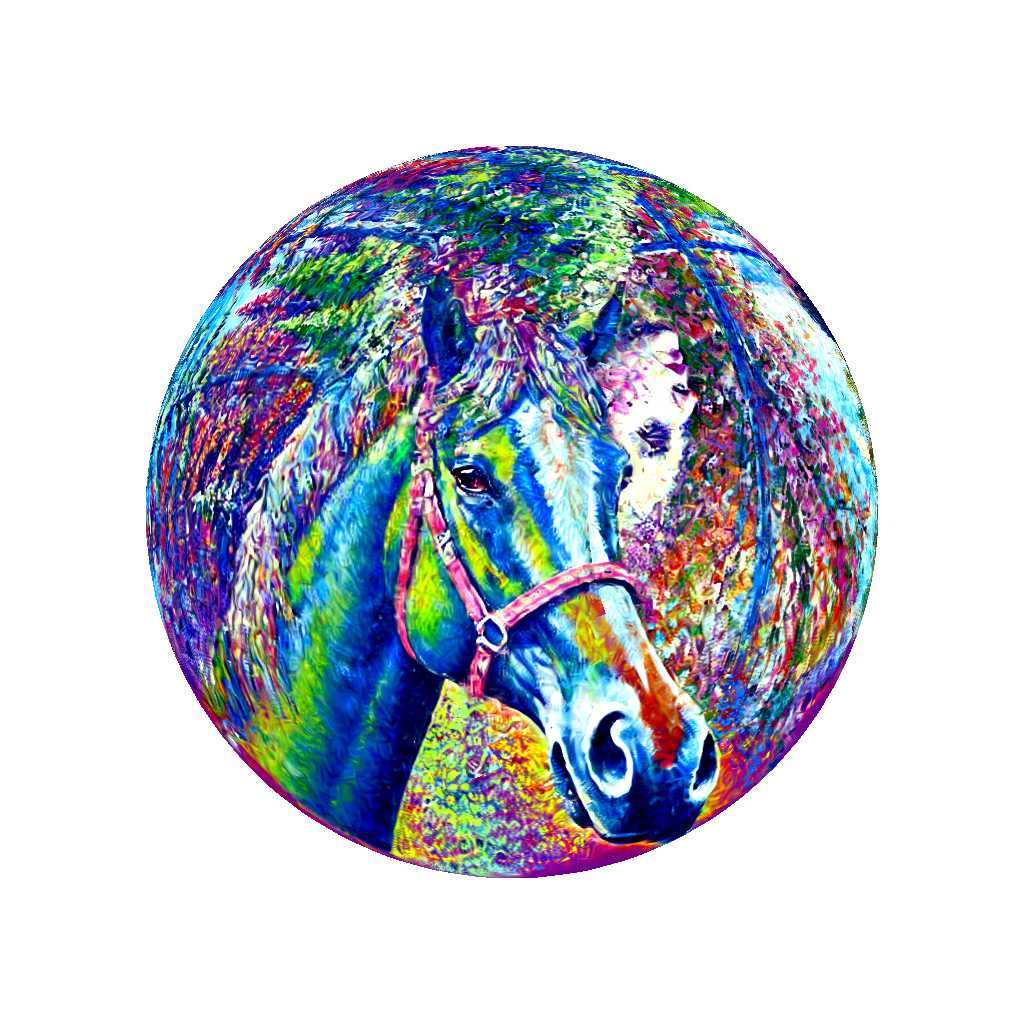} &
        \includegraphics[width=0.125\textwidth, trim=140 140 140 140, clip]{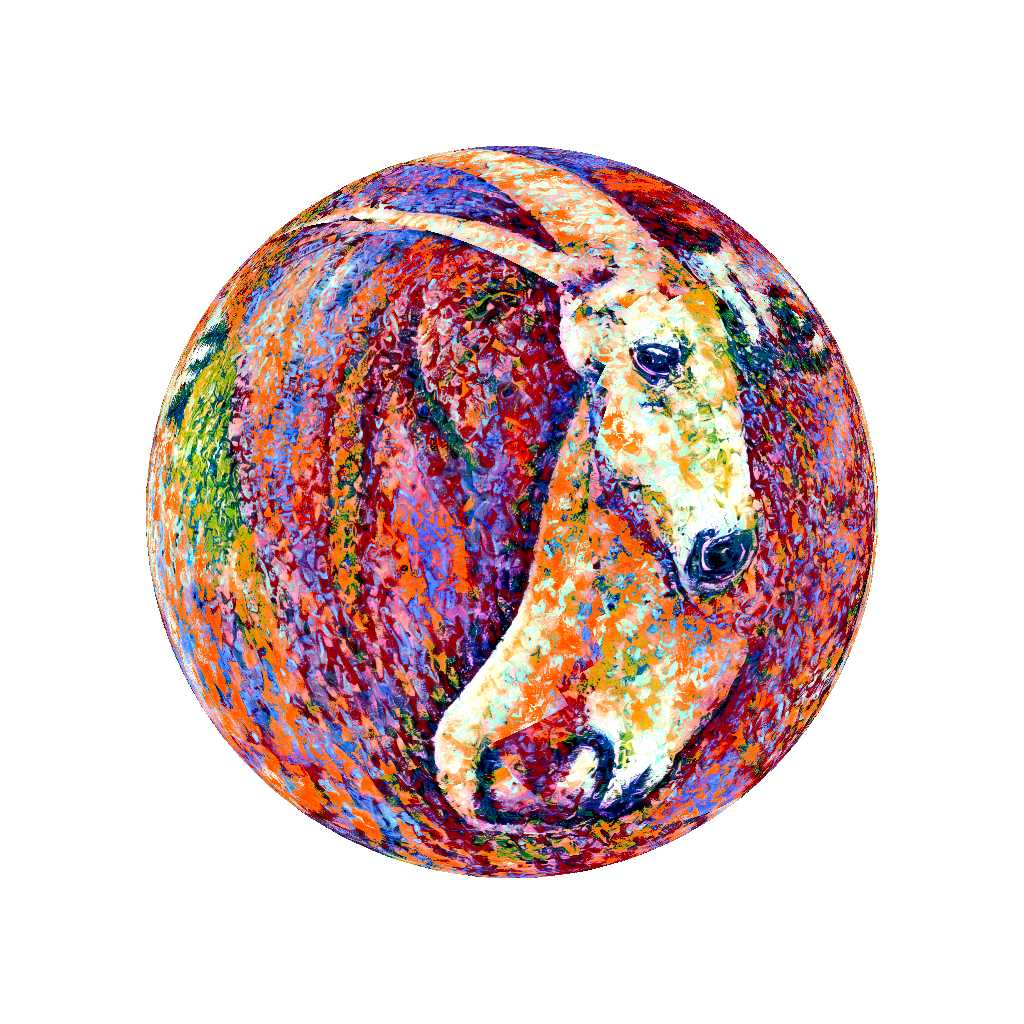} &
        \includegraphics[width=0.125\textwidth, trim=140 140 140 140, clip]{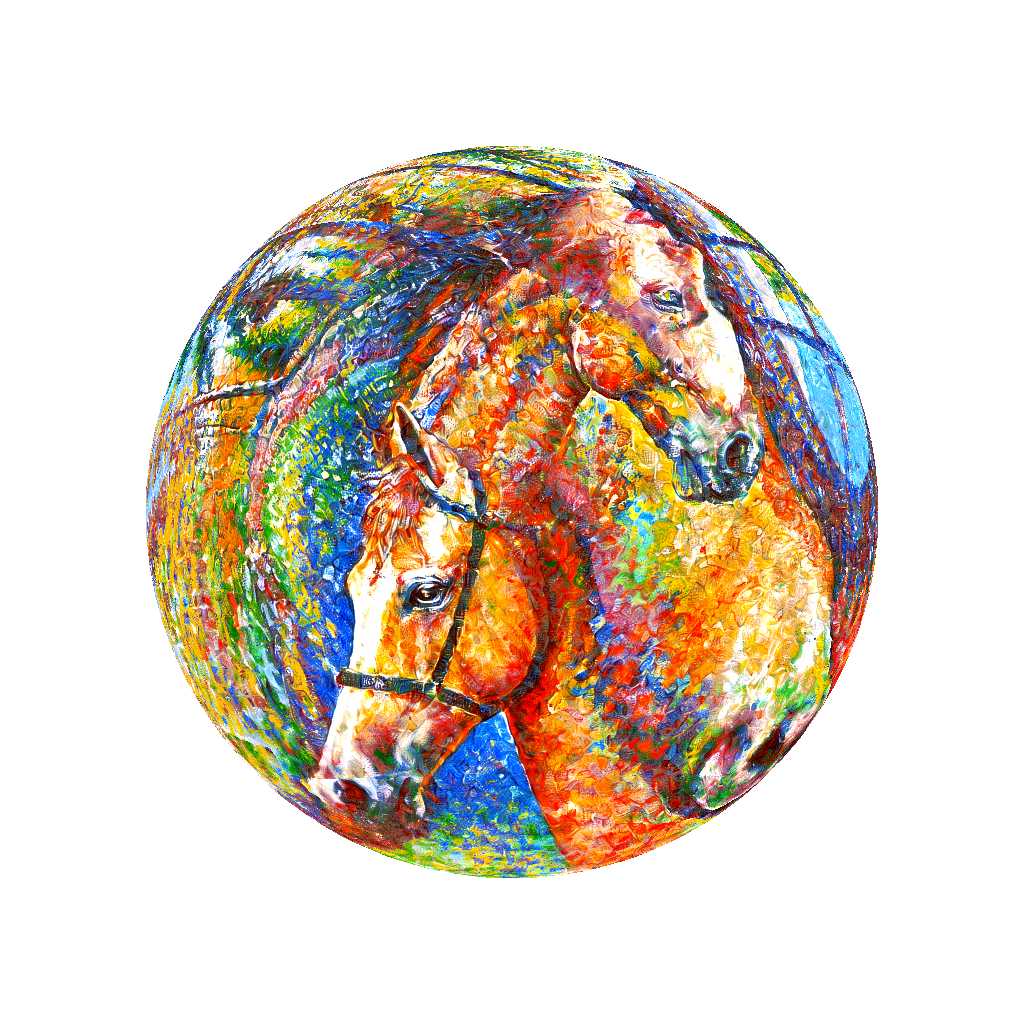} &
        \includegraphics[width=0.125\textwidth, trim=140 140 140 140, clip]{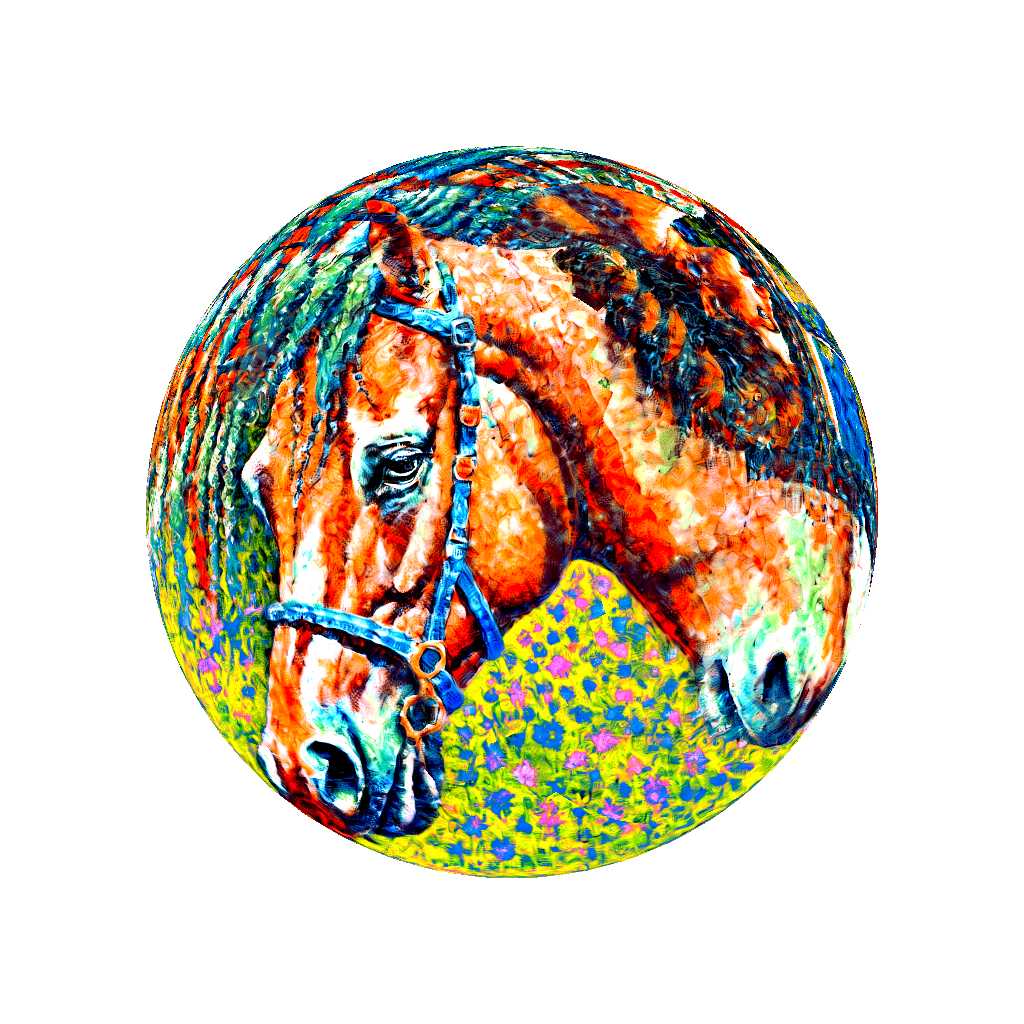} &
        \includegraphics[width=0.125\textwidth, trim=140 140 140 140, clip]{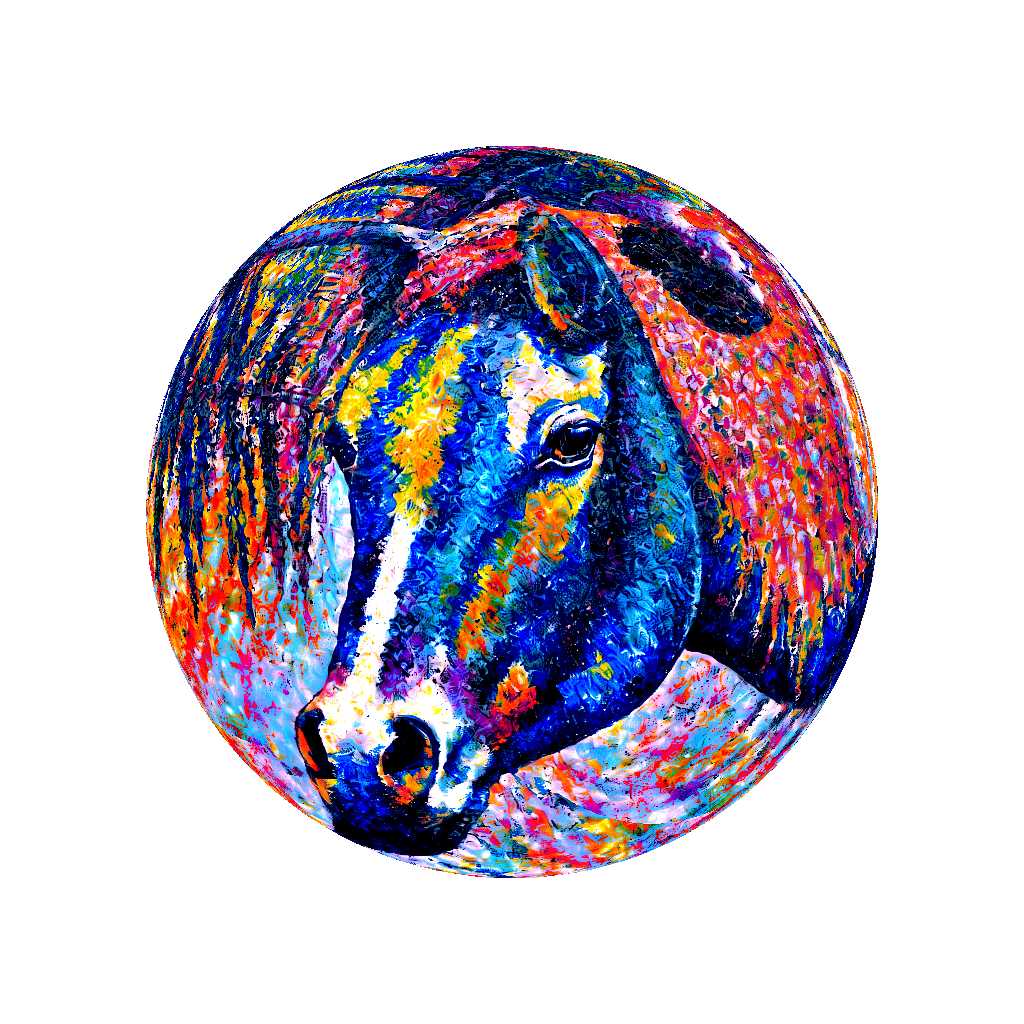} \\
        
        \rotatebox{90}{\includegraphics[width=0.125\textwidth, trim=150 20 150 30, clip]{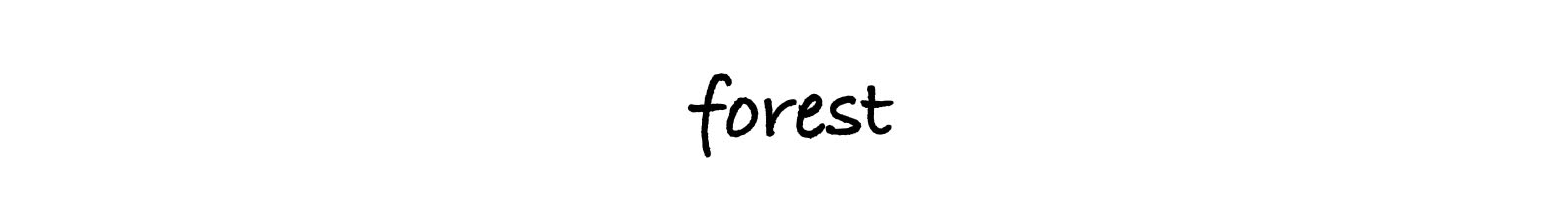}}
        & \includegraphics[width=0.125\textwidth, trim=70 70 70 70, clip]{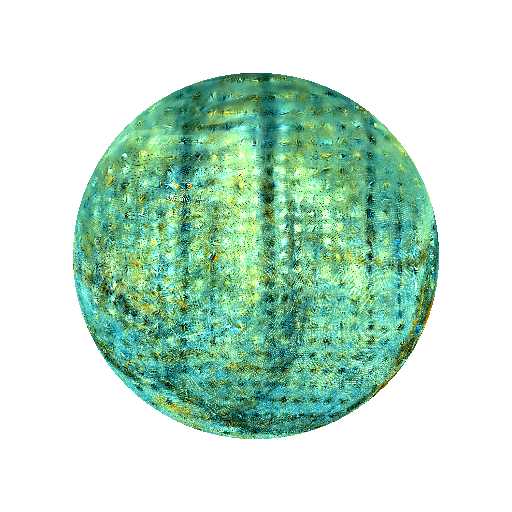} &
        \includegraphics[width=0.125\textwidth, trim=70 70 70 70, clip]{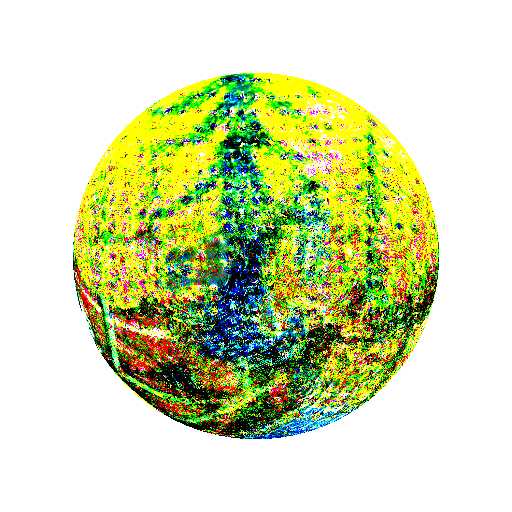} &
        \includegraphics[width=0.125\textwidth, trim=140 140 140 140, clip]{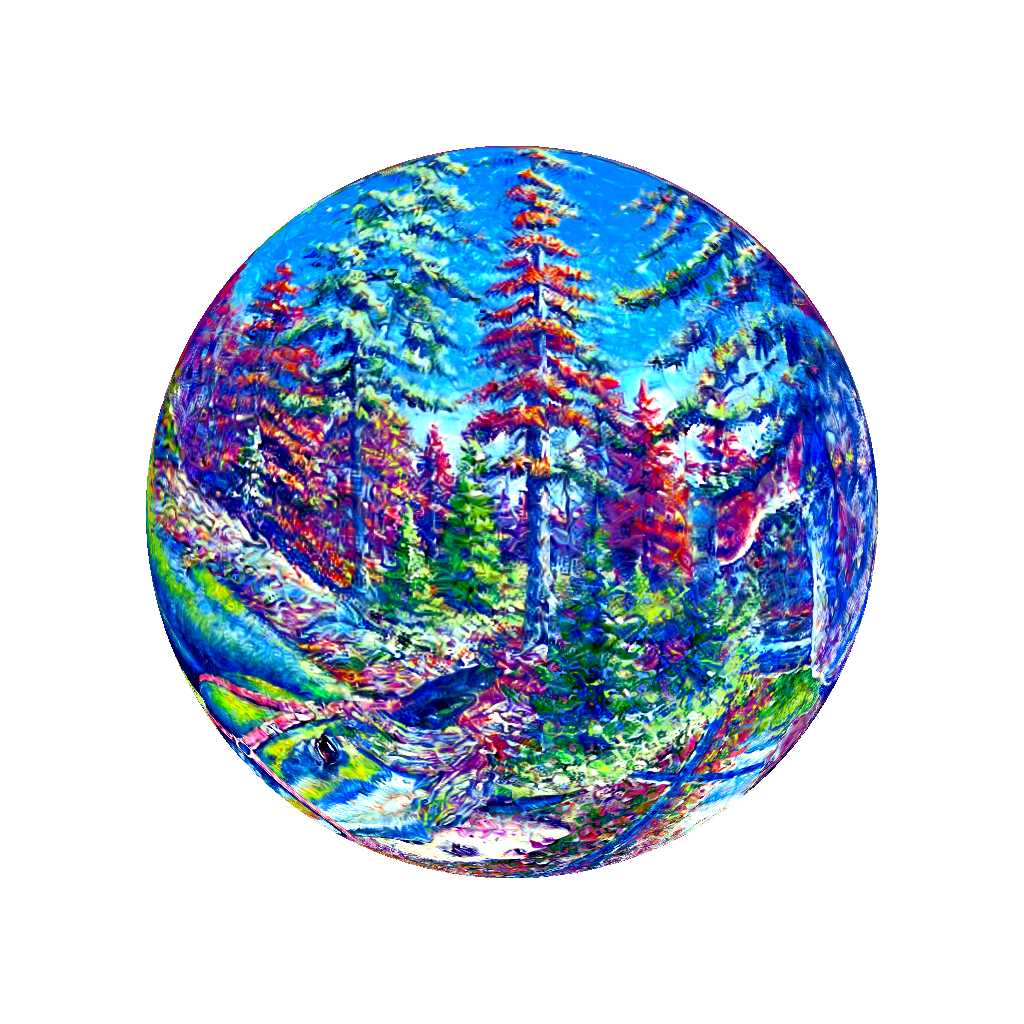} &
        \includegraphics[width=0.125\textwidth, trim=140 140 140 140, clip]{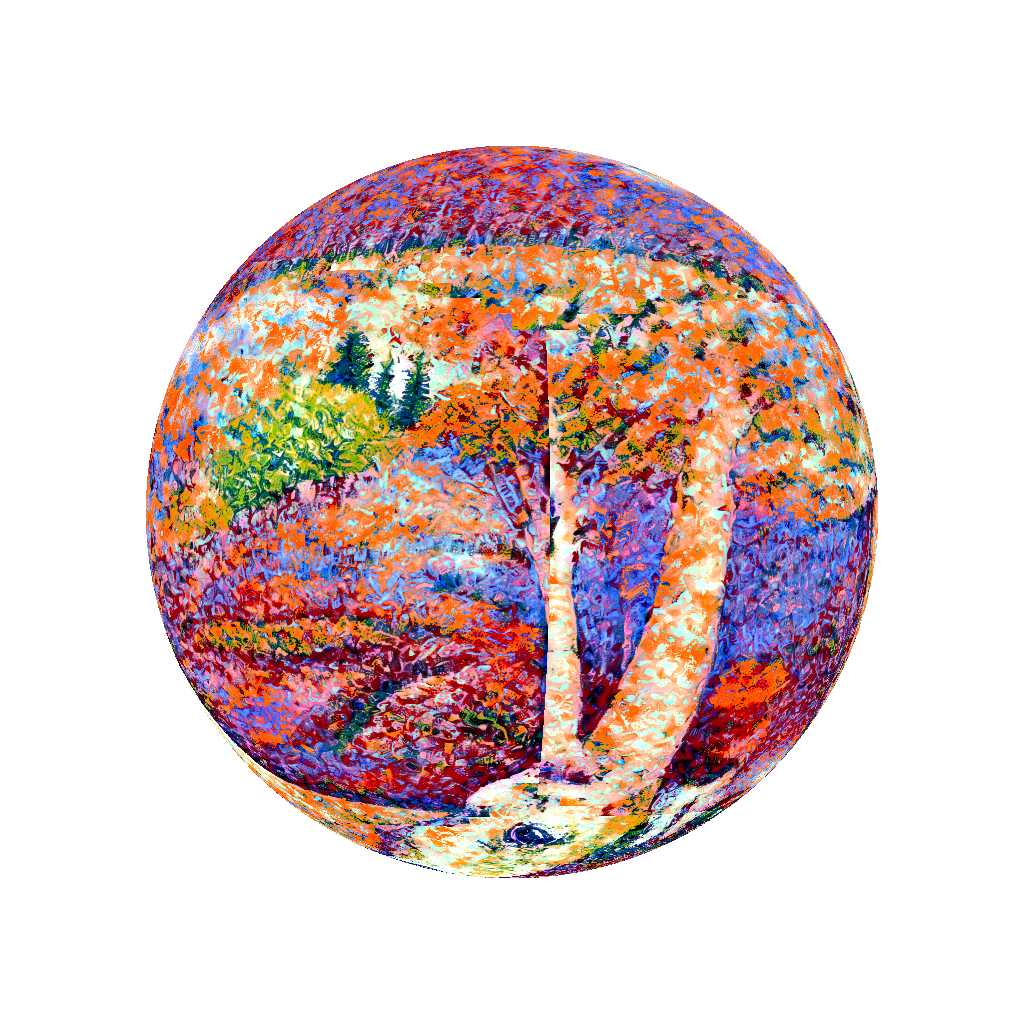} &
        \includegraphics[width=0.125\textwidth, trim=140 140 140 140, clip]{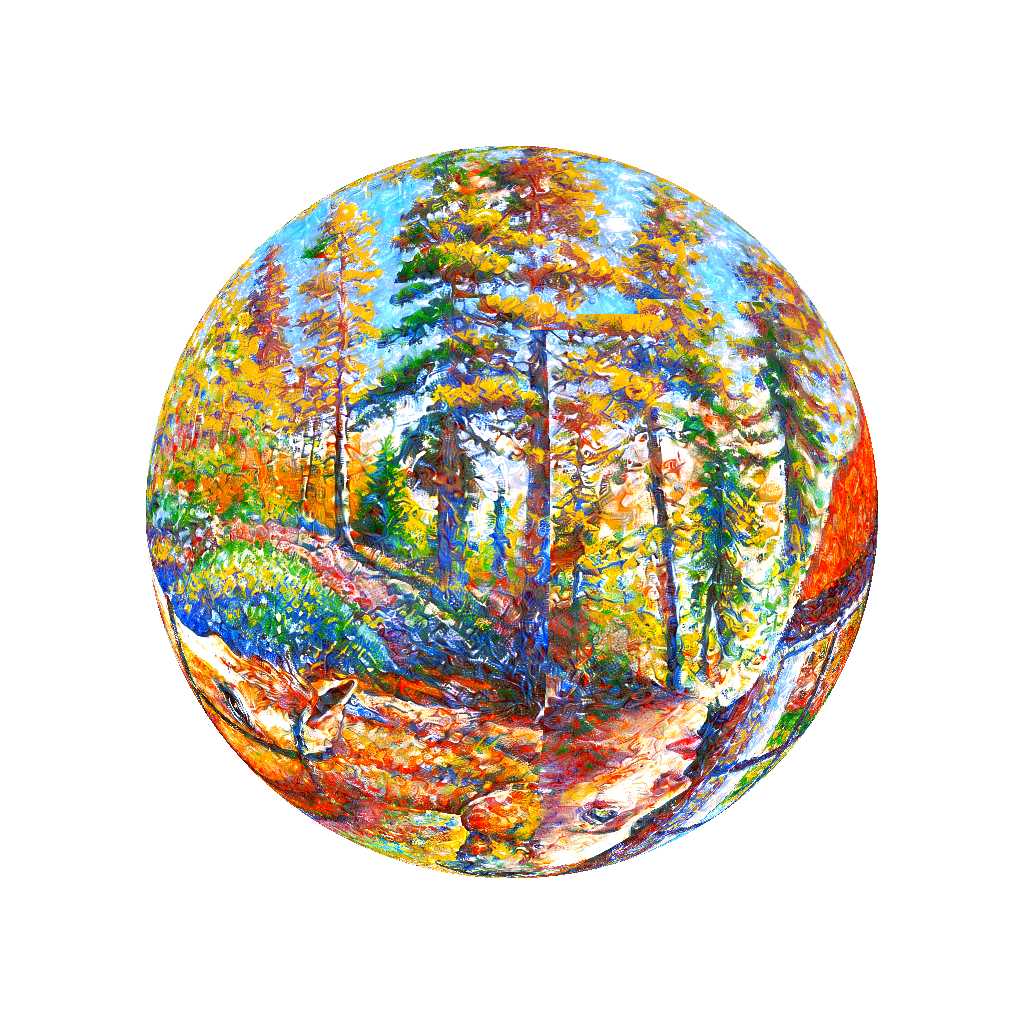} &
        \includegraphics[width=0.125\textwidth, trim=140 140 140 140, clip]{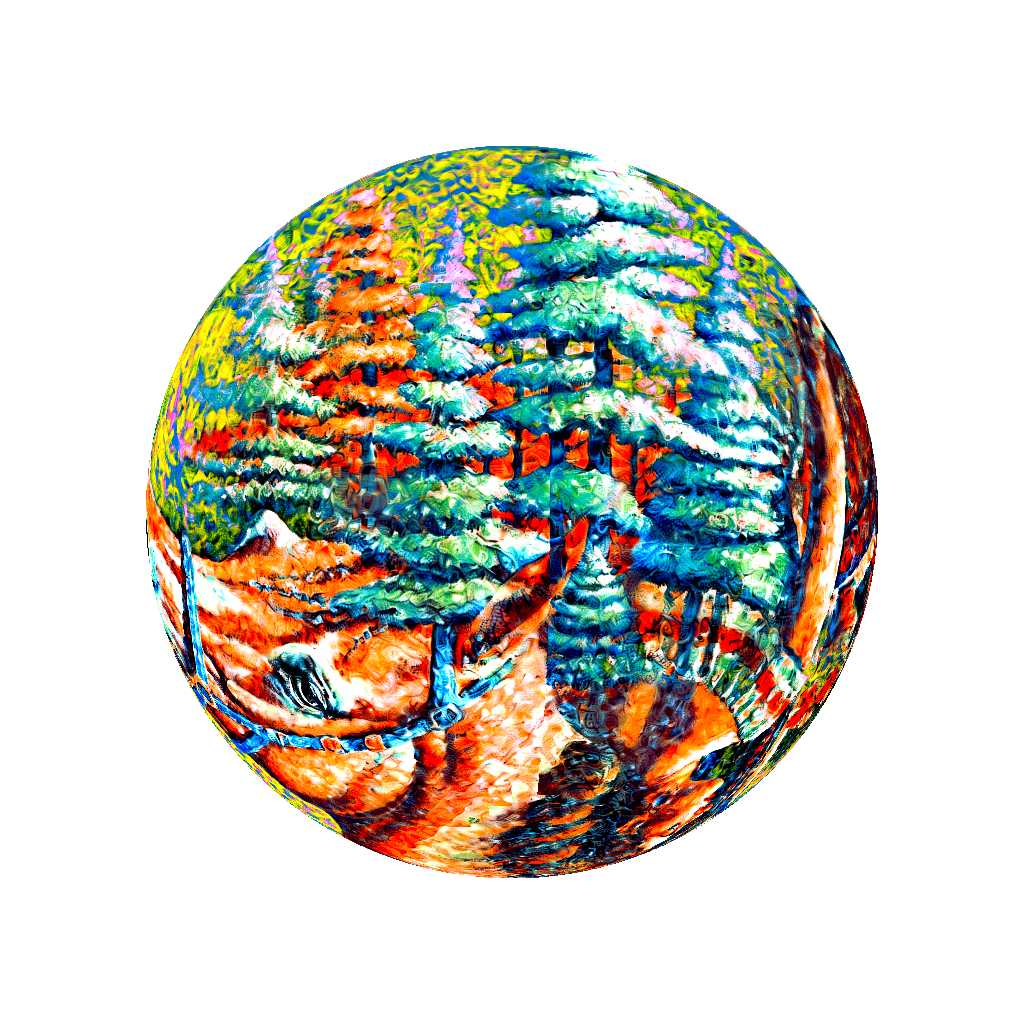} &
        \includegraphics[width=0.125\textwidth, trim=140 140 140 140, clip]{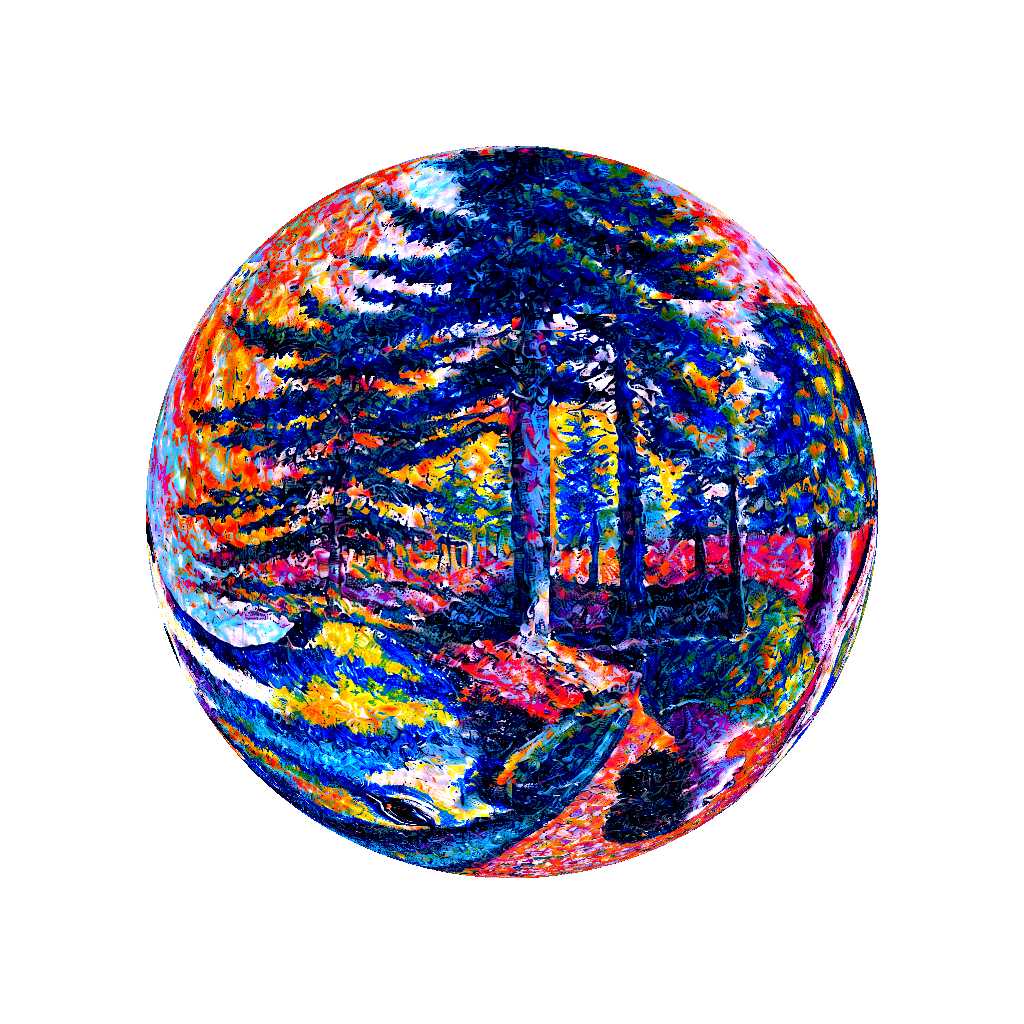} \\
        \toprule    
        \textbf{Camera jitter}        & \textcolor{red}{$\times$}& \textcolor{red}{$\times$}& \textcolor{red}{$\times$} & random & scheduled & \textcolor{red}{$\times$} & scheduled \\
        \textbf{Patch denoising}       &\textcolor{red}{$\times$} &\textcolor{red}{$\times$} & \textcolor{green}{$\checkmark$} & \textcolor{green}{$\checkmark$} & \textcolor{green}{$\checkmark$} & \textcolor{green}{$\checkmark$} & \textcolor{green}{$\checkmark$} \\
        \textbf{Resolution scaling}     & \textcolor{red}{$\times$}&\textcolor{red}{$\times$}& \textcolor{red}{$\times$} & \textcolor{red}{$\times$} & \textcolor{red}{$\times$} & \textcolor{green}{$\checkmark$} & \textcolor{green}{$\checkmark$}\\
        \bottomrule
    \end{tabular}
    
    \caption{\textbf{Comparison of different design choices on sphere case.} Our method can have the best primary content fusion while obtaining visual quality. The baseline method suffers null space of VAE encoder without camera jitter. Random patch denoising can improve the resolution of results while introducing multiple duplicate primary content. Random camera jitter can make the transition smoother, but it still has duplicate pattern artifacts. The scheduled camera jitter resolved the duplicate pattern issue a bit but couldn't resolve it. Progressive render resolution scaling can also smooth the result, but the fusion of the primary content is less and has minor duplicate pattern issues on the horse. Our method combines progressive resolution scaling with scheduled camera jitter to produce results that center the primary content and diminish duplicate pattern artifacts.}
    \label{fig:method-supp}
\end{figure*}

We show more examples of ablation among different methods in Fig. \ref{fig:method-supp} in addition to Fig. \ref{fig:comparison}. 

\begin{figure*}[htbp]
    \centering
    \setlength{\tabcolsep}{1pt} 
    \renewcommand{\arraystretch}{1}

    \begin{tabular}{cccccccc}

        \begin{minipage}[t]{0.12\textwidth}
            \includegraphics[width=\textwidth, trim=240 205 240 205, clip]{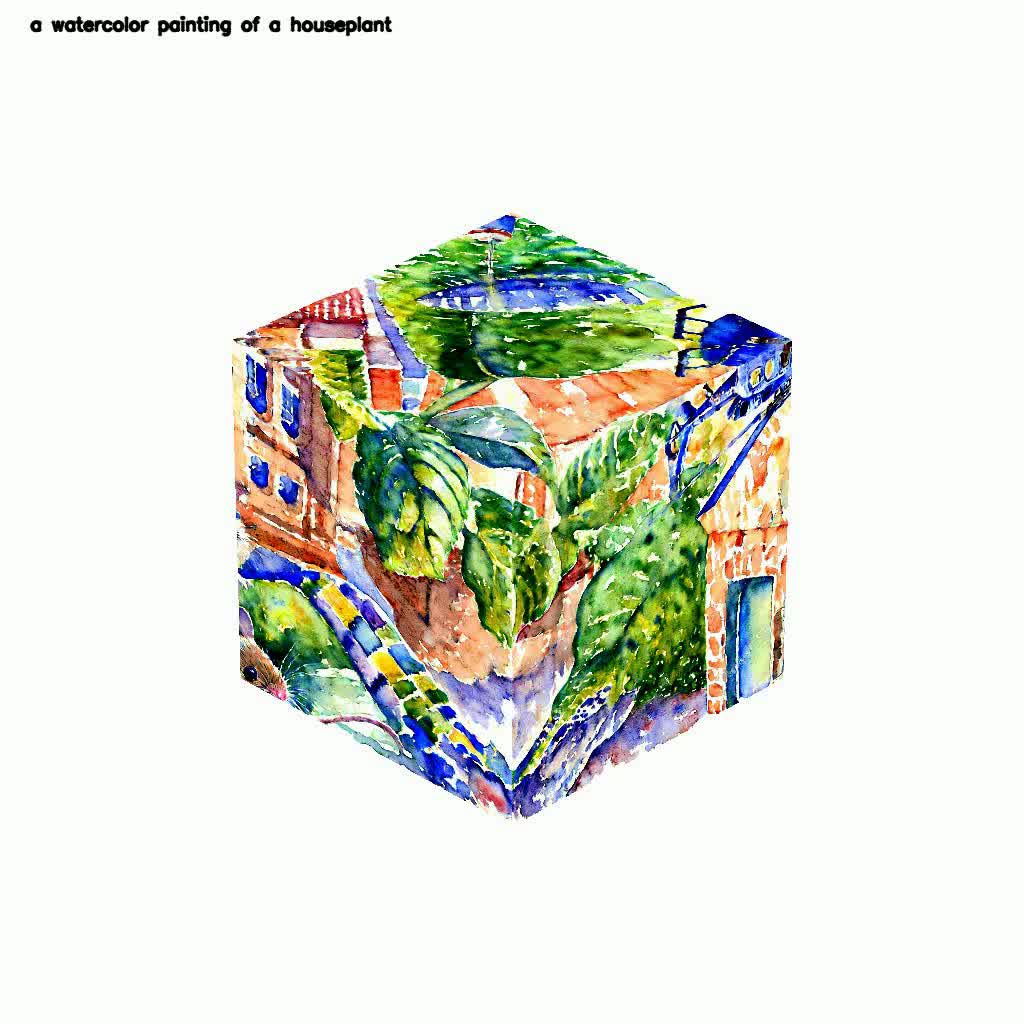}
        \end{minipage}%
        \begin{minipage}[t]{0.12\textwidth}
            \includegraphics[width=\textwidth, trim=240 205 240 205, clip]{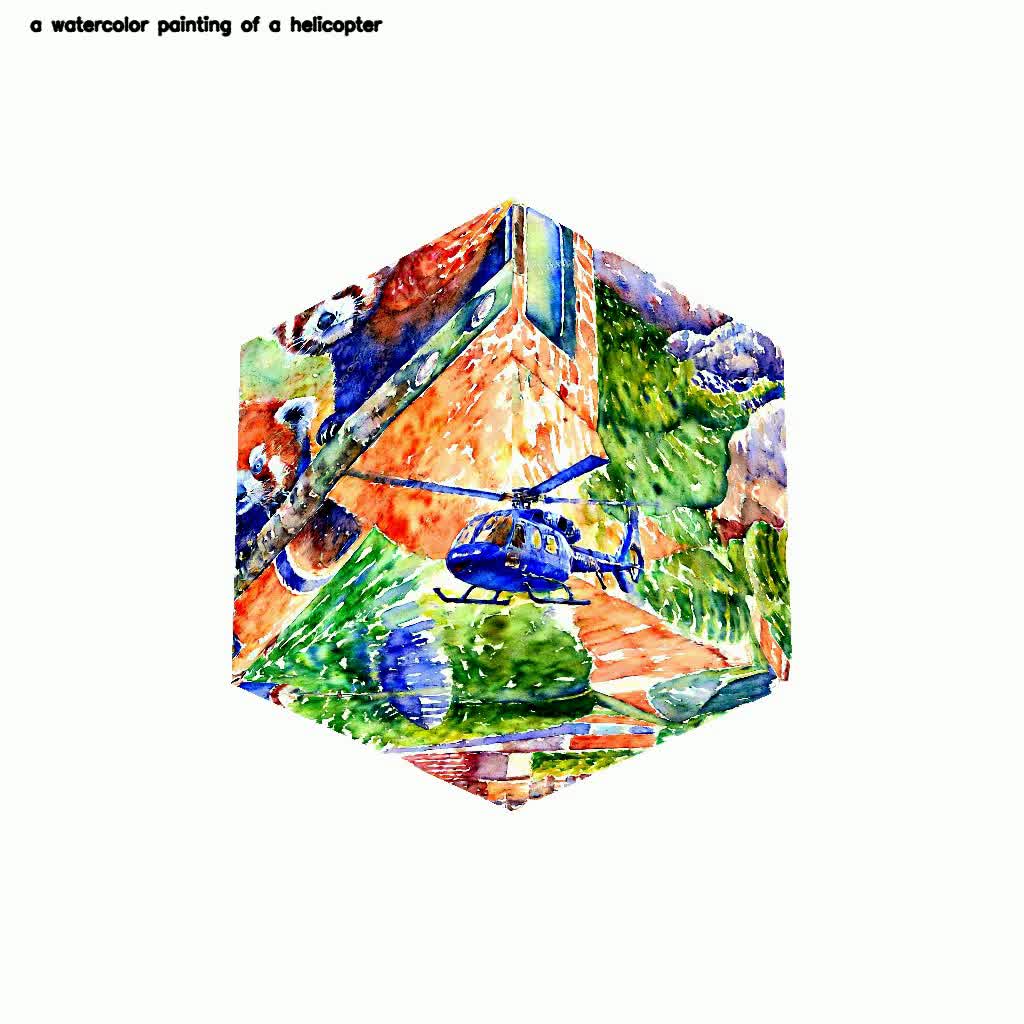}
        \end{minipage}%
        \begin{minipage}[t]{0.12\textwidth}
            \includegraphics[width=\textwidth, trim=240 205 240 205, clip]{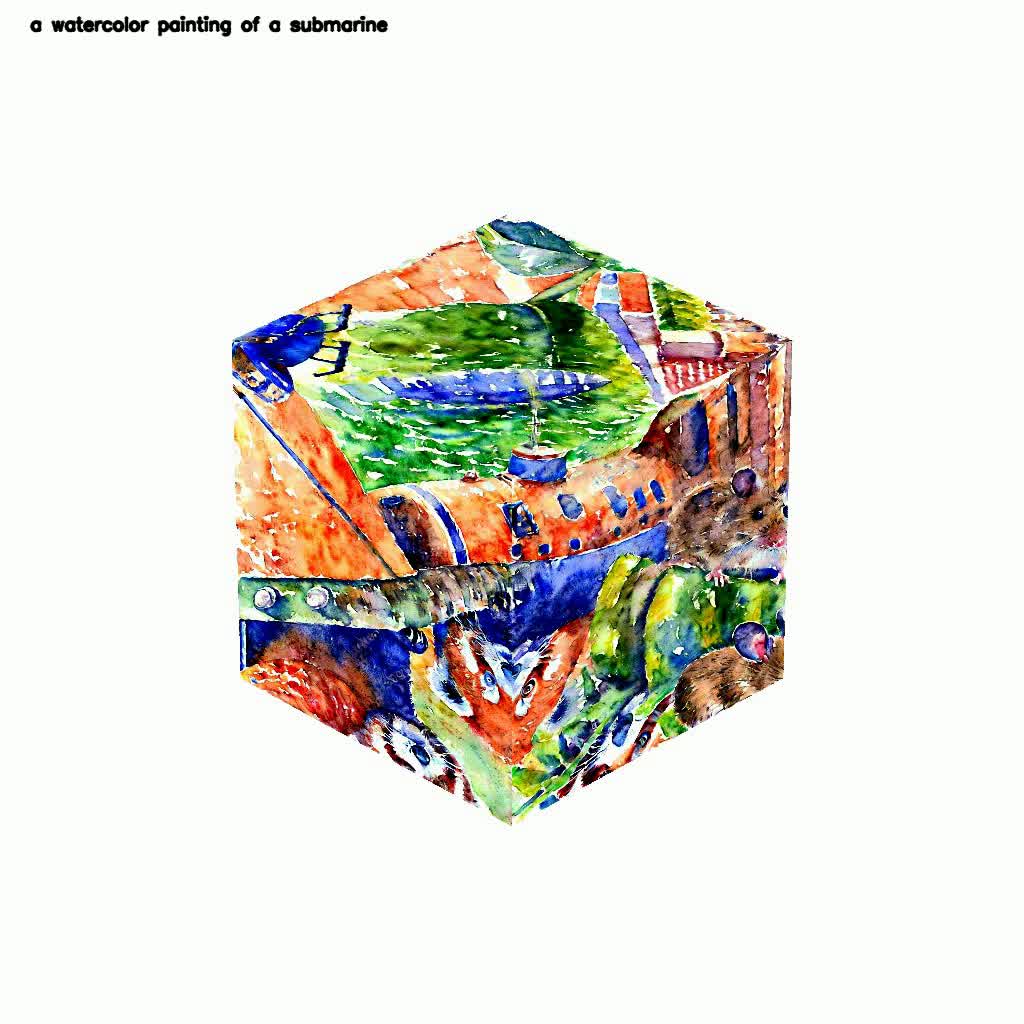}
        \end{minipage}%
        \begin{minipage}[t]{0.12\textwidth}
            \includegraphics[width=\textwidth, trim=240 205 240 205, clip]{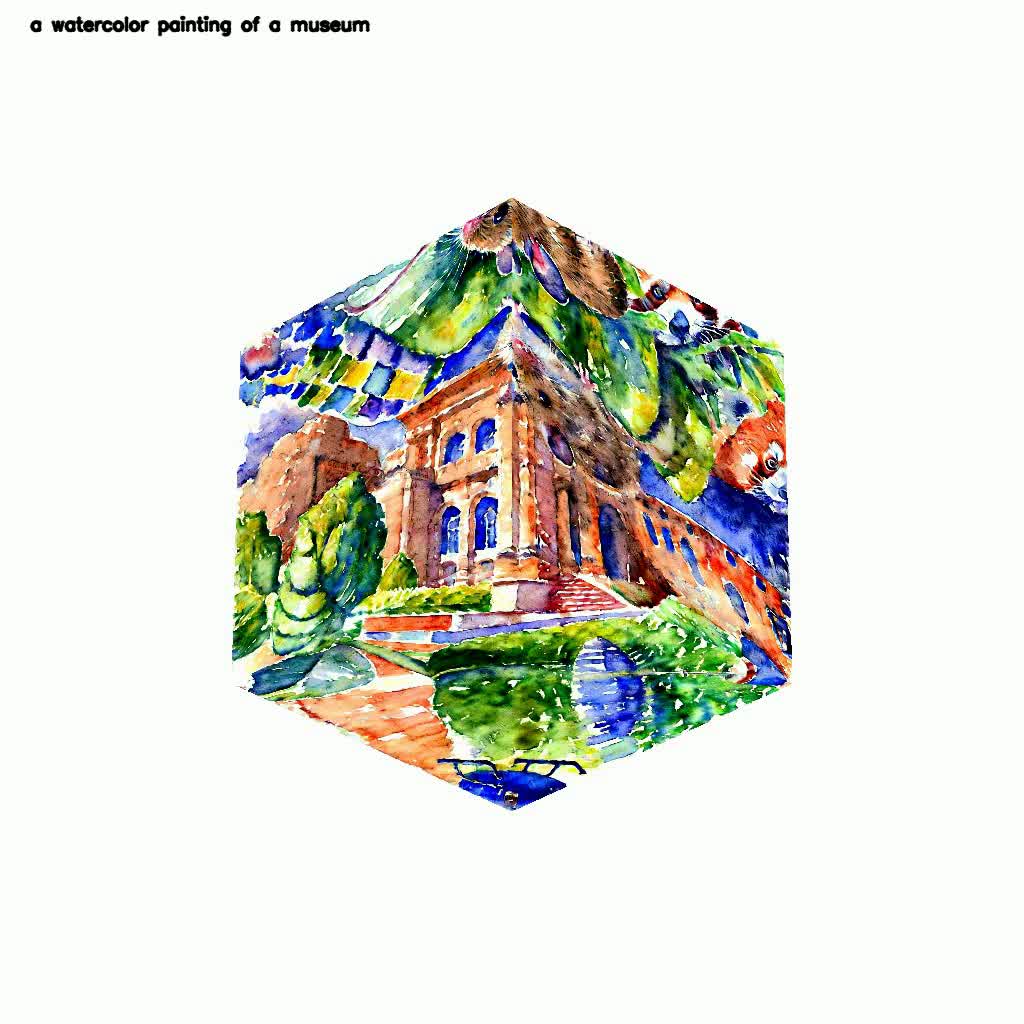}
        \end{minipage}%
        \begin{minipage}[t]{0.12\textwidth}
            \includegraphics[width=\textwidth, trim=240 205 240 205, clip]{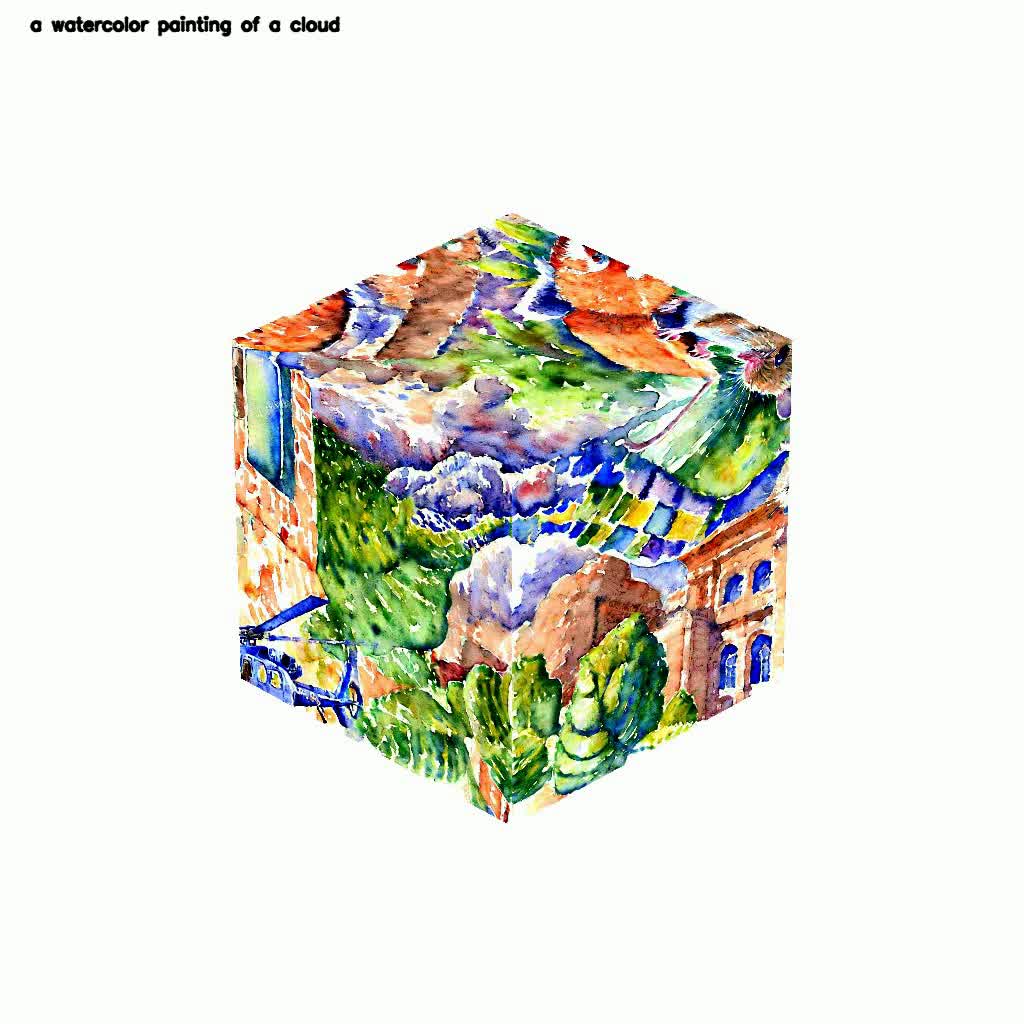}
        \end{minipage}%
        \begin{minipage}[t]{0.12\textwidth}
            \includegraphics[width=\textwidth, trim=240 205 240 205, clip]{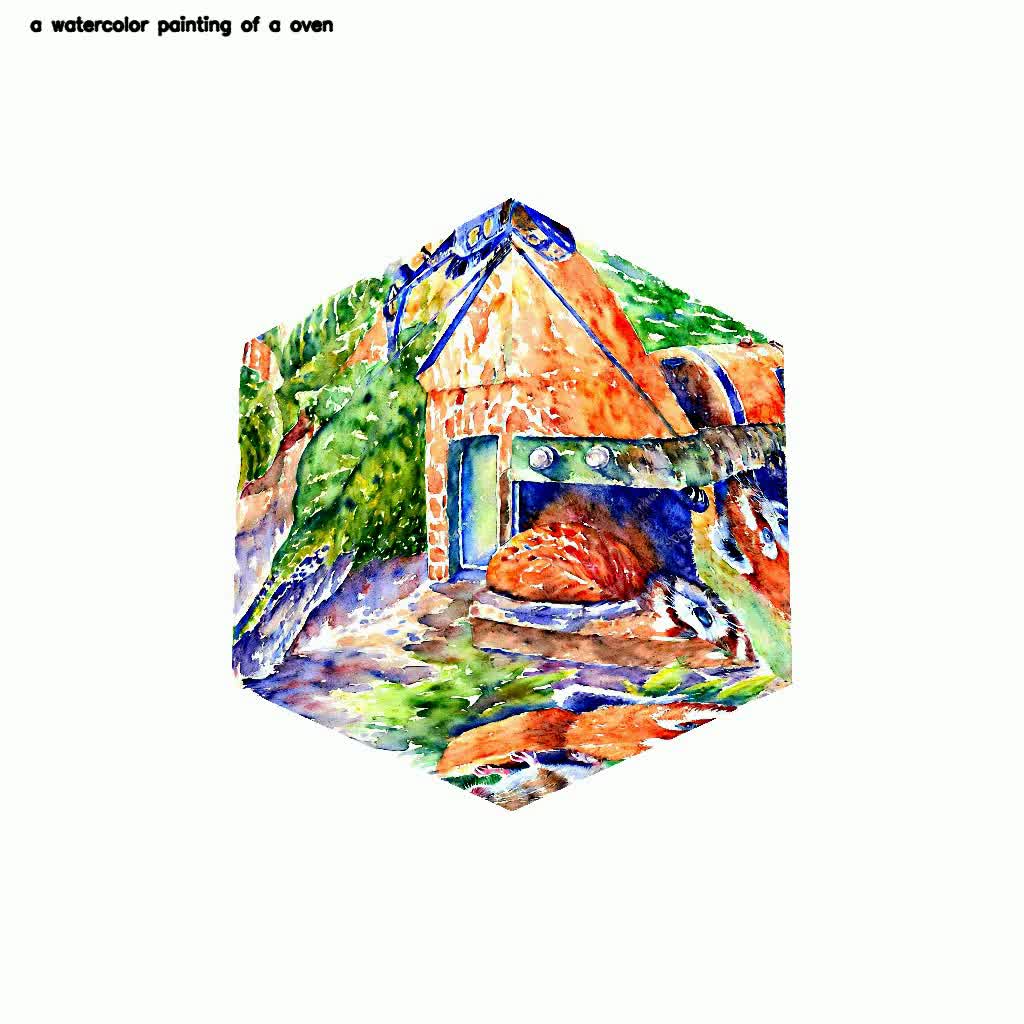}
        \end{minipage}%
        \begin{minipage}[t]{0.12\textwidth}
            \includegraphics[width=\textwidth, trim=240 205 240 205, clip]{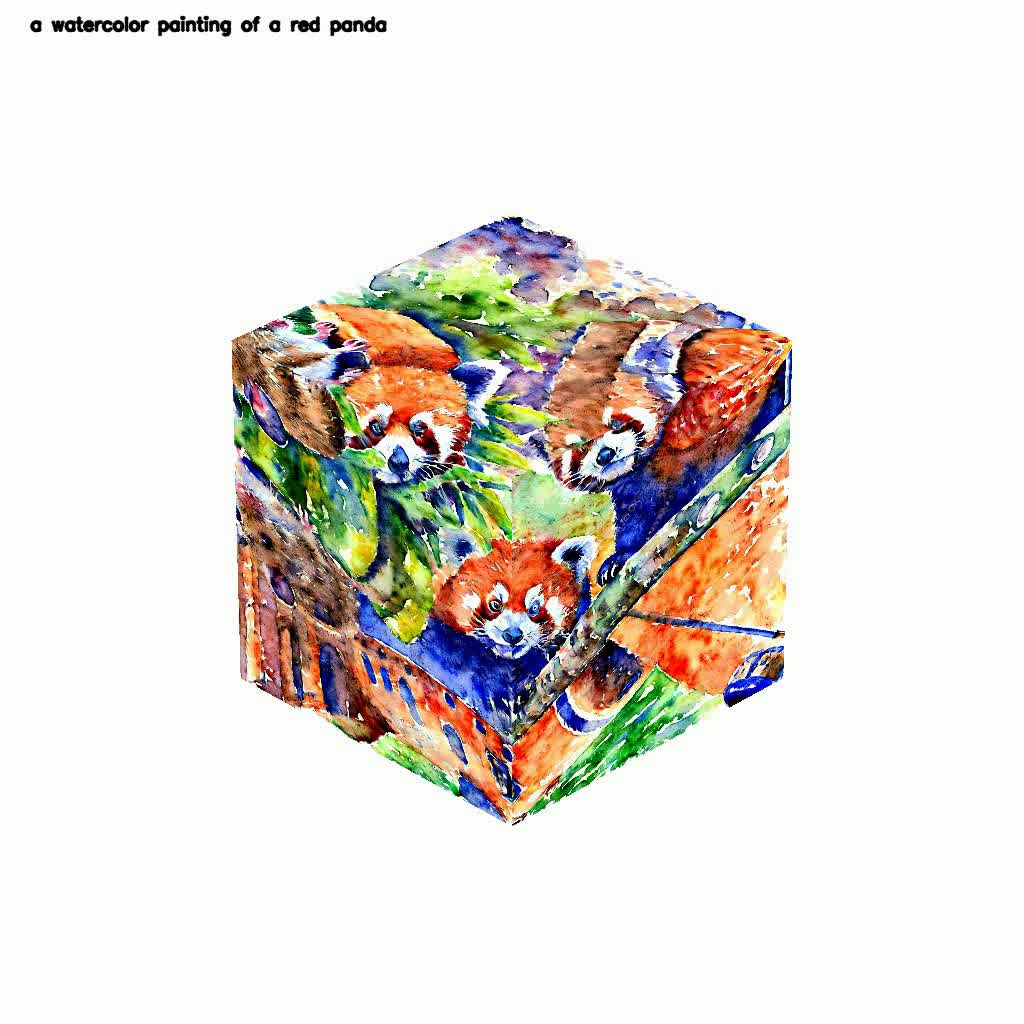}
        \end{minipage}%
        \begin{minipage}[t]{0.12\textwidth}
            \includegraphics[width=\textwidth, trim=240 205 240 205, clip]{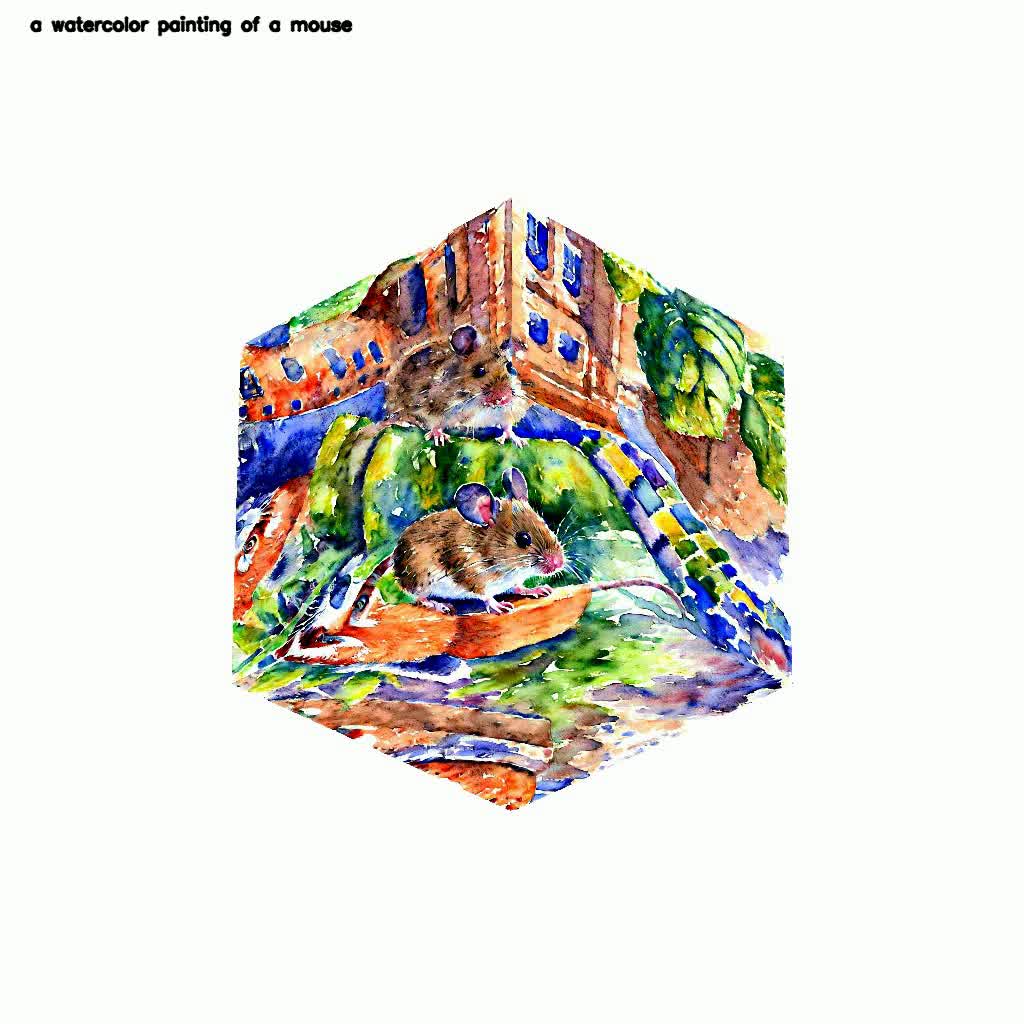}
        \end{minipage}\\

        \begin{minipage}[t]{0.12\textwidth}
        \centering
            \includegraphics[width=\linewidth, trim=165 20 165 30, clip]{figures/prompt/houseplant.jpg}
            \end{minipage}\hfill
        \begin{minipage}[t]{0.12\textwidth}
            \centering
            \includegraphics[width=\linewidth, trim=165 20 165 30, clip]{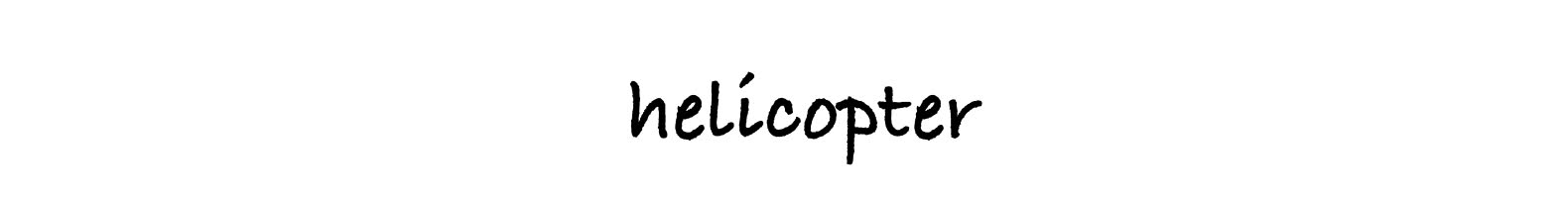}
            \end{minipage}\hfill
        \begin{minipage}[t]{0.12\textwidth}
            \centering
            \includegraphics[width=\linewidth, trim=165 20 165 30, clip]{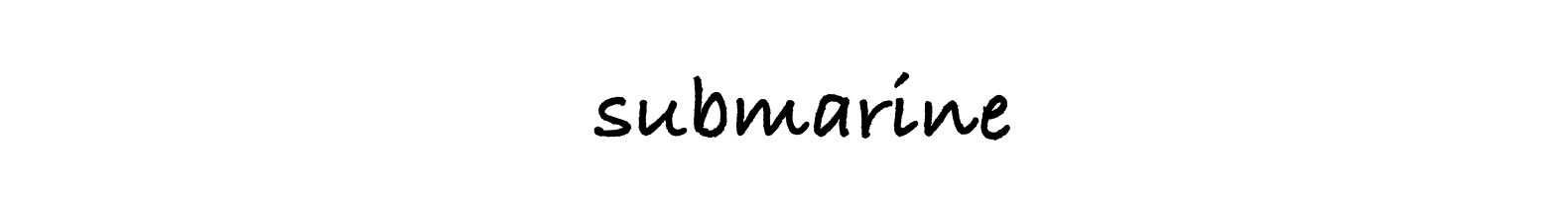}
        \end{minipage}\hfill
        \begin{minipage}[t]{0.12\textwidth}
            \centering
            \includegraphics[width=\linewidth, trim=165 20 165 30, clip]{figures/prompt/museum.jpg}
        \end{minipage}
        \begin{minipage}[t]{0.12\textwidth}
            \centering
            \includegraphics[width=\linewidth, trim=165 20 165 30, clip]{figures/prompt/cloud.jpg}
        \end{minipage}
        \begin{minipage}[t]{0.12\textwidth}
            \centering
            \includegraphics[width=\linewidth, trim=165 20 165 30, clip]{figures/prompt/oven.jpg}
        \end{minipage}
        \begin{minipage}[t]{0.12\textwidth}
            \centering
            \includegraphics[width=\linewidth, trim=165 20 165 30, clip]{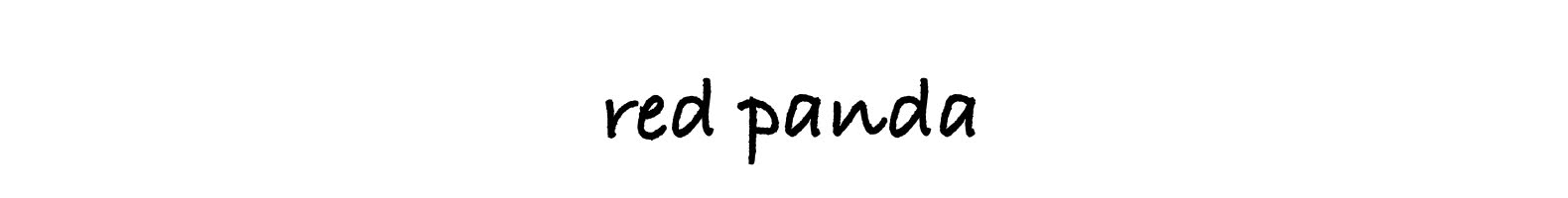}
        \end{minipage}
        \begin{minipage}[t]{0.12\textwidth}
            \centering
            \includegraphics[width=\linewidth, trim=165 20 165 30, clip]{figures/prompt/mouse.jpg}
        \end{minipage}\\

        \begin{minipage}[t]{0.12\textwidth}
            \includegraphics[width=\textwidth, trim=145 145 145 145, clip]{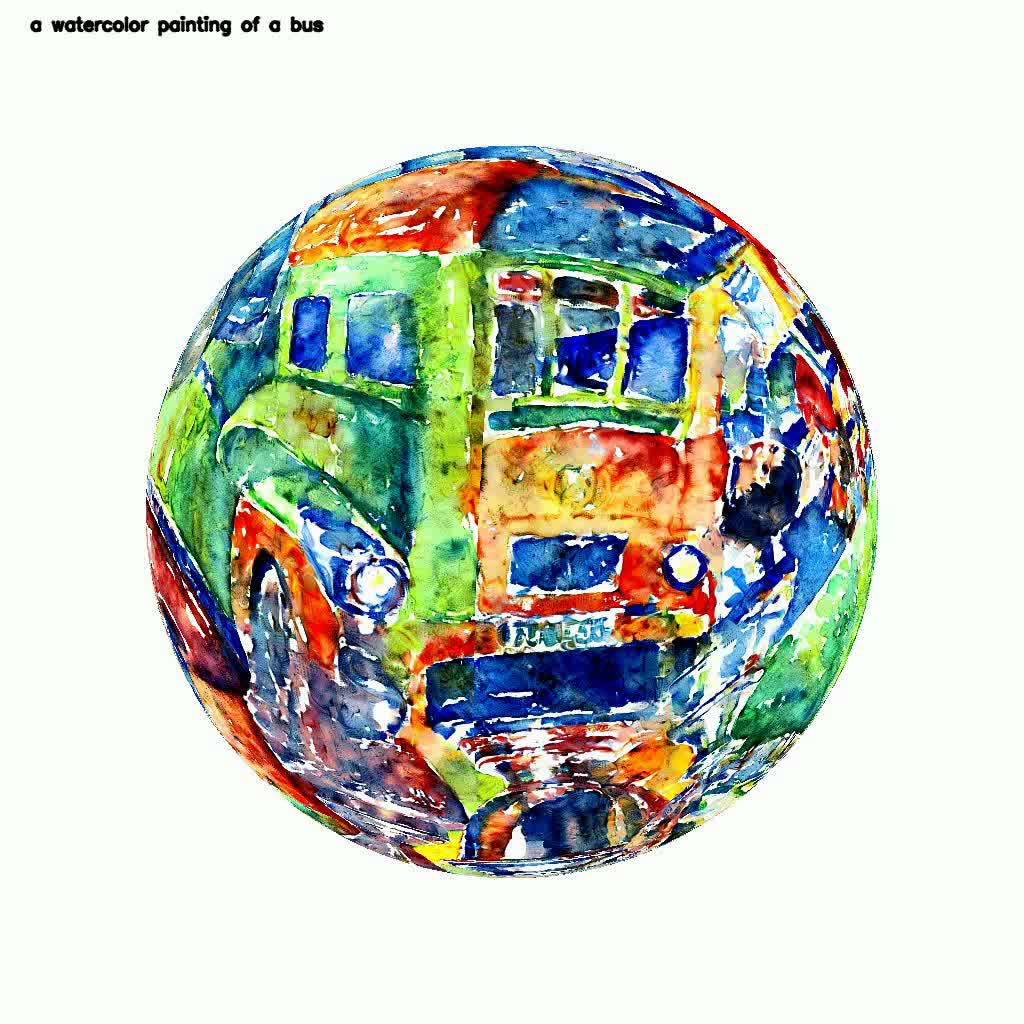}
        \end{minipage}%
        \begin{minipage}[t]{0.12\textwidth}
            \includegraphics[width=\textwidth, trim=145 145 145 145, clip]{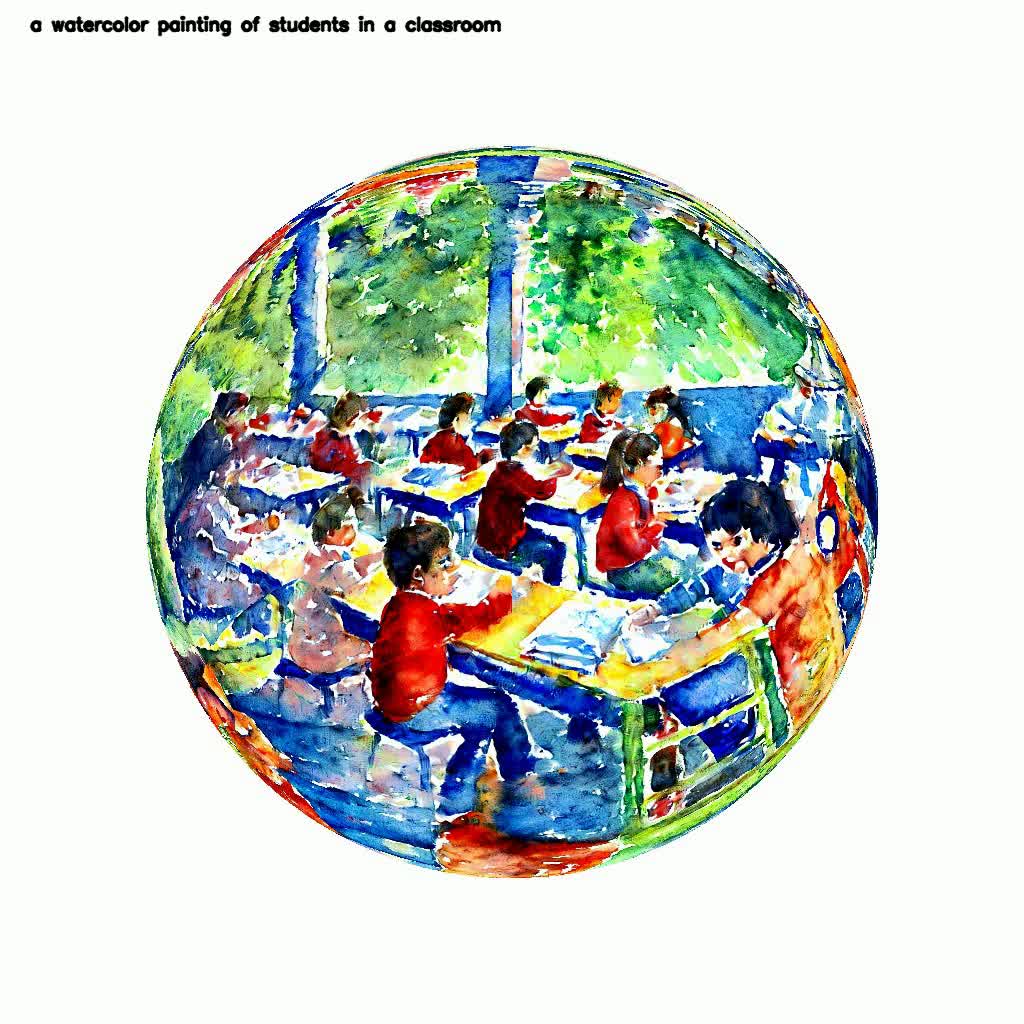}
        \end{minipage}%
        \begin{minipage}[t]{0.12\textwidth}
            \includegraphics[width=\textwidth, trim=145 145 145 145, clip]{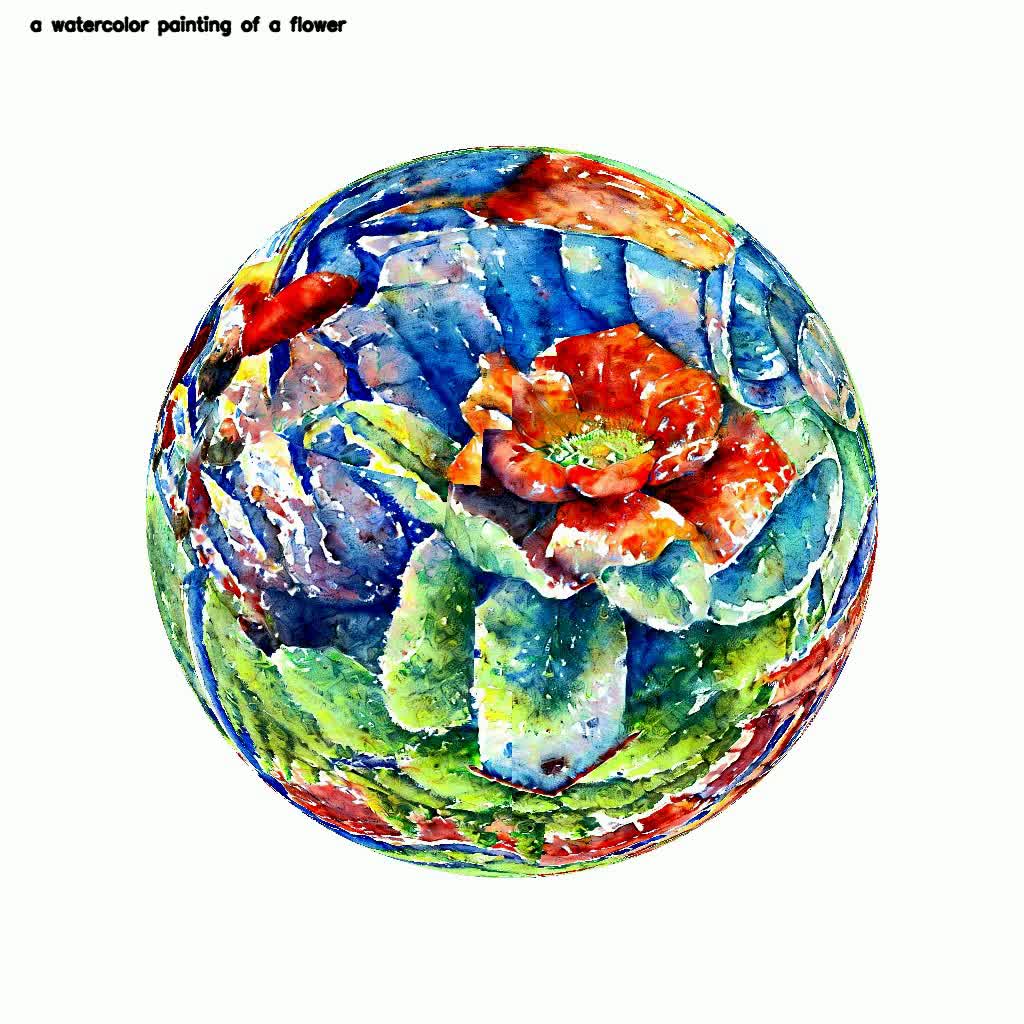}
        \end{minipage}%
        \begin{minipage}[t]{0.12\textwidth}
            \includegraphics[width=\textwidth, trim=145 145 145 145, clip]{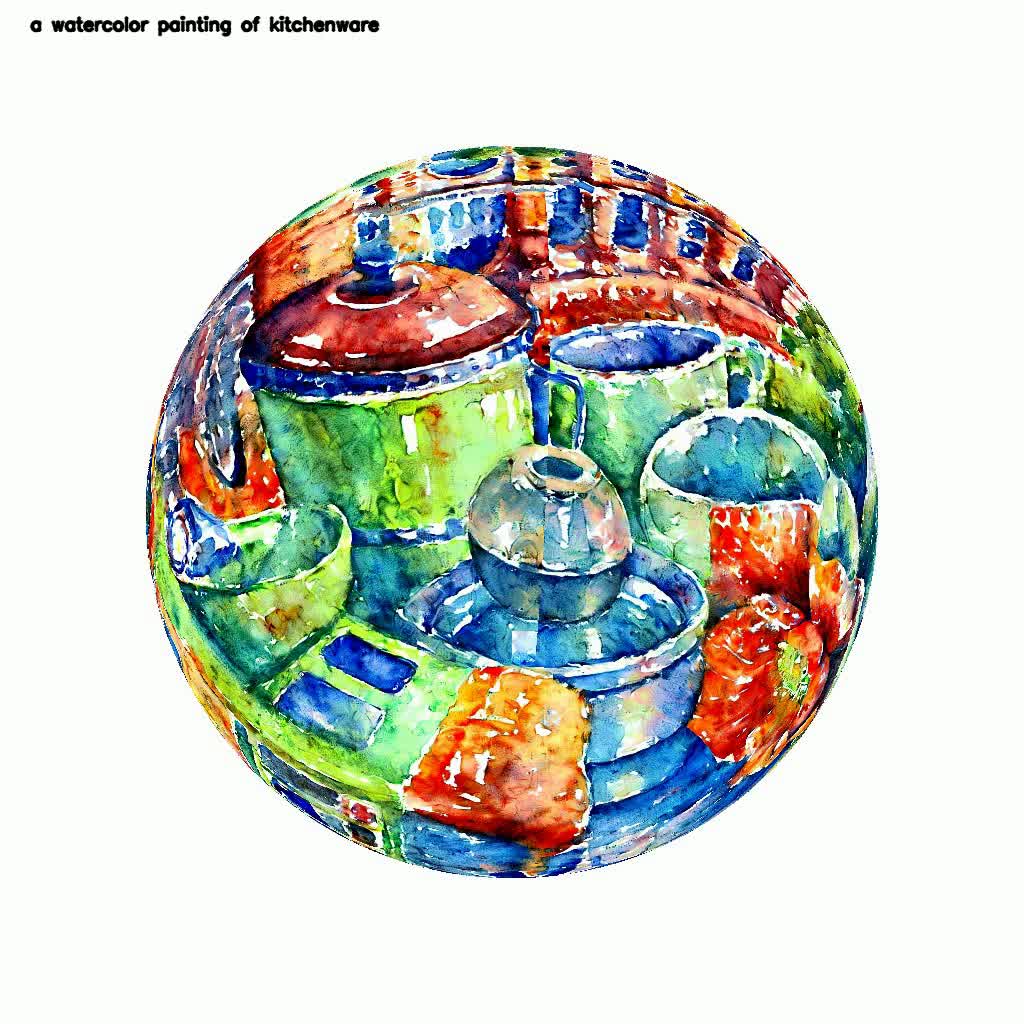}
        \end{minipage}%
        \begin{minipage}[t]{0.12\textwidth}
            \includegraphics[width=\textwidth, trim=145 145 145 145, clip]{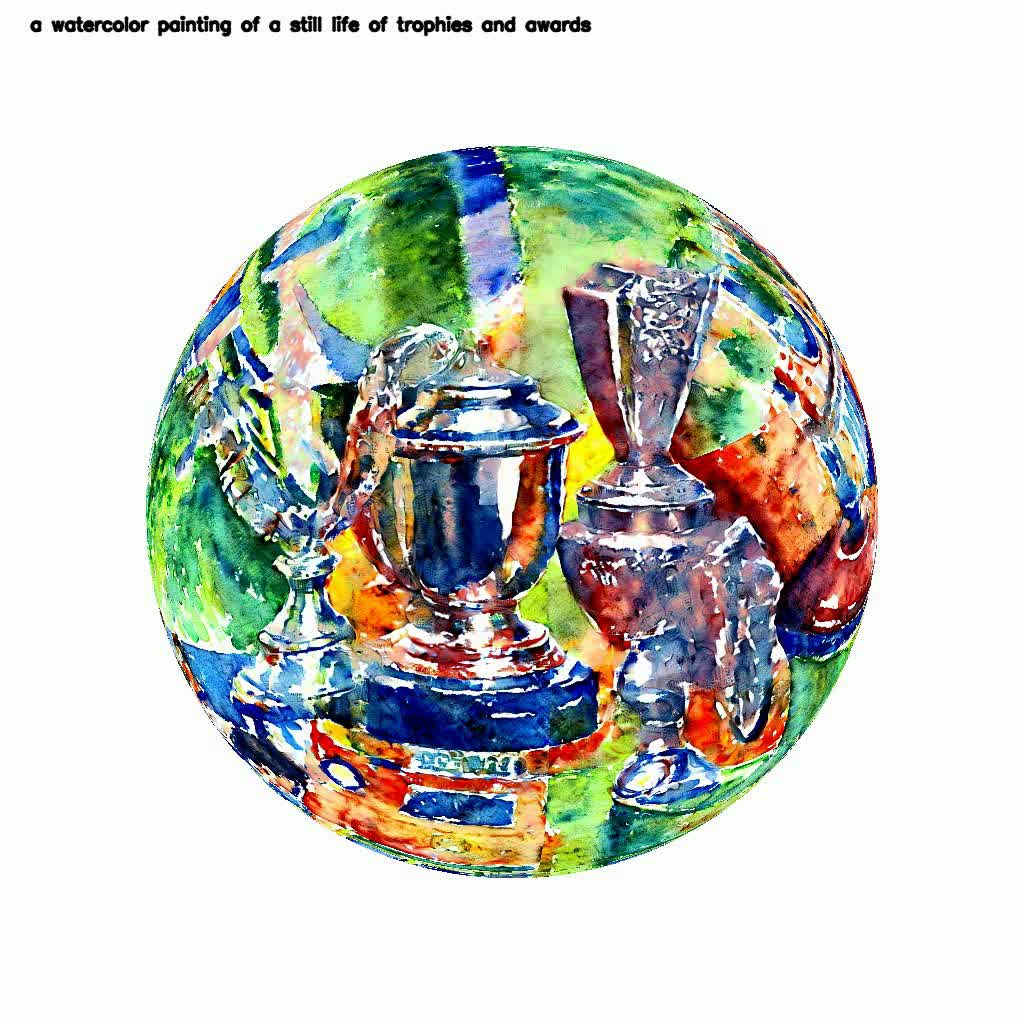}
        \end{minipage}%
        \begin{minipage}[t]{0.12\textwidth}
            \includegraphics[width=\textwidth, trim=145 145 145 145, clip]{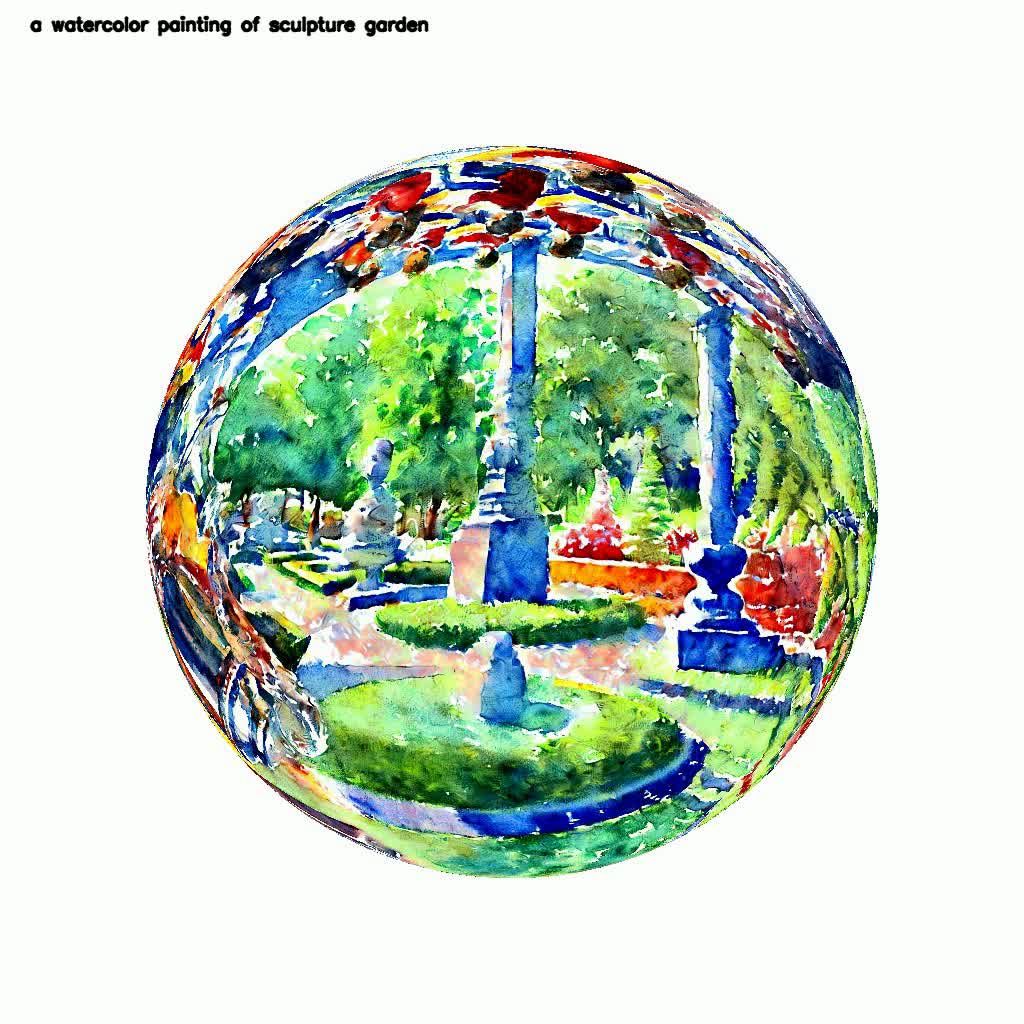}
        \end{minipage}%
        \begin{minipage}[t]{0.12\textwidth}
            \includegraphics[width=\textwidth, trim=145 145 145 145, clip]{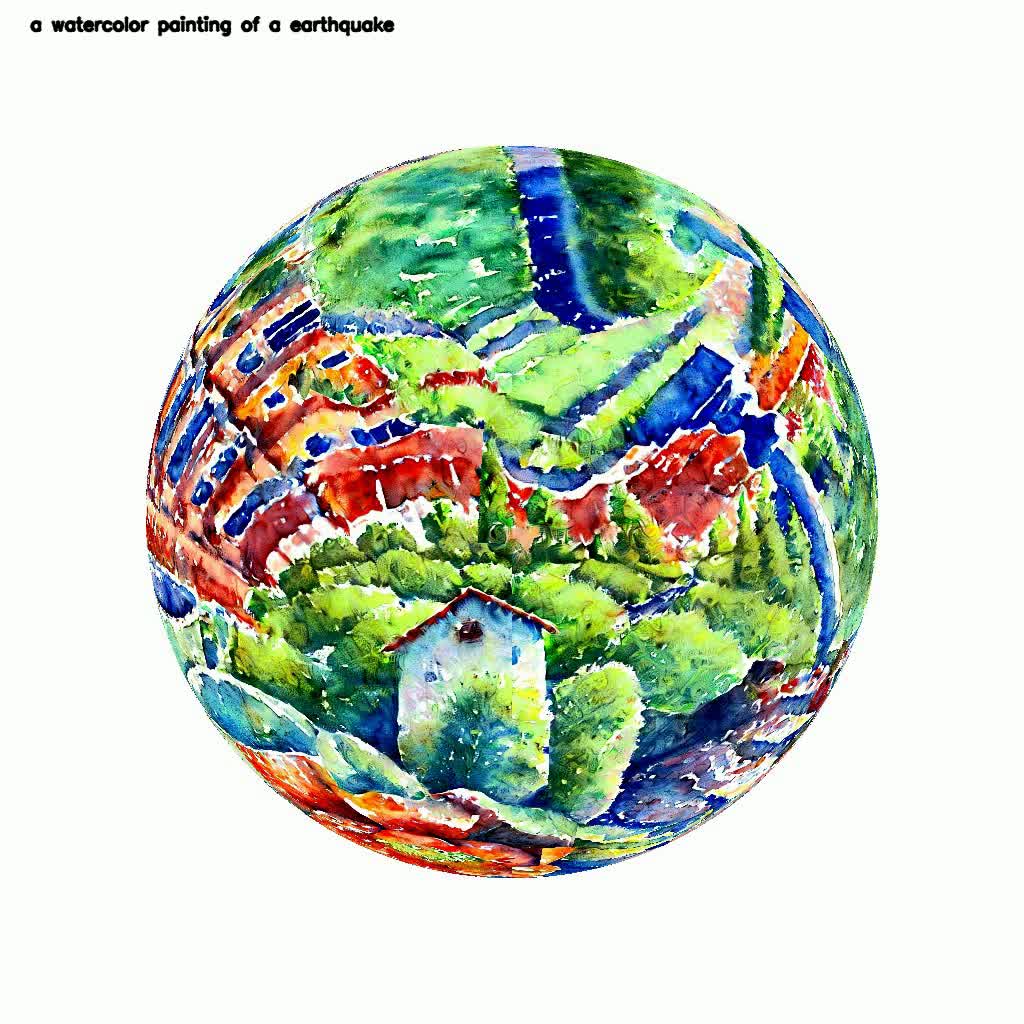}
        \end{minipage}%
        \begin{minipage}[t]{0.12\textwidth}
            \includegraphics[width=\textwidth, trim=145 145 145 145, clip]{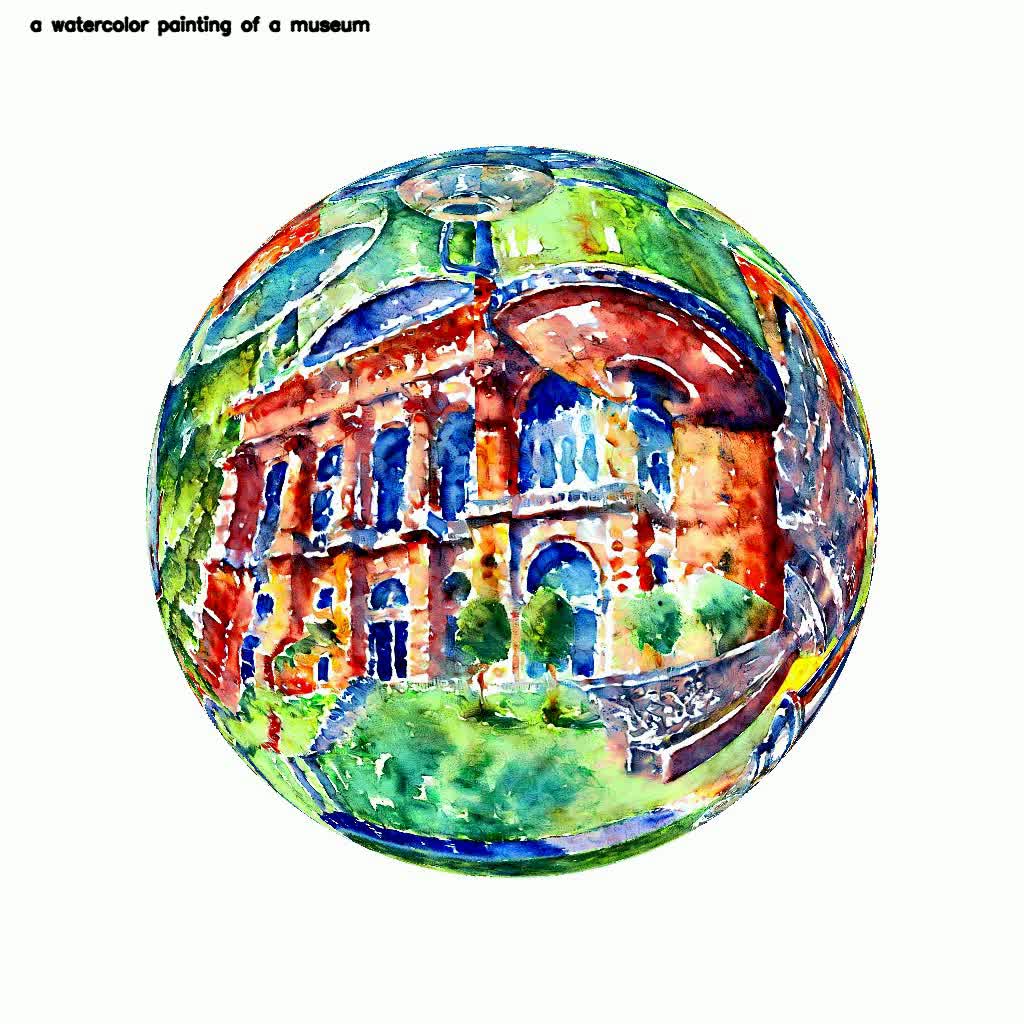}
        \end{minipage}\\

        \begin{minipage}[t]{0.12\textwidth}
        \centering
            \includegraphics[width=\linewidth, trim=165 20 165 30, clip]{figures/prompt/bus.jpg}
            \end{minipage}\hfill
        \begin{minipage}[t]{0.12\textwidth}
            \centering
            \includegraphics[width=\linewidth, trim=165 20 165 30, clip]{figures/prompt/students.jpg}
            \end{minipage}\hfill
        \begin{minipage}[t]{0.12\textwidth}
            \centering
            \includegraphics[width=\linewidth, trim=165 20 165 30, clip]{figures/prompt/flower.jpg}
        \end{minipage}\hfill
        \begin{minipage}[t]{0.12\textwidth}
            \centering
            \includegraphics[width=\linewidth, trim=165 20 165 30, clip]{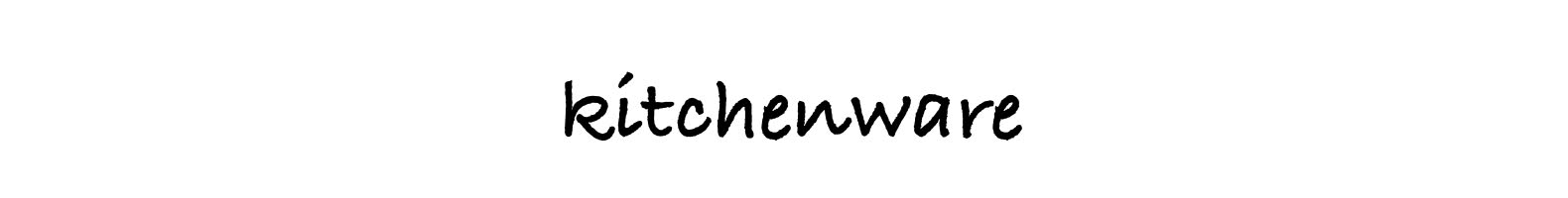}
        \end{minipage}
        \begin{minipage}[t]{0.12\textwidth}
            \centering
            \includegraphics[width=\linewidth, trim=165 20 165 30, clip]{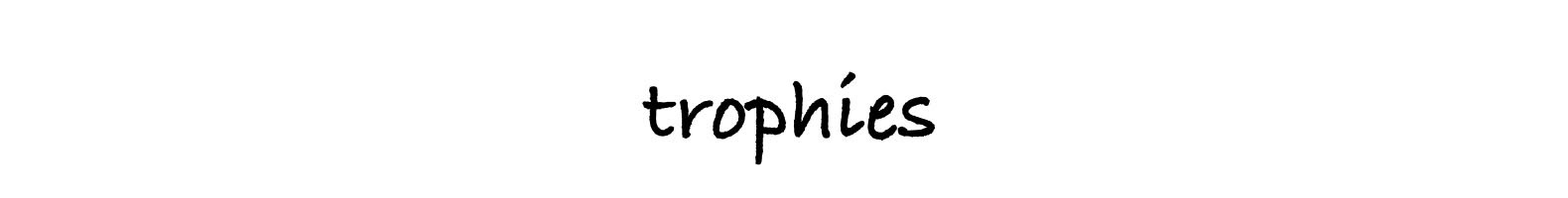}
        \end{minipage}
        \begin{minipage}[t]{0.12\textwidth}
            \centering
            \includegraphics[width=\linewidth, trim=165 20 165 30, clip]{figures/prompt/sculpturegarden.jpg}
        \end{minipage}
        \begin{minipage}[t]{0.12\textwidth}
            \centering
            \includegraphics[width=\linewidth, trim=165 20 165 30, clip]{figures/prompt/earthquake.jpg}
        \end{minipage}
        \begin{minipage}[t]{0.12\textwidth}
            \centering
            \includegraphics[width=\linewidth, trim=165 20 165 30, clip]{figures/prompt/museum.jpg}
        \end{minipage}\\

    \end{tabular}

    \caption{\textbf{Ablation on more views.} We extend the multiview illusion to 8 corners of a cube, each view containing three faces of a cube, and each adjacent corner has {$180^\circ$} flip viewing directions. We demonstrate our ability to generate illusions on more views on cube and sphere cases. }
    \label{fig:8prompts}
\end{figure*}

\topic{More views}
To push the ability to generate illusion, we increase the number of views to a more challenging setting in Fig. \ref{fig:8prompts}. We look at every corner of the cube, and for each view, we look at three faces. In total, we have eight views of a cube, and each face is shared by four corners, i.e., four prompts. Videos can be found at the ablation part in \href{index.html}{index.html}.

\begin{figure}
    \centering
    \vspace{-8mm}

    \includegraphics[width=0.28\linewidth, trim=0 0 0 0, clip]{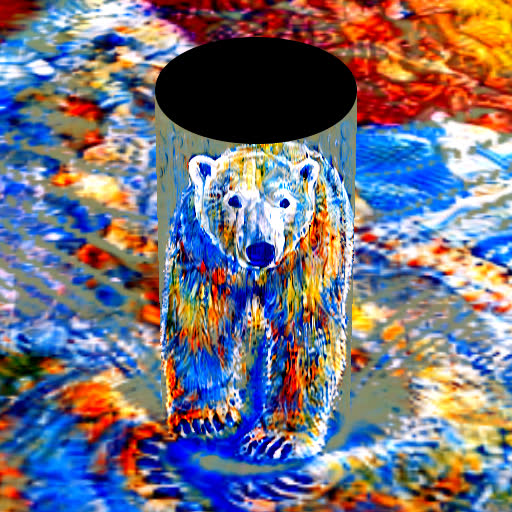}
    \hspace{0.01\linewidth}
    \includegraphics[width=0.28\linewidth, trim=0 0 0 0, clip]{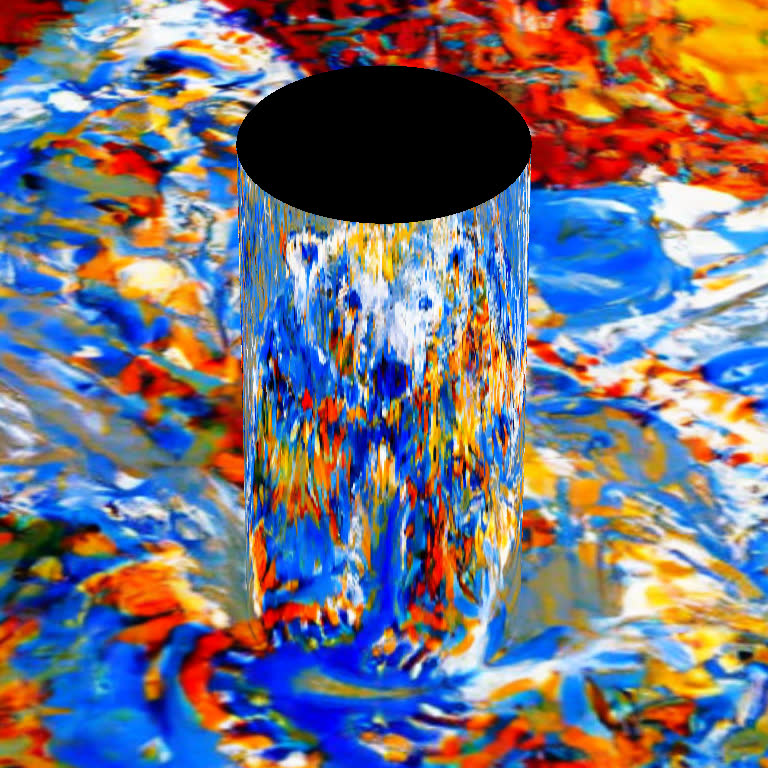}
    \hspace{0.01\linewidth}
    \includegraphics[width=0.28\linewidth, trim=0 0 0 0, clip]{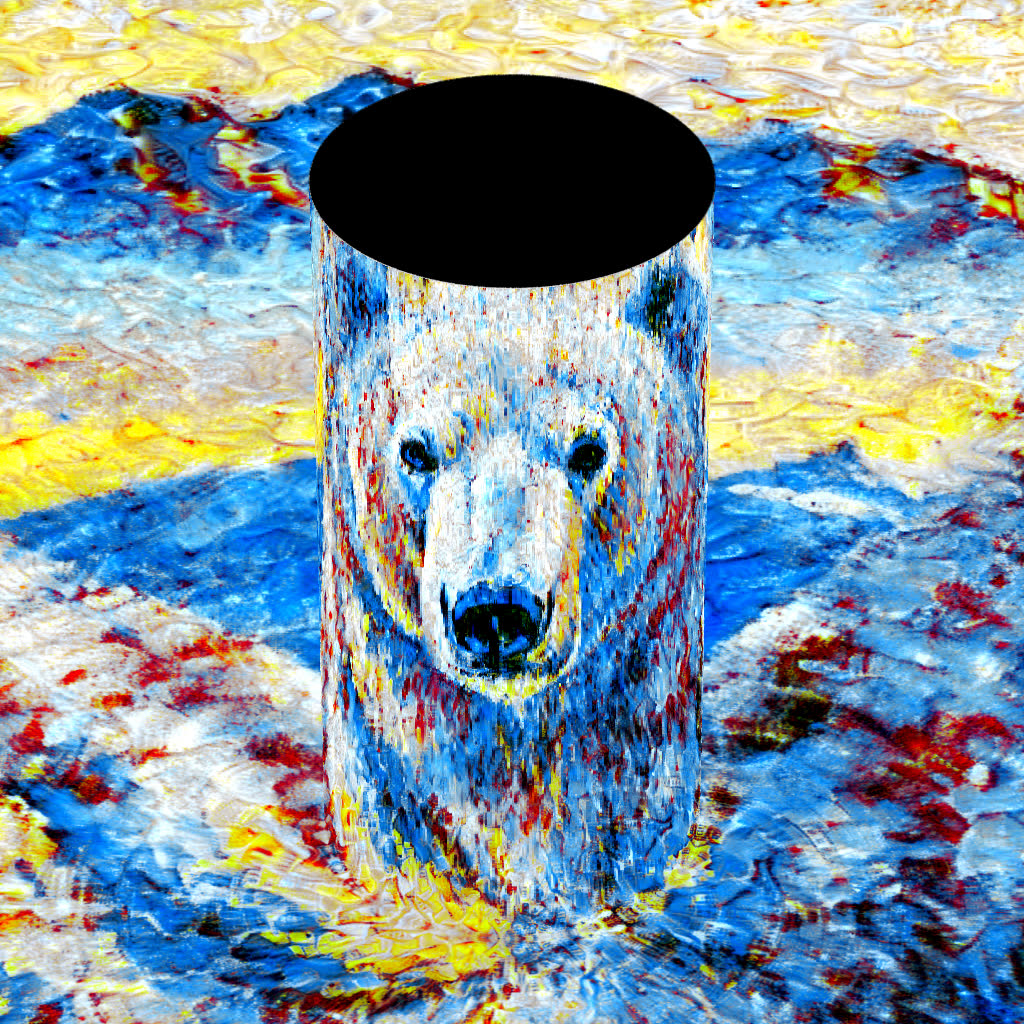}
    \\

    \begin{minipage}[c]{0.28\linewidth} 
        \centering
        \footnotesize{Baseline}
    \end{minipage}
    \hspace{0.01\linewidth}
    \begin{minipage}[c]{0.28\linewidth} 
        \centering
        \footnotesize{Upcaled}
    \end{minipage}
    \hspace{0.01\linewidth}
        \begin{minipage}[c]{0.28\linewidth} 
        \centering
        \footnotesize{Ours}
    \end{minipage}


    \vspace{-2mm}

    \caption{
    \textbf{Ablation: apply super-resolution model on texture map.} 
    Left: 512-resolution illusion generation on single-cylinder reflective case. Middle: rendered image after upcaling the texture map of baseline from  \(512\times512\) to  \(1024\times1024\). Right: our result on \(1024\times1024\) texture map (a new training with same prompt input). Our method can generate high-resolution results and preserve the image content of both views.
    }
    
    \label{fig:SR}
\end{figure}

\topic{Super-resolution model} 
Fig. \ref{fig:SR} demonstrates the limitation of using a super-resolution model (Stable Difision upscaler ~\cite{Rombach_2022_CVPR}) on one single view (2D plane in reflective cylinder setting), as it will degrade the content (polar bear) in the reflective cylinder view. Our result can generate \(2048\times2048\) resolution texture map without using a pre-trained super-resolution model.

\section{Discussion}
\label{sec:discussionsupp}

\topic{Texture map representation}
We tested various texture map representations: plain RGB images, MLP-represented images, and multiresolution texture maps. Plain RGB images failed to produce consistent 2D illusions. MLP representations ~\cite{Burgert2023DiffusionIH} generated 3D illusions but were limited in resolution due to GPU memory constraints. Multiresolution texture maps successfully produced high-quality 3D illusions with higher render resolutions, highlighting the importance of appropriate texture representation.

\topic{Image input}
Given an image input, we use L2 loss in the Waldo case in Fig. \ref{fig:waldo}, supervising one view to a ground-truth image pixel-wise. Because of the resolution mismatch of the low sampling frequency of the texture map, we will have little stripe-like artifacts on the texture map around the Waldo area on the texture map. To alleviate this problem, we query a patch of the pixel around that area where the reflected ray from the cylinder hits the plane. It can reduce the artifact, but it cannot fully solve it. This makes sense because, technically, we are always under-sampling the multi-resolution texture map.

\begin{figure*}[htbp]
    \centering
    \vspace{-16mm}
    \setlength{\tabcolsep}{1pt} 
    \renewcommand{\arraystretch}{1}

    \begin{tabular}{cccccc}
        \vspace{-1mm}
        \begin{minipage}[t]{0.16\textwidth}
            \includegraphics[width=\textwidth, trim=140 140 140 140, clip]{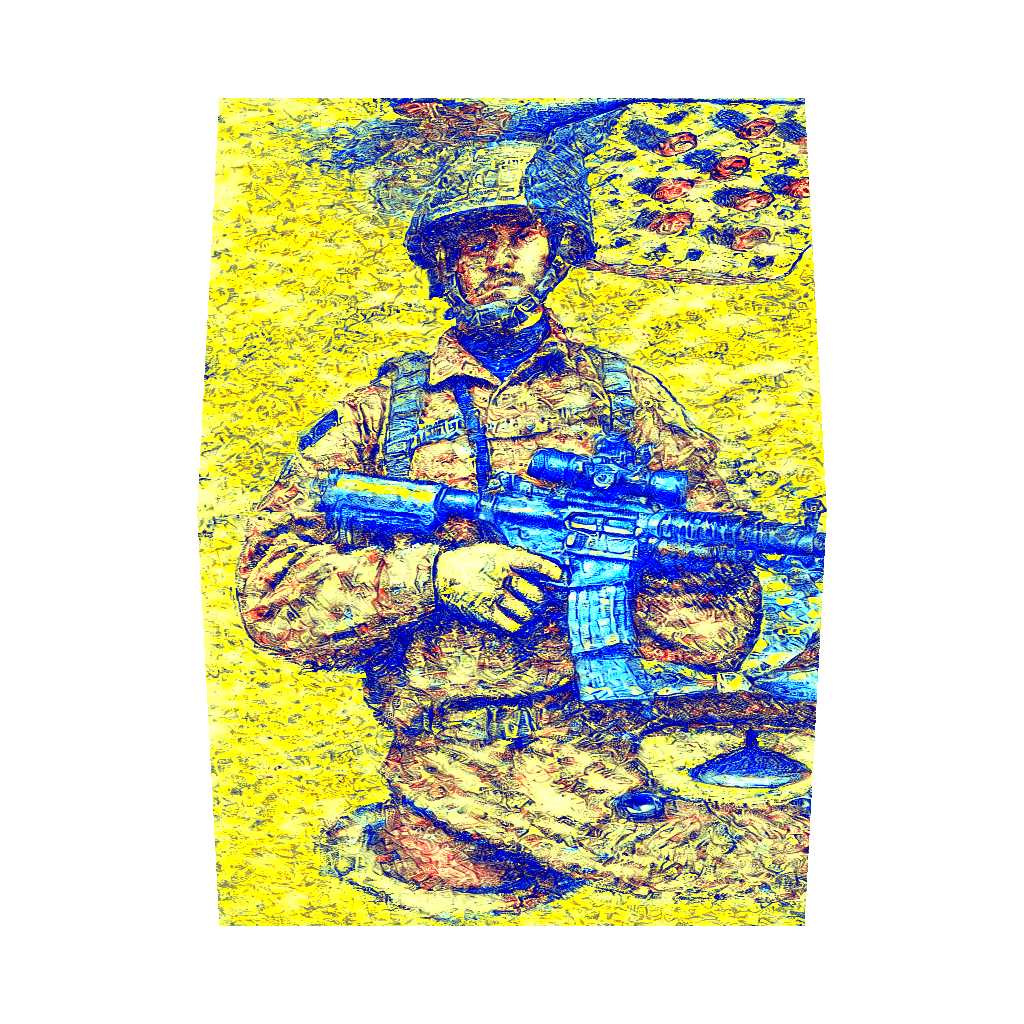}
        \end{minipage}%
        \begin{minipage}[t]{0.16\textwidth}
            \includegraphics[width=\textwidth, trim=140 140 140 140, clip]{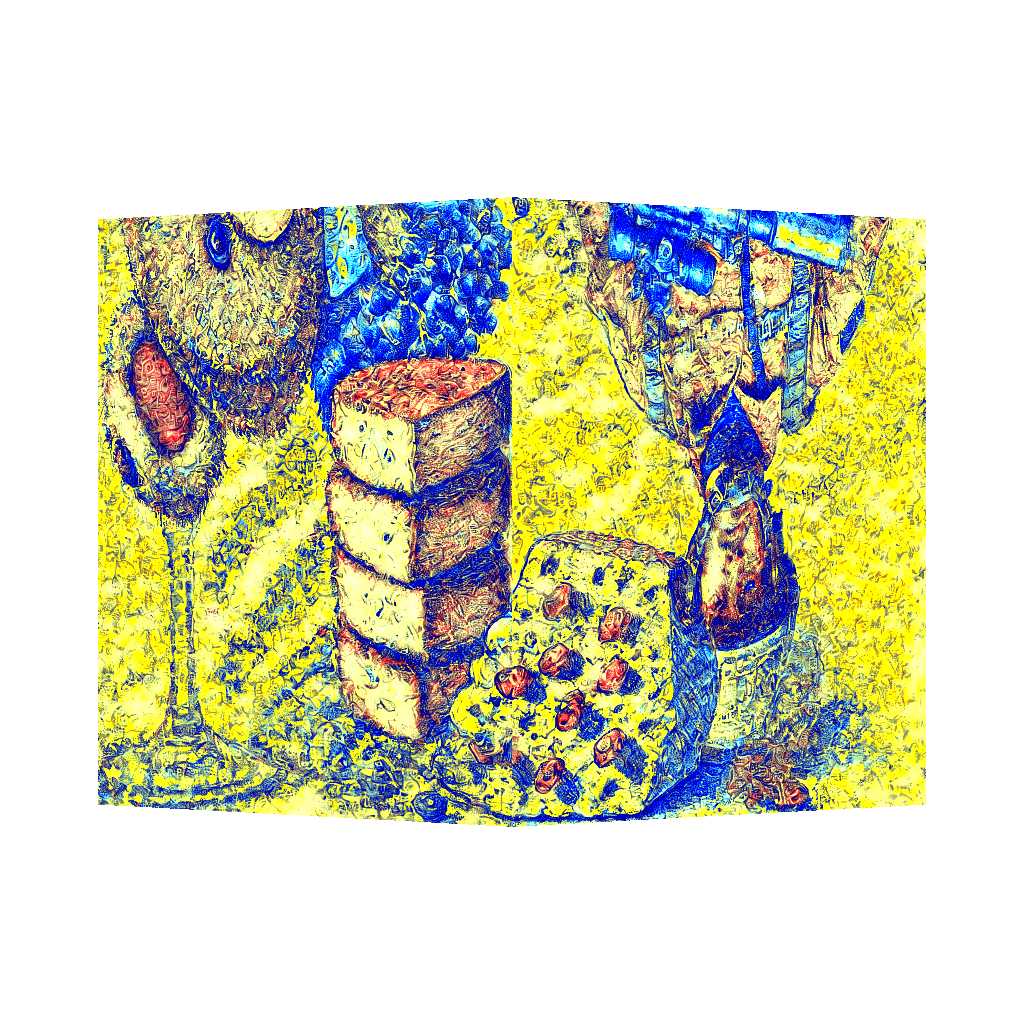}
        \end{minipage}%
        \begin{minipage}[t]{0.16\textwidth}
            \includegraphics[width=\textwidth, trim=140 140 140 140, clip]{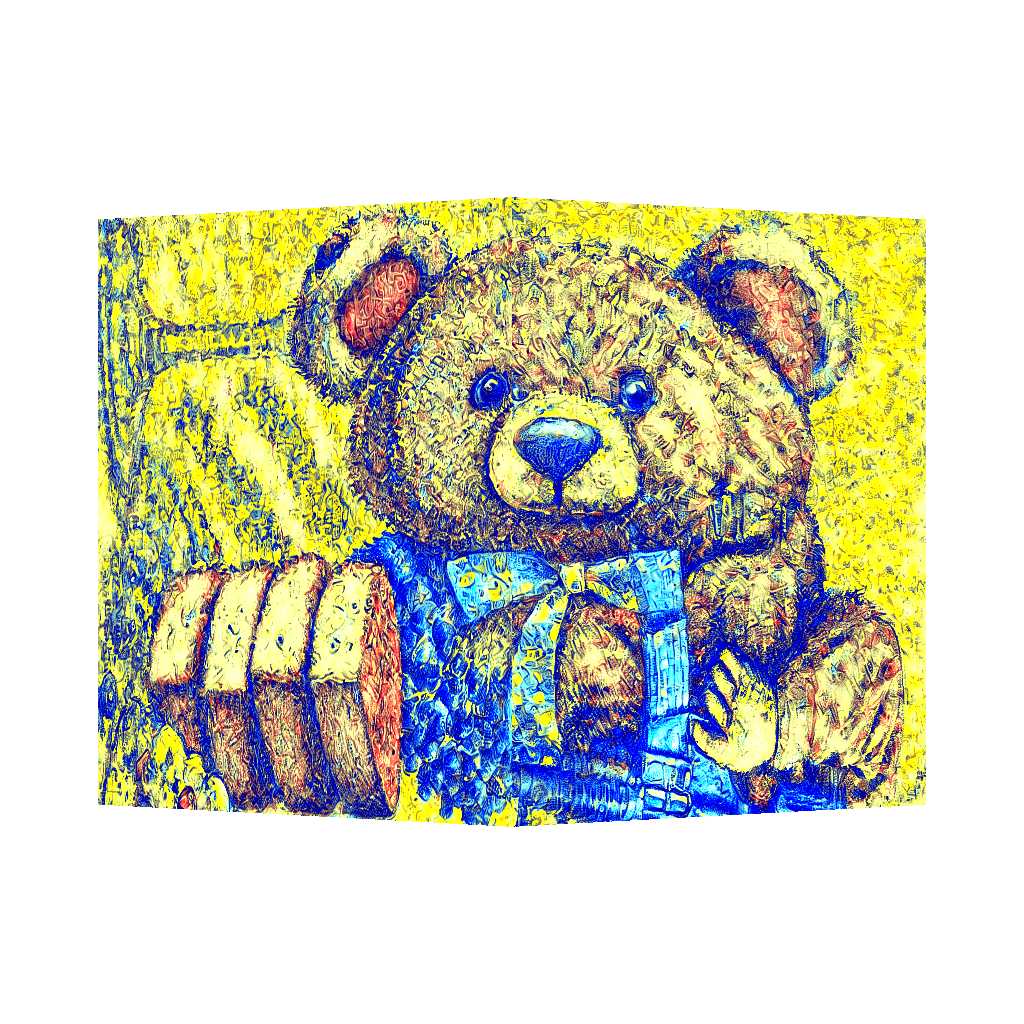}
        \end{minipage}%
        \begin{minipage}[t]{0.16\textwidth}
            \includegraphics[width=\textwidth, trim=140 140 140 140, clip]{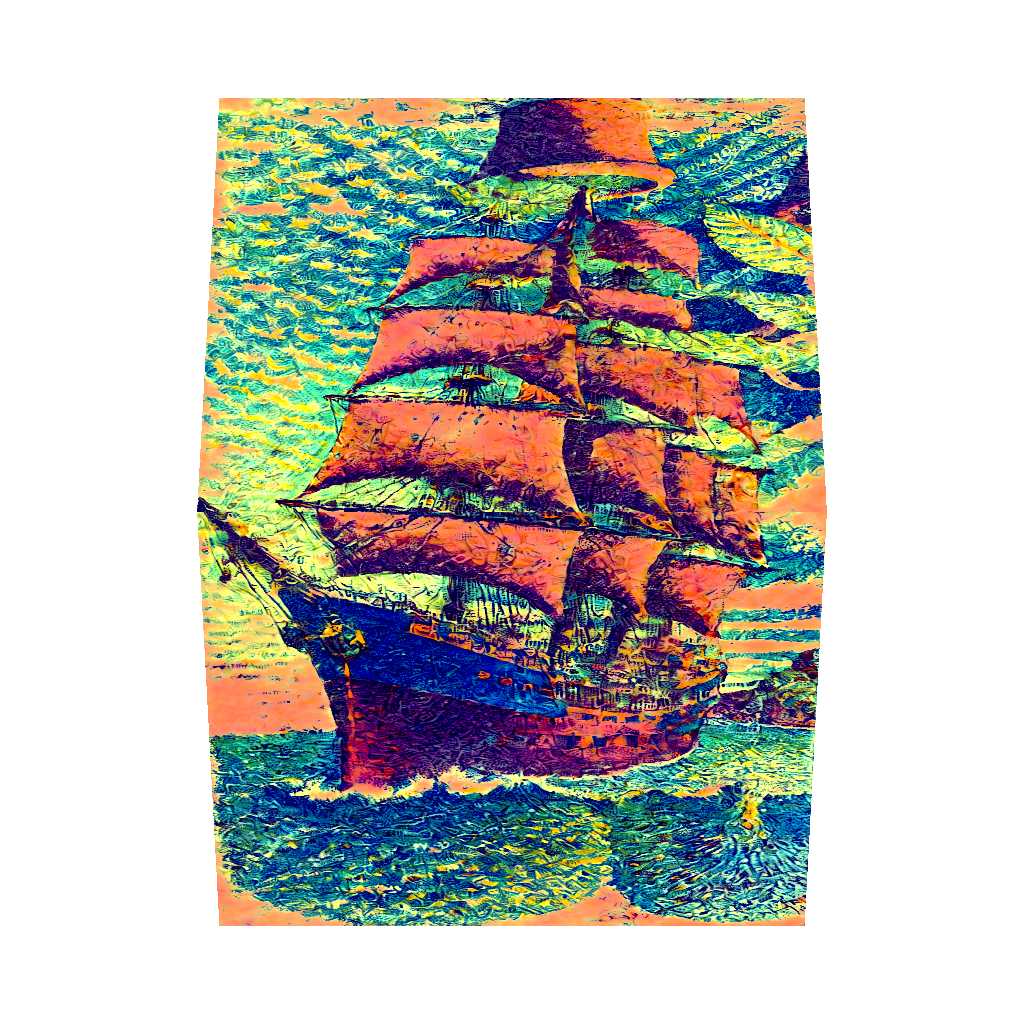}
        \end{minipage}%
        \begin{minipage}[t]{0.16\textwidth}
            \includegraphics[width=\textwidth, trim=140 140 140 140, clip]{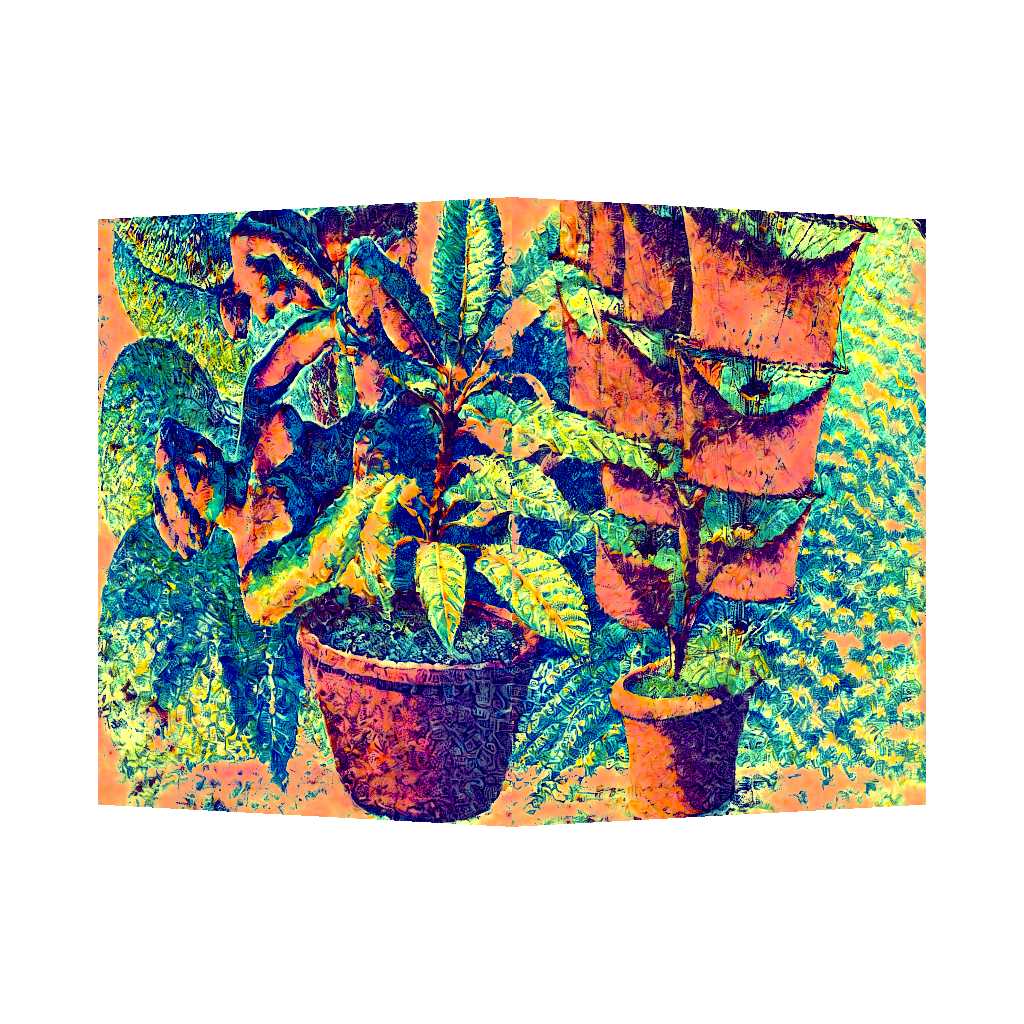}
        \end{minipage}%
        \begin{minipage}[t]{0.16\textwidth}
            \includegraphics[width=\textwidth, trim=140 140 140 140, clip]{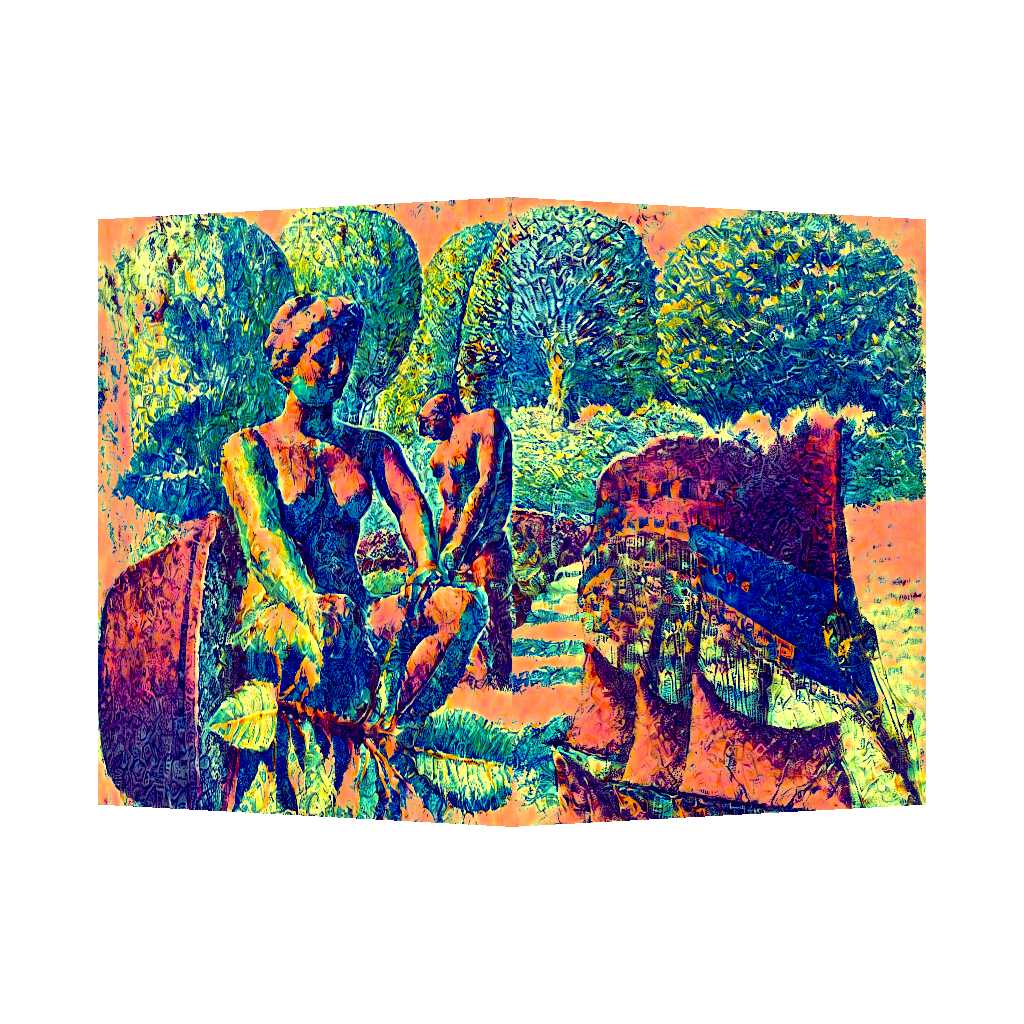}
        \end{minipage}\\
        \vspace{-1mm}
        \begin{minipage}[t]{0.16\textwidth}
        \centering
            \includegraphics[width=\linewidth, trim=185 20 185 30, clip]{figures/prompt/soldier.jpg}
            \end{minipage}\hfill
        \begin{minipage}[t]{0.16\textwidth}
            \centering
            \includegraphics[width=\linewidth, trim=185 20 185 30, clip]{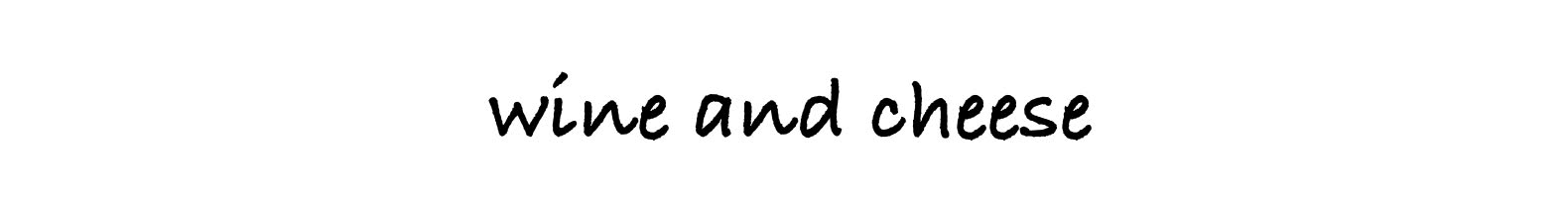}
            \end{minipage}\hfill
        \begin{minipage}[t]{0.16\textwidth}
            \centering
            \includegraphics[width=\linewidth, trim=185 15 185 30, clip]{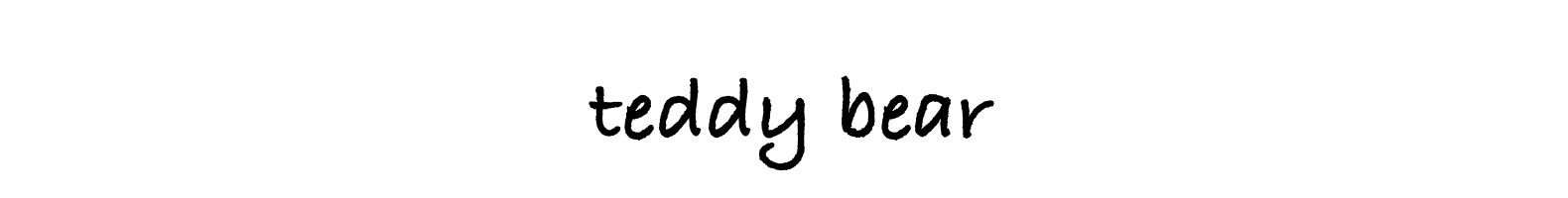}
        \end{minipage}\hfill
        \begin{minipage}[t]{0.16\textwidth}
            \centering
            \includegraphics[width=\linewidth, trim=185 20 185 30, clip]{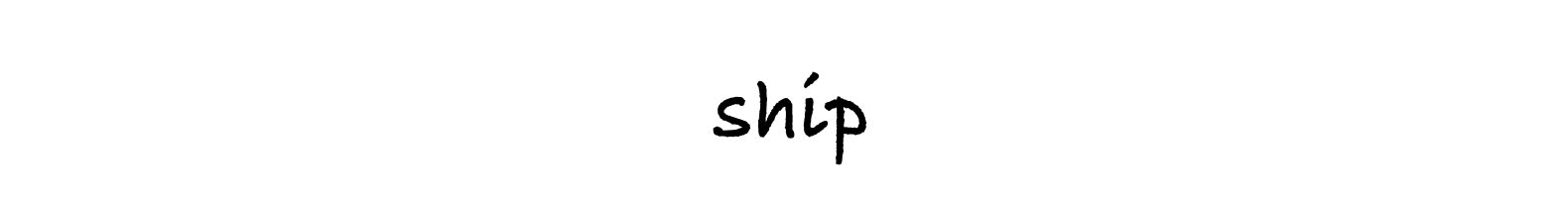}
        \end{minipage}
        \begin{minipage}[t]{0.16\textwidth}
            \centering
            \includegraphics[width=\linewidth, trim=185 20 185 30, clip]{figures/prompt/houseplant.jpg}
        \end{minipage}
        \begin{minipage}[t]{0.16\textwidth}
            \centering
            \includegraphics[width=\linewidth, trim=185 20 185 30, clip]{figures/prompt/sculpturegarden.jpg}
        \end{minipage}\\

        \vspace{-1mm}
        \begin{minipage}[t]{0.16\textwidth}
            \includegraphics[width=\textwidth, trim=140 140 140 140, clip]{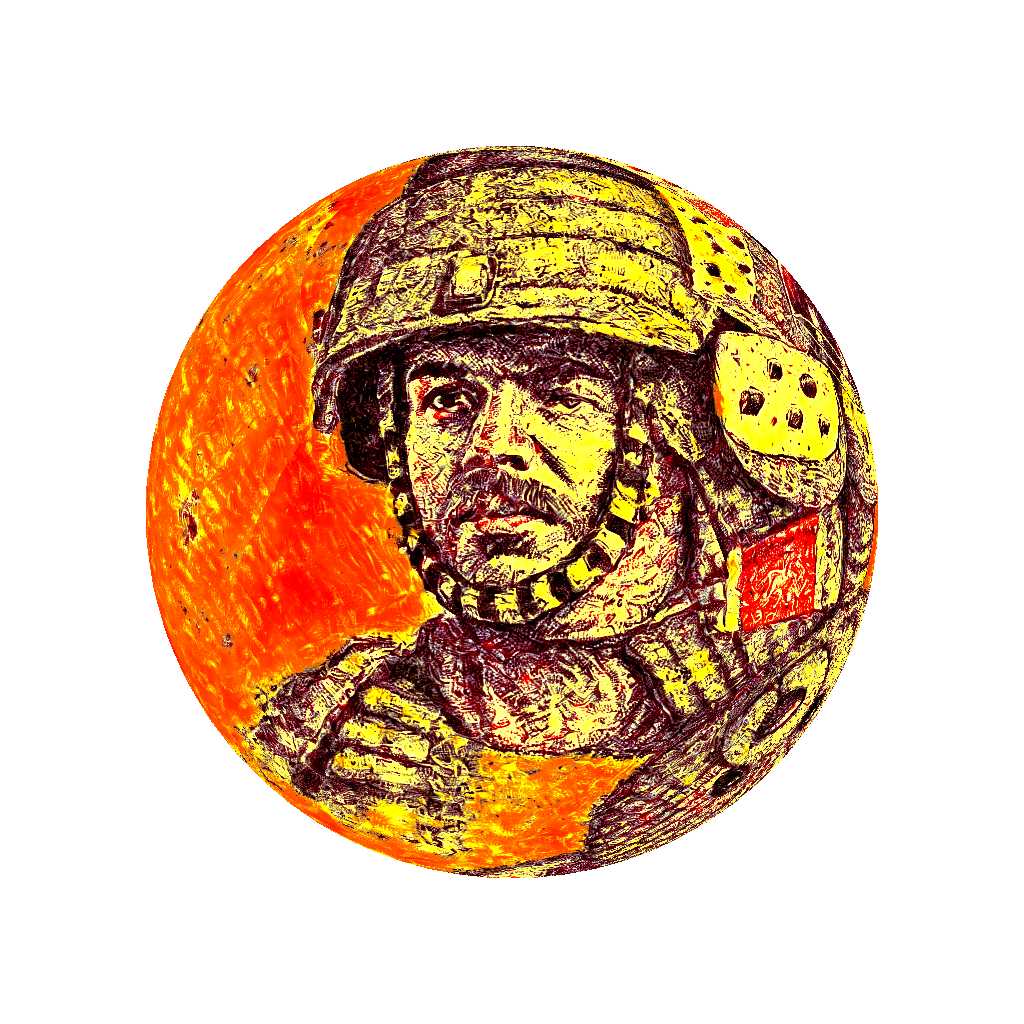}
        \end{minipage}%
        \begin{minipage}[t]{0.16\textwidth}
            \includegraphics[width=\textwidth, trim=140 140 140 140, clip]{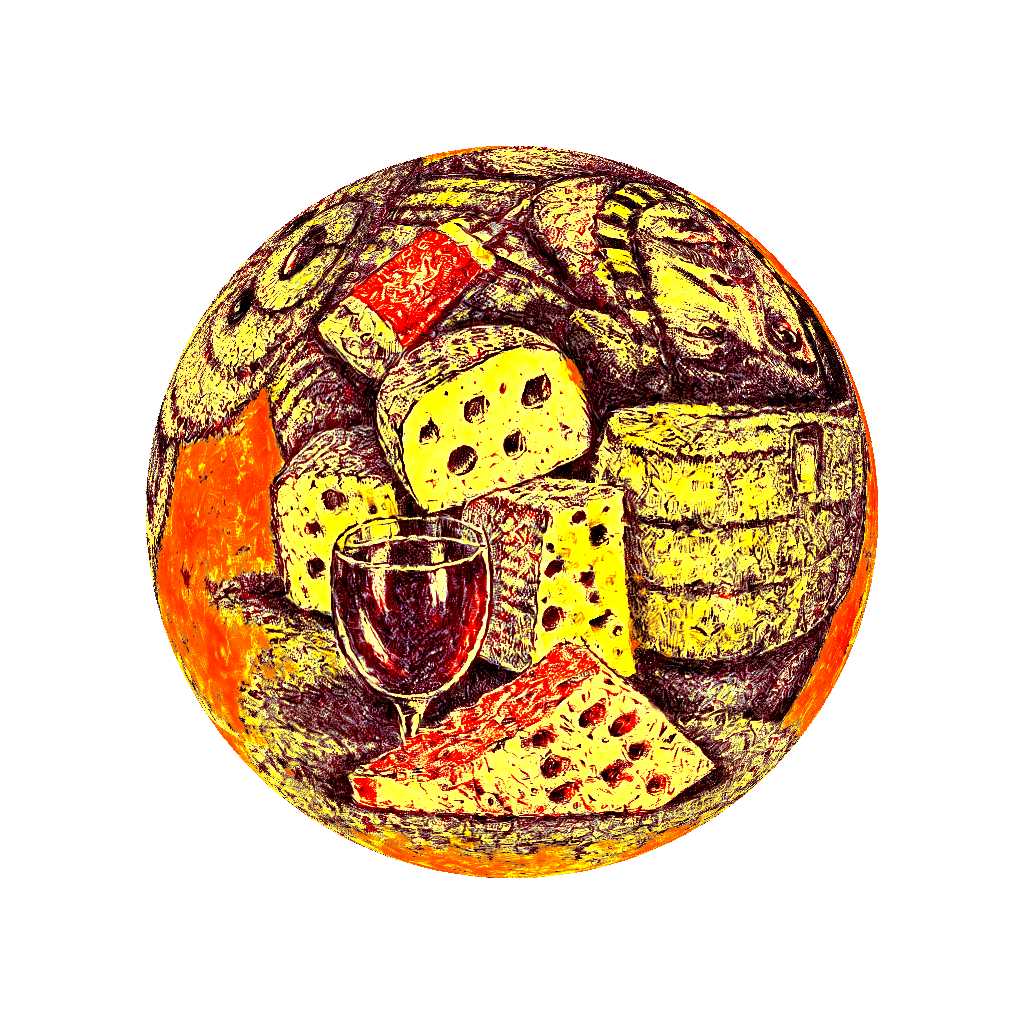}
        \end{minipage}%
        \begin{minipage}[t]{0.16\textwidth}
            \includegraphics[width=\textwidth, trim=140 140 140 140, clip]{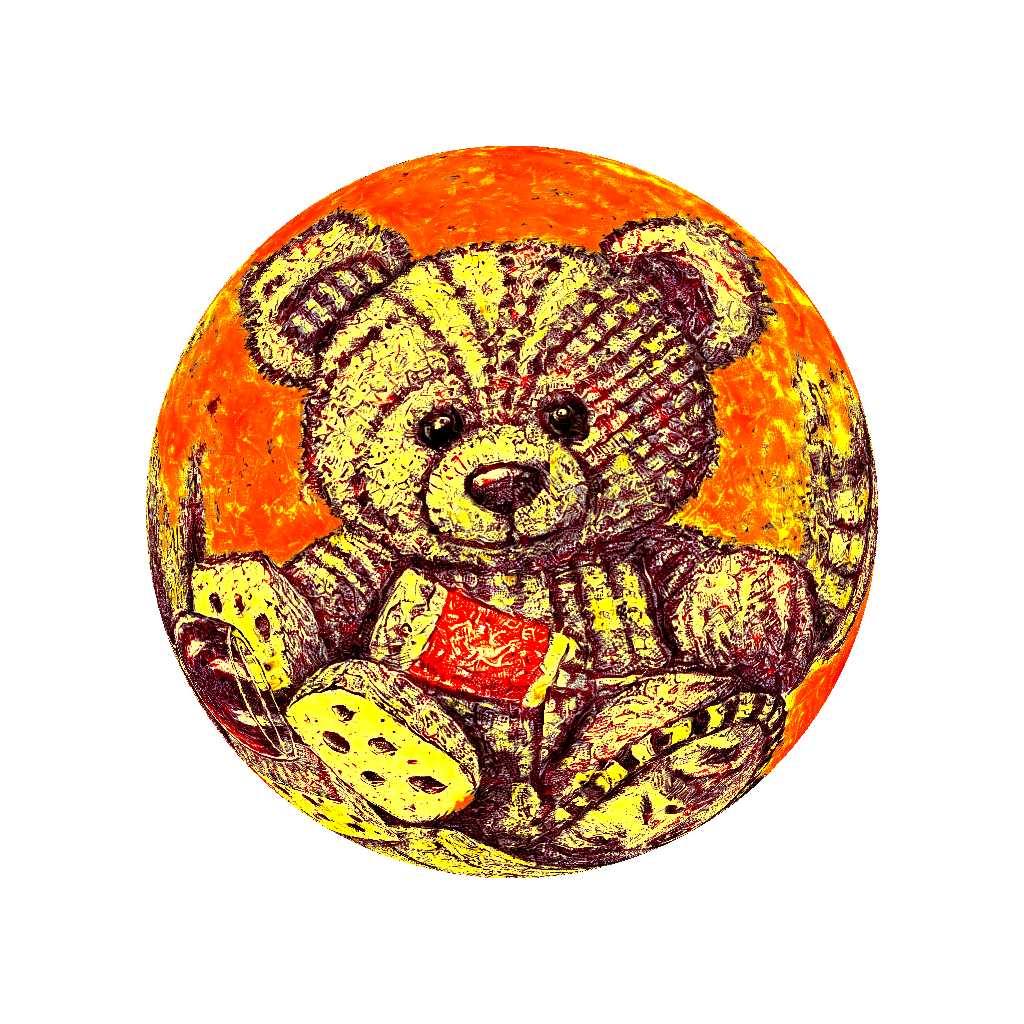}
        \end{minipage}%
        \begin{minipage}[t]{0.16\textwidth}
            \includegraphics[width=\textwidth, trim=140 140 140 140, clip]{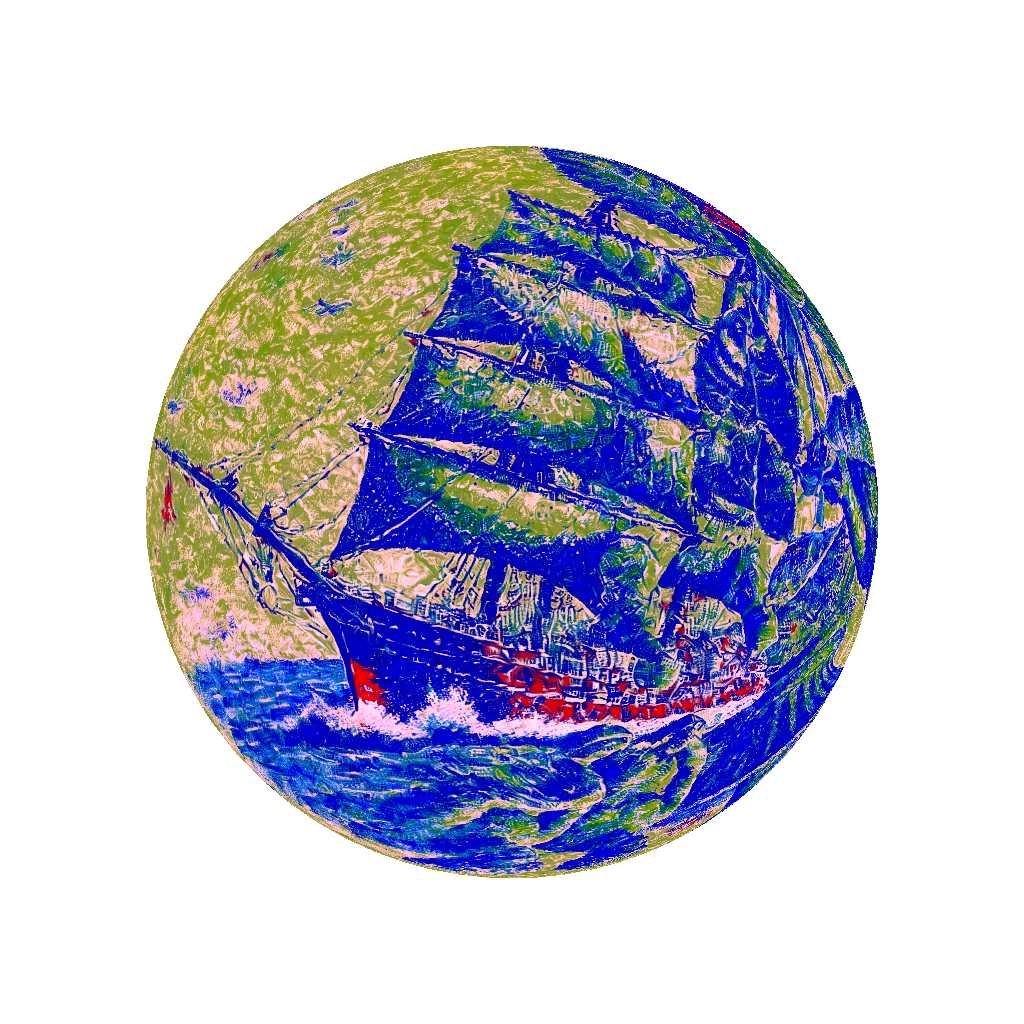}
        \end{minipage}%
        \begin{minipage}[t]{0.16\textwidth}
            \includegraphics[width=\textwidth, trim=140 140 140 140, clip]{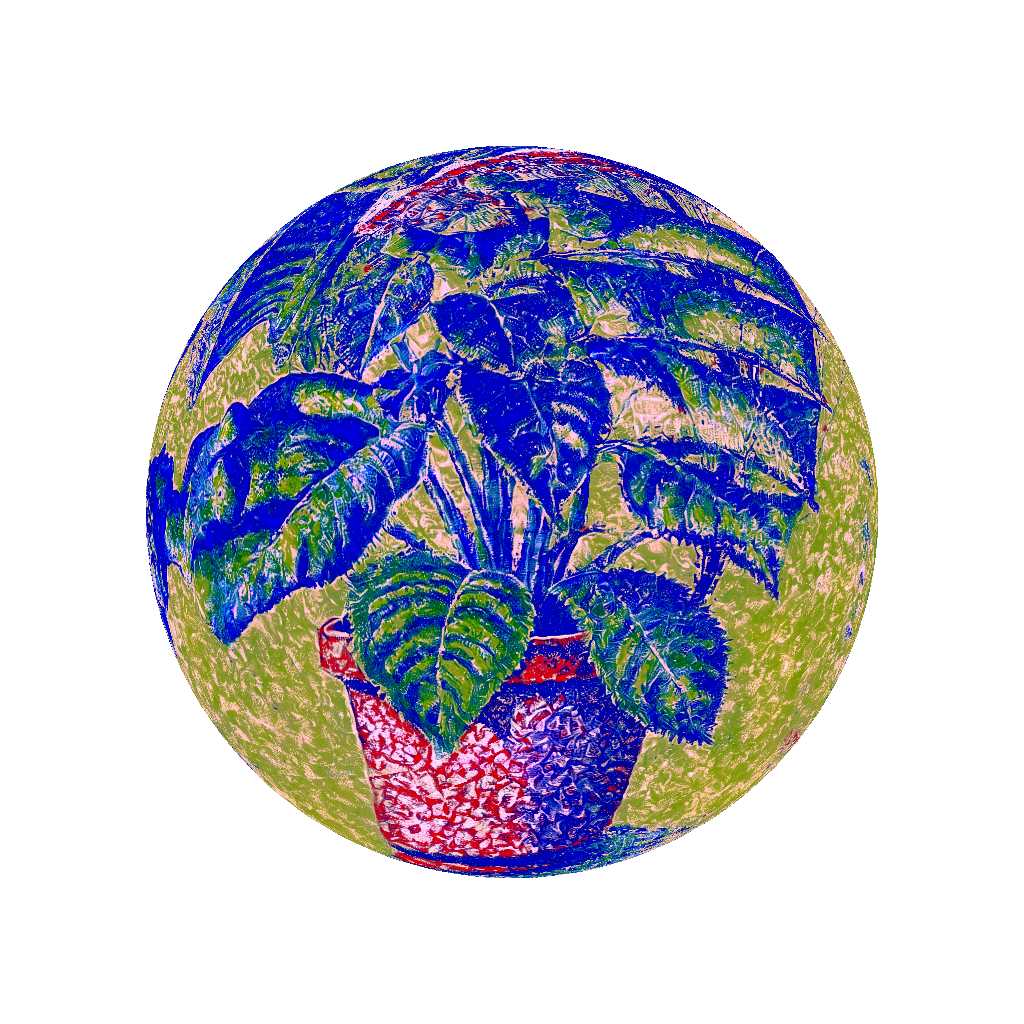}
        \end{minipage}%
        \begin{minipage}[t]{0.16\textwidth}
            \includegraphics[width=\textwidth, trim=140 140 140 140, clip]{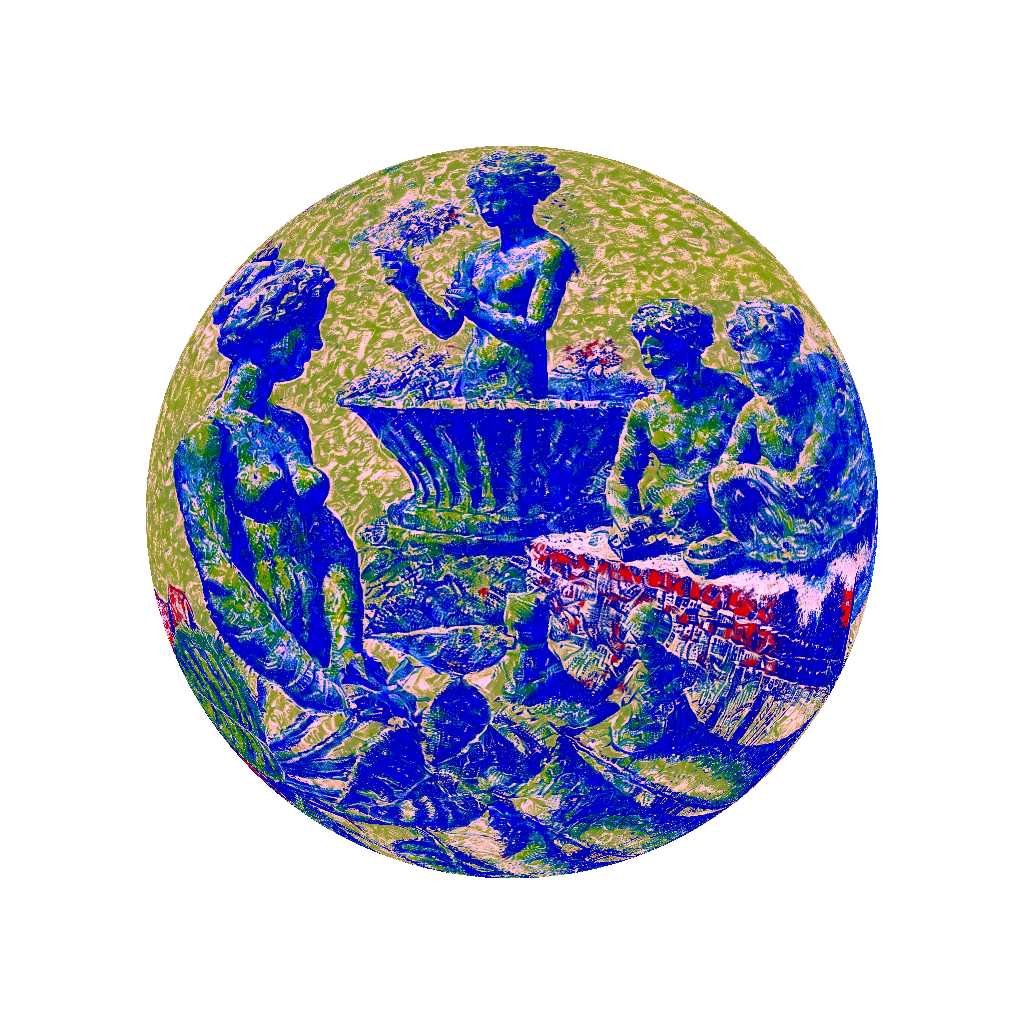}
        \end{minipage}\\

        \vspace{-1mm}
        \begin{minipage}[t]{0.16\textwidth}
            \includegraphics[width=\textwidth, trim=140 140 140 140, clip]{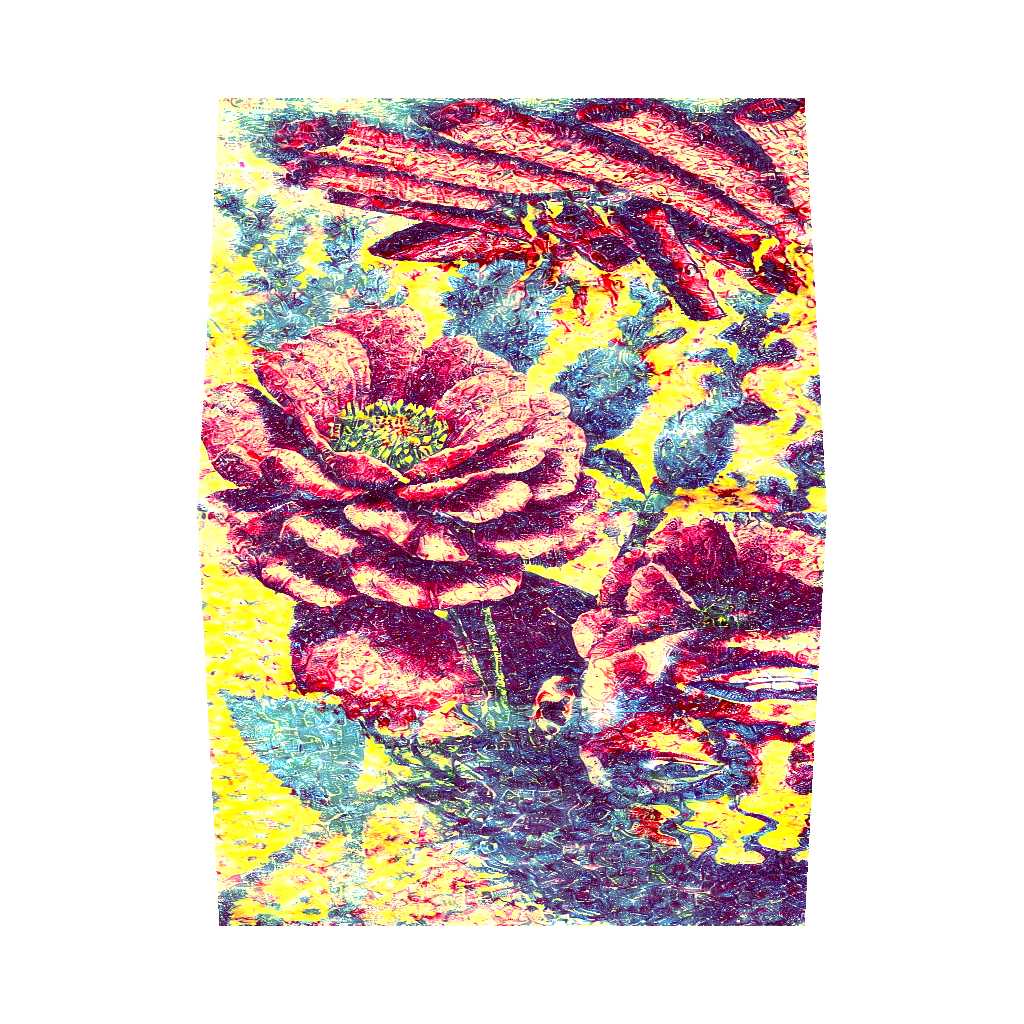}
        \end{minipage}%
        \begin{minipage}[t]{0.16\textwidth}
            \includegraphics[width=\textwidth, trim=140 140 140 140, clip]{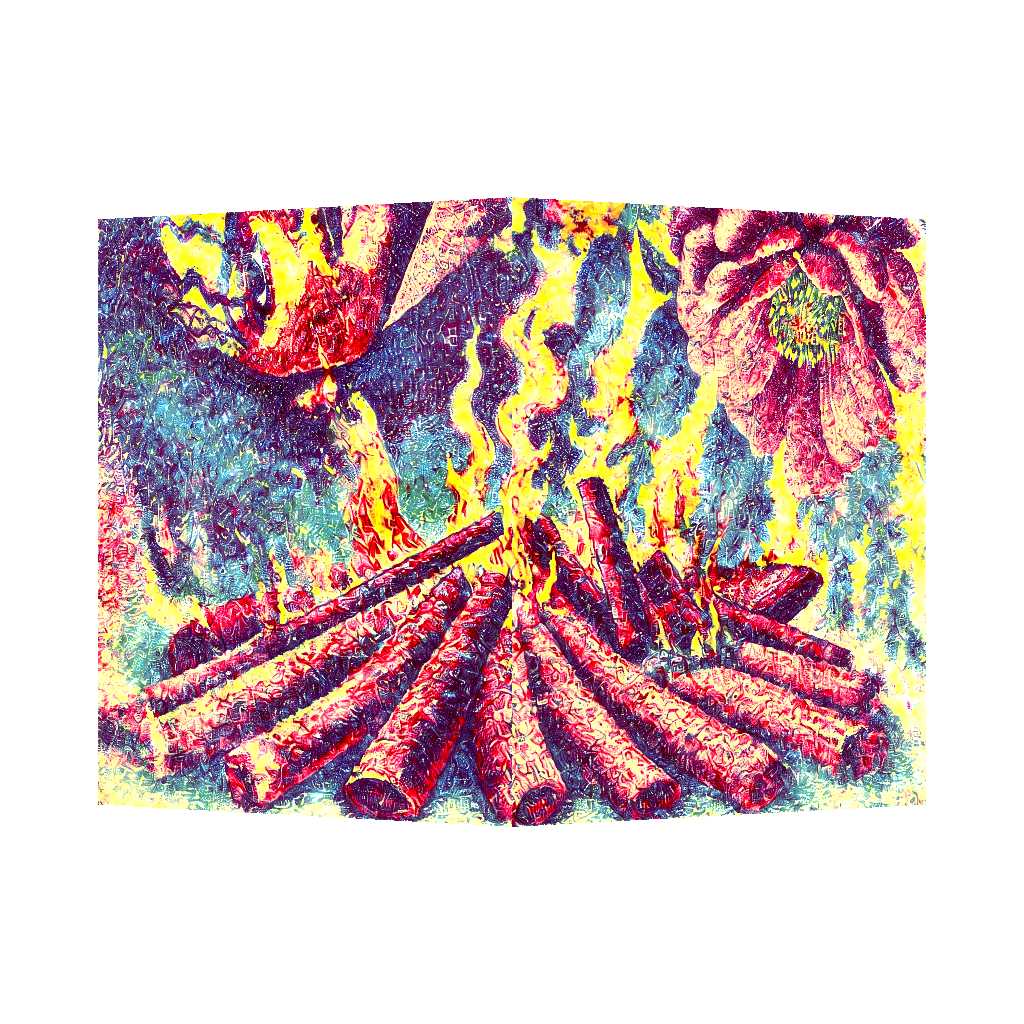}
        \end{minipage}%
        \begin{minipage}[t]{0.16\textwidth}
            \includegraphics[width=\textwidth, trim=140 140 140 140, clip]{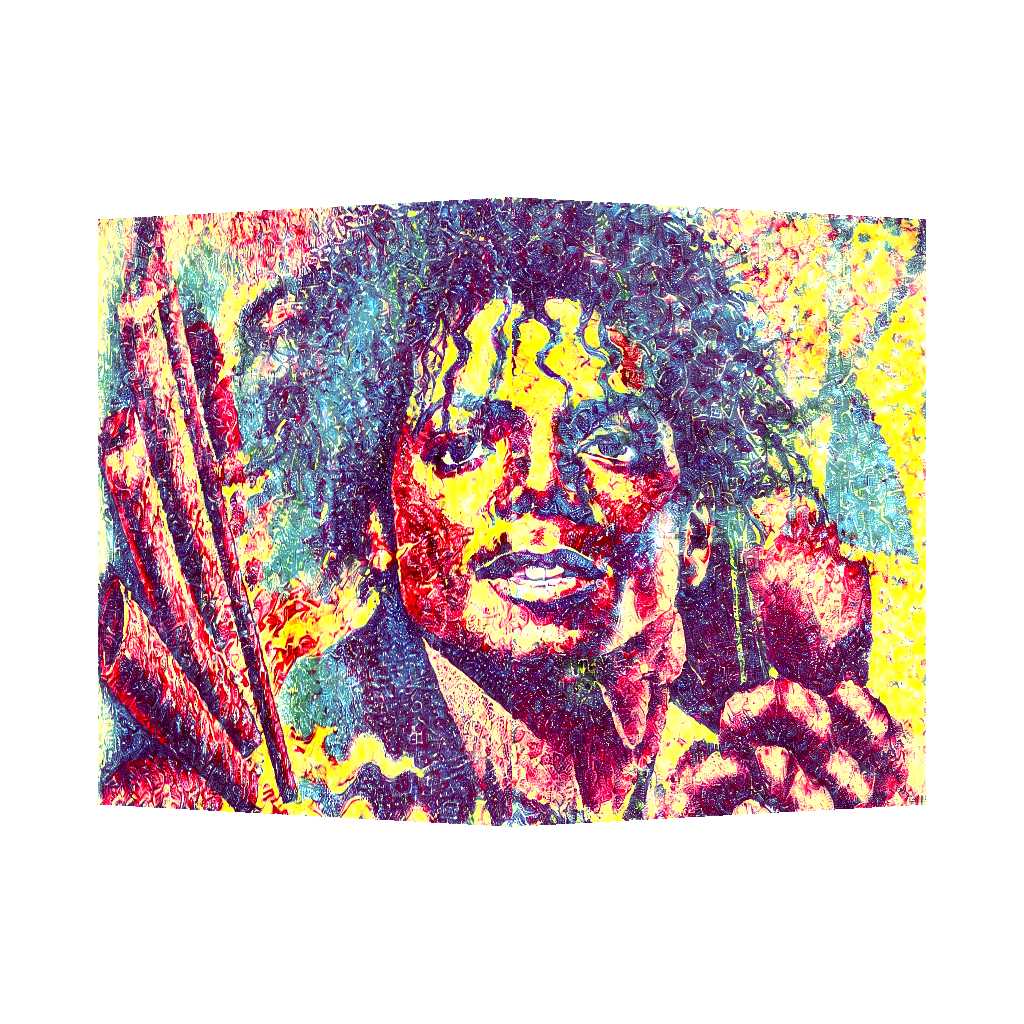}
        \end{minipage}%
        \begin{minipage}[t]{0.16\textwidth}
            \includegraphics[width=\textwidth, trim=140 140 140 140, clip]{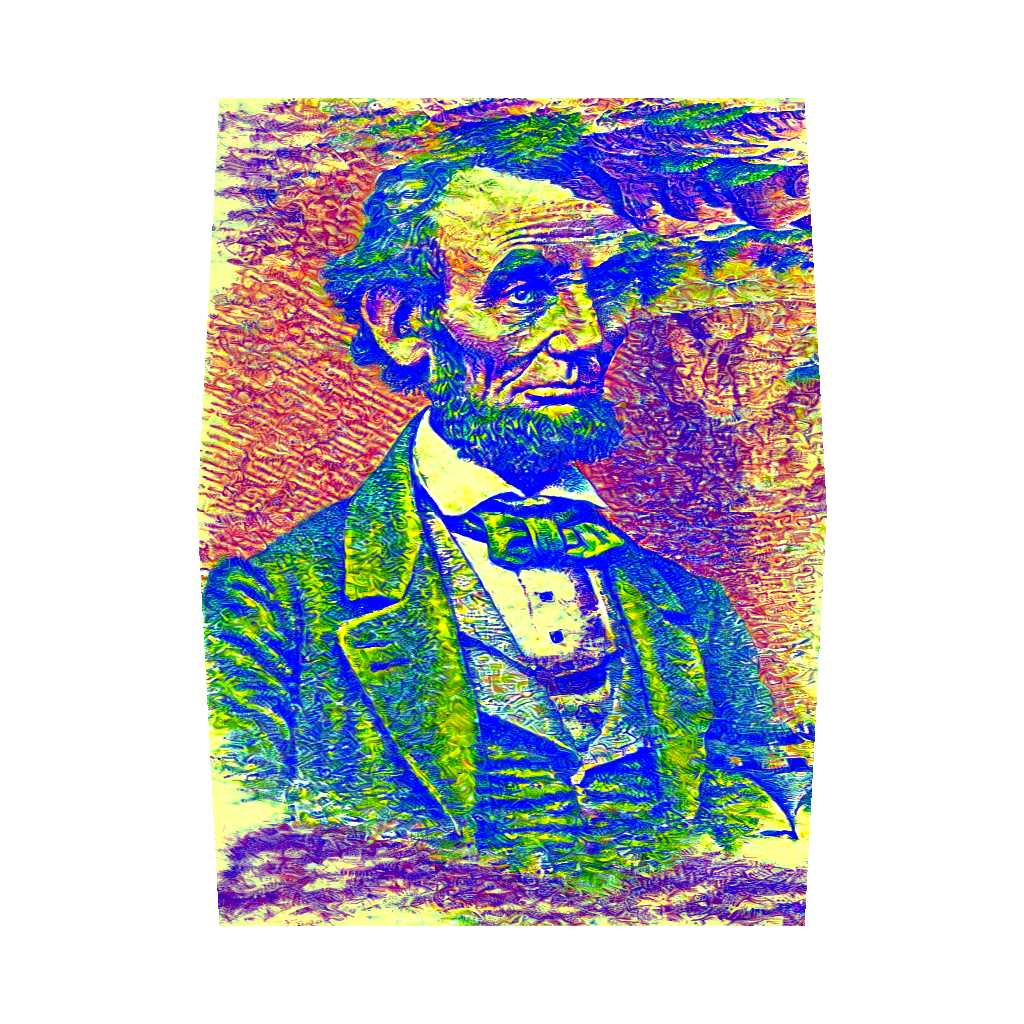}
        \end{minipage}%
        \begin{minipage}[t]{0.16\textwidth}
            \includegraphics[width=\textwidth, trim=140 140 140 140, clip]{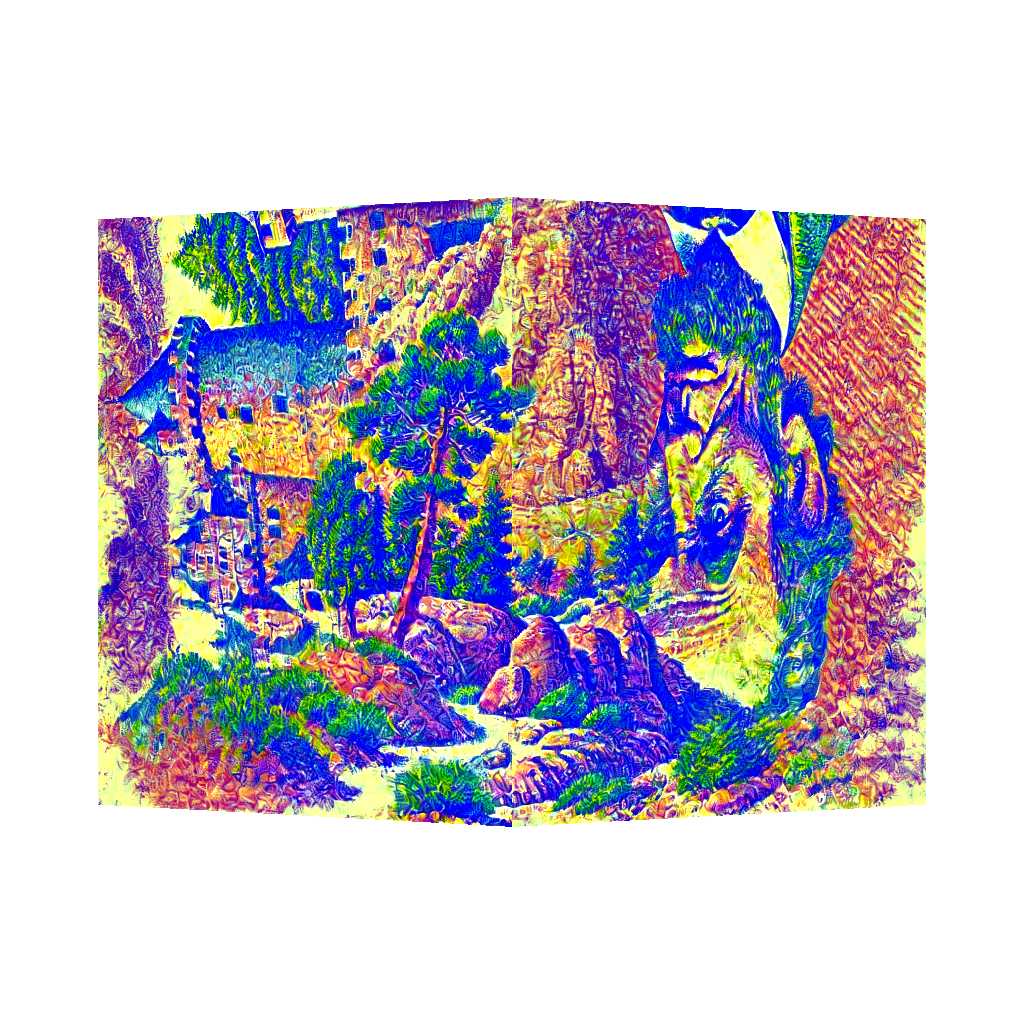}
        \end{minipage}%
        \begin{minipage}[t]{0.16\textwidth}
            \includegraphics[width=\textwidth, trim=140 140 140 140, clip]{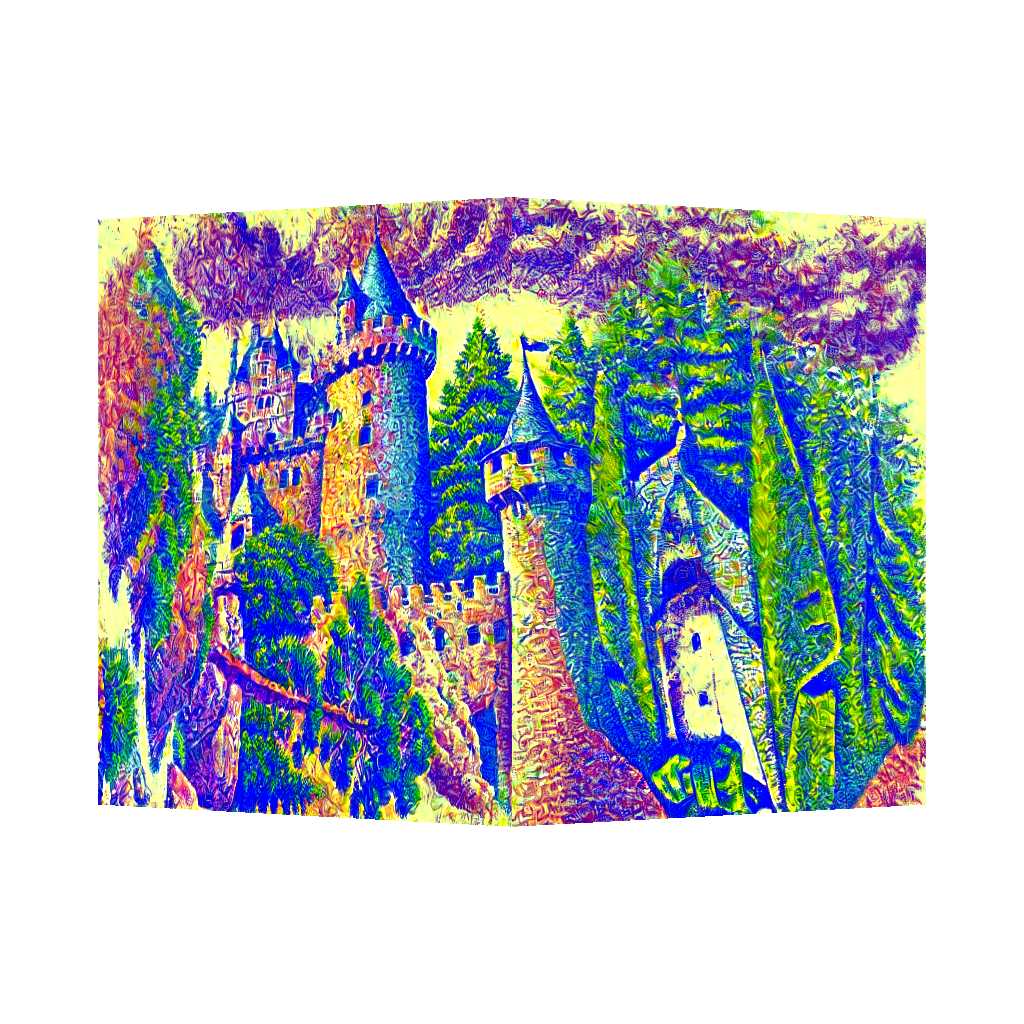}
        \end{minipage}\\

        \vspace{-1mm}
        \begin{minipage}[t]{0.16\textwidth}
        \centering
            \includegraphics[width=\linewidth, trim=185 20 185 30, clip]{figures/prompt/flower.jpg}
            \end{minipage}\hfill
        \begin{minipage}[t]{0.16\textwidth}
            \centering
            \includegraphics[width=\linewidth, trim=185 20 185 30, clip]{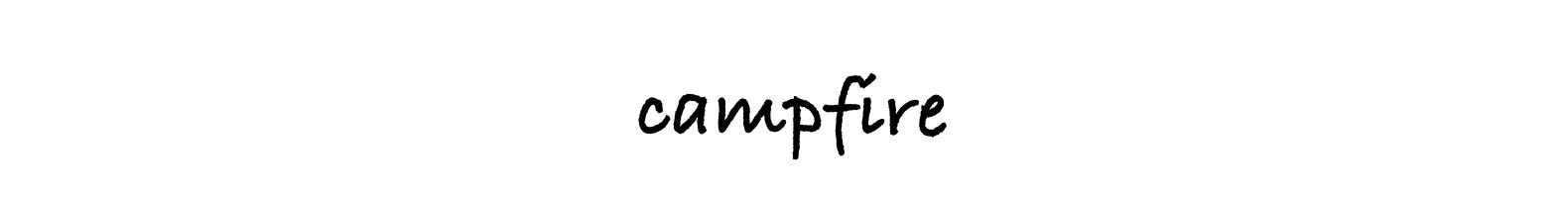}
            \end{minipage}\hfill
        \begin{minipage}[t]{0.16\textwidth}
            \centering
            \includegraphics[width=\linewidth, trim=185 20 185 30, clip]{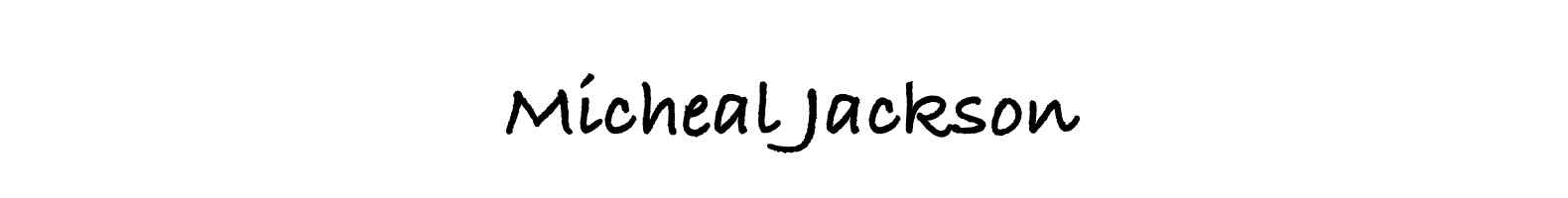}
        \end{minipage}\hfill
        \begin{minipage}[t]{0.16\textwidth}
            \centering
            \includegraphics[width=\linewidth, trim=185 20 185 30, clip]{figures/prompt/al.jpg}
        \end{minipage}
        \begin{minipage}[t]{0.16\textwidth}
            \centering
            \includegraphics[width=\linewidth, trim=185 20 185 30, clip]{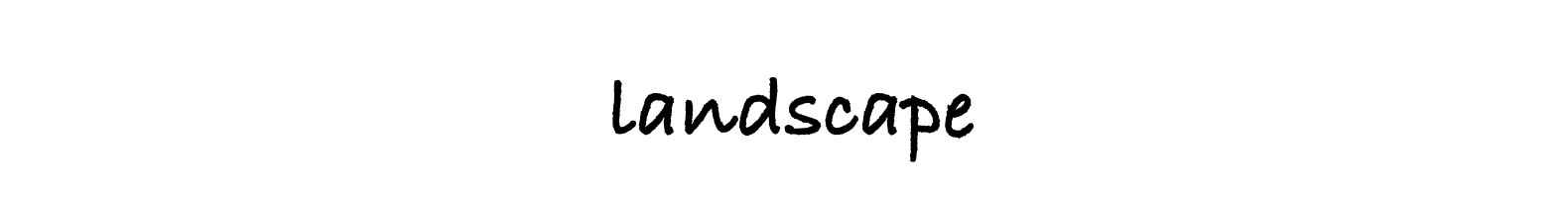}
        \end{minipage}
        \begin{minipage}[t]{0.16\textwidth}
            \centering
            \includegraphics[width=\linewidth, trim=185 20 185 30, clip]{figures/prompt/castle.jpg}
        \end{minipage}\\

        \vspace{-1mm}
        \begin{minipage}[t]{0.16\textwidth}
            \includegraphics[width=\textwidth, trim=140 140 140 140, clip]{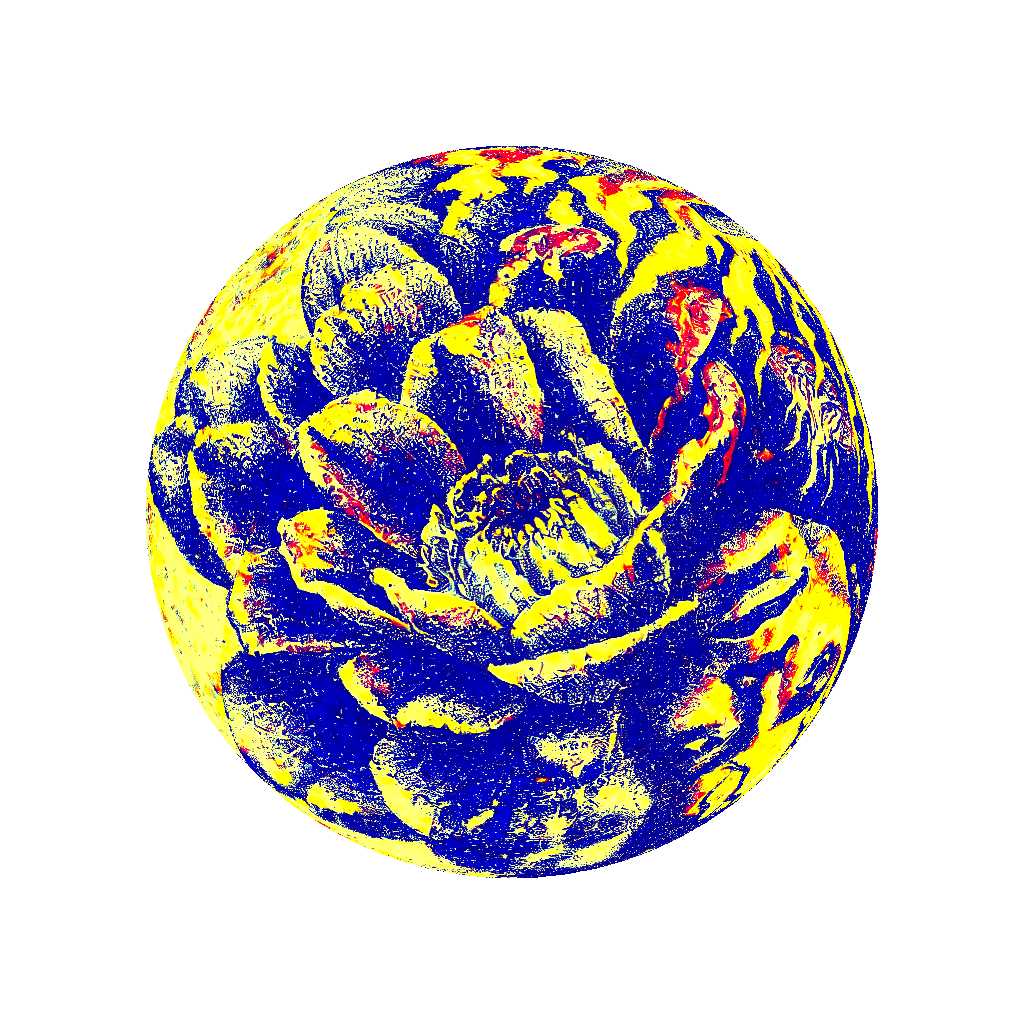}
        \end{minipage}%
        \begin{minipage}[t]{0.16\textwidth}
            \includegraphics[width=\textwidth, trim=140 140 140 140, clip]{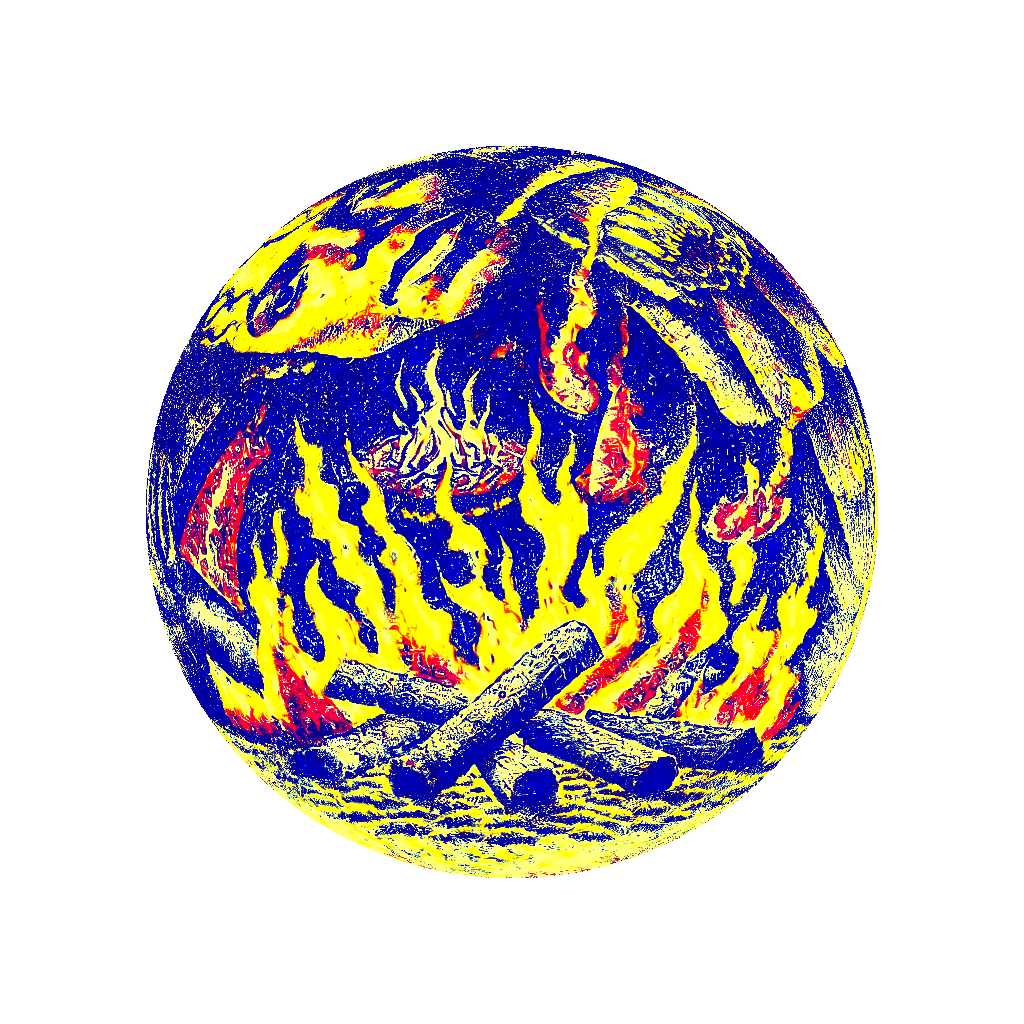}
        \end{minipage}%
        \begin{minipage}[t]{0.16\textwidth}
            \includegraphics[width=\textwidth, trim=140 140 140 140, clip]{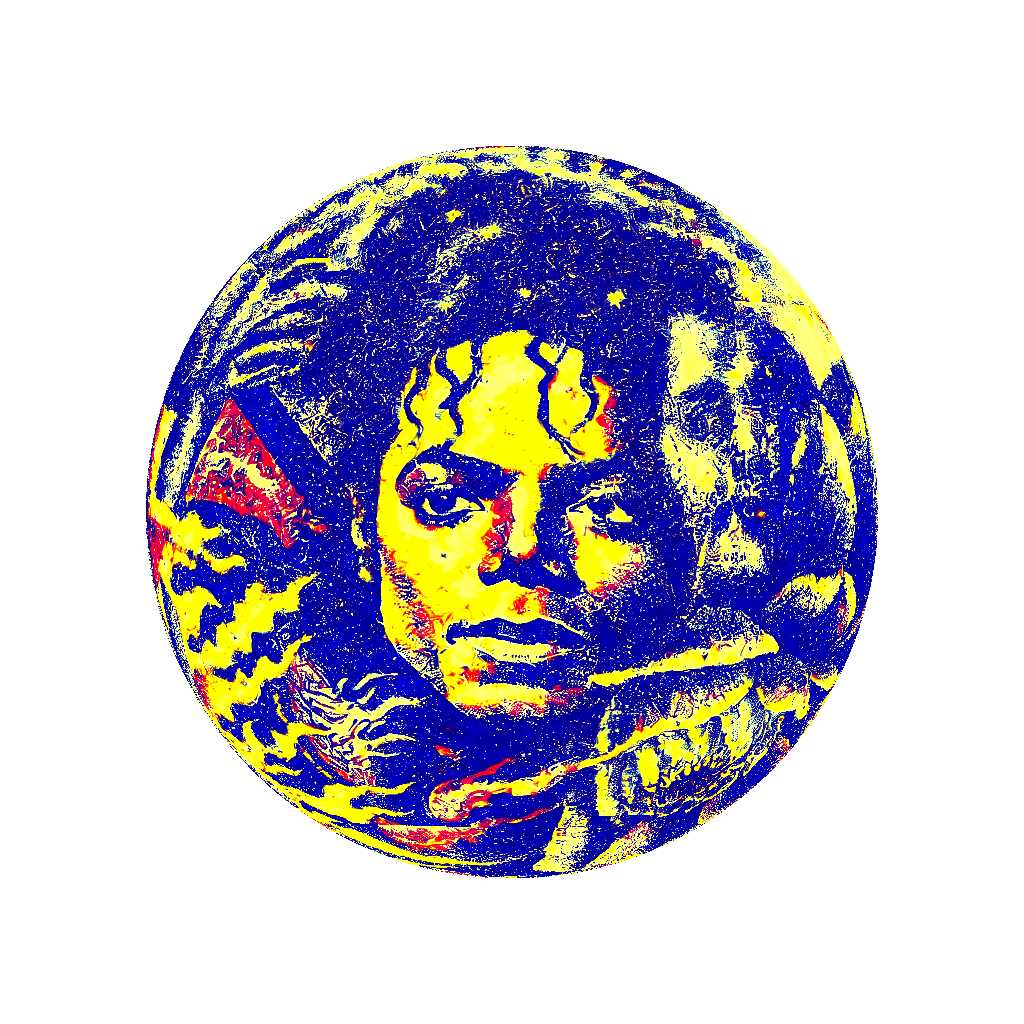}
        \end{minipage}%
        \begin{minipage}[t]{0.16\textwidth}
            \includegraphics[width=\textwidth, trim=140 140 140 140, clip]{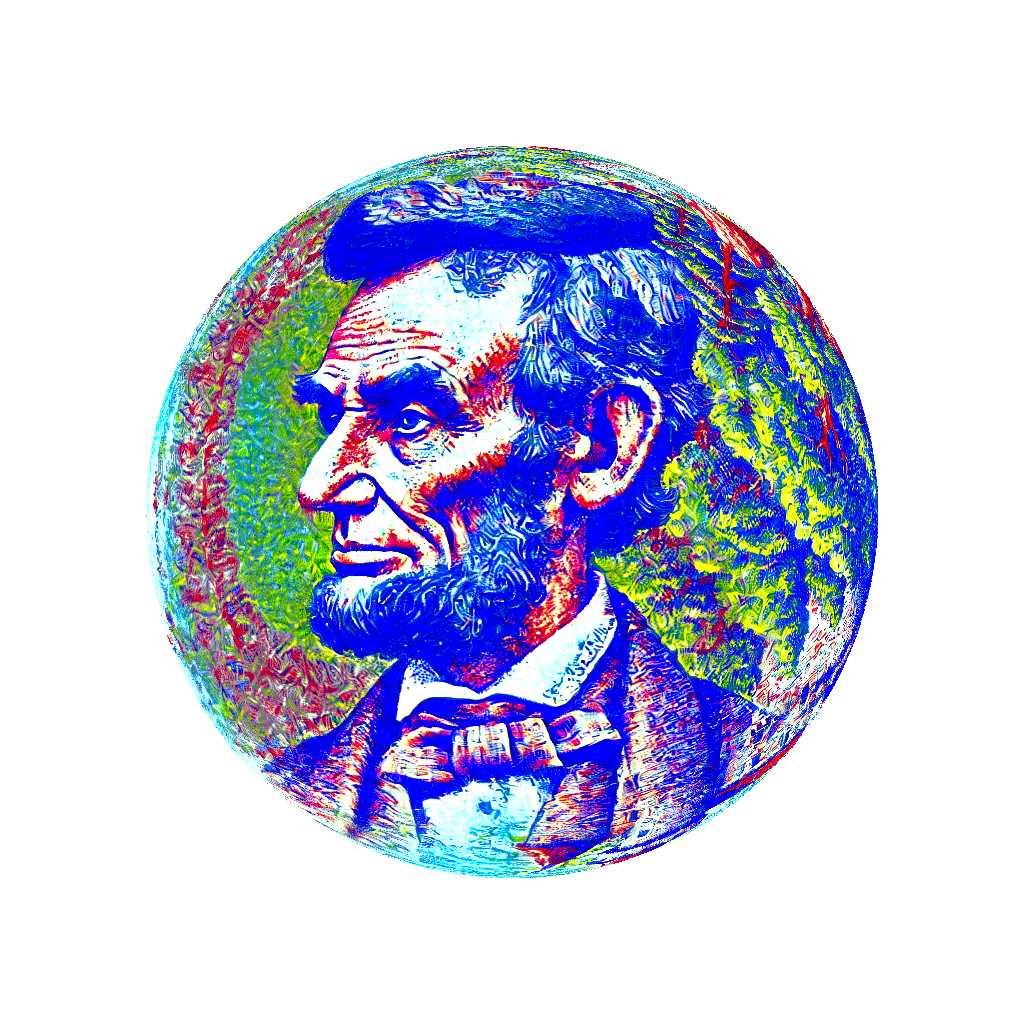}
        \end{minipage}%
        \begin{minipage}[t]{0.16\textwidth}
            \includegraphics[width=\textwidth, trim=140 140 140 140, clip]{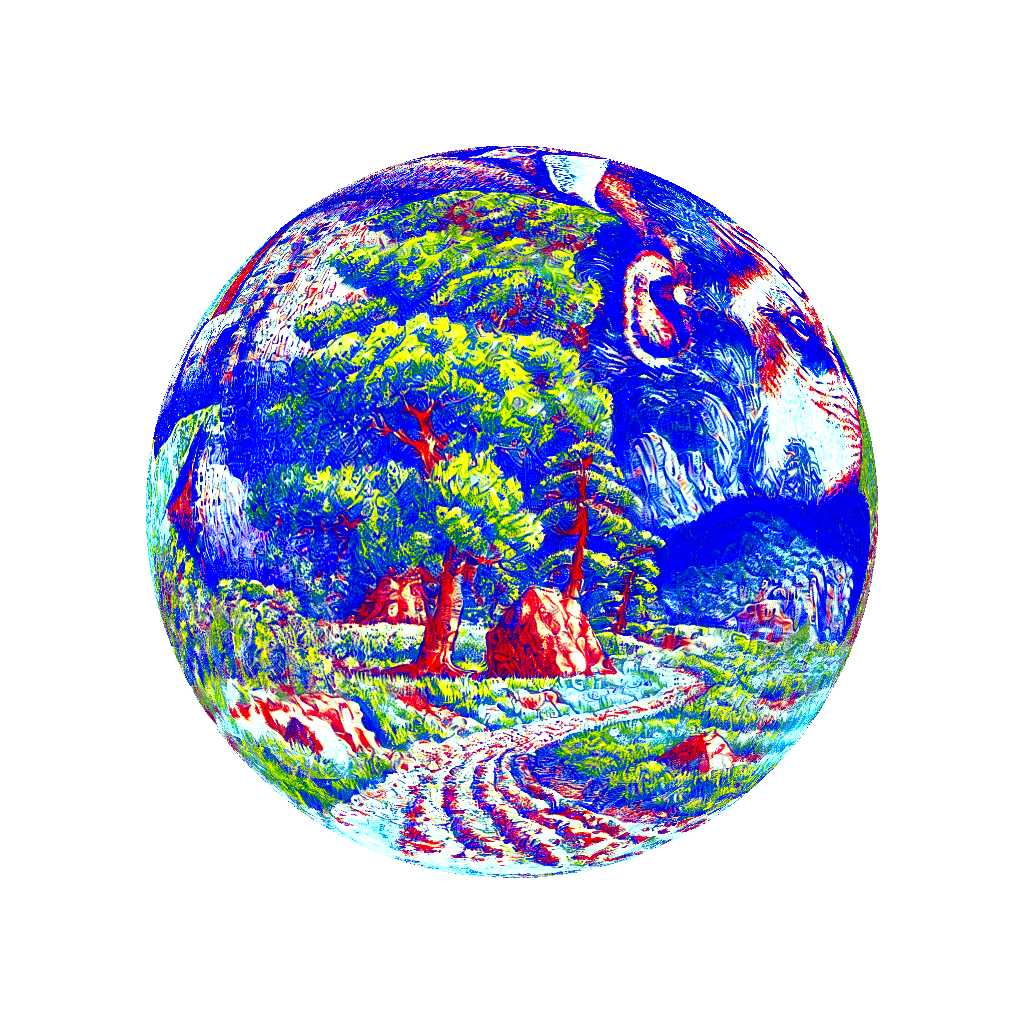}
        \end{minipage}%
        \begin{minipage}[t]{0.16\textwidth}
            \includegraphics[width=\textwidth, trim=140 140 140 140, clip]{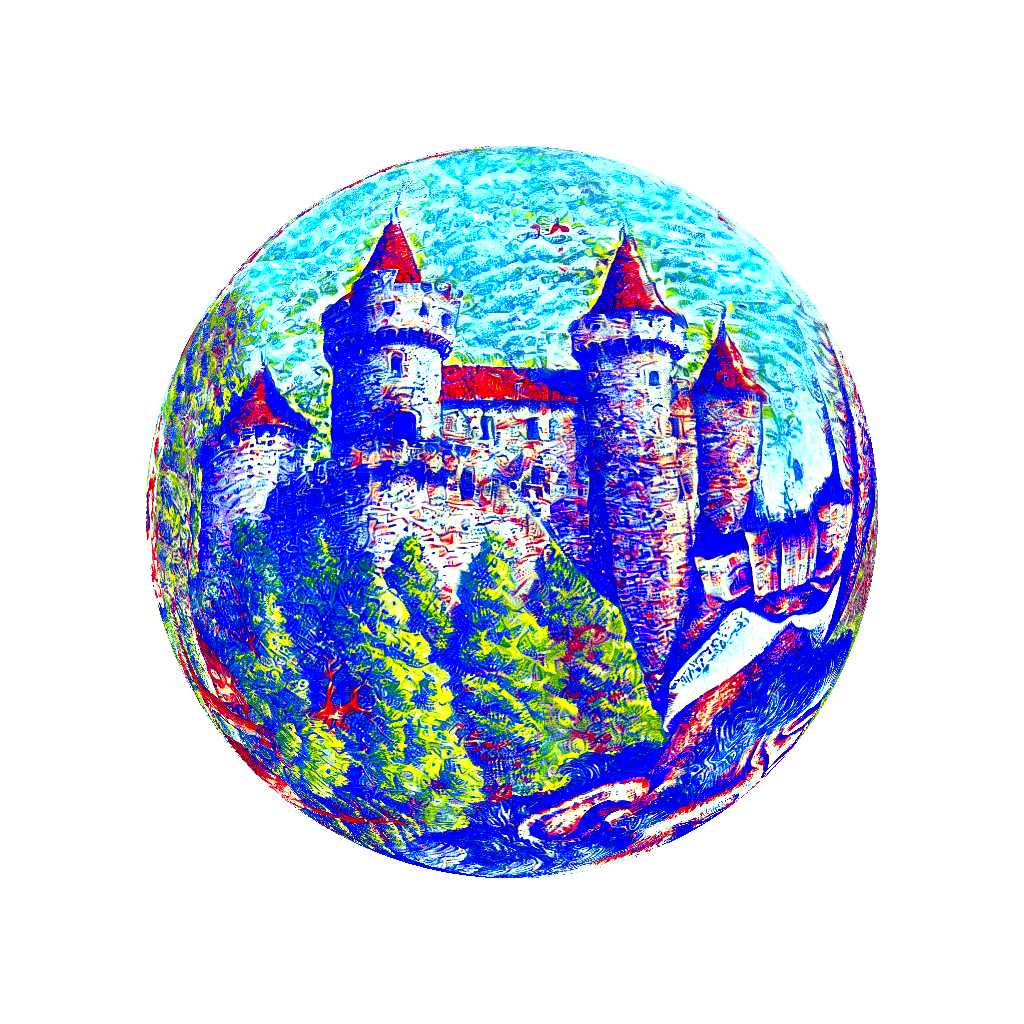}
        \end{minipage}\\
        \vspace{-1mm}
        \begin{minipage}[t]{0.16\textwidth}
            \includegraphics[width=\textwidth, trim=140 140 140 140, clip]{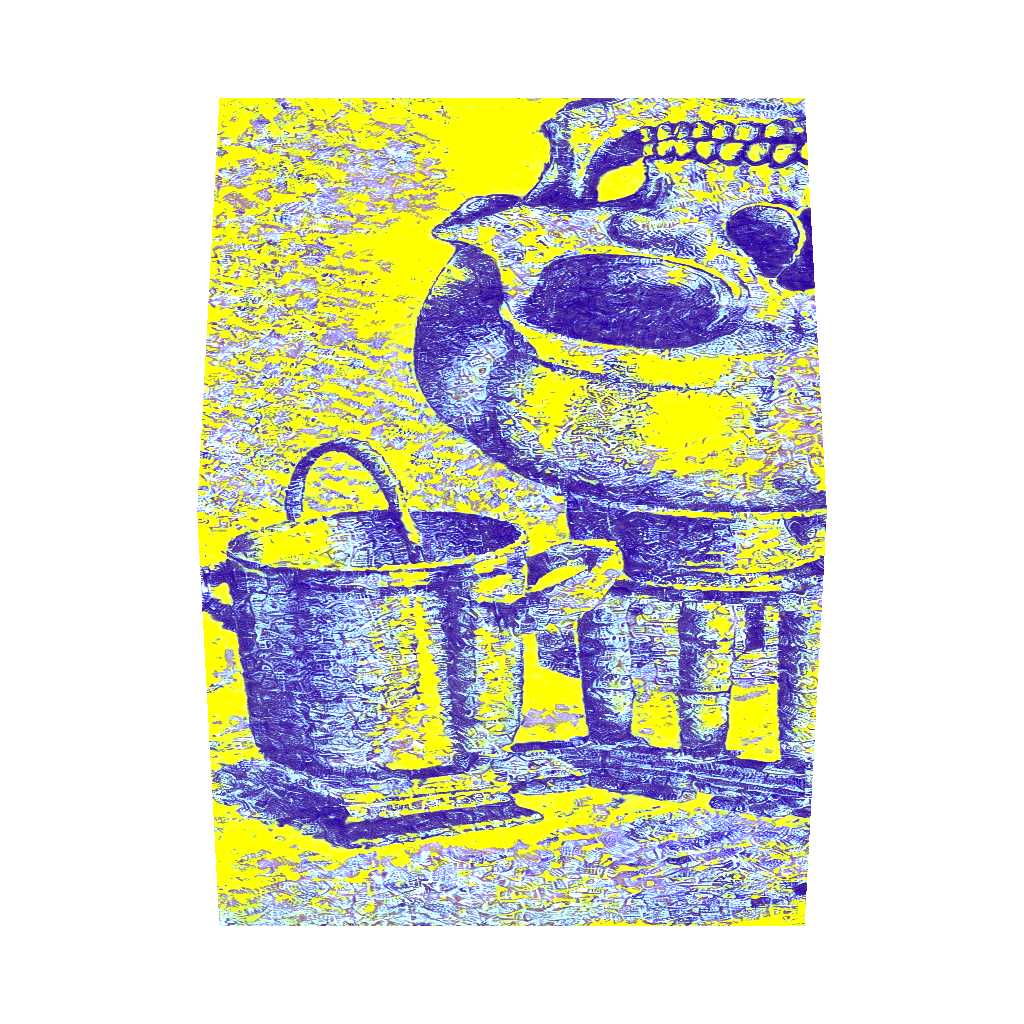}
        \end{minipage}%
        \begin{minipage}[t]{0.16\textwidth}
            \includegraphics[width=\textwidth, trim=140 140 140 140, clip]{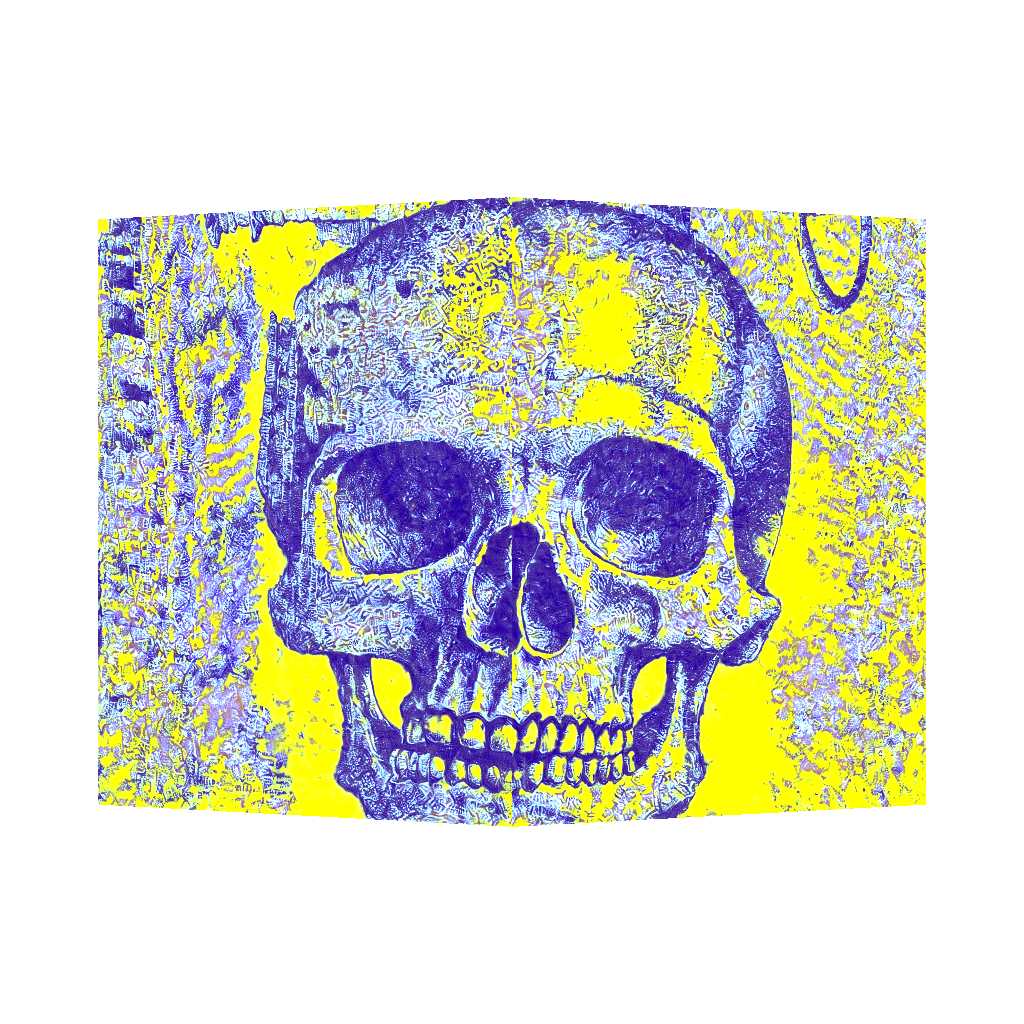}
        \end{minipage}%
        \begin{minipage}[t]{0.16\textwidth}
            \includegraphics[width=\textwidth, trim=140 140 140 140, clip]{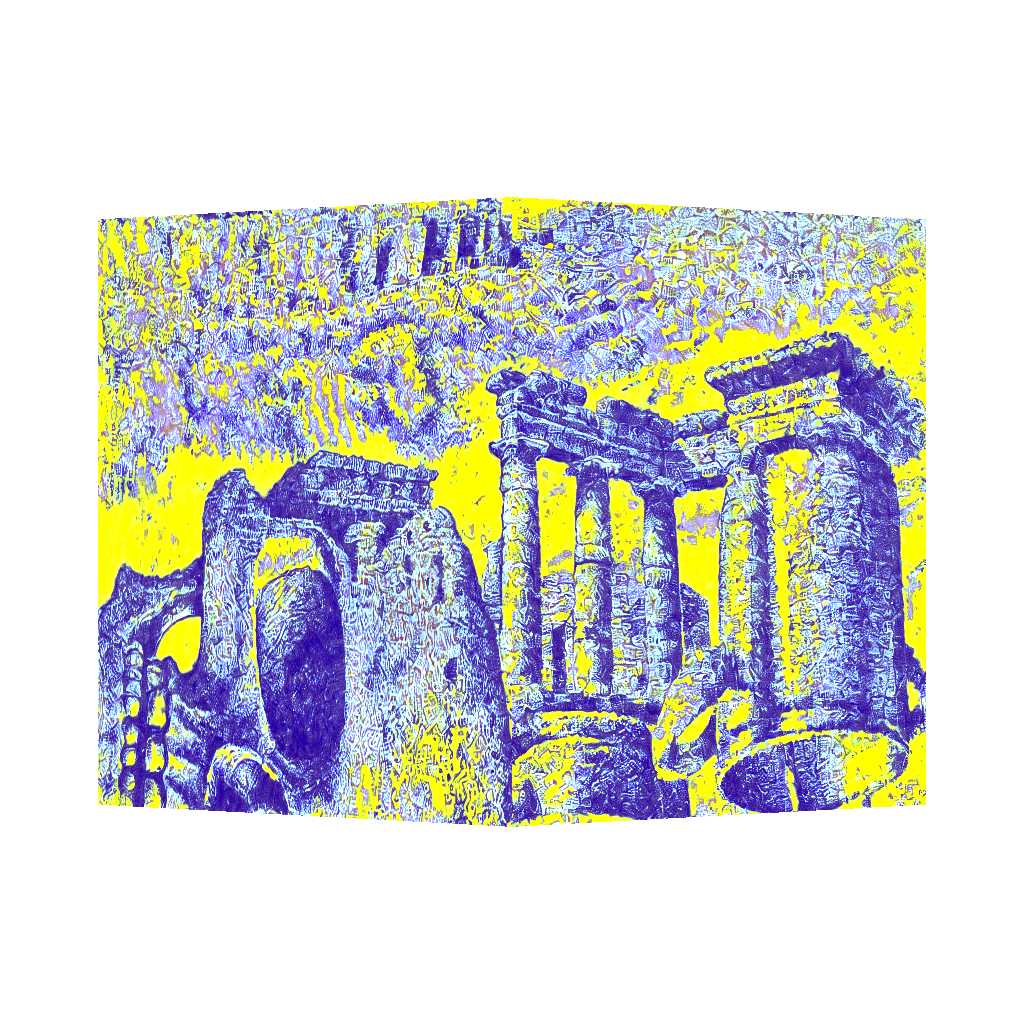}
        \end{minipage}%
        \begin{minipage}[t]{0.16\textwidth}
            \includegraphics[width=\textwidth, trim=140 140 140 140, clip]{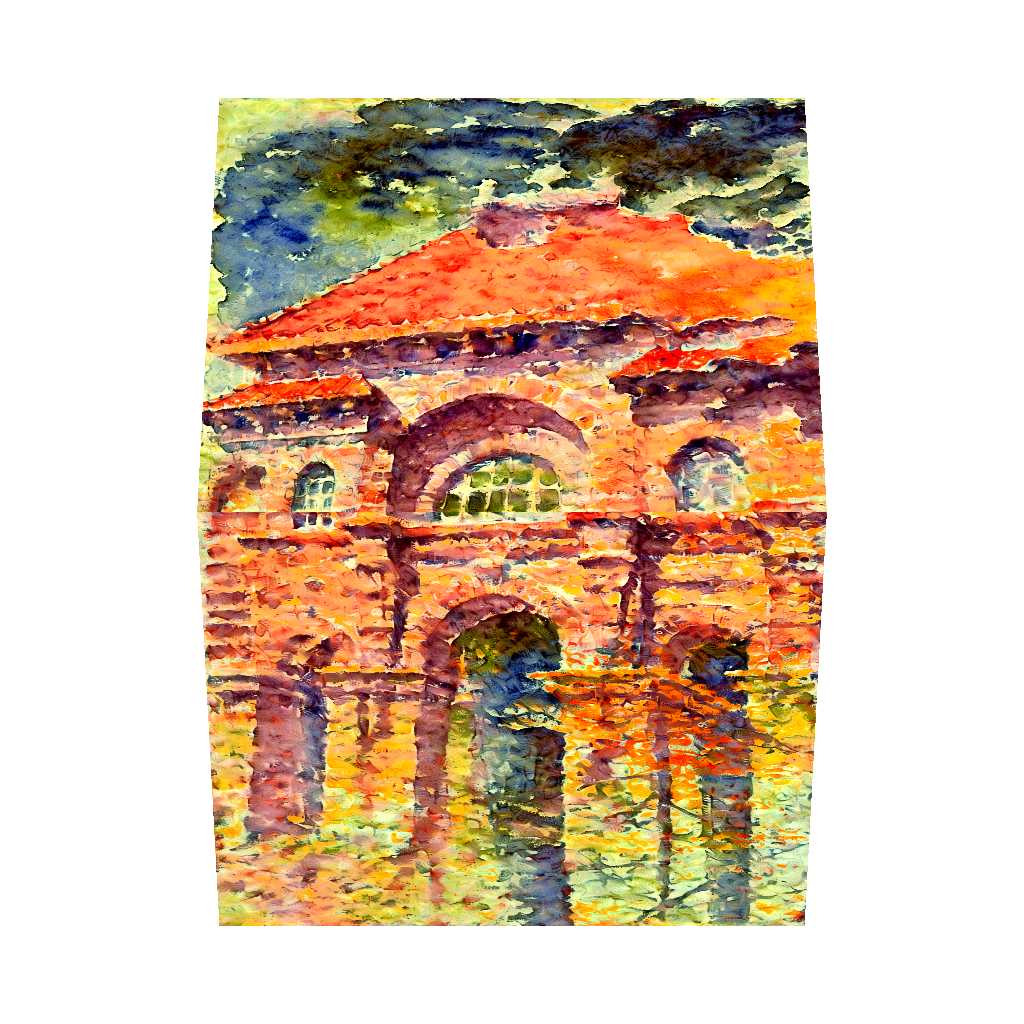}
        \end{minipage}%
        \begin{minipage}[t]{0.16\textwidth}
            \includegraphics[width=\textwidth, trim=140 140 140 140, clip]{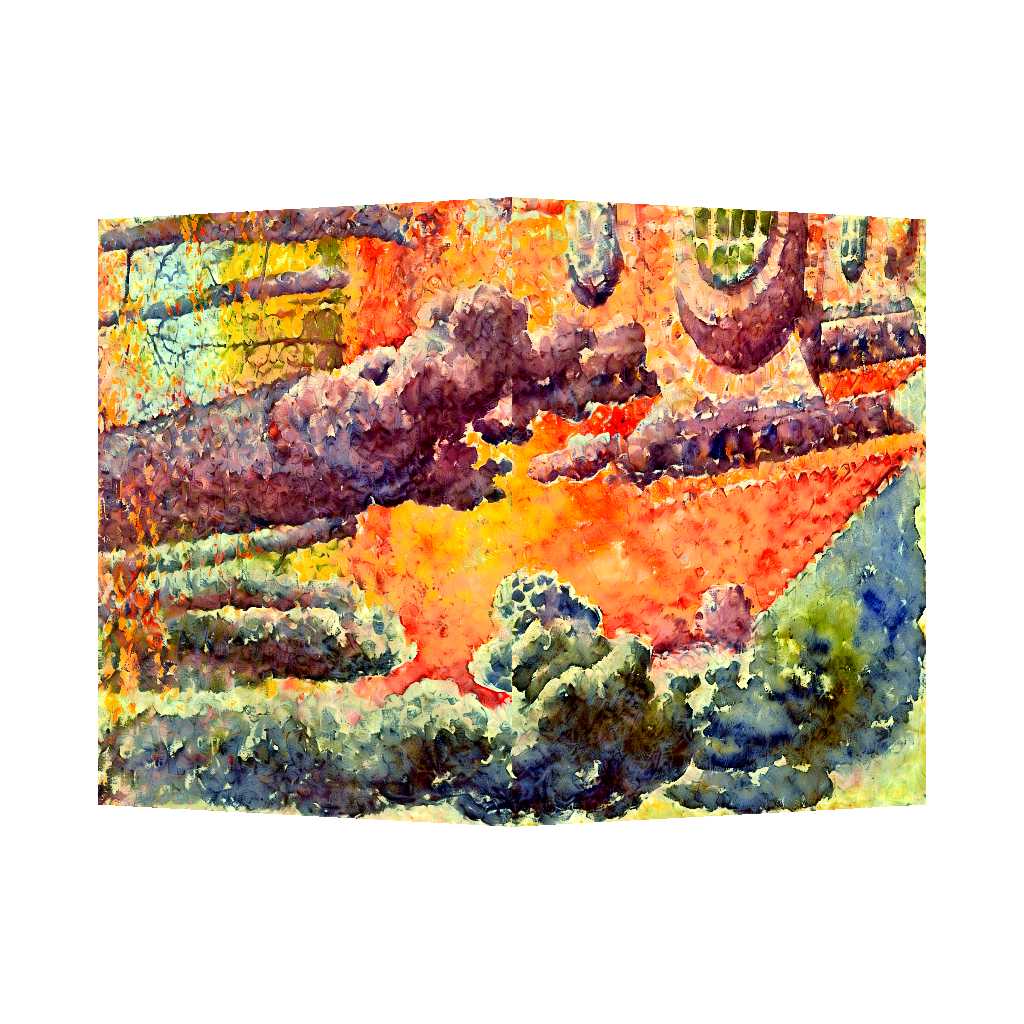}
        \end{minipage}%
        \begin{minipage}[t]{0.16\textwidth}
            \includegraphics[width=\textwidth, trim=140 140 140 140, clip]{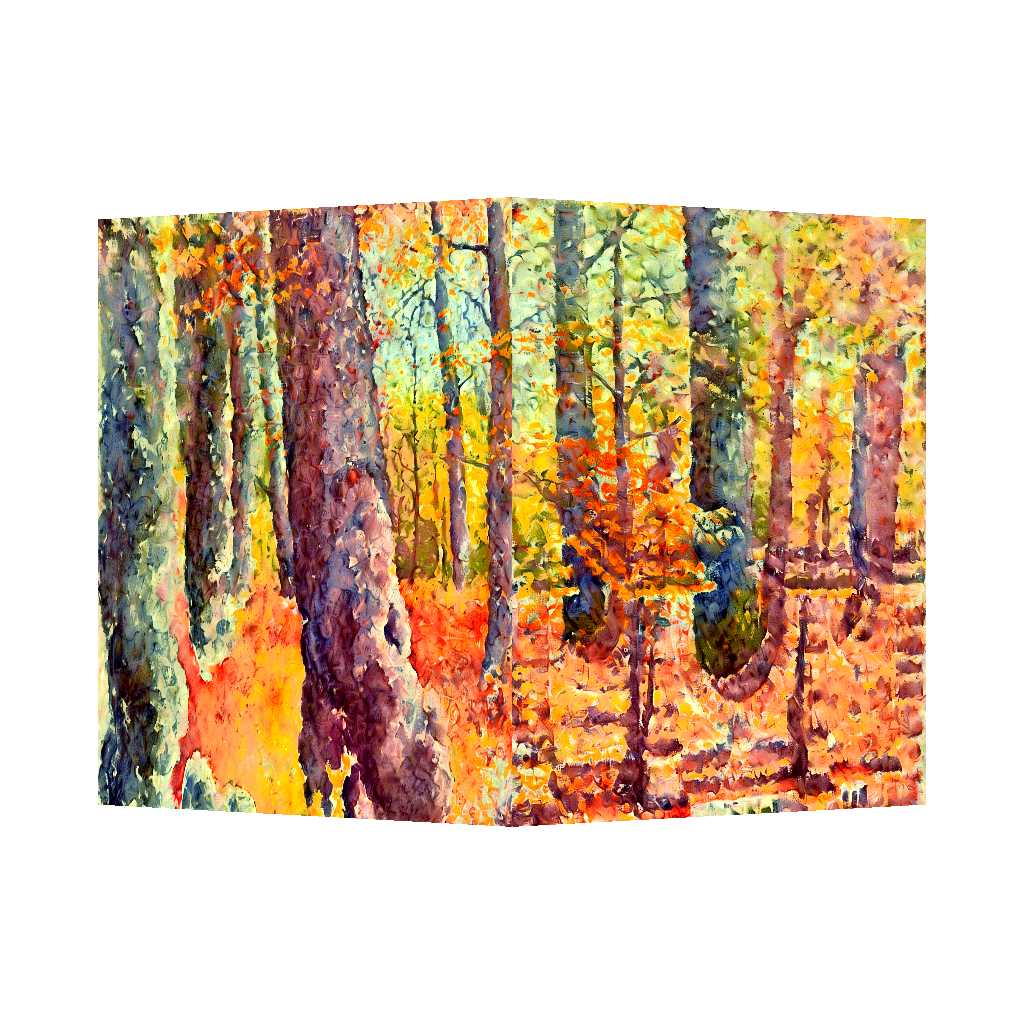}
        \end{minipage}\\

        \vspace{-1mm}
        \begin{minipage}[t]{0.16\textwidth}
        \centering
            \includegraphics[width=\linewidth, trim=185 20 185 30, clip]{figures/prompt/kitchenware.jpg}
            \end{minipage}\hfill
        \begin{minipage}[t]{0.16\textwidth}
            \centering
            \includegraphics[width=\linewidth, trim=185 20 185 30, clip]{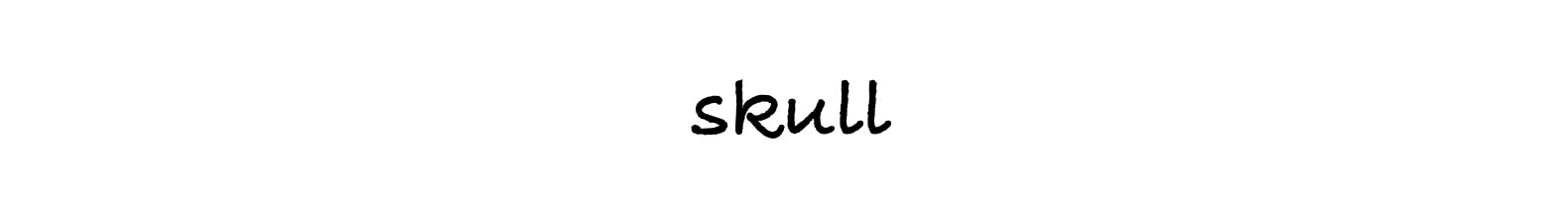}
            \end{minipage}\hfill
        \begin{minipage}[t]{0.16\textwidth}
            \centering
            \includegraphics[width=\linewidth, trim=185 20 185 30, clip]{figures/prompt/ar.jpg}
        \end{minipage}\hfill
        \begin{minipage}[t]{0.16\textwidth}
            \centering
            \includegraphics[width=\linewidth, trim=185 20 185 30, clip]{figures/prompt/museum.jpg}
        \end{minipage}
        \begin{minipage}[t]{0.16\textwidth}
            \centering
            \includegraphics[width=\linewidth, trim=185 20 185 30, clip]{figures/prompt/cloud.jpg}
        \end{minipage}
        \begin{minipage}[t]{0.16\textwidth}
            \centering
            \includegraphics[width=\linewidth, trim=185 20 185 30, clip]{figures/prompt/forest.jpg}
        \end{minipage}\\

        \vspace{-1mm}
        \begin{minipage}[t]{0.16\textwidth}
            \includegraphics[width=\textwidth, trim=140 140 140 140, clip]{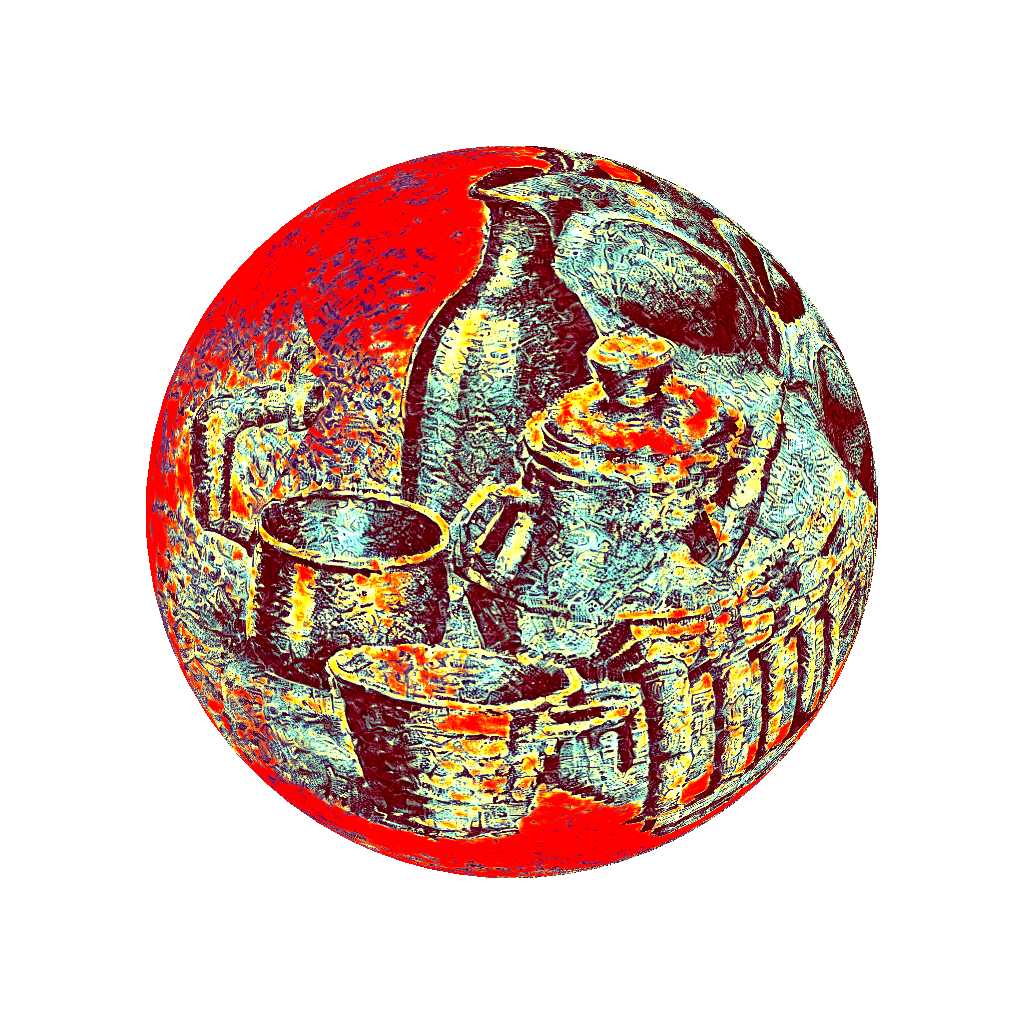}
        \end{minipage}%
        \begin{minipage}[t]{0.16\textwidth}
            \includegraphics[width=\textwidth, trim=140 140 140 140, clip]{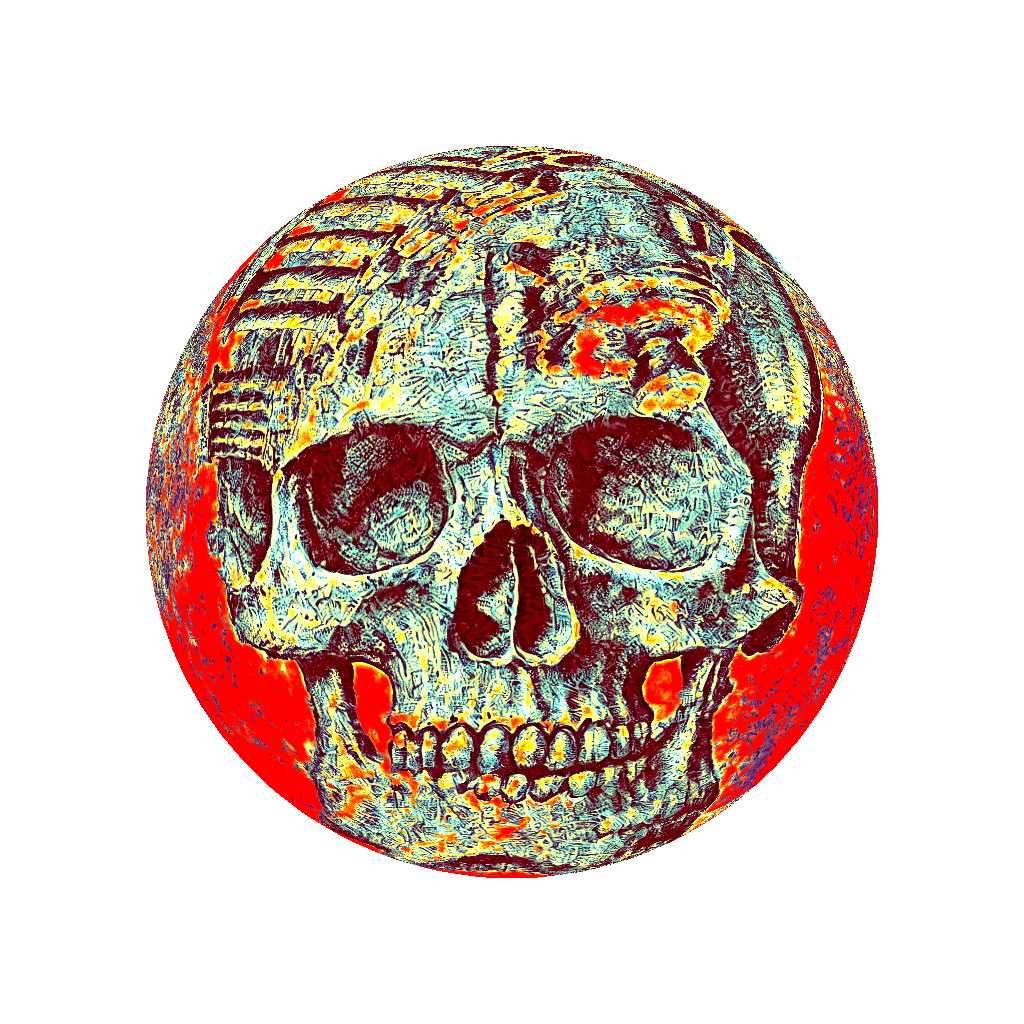}
        \end{minipage}%
        \begin{minipage}[t]{0.16\textwidth}
            \includegraphics[width=\textwidth, trim=140 140 140 140, clip]{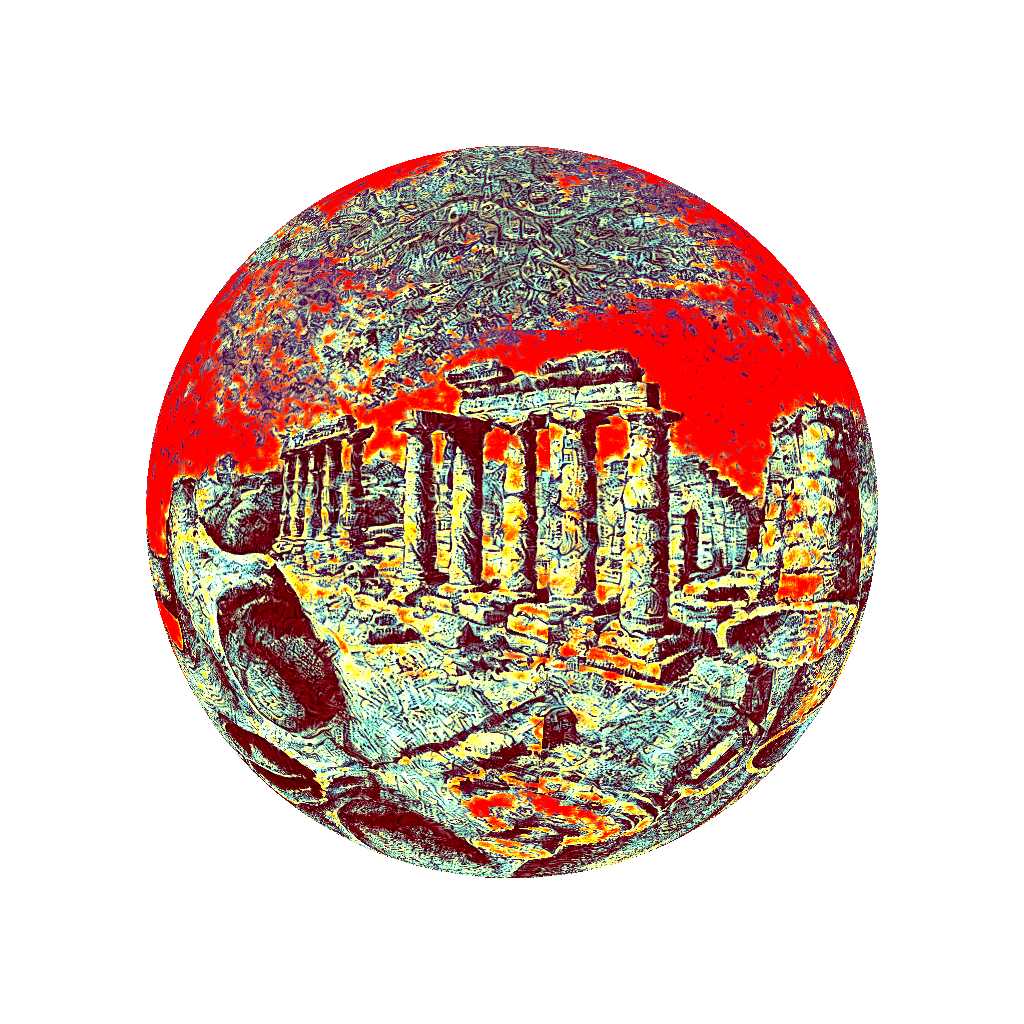}
        \end{minipage}%
        \begin{minipage}[t]{0.16\textwidth}
            \includegraphics[width=\textwidth, trim=140 140 140 140, clip]{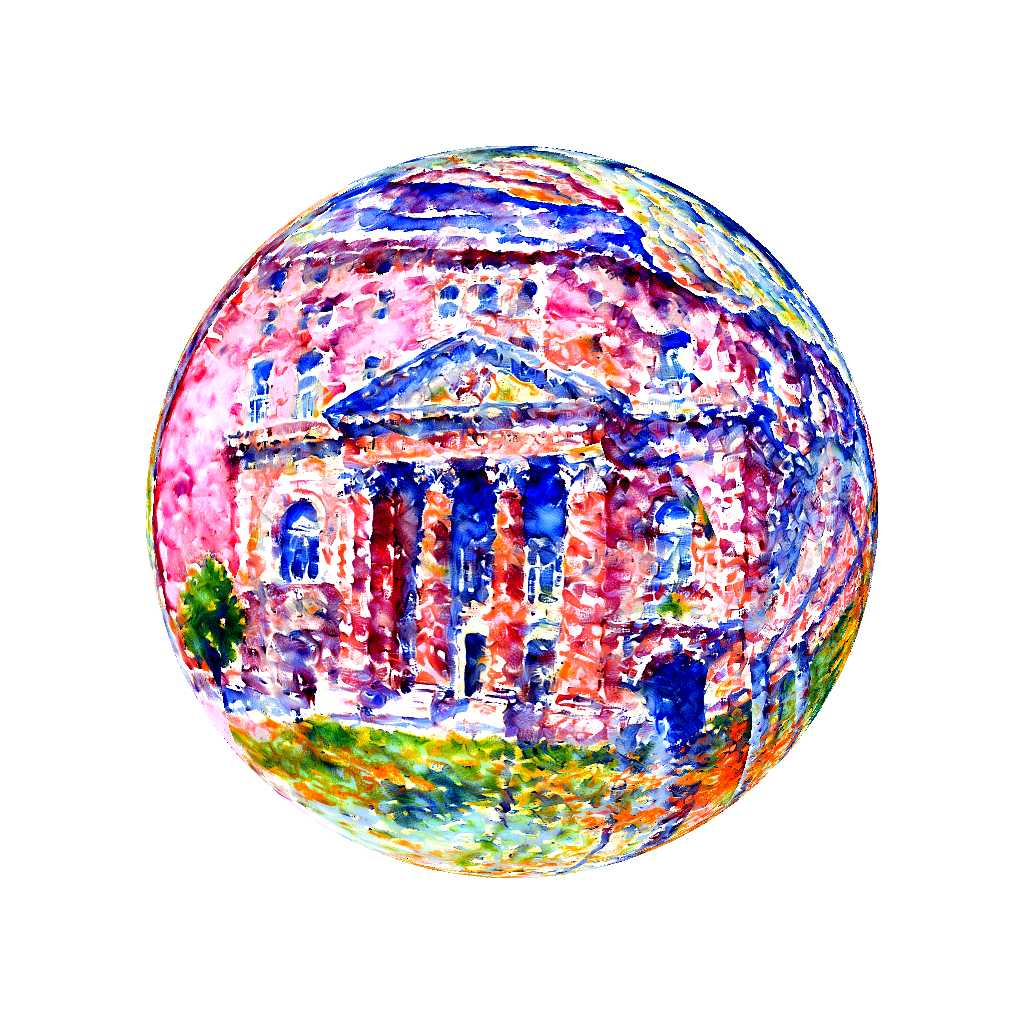}
        \end{minipage}%
        \begin{minipage}[t]{0.16\textwidth}
            \includegraphics[width=\textwidth, trim=140 140 140 140, clip]{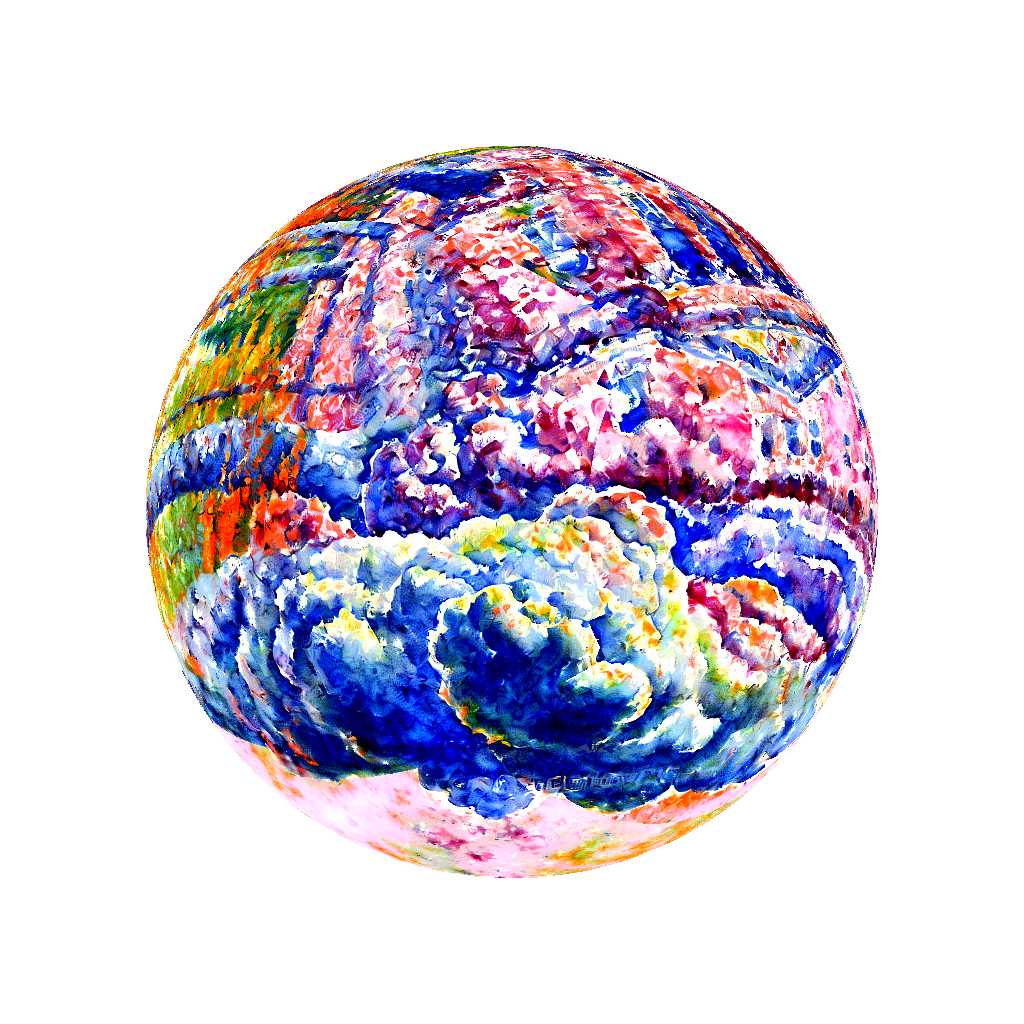}
        \end{minipage}%
        \begin{minipage}[t]{0.16\textwidth}
            \includegraphics[width=\textwidth, trim=140 140 140 140, clip]{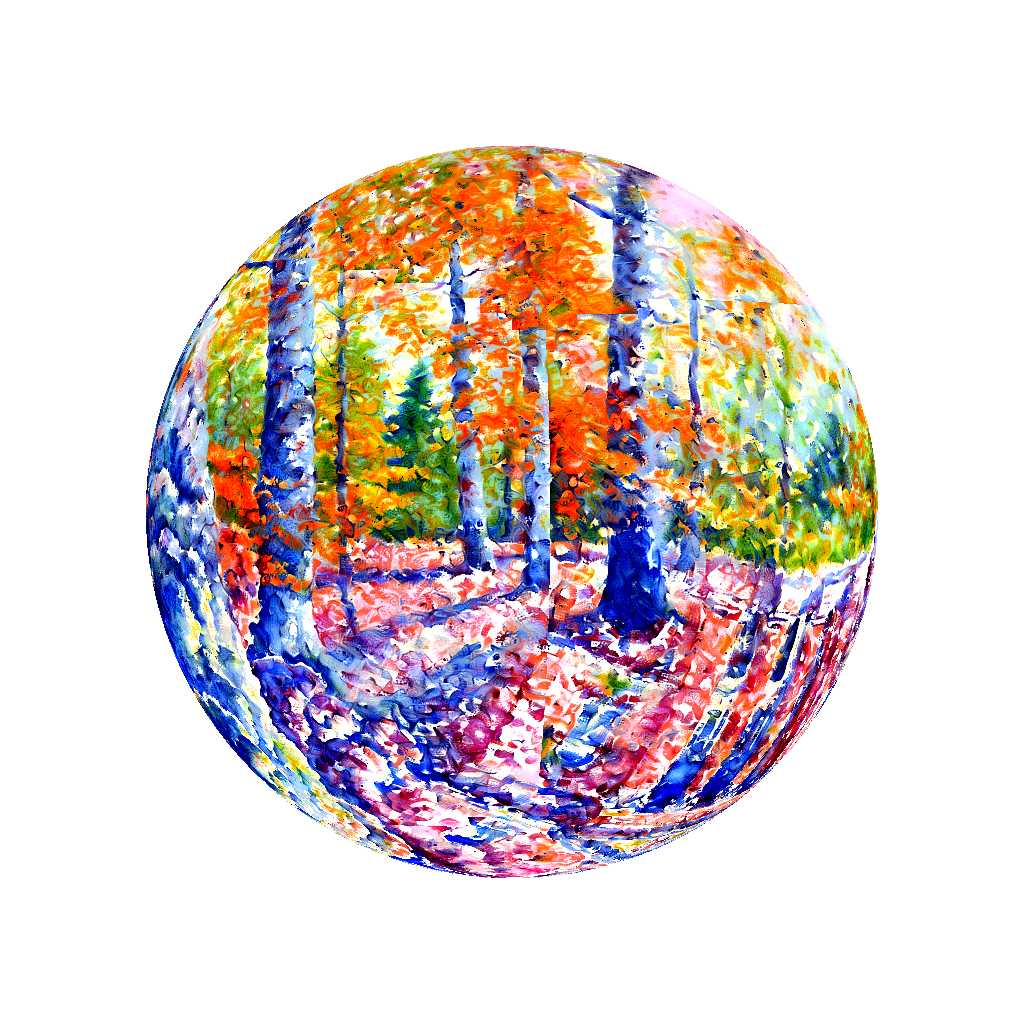}
        \end{minipage}\\
        \vspace{-1mm}
        \begin{minipage}[t]{0.16\textwidth}
            \includegraphics[width=\textwidth, trim=140 140 140 140, clip]{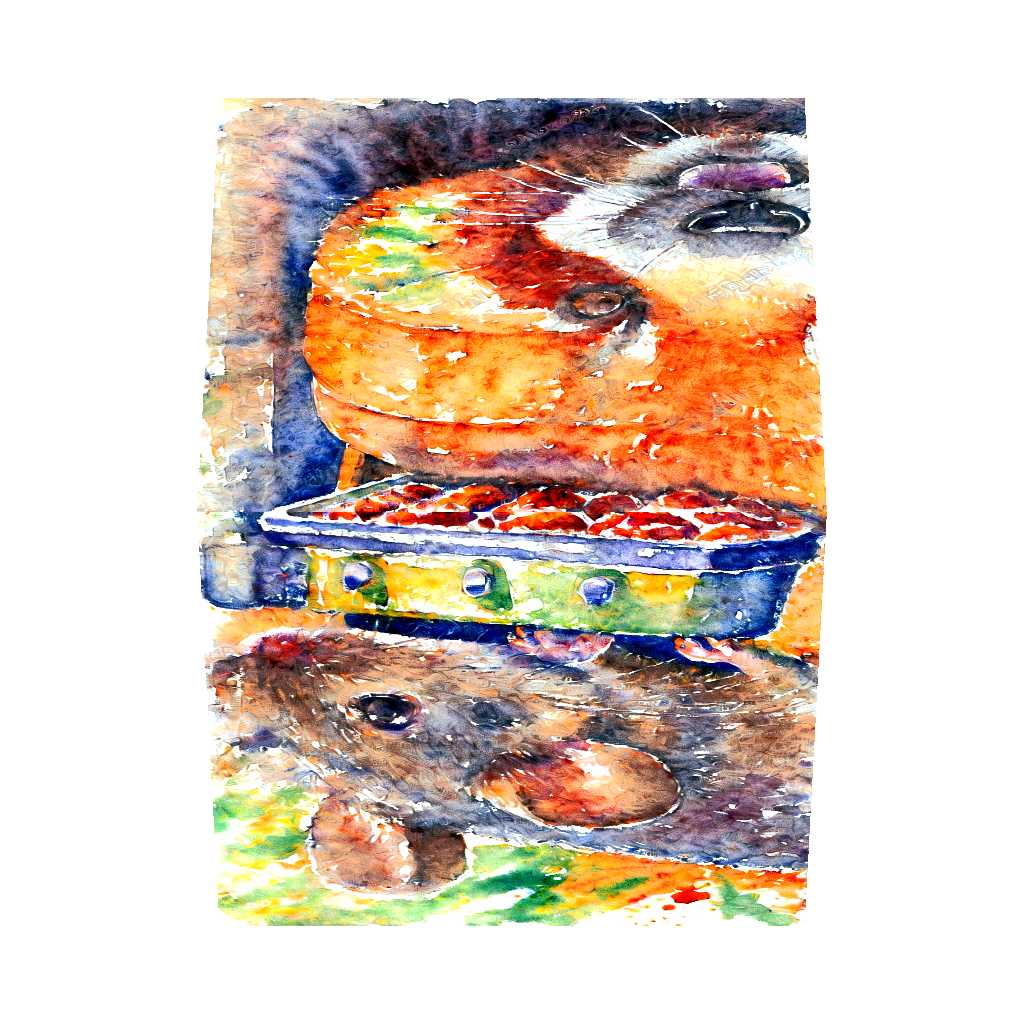}
        \end{minipage}%
        \begin{minipage}[t]{0.16\textwidth}
            \includegraphics[width=\textwidth, trim=140 140 140 140, clip]{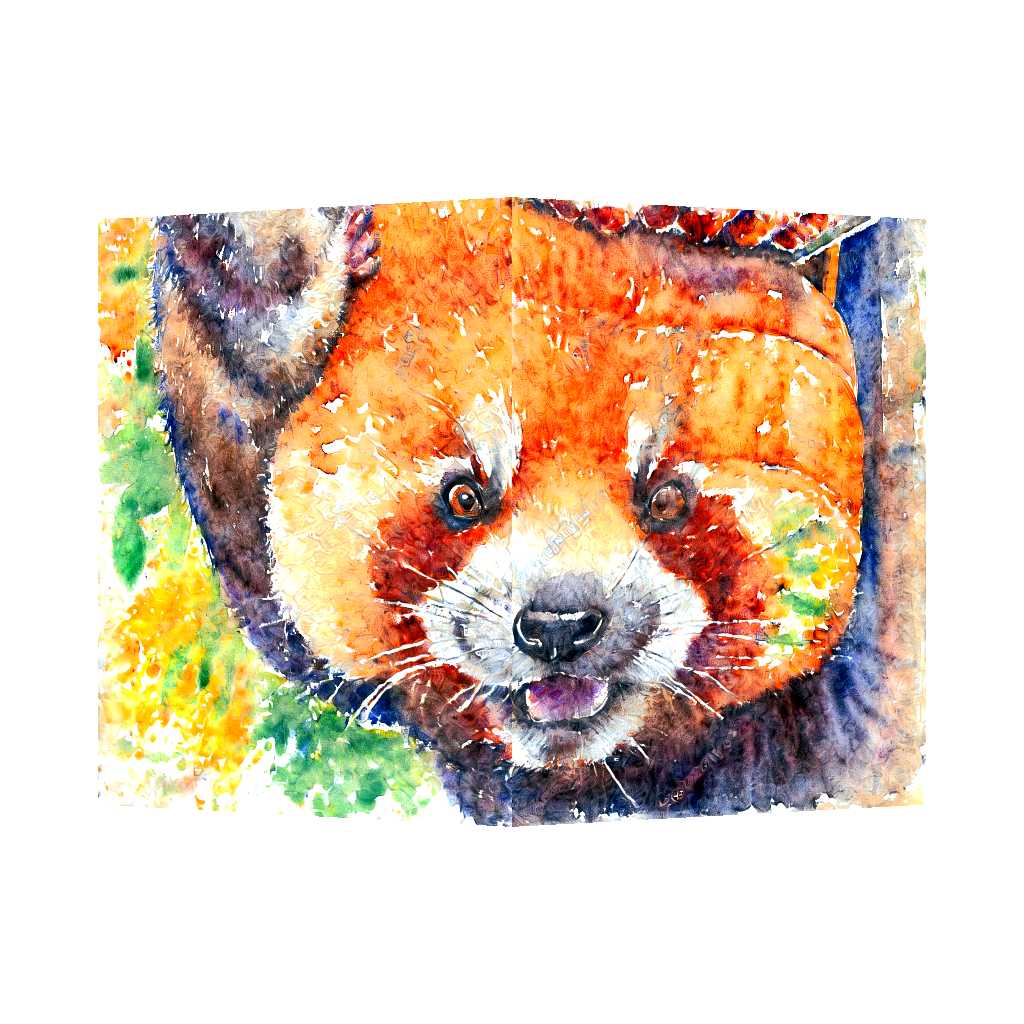}
        \end{minipage}%
        \begin{minipage}[t]{0.16\textwidth}
            \includegraphics[width=\textwidth, trim=140 140 140 140, clip]{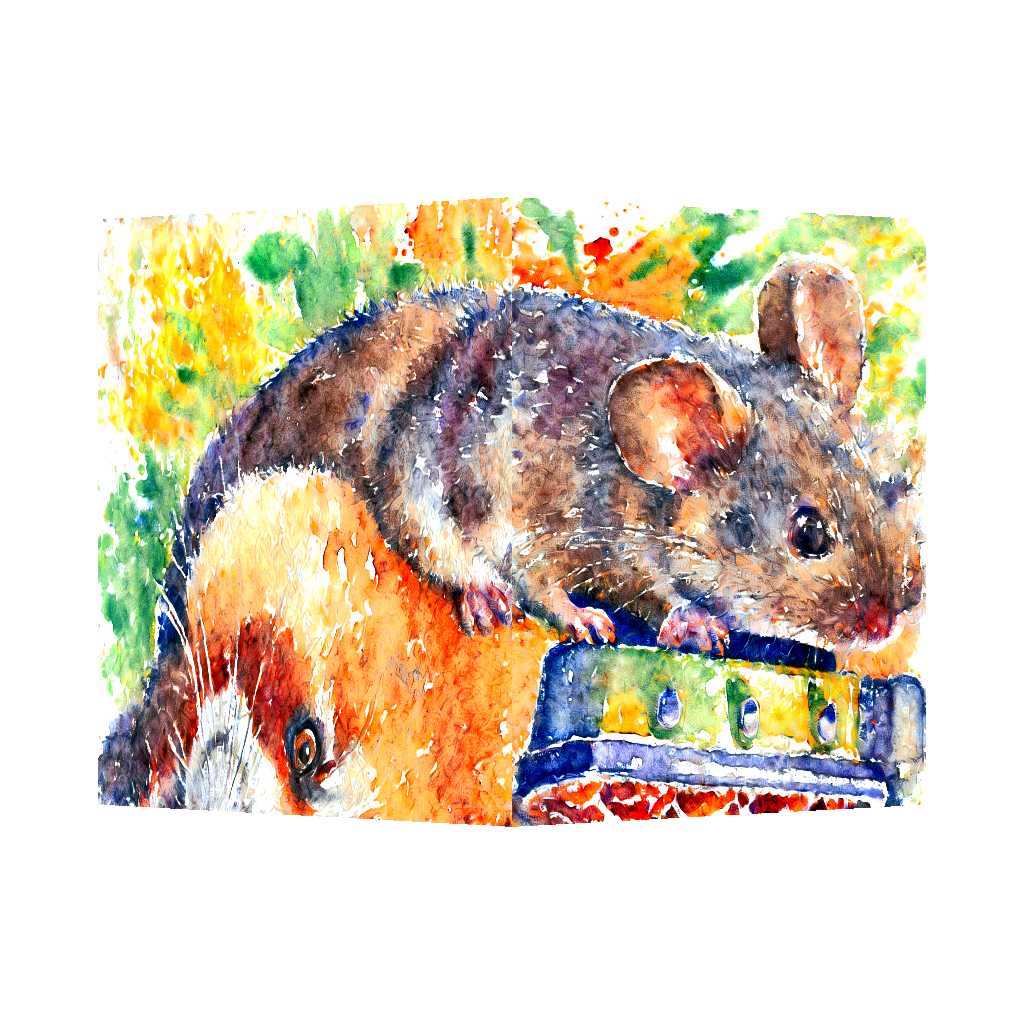}
        \end{minipage}%
        \begin{minipage}[t]{0.16\textwidth}
            \includegraphics[width=\textwidth, trim=140 140 140 140, clip]{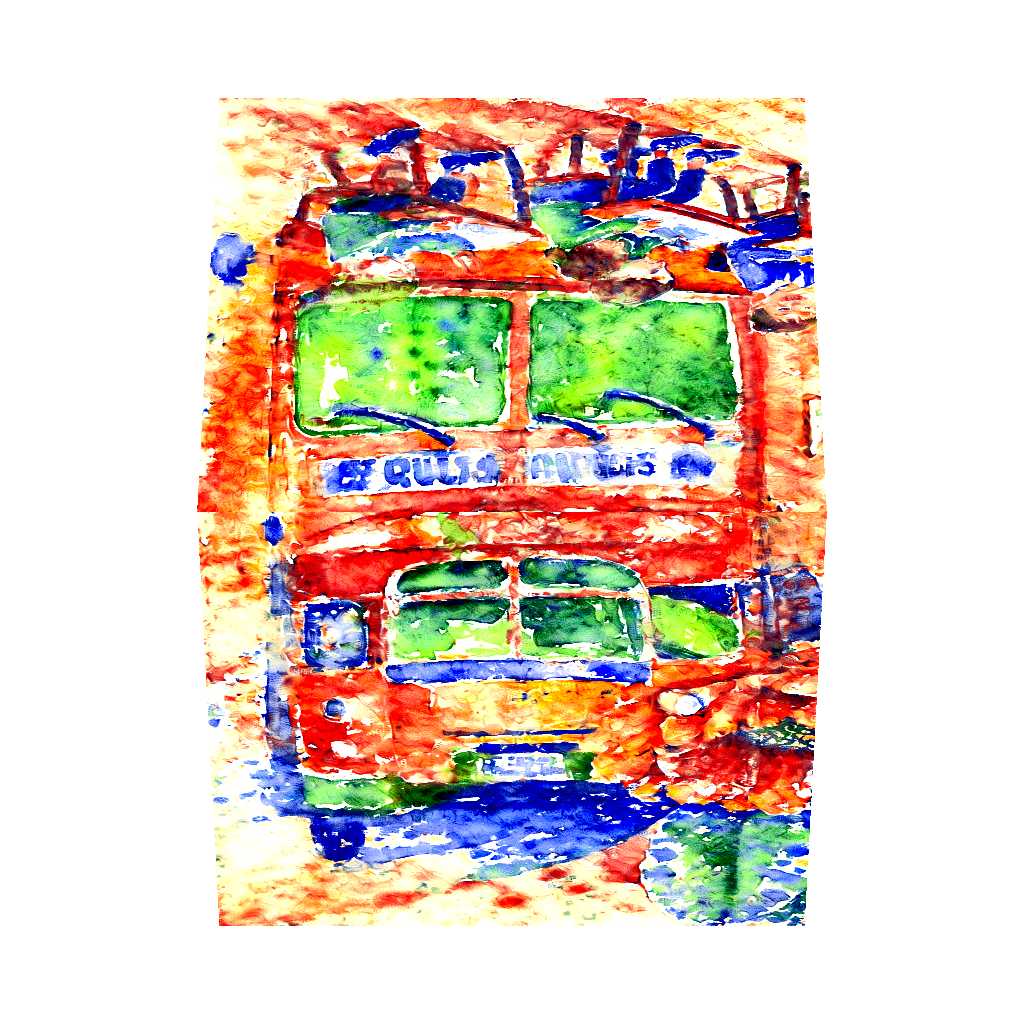}
        \end{minipage}%
        \begin{minipage}[t]{0.16\textwidth}
            \includegraphics[width=\textwidth, trim=140 140 140 140, clip]{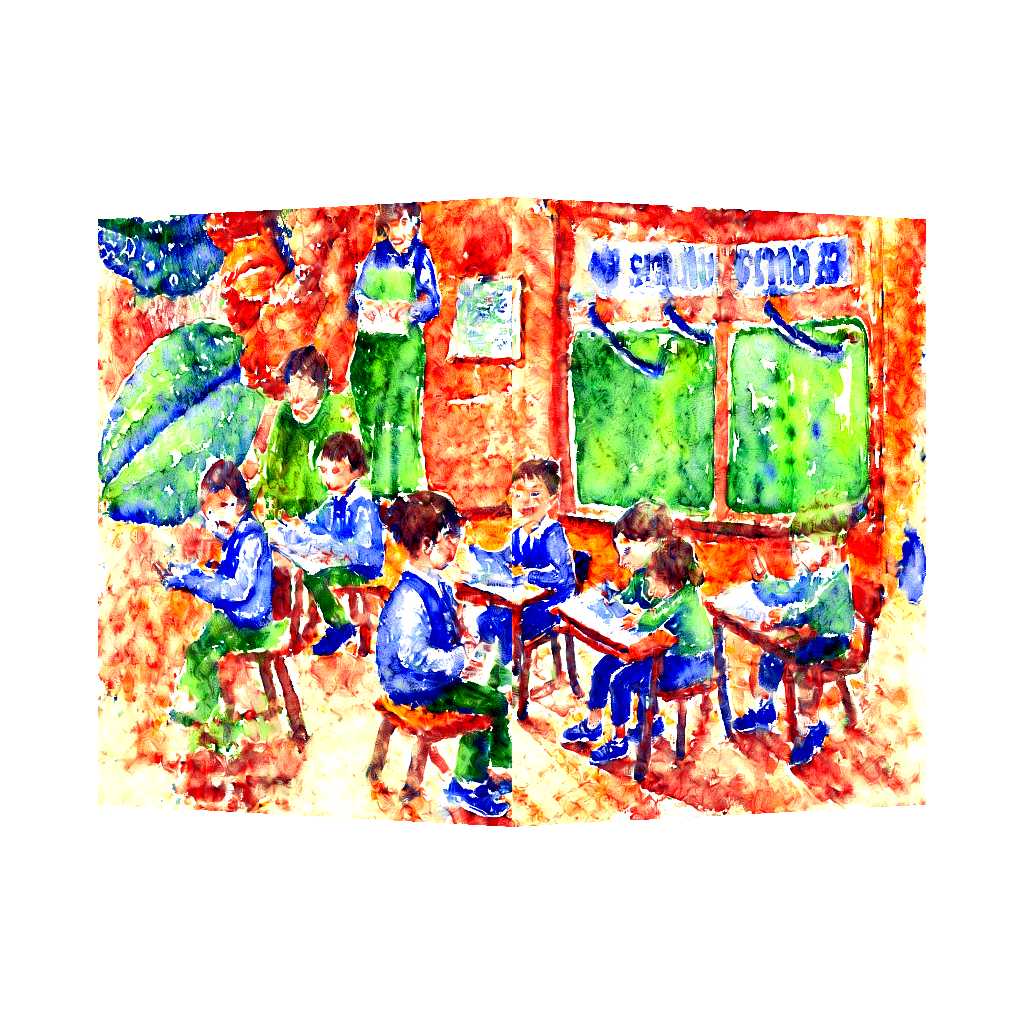}
        \end{minipage}%
        \begin{minipage}[t]{0.16\textwidth}
            \includegraphics[width=\textwidth, trim=140 140 140 140, clip]{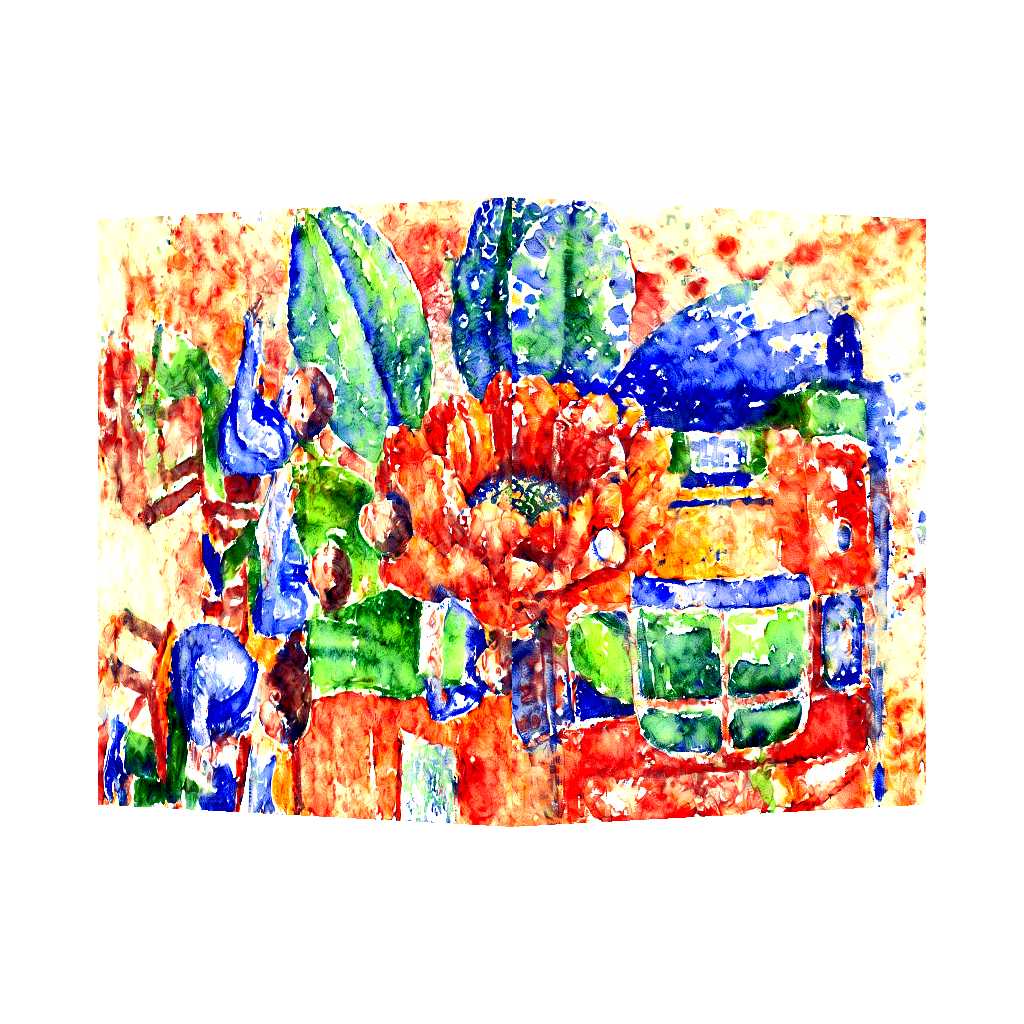}
        \end{minipage}\\

        \vspace{-1mm}
        \begin{minipage}[t]{0.16\textwidth}
        \centering
            \includegraphics[width=\linewidth, trim=185 20 185 30, clip]{figures/prompt/oven.jpg}
            \end{minipage}\hfill
        \begin{minipage}[t]{0.16\textwidth}
            \centering
            \includegraphics[width=\linewidth, trim=185 20 185 30, clip]{figures/prompt/redpanda.jpg}
            \end{minipage}\hfill
        \begin{minipage}[t]{0.16\textwidth}
            \centering
            \includegraphics[width=\linewidth, trim=185 20 185 30, clip]{figures/prompt/mouse.jpg}
        \end{minipage}\hfill
        \begin{minipage}[t]{0.16\textwidth}
            \centering
            \includegraphics[width=\linewidth, trim=185 20 185 30, clip]{figures/prompt/bus.jpg}
        \end{minipage}
        \begin{minipage}[t]{0.16\textwidth}
            \centering
            \includegraphics[width=\linewidth, trim=165 20 165 30, clip]{figures/prompt/students.jpg}
        \end{minipage}
        \begin{minipage}[t]{0.16\textwidth}
            \centering
            \includegraphics[width=\linewidth, trim=185 20 185 30, clip]{figures/prompt/flower.jpg}
        \end{minipage}\\

        \vspace{-1mm}
        \begin{minipage}[t]{0.16\textwidth}
            \includegraphics[width=\textwidth, trim=140 140 140 140, clip]{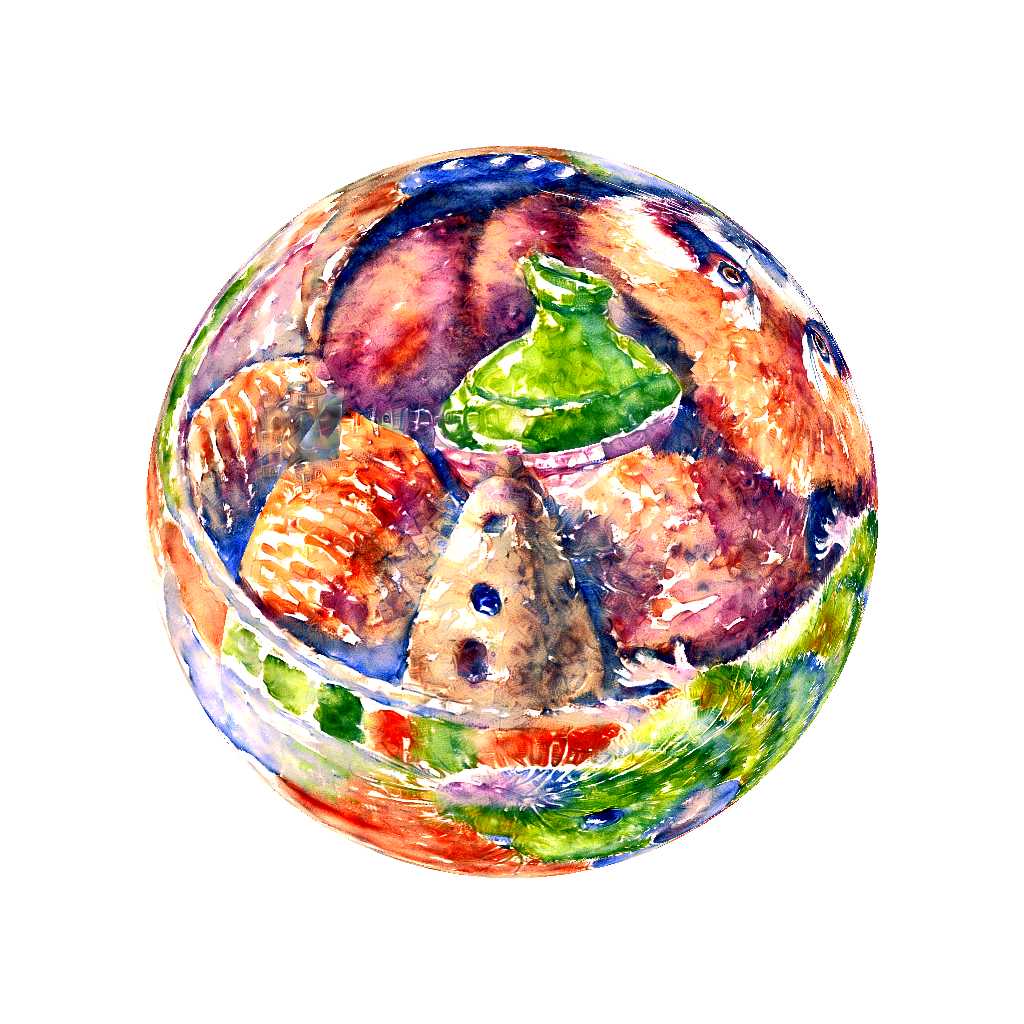}
        \end{minipage}%
        \begin{minipage}[t]{0.16\textwidth}
            \includegraphics[width=\textwidth, trim=140 140 140 140, clip]{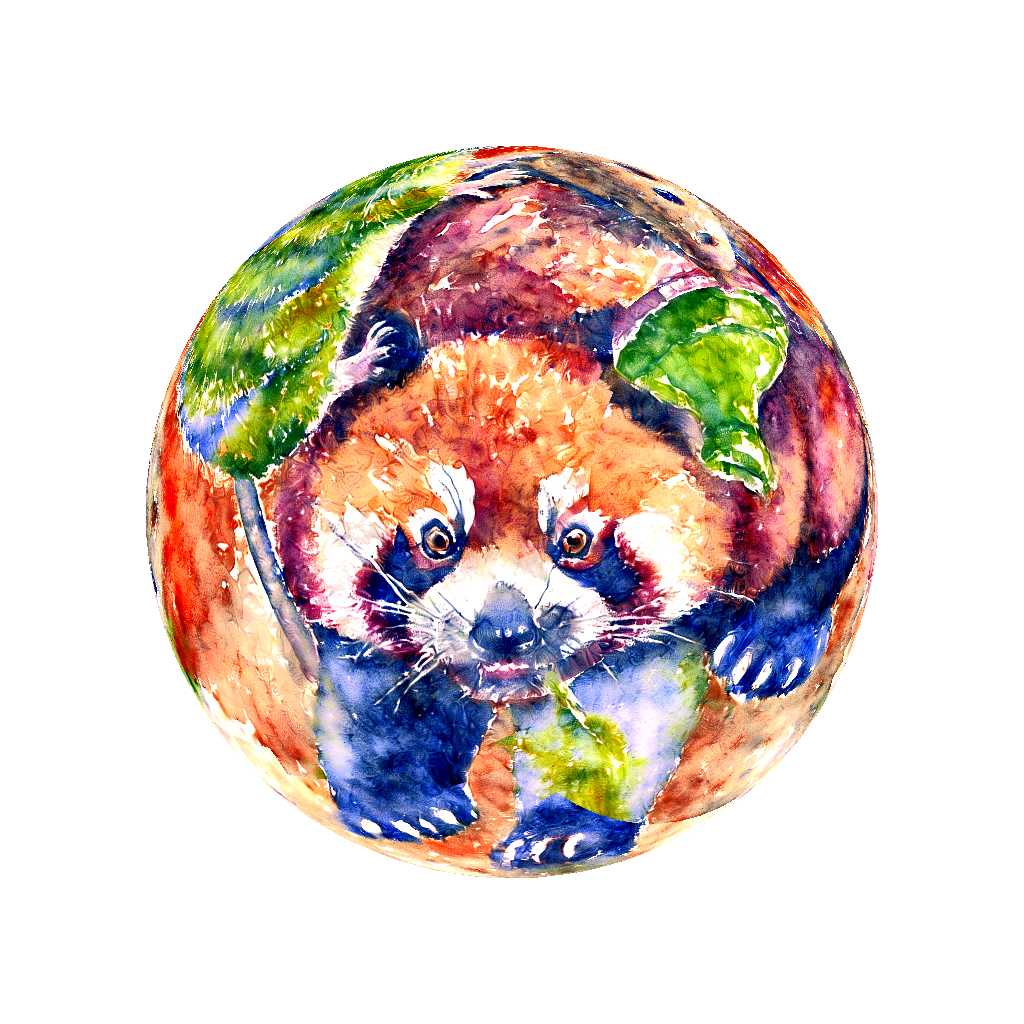}
        \end{minipage}%
        \begin{minipage}[t]{0.16\textwidth}
            \includegraphics[width=\textwidth, trim=140 140 140 140, clip]{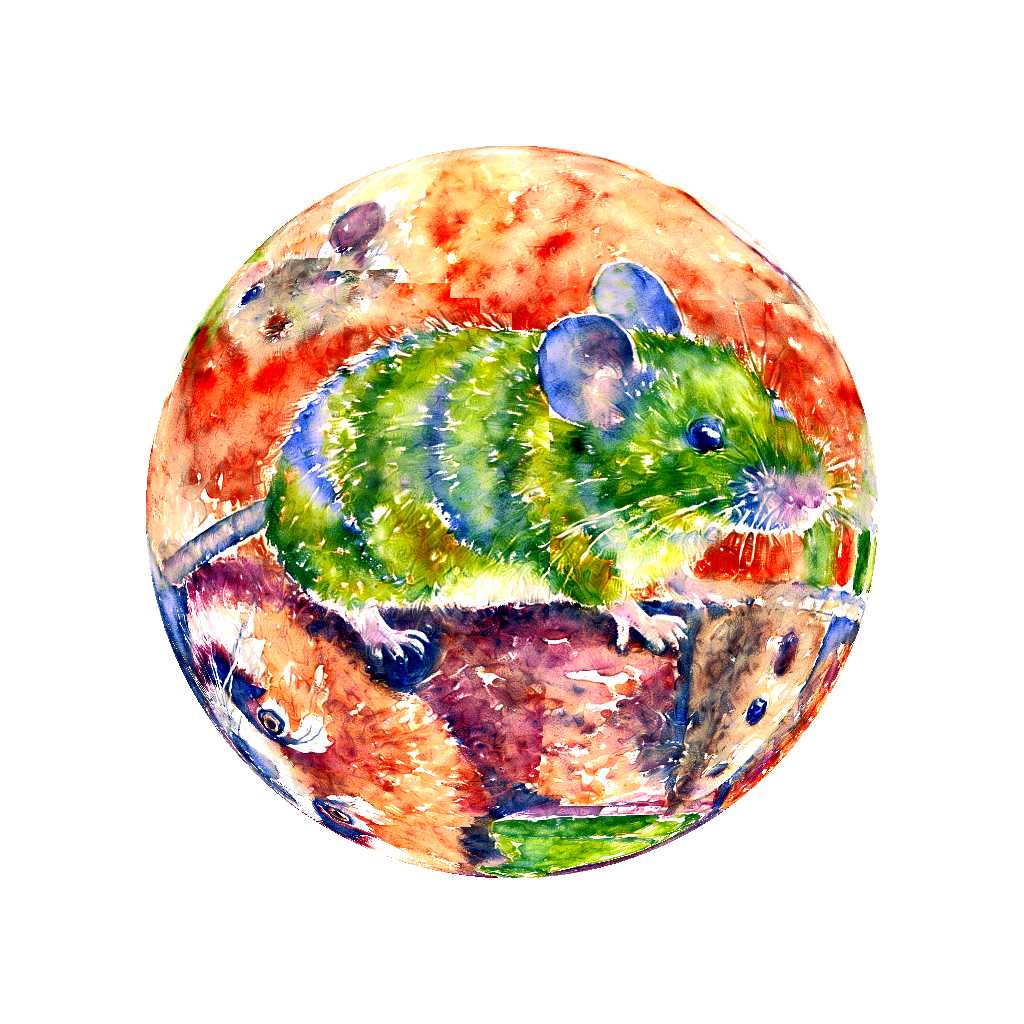}
        \end{minipage}%
        \begin{minipage}[t]{0.16\textwidth}
            \includegraphics[width=\textwidth, trim=140 140 140 140, clip]{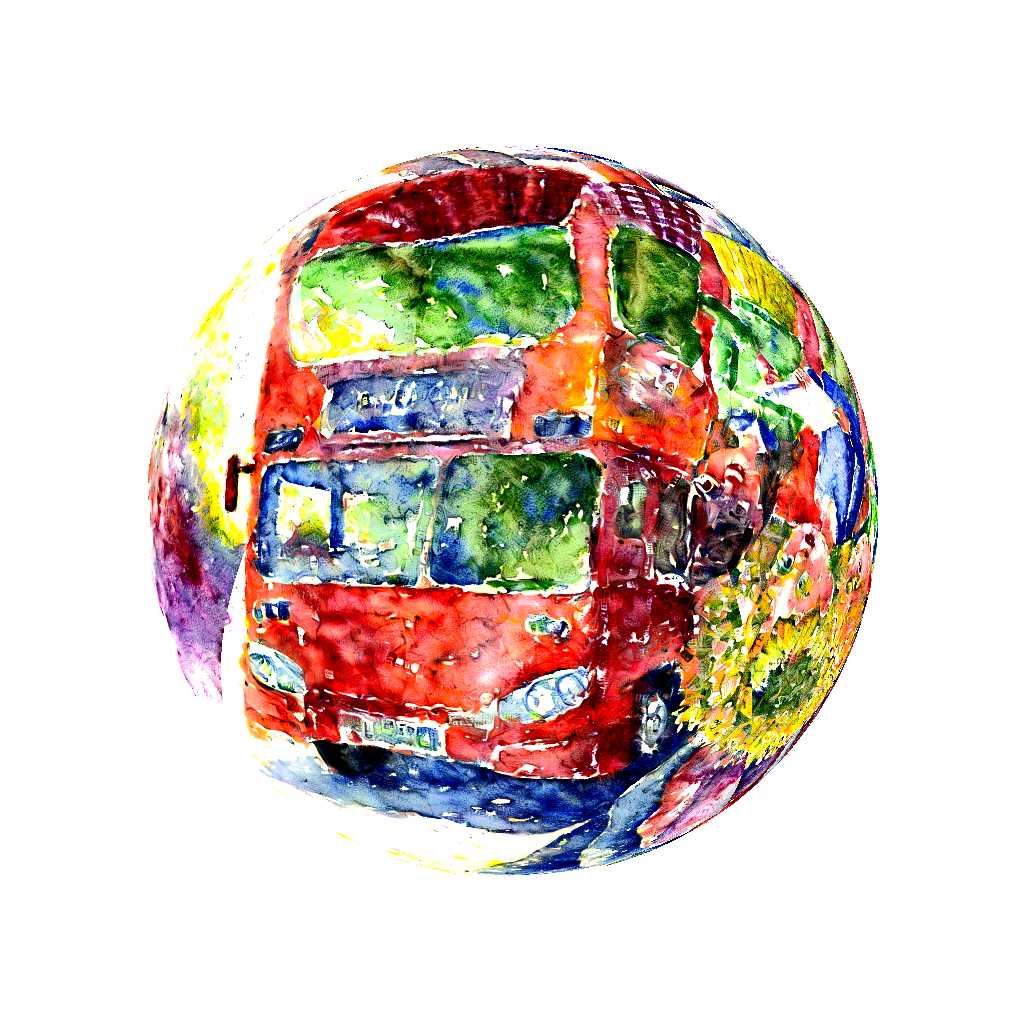}
        \end{minipage}%
        \begin{minipage}[t]{0.16\textwidth}
            \includegraphics[width=\textwidth, trim=140 140 140 140, clip]{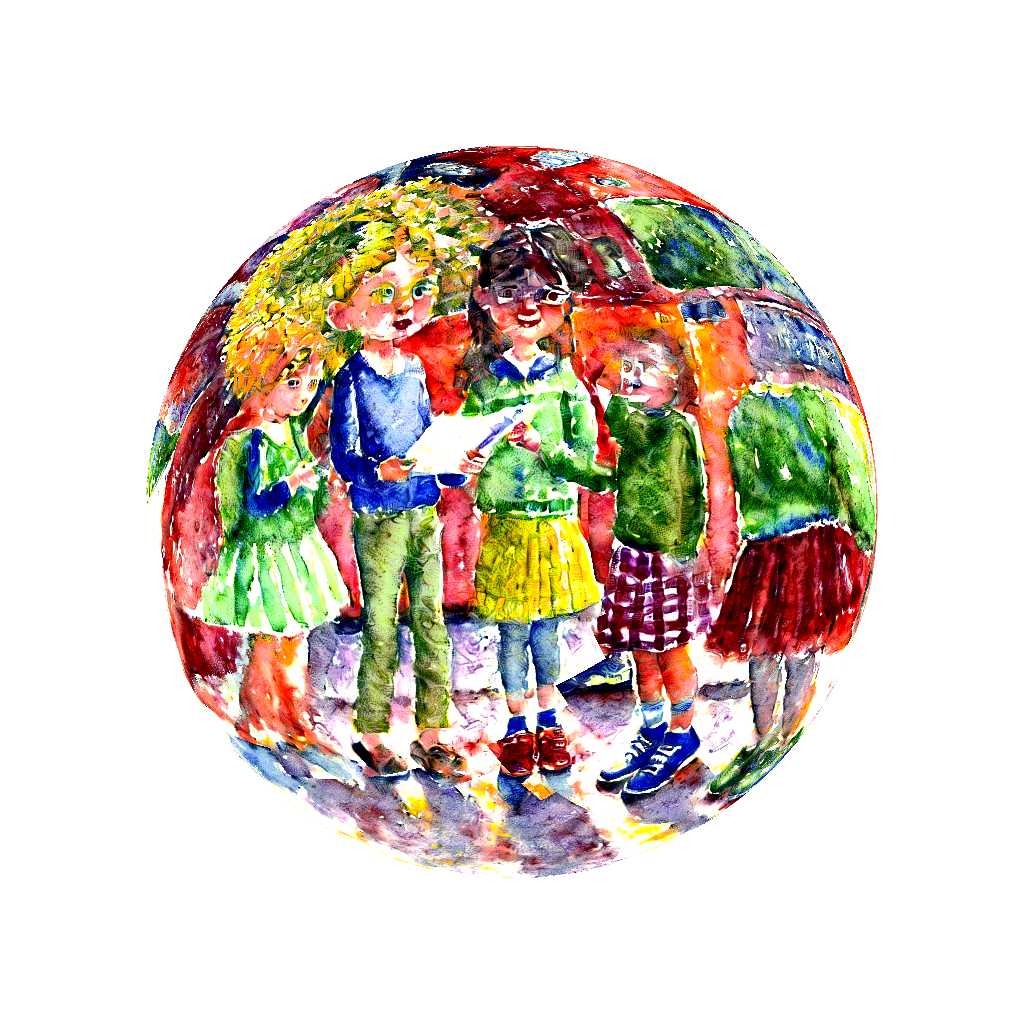}
        \end{minipage}%
        \begin{minipage}[t]{0.16\textwidth}
            \includegraphics[width=\textwidth, trim=140 140 140 140, clip]{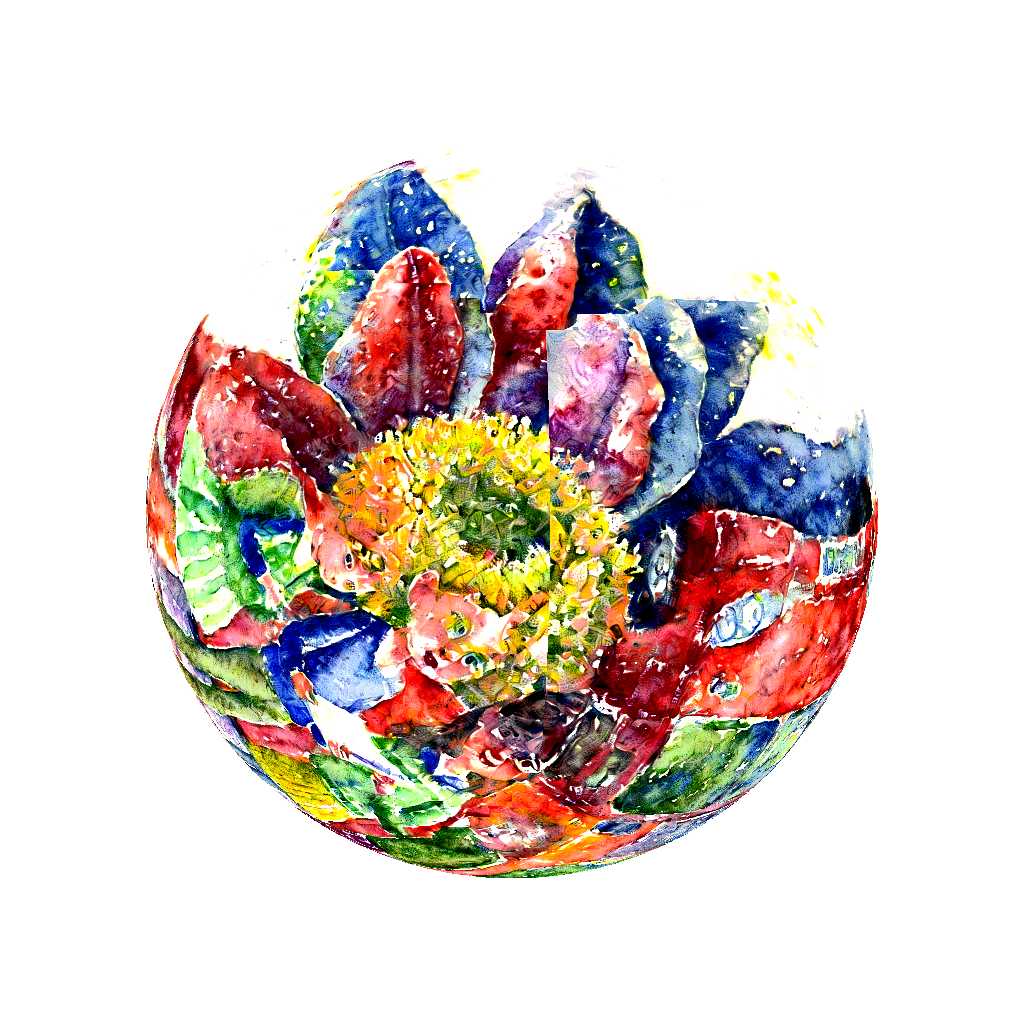}
        \end{minipage}\\

    \end{tabular}

    \vspace{-2mm}
    \caption{\textbf{Random samples.} We show random samples with the same prompt pairs on cube and sphere cases. }
    \label{fig:randomsample}
\end{figure*}

\topic{Complexity of shapes}
The cube scenario was more challenging due to minimal surface distortion, which complicates the concealment of the content, which can be observed in Fig. \ref{fig:randomsample}.
In contrast, scenes with more significant view distortion (sphere, cylinder, beanbag, Lego, and curved mirror) did not have this issue, indicating that high overlap and low distortion make content hiding more difficult.

\topic{Negative prompt of other views}
To make the content of one view hide better from another view, i.e., improve the concealment of the content. One may also propose that, for example, we add the text prompt of view 1 to the negative prompt of view 2. We explored this method but figured it didn't work as expected.

\begin{figure*}[h]
    \centering
    \tiny
    \vspace{7pt}

    \begin{minipage}[t]{0.49\linewidth}
        \centering
        \textbf{Prompt}\\[2pt]
        \includegraphics[trim=240 240 240 245, clip, width=0.32\linewidth]{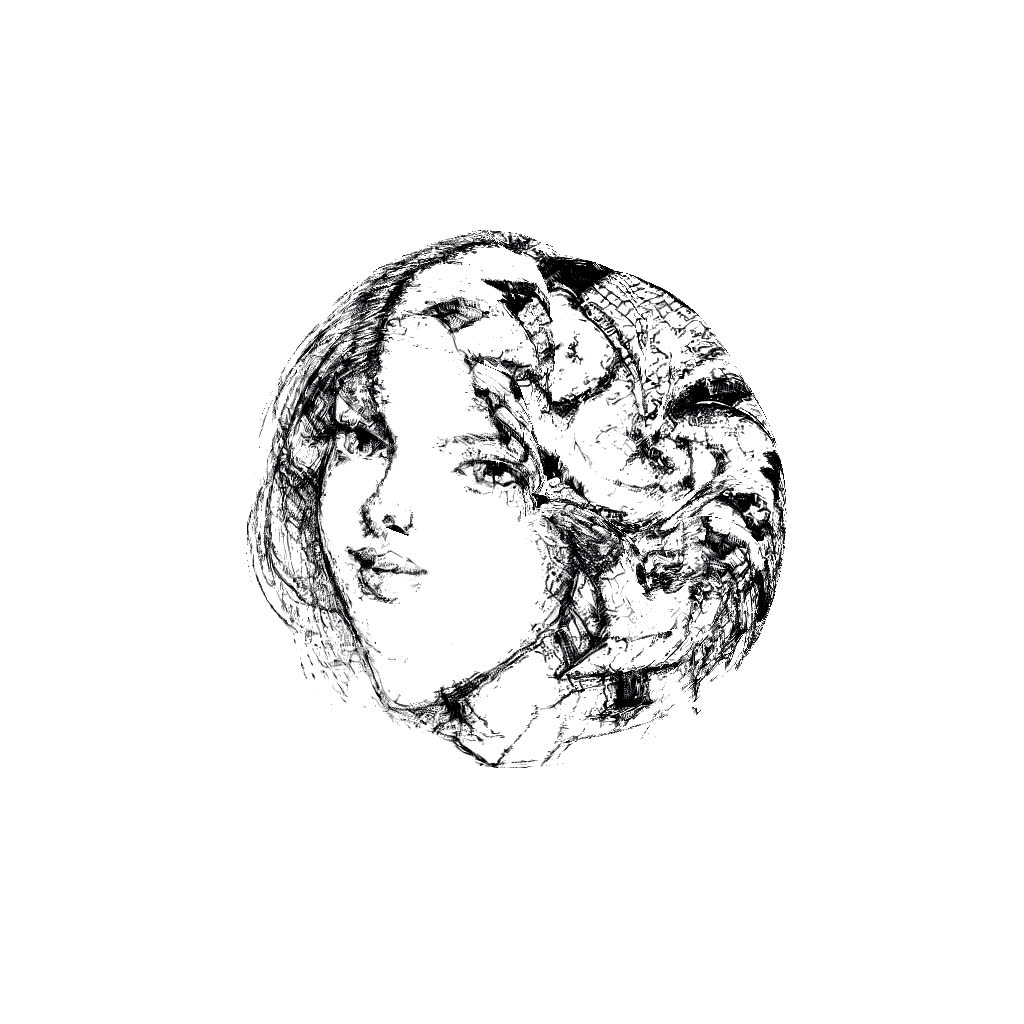}%
        \includegraphics[trim=240 210 240 220, clip, width=0.32\linewidth]{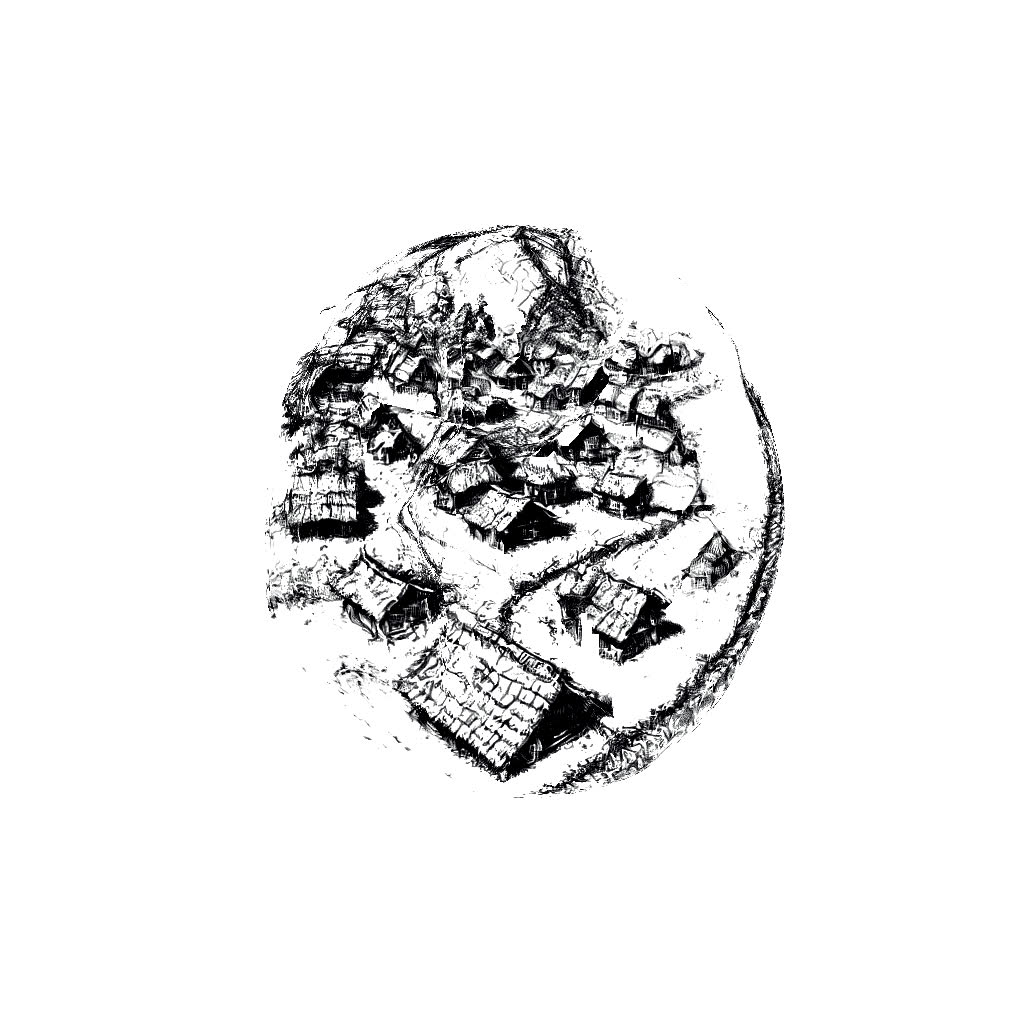}%
        \includegraphics[trim=220 240 220 240, clip, width=0.32\linewidth]{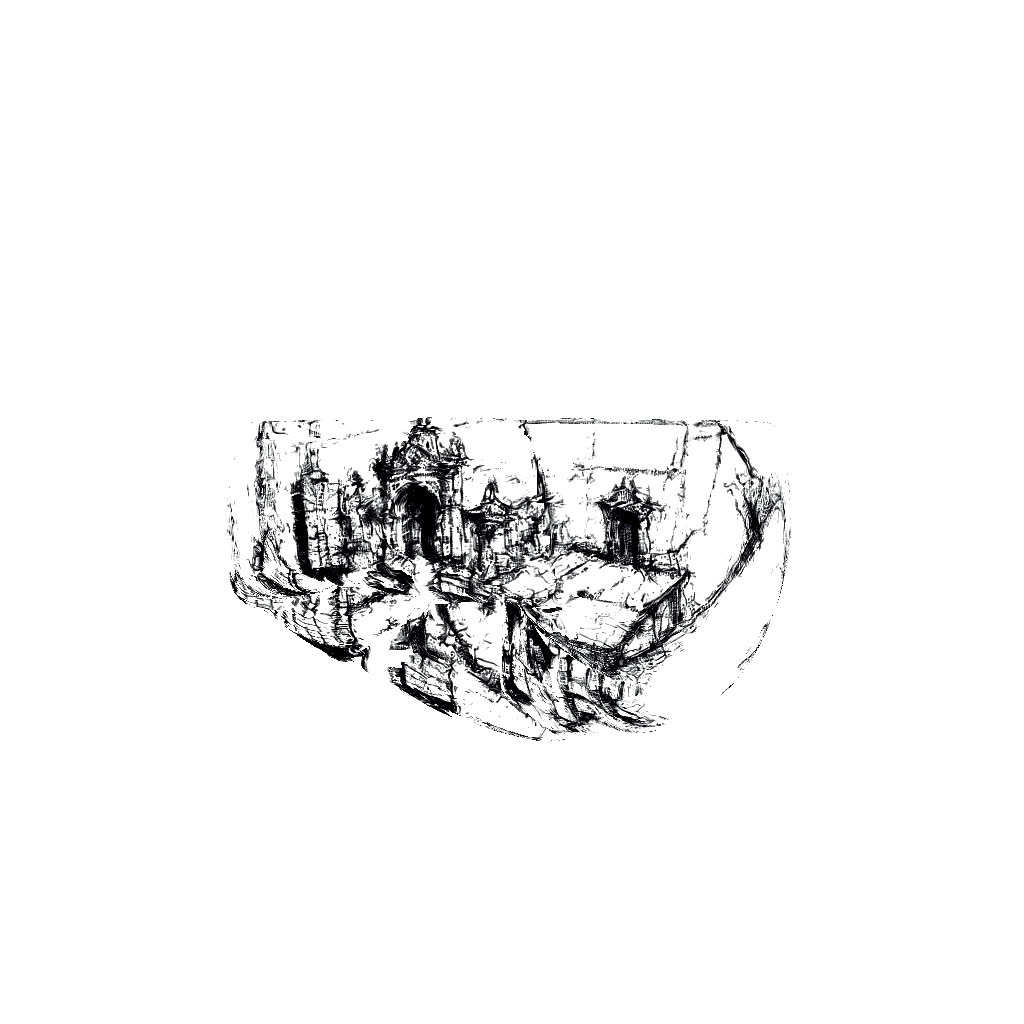}\\[-1pt]
        \includegraphics[trim=220 50 220 5, clip, width=0.33\linewidth]{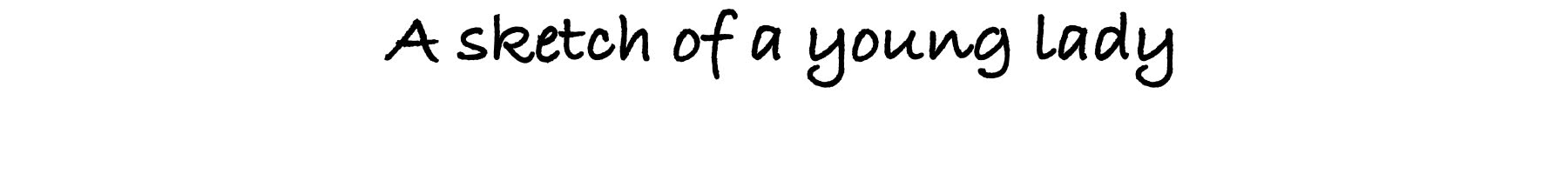}%
        \includegraphics[trim=230 50 230 5, clip, width=0.33\linewidth]{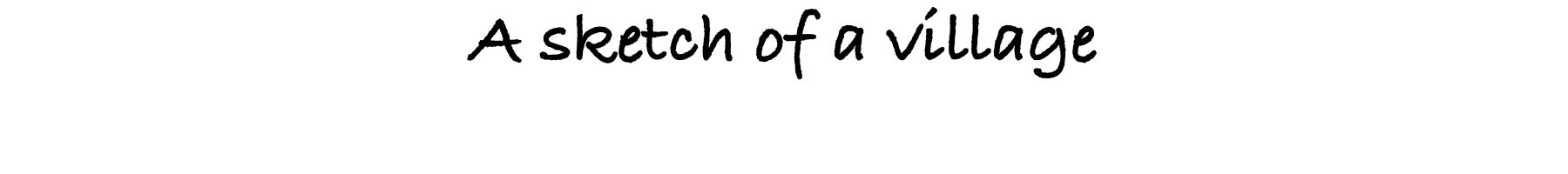}%
        \includegraphics[trim=230 50 220 5, clip, width=0.33\linewidth]{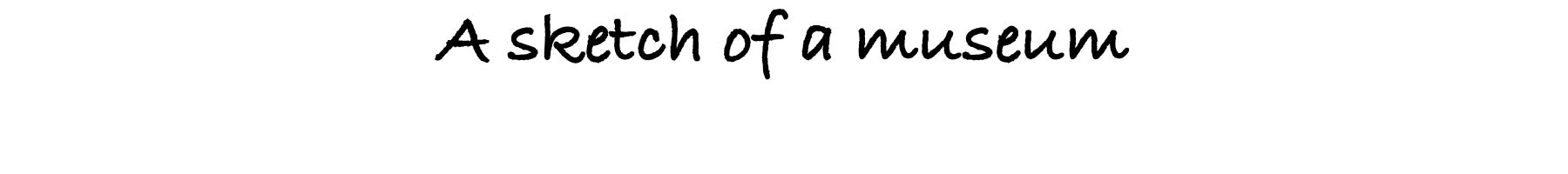}
    \end{minipage}\hspace{-4pt}
    \begin{minipage}[t]{0.49\linewidth}
        \centering
        \textbf{Prompt Variant 1}\\[2pt]  
        \includegraphics[trim=240 240 240 245, clip, width=0.32\linewidth]{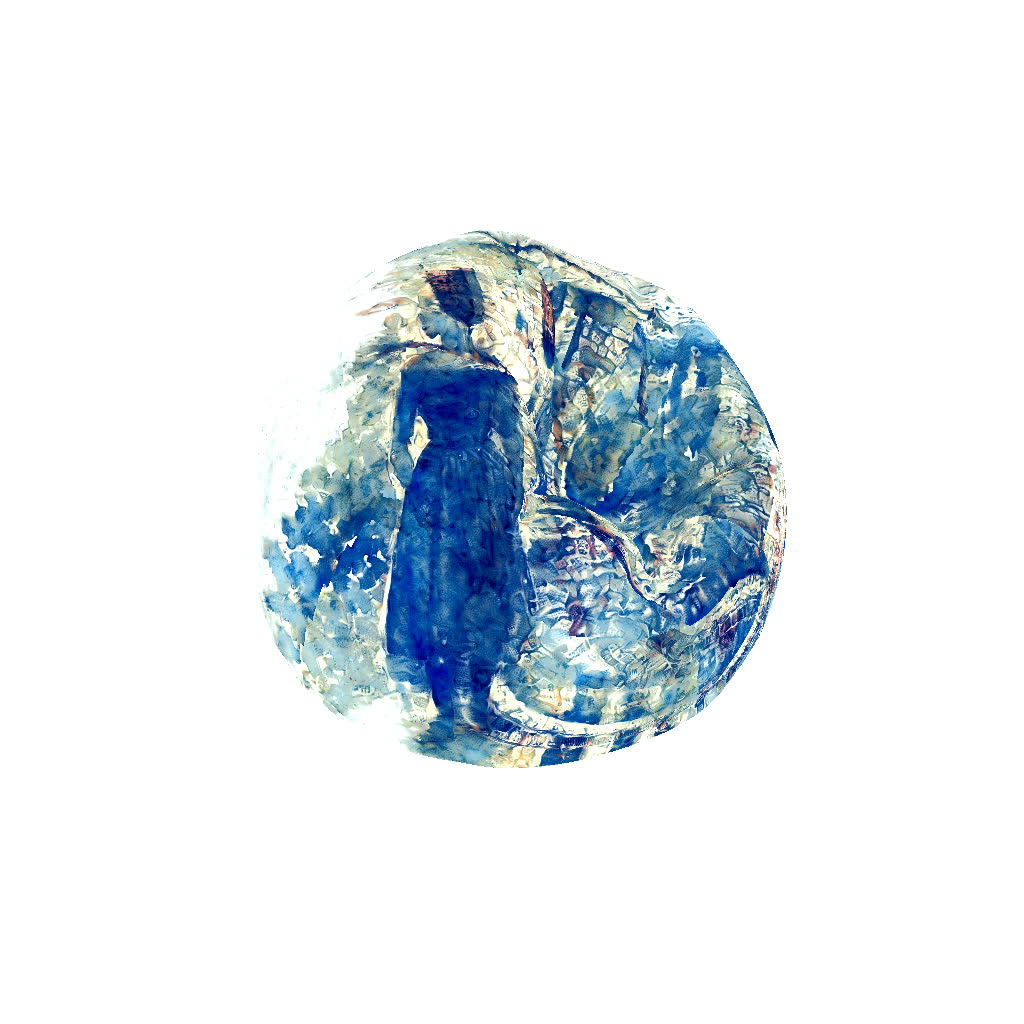}%
        \includegraphics[trim=240 210 240 220, clip, width=0.32\linewidth]{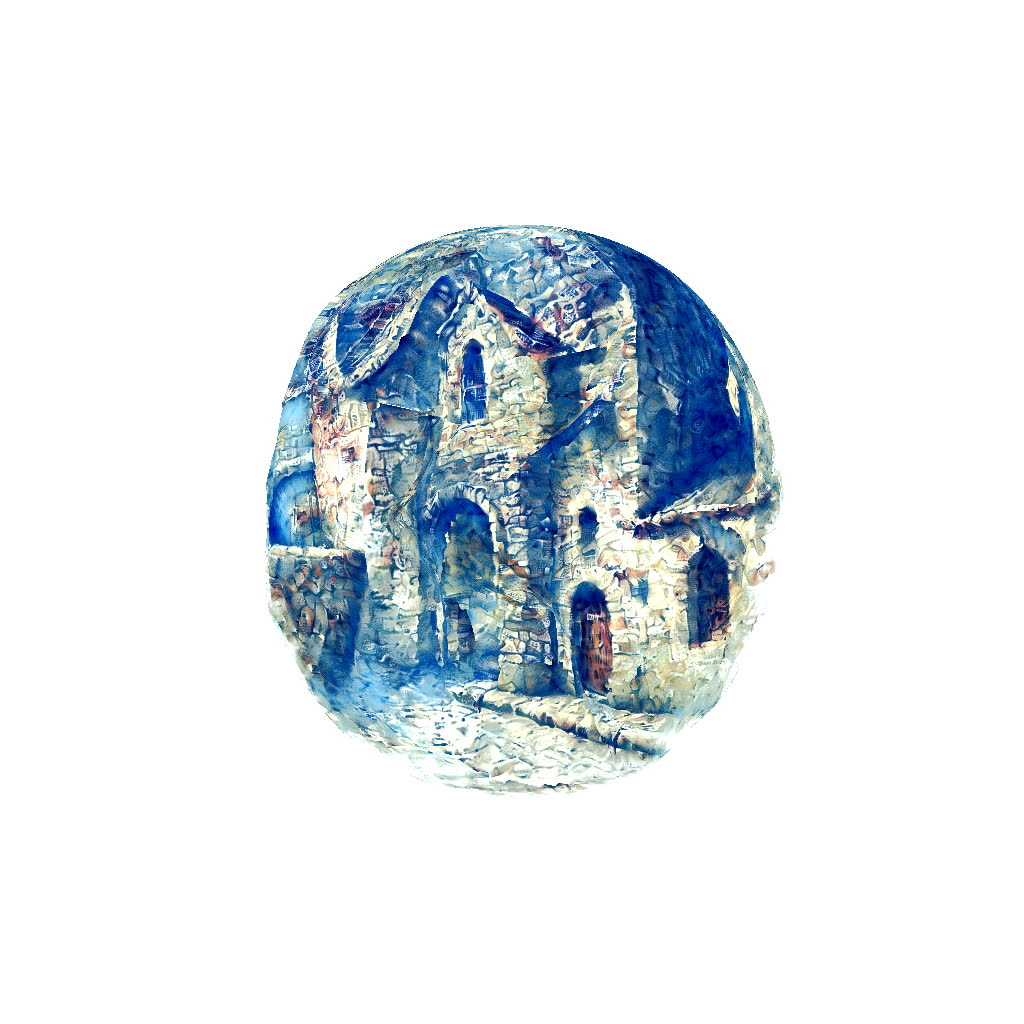}%
        \includegraphics[trim=220 240 220 240, clip, width=0.32\linewidth]{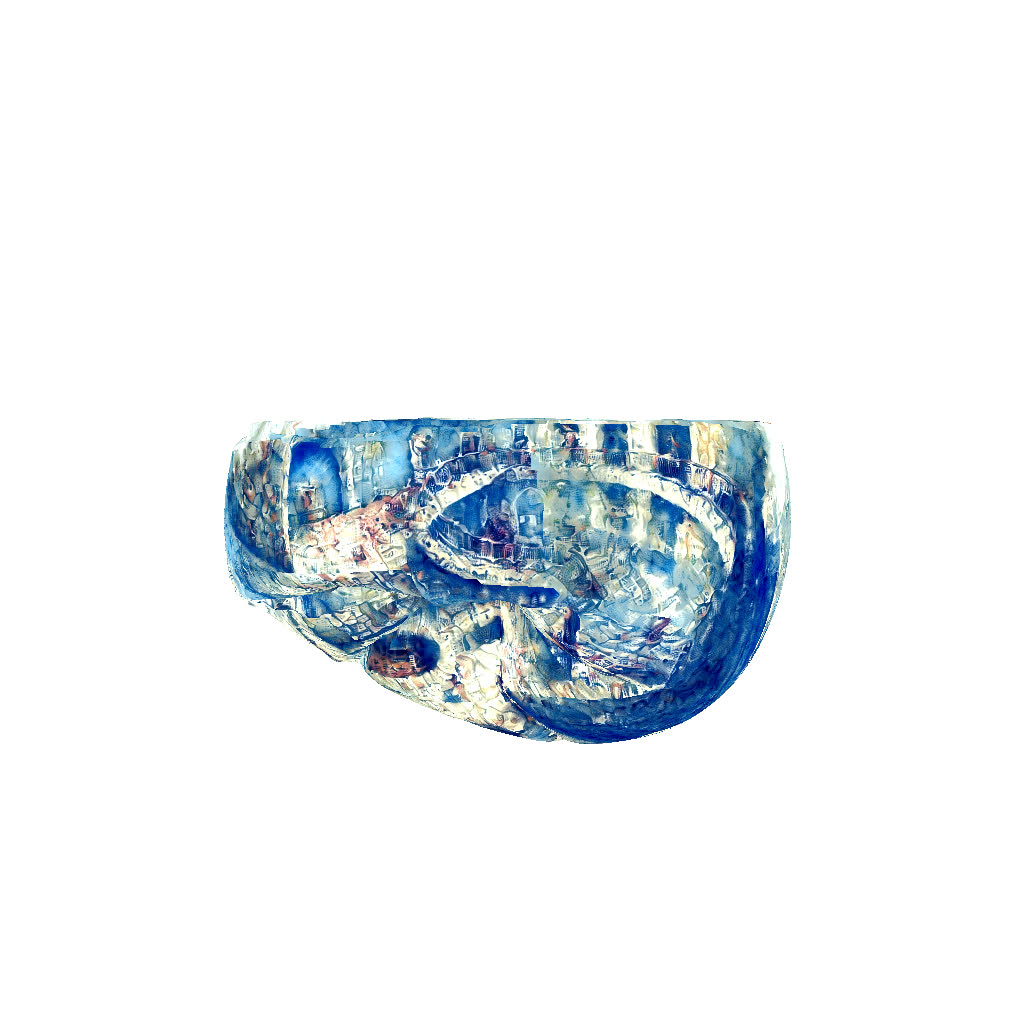}\\[-1pt]
        \includegraphics[trim=160 0 170 5, clip, width=0.33\linewidth]{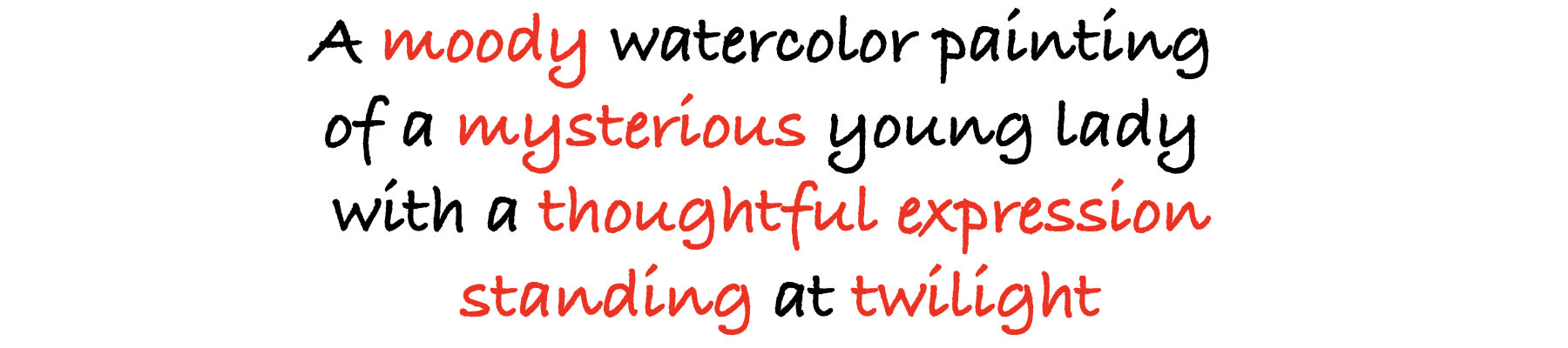}%
        \includegraphics[trim=160 0 170 5, clip, width=0.33\linewidth]{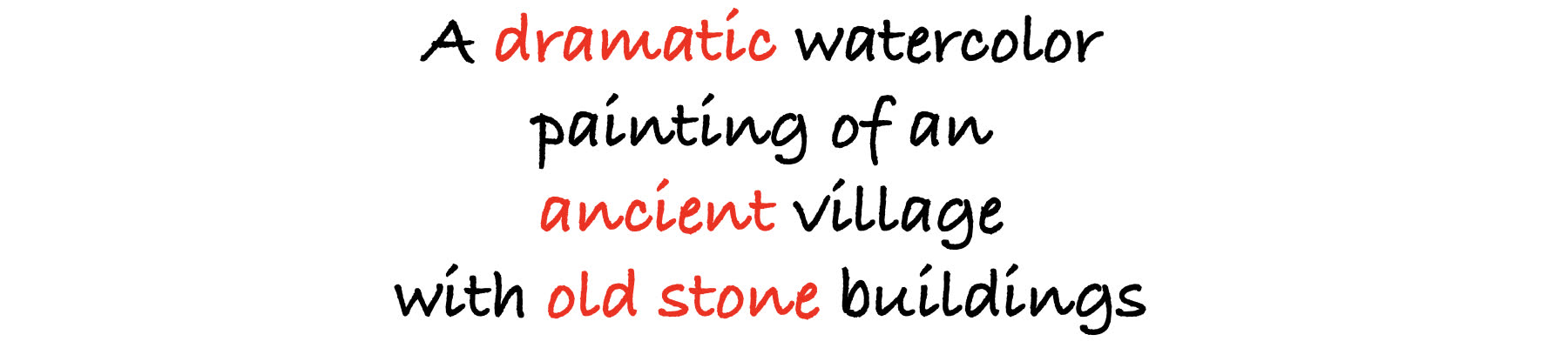}%
        \includegraphics[trim=160 0 170 5, clip, width=0.33\linewidth]{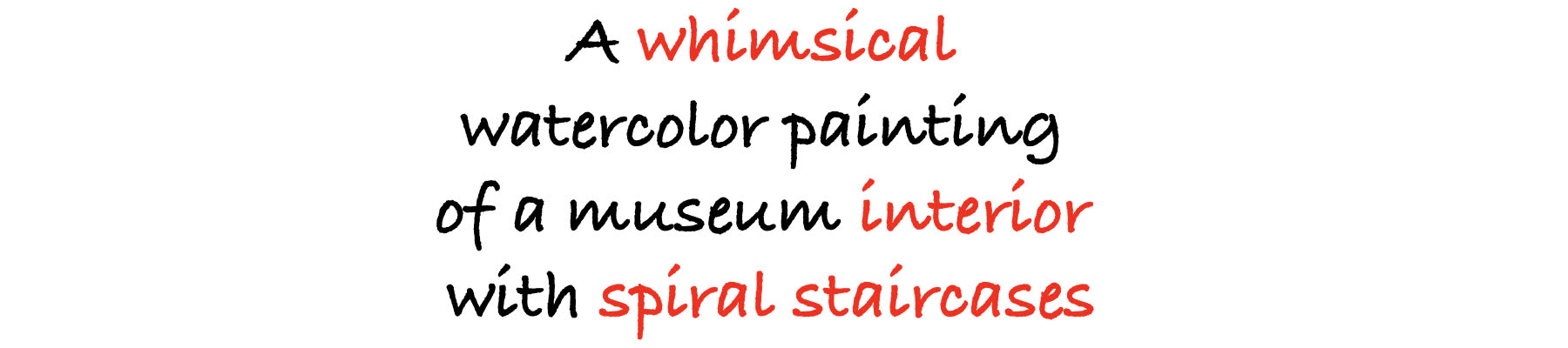}
    \end{minipage}

    \vspace{9pt}

    \begin{minipage}[t]{0.49\linewidth}
        \centering
        \textbf{Prompt Variant 2}\\[2pt]  
        \includegraphics[trim=240 240 240 245, clip, width=0.32\linewidth]{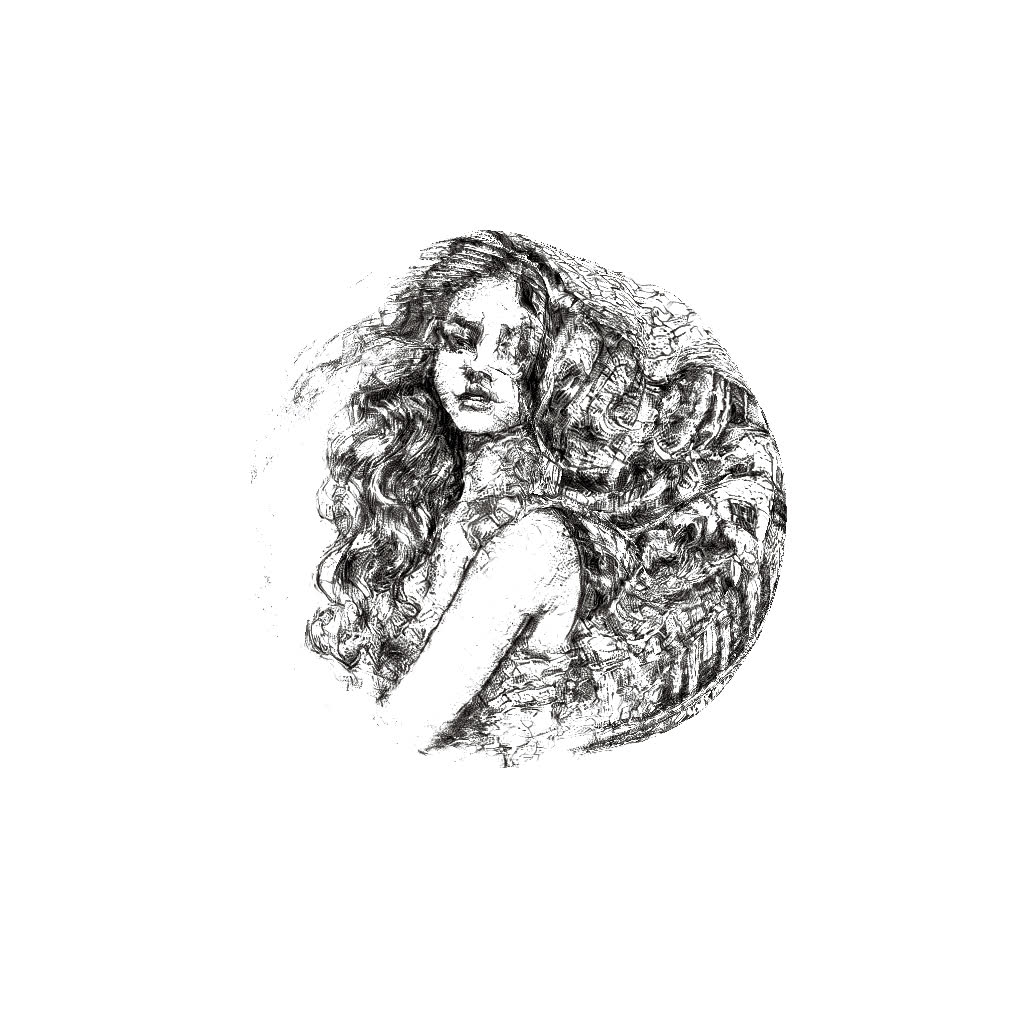}%
        \includegraphics[trim=240 210 240 220, clip, width=0.32\linewidth]{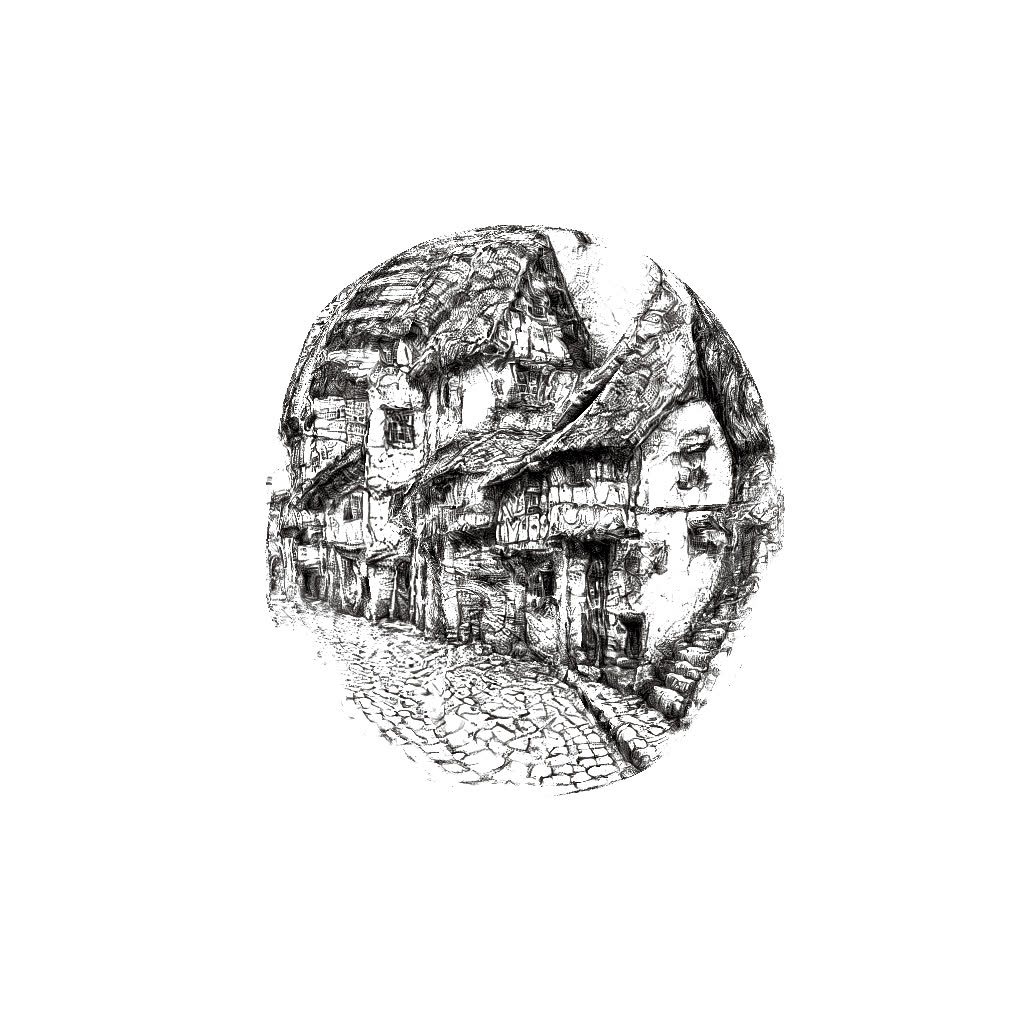}%
        \includegraphics[trim=220 240 220 240, clip, width=0.32\linewidth]{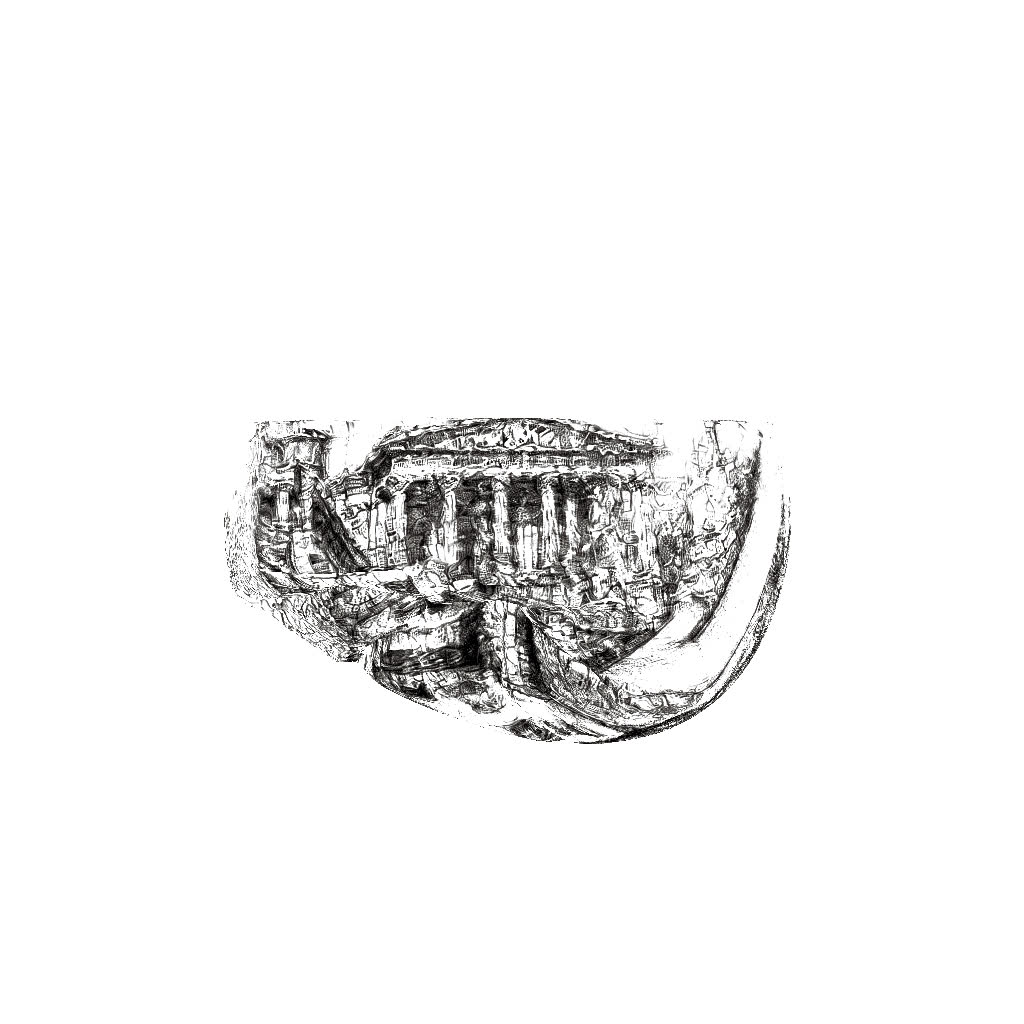}\\[-1pt]
        \includegraphics[trim=160 0 170 5, clip, width=0.33\linewidth]{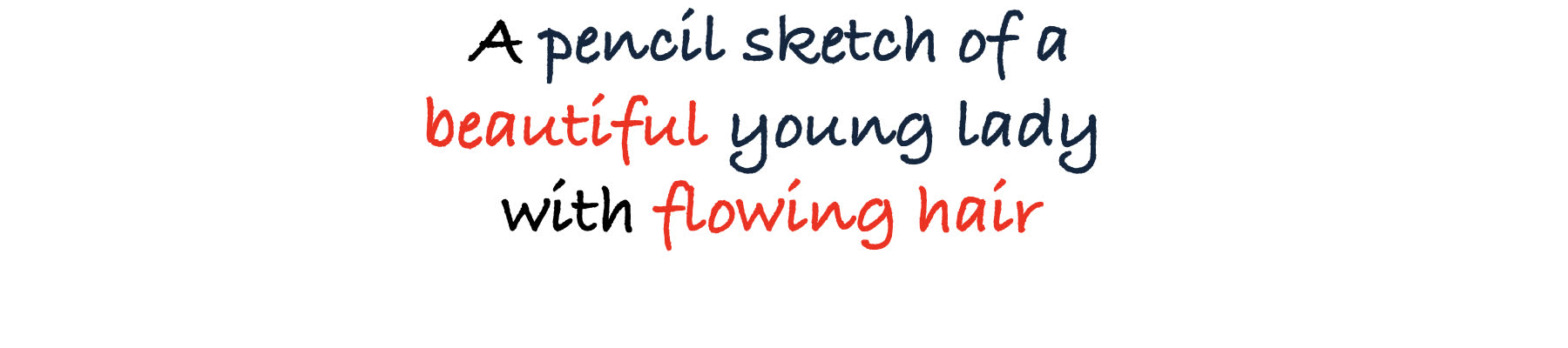}%
        \includegraphics[trim=160 0 170 5, clip, width=0.33\linewidth]{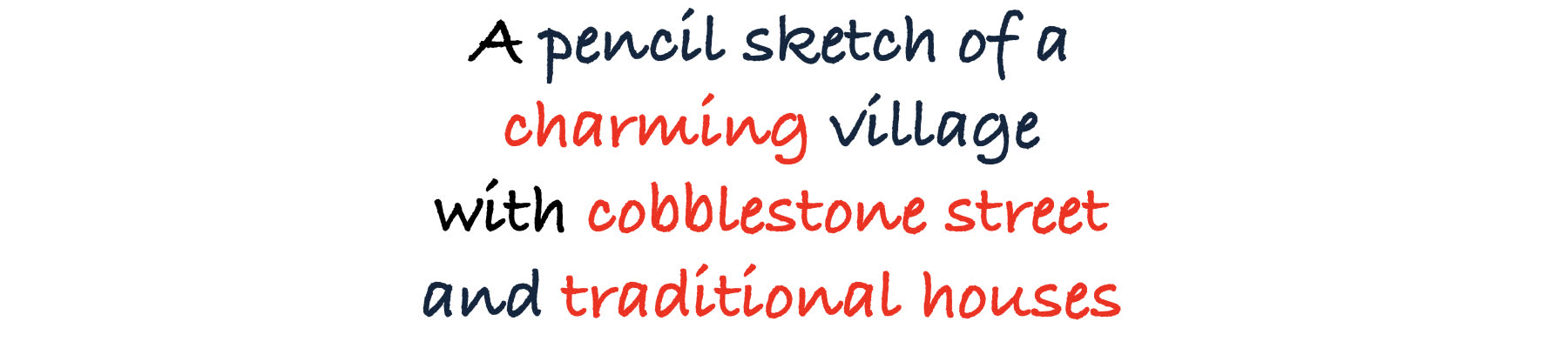}%
        \includegraphics[trim=160 0 170 5, clip, width=0.33\linewidth]{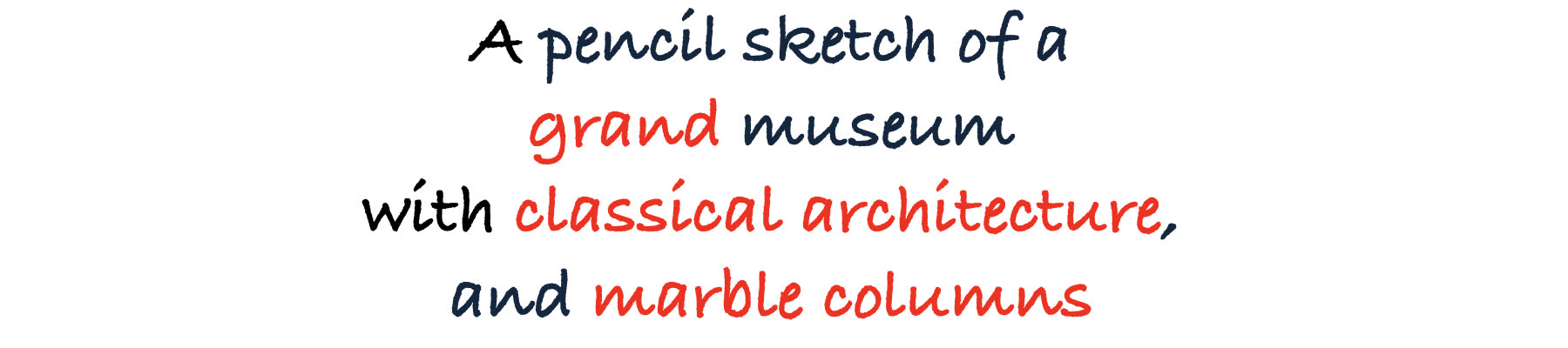}
    \end{minipage}\hspace{-4pt}
    \begin{minipage}[t]{0.49\linewidth}
        \centering
        \textbf{Prompt Variant 3}\\[2pt]
        \includegraphics[trim=240 240 240 245, clip, width=0.32\linewidth]{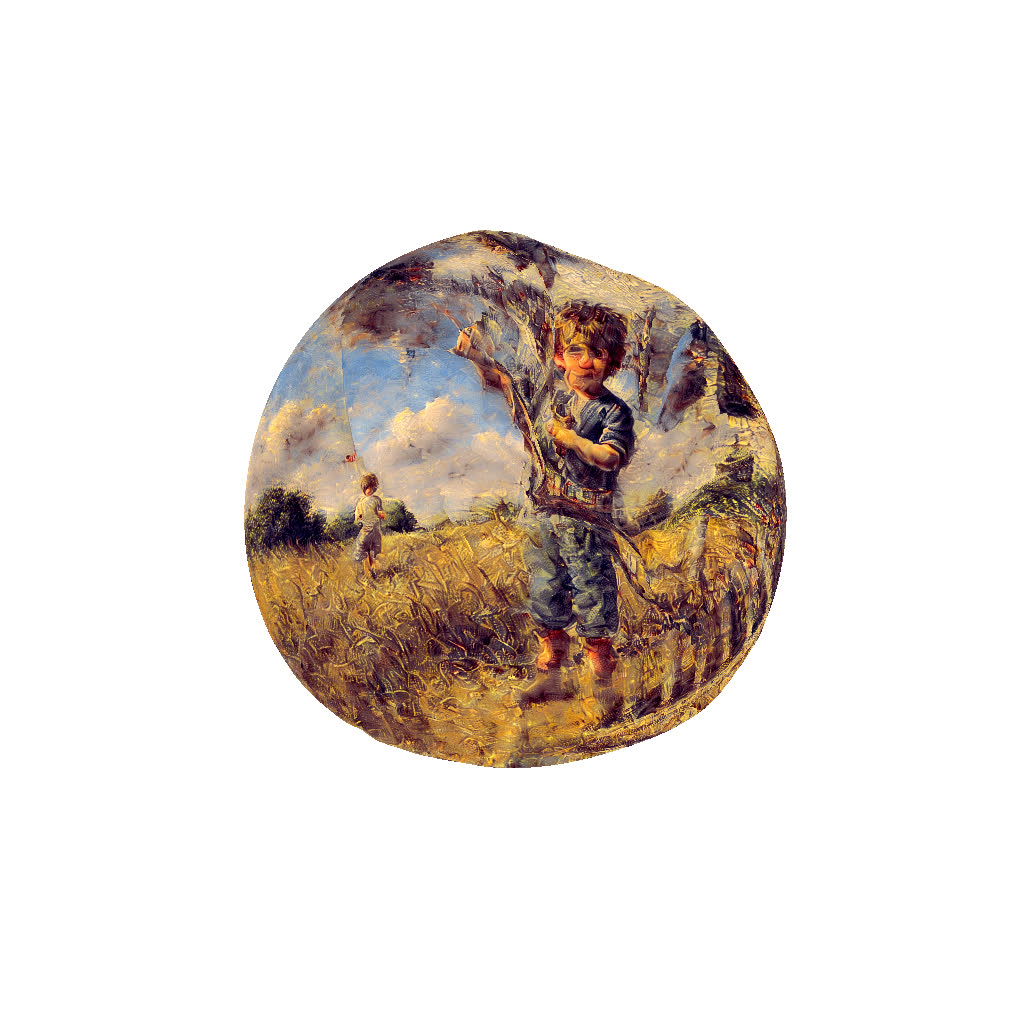}%
        \includegraphics[trim=240 210 240 220, clip, width=0.32\linewidth]{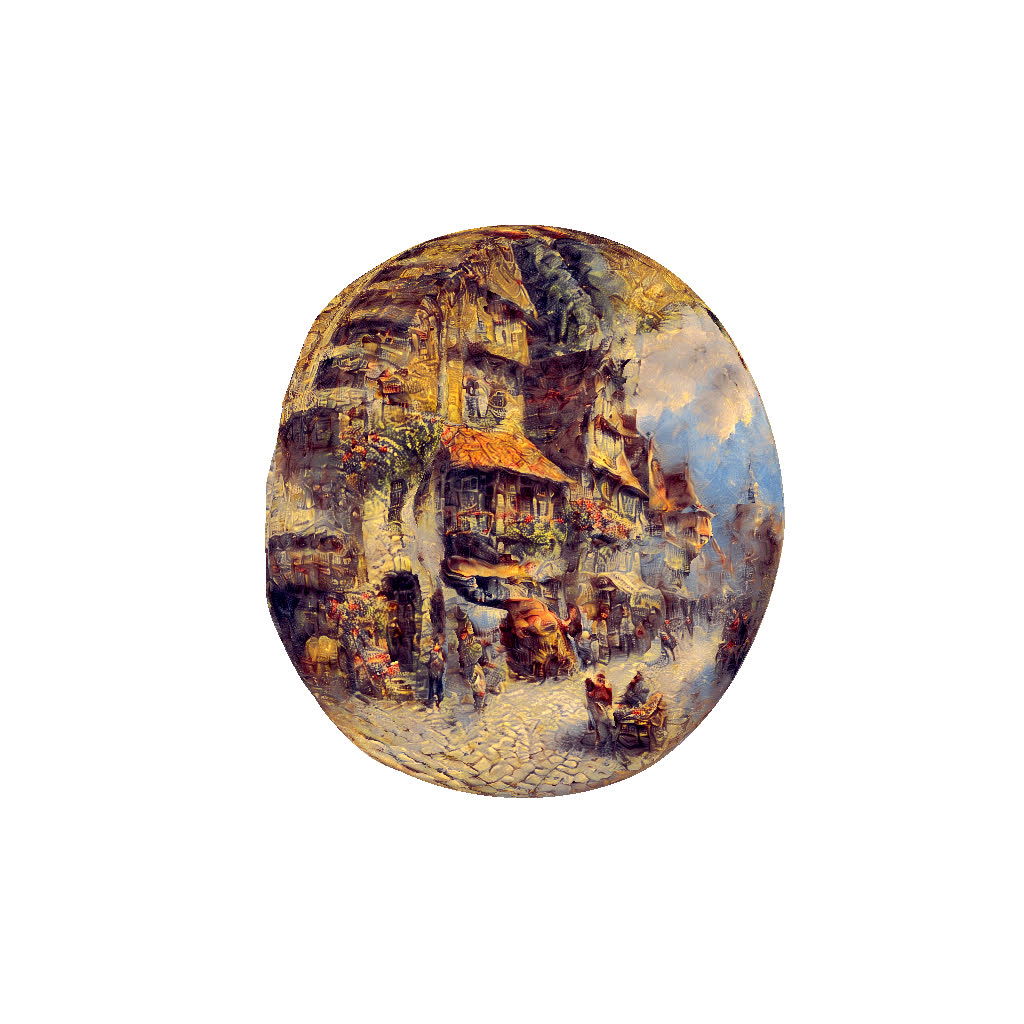}%
        \includegraphics[trim=220 240 220 240, clip, width=0.32\linewidth]{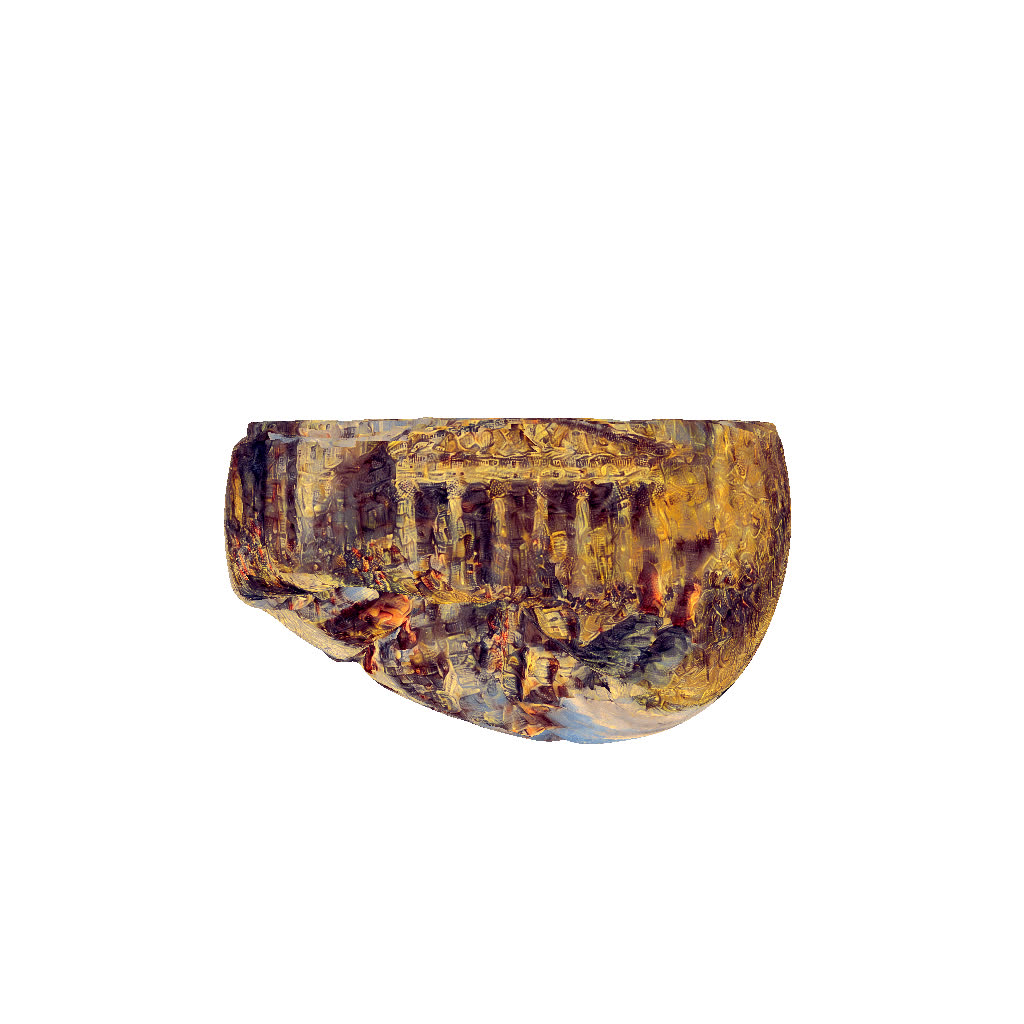}\\[-1pt]
        \includegraphics[trim=160 0 170 5, clip, width=0.33\linewidth]{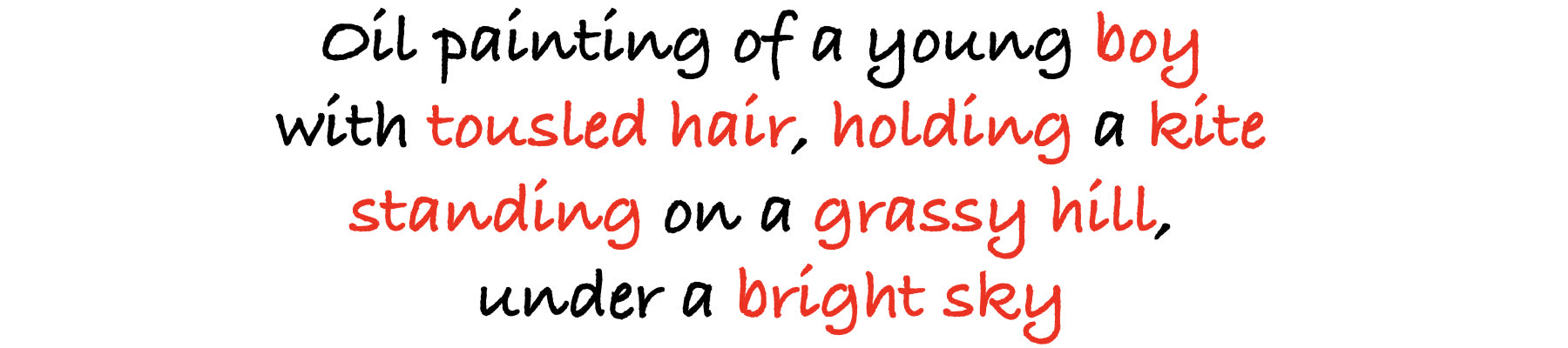}%
        \includegraphics[trim=160 0 170 5, clip, width=0.33\linewidth]{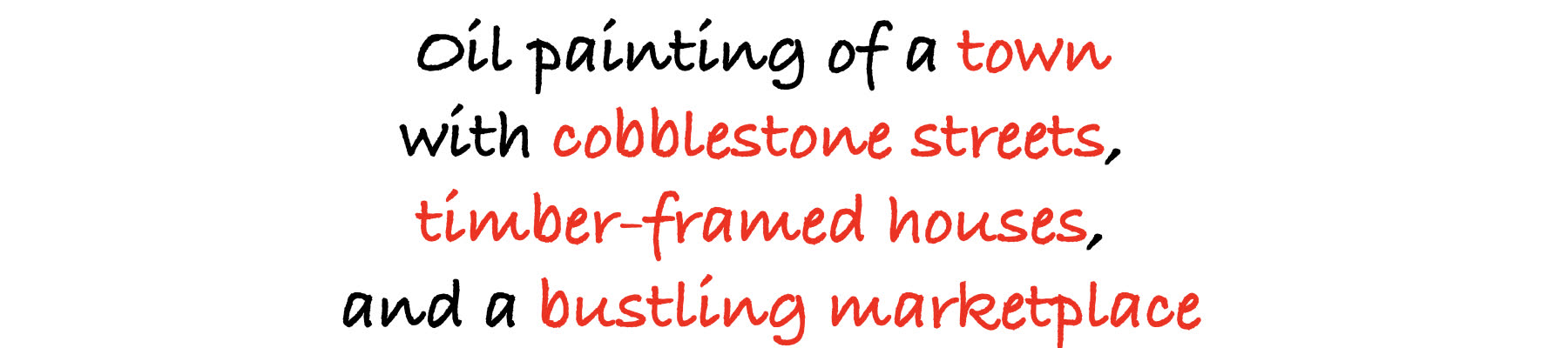}%
        \includegraphics[trim=160 0 170 5, clip, width=0.33\linewidth]{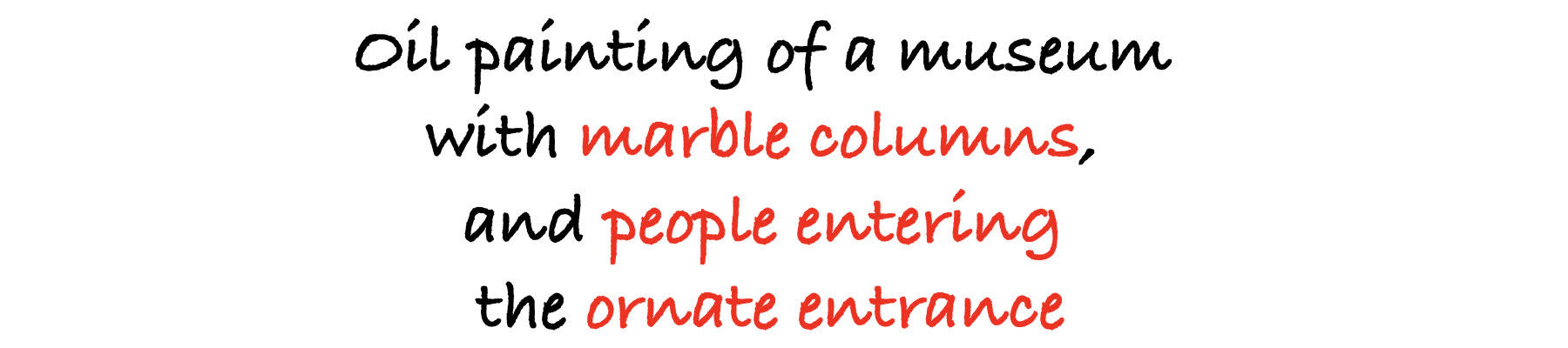}
    \end{minipage}
    \captionsetup{skip=1pt}
    \caption{\scriptsize \textbf{Prompt variance.} We added more descriptions to the input prompts on the bean bag example.}
    \label{fig:prompt_noise}
\end{figure*}

\topic{Prompt selection}
Illusion generation struggles to converge without style constraints, yielding lower success rates. Style-constrained prompts like ``\textsl{an oil painting of}'' and ``\textsl{a lithograph of}'' perform slightly better due to higher pixel consistency, aligning with findings in Diffusion Illusion \cite{Burgert2023DiffusionIH}. Styles or adding negative prompts do not significantly affect the generation. Illusion on real prompts is hard to succeed as described in 2D illusion models ~\cite{Burgert2023DiffusionIH,Geng2023VisualAG}, comparison of prompt variance is in Fig. ~\ref{fig:prompt_noise}.

\topic{Gradient mask}
Applying a gradient mask during optimization focusing on the central region can also enhance the fusion of input views, especially when the object is off-center. We explored this method in a 512-resolution case. We didn't include it in the 1024-resolution version because the render resolution scaling can do the same while increasing the rendered resolution.

\topic{Dream target loss}
Dream target loss from Burgert \textit{et al.} ~\cite{Burgert2023DiffusionIH} uses a target image to help generate the illusion; this is useful in 2D cases when the view of the images does nott change. In the 3D case, this method fails as it only supervises pixels within a fixed space that a VAE encoder can see. If we move the camera, the training will fail as the L2 loss part of Dream Target loss only supervises images pixel-wise with the target image. VSD loss ~\cite{wang2023prolificdreamer} does not have this problem because it supervises the content based on the whole input image and the text prompt. So, if we jitter the view a bit, the training will still succeed, and we train part of the null space of a VAE encoder.

\section{User Study}
\label{sec:userstudy}
\begin{figure}[h]
\vspace{-6mm}
    \centering
    \includegraphics[width=0.4\textwidth]{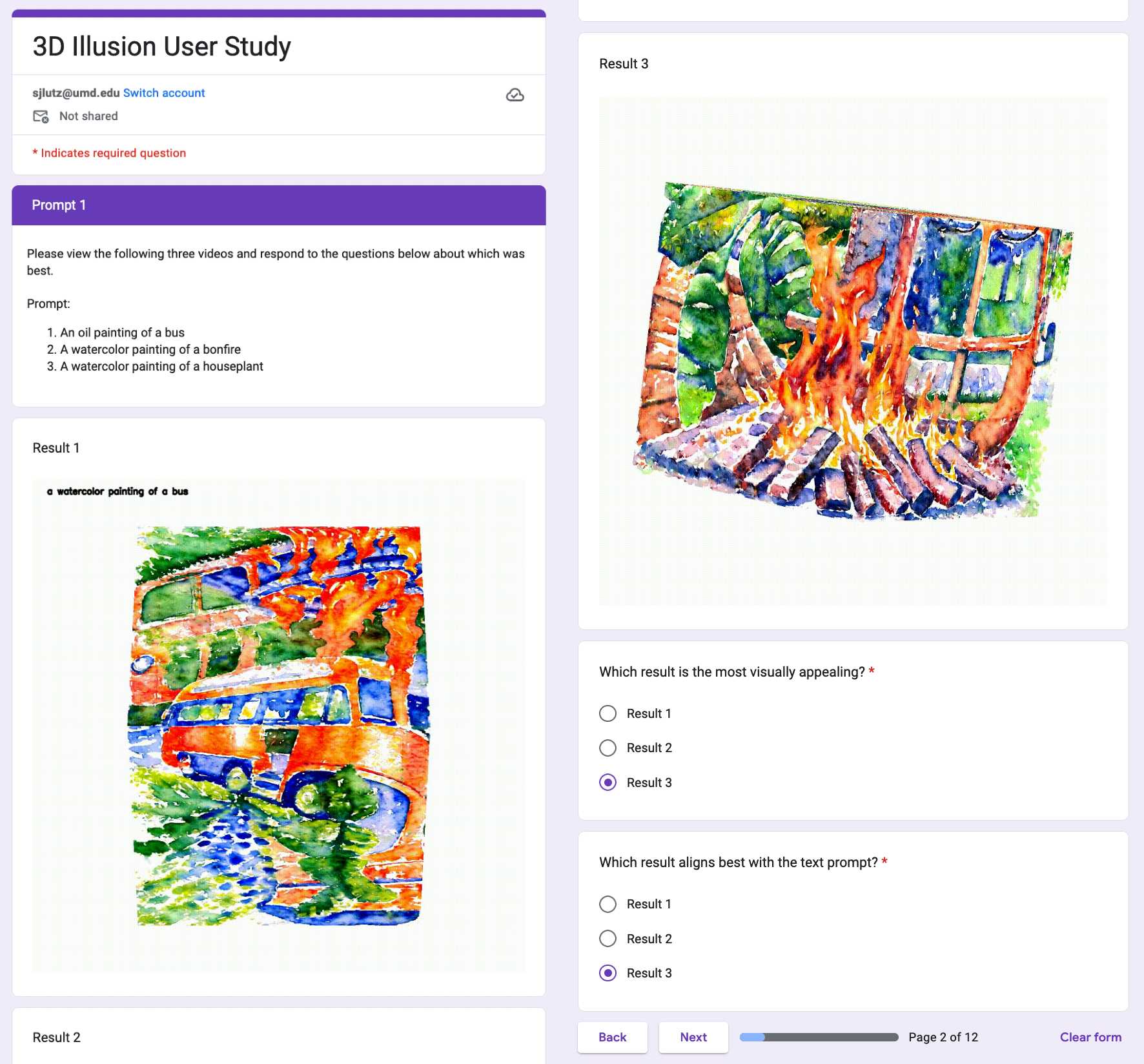}
    \caption{\textbf{Preview of User Study.}
    The user study was distributed over Google Forms. The user is provided with some context and the relevant prompts, followed by three animated GIFs, one for each method. They are then prompted with two multiple choice questions to evaluate the quality of the results.}
    \label{fig:userstudy}
\end{figure}

We present details of user study. We distributed three surveys, each containing 11 sections of method comparisons to 40 users. Each section presented the user with three results generated on the same prompt but with different methods. The results are animated GIFs, allowing the user to evaluate all views of each result. The user is then prompted to answer two questions: ``\textsl{Which result is the most visually appealing?}'' and ``\textsl{Which result aligns best with the text prompt?}''.  A screenshot of user study is in Fig. \ref{fig:userstudy}.

To compute the result, we divide the number of times each method was selected, \( T_{\text{selected}} \), by the number of times it was shown to the user, denoted as \( T_{\text{shown}} \): 
\begin{equation}
\begin{aligned}
\text{Score} = \frac{T_{\text{selected}} }{T_\text{shown}}
\end{aligned}
\label{eq:usetudy}
\end{equation}

This gives us a metric of the probability that a user prefers a method's result over others, given that the result is displayed.

\end{document}